\documentclass[11pt]{article}

\PassOptionsToPackage{dvipsnames,table}{xcolor}
\usepackage[preprint]{acl}

\usepackage{times}
\usepackage{latexsym}
\usepackage[T1]{fontenc}
\usepackage[utf8]{inputenc}
\usepackage{newunicodechar}
\usepackage{microtype}
\usepackage[table]{xcolor}
\usepackage{inconsolata}
\usepackage{graphicx}
\usepackage{booktabs}
\usepackage{adjustbox}
\usepackage{array}
\usepackage{xcolor}
\usepackage{wrapfig}
\usepackage{subcaption}
\usepackage{amsmath}   
\usepackage{amssymb}   
\usepackage{amsfonts}  
\usepackage{mathtools} 
\usepackage{mdframed}
\usepackage{makecell} 
\usepackage{tikz}
\usetikzlibrary{arrows.meta, positioning}
\usepackage{colortbl}
\usepackage{xspace}
\usepackage{ulem}
\usepackage{bbm}
\usepackage{multirow}
\usepackage{adjustbox}
\usepackage{array}

\newcommand{\rate}[1]{~{\color{gray}\scriptsize(#1)}}

\definecolor{darkblue}{rgb}{0, 0, 0.5}
\hypersetup{colorlinks=true, citecolor=darkblue, linkcolor=darkblue, urlcolor=darkblue}
\definecolor{mygreen}{RGB}{0,100,0}   
\definecolor{myred}{RGB}{139,0,0}     
\definecolor{mygray}{RGB}{80,80,80}   

\newcommand{\auto}{\textsc{Self-decision}\xspace}
\newcommand{\notool}{\textsc{No Tool}\xspace}
\newcommand{\withtool}{\textsc{Always Tool}}
\newcommand{\optimized}{\textsc{Optimal}}

\newcommand{\change}[1]{\textcolor{black}{#1}}

\title{To Call or Not to Call:\\ A Framework to Assess and Optimize LLM Tool Calling}

\author{
\textbf{Qinyuan Wu}$^{1}$~\;~
\textbf{Seungeon Lee}$^{1}$~\;~
\textbf{Soumi Das}$^{1}$~\;~
\textbf{Mahsa Amani}$^{1}$~\;~
\textbf{Arijit Nag}$^{1}$~\;~\\
\textbf{Krishna Gummadi}$^{1}$~\;~
\textbf{Abhilasha Ravichander}$^{1}$~\;~
\textbf{Muhammad Bilal Zafar}$^{2,3}$\\
$^1$Max Planck Institute for Software Systems\quad
$^2$Ruhr University Bochum \quad
$^3$UAR RC Trust \quad\\
\small{
   \textbf{Correspondence:} 
   \href{mailto:qwu@mpi-sws.org}{qwu@mpi-sws.org}
 }
}

\begin{document}

\maketitle

\begin{abstract}
Agentic AI architectures augment LLMs with external tools, unlocking strong capabilities but potentially incurring substantial costs. 
Moreover, tool use is not always beneficial: redundant or low-utility calls can even harm task performance. 
Effective tool use, therefore, hinges on a core LLM decision: \textit{whether to call or not call a tool, when performing a task}. 
We introduce a principled framework inspired by decision-making theory
to understand tool-use decisions along three key factors: \textit{necessity, utility, and affordability}. 
Our analysis combines two complementary lenses: a normative perspective that infers true need and utility for optimal tool calls, and a descriptive perspective that infers the model's self-perceived need and utility from their observed behaviors. 
We evaluate six open models and a proprietary OpenAI model across native and customized harnesses, two tools, and six tasks. Models' perceived need and utility remain misaligned with their true values, particularly under budget constraints. This misalignment produces both costly overuse and performance-degrading calls.
To improve the tool decisions, we train lightweight latent estimators of need (LNEs) from model hidden states. LNEs generally predict true need more accurately than model self-reports and improve budgeted tool allocation across model scales and tool types. \footnote{Code and dataset available at \url{https://github.com/QinyuanWu0710/ToCall_or_NotToCall}.}
\end{abstract}

\section{Introduction}

\begin{figure*}[t]
    \centering
    \includegraphics[width=0.95\textwidth]{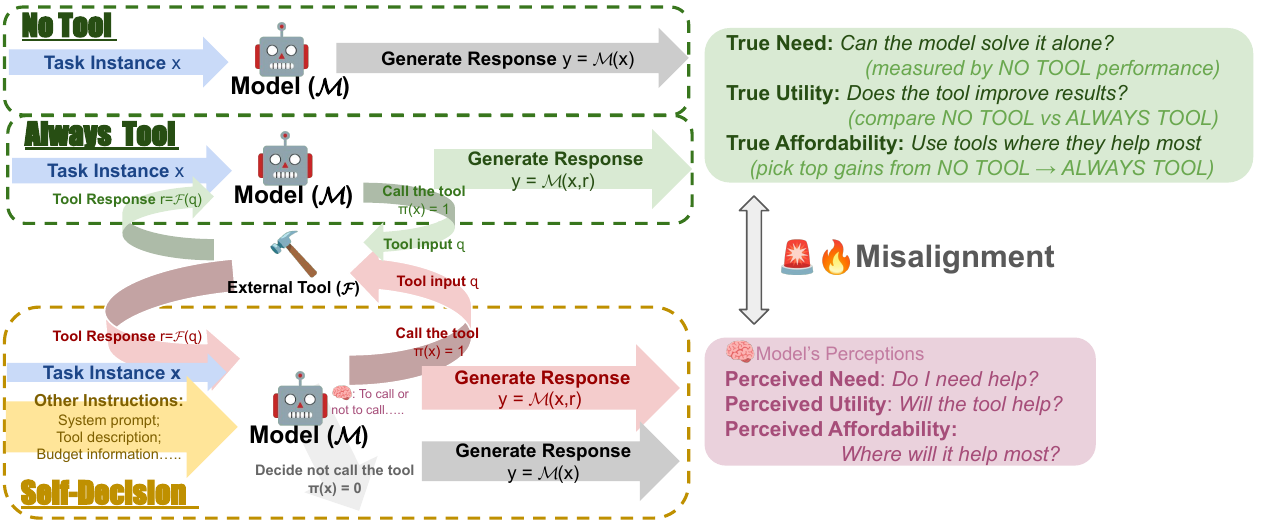}
    \caption{Given input $x$, the model $\mathcal{M}$ decides $\pi(x)\in\{0,1\}$ to call a tool (response $r$) or not, producing $y=\mathcal{M}(x,r)$ or $y=\mathcal{M}(x)$. We compare NO TOOL and ALWAYS TOOL, and evaluate SELF-DECISION decisions via need (requires help), utility (performance gain), and affordability (cost vs. gain).}
    \label{fig:overview}
\end{figure*}

\noindent 
At their core, agentic AI architectures enable AI models, such as LLMs, to extend their functionality by calling external tools~\citep{yao2023react,singh2025agentic}. 
When solving challenging tasks, LLMs can strategically augment their internal (parametric) knowledge, skills, and abilities by utilizing external tools. 
The crucial decision \textit{to call or not to call} a tool lies with the LLM.
Our focus in this paper is on understanding how LLMs exercise their agency in making tool-calling decisions.  

Most evaluations assess tool use through aggregate end-to-end task performance~\citep{yao2023react, qian2025toolrl, feng2025retoolreinforcementlearningstrategic}, which obscures whether individual calls are warranted. 
Table~\ref{tab:main} illustrates why such fine-grained assessment is needed: when models are given access to a Web search tool for answering factual questions, tool use helps on average but is far from optimal. In fact, four of the seven models we evaluate make redundant tool calls and perform worse.
This inefficiency matters beyond final-task accuracy: unnecessary calls add billed inference tokens and latency~\citep{artola2025overcharging}, and paid external APIs can dominate the cost of an agent workflow altogether~\citep{jagtap2026agentbudget}. In our GPT-5.5 \auto{} setup, tool calls cost \$0.94 and \$5.80 for the whole task. 
This gap between aggregate benefit and per-instance optimality shows that current tool-calling decisions made by the model itself are not yet principled, motivating a systematic account of when a tool call is actually warranted and worth its cost. Our goal is to develop a framework to understand and improve these sub-optimal tool calling or function calling decisions by LLM agents. We use tool calling and function calling interchangeably throughout the paper.

Inspired by the theory of rational choice~\citep{peterson2017introduction, tversky1974judgment}, we posit that the optimal goal of tool calling should be to help the model \textit{maximize the expected improvement in task performance (utility gain) afforded by tool use, under cost constraints}.
Tool use should begin with \textit{need}: whether the model can solve an instance adequately without external help. When it cannot, a call is warranted only if its expected \textit{utility} is positive; under a limited budget, calls should prioritize instances with the greatest expected gains.

As shown in Figure~\ref{fig:overview}, we now propose a framework for assessing LLM function calling decisions along three dimensions:\\     
{\bf 1. Necessity:} Does a model $\mathcal{M}$ {\bf require} external help to solve a task $\mathcal{T}$? That is, can $\mathcal{M}$ not solve $\mathcal{T}$ satisfactorily using its own internal parametric knowledge?\\
{\bf 2. Utility:} Will a model $\mathcal{M}$ {\bf benefit} from calling a function $\mathcal{F}$, when solving a task $\mathcal{T}$? That is, to what extent will $\mathcal{M}$'s performance on $\mathcal{T}$ improve or worsen by calling $\mathcal{F}$?\\
{\bf 3. Affordability:} Is it {\bf cost-effective} for a model $\mathcal{M}$ to call a function $\mathcal{F}$ when solving a task $\mathcal{T}$? That is, does the gain in utility from calling $\mathcal{F}$ justify the additional cost of calling $\mathcal{F}$? 

\begin{table}[t]
\centering
\scriptsize
\begin{adjustbox}{
  max width=\linewidth,
  max totalheight=0.3\textheight,
  keepaspectratio
}
\begin{tabular}{lcccc}
\toprule
\textbf{Model}
& \textbf{\notool}
& \textbf{\withtool}
& \textbf{\auto}
& \textsc{\textbf{Optimal}}
 \\
& \textbf{Score} & \textbf{Score} & \textbf{Score} & \textbf{Score}  \\
\midrule

\rowcolor{blue!8}
GPT-OSS-120B & 0.61\rate{0} & 0.76\rate{100} & 0.72\rate{30} & \textbf{0.81}\rate{61}\\
\rowcolor{blue!8}
Qwen3-30B-A3B & 0.70\rate{0} & 0.81\rate{100} & 0.80\rate{56} & \textbf{0.88}\rate{51} \\
\rowcolor{blue!8}
Qwen-3-30B-IT & 0.68\rate{0} & 0.82\rate{100} & 0.82\rate{95} & \textbf{0.87}\rate{60} \\
\rowcolor{blue!8}
Llama3.2-3B-IT & 0.58\rate{0} & 0.70\rate{100} & 0.70\rate{100} & \textbf{0.83}\rate{57} \\
\rowcolor{blue!8}
Mistral3.1-24B-IT & 0.70\rate{0} & 0.83\rate{100} & 0.70\rate{0} & \textbf{0.88}\rate{51} \\
\rowcolor{orange!12}
Gemma3-27B-IT & 0.60\rate{0} & 0.80\rate{100} & 0.80\rate{92} & \textbf{0.85}\rate{59} \\
\rowcolor{green!12}
GPT-5.5 & 0.85\rate{0} & 0.86\rate{100} & 0.85\rate{32} & \textbf{0.94}\rate{39}\\

\bottomrule
\end{tabular}
\end{adjustbox}

\caption{
Entity-task performance under four tool-use policies. Each entry reports the task score, followed in parentheses by the actual tool-call rate (\%). Row colors indicate the harness:
\colorbox{blue!8}{\strut Trained},
\colorbox{orange!12}{\strut Custom}, and
\colorbox{green!12}{\strut OpenAI}. In the GPT-5.5 \auto{} setups, the web-search tool cost 0.94\$ and 5.80\$ for the whole task.
}
\label{tab:main}
\vspace{-10pt}
\end{table}

{\bf Theoretically}, we make two important observations about \textit{need} and \textit{utility} dimensions: First, a model’s utility from calling a tool on a given task is upper bounded by its need under a defined metric, i.e., how poorly it performs when solving the task independently.
\emph{If a model can perfectly solve a task on its own, tool use cannot yield positive utility.}
Second, \textit{estimating need} for tool calling depends only on the model and the task, while \textit{estimating utility} requires additional knowledge about the tool.
However, \emph{learning to predict the behavior of a complex tool can itself be highly challenging.}
As we will show later in Section~\ref{sec:controller}, this distinction is important for assessing and optimizing tool calling decisions in practice.

{\bf Practically}, we study LLM function calling through three lenses~\citep{peterson2017introduction,tversky1974judgment}: \textit{normatively}, when a call is actually warranted, as defined above; \textit{descriptively}, how LLMs behave in practice (the \auto setting), including suboptimal tool use and adherence to cost constraints; and \textit{prescriptively}, how to improve tool calling so that decisions in practice better match the normative ideal.

We operationalize the framework on six open-source models spanning 3B--120B parameters, including instruction-tuned and reasoning models, and one closed-source frontier model, evaluated across two tool types and six tasks under a fixed harness per model (Section~\ref{sec:experiments} gives full details on tasks, search providers, and harnesses). Our experimental setup measures models' \textit{true} need and utility, which define normative (oracle) tool-calling behavior, against their \textit{(self-)perceived} need and utility, which guide their own decisions. Comparing these perspectives reveals that \textit{tool-calling is not universally beneficial}: a call may provide no gain or even hurt performance. Moreover, tool-use behavior is strongly model-dependent, and perceived need and utility often diverge from their true values. This misalignment explains much of the persistent gap between self-decisions and the optimal. Under affordability constraints, models struggle to prioritize the calls with the greatest benefit, resulting in worse task performance under a limited budget.

To improve tool-call decisions, we train lightweight latent estimators of true need (LNE) and utility (LUE) from model hidden states~\citep{orgad2025llms,snyder2024early}. LNE gets higher accuracy than the model's self-perception and ranks instances by predicted need and improves average budgeted performance across all six open-source models, for both web search and calculator tasks. In contrast, LUEs do not reliably predict the true utility, leaving utility estimation as a key open challenge.

In summary, our paper makes three contributions:
(i) Inspired by Rational Choice Theory, we introduce a framework that separates \textit{need}, \textit{utility}, and \textit{affordability}, enabling instance-level comparisons between normative (oracle) and model-perceived tool-calling decisions.
(ii) Across seven models, two tools, and six tasks, we show that tool use is not universally beneficial and that models systematically misjudge when tools are needed and useful. These errors lead to unnecessary costs, harmful calls, and poor allocation under limited budgets.
(iii) We develop lightweight latent controllers from model hidden states. LNEs improve budgeted allocation across models and tools, while LUEs remain smaller improvements, highlighting utility estimation as a key open challenge.

\section{Related Work}

\textbf{Tool-Augmented and Cost-Aware LLMs.}
LLMs can use external tools such as web search, calculators, and APIs \citep{10.5555/3666122.3669119,qin2023toolllm}. Research has progressed from integrating tools into reasoning \citep{yao2023react} to learning self-directed selection policies through prompting, grounding, execution feedback, fine-tuning, and reinforcement learning \citep{lu-etal-2024-gear,qiao-etal-2024-making,singh2025agentic,qian2025toolrl,feng2025retoolreinforcementlearningstrategic,jin2025search,li2025torl,wang2025otc,eisenstein2025don}. Benchmarks evaluate aggregate tool-use performance and calling decisions \citep{qin2023toolllm,li2023api,patil2024gorilla,huang2023metatool,ross2025when2call}; see \citet{qu2025tool} for a survey. Recent work also considers cost-aware planning and budget constraints \citep{wu2025catp,liu2025budget}. Concurrent work identifies overuse of task-irrelevant tools \citep{zeng2026tool}; we show that models can also overuse relevant tools. \textit{Unlike evaluations centered on aggregate success or global cost--performance trade-offs, we ask whether each call is necessary, beneficial, and worth its cost.}

\textbf{Web Search and Retrieval Augmentation.}
Web search and retrieval-augmented generation can improve grounding \citep{lewis2020retrieval,liu2023webglm,xue2025illusion,kirsten2025characterizing}, but retrieval can also add noise, latency, and nondeterminism \citep{3766078.3766168,kirsten2025characterizing,mallen-etal-2023-trust}. Adaptive retrieval therefore uses uncertainty to decide when to retrieve \citep{jiang2023active,su2024dragin}, drawing on fine-tuned uncertainty signals \citep{kadavath2022language,amayuelas2024knowledge,kapoor2024large,moskvoretskii-etal-2025-adaptive,qian2025smart} or auxiliary measures such as factuality, semantic entropy, and self-assessed confidence \citep{gottesman2024estimating,kossen2024semantic,chen2025query}. These signals support adaptive RAG systems \citep{jeong2024adaptive,ding2025rowen,yao2025seakr,searcho1,asai2024selfrag}, and evaluation typically follows retrieval, judging the retrieved evidence itself \citep{gou2023critic}. This line of work targets one stochastic, knowledge-seeking tool, using uncertainty as a proxy for \textit{whether to call}. \textit{Our framework instead spans any tool, stochastic (web search) or deterministic (calculator), evaluated through need, utility, and affordability.}

\section{Operating Frameworks for Assessing and Optimizing Tool Calling}
\label{sec:framework}

\paragraph{Formal Problem Setup}
We now formalize the tool-calling problem motivated in the introduction and introduce the notation used throughout our analysis. For each task instance $x$ and available tool $\mathcal{F}$, we evaluate the model under three setups that differ in whether tool use is permitted, required, or chosen autonomously:\\
\textbf{(1) \notool}: the model receives only $x$ and responds without access to the tool;\\
\textbf{(2) \withtool}: the model receives $x$ and is required to use the tool;\\
\textbf{(3) \auto}: the model receives $x$ and the tool-related instructions, then decides autonomously whether to invoke the tool.

The \notool{} and \withtool{} setups provide counterfactual reference outcomes for evaluating the decision made under \auto{}. Let $s^{\textsc{NT}}(x)$ and $s^{\textsc{AT}}(x)$ denote the model's performance scores under \notool{} and \withtool{}, respectively, normalized to $[0,1]$. Comparing these scores reveals whether tool use improves performance on instance $x$, while comparing this evidence with the model's \auto{} decision reveals whether the model calls the tool when it is beneficial. Together, the three setups separate the value of tool use from the model's ability to decide when to use it.

These setups support three complementary analyses: the \emph{normative} lens defines optimal decisions, the \emph{descriptive} lens examines actual decisions, and the \emph{prescriptive} lens develops controllers to better align the two.

\paragraph{Normative lens: what should the model do?}
The two counterfactual outcomes define true need and true utility. We label an instance as requiring the tool, $\mathrm{N}^\star(x)=1$, when its \notool{} score falls below the acceptable-quality threshold. We define the tool's utility as the performance difference $\Delta^\star(x)=s^{\textsc{AT}}(x)-s^{\textsc{NT}}(x)$.

\textbf{True Need} $\mathrm{N}^\star(x)$ indicates whether the model can attain acceptable quality without assistance, whereas \textbf{True Utility} $\Delta^\star(x)$ measures the realized effect of using the available tool: utility is positive when $\Delta^\star(x)>0$, neutral when $\Delta^\star(x)=0$, and negative when $\Delta^\star(x)<0$.

Absent costs, the hindsight normative policy calls the tool when $\Delta^\star(x)>0$. This is an upper bound that observes both realized outcomes, not a deployable policy with advance knowledge of the better outcome. When calls have a cost, it uses the tool only when the expected gain exceeds that cost. Under a budget of at most $K$ calls, the optimal allocation selects the $K$ largest positive gains. Denoting this set by $\mathcal{S}_K^\star$, its total gain is $\mathrm{Gain}_K^\star=\sum_{x\in\mathcal{S}_K^\star}\Delta^\star(x)$. We refer to this cost-constrained allocation as \textbf{True Affordability}.

\paragraph{Descriptive lens: what does the model do?}
We characterize the model's beliefs and decisions under self-directed tool use. Under \notool{}, we ask whether it needs external assistance; under \auto{}, we observe whether it invokes the available tool.

\textbf{Perceived Need} reflects whether the model believes it can answer $x$ without assistance. \textbf{Perceived Utility} is inferred from its action: calling the tool indicates that the model expects the tool to be useful.

Under a budget of $K$ calls, we retain the first $K$ instances for which the model invokes the tool. Denoting this set by $\widehat{\mathcal{S}}_K$, its realized gain is $\widehat{\mathrm{Gain}}_K=\sum_{x\in\widehat{\mathcal{S}}_K}\Delta^\star(x)$. We refer to this allocation of the model's calls under a limited budget as \textbf{Perceived Affordability}.

\paragraph{Prescriptive lens: how can decisions be improved?}
The gap between normative and descriptive decisions motivates a controller that predicts true need and utility before a tool call. We train lightweight multilayer perceptron (MLP) classifiers using the model's final-token hidden state.

\textbf{Latent Need Estimator (LNE).}
LNE predicts true need $\mathrm{N}^\star(x)$ from the hidden state obtained under \notool{}.

\textbf{Latent Utility Estimators (LUEs).}
LUEs predict whether a tool call will have positive utility.

We do not train a separate affordability estimator. Under a budget of $K$ calls, we rank instances by LNE and LUE confidence and allocate calls to the top $K$, then compute the realized gain from their true marginal gains $\Delta^\star(x)$.
Detailed definitions and training details are provided in Appendix~\ref{sec:controller_details}.

 \section{Experiments and Findings}
\label{sec:experiments}

We first describe our experimental setup and then analyze \textit{need}, \textit{utility}, and \textit{affordability} through normative (optimal) and descriptive lenses. Their systematic misalignment explains the performance gap between \optimized{} and \auto{} in Table~\ref{tab:main}. Finally, we show that a simple binary classifier trained on internal representations improves the accuracy of tool-use decisions and task performance under budget constraints.

\textbf{Models and Tools.}
We evaluate six open-source models from five families (3B--120B parameters), covering instruction-tuned and reasoning variants, and one closed-source frontier model from OpenAI. Table~\ref{tab:model_names} lists all models and their links where applicable.
We study two complementary tools. \textit{Web search} tool is complex, stochastic, and costly, and may return noisy evidence to the model. Our FastMCP server uses Google Search (SerpApi) and returns up to five results, each containing a title, snippet, and URL. We replicate the Entity Task with Perplexity Search, Brave Search, and Tavily Search and obtain consistent results (Appendix~\ref{sec:perplexity}).
The \textit{calculator} tool is local, deterministic, and free. Its output is unambiguous, but invoking it still requires the model to identify the relevant computation and construct a valid expression under a certain schema. The two tools, therefore, test distinct capabilities that will influence their utility: retrieval and evidence use in web search, and precise input formulation for the calculator.

\textbf{Datasets and Performance Metrics.} We evaluate each tool on three datasets. For web search, \textit{Entity} (500 samples) asks models to describe entities drawn from real-world chat logs, while \textit{InVivoQuery} (500 samples) contains factual questions derived from real user requests~\citep{karnam2026bowling}. Both are open-ended and reflect realistic variations. We evaluate their factuality, completeness, and relevance~\cite{li2024llms}. For factuality, we
follow the long-form factuality work~\citep{song-etal-2024-veriscore}; an LLM judge extracts and verifies claims. A limited human check found high class-imbalanced raw endorsement, as shown in Appendix~\ref{sec:human}. The same judge rates completeness and relevance on five-point Likert scales.
The third dataset, \textit{BFCL} (Berkeley Function Calling Leaderboard), tests function calling and tool use~\citep{patil2025bfcl}. We decompose its multi-hop search questions into 314 atomic questions. Because the reference answers are short and unambiguous, we score them with an LLM judge against ground truth without additional human annotation~\citep{zheng2023judging}.

For the calculator, we use three arithmetic tasks. \textit{GSM-Hard}~\citep{gao2023pal} (1319 samples) combines multi-step reasoning with exact arithmetic by replacing the small integers in GSM8K-style problems with large operands. \textit{Synthetic Multiplication} (1000 samples) isolates calculation by presenting products of 2--7 factors, each 1--4 digits long. \textit{Synthetic Large-Digit Multiplication} (1000 samples) asks models to square a 4--40-digit operand (22 digits on average), pushing calculation beyond reliable unaided computation. Together, these tasks span reasoning-bound to scale-bound need, complementing the knowledge-bound need studied with web search. We score all calculator tasks by exact match, allowing true need and utility to be computed directly from correctness.

Dataset construction details and examples are provided in Appendices~\ref{sec:entity_dataset} and~\ref{sec:calculator_dataset} for the web-search and calculator tasks, respectively. Implementation details of the scoring procedures are provided in Appendix~\ref{sec:score}.

\textbf{Experimental Setup.} We serve open-source models with vLLM~\citep{kwon2023efficient} and query the closed-source model through OpenAI's Responses API.\footnote{\url{https://developers.openai.com/api/reference/responses/}} We use temperature 0 for open-source models and temperature 1 for the closed-source model, for which temperature 0 is unavailable, and cap generations at 512 tokens. We apply the same two-stage protocol to every task: the model first decides whether to invoke the available tool and then generates a final response conditioned on the tool output when invoked. Each model uses one fixed harness throughout; only the tool description, input schema, and returned output vary across tools. GPT-OSS-120B, Qwen3-30B-A3B, Qwen3-30B-IT, Llama3.2-3B-IT, and Mistral-Small-3.1-24B-IT use a \textit{trained harness} based on their native tool-calling chat templates, which preserve the full execution trace. Gemma3-27B-IT uses a \textit{customized harness} with a manually prompted decision step because it does not reliably expose a native function-calling schema. For GPT-5.5, the Responses API constructs the model-facing harness and manages the tool-call protocol and execution trace. All harnesses expose the same \notool{}/\auto{}/\withtool{} modes.
Additional details are provided in Appendix~\ref{apd:two-stage}.

\subsection{Need and Utility}
\label{sec:need_utility}

\begin{figure}[t]
    \centering
    \vspace{-10pt}
    \includegraphics[width=0.7\linewidth]{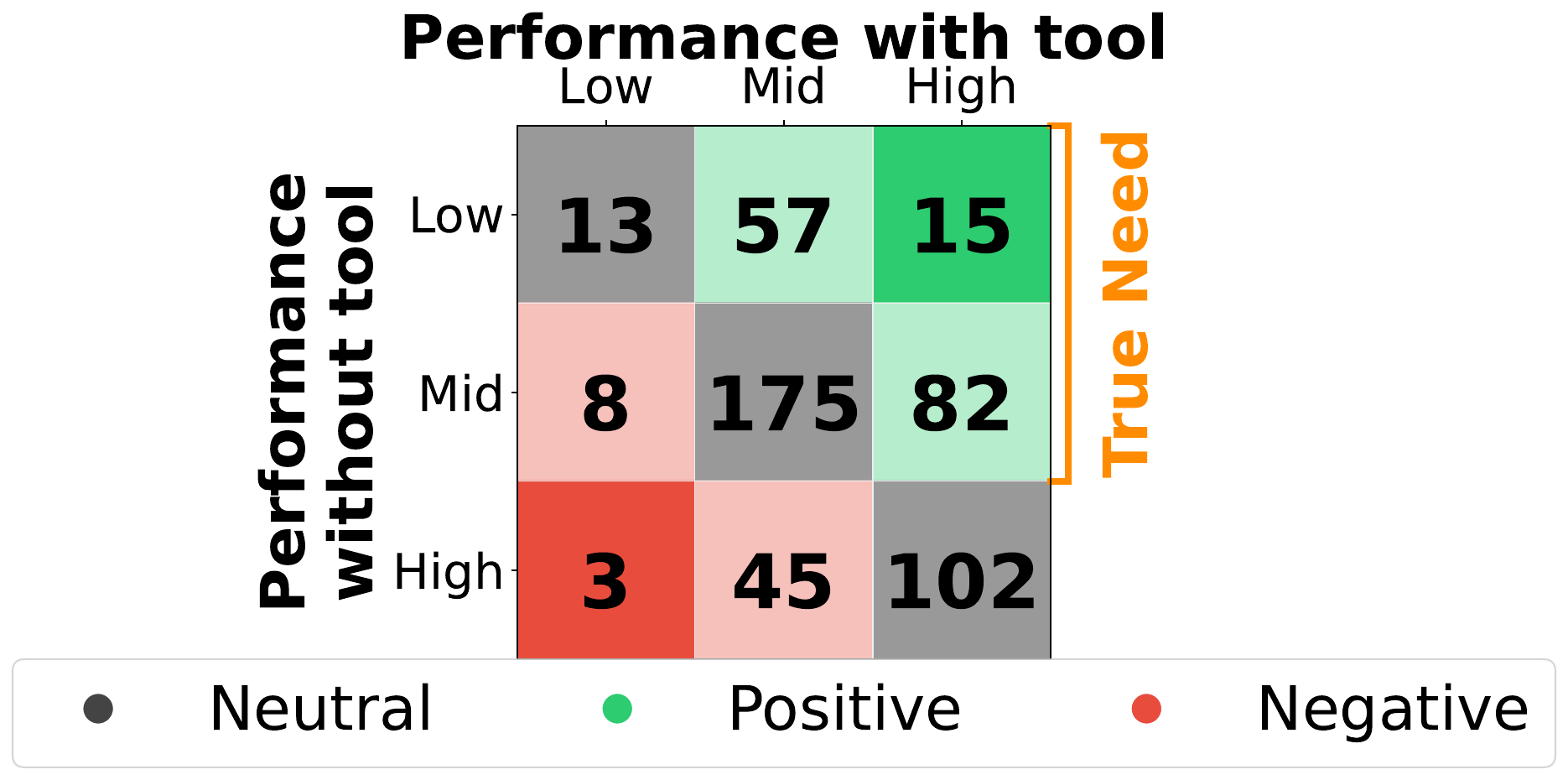}
\caption{
\textbf{True need and true positive utility are correlated, but not perfectly aligned.}
Rows are grouped by the model’s (GPT-OSS-120B) factuality scores under \notool{} (parametric knowledge), while columns show scores under \withtool{}. Scores are bucketed into low (0--0.1), mid (0.1--0.9), and high (0.9--1).
Cells above the diagonal indicate {\color{mygreen}\textit{positive utility}}, while those below indicate {\color{red}\textit{negative utility}}. The bracket highlights the {\color{orange}\textit{true need}} region, comprising both low and mid \notool{} scores.
}
    \label{fig:main-actual-need-utility}
    \vspace{-10pt}
\end{figure}

\textbf{Normative Lens: Measuring True Need and Utility.}
We operationalize \textit{True Need} and \textit{True Utility} by comparing model performance under \notool{} and \withtool{}. For open-ended tasks (the Entity and InvivoQuery task), \textit{True Need} corresponds to instances where \notool{} performance is \textit{Low} or \textit{Mid}, indicating that an external tool is necessary (Figure~\ref{fig:main-actual-need-utility}). \textit{True Positive Utility} captures performance improvements from no tool to with tool (e.g., \textit{Low} $\rightarrow$ \textit{Mid/High}), while \textit{True Negative Utility} reflects performance degradation; instances with no change are categorized as neutral utility. For tasks with concrete, verifiable answers (BFCL and the three calculator tasks), we instead use correctness (0 or 1) to define \textit{True Need}.

\textbf{Tool calls can hurt performance when there is no true need.}
Across models and tasks, we observe three consistent patterns (Figure~\ref{fig:main-actual-need-utility}).
First, tool calling is often effective in the \textit{True Need} regime, where \notool{} performance is low or mid. In this region, 44\% instances improve ($({\color{mygreen}57}+{\color{mygreen}15}+{\color{mygreen}82})/350$).
Second, tool calls can degrade performance when the model already performs well without them. When \notool{} performance is \textit{High}, 32\% instances degrade ($({\color{red}3}+{\color{red}45})/150$), indicating unnecessary or harmful intervention.
Third, 58\% instances ($({\color{mygray}13}+{\color{mygray}175}+{\color{mygray}102})/500$) lie on the diagonal, suggesting that tool calling is often redundant.
These patterns are observed across all models and web-search tasks (Figures~\ref{fig:entity_all_actul_need_utility}, \ref{fig:all_invivo_actul_need_utility}, and \ref{fig:all_bfcl_actul_need_utility}). The calculator tasks show the same central association between true need and positive utility (Figures~\ref{fig:gsmhard_actual_need_utility}, \ref{fig:synthetic_mult_actual_need_utility}, and \ref{fig:synthetic_nn_actual_need_utility}), but also reveal task-specific behavior. On the two synthetic multiplication tasks, a successfully used calculator usually eliminates arithmetic error and therefore produces little negative utility. GSM-Hard, which also requires reasoning and tool-output integration, retains both gains and regressions. Llama3.2-3B-IT is the clearest exception: its calculator integration frequently fails even on high-need instances, showing that tool availability alone does not guarantee positive utility. Detailed score distributions for the web-search datasets are shown in Figures~\ref{fig:entity_hist}, \ref{fig:invivo_hist}, and \ref{fig:bfcl_hist}.

 Taken together, these results show that the observation holds across all evaluated tasks, tools, and models.
 We focus on factuality in the main paper. To demonstrate that our evaluation framework generalizes beyond a single metric, we additionally report completeness and relevance results for GPT-OSS-120B on the Entity task in Appendix~\ref{sec:other_metrics}. These key findings remain consistent across evaluation metrics.
Overall, true need and utility define a normative ideal for tool use: a model should invoke tools only when they yield positive utility. We provide an example where web search has negative utility in Appendix~\ref{sec:hurts}.  

\begin{figure}[t]
\centering
\vspace{-5pt}
\begin{subfigure}{0.3\linewidth}
    \centering
    \includegraphics[width=\linewidth]{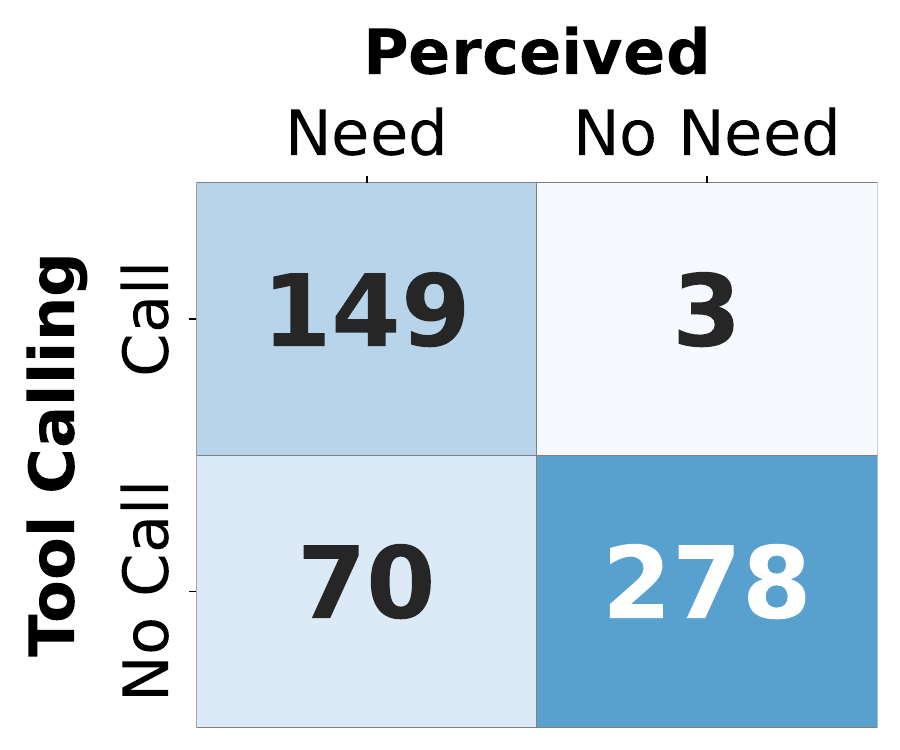}
    \caption{GPT-OSS}
\end{subfigure}\hfill
\begin{subfigure}{0.3\linewidth}
    \centering
    \includegraphics[width=\linewidth]{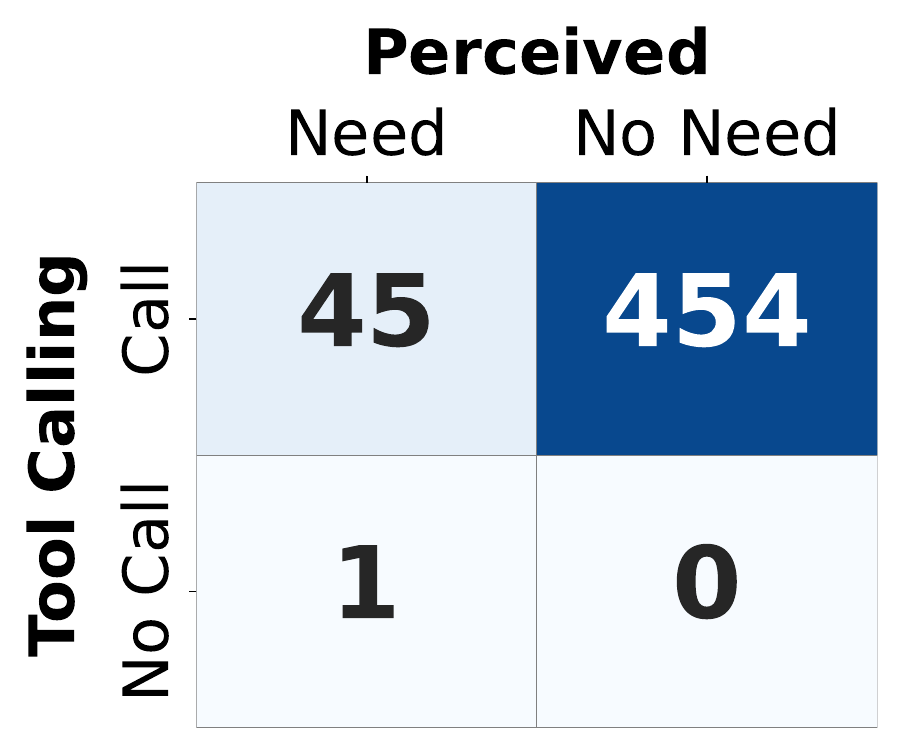}
    \caption{Llama3.2}
\end{subfigure}\hfill
\begin{subfigure}{0.3\linewidth}
    \centering
    \includegraphics[width=\linewidth]{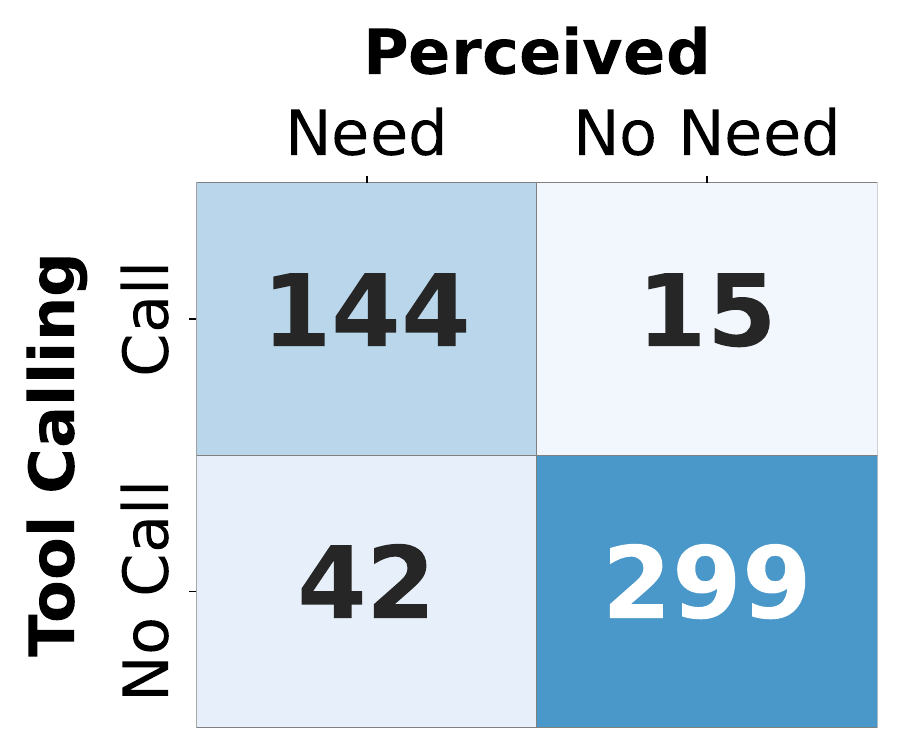}
    \caption{GPT-5.5}
\end{subfigure}

\caption{\textbf{The alignment between perceived need and tool calling is model-dependent.} The x-axis shows perceived need; the y-axis shows the tool-call decision (perceived utility).}
\label{fig:main_figure_need_utility}
\vspace{-10pt}
\end{figure}

 \textbf{Descriptive Lens: Measuring Perceived Need and Utility.}
To estimate perceived need, we test whether the model requires external assistance under the \notool{} setting. We design three prompt variants: (1) a structured JSON response, (2) a direct question (“Do you know the answer?”), and (3) “Do you need help?” We then analyze the responses to measure perceived need. All prompts exclude tool-related information, isolating the model’s knowledge of the tool.
We measure \textit{perceived utility} through tool-calling behavior under the \auto{} setting, where the model is provided with both the tool description and the task input. \textit{A tool invocation is interpreted as an indication of perceived positive utility}.
Full prompts are provided in Appendix~\ref{sec:experimental_setup}.

\textbf{Models’ perceived need does not consistently predict their tool-use behavior (perceived utility).}
Figure~\ref{fig:main_figure_need_utility} highlights three different patterns: GPT-OSS-120B largely couples tool calls to stated need, Llama3.2-3B-IT shows a markedly different joint distribution, and GPT-5.5 sometimes calls despite reporting no need. Qwen3-30B-A3B, Qwen3-30B-IT, Gemma3-27B-IT, and Mistral3.1-24B-IT exhibit further model-specific patterns; their full distributions are reported in Appendix Figure~\ref{fig:main_need_utility_combined_v12}. Thus, perceived need can correlate with action without reliably predicting it for every model.
This model-dependent behavior generalizes beyond the Entity task to InVivoQuery and BFCL (Figures~\ref{fig:invivo_need_utility_combined_v12} and~\ref{fig:bfcl_need_utility_combined_v12}) and to the calculator tasks GSM-Hard, Synthetic Multiplication, and Synthetic Large-Digit Multiplication (Figures~\ref{fig:gsmhard_perceived_need_utility}, \ref{fig:synthetic_mult_perceived_need_utility}, and~\ref{fig:synthetic_nn_perceived_need_utility}, respectively); the complete results are provided in Appendix~\ref{sec:appendix_descriptive_lens}.

\begin{figure}[t]
    \centering
    \vspace{-10pt}
    \includegraphics[width=0.6\linewidth]{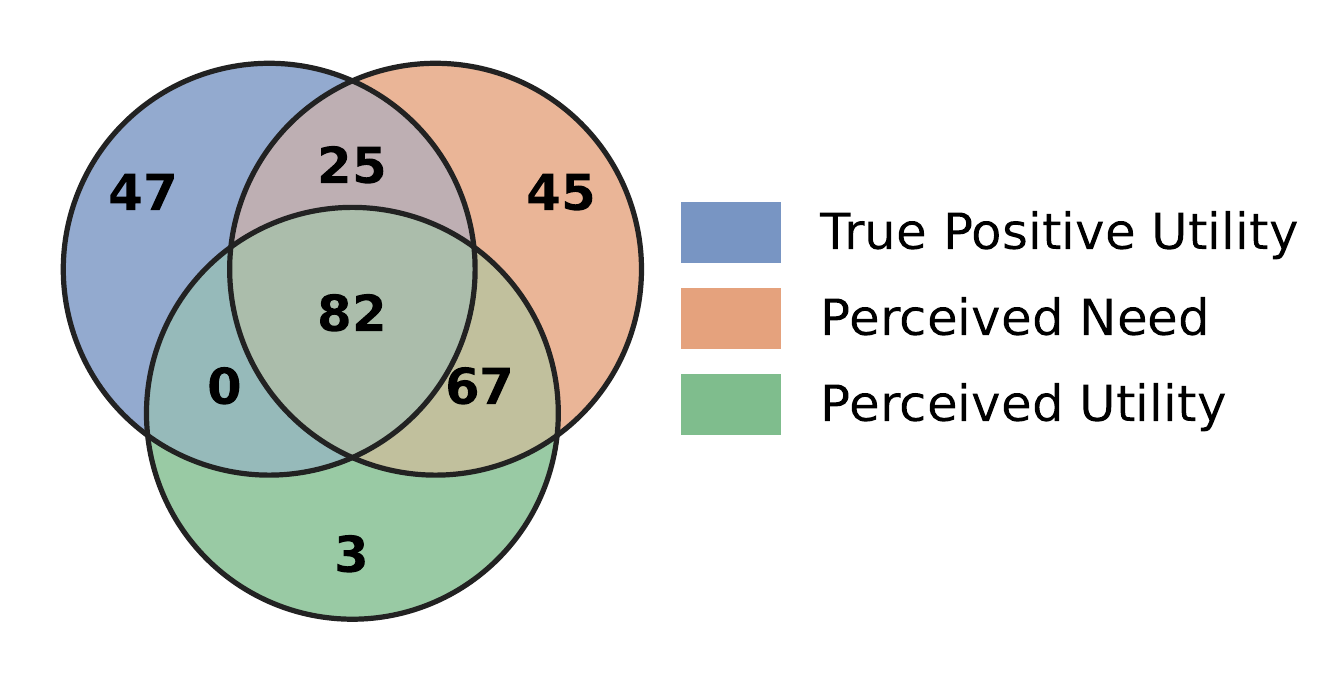}
    \caption{
    \textbf{Perceived signals only partially align with true utility.}
    Venn diagrams of \textit{True Positive Utility}, \textit{Perceived Need}, and \textit{Perceived Utility} for GPT-OSS-120B on the entity task. Calls outside true positive utility are non-beneficial, while positive-utility instances outside perceived utility are missed opportunities. Perceived need is a separate self-assessment and is not assumed to be nested within true positive utility.}
    \label{fig:venn-gpt}
    \vspace{-5pt}
\end{figure}

\begin{figure}[t]
\centering
{\small\textbf{Need}\par}
\begin{subfigure}{0.3\linewidth}
    \centering
    \includegraphics[width=\linewidth]{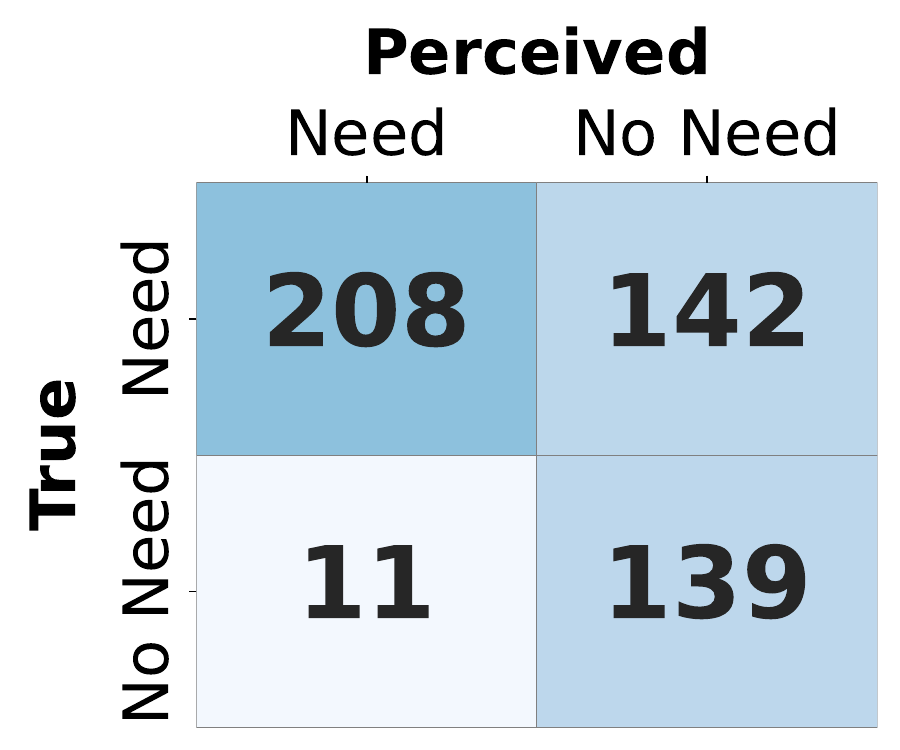}
    \caption{GPT-OSS}
\end{subfigure}\hfill
\begin{subfigure}{0.3\linewidth}
    \centering
    \includegraphics[width=\linewidth]{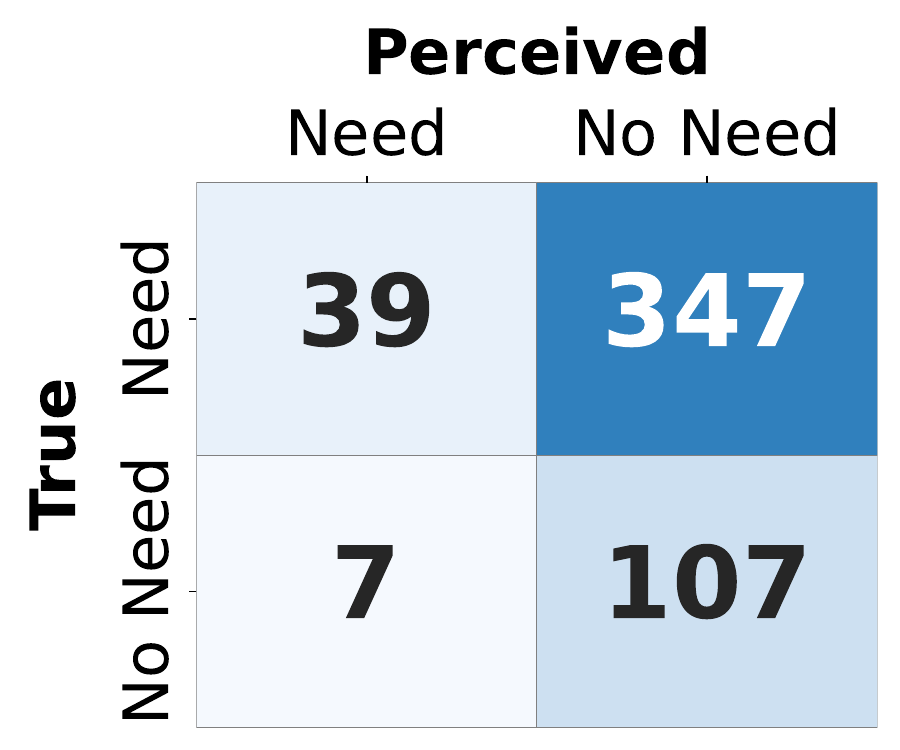}
    \caption{Llama3.2}
\end{subfigure}\hfill
\begin{subfigure}{0.3\linewidth}
    \centering
    \includegraphics[width=\linewidth]{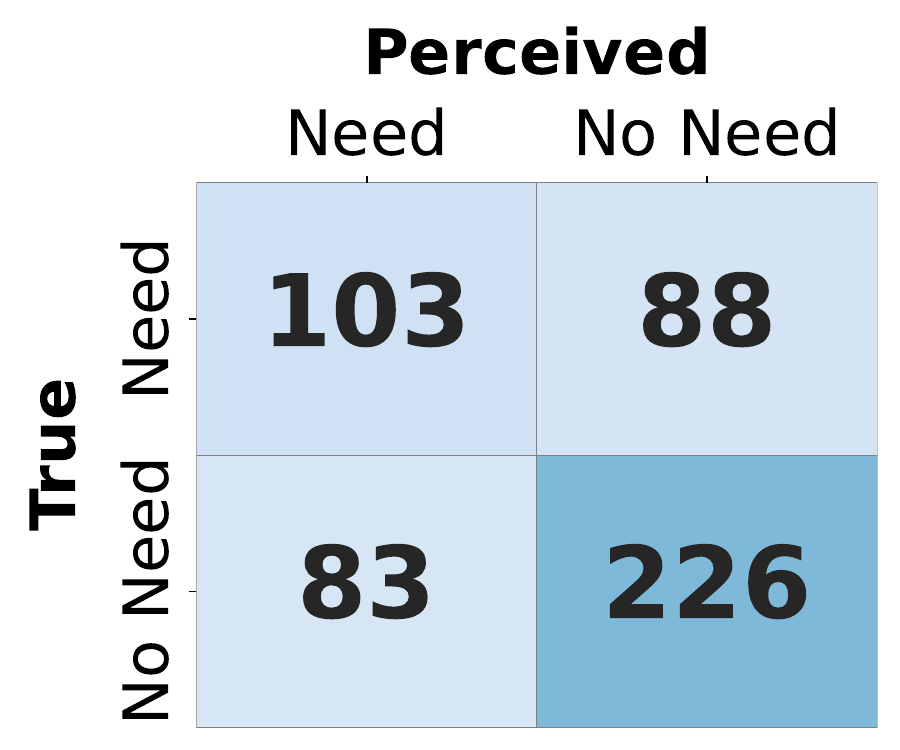}
    \caption{GPT-5.5}
\end{subfigure}

{\small\textbf{Utility}\par}
\begin{subfigure}{0.3\linewidth}
    \centering
    \includegraphics[width=\linewidth]{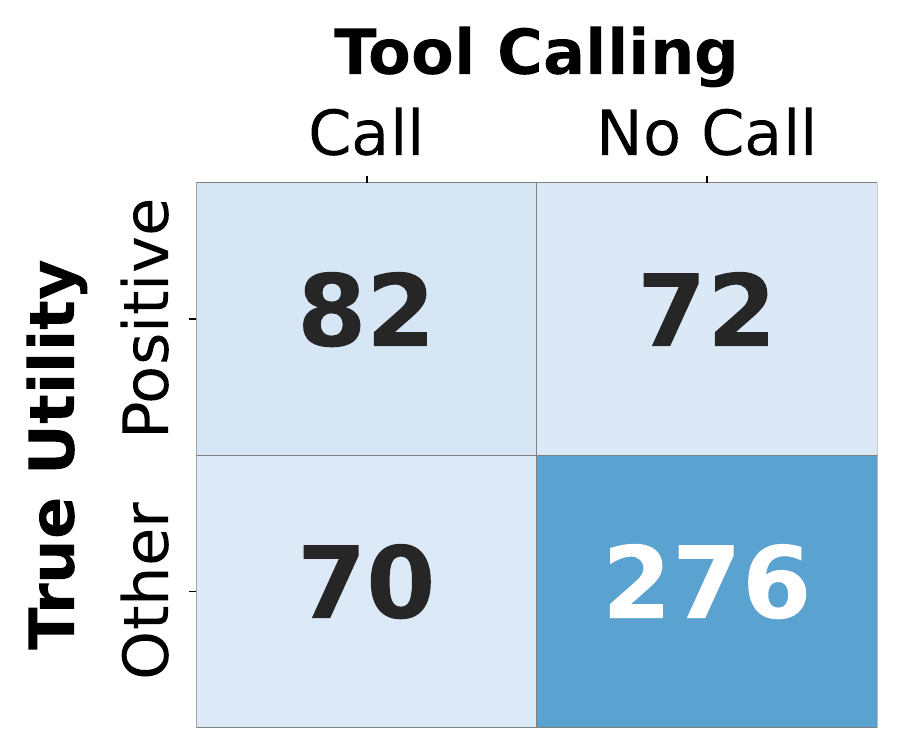}
    \caption{GPT-OSS}
\end{subfigure}\hfill
\begin{subfigure}{0.3\linewidth}
    \centering
    \includegraphics[width=\linewidth]{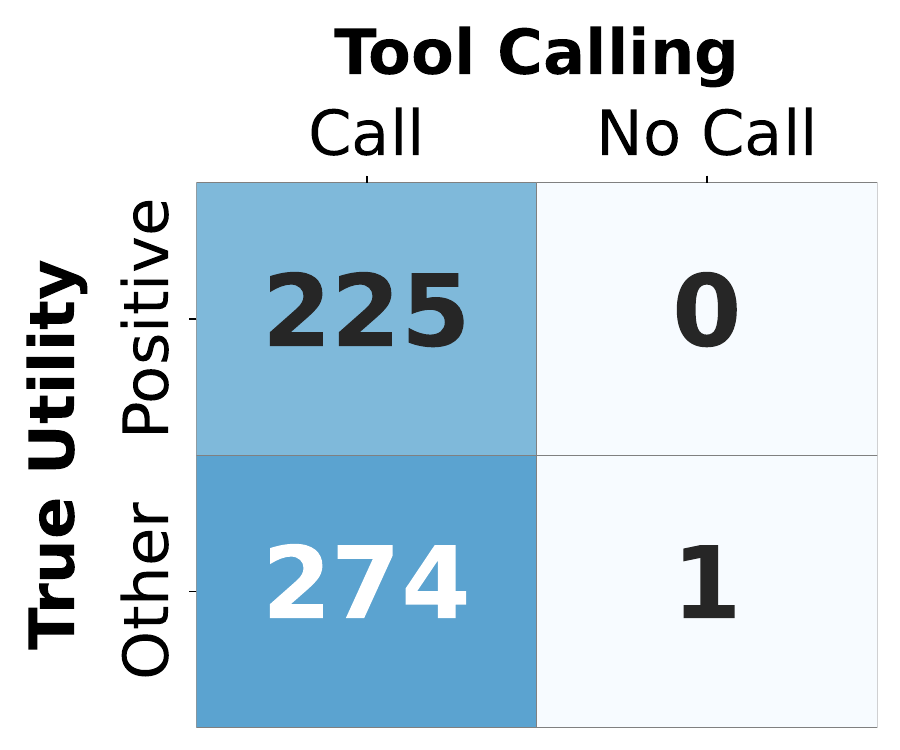}
    \caption{Llama3.2}
\end{subfigure}\hfill
\begin{subfigure}{0.3\linewidth}
    \centering
    \includegraphics[width=\linewidth]{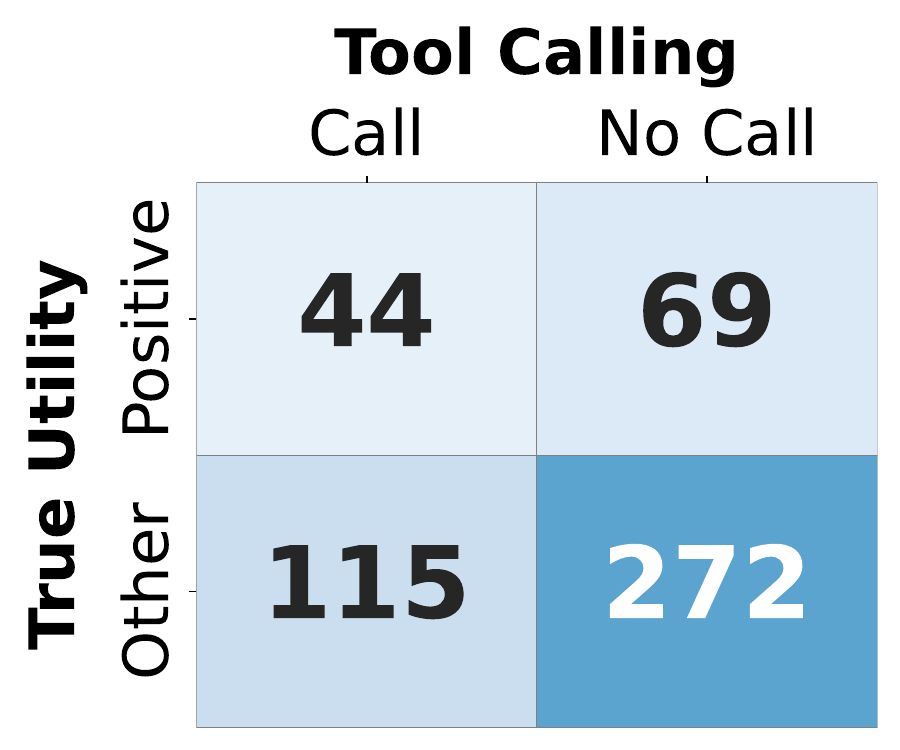}
    \caption{GPT-5.5}
\end{subfigure}

\caption{
\textbf{The perceived need and utility are not aligned with the true need and utility.}
Entity task. Additional models are shown in Figure~\ref{fig:entity_true_perceived_additional}.
}
\label{fig:entity_true_perceived}
\vspace{-10pt}
\end{figure}

\textbf{Models' perceptions do not align with true need and utility.}
As illustrated in Figure~\ref{fig:venn-gpt}, there is a clear mismatch between perceived need/utility and \textit{true positive utility}. Consequently, the model’s perceptions are insufficient for optimal decision-making. 
The same observation holds for other models and web-search tasks in Figures~\ref{fig:venn-all}, \ref{fig:venn-invivo}, \ref{fig:venn-bfcl}, and for calculator tasks in Figures~\ref{fig:venn-gsmhard}, \ref{fig:venn-synthetic-mult}, and~\ref{fig:venn-synthetic-nn}.
To better understand this phenomenon, we provide task-level breakdowns in Figure~\ref{fig:entity_true_perceived}. Llama3.2's perceived need is decoupled from its actual calling behavior; GPT-OSS judges need well but executes utility poorly; GPT-5.5 misjudges need in both directions, yielding both wasted and missed calls. Aggregate accuracy would mask all three of these distinct failure patterns, which is exactly the motivation for the per-instance, per-model diagnostic lens.
The similar observation holds for other tasks: \textit{InVivoQuery} (Figure~\ref{fig:invivo_true_perceived}), \textit{BFCL} (Figure~\ref{fig:bfcl_true_perceived}), \textit{GSM-Hard} (Figure~\ref{fig:gsmhard_true_perceived}), \textit{Synthetic Multiplication} (Figure~\ref{fig:synthetic_mult_true_perceived}), and \textit{Synthetic Large-Digit Multiplication} (Figure~\ref{fig:synthetic_nn_true_perceived}).

Across all settings, the reported accuracy results and joint distributions reveal consistent misalignment between perceived and true need, as well as between perceived utility (i.e., tool-calling decisions) and true utility (i.e., positive versus negative or neutral). The calculator results make the latter gap especially stark: several models call on nearly every synthetic-arithmetic instance, including instances with no need and negative utility. Thus, even when the tool is deterministic and directly relevant, models do not reliably distinguish potential benefit from mere tool availability.
Overall, these findings indicate that \textit{LLMs are unreliable judges of when tool call is necessary or beneficial}.

\subsection{Cost and Affordability}
\label{sec:main-affordability}
When the available budget permits tool use on only a subset of instances, effective tool calling requires deciding not only \emph{whether} a tool is useful, but also \emph{which} instances should receive the limited calls. Following Section~\ref{sec:framework}, for a budget of at most $K$ calls, \emph{normative affordability} selects the set $\mathcal{S}_K^\star$ containing the $K$ largest positive true utility gains $\Delta^\star(x)$ and achieves $\mathrm{Gain}_K^\star=\sum_{x\in\mathcal{S}_K^\star}\Delta^\star(x)$. In contrast, \emph{descriptive affordability} retains the first $K$ instances on which the model autonomously invokes the tool, denoted $\widehat{\mathcal{S}}_K$, and realizes $\widehat{\mathrm{Gain}}_K=\sum_{x\in\widehat{\mathcal{S}}_K}\Delta^\star(x)$. Comparing these allocations reveals four recurring limitations: (1) a persistent gap from the normative utility gain, (2) uneven and sometimes counterproductive responses to cost information, (3) increasingly poor prioritization of high-utility instances as the budget grows, and (4) frequent violations of explicit budgets or failures to track implicit ones. These results suggest that models need external guidance to select and enforce tool calls under a budget. We provide the experimental setup and detailed affordability results in Appendix~\ref{sec:affordability}.

\subsection{The Controller Framework}
\label{sec:controller}

\begin{table*}[t]
\centering
\scriptsize
\begin{adjustbox}{
  max width=0.99\textwidth,
  max totalheight=0.84\textheight,
  keepaspectratio
}
\begin{tabular}{
  >{\raggedright\arraybackslash}p{2.25cm}
  >{\centering\arraybackslash}p{1.8cm}
  cccccccc
}
\toprule
\textbf{Model}
& \textbf{Task}
& \multicolumn{2}{c}{\textbf{Natural}}
& \multicolumn{2}{c}{\textbf{20\% budget}}
& \multicolumn{2}{c}{\textbf{40\% budget}}
& \multicolumn{2}{c}{\textbf{80\% budget}} \\
\cmidrule(lr){3-4}
\cmidrule(lr){5-6}
\cmidrule(lr){7-8}
\cmidrule(lr){9-10}
&
& \textbf{Self} & \textbf{LNE}
& \textbf{Self} & \textbf{LNE}
& \textbf{Self} & \textbf{LNE}
& \textbf{Self} & \textbf{LNE} \\
\midrule

\rowcolor{blue!8}
& Entity
& 0.72\rate{30}
& \textbf{0.76}\rate{72}
& \textbf{0.68}\rate{20}
& 0.66\rate{20}
& \textbf{0.72}\rate{30}
& \textbf{0.72}\rate{40}
& 0.72\rate{30}
& \textbf{0.76}\rate{72} \\
\rowcolor{blue!8}
\multirow{-2}{*}{\cellcolor{blue!8}\strut GPT-OSS-120B}
& GSM-Hard
& 0.65\rate{31}
& \textbf{0.67}\rate{27}
& 0.65\rate{20}
& \textbf{0.66}\rate{20}
& 0.65\rate{31}
& \textbf{0.67}\rate{27}
& 0.65\rate{31}
& \textbf{0.67}\rate{27} \\
\midrule

\rowcolor{blue!8}
& Entity
& 0.80\rate{56}
& \textbf{0.81}\rate{70}
& 0.74\rate{20}
& \textbf{0.75}\rate{20}
& 0.77\rate{40}
& \textbf{0.79}\rate{40}
& 0.80\rate{56}
& \textbf{0.81}\rate{70} \\
\rowcolor{blue!8}
\multirow{-2}{*}{\cellcolor{blue!8}\strut Qwen3-30B-A3B}
& GSM-Hard
& 0.61\rate{63}
& \textbf{0.62}\rate{34}
& \textbf{0.62}\rate{20}
& \textbf{0.62}\rate{20}
& \textbf{0.62}\rate{40}
& \textbf{0.62}\rate{34}
& 0.61\rate{63}
& \textbf{0.62}\rate{34} \\
\midrule

\rowcolor{blue!8}
& Entity
& \textbf{0.82}\rate{95}
& 0.81\rate{68}
& 0.71\rate{20}
& \textbf{0.73}\rate{20}
& 0.74\rate{40}
& \textbf{0.78}\rate{40}
& 0.80\rate{80}
& \textbf{0.81}\rate{68} \\
\rowcolor{blue!8}
\multirow{-2}{*}{\cellcolor{blue!8}\strut Qwen-3-30B-IT}
& GSM-Hard
& 0.59\rate{56}
& \textbf{0.60}\rate{38}
& \textbf{0.60}\rate{20}
& \textbf{0.60}\rate{20}
& \textbf{0.60}\rate{40}
& \textbf{0.60}\rate{38}
& 0.59\rate{56}
& \textbf{0.60}\rate{38} \\
\midrule

\rowcolor{blue!8}
& Entity
& 0.70\rate{0}
& \textbf{0.79}\rate{73}
& 0.70\rate{0}
& \textbf{0.73}\rate{20}
& 0.70\rate{0}
& \textbf{0.75}\rate{40}
& 0.70\rate{0}
& \textbf{0.79}\rate{73} \\
\rowcolor{blue!8}
\multirow{-2}{*}{\cellcolor{blue!8}\strut Mistral3.1-24B-IT}
& GSM-Hard
& \textbf{0.51}\rate{5}
& 0.43\rate{43}
& \textbf{0.51}\rate{5}
& 0.48\rate{20}
& \textbf{0.51}\rate{5}
& 0.43\rate{40}
& \textbf{0.51}\rate{5}
& 0.43\rate{43} \\
\midrule

\rowcolor{blue!8}
& Entity
& \textbf{0.70}\rate{100}
& 0.69\rate{89}
& 0.61\rate{20}
& \textbf{0.62}\rate{20}
& 0.64\rate{40}
& \textbf{0.67}\rate{40}
& 0.66\rate{80}
& \textbf{0.69}\rate{80} \\
\rowcolor{blue!8}
\multirow{-2}{*}{\cellcolor{blue!8}\strut Llama3.2-3B-IT}
& GSM-Hard
& 0.02\rate{97}
& \textbf{0.10}\rate{86}
& 0.14\rate{20}
& \textbf{0.17}\rate{20}
& 0.12\rate{40}
& \textbf{0.16}\rate{40}
& 0.04\rate{80}
& \textbf{0.11}\rate{80} \\
\midrule

\rowcolor{orange!12}
& Entity
& \textbf{0.80}\rate{92}
& 0.79\rate{89}
& 0.63\rate{20}
& \textbf{0.66}\rate{20}
& 0.67\rate{40}
& \textbf{0.71}\rate{40}
& \textbf{0.77}\rate{80}
& \textbf{0.77}\rate{80} \\
\rowcolor{orange!12}
\multirow{-2}{*}{\cellcolor{orange!12}\strut Gemma3-27B-IT}
& GSM-Hard
& 0.56\rate{94}
& \textbf{0.59}\rate{29}
& 0.58\rate{20}
& \textbf{0.59}\rate{20}
& 0.58\rate{40}
& \textbf{0.59}\rate{29}
& 0.56\rate{80}
& \textbf{0.59}\rate{29} \\

\bottomrule
\end{tabular}
\end{adjustbox}

\caption{
Task performance under \auto{} and LNE. We show only the Entity
(Web Search) and GSM-Hard (Calculator) tasks here; the full results for
all six tasks can be found in Table~\ref{tab:all}.
Each entry reports the task score, followed in parentheses by the actual
tool-call rate (\%). \auto{} and LNE are compared within the natural
setting and at each tool-call budget. The higher task score in each
paired comparison is bolded; ties are bolded for both policies.
Row colors indicate the harness:
\colorbox{blue!8}{\strut Trained} and
\colorbox{orange!12}{\strut Custom}.
}
\label{tab:lne-average-improvement}
\vspace{-10pt}
\end{table*}

\begin{figure}[t]
    \centering
    \begin{subfigure}{0.3\textwidth}
        \centering
        \includegraphics[width=\linewidth]{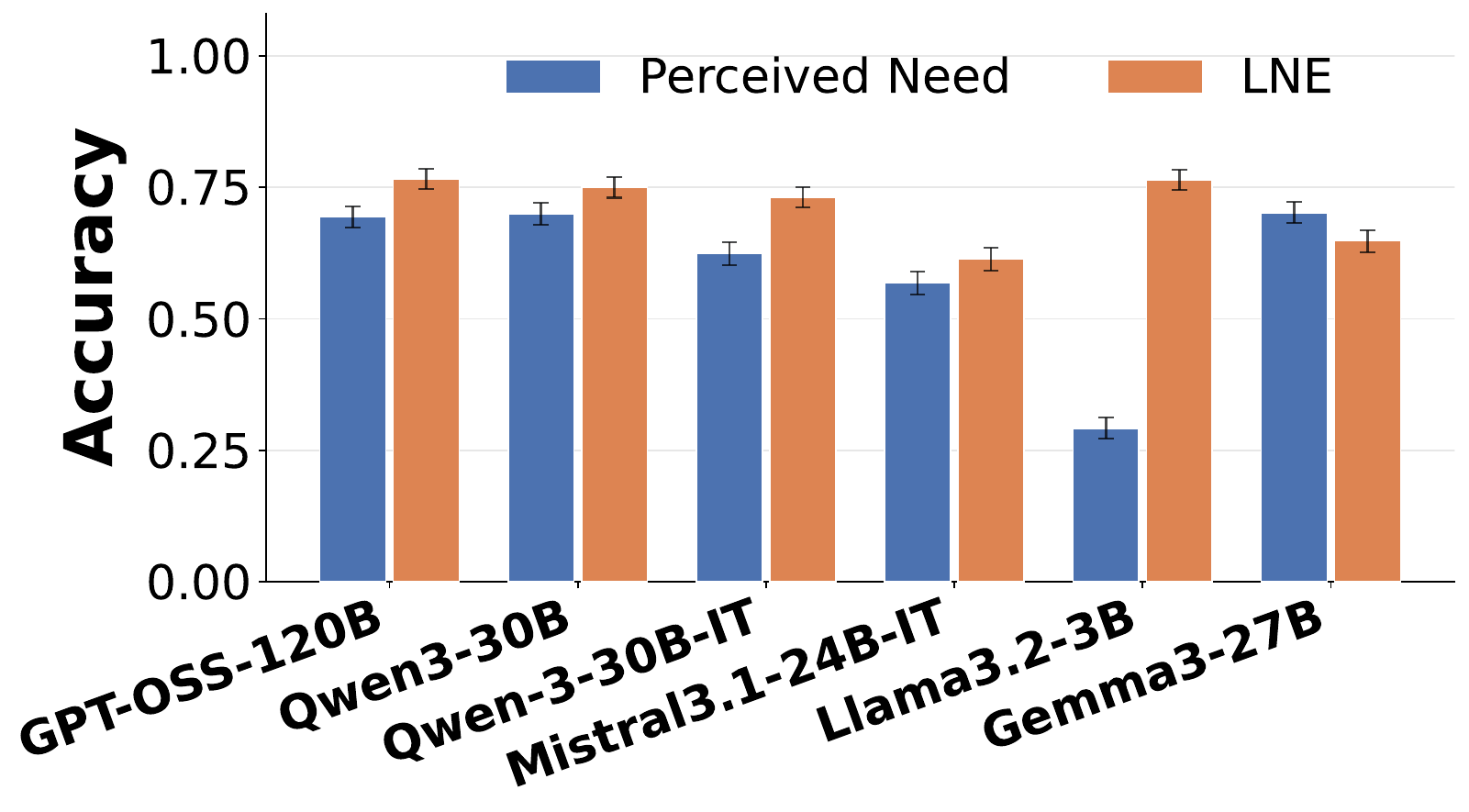}
        \caption{Entity}
        \label{fig:entity_lne_accuracy}
    \end{subfigure}\hfill
    \begin{subfigure}{0.3\textwidth}
        \centering
        \includegraphics[width=\linewidth]{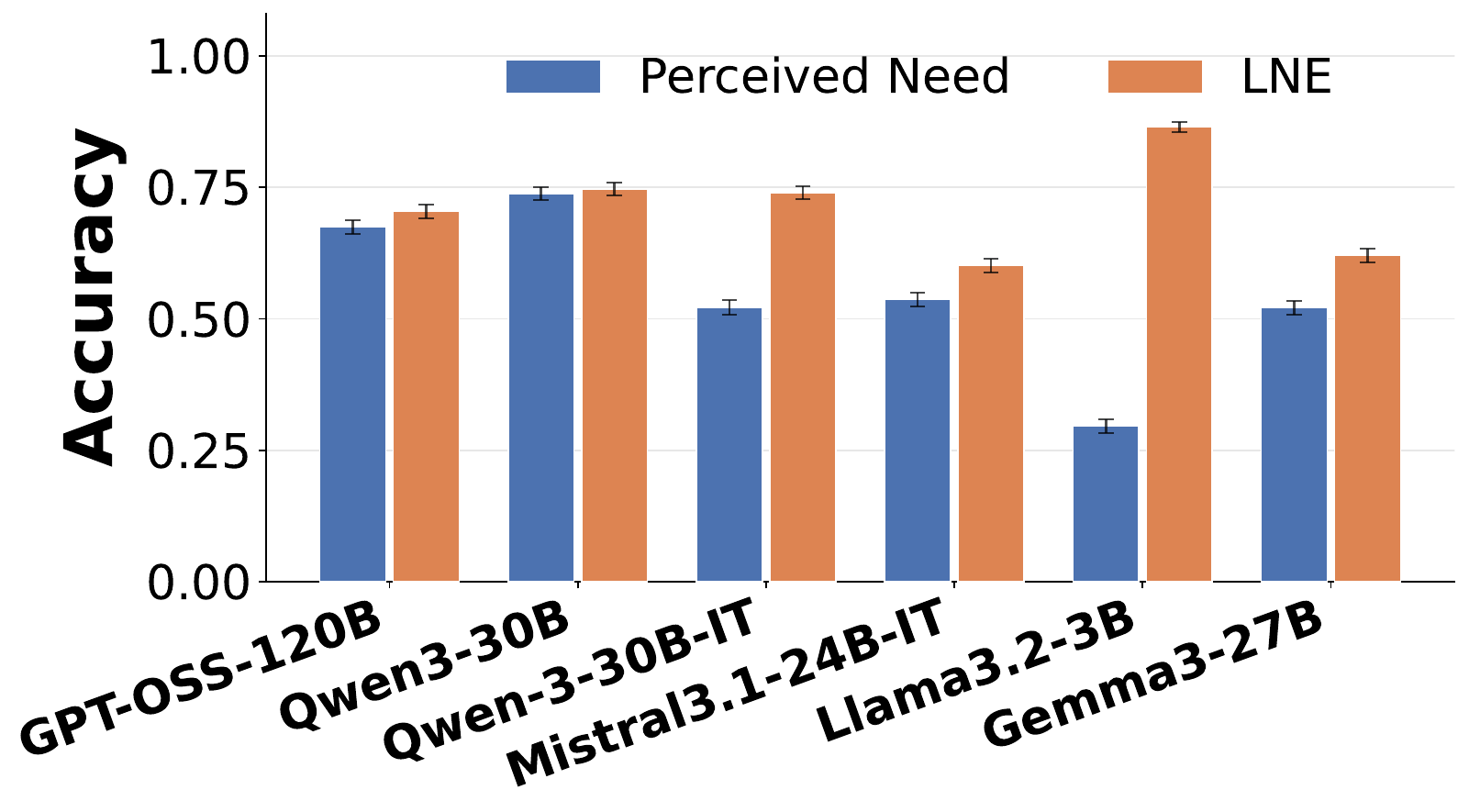}
        \caption{GSM-Hard}
        \label{fig:gsmhard_lne_accuracy}
    \end{subfigure}
    \caption{\textbf{LNE improves true-need prediction accuracy across most models, with the largest gains for smaller models.} Other tasks' results appear in Figures~\ref{fig: bfcl_lne} and~\ref{fig: invivo_lne}, and ~\ref{fig:calculator_lne_accuracy}.}
    \label{fig:lne}
    \vspace{-5pt}
\end{figure}

Prior results reveal a consistent misalignment between \textit{perceived} and \textit{true} need and utility, leading to suboptimal tool-call decisions, particularly under budget constraints. At the same time, prior works~\citep{snyder2024early, orgad2025llms} show that models encode useful signals about whether they know the answer, but fail to reliably express them in their outputs.

Motivated by this gap, we propose a \textit{prescriptive or control mechanism} that operates directly on latent representations to guide tool-call decisions. Our approach requires no fine-tuning and is applicable to any pretrained LLM. 
We train lightweight predictors on hidden states using supervision from our normative framework—specifically, binary multilayer perceptron (MLP) classifiers for \textit{True Need} and \textit{True Utility}.
For each model, we use its final transformer layer as the fixed representation. \change{The resulting latent estimators outperform the model’s explicit decisions in predicting true need for most open-source models. Implementation details are provided in Appendix~\ref{sec:controller_details}, and we compare our latent estimators with AdaptiveRAG~\citep{jeong2024adaptive} in Appendix~\ref{sec:baseline}.}

\textbf{Latent Need Estimator (LNE).}
The latent need estimator (LNE) uses the model’s final-token representation from the final transformer layer to predict \textit{true need} via an MLP. As shown in Figure~\ref{fig:lne}, LNE predicts true need more accurately than the model's perceived-need judgment for most models on Entity and GSM-Hard Tasks, with the largest gains for smaller models. Table~\ref{tab:lne-average-improvement} summarizes its downstream budgeted allocation performance, with complete task-level results in Appendix Table~\ref{tab:all}. LNE consistently improves budgeted tool allocation across web-search and calculator tasks. The controller prioritizes examples using LNE’s confidence scores. This improves allocation because high-need examples are more likely to benefit from tool use, although need alone does not guarantee positive utility.

\textbf{The challenge of modeling the tool's utility.}
Utility estimation remains substantially harder than need estimation, and utility-based controllers exhibit small improvement over \auto{}. Need depends only on the model and task and is readable from hidden states; utility additionally depends on what the tool will return and how well the model will use it. Consequently, adding a tool description does not consistently improve utility estimation, and neither LUE variant reliably recovers the oracle ordering of instances by marginal gain. We retain this high-level limitation in the main paper and report the LUE definitions, natural-setting results, prediction analyses, and budget-allocation results in Appendix~\ref{sec:controller_details}.
\section{Concluding Discussion}
We decompose the tool-calling decision into \textit{necessity}, \textit{utility}, and \textit{affordability}, each judged normatively, descriptively, and prescriptively. Across seven models, two tools, and six tasks, models call tools they do not need and skip tools that would help, since perceived need and utility track their true, outcome-defined counterparts only weakly; this gap widens under a fixed budget, where models misprioritize calls and exceed their own limits. Without touching the base model, we train latent estimators on its hidden states: a need estimator (LNE) beats self-report and outperforms \auto{} across most models and tools, while a utility estimator (LUE) is less reliable, hidden states signal what a model knows better than how a tool will behave. Need estimation appears substantially easier than utility estimation; utility estimation is the harder problem, and the better target for future work.

\section*{Ethics Statement}

The Entity and InVivoQuery datasets are both derived from the InVivoGPT dataset~\citep{karnam2026bowling}, a collection of real-world ChatGPT conversation logs. Because these logs originate from real users, we take explicit steps to protect user privacy before any conversation content is used in our pipeline or released as part of this work. Prior to entity extraction and query sampling, we filter the source utterances to remove personally identifiable information (e.g., names, contact details, addresses, and other content that could identify an individual), and we retain only the sanitized entity spans (Entity Task) or de-identified query text (InVivoQuery Task) needed for our evaluation, rather than full raw conversations. We do not publish or redistribute the underlying InVivoGPT logs; only the filtered, task-derived resources described in Appendix~\ref{sec:entity_dataset} are used in our experiments.
\section*{Limitations}

Our study has several limitations. Our latent need estimator (LNE) reliably improves task performance especially under budgeted tool allocation, but the utility estimator (LUE) shows only small, inconsistent gains, utility estimation remains an open problem. Our controllers also use a single fixed representation (the final-token hidden state of each model's last layer) without searching over layers, so our results are a lower bound on what hidden-state probing could achieve.

Our factuality scores rely on an LLM-as-judge pipeline, validated against human judgments on only a small ($n=100$), class-imbalanced sample, which limits how precisely it characterizes judge reliability overall. Models were served with a 4,096-token context window and 512-token generation cap, which may understate the value of tool use for models capable of longer reasoning or evidence integration.

\section*{Contribution}
The overall research direction and experimental design were conceived and discussed collectively by all co-authors. Specific contributions are as follows:\\
Q. Wu \& S. Das led the core framework design. Q. Wu also implemented the primary codes, conducted the main entity-task and parts of the calculators' experiments, and wrote the draft. S. Lee conducted the Perplexity, Brave, and Tavily Search results for the entity task and parts of the calculators' experiments. M. Amani was responsible for the InVivoQuery task data set construction and its corresponding experiments. A. Nag handled the BFCL task data set construction and experiments. K. Gummadi, A. Ravichander, and B. Zafar provided senior guidance and feedback on framework design, experimental methodology, data analysis, and manuscript writing.

\nolinenumbers
\bibliography{colm2026_conference}

\appendix

\section{Disclosure of LLM use in Research}

We used generative AI tools, including GitHub Copilot~\footnote{https://github.com/features/copilot} and Claude Code~\footnote{https://github.com/anthropics/claude-code}, to assist with selected aspects of this work. Specifically, these tools were used to support code development by generating implementation details based on high-level designs specified by the authors, and to help draft portions of the experimental setup descriptions from code written by the authors.

We also employ LLM-based evaluators (LLMs-as-a-judge) to assess model outputs. To ensure reliability, we conduct human evaluation on a randomly sampled subset of instances and verify that the automated judgments are well-aligned with human annotations.

For manuscript preparation, generative AI tools were used in a limited capacity for editing purposes, including shortening paragraphs and correcting grammar and spelling. All core ideas, methodological design, experimental decisions, and interpretations of results were developed and verified by the authors.

All AI-assisted outputs were carefully reviewed and validated by the authors to ensure correctness, originality, and alignment with the intended scientific contributions. The authors take full responsibility for the content of this paper.

\section{Experimental Setup}
\label{sec:experimental_setup}
 
\subsection{Models}
 
\paragraph{Open-source models.}
Locally-hosted models are served with vLLM~\cite{kwon2023efficient}.
The context window is set to 4{,}096 tokens, GPU memory utilisation to 90\%,
and tensor parallelism is configurable via \texttt{--tensor-parallel-size}.
Unless a model-family override is specified, all open-source models are
generated with a maximum of 512 output tokens, temperature set to 0.

\paragraph{Closed-source model.}
We query one closed-source model from OpenAI through the Responses API. We set the maximum generation length to 512 tokens and the temperature to 1, since temperature 0 is unavailable for this model.

We provide the details and links to the models in the Table~\ref{tab:model_names}.

 \begin{table*}[h]
\centering
\small
\resizebox{\textwidth}{!}{
\begin{tabular}{llll}
\toprule
\textbf{Name in paper} & \textbf{Shorten for space}& \textbf{Link} & \textbf{Trained for Tool-Use} \\
\midrule
 
\rowcolor{blue!8}
GPT-OSS-120B
  & GPT-OSS
  & {https://huggingface.co/openai/gpt-oss-120b} 
  & Yes \\
 
\rowcolor{blue!8}
Qwen3-30B-A3B
  & Qwen3-A3B
  & {https://huggingface.co/Qwen/Qwen3-30B-A3B} 
    & Yes \\
 
\rowcolor{blue!8}
Qwen3-30B-IT
  & Qwen3-IT
  & {https://huggingface.co/Qwen/Qwen3-30B-A3B-Instruct-2507} 
    & Yes \\

\rowcolor{blue!8}
Mistral3.1-24B-IT
  & Mistral3.1-IT
  & {https://huggingface.co/mistralai/Mistral-Small-3.1-24B-Instruct-2503}
    & Yes \\
 
\rowcolor{blue!8}
Llama3.2-3B-IT
  & Llama3.2-IT
  & {https://huggingface.co/meta-llama/Llama-3.2-3B-Instruct}
    & Yes \\

\rowcolor{orange!12}
Gemma3-27B-IT
  & Gemma3-IT
  & {https://huggingface.co/google/gemma-3-27b-it}
  & No \\

\rowcolor{green!12}
GPT-5.5
  & gpt-5.5-2026-04-23
  & {https://developers.openai.com/api/docs/models/gpt-5.5}
    & Yes \\

\bottomrule
\end{tabular}
}
\caption{Model names used in this paper and their corresponding identifiers.}
\label{tab:model_names}
\end{table*}

\subsection{Task and Dataset}
\label{sec:task_dataset}

We introduce three datasets used in our evaluation.

\subsubsection{Main: Entity Task Dataset Construction}
\label{sec:entity_dataset}

In real-world applications, search is typically associated with entity-centric factual questions. Successfully using search requires the model to determine when external information is needed, how to formulate an effective query, and how to incorporate retrieved results into its final answer. 
Therefore, we focus on the \textit{Entity Task}, where we evaluate model factuality on an entity-centric question-answering benchmark.

For each entity $e$ in the dataset, the model is given the following prompt template:
\begin{mdframed}[backgroundcolor=SkyBlue!20, linewidth=0pt]
\small
\textit{``In a paragraph, could you tell me what you know about \{entity\}?''}
\end{mdframed}
and is asked to generate a free-form paragraph. We adopt this entity-centered question format, inspired by prior work~\citep{zhao2024wildhallucinations} that evaluates LLMs’ knowledge of individual entities and shows that LLMs perform poorly on long-tail entities.

\begin{figure}
    \centering
    \includegraphics[width=0.65\linewidth]{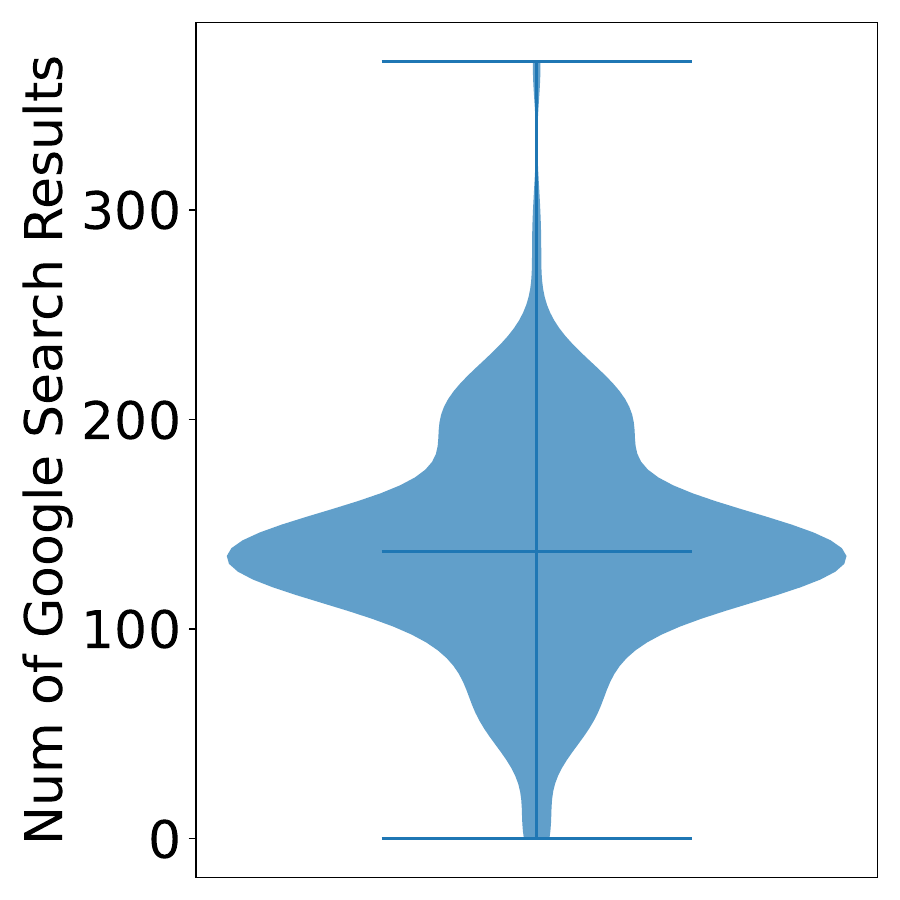}
    \caption{Distribution of the number of Google search results across all entities in the entity dataset. The violin depicts the kernel density estimate of the result count distribution. The horizontal line indicates the median. Extreme outliers, defined as values beyond 1.5× the interquartile range, are excluded for visual clarity.}
    \label{fig:entity-dist}
\end{figure}

To construct a realistic set of entities, we employ a two-stage, claim-based extraction pipeline using GPT-4o on the InVivoGPT dataset~\cite{karnam2026bowling}, which consists of real-world ChatGPT conversation logs. In the first stage, GPT-4o (temperature=1) extracts all named entities from user and assistant messages and classifies them according to the Wolfram Language Entity Types, a taxonomy covering 21 high-level categories (e.g., geographic entities, people, computing-related entities, and medical entities) with fine-grained subtypes. Each entity is returned as a structured JSON object containing its text span, specific type (e.g., \textit{City}, \textit{ProgrammingLanguage}), and high-level category.

In the second stage, an optional verification step checks whether each extracted entity exists in the real world (on the internet) and whether its assigned category is correct, correcting it if necessary. To scale this process efficiently, we use the OpenAI Batch API to process up to 25,000 requests per batch, storing results as JSON files per message. From the extracted entities, we randomly sampled 500 entities for this study. Among them, 262 entities have corresponding Wikipedia pages, while 238 do not. Figure~\ref{fig:entity-dist} shows the distribution of the number of Google search results for these entities. The majority of entities have relatively low search result counts.

\paragraph{Example prompts.}

\begin{mdframed}[backgroundcolor=SkyBlue!20, linewidth=0pt]
\small
\textit{``In a paragraph, could you tell me what you know about Adréa?''}\\[4pt]
\textit{``In a paragraph, could you tell me what you know about National Guards?''}\\[4pt]
\textit{``In a paragraph, could you tell me what you know about Damodaran?''}\\[4pt]
\textit{``In a paragraph, could you tell me what you know about CDA01?''}
\end{mdframed}

\subsubsection{InVivoQuery Task Dataset.}
We further construct evaluation queries based on the same source dataset. Specifically, we select entity-centered user prompts (i.e., prompts containing extracted entities), retain only English queries, and randomly sample 250 queries that triggered a web search and 250 that did not, yielding a total of 500 queries.

We focus on entity-centered queries because they better reflect realistic user behavior, where information needs are typically anchored around identifiable entities such as people, organizations, locations, or products. Such queries provide a more grounded basis for evaluation, as they are associated with verifiable facts and reduce ambiguity in assessing response correctness. Moreover, entity-centric prompts are particularly well-suited for analyzing retrieval and tool-use behavior: queries involving well-known or static entities can often be resolved from internal knowledge, whereas those involving long-tail, ambiguous, or rapidly evolving entities are more likely to require external retrieval (e.g., web search). This distinction allows us to more effectively evaluate a system’s ability to decide when to rely on parametric knowledge versus when to invoke external tools.

\paragraph{Example prompts.}

\begin{mdframed}[backgroundcolor=SkyBlue!20, linewidth=0pt]
\small
\textit{``what happened to chaosium around the kickstarter for 7th edition?''}\\[4pt]
\textit{``Does Sam's club accept EBT online for grocery delivery same Day or next day because I wanted to purchase the Sam's club plus membership''}\\[4pt]
\textit{``should i get sugar free metamucil or the NOW psyllium husk''}\\[4pt]
\textit{``how does walmart 401k match work''}\\[4pt]
\textit{``Trump admin cutting \$20M in DC security funding after federal law enforcement ordered to increase presence By Landon Mion, 3 hrs ago Fox Fox News Follow The Trump administration plans to cut millions in security funding for Washington, D.C., despite the president also directing federal law enforcement to increase its presence in the city because of its ``totally out of control'' crime. In a grant notice posted last week, the Federal Emergency Management Agency (FEMA) said that D.C.'s urban security fund would receive \$25.2 million, a 44\% year-over-year reduction. The Department of Homeland Security, which oversees FEMA, said on Friday it slashed funds to multiple cities to be consistent with the ``current threat landscape.'' Chicago, New York City, Los Angeles, San Francisco and Jersey City also had their security funds cut, but the decrease in D.C. was the largest for any urban area that received funding from the program last fiscal year''}\\[4pt]
\textit{``What are the top 5 soulslike games? Don't include titles by fromsoftware''}

\end{mdframed}

\subsubsection{BFCL Task Dataset}

We additionally evaluate models on the BFCL V4 Web Search dataset from the Berkeley Function Calling Leaderboard (BFCL)~\cite{patil2025bfcl}. This dataset contains 100 multi-hop questions spanning diverse real-world topics, each annotated with single-hop sub-questions and corresponding ground-truth answers. Rather than evaluating full reasoning chains, we treat these sub-questions as standalone queries, resulting in 314 atomic questions.

We focus on sub-questions to isolate and more precisely evaluate a model’s retrieval and tool-use capabilities. Full multi-hop questions entangle multiple factors, such as reasoning quality, intermediate decomposition, and retrieval, making it difficult to attribute errors to specific components. By decomposing them into atomic sub-questions, we reduce this confounding effect and enable a more controlled assessment of whether the model can correctly decide when to invoke web search and retrieve relevant information for a single, well-defined information need.

\paragraph{Example prompts.}

\begin{mdframed}[backgroundcolor=SkyBlue!20, linewidth=0pt]
\small
\textit{``What is the most expensive tea in the world?''}\\[4pt]
\textit{``Which country produces DaHong Pao?''}\\[4pt]
\textit{``Who is the richest billionaire in China?''}\\[4pt]
\textit{``Who is the performer of 2024 Super Bowl halftime show?''}\\[4pt]
\textit{``What is the birthplace of Usher?''}\\[4pt]
\textit{``Which is the NFL team based in Dallas?''}

\end{mdframed}

\subsubsection{Calculator Task Dataset Construction}
\label{sec:calculator_dataset}

We consider three tasks that can benefit from calculator use and are designed to probe arithmetic execution rather than difficult multi-step reasoning.

\medskip\noindent\textbf{GSM-Hard.}\par
This task~\citep{gao2023pal} starts from GSM8K word problems and replaces the small, easily mentally computed constants with large randomly sampled integers while preserving each problem's original reasoning template and executable Python \texttt{solution()} function. The numeric target is obtained by running the perturbed program rather than calculating the answer by hand. Thus, the underlying reasoning remains simple while the arithmetic becomes difficult.

\paragraph{Example prompt.}
\begin{mdframed}[backgroundcolor=SkyBlue!20, linewidth=0pt]
\small
\textit{``A robe takes 2287720 bolts of blue fiber and half that much white fiber. How many bolts in total does it take?''}\\[4pt]
\textbf{Target:} \texttt{3431580.0}
\end{mdframed}

\medskip\noindent\textbf{Synthetic Multiplication.}\par
This dataset (\texttt{synthetic\_multiplication.jsonl}) is generated procedurally by \texttt{generate\_multiplication\_dataset.py} in its ``legacy'' mode. Each prompt has the form \texttt{``Compute the product: $o_1 * o_2 * \cdots * o_k$''}. Its \texttt{easy}, \texttt{medium}, or \texttt{hard} tier controls both the number of multiplications ($k-1 \in \{1\text{--}2, 3\text{--}4, 5\text{--}7\}$) and the operand width (1--2, 2--3, or 3--4 digits, respectively). The exact integer product is stored in the form \texttt{\#\#\#\# <result>}.

\paragraph{Example prompt.}
\begin{mdframed}[backgroundcolor=SkyBlue!20, linewidth=0pt]
\small
\textit{``Compute the product: 635 * 270 * 229 * 850 * 13''}\\[4pt]
\textbf{Target:} \texttt{\#\#\#\# 433845652500}\\[4pt]
\textbf{Difficulty:} \texttt{medium}
\end{mdframed}

\medskip\noindent\textbf{Synthetic Large-Digit Multiplication.}\par
This dataset (\texttt{synthetic\_nn.jsonl}) is produced by the same script in its ``two-number'' mode. Each example asks for the square of one large operand, $N \times N$, where the operand's digit count is sampled from a truncated Gaussian over the configured range and is recorded directly as the difficulty label. Some downstream result files internally label this dataset as \texttt{synthetic\_multiplication}; throughout the paper, we call it Synthetic Large-Digit Multiplication to distinguish it from the chained-product task.

\paragraph{Example prompt.}
\begin{mdframed}[backgroundcolor=SkyBlue!20, linewidth=0pt]
\small
\textit{``Compute the product: 4834314033913 * 4834314033913''}\\[4pt]
\textbf{Target:} \texttt{\#\#\#\# 23370592178488182514091569}\\[4pt]
\textbf{Difficulty:} \texttt{13-digit}
\end{mdframed}

Together, the three datasets isolate complementary sources of computational difficulty: GSM-Hard tests arithmetic embedded in natural-language reasoning, Synthetic Multiplication scales the \emph{number} of chained operations, and Synthetic Large-Digit Multiplication scales the \emph{operand magnitude} of a single multiplication.

\subsection{Scoring}
\label{sec:score}

\subsubsection{Entity and InVivoQuery task}
\change{
We use factuality, completeness, and relevance as the three metrics for the open-ended tasks. We focused on factuality in our analysis and provided completeness and relevance results for the GPT-OSS-120B model on the Entity task to show that our evaluation framework can be generalized to other metrics.
}

\paragraph{Factuality}
Factuality is measured with an automated two-stage pipeline built on top of the OpenAI Responses API:
 
\begin{enumerate}
  \item \textbf{Claim extraction.}
        An extraction model (default: GPT-4o) is prompted with a structured
        JSON schema to decompose the model's response into atomic, checkable
        claims (\textit{factual}, \textit{numerical}, \textit{historical},
        \textit{definition}, \textit{other}).
 
  \item \textbf{Claim verification.}
        A verification model (default: GPT-4o) is given the extracted claims
        and uses its built-in \texttt{web\_search} tool to assess each claim
        against live web sources.
        It returns a boolean \texttt{is\_correct} flag and one-to-three
        source URLs per claim.
\end{enumerate}
 
The factuality score for a response is
\[
  s = \frac{\text{correct claims}}{\text{total claims}} \in [0, 1].
\]
If the extractor returns zero claims, we assign a factuality score of 0 rather than evaluating an undefined $0/0$ ratio. This rule covers refusals, empty or malformed answers, and extraction failures that persist after retry, and is applied identically in all tool-use conditions.
Both the extraction and verification calls use exponential-backoff retry
(up to five attempts) to handle transient API errors.

\paragraph{Claim Extraction and Verification Prompts.}

To evaluate factuality, we adopt a two-stage pipeline consisting of claim extraction followed by claim verification.

\paragraph{(1) Claim Extraction.}
\begin{mdframed}[backgroundcolor=SkyBlue!20, linewidth=0pt]
\small
\textbf{Extraction Prompt:}\\
You are a claim extraction engine. Extract all distinct, checkable claims from the RESPONSE.
\begin{itemize}
  \item A claim is an assertion that could be true or false
  \item Split compound sentences into atomic claims
  \item Do NOT add new claims
  \item Prefer recall over precision
\end{itemize}
Return \textbf{only} valid JSON matching the schema.

\vspace{4pt}
\textbf{Schema:}
\begin{verbatim}
_EXTRACTION_SCHEMA = {
    "type": "object",
    "properties": {
        "entity": {"type": "string"},
        "question": {"type": "string"},
        "claims": {
            "type": "array",
            "items": {
                "type": "object",
                "properties": {
                    "id": {"type": "string"},
                    "claim": {"type": "string"},
                    "type": {
                        "type": "string",
                        "enum": [
                                    "factual", 
                                    "numerical", 
                                    "historical", 
                                    "definition", 
                                    "other"],
                    },
                    "span": {"type": "string"},
                },
                "required": ["id", "claim", "type", "span"],
                "additionalProperties": False,
            },
        },
    },
    "required": ["entity", "question", "claims"],
    "additionalProperties": False,
}

\end{verbatim}
\end{mdframed}

\paragraph{(2) Claim Verification.}
\begin{mdframed}[backgroundcolor=SkyBlue!20, linewidth=0pt]
\small
\textbf{Verification Prompt:}\\
You are a claim verification engine. Verify each claim using the \texttt{web\_search} tool when needed, prioritizing reliable sources.
\begin{itemize}
  \item Preserve the original id and claim text exactly
  \item Set \texttt{is\_correct=true} only if the claim is clearly correct
  \item If uncertain, set \texttt{is\_correct=false} and explain briefly
  \item Include 1--3 plain-text source UR\textbf{}Ls in the reason
\end{itemize}
Return \textbf{only} valid JSON matching the schema.

\vspace{4pt}
\textbf{Schema:}
\begin{verbatim}
_VERIFICATION_SCHEMA = {
    "type": "object",
    "properties": {
        "entity": {"type": "string"},
        "question": {"type": "string"},
        "verifications": {
            "type": "array",
            "items": {
                "type": "object",
                "properties": {
                    "id": {"type": "string"},
                    "claim": {"type": "string"},
                    "is_correct": {"type": "boolean"},
                    "reason": {"type": "string"},
                },
                "required": ["id", "claim", "is_correct", "reason"],
                "additionalProperties": False,
            },
        },
    },
    "required": ["entity", "question", "verifications"],
    "additionalProperties": False,
}
\end{verbatim}
\end{mdframed}

\paragraph{Completeness}

We use the LLM-as-a-judge to give a 5-likert score to evaluate the completeness. The full prompt we used is:
\begin{mdframed}[backgroundcolor=SkyBlue!20, linewidth=0pt]
You are an evaluator assessing the completeness of an AI-generated response to a user query.
\\
Evaluate:
- Does the response fully address and cover all parts of the user’s question?
\\
Return JSON:
\\
\{\{
  "score": 1-5,
  "reasoning": "1-2 sentence explanation"
\}\}
\\
Scoring guide:\\
1 = Very incomplete; misses most parts of the question or fails to address the main request\\
2 = Partially incomplete; addresses some parts but omits major components of the question\\
3 = Moderately complete; covers the main request but misses some secondary aspects or details\\
4 = Mostly complete; addresses nearly all parts with only minor omissions\\
5 = Fully complete; covers all aspects of the question thoroughly\\
\\\\
Before scoring, consider the query type:\\
- For open-ended queries, interpret completeness as reasonable coverage of key aspects, not exhaustiveness.
\end{mdframed}

\paragraph{Relevance}

We use the LLM-as-a-judge to give a 5-likert score to evaluate the relevance. The full prompt we used is:
\begin{mdframed}[backgroundcolor=SkyBlue!20, linewidth=0pt]
You are an evaluator assessing how relevant an AI-generated response is to a user query.
\\
Evaluate:\\
- Does the response directly address the user's question or intent?\\
- Is the response concise, to the point, and free from off-topic or unnecessary information?\\
\\
Return JSON:\\
\\
\{\{
"score": 1-5,
"reasoning": "1-2 sentence explanation"
\}\}
\\
Scoring guide:\\
1 = Irrelevant; does not address the user’s question or intent at all\\
2 = Weakly relevant; touches on the topic but largely misses the user’s intent or includes substantial off-topic content\\
3 = Partially relevant; addresses the main intent but includes noticeable irrelevance or digressions\\
4 = Mostly relevant; well-aligned with the intent with only minor off-topic or unnecessary details\\
5 = Fully relevant; directly and precisely addresses the user’s intent with no unnecessary content\\
\end{mdframed}

\subsubsection{Calculator Task Scorer}

We score all three calculator datasets using normalized exact-match accuracy. For each response, we extract the final numeric answer and normalize its numeric representation before comparing it with the stored target; formatting differences such as surrounding text, separators, or an integer-equivalent decimal representation do not change correctness. A response receives $s=1$ if the normalized value equals the target and $s=0$ otherwise. We apply the same binary scorer in the \notool{}, \auto{}, and \withtool{} conditions. This paired correctness signal directly determines true need and utility: calculator use has positive utility when the no-tool answer is incorrect and the with-tool answer is correct, negative utility in the reverse case, and neutral utility when correctness is unchanged.

\subsubsection{BFCL Scorer}

An additional BFCL LLM-as-Judge scorer is provided for
BFCL benchmark questions.
We're using this prompt to guide the model to make the judgment:

\begin{mdframed}[backgroundcolor=SkyBlue!20, linewidth=0pt]
\small
\texttt{<bos><start\_of\_turn>user}\\
You are an expert evaluator for question answering systems.

You will receive:
\begin{itemize}
    \item A question
    \item A ground truth answer
    \item A model's answer
\end{itemize}

Your task is to determine whether the model's answer correctly contains or conveys the ground truth answer as the answer to the question.

Evaluation criteria:

Score 1 only if the model answer explicitly, unambiguously, and correctly provides the ground truth answer (or a semantically equivalent answer) in response to the question.
Score 0 if the model answer is incorrect, contradictory, ambiguous, speculative, hedged, or gives multiple possible answers without clearly identifying the correct one.\\
Accept semantically equivalent answers, paraphrases, standard aliases, and unambiguous abbreviations when they clearly refer to the same answer.
Ignore differences in capitalization, punctuation, articles, and minor formatting.
If the question asks for a specific entity, number, date, or fact, the model answer must match that fact accurately enough to be considered correct in context.
If the model answer includes extra information, that is acceptable only if the final answer remains clearly correct and not contradicted.\\
Return ONLY valid JSON in exactly this format:\\
{ "score": 0 or 1 } 

0 = incorrect \\
1 = correct\\
<end\_of\_turn>\\
<start\_of\_turn>model
\end{mdframed}

\subsection{Agentic Tool-Use Framework}
\label{apd:two-stage}
 
Our agentic framework uses the same two-stage protocol for every task: the model first decides whether to invoke the available tool and then generates an answer conditioned on the tool output when invoked~\cite{yao2023react}. Harness assignment depends only on the model and remains fixed across tools; only the tool description, input schema, and returned output change. Gemma3-27B-IT uses a \textit{customized harness} with manually constructed prompts because it does not reliably expose a native function-calling schema. The other open-source models use a \textit{trained harness} based on their native tool-calling chat templates, which retain the full execution trace---including the system prompt, user query, tool call, and tool result. For GPT-5.5, the OpenAI Responses API constructs and manages the tool-use harness and execution trace. All three harnesses expose the same \notool{}/\auto{}/\withtool{} modes.
 
\subsubsection{Stage 1: Tool decision.}
\label{sec:prompts}

We first present the customized Gemma harness using web search as an illustrative instantiation. The calculator uses the same decision and response-generation protocol, with only the tool description, input schema, and output substituted. In the trained open-source harnesses, equivalent instructions and schemas are rendered by each model's native chat template. For GPT-5.5, the Responses API handles their construction and serialization.

\paragraph{The \auto setup.}
In the illustrated web-search instantiation, the model receives a structured JSON schema and decides whether to invoke \texttt{web\_search}. For the example entity \textit{CDA01}, the \textbf{self} tool-selection prompt is:

\begin{mdframed}[backgroundcolor=SkyBlue!20, linewidth=0pt]
\small
\texttt{<bos><start\_of\_turn>user}\\
You are an intelligent agent that decides when to use tools to answer questions.\\
You have access to the following tools:
\begin{itemize}
  \item \textbf{web\_search}: Search the web for current information about entities, facts, or topics. Use this when you need up-to-date or factual information you don't have.
\end{itemize}
Given the user's question: \textit{``In a paragraph, could you tell me what you know about CDA01?''}\\[6pt]
Decide if you need to use any tools. Respond with a JSON object:

\{
    "needs\_tool": true\/false,
    "tool\_name": "tool\_name" or null,
    "tool\_input": "the input you need to give to the tool",
    "reasoning": "why you need this tool or why you don't need tools"
\}

Rules:
\begin{itemize}
  \item Only use tools when you genuinely need external information
  \item If you already know the answer, set needs\_tool to false
  \item Only select ONE tool at a time
\end{itemize}
\texttt{<end\_of\_turn>}\\
\texttt{<start\_of\_turn>model}
\end{mdframed}

We adopt a simple and general tool description for our web search tool to avoid over constraining the model’s behavior while still providing sufficient guidance for effective use. Rather than specifying detailed rules or heuristics, the description emphasizes the tool's core capability: retrieving current, factual information, along with a high-level condition for use when knowledge is missing or uncertain. This design is broadly consistent with prior work such as ReAct~\cite{yao2023react} and Toolformer~\cite{10.5555/3666122.3669119}, which suggest that relatively lightweight tool descriptions, when combined with appropriate prompting or training, can enable effective tool selection.

Then we will get the response from the model as follows:

\begin{mdframed}[backgroundcolor=SkyBlue!20, linewidth=0pt]
\small
\texttt{<bos><start\_of\_turn>model}\\[6pt]
\texttt{\{}\\
\texttt{\hspace{1em}"needs\_tool": true,}\\
\texttt{\hspace{1em}"tool\_name": "web\_search",}\\
\texttt{\hspace{1em}"tool\_input": "CDA01",}\\
\texttt{\hspace{1em}"reasoning": "I do not have pre-existing knowledge about 'CDA01'. It is likely a specific code, identifier, or acronym that requires a web search to understand its meaning and provide a relevant response."}\\
\texttt{\}}\\[6pt]
\texttt{<end\_of\_turn>}
\end{mdframed}

The customized harness parses the model's response and invokes the available tool when requested.

If the model declines the tool, the harness proceeds with the same task prompt used in the \notool{} setup.

If the model invokes the tool, the harness executes the generated input and supplies the result during response generation. For web search, this means sending the generated query to the MCP server and adding the returned results to the prompt:

\begin{mdframed}[backgroundcolor=SkyBlue!20, linewidth=0pt]
\small
\texttt{<|im\_start|>system}\\[6pt]
You are a helpful assistant.\\
\texttt{<|im\_end|>}\\[6pt]

\texttt{<|im\_start|>user}\\[6pt]
Based on the following search results, please answer the question.\\
Search Results:\\
1. Current Transformer | Submetering | Submeter | CDA01-K12\\
Current Transformer, Solid Core, 100A, 100mA, 0.72" Opening, 48” Leads, +/-0.3\% Accuracy, Black, For Submetering.\\
Source: https://leviton.com/products/cda01-k12\\

2. Leviton CDA01-K12 Current Transformer, 100:0.1A, Solid ...\\
The Leviton CDA01-K12 solid core current transformer (CT) is cost effective and less susceptible to damage during installation.\\
Source: https://www.powermeterstore.com/product/....\\

3. VerifEye™, Sub-Metering Current Transformer, 100A, Solid ...\\
Designed for accurately capturing measurements of power consumption, CTs are easy to specify and install.\\
Source: https://www.graybar.com/...\\

\text{4. CDA01-K12.pdf}\\
Leviton solid core CTs are cost effective and less susceptible to damage during installation.\\
Source: https://www.bulbspro.com/media/pdf/CDA01-K12.pdf\\

\text{5. Leviton® CDA01-R12 Solid Core Sub-Metering Current ...}\\
100:0.1 current ratio, 100 A primary, 0.1 A secondary, 0.3\% accuracy.\\
Source: https://www.steinerelectric.com/...\\

Question: In a paragraph, could you tell me what you know about CDA01?\\
\texttt{<|im\_end|>}\\

\texttt{<|im\_start|>assistant}\\[6pt]
\end{mdframed}

\paragraph{The \notool{} setup.}

In the \notool{} setup, the harness skips the decision stage and presents the task query directly. The following example uses web search:

\begin{mdframed}[backgroundcolor=SkyBlue!20, linewidth=0pt]
\small
\texttt{<|im\_start|>system}\\[6pt]
You are a helpful assistant.\\
\texttt{<|im\_end|>}\\[6pt]

\texttt{<|im\_start|>user}\\[6pt]
In a paragraph, could you tell me what you know about Damodaran?\\
\texttt{<|im\_end|>}\\[6pt]

\texttt{<|im\_start|>assistant}\\
\texttt{<|im\_end|>}
\end{mdframed}

\paragraph{The \withtool{} setup.} 

In the \withtool{} setup, the harness forces tool invocation while allowing the model to generate the tool input. The following example shows this setup for web search:

\begin{mdframed}[backgroundcolor=SkyBlue!20, linewidth=0pt]
\small
\texttt{<bos><start\_of\_turn>user}\\
You are an intelligent agent that decides when to use tools to answer questions.\\
You have access to the following tools:
\begin{itemize}
  \item \textbf{web\_search}: Search the web for current information about entities, facts, or topics. Use this when you need up-to-date or factual information you don't have.
\end{itemize}

Given the user's question: \textit{``In a paragraph, could you tell me what you know about CDA01?''}\\[6pt]
Use this tool. Respond with a JSON object:

\{
    "needs\_tool": true,
    "tool\_input": "the input you need to give to the tool"
\}
\end{mdframed}

Then we will get the model response as:

\begin{mdframed}[backgroundcolor=SkyBlue!20, linewidth=0pt]
\small
\texttt{<bos><start\_of\_turn>model}\\[6pt]
\texttt{\{}\\
\texttt{\hspace{1em}"needs\_tool": true,}\\
\texttt{\hspace{1em}"tool\_input": "CDA01",}\\
\texttt{\}}\\[6pt]
\texttt{<end\_of\_turn>}
\end{mdframed}

We perform an exact-match check: if the model follows the instruction to set \textit{needs\_tool} to \textit{True}, we leave it as \textit{True}; otherwise, we set it to \textit{True} and use the original query as the tool input.

\paragraph{Perceived need prompts}

To test the model's perception of need, we prompt the model using three different strategies to decide whether it needs help from an external tool without giving the exact information about the tool.

\textbf{Perceived need prompts - v1}
\begin{mdframed}[backgroundcolor=SkyBlue!20, linewidth=0pt]
\small
\texttt{<bos><start\_of\_turn>user}\\
You are an intelligent agent that decides when to use tools to answer questions.\\

Given the user's question: \textit{``In a paragraph, could you tell me what you know about CDA01?''}\\[6pt]
You can call for help from a tool.
Respond with ONLY a JSON object in this exact schema:

\{
    "needs\_tool": true,
\}

or 

\{
    "needs\_tool": false,
\}
\end{mdframed}
\textbf{Perceived need prompts - v2}
\begin{mdframed}[backgroundcolor=SkyBlue!20, linewidth=0pt]
\small
\texttt{<bos><start\_of\_turn>user}\\
You are an intelligent agent that decides when to use tools to answer questions.\\

Given the user's question: \textit{``In a paragraph, could you tell me what you know about CDA01?''}\\[6pt]
Do you need help to answer the question?\\
Answer:
\end{mdframed}
\textbf{Perceived need prompts - v3}
\begin{mdframed}[backgroundcolor=SkyBlue!20, linewidth=0pt]
\small
\texttt{<bos><start\_of\_turn>user}\\
You are an intelligent agent that decides when to use tools to answer questions.\\

Given the user's question: \textit{``In a paragraph, could you tell me what you know about CDA01?''}\\[6pt]
Do you know the answer to the question?\\
Answer:
\end{mdframed}

\paragraph{Cost-Aware Tool Description Variants}
\label{sec:tool_descriptions}
 
To study how the framing of tool cost affects search behaviour, we vary the
natural-language description of the \texttt{web\_search} tool injected into
the tool-selection prompt.
All variants share the same base description.

We consider two variants of cost-aware tool descriptions, differing in whether the model is explicitly given the remaining tool-call budget.

\paragraph{(1) Explicit Budget (No Implicit Calculation Required).}
\begin{mdframed}[backgroundcolor=SkyBlue!20, linewidth=0pt]
\label{prompt:cost1}
\small
\texttt{<bos><start\_of\_turn>user}\\
You are an intelligent agent that decides when to use tools to answer questions.\\
You have access to the following tools:
\begin{itemize}
  \item \textbf{web\_search}: Search the web for current information about entities, facts, or topics. Use this when you need up-to-date or factual information you don't have.
  \item {\color{blue}\textbf{Each tool call costs \$$X$. You have a total budget of \$10000. You have 500 questions in total, have already answered $n_{\text{fin}}$, and have made $n_{\text{call}}$ tool calls so far. You have $y$ tool calls remaining.}}
\end{itemize}
Given the user's question: \textit{``In a paragraph, could you tell me what you know about CDA01?''}\\[6pt]
Decide whether to use a tool. Respond with a JSON object:

\{
    "needs\_tool": true\/false,
    "tool\_name": "tool\_name" or null,
    "tool\_input": "input to the tool",
    "reasoning": "justification"
\}

Rules:
\begin{itemize}
  \item Use tools only when external information is necessary
  \item If the answer is known, set needs\_tool to false
  \item Select at most one tool
\end{itemize}
\texttt{<end\_of\_turn>}\\
\texttt{<start\_of\_turn>model}
\end{mdframed}

\paragraph{(2) Implicit Budget (Requires Internal Calculation).}
\begin{mdframed}[backgroundcolor=SkyBlue!20, linewidth=0pt]
\label{prompt:cost2}
\small
\texttt{<bos><start\_of\_turn>user}\\
You are an intelligent agent that decides when to use tools to answer questions.\\
You have access to the following tools:
\begin{itemize}
  \item \textbf{web\_search}: Search the web for current information about entities, facts, or topics. Use this when you need up-to-date or factual information you don't have.
  \item {\color{blue}\textbf{Each tool call costs \$$X$. You have a total budget of \$10000. You have 500 questions in total, have already answered $n_{\text{fin}}$, and have made $n_{\text{call}}$ tool calls so far.}}
\end{itemize}
Given the user's question: \textit{``In a paragraph, could you tell me what you know about CDA01?''}\\[6pt]
Decide whether to use a tool. Respond with a JSON object:

\{
    "needs\_tool": true\/false,
    "tool\_name": "tool\_name" or null,
    "tool\_input": "input to the tool",
    "reasoning": "justification"
\}

Rules:
\begin{itemize}
  \item Use tools only when external information is necessary
  \item If the answer is known, set needs\_tool to false
  \item Select at most one tool
\end{itemize}
\texttt{<end\_of\_turn>}\\
\texttt{<start\_of\_turn>model}
\end{mdframed}

\paragraph{Budget-Aware Setting.}
Budget-aware variants provide the model with: the total number of questions $N$, the number already answered $n_{\text{fin}}$, and the cumulative number of tool calls $n_{\text{call}}$. These values are updated after each sample and injected dynamically at inference time (i.e., not pre-computed).

We vary the per-call cost $X$ from 0 to 10{,}000 to study whether models adapt their tool-use decisions under different budget constraints. In the explicit-budget variant, the remaining number of tool calls is given by $y = \left\lfloor \frac{10000 - X \cdot n_{\text{call}}}{X} \right\rfloor$.

\subsubsection{Stage 2: Response generation.}

After tool selection and execution, each model's fixed harness returns the tool output for final-response generation. This procedure is identical across tasks; only the content and schema of the tool output differ. The examples below use web search.

\paragraph{Customized harness (Gemma).}
In the web-search example, the customized harness injects the retrieved titles, snippets, and URLs together with the current user query into a new final-response prompt. For the calculator, the same harness position instead contains the calculator result. The manually constructed web-search prompt is:

\begin{mdframed}[backgroundcolor=SkyBlue!20, linewidth=0pt]
\small
\texttt{<bos><start\_of\_turn>user}\\
You are an intelligent agent that decides when to use tools to answer questions.\\

Based on the following search results, please answer the question.
  $\langle$results$\rangle$ Question: \{query\}
  \texttt{<end\_of\_turn>}\\
\texttt{<start\_of\_turn>model}
\end{mdframed}

\paragraph{Trained harnesses (open-source models).}
For the remaining open-source models, the trained harness appends the output of the available tool as a tool-response message to the existing conversation, preserving the full execution trace: the system prompt, user query, tool schema, model tool call, and tool result. The following web-search example illustrates this serialized trace:
\begin{mdframed}[backgroundcolor=SkyBlue!20,linewidth=0pt]
\footnotesize
\begin{minipage}{\linewidth}
\ttfamily

<|im\_start|>system\\
You are a helpful assistant.\\

\# Tools\\
You may call one or more functions to assist with the user query.\\
You are provided with function signatures within
<tools></tools> XML tags.\\

<tools>\\
\{"type":"function",\\
"function":\{"name":"web\_search",\\
"description":"Search the web for current information.",\\
"parameters":\{"type":"object",\\
"properties":\{"query":\{"type":"string"\}\},\\
"required":["query"]\}\}\}\\
</tools>\\

For each function call, return a JSON object within
<tool\_call></tool\_call> tags.\\

<tool\_call>\\
\{"name":"web\_search","arguments":\{"query":"Adréa"\}\}\\
</tool\_call>\\

<tool\_response>\\
Search Results:\\
1. Adrea\\
Source: \url{https://www.youtube.com/adreacastiano}\\

2. Adrea\\
Source: \url{https://open.spotify.com/artist/6Xfod4pClV2XvfTkiZumH3}\\

3. Adrea (@adreasings)\\
Source: \url{https://www.instagram.com/adreasings/}\\

4. About - Adrea\\
Source: \url{https://adreacastiano.com/about}\\

5. Adrea\\
Source: \url{https://music.apple.com/us/artist/adrea/410768050}\\

\end{minipage}
\end{mdframed}

\paragraph{OpenAI-managed harness.}
For GPT-5.5, we provide the tool definition through the OpenAI Responses API. The API constructs the model-facing harness, records the tool call and result, and carries the execution trace into final-response generation; we do not manually construct or serialize this harness.

If no tool is invoked, the task query is passed directly. Across all tasks, the system message and tool-call serialization are determined by the model's assigned harness, not by the tool.

\subsection{Web-Search Backend}
 
\paragraph{Google Search via SerpAPI.} Web search is provided via a FastMCP server~\footnote{https://github.com/prefecthq/fastmcp} that exposes a
single \texttt{web\_search(query, count)} tool.
One provider is supported: Google Search (via SerpApi).
The server returns up to $k{=}5$ results per query, each comprising a title,
snippet, and URL.
The MCP client communicates with the server over stdio using the Model Context
Protocol, enabling the agent loop to call the tool asynchronously without
blocking the main generation thread.

\paragraph{Perplexity, Brave and Tavily.}
The experiment integrates Brave, Tavily, and Perplexity search through the MCP protocol by wrapping each provider's REST API in a lightweight FastMCP server. Each server exposes a single `web\_search` tool that the language model can call during inference. When the server starts up, it is configured to use a specific provider — Brave, Tavily, or Perplexity — and all HTTP communication with that provider's API happens server-side, invisible to the model. The model only ever sees one tool with one interface, regardless of which provider is running underneath, so swapping providers is purely a configuration concern: set the right API key and pass the provider name at launch.

On the client side, the agent loop connects to the running MCP server at the start of each experiment, retrieves the tool schema, and injects it into the model's context. After each generation step, if the model emits a tool call, the loop dispatches it to the MCP server, gets back formatted search results, appends them as a tool-role message, and continues the conversation until the model produces a final answer. Because the tool interface is identical across providers, none of the agent logic, scoring, or evaluation code needs to change when switching between Brave, Tavily, and Perplexity — adding a new provider only requires implementing its API call server-side and registering it under a new provider name, and the rest of the system picks it up automatically.

 \subsection{The Controller Framework}
 \label{sec:controller_details}

For every model and predictor, we use the last-token representation from the final transformer layer. This layer is fixed before evaluation; we do not search over layers or select a layer using evaluation performance. Given model responses and their corresponding task scores, prompts are encoded in batches using Hugging Face Transformers, producing one embedding matrix of shape $(N \times H)$ per model.

We train a multilayer perceptron (MLP) classifier using the \texttt{MLPClassifier} implementation from scikit-learn. The input features are fixed-length final-layer embedding vectors. Prior to training, all features are standardized using \texttt{StandardScaler} fitted within each training fold.
The MLP was trained with a maximum of 100 iterations, and early stopping was enabled, where training terminates if the validation performance does not improve for five consecutive iterations. The optimization follows the default settings of scikit-learn, which employs the Adam optimizer. A fixed random seed of 42 was used to ensure reproducibility.

Hyperparameters were selected via grid search over a set of candidate architectures and learning rates. Specifically, we considered hidden layer configurations of $\{\emptyset, (128), (256), (128,64), (1024,64)\}$, where $\emptyset$ denotes a model without hidden layers, and initial learning rates of $\{10^{-3}, 10^{-4}\}$. Model selection was performed using 5-fold stratified cross-validation on the training data, preserving the class distribution across folds.

For evaluation, we employed 5-fold stratified cross-validation over the entire dataset and reported out-of-fold predictions. Predictor quality is reported using accuracy. All reported results are based on these cross-validated predictions.

All reported controller scores and downstream allocations use this fixed final-layer representation. Hyperparameter selection remains confined to the training data within each evaluation fold; there is no representation-layer selection step.

 We further assess whether the latent estimators learn signals that generalize beyond their training folds. Table~\ref{tab:oof_auc_detailed} reports five-fold stratified out-of-fold AUROC for the three web-search tasks. LNE performs reliably above chance in most model–task combinations, with particularly strong results on Entity and BFCL, whereas utility prediction is less consistent and more model-dependent.

\begin{table*}[t]
\centering
\small
\begin{tabular}{llccc}
\toprule
Task & Model & LNE & $LUE_x$ & $LUE_{x,d_{\mathcal{F}}}$ \\
\midrule

\multirow{6}{*}{Entity}
& GPT-OSS-120B
& $0.782 \pm 0.040$ \checkmark
& $0.691 \pm 0.067$ \checkmark
& $0.667 \pm 0.029$ \checkmark \\

& Qwen3-30B-A3B
& $0.797 \pm 0.061$ \checkmark
& $0.677 \pm 0.044$ \checkmark
& $0.623 \pm 0.048$ \checkmark \\

& Qwen3-30B-A3B-Instruct
& $0.775 \pm 0.056$ \checkmark
& $0.581 \pm 0.066$ $\sim$
& $0.615 \pm 0.055$ \checkmark \\

& Gemma-3-27B
& $0.623 \pm 0.052$ \checkmark
& $0.566 \pm 0.042$ \checkmark
& $0.496 \pm 0.078$ $\sim$ \\

& Mistral-Small-24B
& $0.545 \pm 0.051$ $\sim$
& $0.556 \pm 0.059$ $\sim$
& $0.539 \pm 0.053$ $\sim$ \\

& Llama-3.2-3B
& $0.651 \pm 0.051$ \checkmark
& $0.593 \pm 0.075$ $\sim$
& $0.653 \pm 0.067$ \checkmark \\

\midrule

\multirow{6}{*}{InVivo}
& GPT-OSS-120B
& $0.631 \pm 0.054$ \checkmark
& $0.674 \pm 0.056$ \checkmark
& $0.703 \pm 0.042$ \checkmark \\

& Qwen3-30B-A3B
& $0.579 \pm 0.061$ \checkmark
& $0.526 \pm 0.027$ $\sim$
& $0.523 \pm 0.043$ $\sim$ \\

& Qwen3-30B-A3B-Instruct
& $0.607 \pm 0.059$ \checkmark
& $0.588 \pm 0.045$ \checkmark
& $0.665 \pm 0.026$ \checkmark \\

& Gemma-3-27B
& $0.537 \pm 0.065$ $\sim$
& $0.554 \pm 0.046$ $\sim$
& $0.471 \pm 0.050$ $\sim$ \\

& Mistral-Small-24B
& $0.582 \pm 0.051$ \checkmark
& $0.534 \pm 0.042$ $\sim$
& $0.542 \pm 0.032$ \checkmark \\

& Llama-3.2-3B
& $0.668 \pm 0.091$ \checkmark
& $0.544 \pm 0.053$ $\sim$
& $0.571 \pm 0.088$ $\sim$ \\

\midrule

\multirow{6}{*}{BFCL}
& GPT-OSS-120B
& $0.765 \pm 0.050$ \checkmark
& $0.657 \pm 0.068$ \checkmark
& $0.633 \pm 0.040$ \checkmark \\

& Qwen3-30B-A3B
& $0.844 \pm 0.042$ \checkmark
& $0.753 \pm 0.063$ \checkmark
& $0.758 \pm 0.060$ \checkmark \\

& Qwen3-30B-A3B-Instruct
& $0.878 \pm 0.035$ \checkmark
& $0.823 \pm 0.052$ \checkmark
& $0.760 \pm 0.050$ \checkmark \\

& Gemma-3-27B
& $0.502 \pm 0.113$ $\sim$
& $0.623 \pm 0.101$ $\sim$
& $0.526 \pm 0.090$ $\sim$ \\

& Mistral-Small-24B
& $0.604 \pm 0.051$ \checkmark
& $0.709 \pm 0.060$ \checkmark
& $0.640 \pm 0.052$ \checkmark \\

& Llama-3.2-3B
& $0.812 \pm 0.039$ \checkmark
& $0.723 \pm 0.024$ \checkmark
& $0.680 \pm 0.066$ \checkmark \\

\bottomrule
\end{tabular}

\caption{
Detailed out-of-fold AUROC results for each web-search task, model,
and predictor. Values report the mean and standard deviation across
five stratified cross-validation folds. Symbols are determined using
a two-sided 95\% $t$-interval for the mean AUROC:
\checkmark indicates that the lower bound exceeds chance level
($0.50$), $\sim$ indicates that the interval includes $0.50$, and
$\times$ indicates that the upper bound is below $0.50$.
}
\label{tab:oof_auc_detailed}
\end{table*}
\section{Human Alignment with the LLM-as-Judge Evaluation}
\label{sec:human}

We assessed the LLM-as-judge verdicts by having three human annotators independently validate each claim-level judgment on a shared sample of 100 responses. For each claim, annotators indicated whether they agreed (1) or disagreed (0) with the LLM's verdict. Raw endorsement was 99.7\%, 99.9\%, and 92.1\% for Annotators 1--3, respectively; Annotators 1 and 2 agreed with each other on 99.8\% of shared claims. These high raw percentages should be interpreted cautiously because agreement labels were extremely imbalanced: the first two annotators exhibited near-zero label variance, and Annotator 3 had near-chance chance-corrected agreement with them despite disagreeing with 7.9\% of the judge verdicts. We therefore use this exercise as a limited validation check, not as evidence of near-perfect judge reliability.

\section{Example of True Negative Utility}
\label{sec:hurts}
In this section, we show some examples where the factuality is decreasing in the entity task across different models:

\paragraph{Entity: Sky Blue}

\begin{mdframed}[backgroundcolor=SkyBlue!20, linewidth=0pt]
\small
\texttt{<|start|>system<|message|>You are ChatGPT, a large language model trained by OpenAI.}\\
\texttt{Knowledge cutoff: 2024-06}\\
\texttt{Current date: 2026-03-09}\\[6pt]
\texttt{Reasoning: medium}\\[6pt]
\texttt{\# Valid channels: analysis, commentary, final. Channel must be included for every message.<|end|>}\\[6pt]
\texttt{<|start|>developer<|message|>\# Instructions}\\[6pt]
\texttt{<|end|>}\\[6pt]
\texttt{<|start|>user<|message|>In a paragraph, could you tell me what you know about Sky Blue?<|end|>}\\[6pt]
\texttt{<|start|>assistant<|message|>}
\end{mdframed}

The response from GPT-OSS-120B without search, which gets a factuality score of 1:

\begin{mdframed}[backgroundcolor=SkyBlue!20, linewidth=0pt]
\small
"Sky blue is a light, calming shade of blue that resembles the color of a clear daytime sky, typically ranging from a pale cyan to a soft, pastel hue. In color theory, it sits between turquoise and baby blue on the spectrum and is often associated with tranquility, openness, and optimism, making it a popular choice in design, fashion, and interior décor to evoke a sense of spaciousness and serenity. Historically, sky blue has been used in art to depict atmospheric perspective, helping to convey depth and distance, while in cultural contexts it can symbolize freedom, peace, and the infinite possibilities of the horizon. The shade also appears in various branding and corporate identities to convey reliability and calmness, and it is a frequent favorite in wedding palettes, baby nurseries, and wellness spaces for its soothing visual impact."
\end{mdframed}

The prompt with search:

\begin{mdframed}[backgroundcolor=SkyBlue!20, linewidth=0pt]
\small
\texttt{<|im\_start|>system}\\[6pt]
\texttt{You are a helpful assistant.}\\
\texttt{<|im\_end|>}\\[6pt]

\texttt{<|im\_start|>user}\\[6pt]
\texttt{Based on the following search results, please answer the question.}\\[6pt]

\texttt{Search Results:}\\[6pt]

\texttt{1. Coventry City F.C.}\\
\texttt{\hspace{1em}The club plays in the EFL Championship and is nicknamed the Sky Blues after its sky blue colours.}\\
\texttt{\hspace{1em}Source: https://en.wikipedia.org/wiki/Coventry\_City\_F.C.}\\[6pt]

\texttt{2. Club History}\\
\texttt{\hspace{1em}The Sky Blues played their first game at the Ricoh Arena in 2005, winning 3-0.}\\
\texttt{\hspace{1em}Source: https://www.ccfc.co.uk/...}\\[6pt]

\texttt{3. Sky Blue FC Profile}\\
\texttt{\hspace{1em}Sky Blue FC was founded in New Jersey in 2007, with its inaugural season in 2009.}\\
\texttt{\hspace{1em}Source: https://www.planetsport.com/...}\\[6pt]

\texttt{4. Founded in 1883 as Singers F.C.}\\
\texttt{\hspace{1em}Coventry City F.C., now known as "The Sky Blues", was founded in 1883.}\\
\texttt{\hspace{1em}Source: https://www.facebook.com/...}\\[6pt]

\texttt{Question: In a paragraph, could you tell me what you know about Sky Blue?<|im\_end|>}\\[6pt]

\texttt{<|im\_start|>assistant}
\texttt{<|im\_end|>}
\end{mdframed}

The response that gets the factuality score of 0.58.

\begin{mdframed}[backgroundcolor=SkyBlue!20, linewidth=0pt]
\small
"The term \"Sky Blue\" primarily refers to Coventry City Football Club, an English professional football team based in Coventry, which plays in the EFL Championship. The club, originally founded in 1883 as Singers F.C., adopted the nickname \"The Sky Blues\" due to their distinctive sky blue kits. They joined the Football League in 1898 and have played at the Ricoh Arena (formerly known as the Coventry Building Society Arena) since 2005, marking their first game there with a 3-0 victory over QPR. The nickname \"Sky Blue\" is also associated with Sky Blue FC, a women's soccer team based in New Jersey, USA, which was founded in 2007 and began play in the Women's Professional Soccer league in 2009. While both teams share the \"Sky Blue\" name, they are distinct entities from different countries and leagues."
\end{mdframed}

\section{Additional Entity Task Results}

In this section, we show the additional results for the entity task. 

\begin{figure*}[t]
\centering
\vspace{-5pt}
{\small\textbf{Need}\par}
\begin{subfigure}{0.23\textwidth}
    \centering
    \includegraphics[width=\linewidth]{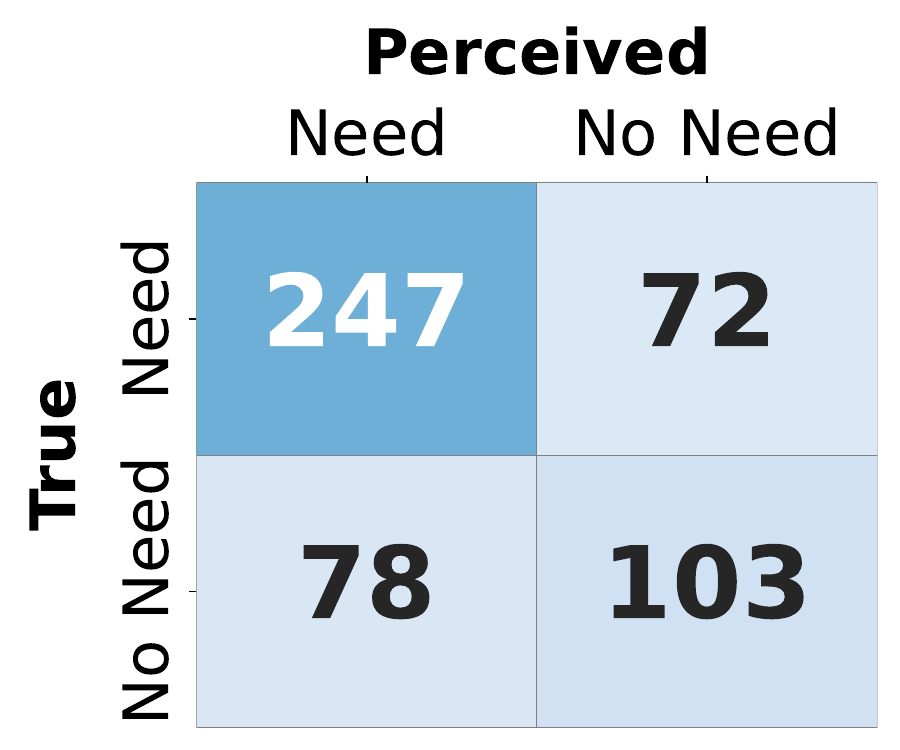}
    \caption{Qwen3-A3B}
\end{subfigure}\hfill
\begin{subfigure}{0.23\textwidth}
    \centering
    \includegraphics[width=\linewidth]{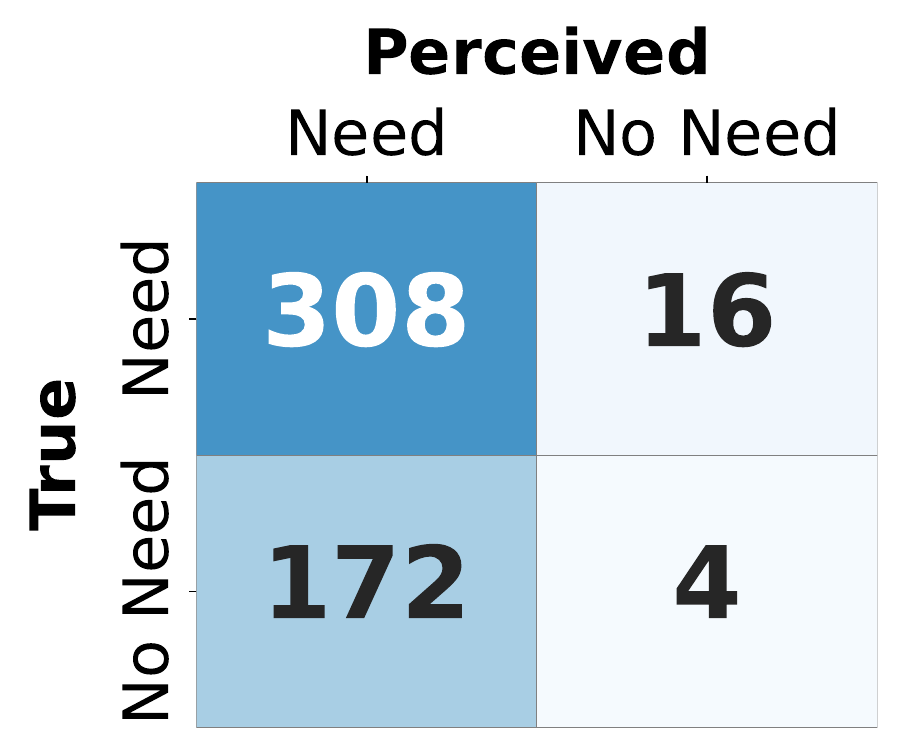}
    \caption{Qwen3-IT}
\end{subfigure}\hfill
\begin{subfigure}{0.23\textwidth}
    \centering
    \includegraphics[width=\linewidth]{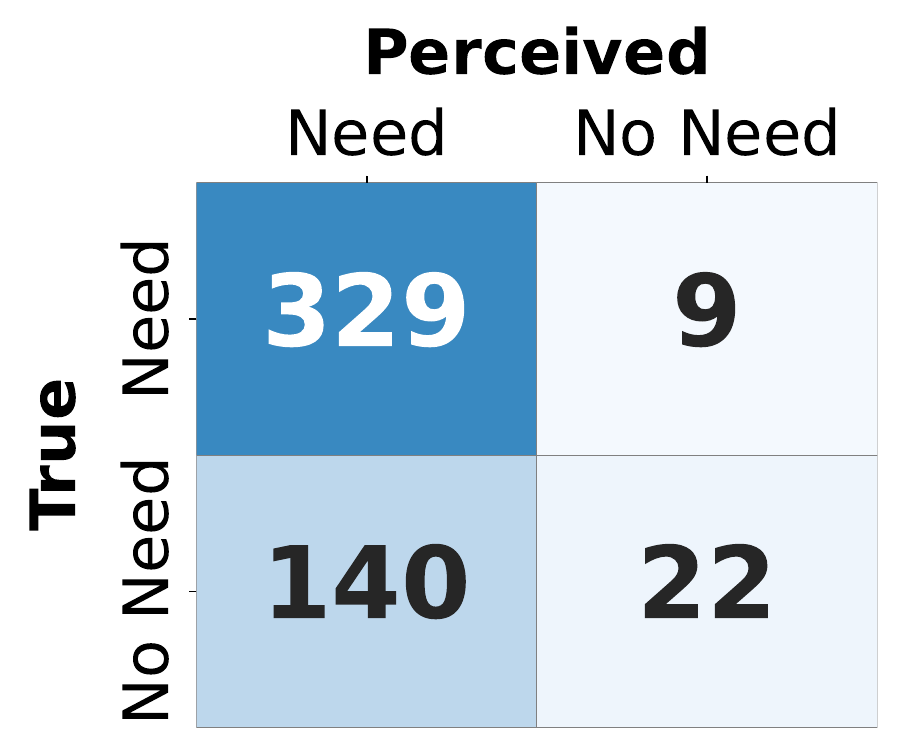}
    \caption{Gemma3-IT}
\end{subfigure}\hfill
\begin{subfigure}{0.23\textwidth}
    \centering
    \includegraphics[width=\linewidth]{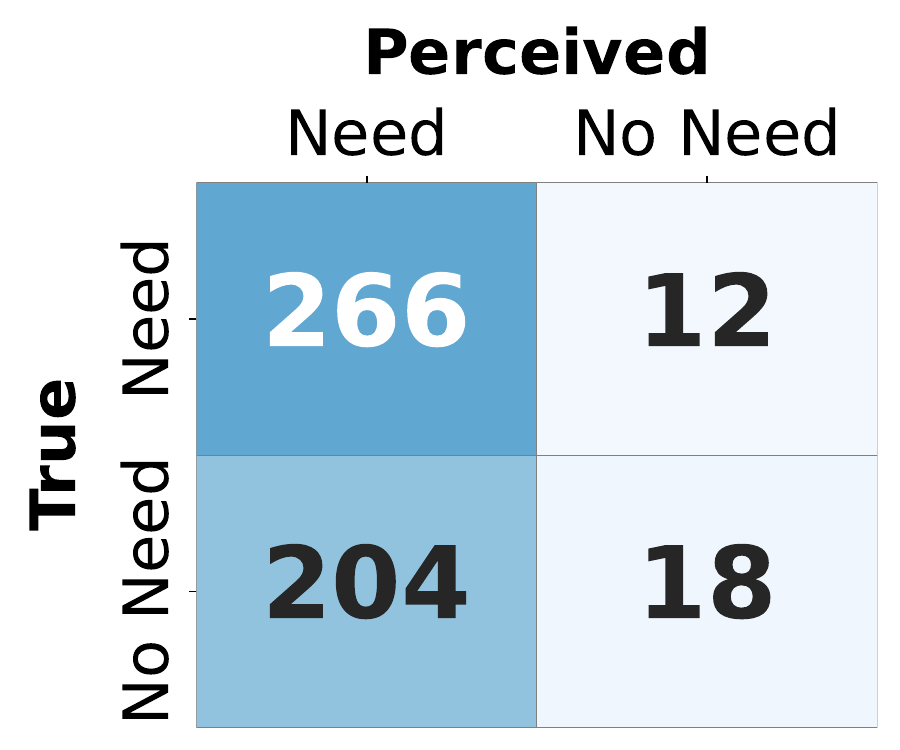}
    \caption{Mistral3.1-IT}
\end{subfigure}

{\small\textbf{Utility}\par}
\begin{subfigure}{0.23\textwidth}
    \centering
    \includegraphics[width=\linewidth]{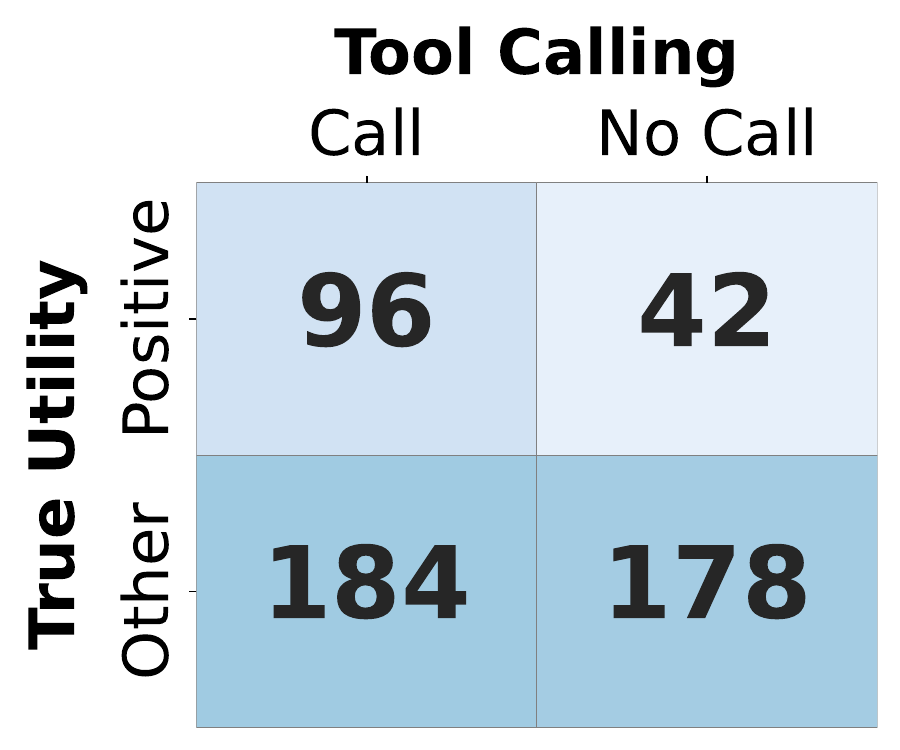}
    \caption{Qwen3-A3B}
\end{subfigure}\hfill
\begin{subfigure}{0.23\textwidth}
    \centering
    \includegraphics[width=\linewidth]{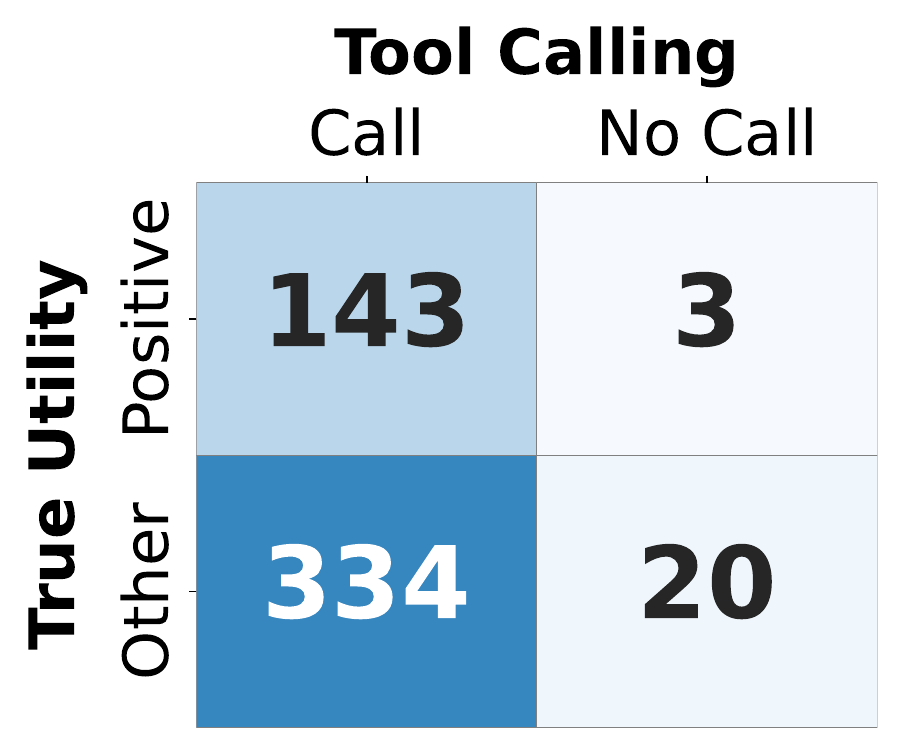}
    \caption{Qwen3-IT}
\end{subfigure}\hfill
\begin{subfigure}{0.23\textwidth}
    \centering
    \includegraphics[width=\linewidth]{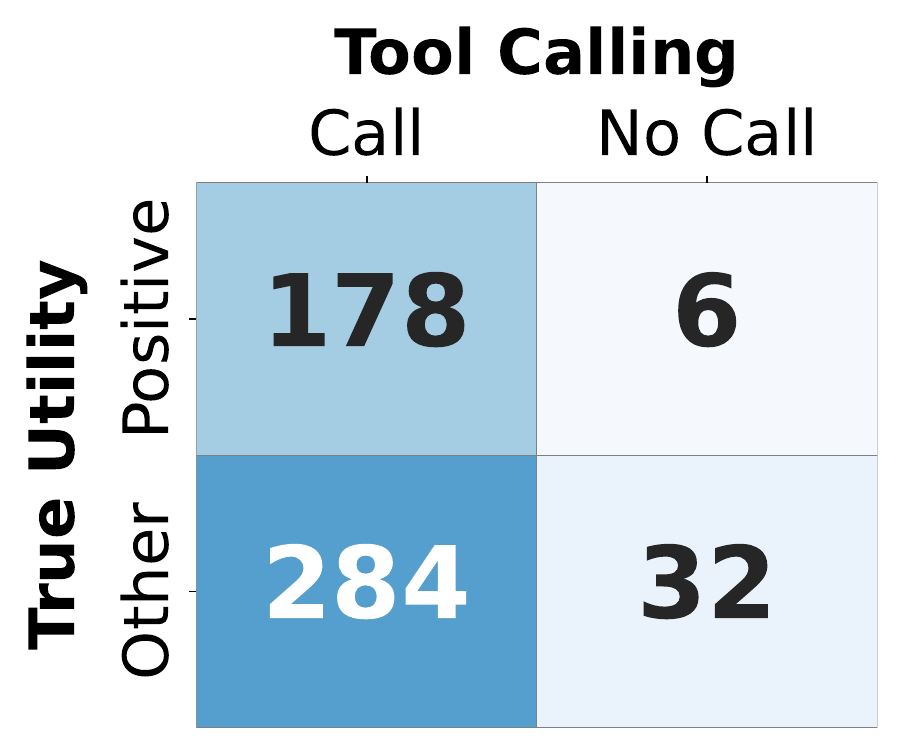}
    \caption{Gemma3-IT}
\end{subfigure}\hfill
\begin{subfigure}{0.23\textwidth}
    \centering
    \includegraphics[width=\linewidth]{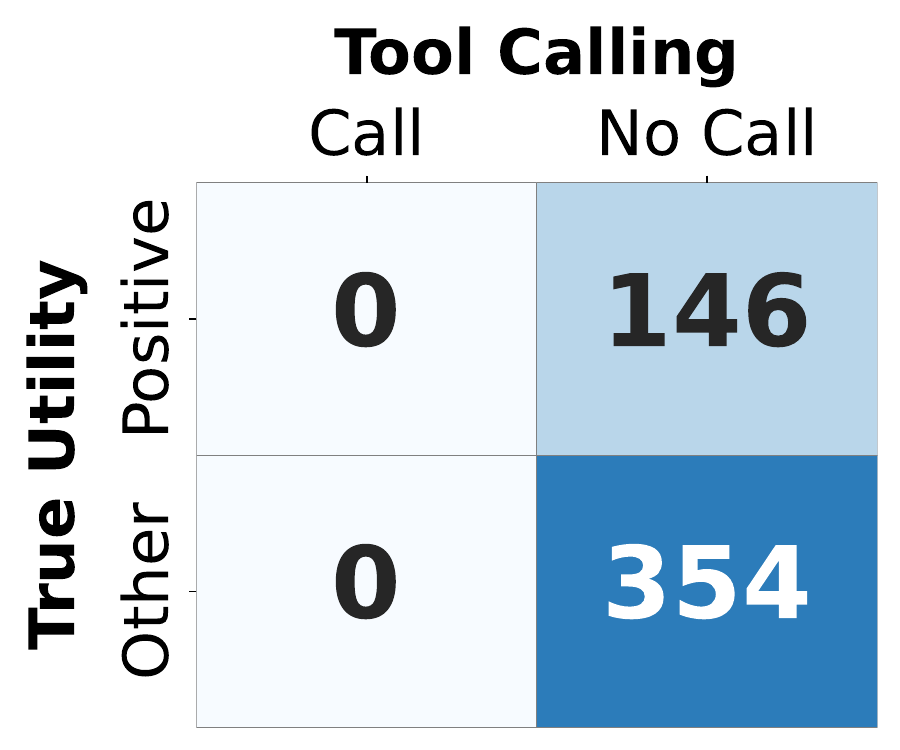}
    \caption{Mistral3.1-IT}
\end{subfigure}

\caption{\textbf{Perceived need and utility do not align with true need and utility for the additional Entity-task models.} Top: true vs.\ perceived need. Bottom: true vs.\ perceived utility.}
\label{fig:entity_true_perceived_additional}
\end{figure*}

\begin{figure*}
    \centering
    \includegraphics[width=\linewidth]{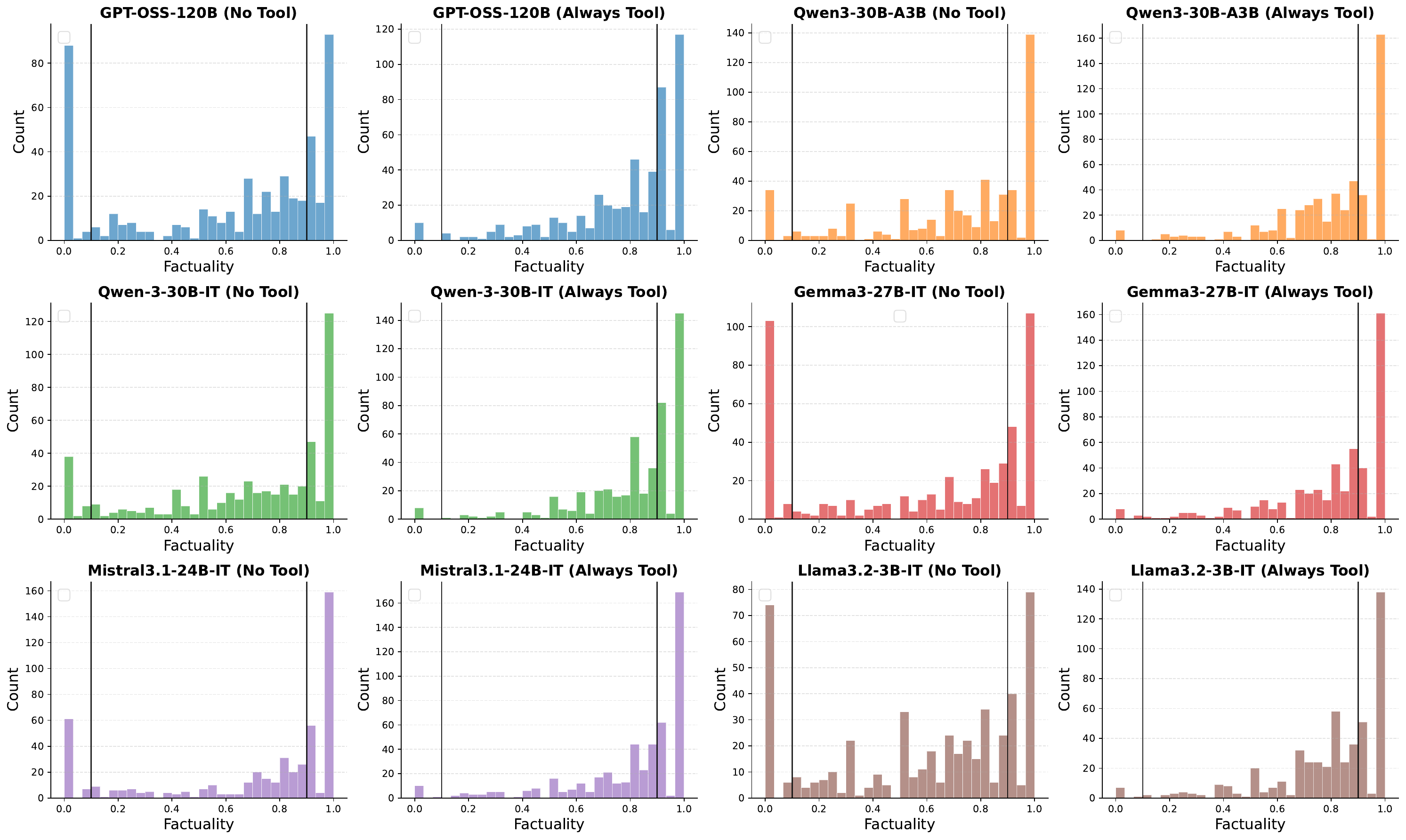}
    \caption{\textbf{Entity Task: factuality distribution across different models.}}
    \label{fig:entity_hist}
\end{figure*}

In Figure~\ref{fig:entity_hist}, we show the factuality score distribution across all the models and entities. Visualizing the distribution is important because aggregate metrics alone (e.g., mean or accuracy) can obscure underlying differences in model behavior. The distribution provides a more fine-grained view of how factuality scores are spread, revealing patterns such as skewness, variance, and the presence of extreme cases.
In particular, this figure allows us to examine how factuality shifts when tool use is enabled versus disabled. Rather than only observing average improvements, the distribution highlights whether gains are consistent across samples or driven by a subset of cases. 

\newcommand{\scorecalls}[2]{#1{\color{gray}\textsubscript{#2}}}

\begin{table*}[t]
\centering
\scriptsize
\setlength{\tabcolsep}{3.5pt}
\begin{adjustbox}{max width=0.99\textwidth,max totalheight=0.90\textheight,keepaspectratio}
\rowcolors{1}{blue!8}{blue!8}
\begin{tabular}{lllcccc}
\toprule
\textbf{Tool} & \textbf{Task} & \textbf{Model} & \textbf{No Tool} & \textbf{Always Tool} & \textbf{Optimal} & \textbf{Self-Decision} \\
\midrule
& & GPT-OSS-120B & \scorecalls{0.61}{0} & \scorecalls{0.76}{100} & \scorecalls{0.81}{61} & \scorecalls{0.72}{30} \\
&& Qwen3-30B-A3B & \scorecalls{0.70}{0} & \scorecalls{0.81}{100} & \scorecalls{0.88}{51} & \scorecalls{0.80}{56} \\
&& Qwen3-30B-IT & \scorecalls{0.68}{0} & \scorecalls{0.82}{100} & \scorecalls{0.87}{60} & \scorecalls{0.82}{95} \\
&& Mistral3.1-24B-IT & \scorecalls{0.70}{0} & \scorecalls{0.83}{100} & \scorecalls{0.88}{51} & \scorecalls{0.70}{0} \\
&& Llama3.2-3B-IT & \scorecalls{0.58}{0} & \scorecalls{0.70}{100} & \scorecalls{0.83}{57} & \scorecalls{0.70}{100} \\
\rowcolor{orange!12} && Gemma3-27B-IT & \scorecalls{0.60}{0} & \scorecalls{0.80}{100} & \scorecalls{0.85}{59} & \scorecalls{0.80}{92} \\
\rowcolor{green!12} & \multirow{-7}{*}{Entity} & GPT-5.5 & \scorecalls{0.85}{0} & \scorecalls{0.86}{100} & \scorecalls{0.94}{39} & \scorecalls{0.85}{32} \\
\cmidrule(lr){2-7}
& & GPT-OSS-120B & \scorecalls{0.53}{0} & \scorecalls{0.45}{100} & \scorecalls{0.66}{36} & \scorecalls{0.57}{38} \\
&& Qwen3-30B-A3B & \scorecalls{0.64}{0} & \scorecalls{0.64}{100} & \scorecalls{0.79}{40} & \scorecalls{0.64}{39} \\
&& Qwen3-30B-IT & \scorecalls{0.54}{0} & \scorecalls{0.61}{100} & \scorecalls{0.72}{51} & \scorecalls{0.61}{67} \\
&& Mistral3.1-24B-IT & \scorecalls{0.56}{0} & \scorecalls{0.60}{100} & \scorecalls{0.74}{46} & \scorecalls{0.56}{44} \\
&& Llama3.2-3B-IT & \scorecalls{0.53}{0} & \scorecalls{0.54}{100} & \scorecalls{0.68}{45} & \scorecalls{0.55}{64} \\
\rowcolor{orange!12} && Gemma3-27B-IT & \scorecalls{0.54}{0} & \scorecalls{0.63}{100} & \scorecalls{0.74}{51} & \scorecalls{0.62}{58} \\
\rowcolor{green!12} & \multirow{-7}{*}{InVivoQuery} & GPT-5.5 & \scorecalls{0.85}{0} & \scorecalls{0.87}{100} & \scorecalls{0.94}{40} & \scorecalls{0.87}{32} \\
\cmidrule(lr){2-7}
& & GPT-OSS-120B & \scorecalls{0.38}{0} & \scorecalls{0.76}{100} & \scorecalls{0.79}{41} & \scorecalls{0.76}{100} \\
&& Qwen3-30B-A3B & \scorecalls{0.20}{0} & \scorecalls{0.62}{100} & \scorecalls{0.66}{46} & \scorecalls{0.62}{100} \\
&& Qwen3-30B-IT & \scorecalls{0.31}{0} & \scorecalls{0.74}{100} & \scorecalls{0.77}{46} & \scorecalls{0.74}{100} \\
&& Mistral3.1-24B-IT & \scorecalls{0.30}{0} & \scorecalls{0.69}{100} & \scorecalls{0.73}{44} & \scorecalls{0.69}{80} \\
&& Llama3.2-3B-IT & \scorecalls{0.17}{0} & \scorecalls{0.65}{100} & \scorecalls{0.68}{51} & \scorecalls{0.65}{100} \\
\rowcolor{orange!12} && Gemma3-27B-IT & \scorecalls{0.44}{0} & \scorecalls{0.69}{100} & \scorecalls{0.76}{32} & \scorecalls{0.69}{100} \\
\rowcolor{green!12} \multirow{-21}{*}{\rotatebox[origin=c]{90}{Web Search}} & \multirow{-7}{*}{BFCL} & GPT-5.5 & \scorecalls{0.65}{0} & \scorecalls{0.57}{100} & \scorecalls{0.76}{11} & \scorecalls{0.66}{49} \\
\midrule
& & GPT-OSS-120B & \scorecalls{0.66}{0} & \scorecalls{0.64}{100} & \scorecalls{0.70}{4} & \scorecalls{0.65}{31} \\
&& Qwen3-30B-A3B & \scorecalls{0.62}{0} & \scorecalls{0.57}{100} & \scorecalls{0.70}{8} & \scorecalls{0.61}{63} \\
&& Qwen3-30B-IT & \scorecalls{0.60}{0} & \scorecalls{0.56}{100} & \scorecalls{0.66}{6} & \scorecalls{0.59}{56} \\
&& Mistral3.1-24B-IT & \scorecalls{0.52}{0} & \scorecalls{0.28}{100} & \scorecalls{0.57}{5} & \scorecalls{0.51}{5} \\
&& Llama3.2-3B-IT & \scorecalls{0.18}{0} & \scorecalls{0.02}{100} & \scorecalls{0.19}{1} & \scorecalls{0.02}{97} \\
\rowcolor{orange!12} && Gemma3-27B-IT & \scorecalls{0.59}{0} & \scorecalls{0.55}{100} & \scorecalls{0.67}{8} & \scorecalls{0.56}{94} \\
\rowcolor{green!12} & \multirow{-7}{*}{GSM-Hard} & GPT-5.5 & \scorecalls{0.72}{0} & \scorecalls{0.75}{100} & \scorecalls{0.76}{4} & \scorecalls{0.73}{29} \\
\cmidrule(lr){2-7}
& & GPT-OSS-120B & \scorecalls{0.90}{0} & \scorecalls{1.00}{100} & \scorecalls{1.00}{10} & \scorecalls{1.00}{61} \\
&& Qwen3-30B-A3B & \scorecalls{0.42}{0} & \scorecalls{1.00}{100} & \scorecalls{1.00}{58} & \scorecalls{1.00}{97} \\
&& Qwen3-30B-IT & \scorecalls{0.78}{0} & \scorecalls{1.00}{100} & \scorecalls{1.00}{22} & \scorecalls{1.00}{100} \\
&& Mistral3.1-24B-IT & \scorecalls{0.31}{0} & \scorecalls{0.29}{100} & \scorecalls{0.43}{12} & \scorecalls{0.29}{11} \\
&& Llama3.2-3B-IT & \scorecalls{0.17}{0} & \scorecalls{0.00}{100} & \scorecalls{0.17}{0} & \scorecalls{0.00}{100} \\
\rowcolor{orange!12} && Gemma3-27B-IT & \scorecalls{0.42}{0} & \scorecalls{1.00}{100} & \scorecalls{1.00}{58} & \scorecalls{1.00}{100} \\
\rowcolor{green!12} & \multirow{-7}{*}{Multiplication} & GPT-5.5 & \scorecalls{0.99}{0} & \scorecalls{1.00}{100} & \scorecalls{1.00}{1} & \scorecalls{1.00}{66} \\
\cmidrule(lr){2-7}
& & GPT-OSS-120B & \scorecalls{0.11}{0} & \scorecalls{1.00}{100} & \scorecalls{1.00}{89} & \scorecalls{0.99}{97} \\
&& Qwen3-30B-A3B & \scorecalls{0.01}{0} & \scorecalls{1.00}{100} & \scorecalls{1.00}{99} & \scorecalls{1.00}{100} \\
&& Qwen3-30B-IT & \scorecalls{0.00}{0} & \scorecalls{0.95}{100} & \scorecalls{0.95}{94} & \scorecalls{0.95}{100} \\
&& Mistral3.1-24B-IT & \scorecalls{0.00}{0} & \scorecalls{0.66}{100} & \scorecalls{0.66}{66} & \scorecalls{0.35}{56} \\
&& Llama3.2-3B-IT & \scorecalls{0.00}{0} & \scorecalls{0.00}{100} & \scorecalls{0.00}{0} & \scorecalls{0.00}{100} \\
\rowcolor{orange!12} && Gemma3-27B-IT & \scorecalls{0.01}{0} & \scorecalls{1.00}{100} & \scorecalls{1.00}{100} & \scorecalls{1.00}{100} \\
\rowcolor{green!12} \multirow{-21}{*}{\rotatebox[origin=c]{90}{Calculator}} & \multirow{-7}{*}{Large-digit mult.} & GPT-5.5 & \scorecalls{0.81}{0} & \scorecalls{1.00}{100} & \scorecalls{1.00}{19} & \scorecalls{1.00}{100} \\
\bottomrule
\end{tabular}
\end{adjustbox}
\caption{Complete natural-setting performance comparison across tools and tasks. Each cell reports task score, with tool-call rate (\%) in gray subscript. Row colors indicate the harness: \colorbox{blue!8}{\strut Trained}, \colorbox{orange!12}{\strut Custom}, and \colorbox{green!12}{\strut OpenAI}. The table excludes budget-constrained variants: \textsc{No Tool}, \textsc{Always Tool}, and \textsc{Optimal} provide reference points, while \textsc{Self-Decision} compares the model's native policy with the latent need controller.}
\label{tab:main-full}
\vspace{-10pt}
\end{table*}

\subsection{Normative Lens}

We operationalize \textit{True Need} and \textit{True Utility} by comparing model performance under the \notool{} and \withtool{} settings. \textit{True Need} corresponds to instances where performance without tool use falls into the \textit{Low} or \textit{Mid} categories, indicating that external tool support is likely necessary. We further define \textit{True Positive Utility} as cases where tool use leads to performance improvements (e.g., \textit{Low} $\rightarrow$ \textit{Mid/High}), while \textit{True Negative Utility} captures cases where performance degrades with tool use. Instances where performance remains unchanged are categorized as neutral utility.

\subsubsection{Factuality}

\begin{figure*}[t]
\centering
\begin{subfigure}{0.25\linewidth}
    \centering
    \includegraphics[width=\linewidth]{entity-openai_gpt-oss-120b_bucket_confusion_matrix.pdf}
    \caption{GPT-OSS-120B}
\end{subfigure}
\hfill
\begin{subfigure}{0.25\linewidth}
    \centering
    \includegraphics[width=\linewidth]{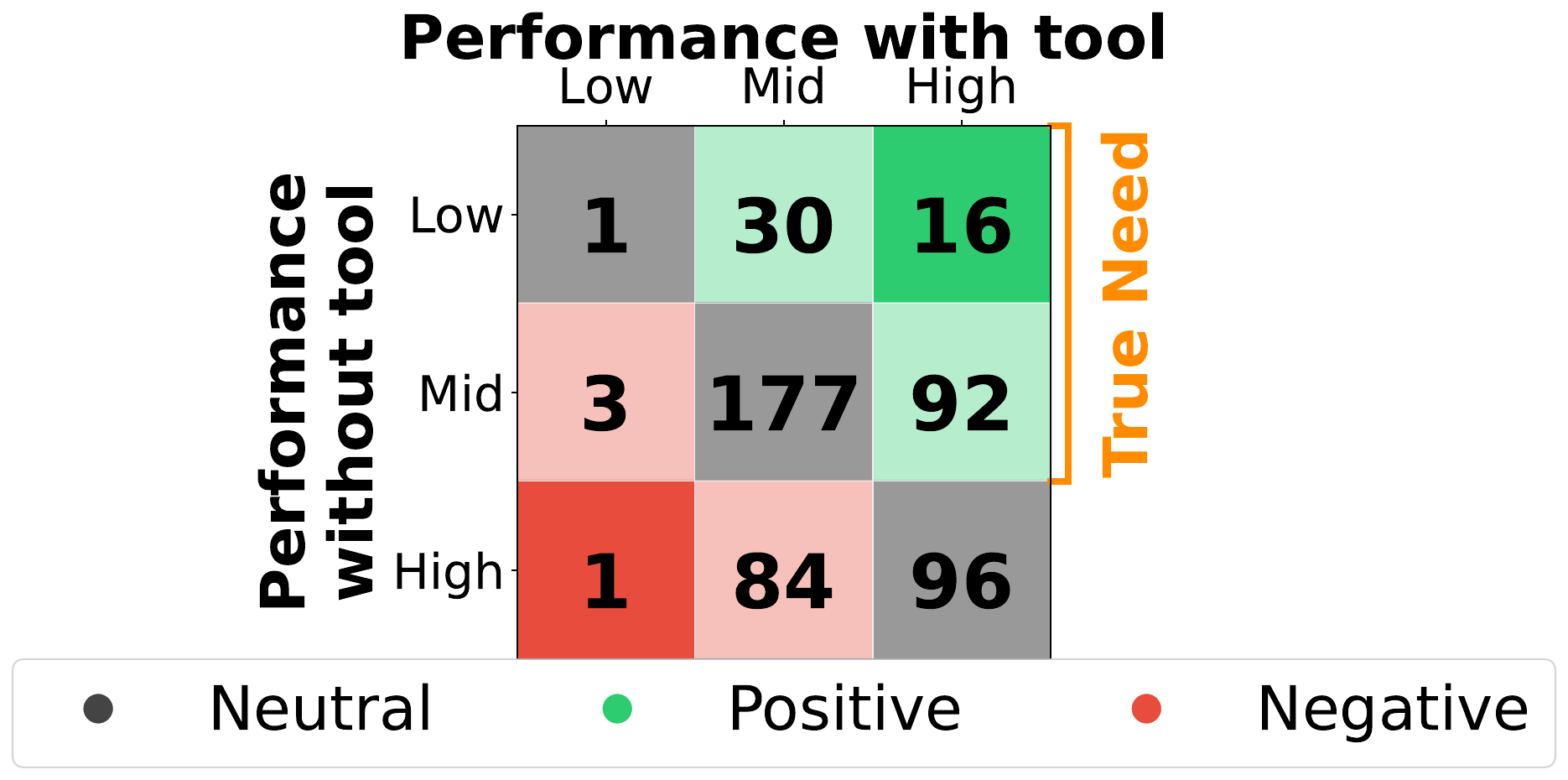}
    \caption{Qwen3-30B-A3B}
\end{subfigure}
\hfill
\begin{subfigure}{0.25\linewidth}
    \centering
    \includegraphics[width=\linewidth]{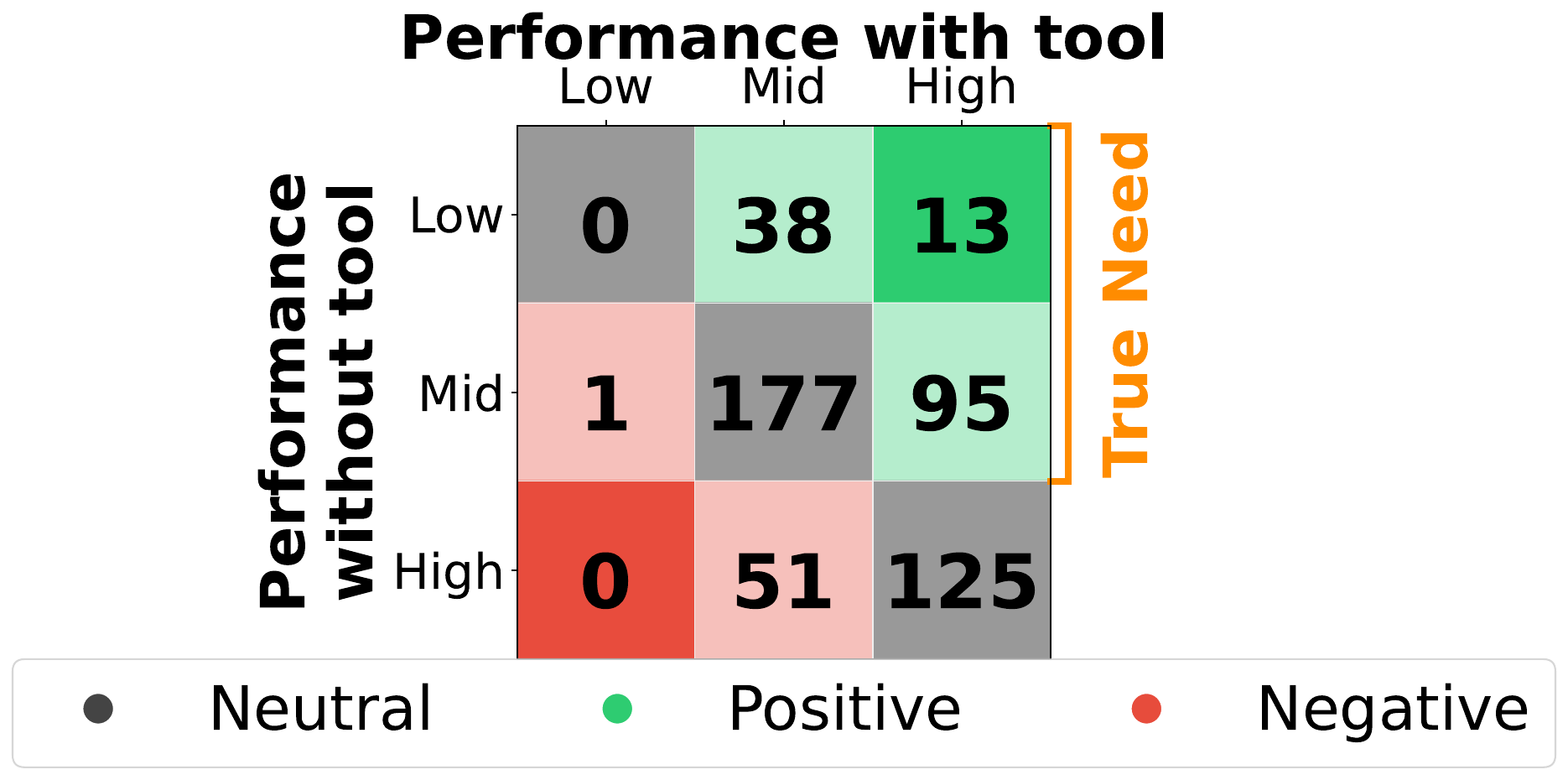}
    \caption{Qwen3-30B-A3B-Instruct}
\end{subfigure}
\begin{subfigure}{0.25\linewidth}
    \centering
    \includegraphics[width=\linewidth]{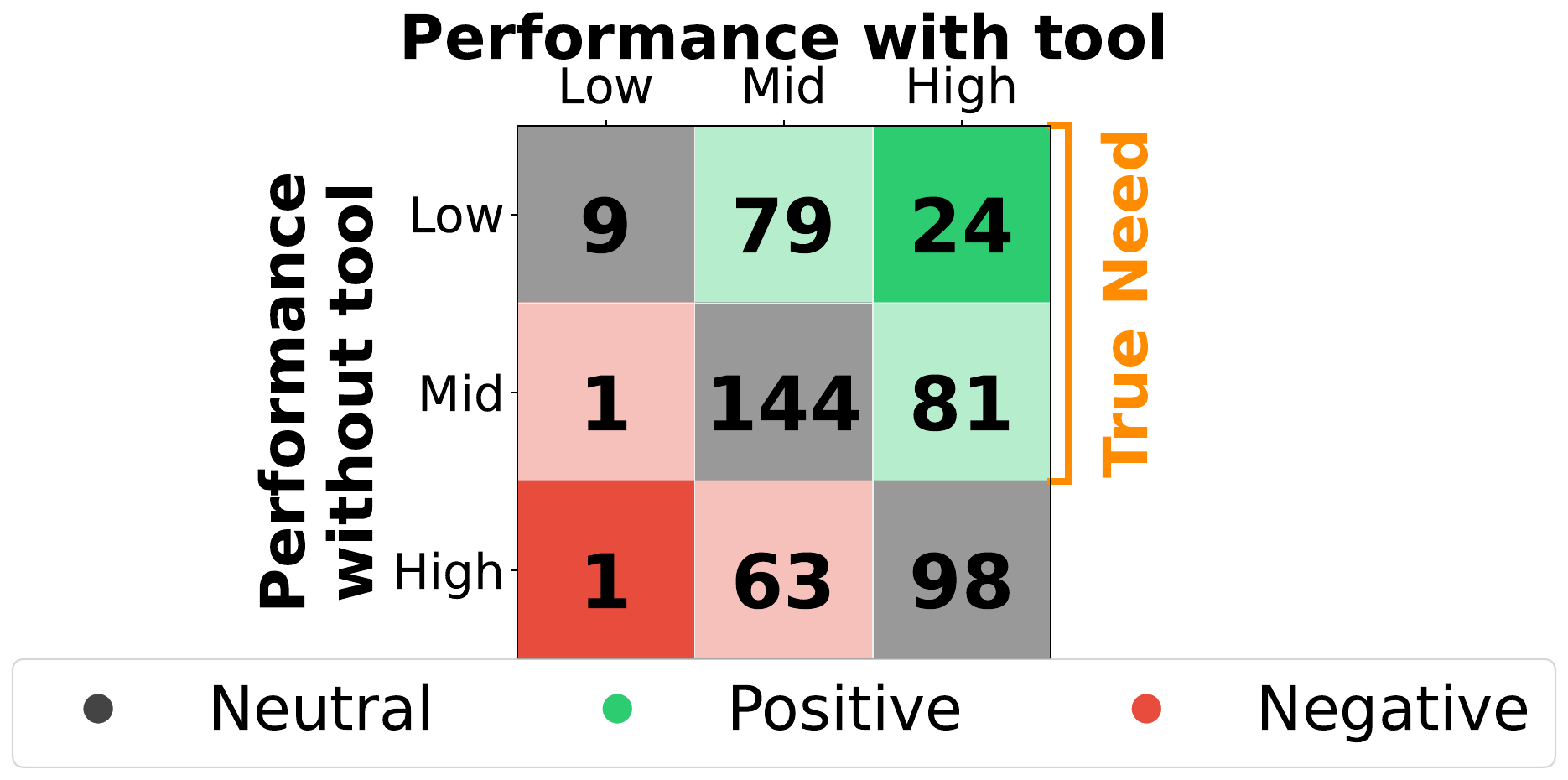}
    \caption{Gemma-3-27B-IT}
\end{subfigure}
\hfill
\begin{subfigure}{0.25\linewidth}
    \centering
    \includegraphics[width=\linewidth]{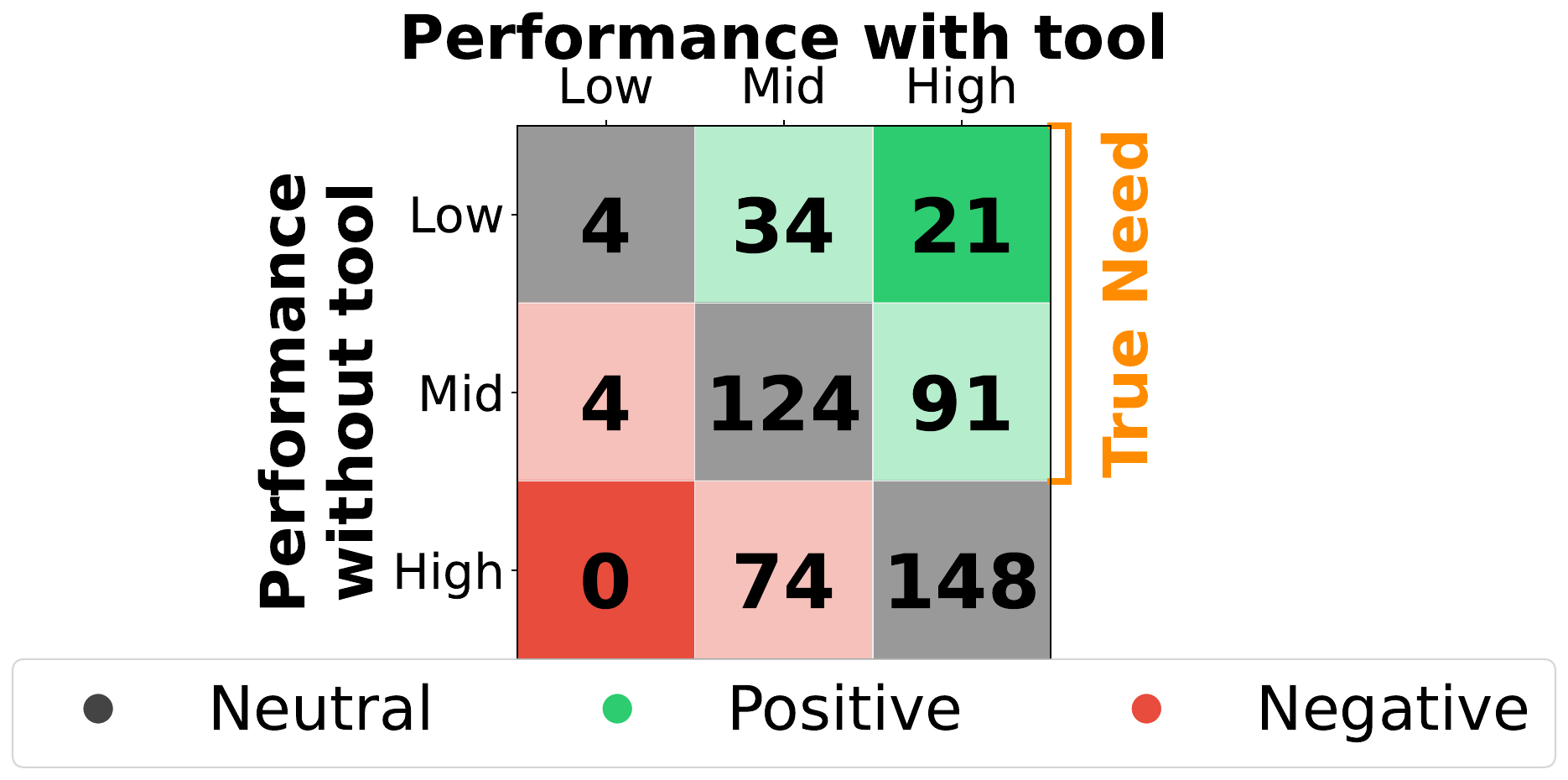}
    \caption{Mistral-3.1-24B-IT}
\end{subfigure}
\hfill
\begin{subfigure}{0.25\linewidth}
    \centering
    \includegraphics[width=\linewidth]{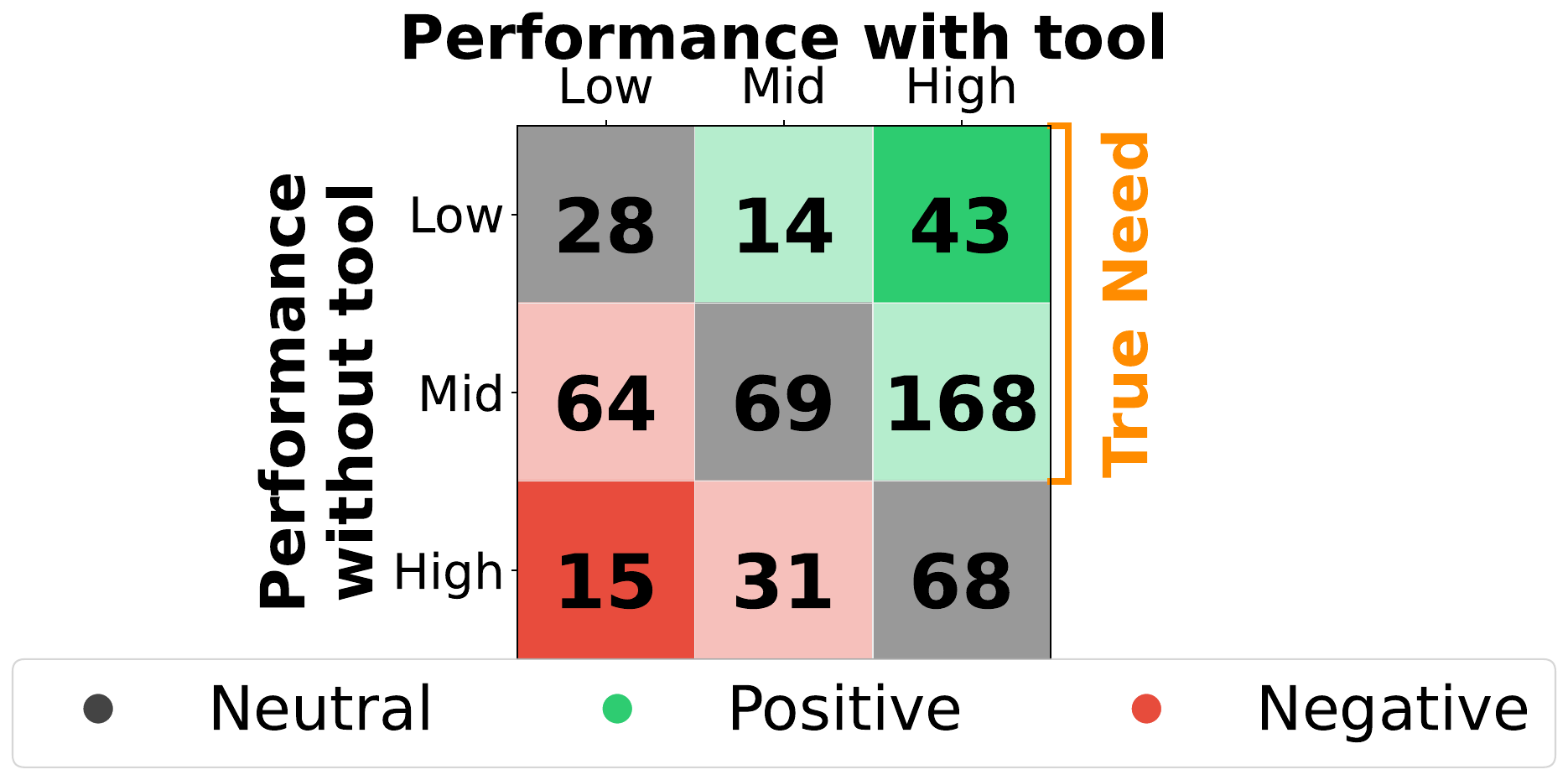}
    \caption{Llama-3.2-3B-IT}
\end{subfigure}

\caption{\textbf{Entity task: No-Tool vs.\ Always-Tool performance.}
        Rows group entities by the model’s factuality score \textit{without a tool} (reflecting parametric knowledge), while columns group scores when tool use is \textit{forced}. Each cell reports the count and the column percentage. Off-diagonal cells indicate performance shifts due to tool use: cells above the diagonal show cases where the tool has {\color{green}\textit{positive utility}}, while cells below the diagonal indicate cases where the tool has {{\color{red}\textit{negative utility}}}. The dashed bracket marks the region of {{\color{orange}\textit{True Need}}}, where \textit{Low} or \textit{Mid} No-Tool scores suggest insufficient parametric knowledge and thus a likely need for an external tool.}
\label{fig:entity_all_actul_need_utility}
\end{figure*}

As shown in Figure~\ref{fig:entity_all_actul_need_utility}, a consistent pattern emerges across all models: \textbf{tool use is most beneficial when it is truly needed, can be harmful when unnecessary, and is often redundant otherwise}. This observation highlights the importance of accurately predicting when to invoke external tools, as indiscriminate usage may introduce noise or errors rather than improving factuality.

\subsubsection{Completeness and Relevance}
\label{sec:other_metrics}
\begin{figure*}[t]
\centering
\begin{subfigure}{0.49\linewidth}
    \centering
    \includegraphics[width=\linewidth]{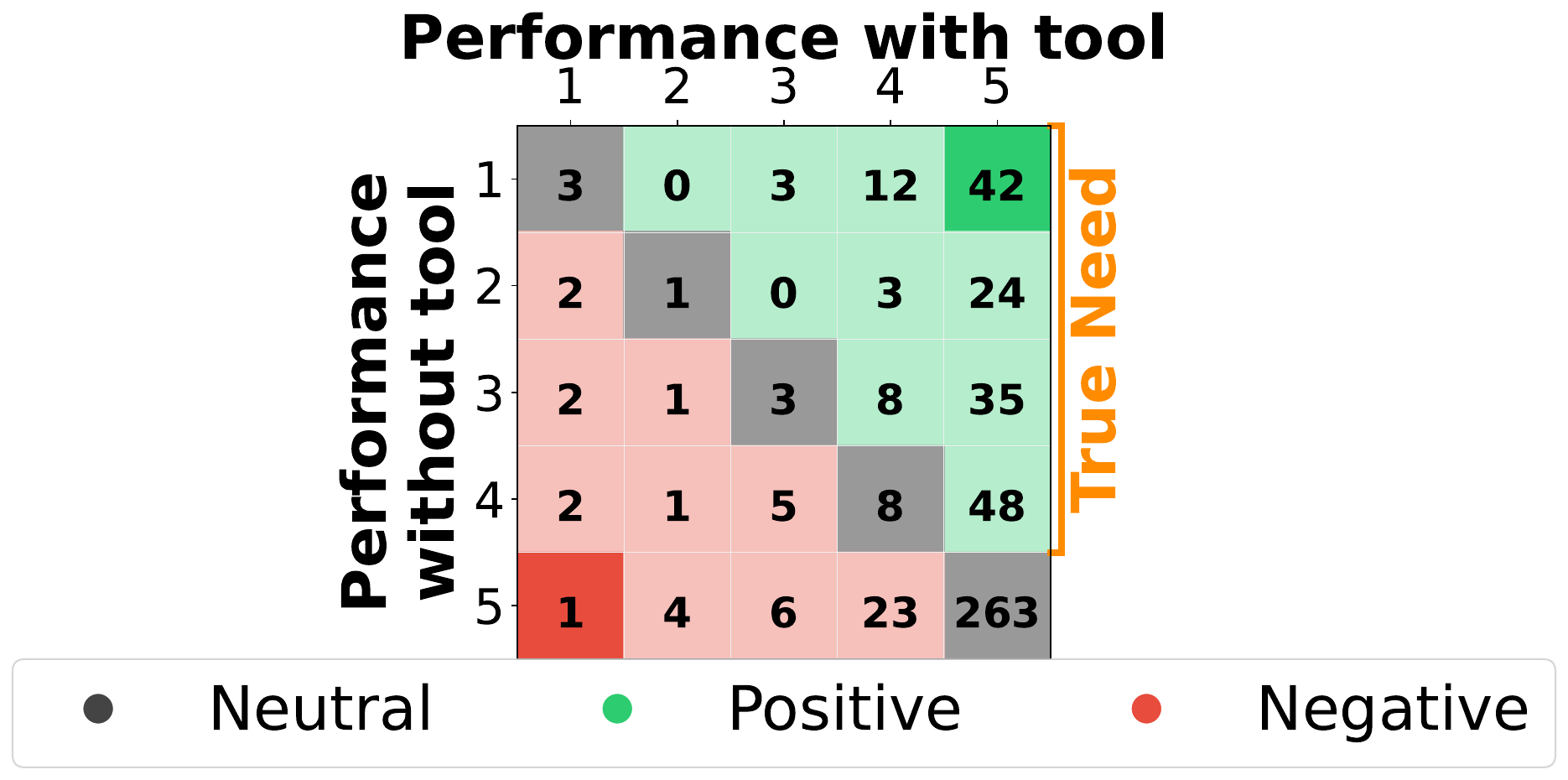}
    \caption{Completeness}
\end{subfigure}
\hfill
\begin{subfigure}{0.49\linewidth}
    \centering
    \includegraphics[width=\linewidth]{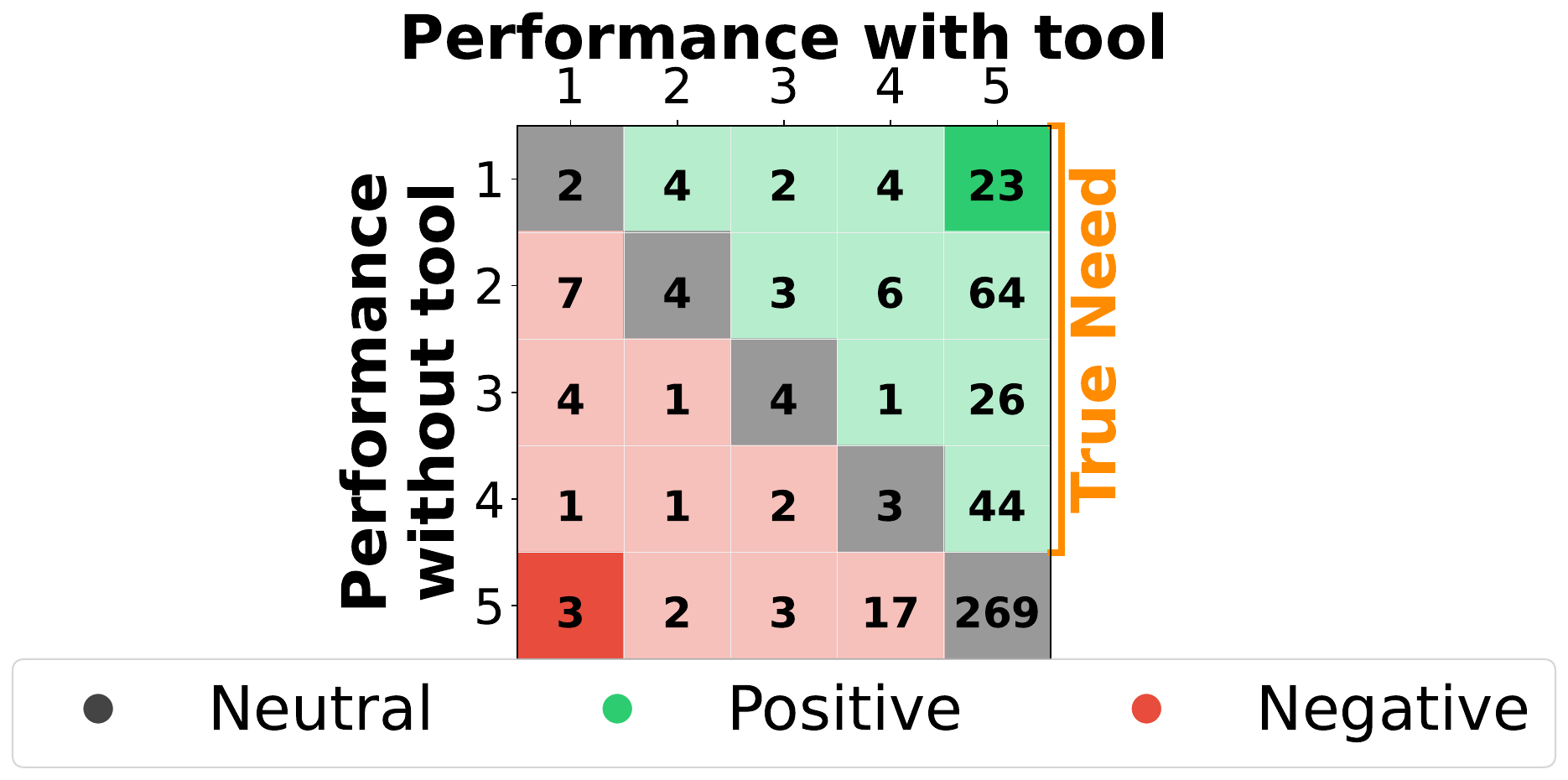}
    \caption{Relevance}
\end{subfigure}

\caption{\textbf{Entity task, GPT-OSS-120B: No-Tool vs.\ Always-Tool performance on completeness and relevance.}
        Rows group entities by the model’s 5-Likert score on completeness and relevance \textit{without a tool} (reflecting parametric knowledge), while columns group scores when tool use is \textit{forced}. Each cell reports the count and the column percentage. Off-diagonal cells indicate performance shifts due to tool use: cells above the diagonal show cases where the tool has {\color{green}\textit{positive utility}}, while cells below the diagonal indicate cases where the tool has {{\color{red}\textit{negative utility}}}. The dashed bracket marks the region of {{\color{orange}\textit{True Need}}}.}
\label{fig:complete_relevance}
\end{figure*}

\change{
The bound generalizes beyond factuality. In our tasks (Entity, InVivoQuery, BFCL), factuality is the natural metric, as the user's goal is to obtain accurate factual knowledge. However, to address the reviewer's concern directly, we additionally verified this on completeness and relevance using LLM-as-a-Judge: when the NO TOOL response already scores perfectly on completeness or relevance, i.e. as shown in Figure~\ref{fig:complete_relevance} when the likert scale reading is 5, retrieval yields negligible gain — and in fact introduces slight degradation (11.4\% ((1+4+6+23)/(1+4+6+23+263)) for high-completeness and 8.5\% ((3+2+3+17)/(3+2+3+17+269)) for relevance responses drop under tool use). This is consistent with our factuality findings and confirms that the bound is not an artifact of the chosen metric.
}

\subsection{Descriptive Lens}
\label{sec:appendix_descriptive_lens}

\begin{figure*}[t]
    \centering
    \includegraphics[width=0.78\textwidth]{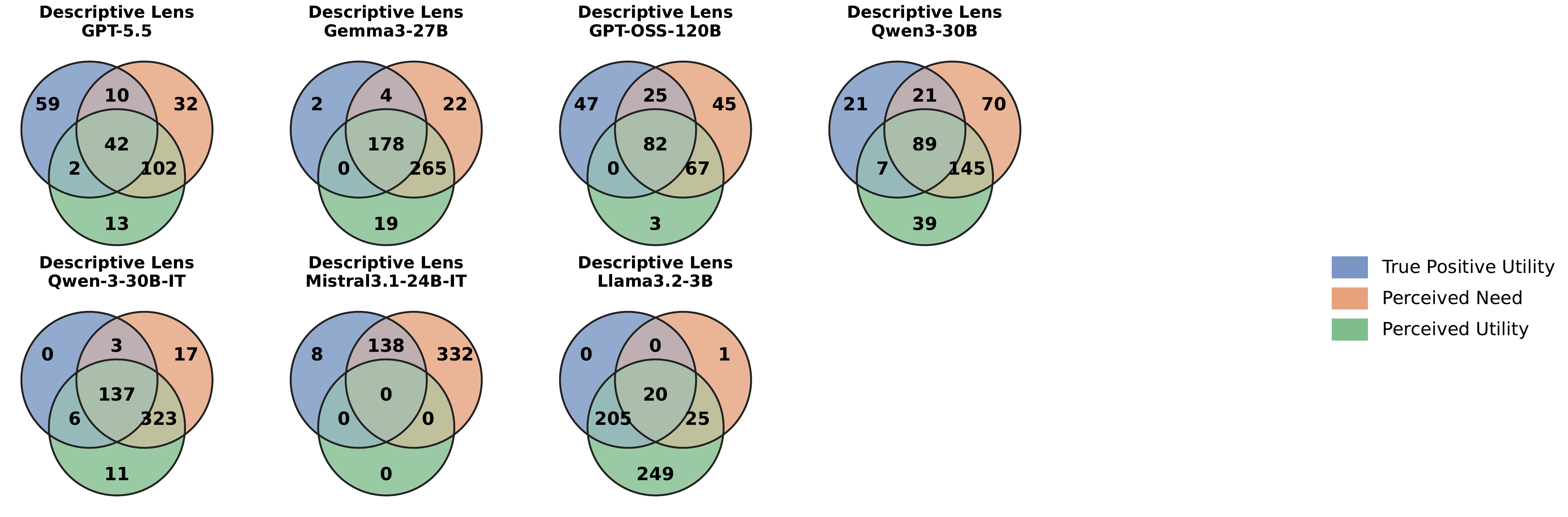}
    \caption{[Entity Task] Venn diagrams of \textbf{True Positive Utility, Perceived Need, and Perceived Utility}. Each panel shows their empirical overlap for one model. Calls outside true positive utility are non-beneficial; true-positive-utility cases outside perceived utility are missed opportunities. Perceived need is a separate self-assessment and need not be nested within either utility set.}
    \label{fig:venn-all}
\end{figure*}

As shown in Figure~\ref{fig:venn-all}, there is a consistent misalignment between the perceived need and utility and the true positive utility across all models. This discrepancy indicates that models often fail to accurately identify when tool use is genuinely beneficial. As a result, none of the models achieve optimal tool-calling performance, since effective tool use critically depends on correctly aligning perceived need with actual utility.

\begin{figure*}
\centering
\vspace{-10pt}

\begin{subfigure}{0.25\linewidth}
    \centering
    \includegraphics[width=\linewidth]{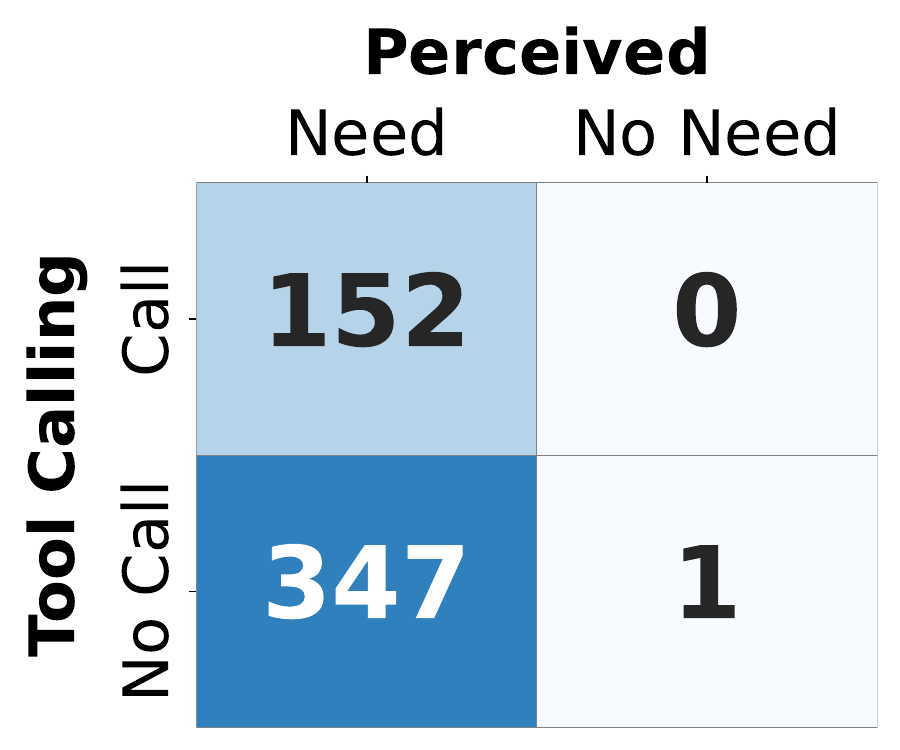}
    \caption{GPT-OSS-120B}
\end{subfigure}\hfill
\begin{subfigure}{0.25\linewidth}
    \centering
    \includegraphics[width=\linewidth]{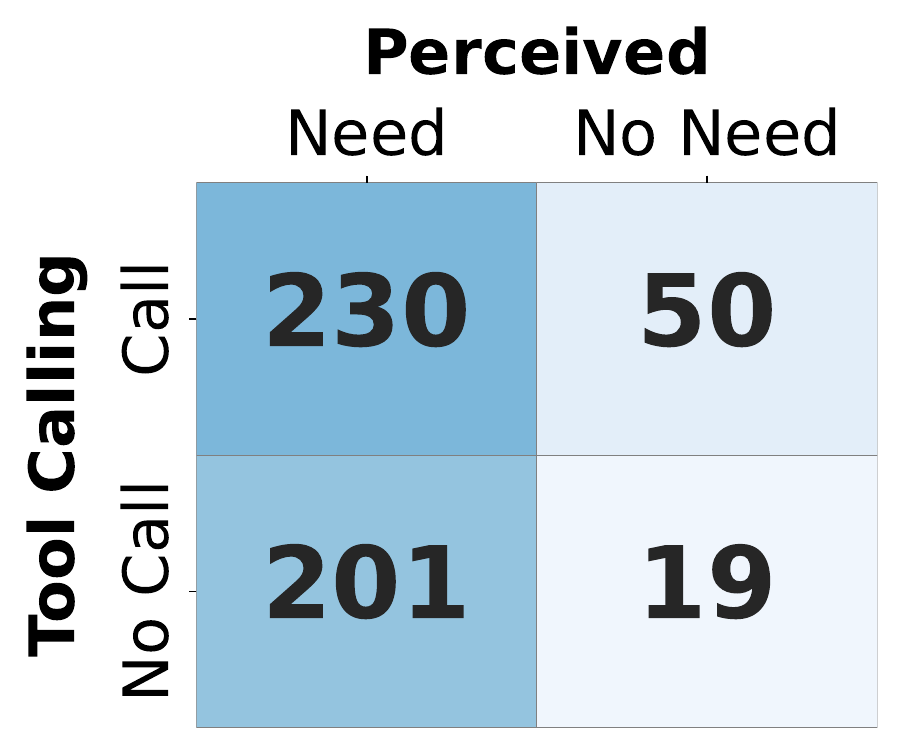}
    \caption{Qwen3-30B-A3B}
\end{subfigure}\hfill
\begin{subfigure}{0.25\linewidth}
    \centering
    \includegraphics[width=\linewidth]{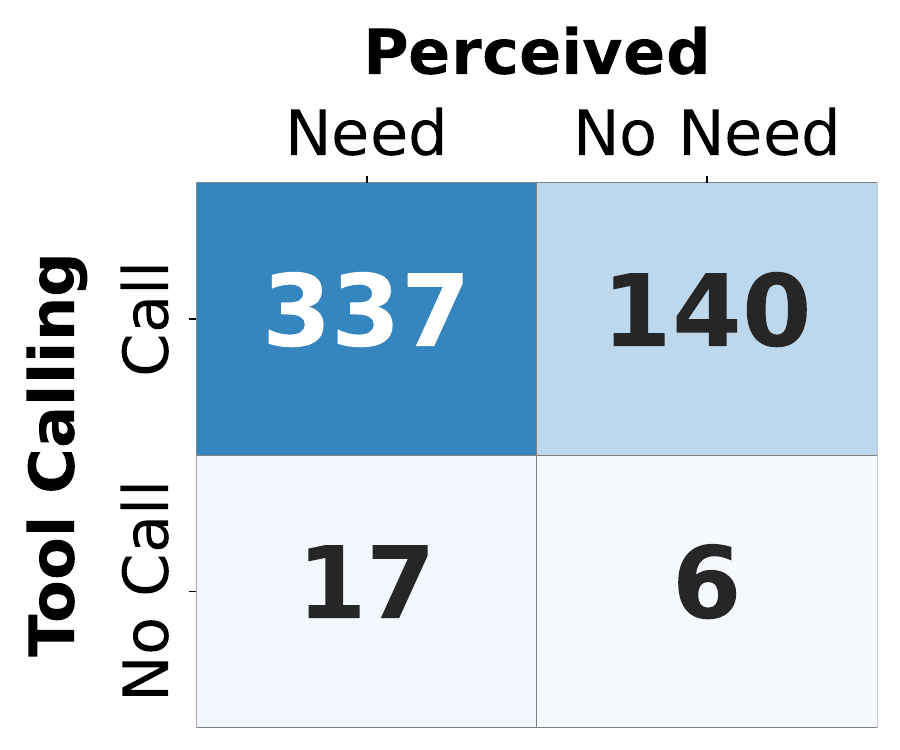}
    \caption{Qwen3-30B-A3B-IT}
\end{subfigure}\hfill
\begin{subfigure}{0.25\linewidth}
    \centering
    \includegraphics[width=\linewidth]{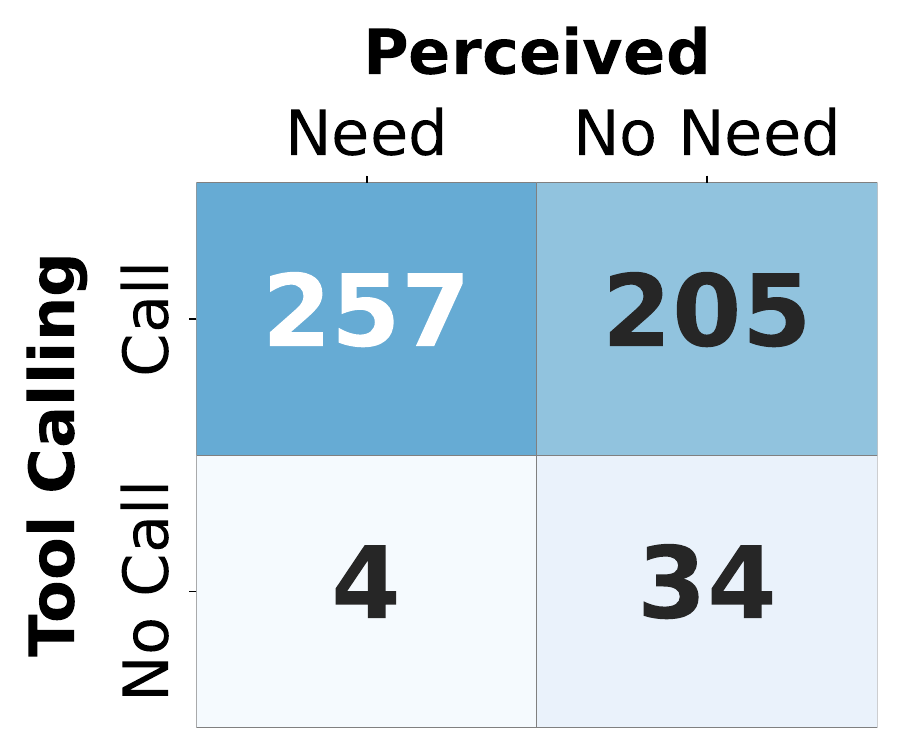}
    \caption{Gemma-3-27B-IT}
\end{subfigure}\hfill
\begin{subfigure}{0.25\linewidth}
    \centering
    \includegraphics[width=\linewidth]{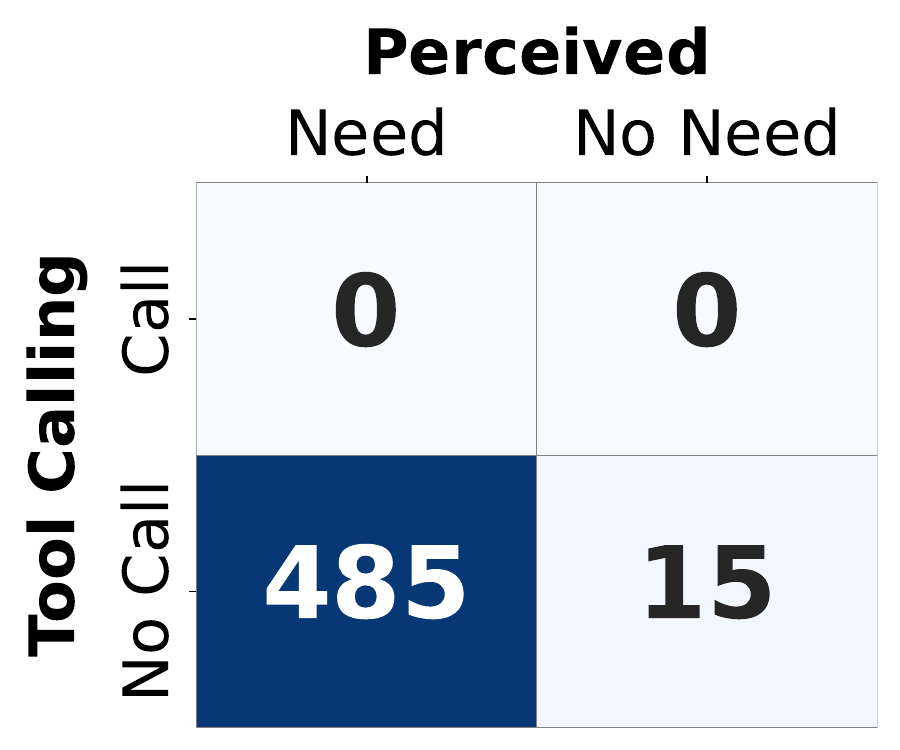}
    \caption{Mistral-3.1-24B-IT}
\end{subfigure}\hfill
\begin{subfigure}{0.25\linewidth}
    \centering
    \includegraphics[width=\linewidth]{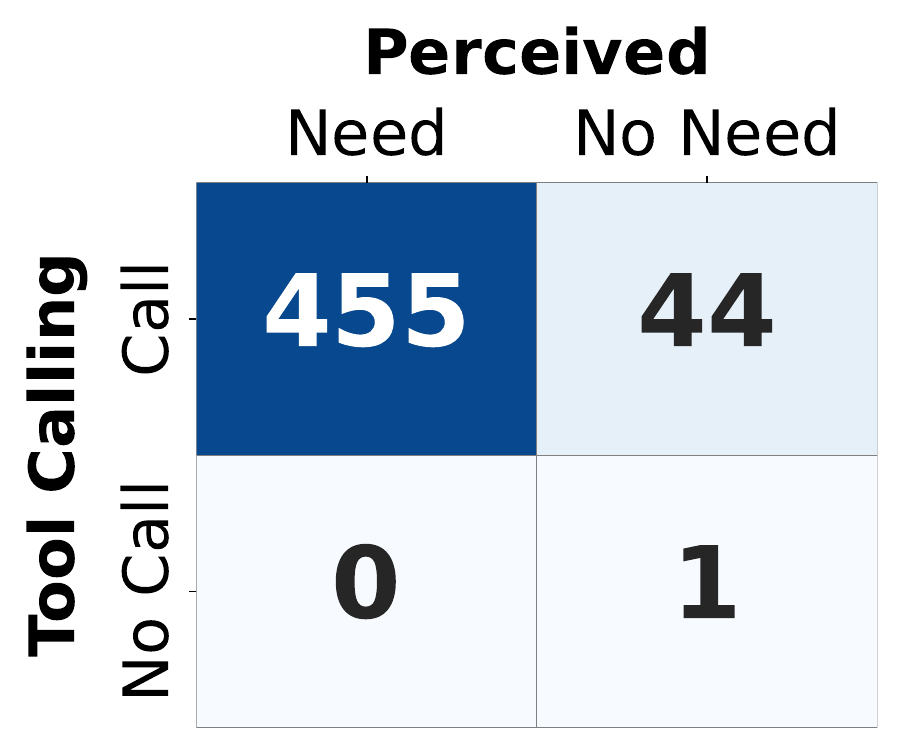}
    \caption{Llama-3.2-3B-IT}
\end{subfigure}

\vspace{6pt}
\textbf{(a) Perceived-need prompt v2}

\vspace{10pt}

\begin{subfigure}{0.25\linewidth}
    \centering
    \includegraphics[width=\linewidth]{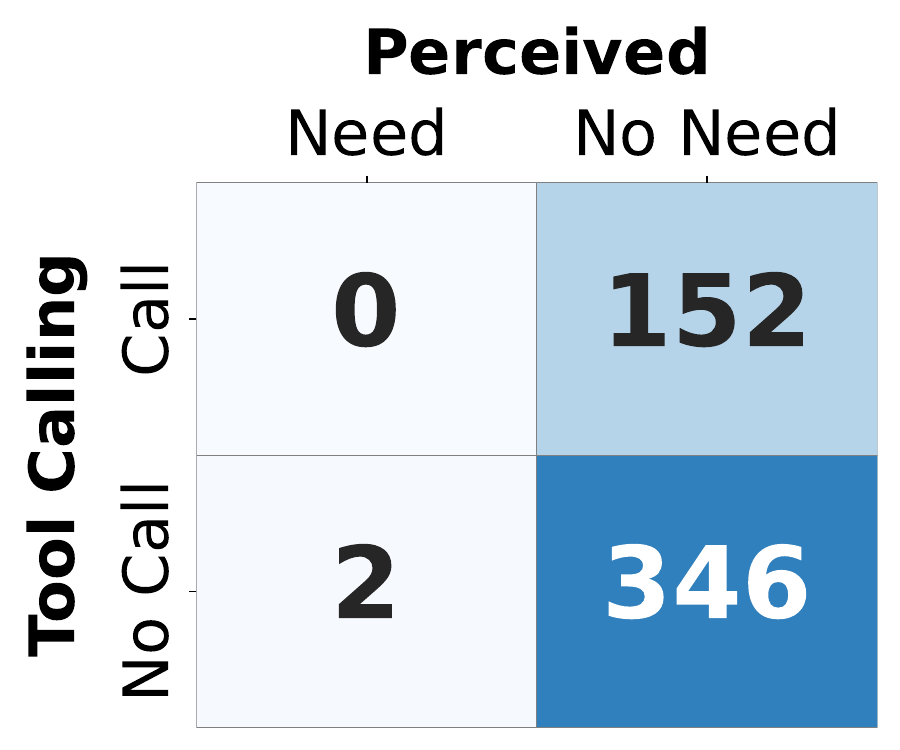}
    \caption{GPT-OSS-120B}
\end{subfigure}\hfill
\begin{subfigure}{0.25\linewidth}
    \centering
    \includegraphics[width=\linewidth]{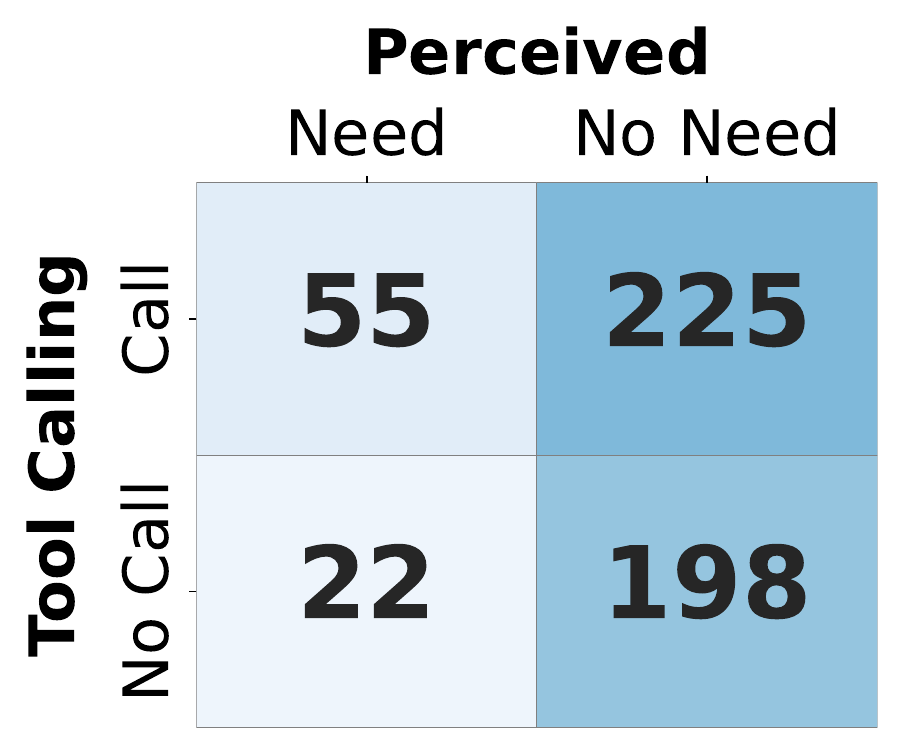}
    \caption{Qwen3-30B-A3B}
\end{subfigure}\hfill
\begin{subfigure}{0.25\linewidth}
    \centering
    \includegraphics[width=\linewidth]{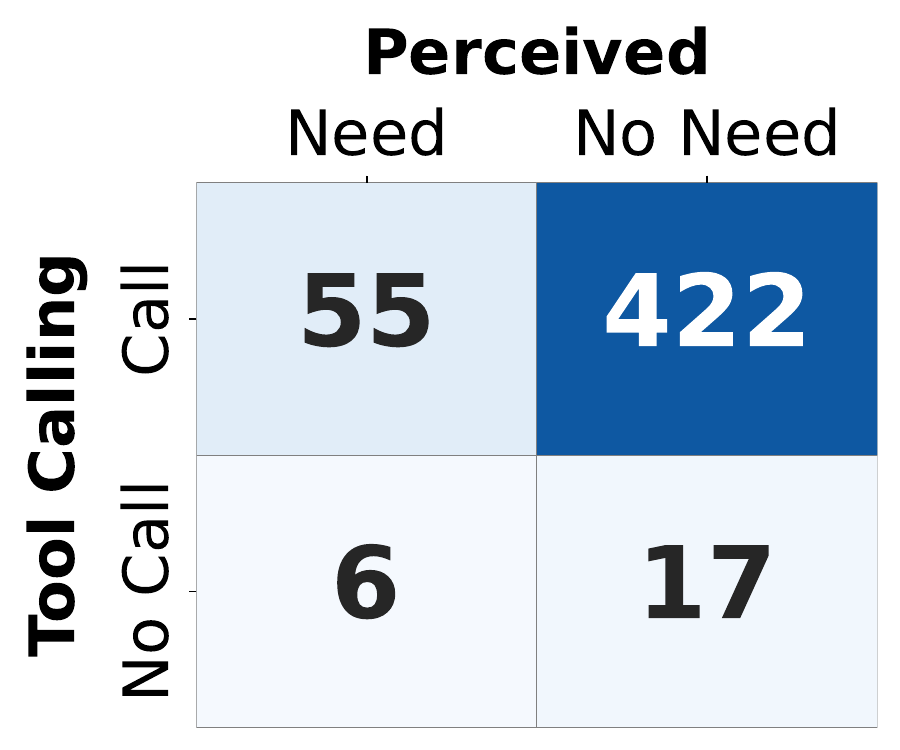}
    \caption{Qwen3-30B-A3B-IT}
\end{subfigure}\hfill
\begin{subfigure}{0.25\linewidth}
    \centering
    \includegraphics[width=\linewidth]{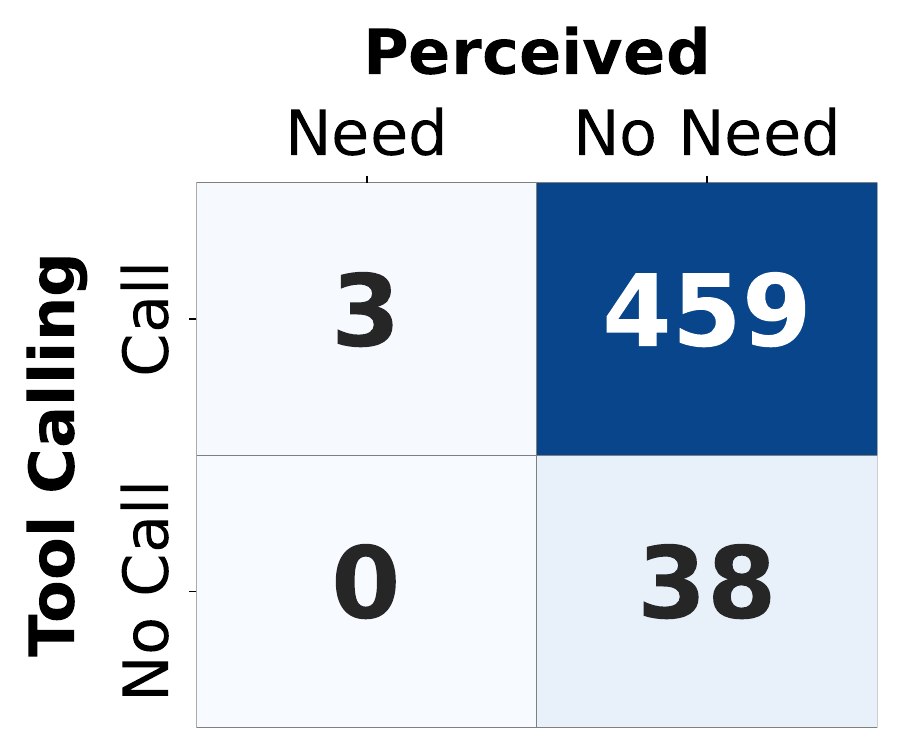}
    \caption{Gemma-3-27B-IT}
\end{subfigure}\hfill
\begin{subfigure}{0.25\linewidth}
    \centering
    \includegraphics[width=\linewidth]{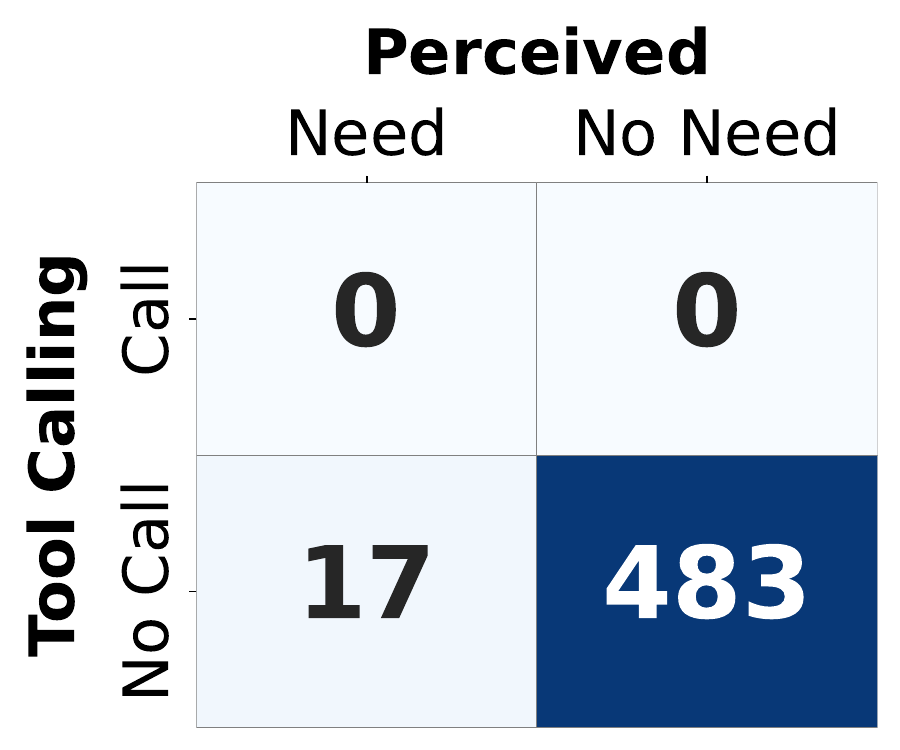}
    \caption{Mistral-3.1-24B-IT}
\end{subfigure}\hfill
\begin{subfigure}{0.25\linewidth}
    \centering
    \includegraphics[width=\linewidth]{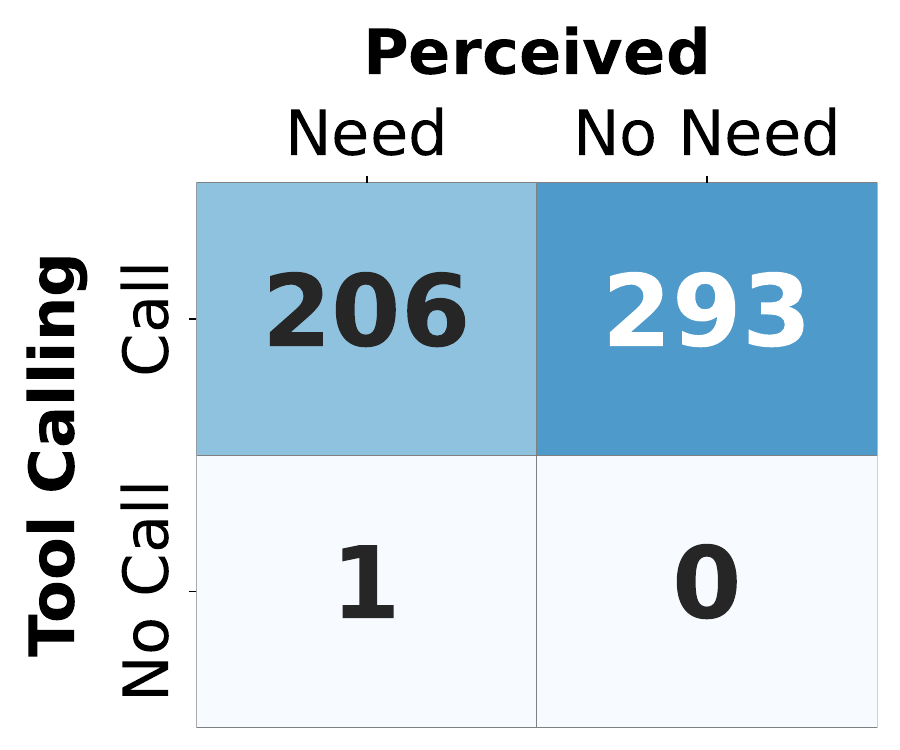}
    \caption{Llama-3.2-3B-IT}
\end{subfigure}

\vspace{6pt}
\textbf{(b) Perceived-need prompt v3}

\caption{\textbf{Entity Task: perceived need is only partially aligned with tool use.} The x-axis shows perceived utility (number of entities predicted to need or not need external information), and the y-axis shows actual tool-use decisions. Percentages indicate how often the model follows its own prediction (call vs.\ not call). Results are shown for two prompt variants (v2 and v3). Some responses are excluded due to parsing failures (i.e., missing explicit yes/no decisions), so the total count is less than 500.}

\label{fig:main_need_utility_combined_v12}
\vspace{-15pt}
\end{figure*}

Figure~\ref{fig:main_need_utility_combined_v12} shows two key observations. First, the model's self-perceived need for tool use is highly sensitive to the prompting format, where even small variations can lead to noticeably different outcomes. Second, across these variations, perceived need and utility (i.e., actual tool-calling decisions) are consistently related but not perfectly aligned. 

Overall, the main takeaway remains robust across different prompt formulations: while prompting influences the model's perception of need, it does not resolve the fundamental misalignment between perceived need and actual utility.

\subsubsection{Affordability}
\label{sec:affordability}

In this section, we present detailed results for both normative and descriptive affordability. To evaluate whether models prioritize the most beneficial entities under constrained tool usage, we employ the normalized discounted cumulative gain (NDCG) rank correlation metric. Specifically, NDCG measures the extent to which the model’s ranking of entities aligns with the optimal ranking based on maximum utility gain, thereby capturing how effectively the model selects high-utility entities when tool calls are limited.

Under the normative lens, a rational policy allocates a budget of $K$ calls to the instances with the largest positive marginal gains $\Delta^\star(x)$; the resulting oracle gain is $\mathrm{Gain}_K^\star$. Under the descriptive lens, we expose the model to the budget and current cost in its prompt, retain the first $K$ instances on which it chooses to call the tool, and measure their realized gain $\widehat{\mathrm{Gain}}_K$. We vary the permitted fraction of tool-using instances from $1\%$ to $100\%$. For the Entity dataset of 500 instances, for example, an $80\%$ budget permits at most 400 calls. We fix the total budget at $\mathcal{B}=\$10{,}000$ and assume a uniform cost per call, so a budget permitting $K$ calls corresponds to a per-call cost of $\mathcal{B}/K$. For each budget, we record both the utility gain over \notool{} and the model's uncapped number of tool calls. Prompt details appear in Appendix~\ref{sec:experimental_setup}.

\begin{figure}[t]
    \centering
    \includegraphics[width=\linewidth]{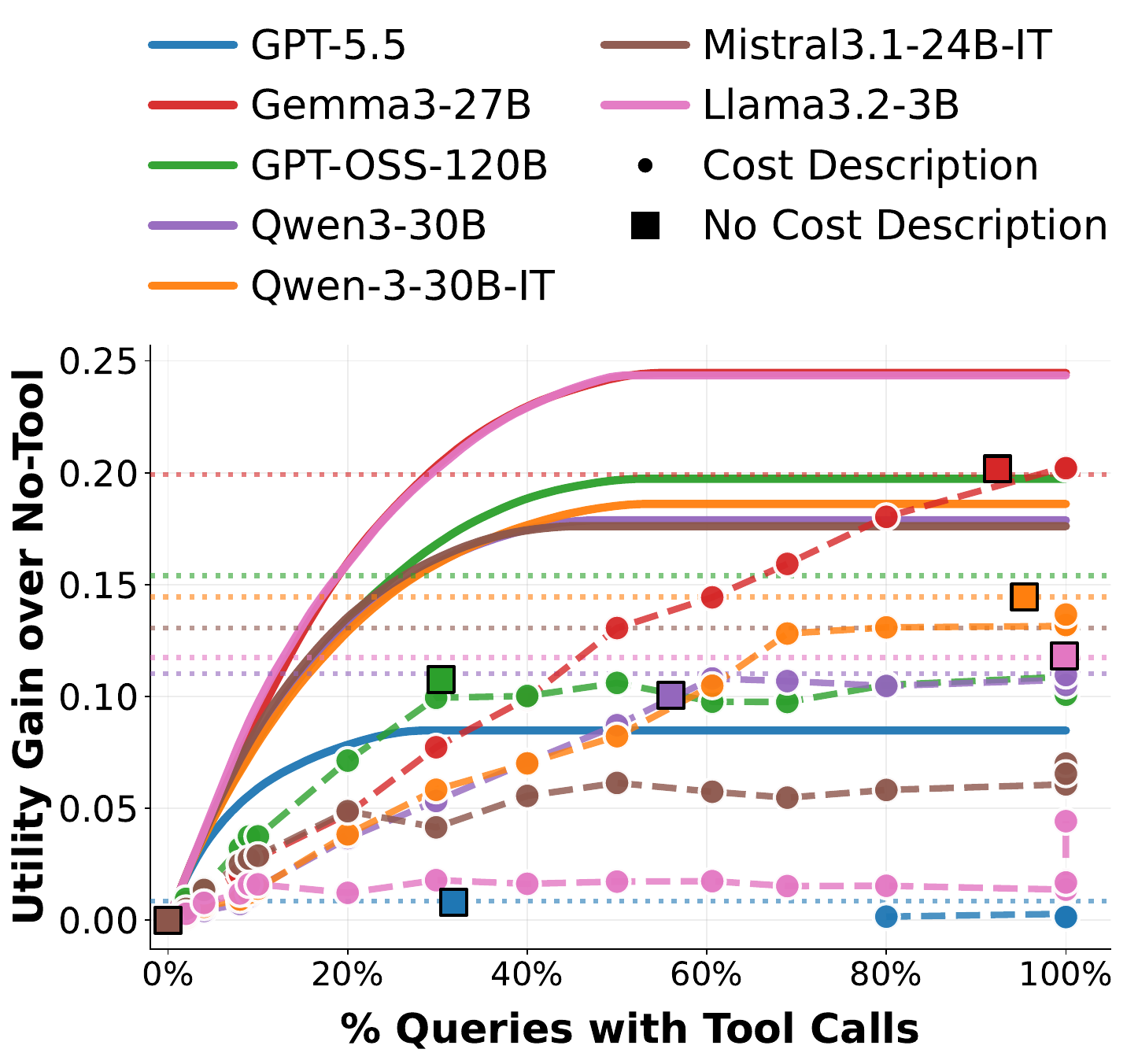}
    \caption{Utility gain over the \notool{} under varying cost constraints. Solid lines show optimal allocation (optimal top-$k$), dashed lines show model performance with cost information, squares denote no cost-awareness, and dotted lines indicate always-calling.}
    \label{fig:affordability_combined_v2}
\end{figure}

\paragraph{Normative--descriptive gap.}
All models exhibit a consistent gap between the ideal gain $\mathrm{Gain}_K^\star$ and their realized gain $\widehat{\mathrm{Gain}}_K$ (Figure~\ref{fig:affordability_combined_v2}). The oracle curve plateaus as the budget grows, reflecting diminishing returns, whereas descriptive gains vary substantially across models. Gemma and GPT-OSS achieve relatively high gains, while Llama and Mistral obtain weaker improvements despite making more calls, indicating inefficient allocation.

\paragraph{Uneven effects of cost information.}
Without a cost description, most models except Mistral and Llama obtain slightly higher utility gains than in the cost-aware condition. The difference is substantially larger for Mistral and Llama, suggesting that these models do not effectively incorporate stated costs into their decisions and may be distracted by this information.

\begin{figure}
    \centering
    \includegraphics[width=\linewidth]{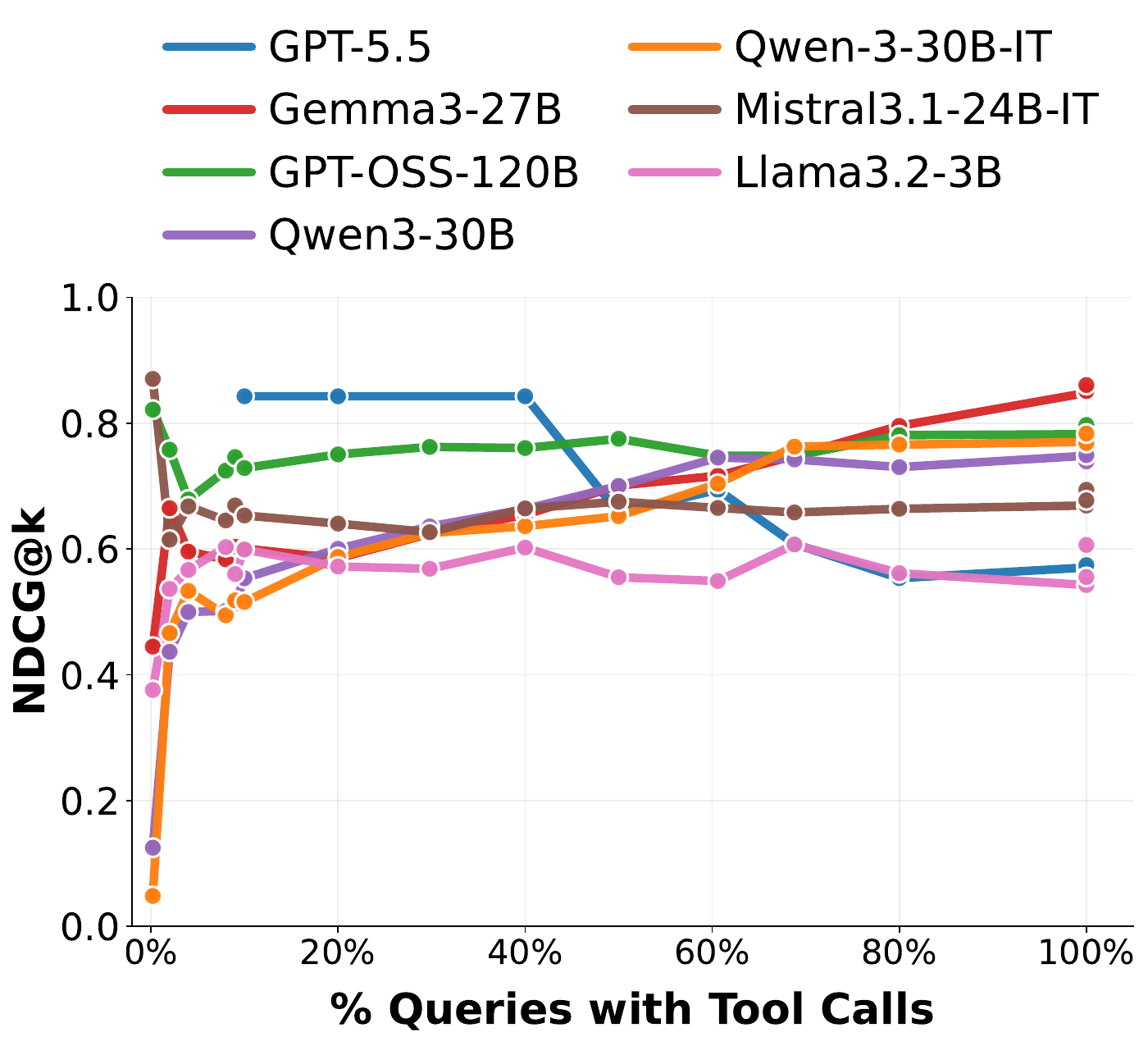}
    \caption{[Entity Task] The NDCG rank correlation under different budgets across different models. The correlation is low, which reflects that the models are not choosing the best utility gain tool calling. Cost prompt v1.}
    \label{fig:entity-ndcg-v1}
\end{figure}

\paragraph{Poor prioritization under budget constraints.}
Figure~\ref{fig:entity-ndcg-v1} compares the model's ordering of calls with the oracle utility ranking using NDCG~\cite{jarvelin2002cumulated}. Overall alignment is weak. Under the tightest budgets, GPT-OSS, Llama, and Mistral can sometimes identify the single highest-utility instance, but this behavior does not scale: NDCG falls sharply as more calls become available, revealing inconsistent prioritization across a sequence of decisions.

\begin{figure}[t]
\centering
    \includegraphics[width=\linewidth]{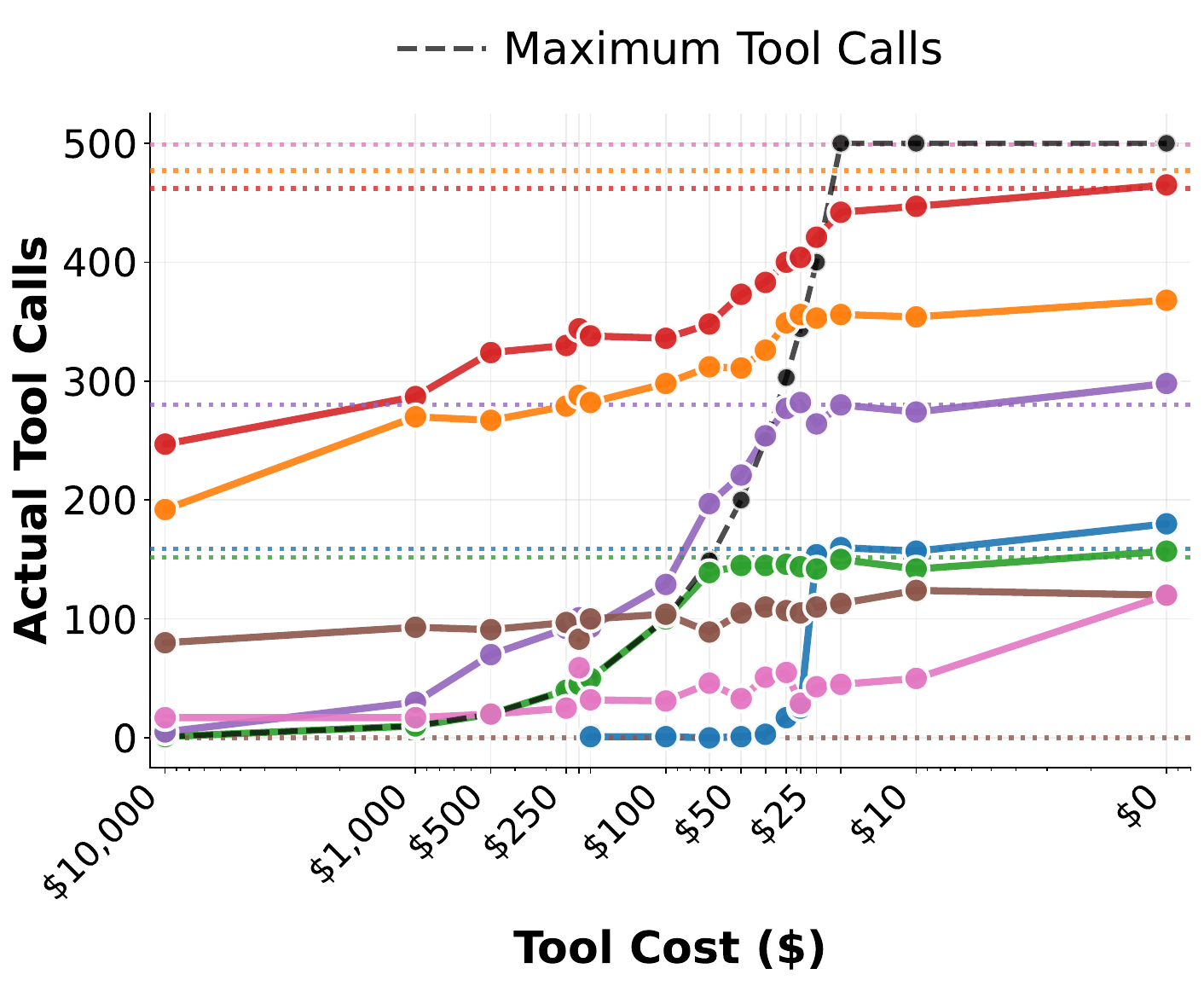}
    \caption{[Entity Task] Actual tool calls without budget enforcement. Models do not reliably reduce or stop calls as cost increases, despite being provided with cost and remaining budget.}
    \label{fig:actual_tool_call_v2}
\end{figure}

\begin{figure}[t]
\vspace{-6pt}
\centering

\begin{subfigure}{\linewidth}
    \centering
    \includegraphics[width=\linewidth]{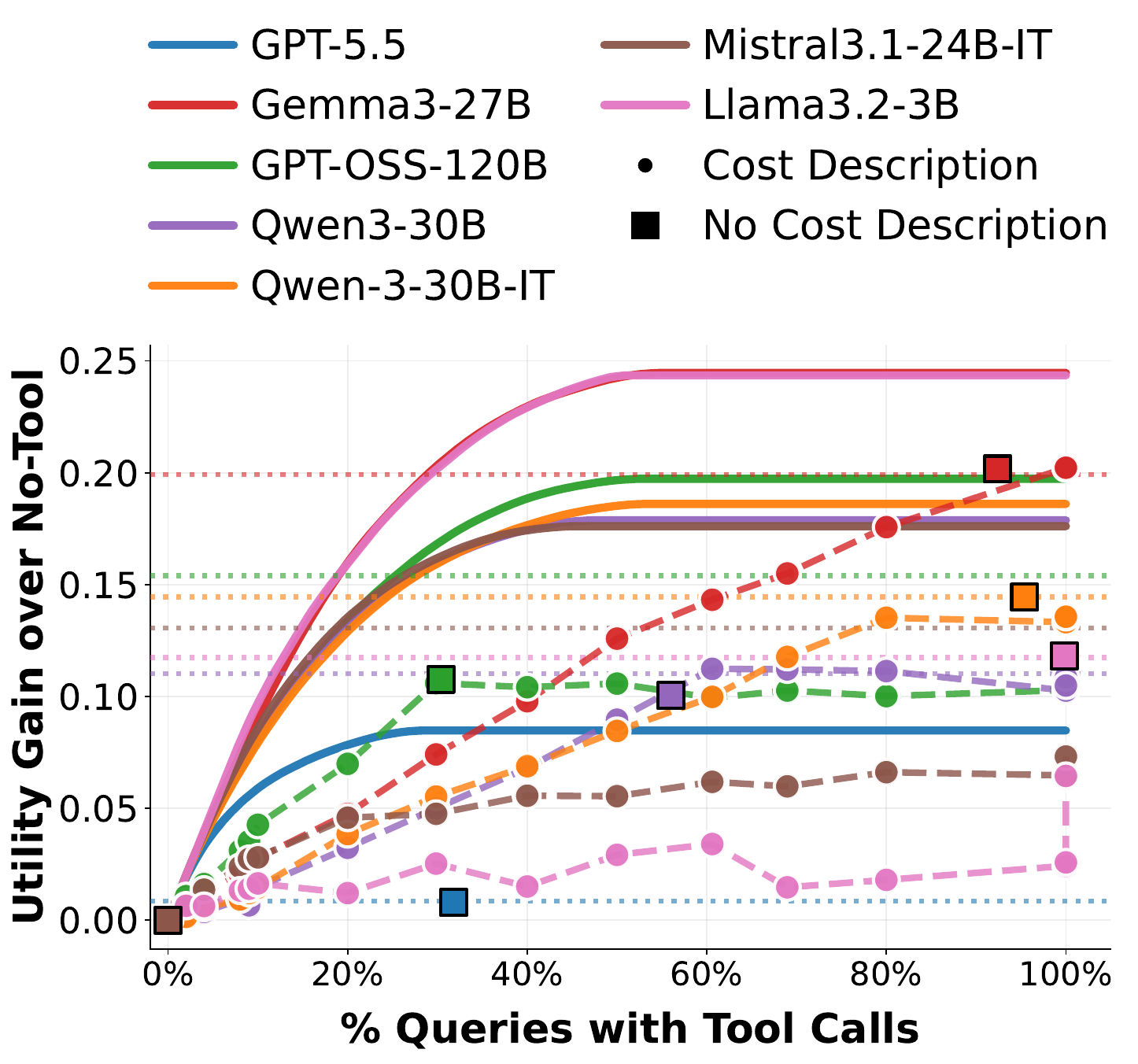}
    \caption{\textbf{Utility gain with hard stop after exceeding the budget.}}
\end{subfigure}\hfill
\begin{subfigure}{\linewidth}
    \centering
    \includegraphics[width=\linewidth]{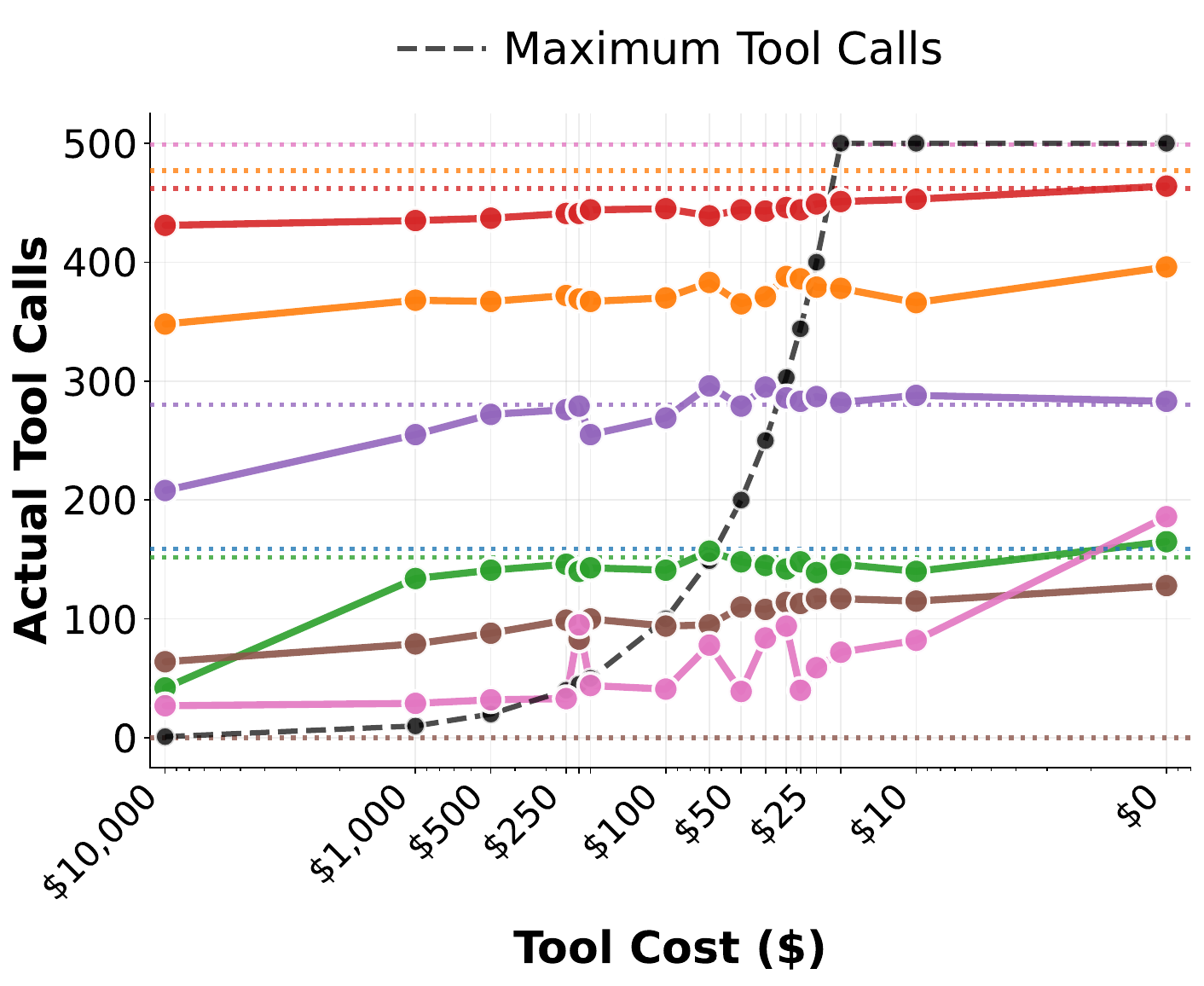}
    \caption{\textbf{Tool-calling behavior without hard stop.}}
\end{subfigure}
\caption{
\textbf{Cost-aware tool use on the Entity Task.}
\textbf{Left:} Utility gain over the no-tool baseline under varying cost constraints. Solid lines show oracle allocation (optimal top-$k$), dashed lines show model performance with cost information, squares denote no cost-awareness, and dotted lines indicate always-calling.
\textbf{Right:} Actual tool calls without budget enforcement. Models do not reliably reduce or stop calls as cost increases, despite being provided with cost and remaining budget.
}
\label{fig:affordability_combined}
\vspace{-8pt}
\end{figure}

\paragraph{Budget violations.}
Figure~\ref{fig:actual_tool_call_v2} compares actual calls with the permitted number under explicit budget prompting (Prompt~\ref{prompt:cost1}). Although most models respond directionally to higher per-call costs, they systematically exceed the budget. Qwen3-30B-A3B adheres most closely, followed by GPT-OSS-120B; Llama and Mistral show weak cost sensitivity, while Gemma and Qwen3-30B-IT frequently violate constraints even above $\$50$ per call. Under implicit cost prompting (Prompt~\ref{prompt:cost2}), all models fail to regulate usage (Figure~\ref{fig:affordability_combined}), indicating poor internal budget tracking. Explicit instructions help only partially (Figure~\ref{fig:affordability_combined_v2}). These four findings motivate the budget-capped LNE policy evaluated in Table~\ref{tab:all}.

\paragraph{NDCG Rank Correlation.}
To evaluate the \emph{quality} of the model's budget-aware tool-call selection,
we measure how well the instances chosen under \emph{perceived affordability}
align with the ideal selection under \emph{true affordability}.

For a given cost level $c$, let $\widehat{\mathcal{S}}$ be the set of instances
for which the model invokes the tool, and let $K = \lfloor B / c \rfloor$ be
the budget-permitted call limit (with $K = n$ when $c = 0$).
Following the same capping logic as the affordability evaluation, we construct
$\widehat{\mathcal{S}}_K$ by retaining only the first $K$ instances in
$\widehat{\mathcal{S}}$ (in the order they appear), discarding any calls that
exceed the budget.

\textbf{Relevance labels.}
The ground-truth relevance of invoking the tool on instance $x_i$ is the
marginal factuality gain $\Delta^\star(x_i) = s^{\textsc{AT}}(x_i) -
s^{\textsc{NT}}(x_i)$.
Since NDCG requires non-negative relevance values, we replace
$\Delta^\star(x_i)$ with its ordinal rank $r_i$ among all $n$ instances
(average rank for ties, ascending), yielding the relevance vector
$\mathbf{r} = (r_1, \ldots, r_n)$.

\textbf{NDCG@$K$.}
We define the budget-capped perceived affordability indicator as
$\widehat{\mathrm{A}}_K(x_i) = \mathbf{1}\{x_i \in \widehat{\mathcal{S}}_K\}
\in \{0,1\}$.
Instances are ranked by $\widehat{\mathrm{A}}_K$ descending (called instances
first), and NDCG@$K$ is computed as:
\begin{equation}
\begin{aligned}
    \mathrm{NDCG}@K
    &= \frac{\mathrm{DCG}@K\!\left(\mathbf{r},\,\widehat{\mathrm{A}}_K\right)}
            {\mathrm{IDCG}@K\!\left(\mathbf{r}\right)},\\
    \mathrm{DCG}@K
    &= \sum_{i=1}^{K}\frac{r_{\pi(i)}}{\log_2(i+1)}.
\end{aligned}
\end{equation}
where $\pi$ is the permutation induced by $\widehat{\mathrm{A}}_K$, and
$\mathrm{IDCG}@K$ is the DCG of the ideal ranking (instances sorted by $r_i$
descending).
A score of $1.0$ means $\widehat{\mathcal{S}}_K = \mathcal{S}_K^\star$,
i.e., the model selected exactly the $K$ instances with the highest marginal
gain $\Delta^\star$.

\textbf{Curve.}
Each cost level $c$ yields one point $\bigl(K/n \times 100\%,\;
\mathrm{NDCG}@K\bigr)$ on the NDCG curve, where the x-axis $K/n \times 100\%$
is the budget-determined coverage — identical to the x-axis of the
affordability plot — enabling direct comparison between the two.
This formulation decouples \emph{how often} the model invokes the tool
(coverage, x-axis) from \emph{how well it prioritises} the right instances
within that budget (NDCG@$K$, y-axis).

Figure~\ref{fig:actual_tool_call_v2} shows the model’s tool-calling frequency under different budgets with explicit budget-aware prompting. Figure~\ref{fig:entity-ndcg-v1} reports the corresponding NDCG-based rank correlation.

\begin{figure}
    \centering
    \includegraphics[width=\linewidth]{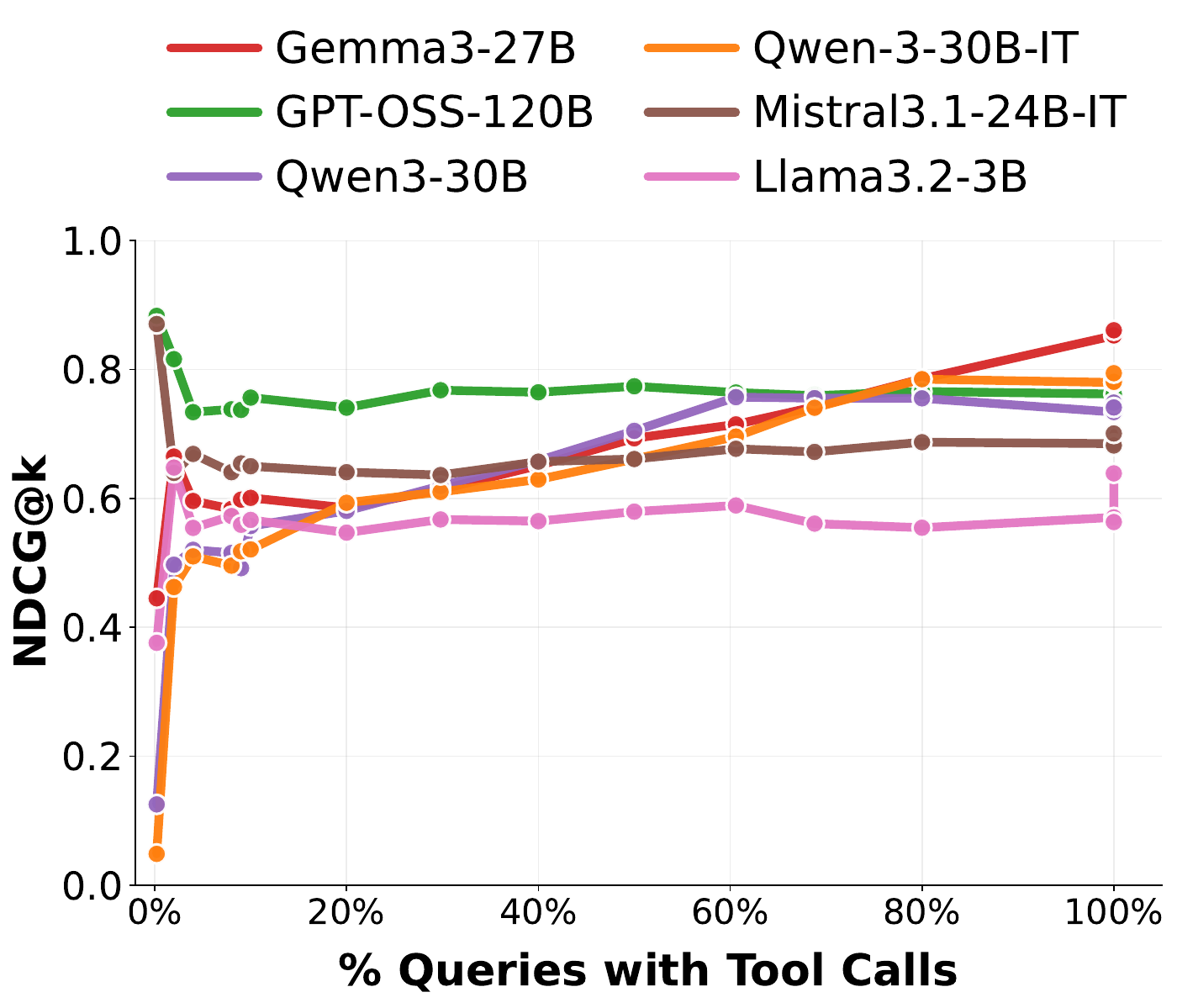}
    \caption{[Entity Task] The NDCG rank correlation under different budgets across different models. The correlation is low, which reflects that the models are not choosing the best utility gain tool calling. Cost prompt v2.}
    \label{fig:entity-ndcg-v2}
\end{figure}

We then evaluate implicit budget-aware prompting in Figures~\ref{fig:affordability_combined} and \ref{fig:entity-ndcg-v2}. We observe that implicit prompting fails to effectively constrain tool use.

\subsection{Controller Framework}
\label{sec:controller_details}

\subsubsection{Full LNE Results}

Table~\ref{tab:all} reports the complete comparison between model self-decision and LNE across models, tasks, natural tool use, and the 20\%, 40\%, and 80\% tool-call budgets.

\begin{table*}[t]
\centering
\scriptsize
\begin{adjustbox}{
  max width=0.99\textwidth,
  max totalheight=0.84\textheight,
  keepaspectratio
}
\begin{tabular}{
  l
  >{\centering\arraybackslash}p{1.7cm}
  >{\centering\arraybackslash}p{1.8cm}
  cccccccc
}
\toprule
\textbf{Model}
& \textbf{Tool}
& \textbf{Task}
& \multicolumn{2}{c}{\textbf{Natural}}
& \multicolumn{2}{c}{\textbf{20\% budget}}
& \multicolumn{2}{c}{\textbf{40\% budget}}
& \multicolumn{2}{c}{\textbf{80\% budget}} \\
\cmidrule(lr){4-5}
\cmidrule(lr){6-7}
\cmidrule(lr){8-9}
\cmidrule(lr){10-11}
&&
& \textbf{Self} & \textbf{LNE}
& \textbf{Self} & \textbf{LNE}
& \textbf{Self} & \textbf{LNE}
& \textbf{Self} & \textbf{LNE} \\
\midrule

\rowcolor{blue!8}

& 
& Entity
& 0.72{\color{gray}\textsubscript{30}} & \textbf{0.76}{\color{gray}\textsubscript{72}} & \textbf{0.68}{\color{gray}\textsubscript{20}} & 0.66{\color{gray}\textsubscript{20}} & \textbf{0.72}{\color{gray}\textsubscript{30}} & \textbf{0.72}{\color{gray}\textsubscript{40}} & 0.72{\color{gray}\textsubscript{30}} & \textbf{0.76}{\color{gray}\textsubscript{72}} \\
\rowcolor{blue!8}

& 
& InVivoQuery
& \textbf{0.57}{\color{gray}\textsubscript{38}} & 0.50{\color{gray}\textsubscript{68}} & \textbf{0.54}{\color{gray}\textsubscript{20}} & 0.53{\color{gray}\textsubscript{20}} & \textbf{0.57}{\color{gray}\textsubscript{38}} & 0.53{\color{gray}\textsubscript{40}} & \textbf{0.57}{\color{gray}\textsubscript{38}} & 0.50{\color{gray}\textsubscript{68}} \\
\rowcolor{blue!8}

& \multirow{-3}{*}{Web Search}
& BFCL
& \textbf{0.76}{\color{gray}\textsubscript{100}} & \textbf{0.76}{\color{gray}\textsubscript{97}} & 0.44{\color{gray}\textsubscript{20}} & \textbf{0.46}{\color{gray}\textsubscript{20}} & 0.52{\color{gray}\textsubscript{40}} & \textbf{0.57}{\color{gray}\textsubscript{40}} & 0.68{\color{gray}\textsubscript{80}} & \textbf{0.73}{\color{gray}\textsubscript{80}} \\
\cline{2-11}
\rowcolor{blue!8}

& 
& GSM-Hard
& 0.65{\color{gray}\textsubscript{31}} & \textbf{0.67}{\color{gray}\textsubscript{27}} & 0.65{\color{gray}\textsubscript{20}} & \textbf{0.66}{\color{gray}\textsubscript{20}} & 0.65{\color{gray}\textsubscript{31}} & \textbf{0.67}{\color{gray}\textsubscript{27}} & 0.65{\color{gray}\textsubscript{31}} & \textbf{0.67}{\color{gray}\textsubscript{27}} \\
\rowcolor{blue!8}

& 
& Multiplication
& \textbf{1.00}{\color{gray}\textsubscript{61}} & 0.91{\color{gray}\textsubscript{4}} & \textbf{0.93}{\color{gray}\textsubscript{20}} & 0.91{\color{gray}\textsubscript{4}} & \textbf{0.97}{\color{gray}\textsubscript{40}} & 0.91{\color{gray}\textsubscript{4}} & \textbf{1.00}{\color{gray}\textsubscript{61}} & 0.91{\color{gray}\textsubscript{4}} \\
\rowcolor{blue!8}
\multirow{-6}{*}{GPT-OSS-120B}
& \multirow{-3}{*}{Calculator}
& \shortstack{Multiplication\\($N{\times}N$)}
& \textbf{0.99}{\color{gray}\textsubscript{97}} & 0.98{\color{gray}\textsubscript{93}} & 0.30{\color{gray}\textsubscript{20}} & \textbf{0.31}{\color{gray}\textsubscript{20}} & 0.48{\color{gray}\textsubscript{40}} & \textbf{0.51}{\color{gray}\textsubscript{40}} & 0.84{\color{gray}\textsubscript{80}} & \textbf{0.89}{\color{gray}\textsubscript{80}} \\
\midrule

\rowcolor{blue!8}

& 
& Entity
& 0.80{\color{gray}\textsubscript{56}} & \textbf{0.81}{\color{gray}\textsubscript{70}} & 0.74{\color{gray}\textsubscript{20}} & \textbf{0.75}{\color{gray}\textsubscript{20}} & 0.77{\color{gray}\textsubscript{40}} & \textbf{0.79}{\color{gray}\textsubscript{40}} & 0.80{\color{gray}\textsubscript{56}} & \textbf{0.81}{\color{gray}\textsubscript{70}} \\
\rowcolor{blue!8}

& 
& InVivoQuery
& \textbf{0.64}{\color{gray}\textsubscript{39}} & \textbf{0.64}{\color{gray}\textsubscript{86}} & \textbf{0.65}{\color{gray}\textsubscript{20}} & \textbf{0.65}{\color{gray}\textsubscript{20}} & \textbf{0.64}{\color{gray}\textsubscript{39}} & \textbf{0.64}{\color{gray}\textsubscript{40}} & \textbf{0.64}{\color{gray}\textsubscript{39}} & 0.63{\color{gray}\textsubscript{80}} \\
\rowcolor{blue!8}

& \multirow{-3}{*}{Web Search}
& BFCL
& \textbf{0.62}{\color{gray}\textsubscript{100}} & \textbf{0.62}{\color{gray}\textsubscript{100}} & 0.29{\color{gray}\textsubscript{20}} & \textbf{0.30}{\color{gray}\textsubscript{20}} & 0.36{\color{gray}\textsubscript{40}} & \textbf{0.39}{\color{gray}\textsubscript{40}} & 0.54{\color{gray}\textsubscript{80}} & \textbf{0.59}{\color{gray}\textsubscript{80}} \\
\cline{2-11}
\rowcolor{blue!8}

& 
& GSM-Hard
& 0.61{\color{gray}\textsubscript{63}} & \textbf{0.62}{\color{gray}\textsubscript{34}} & \textbf{0.62}{\color{gray}\textsubscript{20}} & \textbf{0.62}{\color{gray}\textsubscript{20}} & \textbf{0.62}{\color{gray}\textsubscript{40}} & \textbf{0.62}{\color{gray}\textsubscript{34}} & 0.61{\color{gray}\textsubscript{63}} & \textbf{0.62}{\color{gray}\textsubscript{34}} \\
\rowcolor{blue!8}

& 
& Multiplication
& \textbf{1.00}{\color{gray}\textsubscript{97}} & 0.96{\color{gray}\textsubscript{59}} & 0.54{\color{gray}\textsubscript{20}} & \textbf{0.62}{\color{gray}\textsubscript{20}} & 0.66{\color{gray}\textsubscript{40}} & \textbf{0.82}{\color{gray}\textsubscript{40}} & 0.90{\color{gray}\textsubscript{80}} & \textbf{0.96}{\color{gray}\textsubscript{59}} \\
\rowcolor{blue!8}
\multirow{-6}{*}{Qwen3-30B-A3B}
& \multirow{-3}{*}{Calculator}
& \shortstack{Multiplication\\($N{\times}N$)}
& \textbf{1.00}{\color{gray}\textsubscript{100}} & \textbf{1.00}{\color{gray}\textsubscript{100}} & \textbf{0.21}{\color{gray}\textsubscript{20}} & \textbf{0.21}{\color{gray}\textsubscript{20}} & \textbf{0.40}{\color{gray}\textsubscript{40}} & \textbf{0.40}{\color{gray}\textsubscript{40}} & \textbf{0.80}{\color{gray}\textsubscript{80}} & \textbf{0.80}{\color{gray}\textsubscript{80}} \\
\midrule

\rowcolor{blue!8}

& 
& Entity
& \textbf{0.82}{\color{gray}\textsubscript{95}} & 0.81{\color{gray}\textsubscript{68}} & 0.71{\color{gray}\textsubscript{20}} & \textbf{0.73}{\color{gray}\textsubscript{20}} & 0.74{\color{gray}\textsubscript{40}} & \textbf{0.78}{\color{gray}\textsubscript{40}} & 0.80{\color{gray}\textsubscript{80}} & \textbf{0.81}{\color{gray}\textsubscript{68}} \\
\rowcolor{blue!8}

& 
& InVivoQuery
& \textbf{0.61}{\color{gray}\textsubscript{67}} & 0.60{\color{gray}\textsubscript{88}} & \textbf{0.56}{\color{gray}\textsubscript{20}} & \textbf{0.56}{\color{gray}\textsubscript{20}} & 0.58{\color{gray}\textsubscript{40}} & \textbf{0.59}{\color{gray}\textsubscript{40}} & \textbf{0.61}{\color{gray}\textsubscript{67}} & 0.59{\color{gray}\textsubscript{80}} \\
\rowcolor{blue!8}

& \multirow{-3}{*}{Web Search}
& BFCL
& \textbf{0.74}{\color{gray}\textsubscript{100}} & 0.73{\color{gray}\textsubscript{96}} & 0.40{\color{gray}\textsubscript{20}} & \textbf{0.43}{\color{gray}\textsubscript{20}} & 0.48{\color{gray}\textsubscript{40}} & \textbf{0.54}{\color{gray}\textsubscript{40}} & 0.63{\color{gray}\textsubscript{80}} & \textbf{0.71}{\color{gray}\textsubscript{80}} \\
\cline{2-11}
\rowcolor{blue!8}

& 
& GSM-Hard
& 0.59{\color{gray}\textsubscript{56}} & \textbf{0.60}{\color{gray}\textsubscript{38}} & \textbf{0.60}{\color{gray}\textsubscript{20}} & \textbf{0.60}{\color{gray}\textsubscript{20}} & \textbf{0.60}{\color{gray}\textsubscript{40}} & \textbf{0.60}{\color{gray}\textsubscript{38}} & 0.59{\color{gray}\textsubscript{56}} & \textbf{0.60}{\color{gray}\textsubscript{38}} \\
\rowcolor{blue!8}

& 
& Multiplication
& \textbf{1.00}{\color{gray}\textsubscript{100}} & 0.90{\color{gray}\textsubscript{19}} & 0.82{\color{gray}\textsubscript{20}} & \textbf{0.90}{\color{gray}\textsubscript{19}} & 0.86{\color{gray}\textsubscript{40}} & \textbf{0.90}{\color{gray}\textsubscript{19}} & \textbf{0.96}{\color{gray}\textsubscript{80}} & 0.90{\color{gray}\textsubscript{19}} \\
\rowcolor{blue!8}
\multirow{-6}{*}{Qwen-3-30B-IT}
& \multirow{-3}{*}{Calculator}
& \shortstack{Multiplication\\($N{\times}N$)}
& \textbf{0.95}{\color{gray}\textsubscript{100}} & \textbf{0.95}{\color{gray}\textsubscript{100}} & \textbf{0.20}{\color{gray}\textsubscript{20}} & 0.19{\color{gray}\textsubscript{20}} & \textbf{0.38}{\color{gray}\textsubscript{40}} & \textbf{0.38}{\color{gray}\textsubscript{40}} & \textbf{0.76}{\color{gray}\textsubscript{80}} & 0.75{\color{gray}\textsubscript{80}} \\
\midrule

\rowcolor{blue!8}

& 
& Entity
& 0.70{\color{gray}\textsubscript{0}} & \textbf{0.79}{\color{gray}\textsubscript{73}} & 0.70{\color{gray}\textsubscript{0}} & \textbf{0.73}{\color{gray}\textsubscript{20}} & 0.70{\color{gray}\textsubscript{0}} & \textbf{0.75}{\color{gray}\textsubscript{40}} & 0.70{\color{gray}\textsubscript{0}} & \textbf{0.79}{\color{gray}\textsubscript{73}} \\
\rowcolor{blue!8}

& 
& InVivoQuery
& 0.56{\color{gray}\textsubscript{44}} & \textbf{0.60}{\color{gray}\textsubscript{14}} & 0.55{\color{gray}\textsubscript{20}} & \textbf{0.60}{\color{gray}\textsubscript{14}} & 0.56{\color{gray}\textsubscript{40}} & \textbf{0.60}{\color{gray}\textsubscript{14}} & 0.56{\color{gray}\textsubscript{44}} & \textbf{0.60}{\color{gray}\textsubscript{14}} \\
\rowcolor{blue!8}

& \multirow{-3}{*}{Web Search}
& BFCL
& \textbf{0.69}{\color{gray}\textsubscript{81}} & 0.51{\color{gray}\textsubscript{47}} & \textbf{0.40}{\color{gray}\textsubscript{20}} & \textbf{0.40}{\color{gray}\textsubscript{20}} & \textbf{0.48}{\color{gray}\textsubscript{40}} & 0.47{\color{gray}\textsubscript{40}} & \textbf{0.69}{\color{gray}\textsubscript{80}} & 0.51{\color{gray}\textsubscript{47}} \\
\cline{2-11}
\rowcolor{blue!8}

& 
& GSM-Hard
& \textbf{0.51}{\color{gray}\textsubscript{5}} & 0.43{\color{gray}\textsubscript{43}} & \textbf{0.51}{\color{gray}\textsubscript{5}} & 0.48{\color{gray}\textsubscript{20}} & \textbf{0.51}{\color{gray}\textsubscript{5}} & 0.43{\color{gray}\textsubscript{40}} & \textbf{0.51}{\color{gray}\textsubscript{5}} & 0.43{\color{gray}\textsubscript{43}} \\
\rowcolor{blue!8}

& 
& Multiplication
& 0.29{\color{gray}\textsubscript{11}} & \textbf{0.39}{\color{gray}\textsubscript{71}} & 0.29{\color{gray}\textsubscript{11}} & \textbf{0.34}{\color{gray}\textsubscript{20}} & 0.29{\color{gray}\textsubscript{11}} & \textbf{0.37}{\color{gray}\textsubscript{40}} & 0.29{\color{gray}\textsubscript{11}} & \textbf{0.39}{\color{gray}\textsubscript{71}} \\
\rowcolor{blue!8}
\multirow{-6}{*}{Mistral3.1-24B-IT}
& \multirow{-3}{*}{Calculator}
& \shortstack{Multiplication\\($N{\times}N$)}
& 0.35{\color{gray}\textsubscript{56}} & -- & 0.12{\color{gray}\textsubscript{20}} & -- & 0.25{\color{gray}\textsubscript{40}} & -- & 0.35{\color{gray}\textsubscript{56}} & -- \\
\midrule

\rowcolor{blue!8}

& 
& Entity
& \textbf{0.70}{\color{gray}\textsubscript{100}} & 0.69{\color{gray}\textsubscript{89}} & 0.61{\color{gray}\textsubscript{20}} & \textbf{0.62}{\color{gray}\textsubscript{20}} & 0.64{\color{gray}\textsubscript{40}} & \textbf{0.67}{\color{gray}\textsubscript{40}} & 0.66{\color{gray}\textsubscript{80}} & \textbf{0.69}{\color{gray}\textsubscript{80}} \\
\rowcolor{blue!8}

& 
& InVivoQuery
& \textbf{0.55}{\color{gray}\textsubscript{64}} & 0.53{\color{gray}\textsubscript{71}} & \textbf{0.55}{\color{gray}\textsubscript{20}} & 0.54{\color{gray}\textsubscript{20}} & \textbf{0.55}{\color{gray}\textsubscript{40}} & 0.54{\color{gray}\textsubscript{40}} & \textbf{0.55}{\color{gray}\textsubscript{64}} & 0.53{\color{gray}\textsubscript{71}} \\
\rowcolor{blue!8}

& \multirow{-3}{*}{Web Search}
& BFCL
& \textbf{0.65}{\color{gray}\textsubscript{100}} & 0.63{\color{gray}\textsubscript{95}} & 0.27{\color{gray}\textsubscript{20}} & \textbf{0.31}{\color{gray}\textsubscript{20}} & 0.36{\color{gray}\textsubscript{40}} & \textbf{0.40}{\color{gray}\textsubscript{40}} & 0.56{\color{gray}\textsubscript{80}} & \textbf{0.59}{\color{gray}\textsubscript{80}} \\
\cline{2-11}
\rowcolor{blue!8}

& 
& GSM-Hard
& 0.02{\color{gray}\textsubscript{97}} & \textbf{0.10}{\color{gray}\textsubscript{86}} & 0.14{\color{gray}\textsubscript{20}} & \textbf{0.17}{\color{gray}\textsubscript{20}} & 0.12{\color{gray}\textsubscript{40}} & \textbf{0.16}{\color{gray}\textsubscript{40}} & 0.04{\color{gray}\textsubscript{80}} & \textbf{0.11}{\color{gray}\textsubscript{80}} \\
\rowcolor{blue!8}

& 
& Multiplication
& 0.00{\color{gray}\textsubscript{100}} & \textbf{0.13}{\color{gray}\textsubscript{85}} & 0.13{\color{gray}\textsubscript{20}} & \textbf{0.17}{\color{gray}\textsubscript{20}} & 0.11{\color{gray}\textsubscript{40}} & \textbf{0.16}{\color{gray}\textsubscript{40}} & 0.04{\color{gray}\textsubscript{80}} & \textbf{0.14}{\color{gray}\textsubscript{80}} \\
\rowcolor{blue!8}
\multirow{-6}{*}{Llama3.2-3B-IT}
& \multirow{-3}{*}{Calculator}
& \shortstack{Multiplication\\($N{\times}N$)}
& 0.00{\color{gray}\textsubscript{100}} & -- & 0.00{\color{gray}\textsubscript{20}} & -- & 0.00{\color{gray}\textsubscript{40}} & -- & 0.00{\color{gray}\textsubscript{80}} & -- \\
\midrule

\rowcolor{orange!12}

& 
& Entity
& \textbf{0.80}{\color{gray}\textsubscript{92}} & 0.79{\color{gray}\textsubscript{89}} & 0.63{\color{gray}\textsubscript{20}} & \textbf{0.66}{\color{gray}\textsubscript{20}} & 0.67{\color{gray}\textsubscript{40}} & \textbf{0.71}{\color{gray}\textsubscript{40}} & \textbf{0.77}{\color{gray}\textsubscript{80}} & \textbf{0.77}{\color{gray}\textsubscript{80}} \\
\rowcolor{orange!12}

& 
& InVivoQuery
& 0.62{\color{gray}\textsubscript{58}} & \textbf{0.63}{\color{gray}\textsubscript{99}} & \textbf{0.56}{\color{gray}\textsubscript{20}} & \textbf{0.56}{\color{gray}\textsubscript{20}} & \textbf{0.60}{\color{gray}\textsubscript{40}} & 0.58{\color{gray}\textsubscript{40}} & \textbf{0.62}{\color{gray}\textsubscript{58}} & 0.61{\color{gray}\textsubscript{79}} \\
\rowcolor{orange!12}

& \multirow{-3}{*}{Web Search}
& BFCL
& \textbf{0.69}{\color{gray}\textsubscript{100}} & \textbf{0.69}{\color{gray}\textsubscript{100}} & \textbf{0.47}{\color{gray}\textsubscript{20}} & \textbf{0.47}{\color{gray}\textsubscript{20}} & \textbf{0.53}{\color{gray}\textsubscript{40}} & 0.52{\color{gray}\textsubscript{40}} & \textbf{0.64}{\color{gray}\textsubscript{80}} & \textbf{0.64}{\color{gray}\textsubscript{80}} \\
\cline{2-11}
\rowcolor{orange!12}

& 
& GSM-Hard
& 0.56{\color{gray}\textsubscript{94}} & \textbf{0.59}{\color{gray}\textsubscript{29}} & 0.58{\color{gray}\textsubscript{20}} & \textbf{0.59}{\color{gray}\textsubscript{20}} & 0.58{\color{gray}\textsubscript{40}} & \textbf{0.59}{\color{gray}\textsubscript{29}} & 0.56{\color{gray}\textsubscript{80}} & \textbf{0.59}{\color{gray}\textsubscript{29}} \\
\rowcolor{orange!12}

& 
& Multiplication
& \textbf{1.00}{\color{gray}\textsubscript{100}} & 0.88{\color{gray}\textsubscript{55}} & 0.53{\color{gray}\textsubscript{20}} & \textbf{0.60}{\color{gray}\textsubscript{20}} & 0.65{\color{gray}\textsubscript{40}} & \textbf{0.78}{\color{gray}\textsubscript{40}} & \textbf{0.89}{\color{gray}\textsubscript{80}} & 0.88{\color{gray}\textsubscript{55}} \\
\rowcolor{orange!12}
\multirow{-6}{*}{Gemma3-27B-IT}
& \multirow{-3}{*}{Calculator}
& \shortstack{Multiplication\\($N{\times}N$)}
& \textbf{1.00}{\color{gray}\textsubscript{100}} & \textbf{1.00}{\color{gray}\textsubscript{100}} & 0.20{\color{gray}\textsubscript{20}} & \textbf{0.21}{\color{gray}\textsubscript{20}} & \textbf{0.40}{\color{gray}\textsubscript{40}} & \textbf{0.40}{\color{gray}\textsubscript{40}} & \textbf{0.80}{\color{gray}\textsubscript{80}} & \textbf{0.80}{\color{gray}\textsubscript{80}} \\

\bottomrule
\end{tabular}
\end{adjustbox}

\caption{
Task performance under self-decision and learned tool-use policies.
Each entry reports the task score; the gray subscript gives the actual
tool-call rate (\%). Self-decision and LNE are compared within the natural
setting and at each tool-call budget. The higher task score in each paired
comparison is bolded; ties are bolded for both policies. Results are grouped
by model, with Web Search and Calculator tasks nested within each group. The 20\%, 40\%, and 80\% columns
report performance at the corresponding maximum tool-call budgets; actual
call rates may be lower when a policy selects fewer positive instances. Row colors
indicate the harness:
\colorbox{blue!8}{\strut Trained} and
\colorbox{orange!12}{\strut Custom}.
A double dash indicates that the corresponding result is unavailable; LNE is omitted for Llama3.2-3B-IT and Mistral3.1-24B-IT on Multiplication ($N{\times}N$) because the true-need labels are degenerate for these models (Section~\ref{par:synthetic-nn-lne-omission}).
}
\label{tab:all}
\vspace{-10pt}
\end{table*}

\subsubsection{Latent Utility Estimators}

The latent utility estimator (LUE) predicts \textit{true utility}. Following Section~\ref{sec:framework}, we consider two variants: $\mathrm{LUE}_x$, which uses the input representation, and $\mathrm{LUE}_{x,d_{\mathcal{F}}}$, which additionally includes the tool description. Table~\ref{tab:web_lue} reports their natural-setting web-search results. Both variants can outperform the model's perceived utility for some models and tasks, indicating that hidden states encode signals of tool usefulness that are not consistently expressed in self-decisions. However, these improvements are uneven, and adding the tool description yields limited and inconsistent gains: $\mathrm{LUE}_{x,d_{\mathcal{F}}}$ is often comparable to or worse than $\mathrm{LUE}_x$.

\begin{figure}
    \centering
    \begin{subfigure}{\linewidth}
    \centering
    \includegraphics[width=\linewidth]{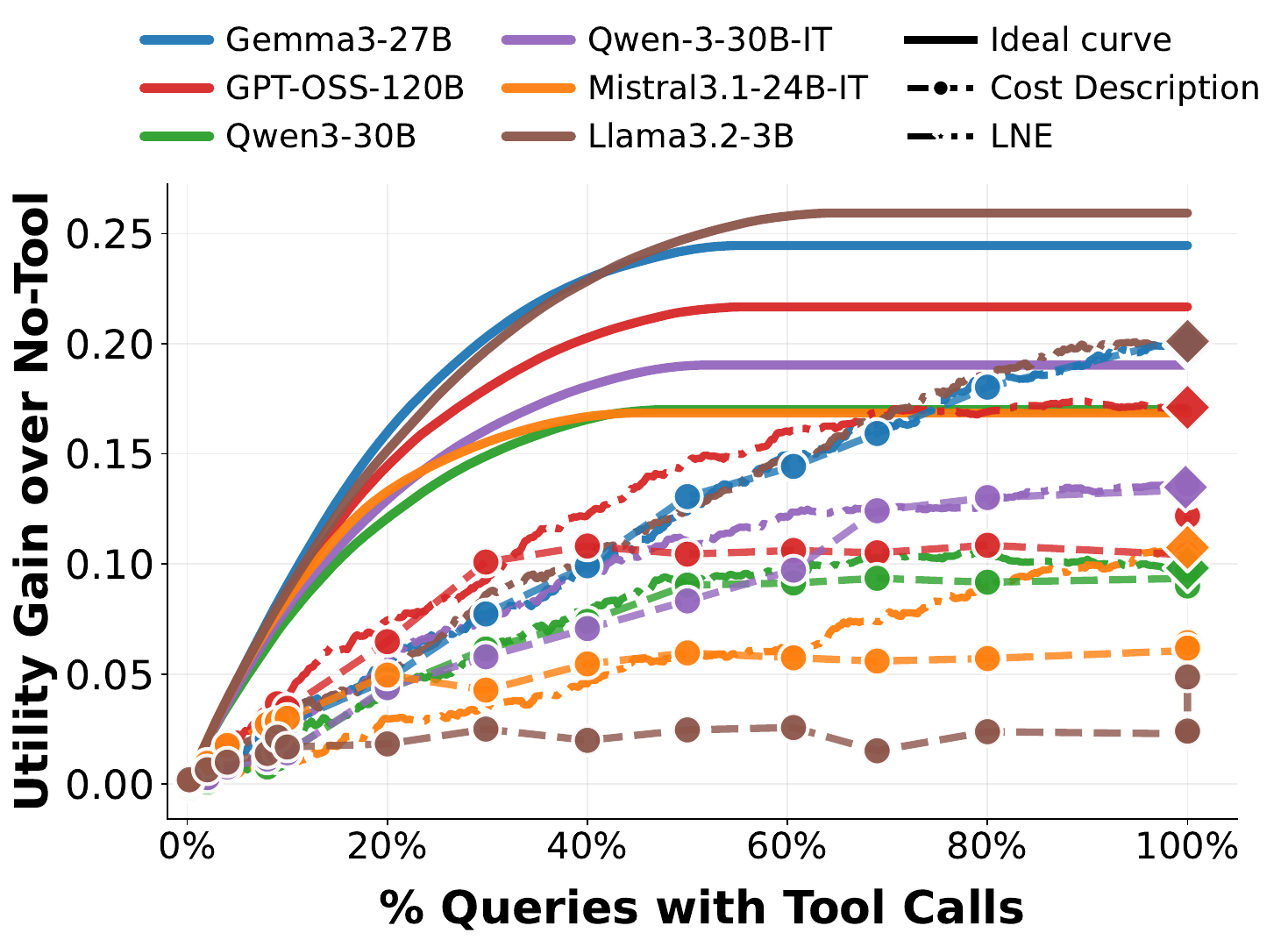}
    \caption{LNE}
    \end{subfigure}
    \begin{subfigure}{\linewidth}
    \centering
    \includegraphics[width=\linewidth]{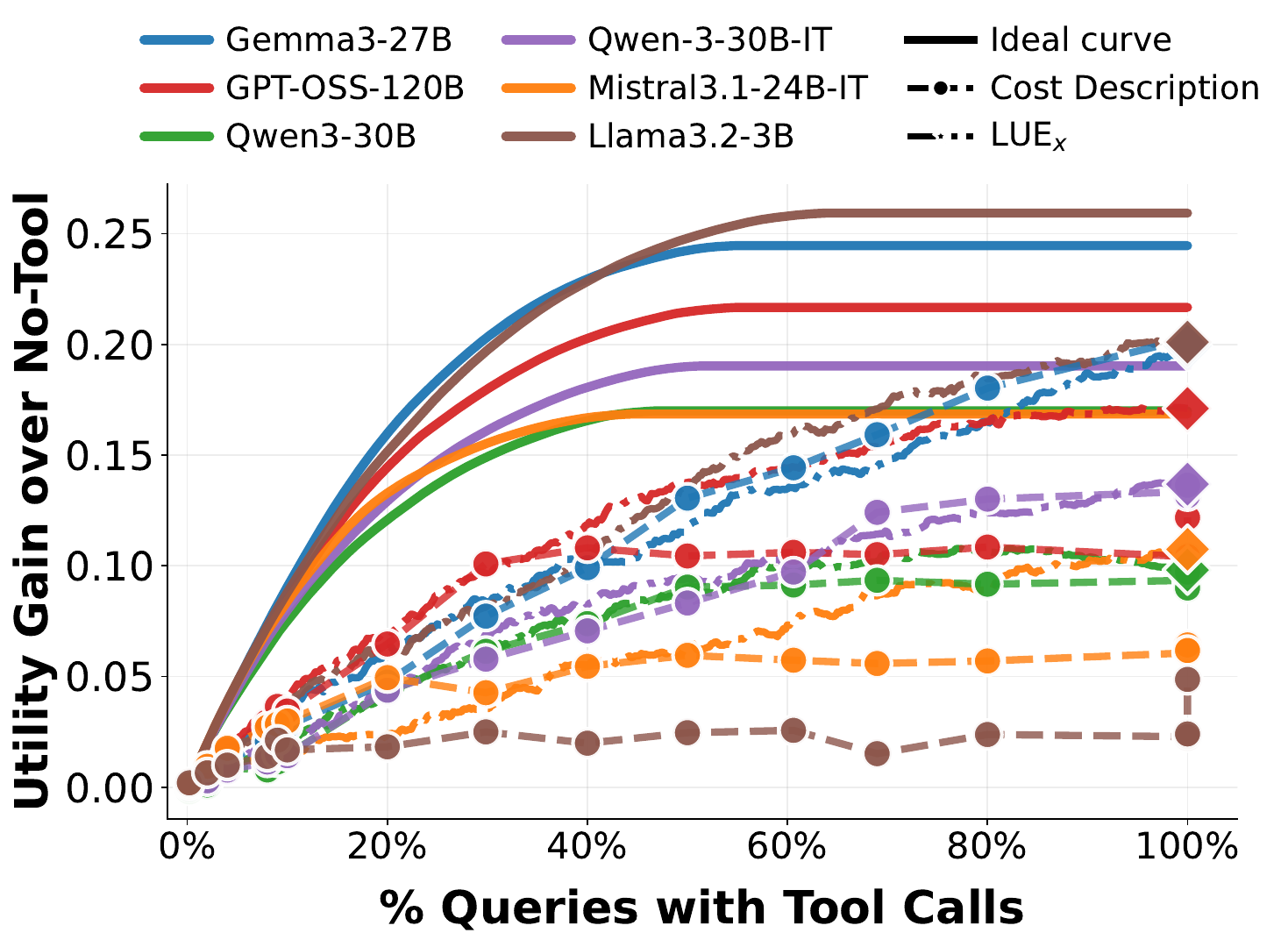}
    \caption{LUE$_x$}
    \end{subfigure}
    \begin{subfigure}{\linewidth}
    \centering
    \includegraphics[width=\linewidth]{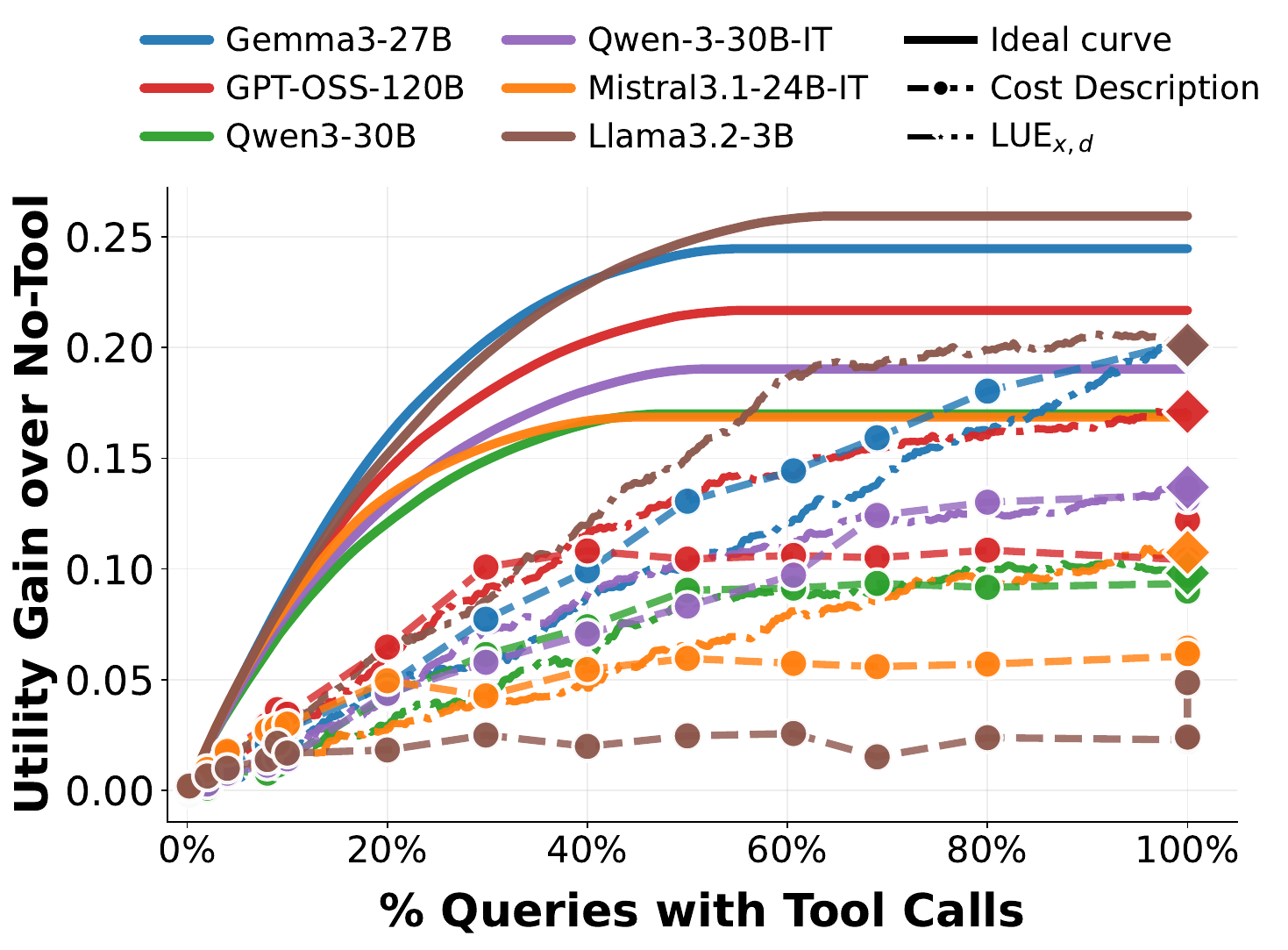}
    \caption{LUE$_{x,d_{\mathcal{F}}}$}
    \end{subfigure}
\caption{[Entity Task] Tool-call decisions are guided by the latent need estimator’s predicted probabilities under a fixed budget constraint.}
\label{fig:prediction_cost}
\end{figure}

Under a fixed tool-call budget, we use the estimators' predicted probabilities to rank positively classified instances and grant tool access to at most the top-$K$. If fewer than $K$ instances are classified as positive, the remaining budget is unused. Figures~\ref{fig:prediction_cost}, \ref{fig:prediction_cost_invivo}, and \ref{fig:prediction_cost_bfcl} report the resulting allocation performance. This ranking can improve utility over native self-decisions, but the LUE policies remain well below \optimized{} because utility depends on the unknown tool response and the model's ability to use it. Thus, unlike need estimation, utility estimation requires an implicit model of tool behavior; a static tool description does not provide enough information to reliably order instances by marginal gain.

\begin{table*}[t]
\centering
\scriptsize
\setlength{\tabcolsep}{4pt}
\begin{adjustbox}{max width=0.99\textwidth,max totalheight=0.9\textheight,keepaspectratio}
\begin{tabular}{llccccc}
\toprule
\textbf{Task} & \textbf{Model} & \textbf{No Tool} & \textbf{Always Tool} & \textbf{Optimal} & \textbf{LUE}$_x$ & \textbf{LUE}$_{x,d_{\mathcal F}}$ \\
\midrule
\rowcolor{blue!8}
\multirow{6}{*}{Entity} & GPT-OSS-120B & \scorecalls{0.61}{0} & \scorecalls{0.76}{100} & \scorecalls{0.81}{61} & \scorecalls{0.74}{58} & \scorecalls{0.73}{69} \\
\rowcolor{blue!8}
& Qwen3-30B-A3B & \scorecalls{0.70}{0} & \scorecalls{0.81}{100} & \scorecalls{0.88}{51} & \scorecalls{0.80}{54} & \scorecalls{0.81}{50} \\
\rowcolor{blue!8}
& Qwen3-30B-IT & \scorecalls{0.68}{0} & \scorecalls{0.82}{100} & \scorecalls{0.87}{60} & \scorecalls{0.79}{61} & \scorecalls{0.80}{63} \\
\rowcolor{blue!8}
& Mistral3.1-24B-IT & \scorecalls{0.70}{0} & \scorecalls{0.83}{100} & \scorecalls{0.88}{51} & \scorecalls{0.75}{45} & \scorecalls{0.78}{47} \\
\rowcolor{blue!8}
& Llama3.2-3B-IT & \scorecalls{0.58}{0} & \scorecalls{0.70}{100} & \scorecalls{0.83}{57} & \scorecalls{0.71}{79} & \scorecalls{0.70}{81} \\
\rowcolor{orange!12}
& Gemma3-27B-IT & \scorecalls{0.60}{0} & \scorecalls{0.80}{100} & \scorecalls{0.85}{59} & \scorecalls{0.75}{72} & \scorecalls{0.75}{69} \\
\midrule
\rowcolor{blue!8}
\multirow{6}{*}{InVivoQuery} & GPT-OSS-120B & \scorecalls{0.53}{0} & \scorecalls{0.45}{100} & \scorecalls{0.66}{36} & \scorecalls{0.46}{99} & \scorecalls{0.49}{62} \\
\rowcolor{blue!8}
& Qwen3-30B-A3B & \scorecalls{0.64}{0} & \scorecalls{0.64}{100} & \scorecalls{0.79}{40} & \scorecalls{0.64}{34} & \scorecalls{0.65}{33} \\
\rowcolor{blue!8}
& Qwen3-30B-IT & \scorecalls{0.54}{0} & \scorecalls{0.61}{100} & \scorecalls{0.72}{51} & \scorecalls{0.59}{55} & \scorecalls{0.58}{29} \\
\rowcolor{blue!8}
& Mistral3.1-24B-IT & \scorecalls{0.56}{0} & \scorecalls{0.60}{100} & \scorecalls{0.74}{46} & \scorecalls{0.59}{82} & \scorecalls{0.57}{8} \\
\rowcolor{blue!8}
& Llama3.2-3B-IT & \scorecalls{0.53}{0} & \scorecalls{0.54}{100} & \scorecalls{0.68}{45} & \scorecalls{0.55}{37} & \scorecalls{0.54}{95} \\
\rowcolor{orange!12}
& Gemma3-27B-IT & \scorecalls{0.54}{0} & \scorecalls{0.63}{100} & \scorecalls{0.74}{51} & \scorecalls{0.55}{8} & \scorecalls{0.56}{12} \\
\midrule
\rowcolor{blue!8}
\multirow{6}{*}{BFCL} & GPT-OSS-120B & \scorecalls{0.38}{0} & \scorecalls{0.76}{100} & \scorecalls{0.79}{41} & \scorecalls{0.76}{98} & \scorecalls{0.57}{54} \\
\rowcolor{blue!8}
& Qwen3-30B-A3B & \scorecalls{0.20}{0} & \scorecalls{0.62}{100} & \scorecalls{0.66}{46} & \scorecalls{0.35}{26} & \scorecalls{0.45}{49} \\
\rowcolor{blue!8}
& Qwen3-30B-IT & \scorecalls{0.31}{0} & \scorecalls{0.74}{100} & \scorecalls{0.77}{46} & \scorecalls{0.74}{94} & \scorecalls{0.34}{5} \\
\rowcolor{blue!8}
& Mistral3.1-24B-IT & \scorecalls{0.30}{0} & \scorecalls{0.69}{100} & \scorecalls{0.73}{44} & \scorecalls{0.45}{43} & \scorecalls{0.44}{28} \\
\rowcolor{blue!8}
& Llama3.2-3B-IT & \scorecalls{0.17}{0} & \scorecalls{0.65}{100} & \scorecalls{0.68}{51} & \scorecalls{0.63}{95} & \scorecalls{0.58}{89} \\
\rowcolor{orange!12}
& Gemma3-27B-IT & \scorecalls{0.44}{0} & \scorecalls{0.69}{100} & \scorecalls{0.76}{33} & \scorecalls{0.64}{80} & \scorecalls{0.52}{31} \\
\bottomrule
\end{tabular}
\end{adjustbox}
\caption{Web-search performance of the two latent utility estimators in the natural setting. Each cell reports task score, with tool-call rate (\%) in gray subscript. Row colors indicate the harness: \colorbox{blue!8}{\strut Trained} and \colorbox{orange!12}{\strut Custom}.}
\label{tab:web_lue}
\vspace{-10pt}
\end{table*}

\begin{figure}[t]
    \centering
    \vspace{-10pt}
    \includegraphics[width=\linewidth]{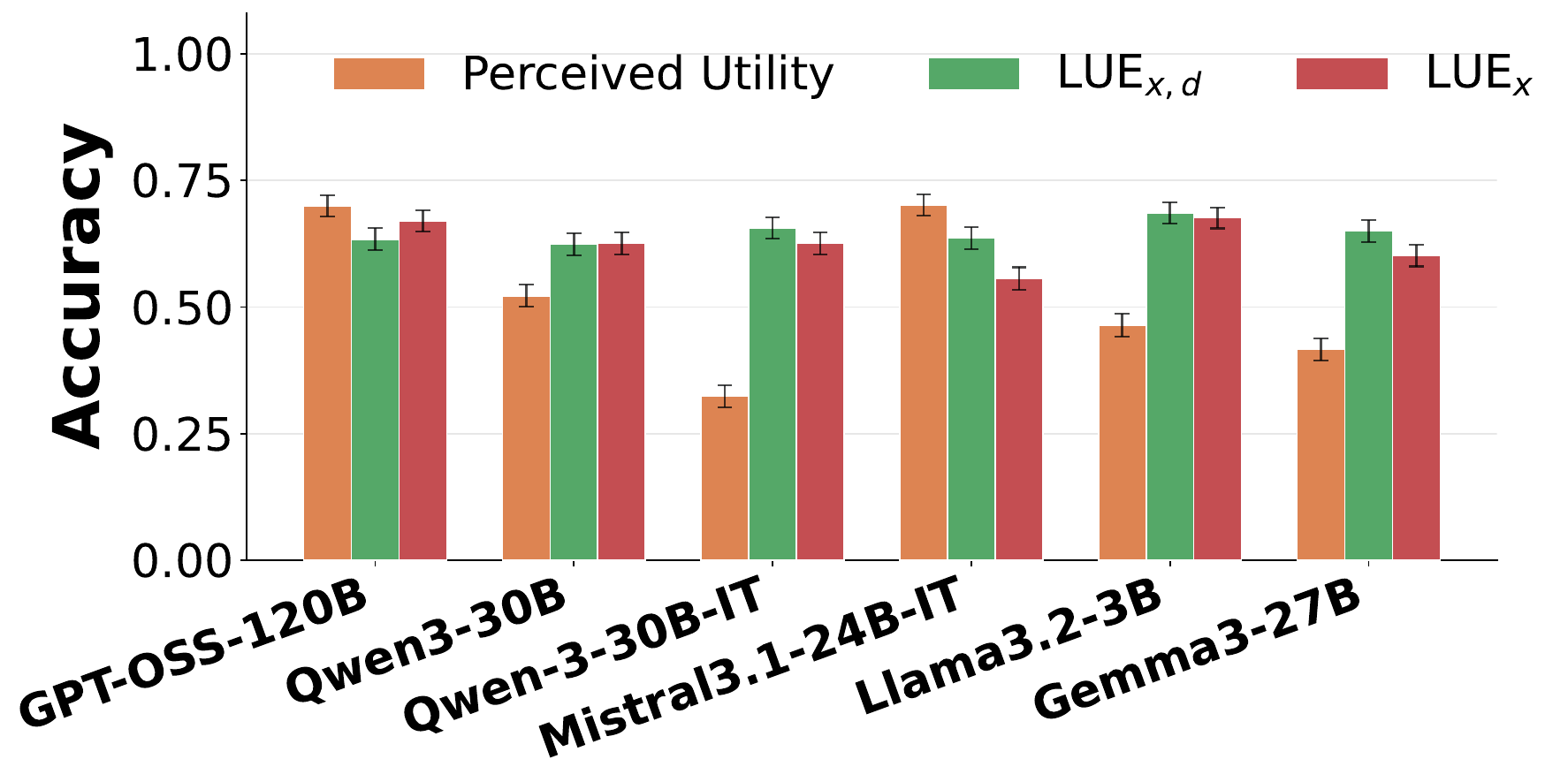}
    \caption{
       \textbf{The LUEs can predict the \textit{True Utility} more accurately across most models, especially for small and weaker models.}
       The same pattern is observed also in the \textit{InVivoQuery Task} in Figure~\ref{fig: invivo_lue} and the \textit{BFCL Task in Figure~\ref{fig: bfcl_lue}}.
    }
    \label{fig:lue}
    \vspace{-15pt}
\end{figure}

Tool-call decisions are guided by the latent need estimator’s predicted probabilities under a fixed budget constraint. We rank positively classified instances by their predicted need and enable tool calling for at most the top-$k$; when fewer than $k$ instances are classified as positive, the policy leaves the remaining budget unused. This strategy can improve performance under the same budget compared with self-decision. Figure~\ref{fig:prediction_cost} illustrates its behavior across budget levels.

\subsubsection{Comparison with the AdaptiveRAG Baseline}
\label{sec:baseline}

We add the adaptive-RAG baseline (Adaptive-RAG~\citep{jeong2024adaptive}) as follows.
The original method trains a T5-large model to classify query complexity into three
categories (zero-shot / single-step / multi-step retrieval) using silver labels derived
from multiple RAG systems on six public QA datasets. Since our task is binary, i.e.\
predicting whether search is needed for a given query, we retrain the same T5-large model
on our own labeled data, mapping \emph{no-search} to label A and \emph{needs-search} to
label B, using the identical training objective and hyperparameters as the original
(AdamW, $\text{lr}=3\times10^{-5}$, batch size $32$, $20$ epochs;
\footnote{https://github.com/starsuzi/Adaptive-RAG}). Labels are derived from our existing
annotation pipeline (hallucination score $>0.9 \rightarrow$ needs search). To ensure a
fair, apples-to-apples comparison with all other predictors, we evaluate using $5$-fold
stratified cross-validation on the full dataset---the same protocol used throughout our
experiments---and report out-of-fold predictions. The results for all models on the
Entity task are shown in Table~\ref{tab:entity-results}.

\begin{table*}[t]
\centering
\scriptsize
\begin{adjustbox}{max width=0.99\textwidth,max totalheight=0.9\textheight,keepaspectratio}
\begin{tabular}{llcccccccc}
\toprule
\textbf{Task} & \textbf{Model} & \textbf{\notool{}} & \textbf{\withtool{}} & \textbf{\optimized{}} & \textbf{\auto{}} & \textbf{LNE} & \textbf{LUE}$_{x}$ & \textbf{LUE}$_{x,d_{\mathcal{F}}}$ & \textbf{AdaptiveRAG} \\
\midrule
\rowcolor{blue!8}
\multirow{6}{*}{Entity Task}
 & GPT-OSS-120B      & 0.61 (0) & 0.76 (500) & 0.81 (300) & 0.72 (152) & \textbf{0.78 (351)} & 0.77 (341) & 0.75 (293)          & 0.71 (348) \\
\rowcolor{blue!8}
 & Qwen3-30B-A3B     & 0.70 (0) & 0.81 (500) & 0.88 (252) & 0.80 (342) & \textbf{0.81 (329)} & 0.79 (249) & 0.78 (248)          & 0.74 (297) \\
\rowcolor{blue!8}
 & Qwen-3-30B-IT     & 0.68 (0) & 0.82 (500) & 0.87 (284) & 0.82 (452) & 0.81 (339)          & 0.79 (306) & \textbf{0.81 (271)} & 0.76 (330) \\
\rowcolor{blue!8}
 & Mistral3.1-24B-IT & 0.70 (0) & 0.83 (500) & 0.88 (345) & 0.70 (172) & 0.78 (279)          & 0.77 (267) & \textbf{0.80 (263)} & 0.76 (357) \\
\rowcolor{blue!8}
 & Llama3.2-3B-IT    & 0.58 (0) & 0.70 (500) & 0.83 (249) & 0.70 (340) & \textbf{0.79 (429)} & 0.78 (383) & 0.78 (354)          & 0.67 (238) \\
\rowcolor{orange!12}
 & Gemma3-27B-IT     & 0.60 (0) & 0.80 (500) & 0.85 (297) & \textbf{0.80 (462)} & 0.75 (292) & 0.75 (284) & 0.78 (308)          & 0.68 (263) \\
\bottomrule
\end{tabular}
\end{adjustbox}
\caption{Results for all models on the Entity task. Each cell reports accuracy with the
number of tool calls in parentheses. Best learned predictor per row in \textbf{bold}. Row colors indicate the harness: \colorbox{blue!8}{\strut Trained} and \colorbox{orange!12}{\strut Custom}.}
\label{tab:entity-results}
\vspace{-10pt}
\end{table*}

\subsection{Results with Other Web Search Tools}
\label{sec:perplexity}

Table~\ref{tab:perplexity} shows the general performance of GPT-OSS-120B with the Perplexity Search~\footnote{https://docs.perplexity.ai/docs/getting-started/integrations/mcp-server}, Brave Search~\footnote{https://github.com/brave/brave-search-mcp-server}, and Tavily Search~\footnote{https://docs.tavily.com/documentation/mcp} as the backend search engine.

\begin{table*}[t]
\centering
\scriptsize
\begin{adjustbox}{max width=0.9\textwidth,max totalheight=0.9\textheight,keepaspectratio}
\rowcolors{1}{blue!8}{blue!8}
\begin{tabular}{lcccccccc}
\toprule
\textbf{Model}
& \multicolumn{2}{c}{\textbf{\notool}}
& \multicolumn{2}{c}{\textbf{\withtool}}
& \multicolumn{2}{c}{\textbf{\auto}}
& \multicolumn{2}{c}{\textsc{\textbf{Optimal}}} \\
\cmidrule(lr){2-3} \cmidrule(lr){4-5} \cmidrule(lr){6-7} \cmidrule(lr){8-9}
& Score & Calls & Score & Calls & Score & Calls & Score & Calls \\
\midrule
GPT-OSS-120B (Perplexity)      & 0.61 & 0   & 0.78  & 500  & 0.72 & 149 & 0.82 & 307 \\
GPT-OSS-120B (Brave)      & 0.61 & 0   & 0.78  & 500  & 0.73 & 150 & 0.82 & 305 \\
GPT-OSS-120B (Tavily)      & 0.61 & 0   & 0.75  & 500  & 0.73 & 150 & 0.81 & 269 \\
GPT-OSS-120B (Google via SerpAPI)  & 0.61 & 0   & 0.76 & 500 & 0.72 & 152 & 0.81 & 300 \\
\bottomrule
\end{tabular}
\end{adjustbox}
\caption{
Performance on the entity task under different tool-use strategies.
\textbf{\notool}: no access to tools.
\textbf{\withtool}: tool is always invoked.
\textbf{\auto}: model autonomously decides.
\textsc{\textbf{Optimal}}: oracle policy selecting the best decision per instance. Row color indicates the harness: \colorbox{blue!8}{\strut Trained}.
}
\label{tab:perplexity}
\vspace{-10pt}
\end{table*}

\begin{figure}[t]
    \centering
    \includegraphics[width=0.88\linewidth]{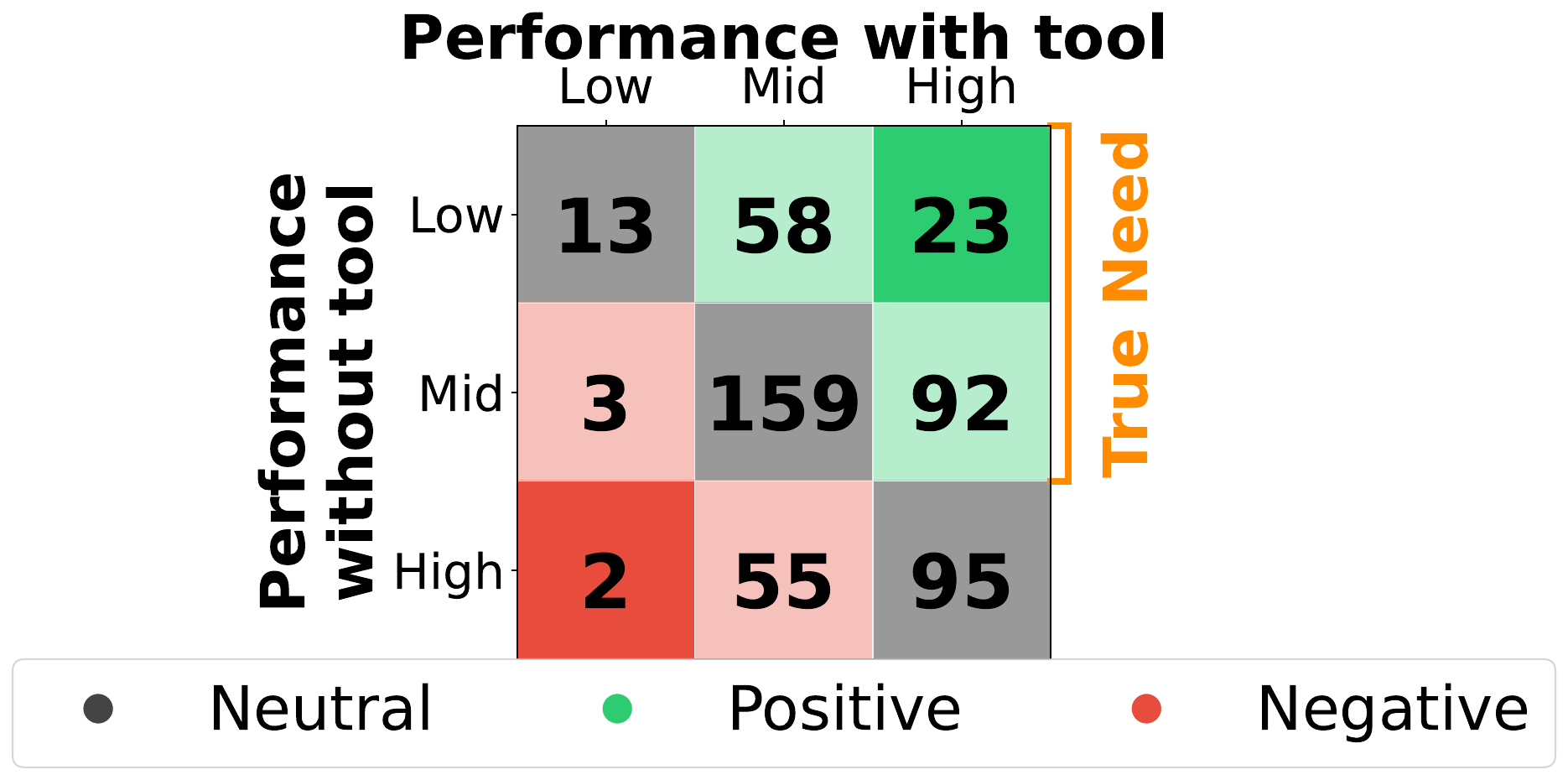}
\caption{
\textbf{[Entity Task, Perplexity Search] True need and positive utility are correlated, but not perfectly aligned.}
Rows group by the model’s (GPT-OSS-120B) factuality scores under \notool{} (parametric knowledge), while columns show scores under \withtool{}. Scores are bucketed into low (0--0.1), mid (0.1--0.9), and high (0.9--1).
Cells above the diagonal indicate {\color{mygreen}\textit{positive utility}}, while those below indicate {\color{red}\textit{negative utility}}. The bracket highlights the {\color{orange}\textit{true need}} region, where low \notool{} scores reflect insufficient parametric knowledge.
}
    \label{fig:perplexity-actual-need-utility}
\end{figure}

Figure~\ref{fig:perplexity-actual-need-utility} shows that, even when using Perplexity search as the backend, not all tool calls yield positive utility. Only when the model genuinely requires external assistance do most tool calls result in a benefit.

\begin{figure}[t]
\centering
\begin{subfigure}{0.32\linewidth}
    \centering
    \includegraphics[width=\linewidth]{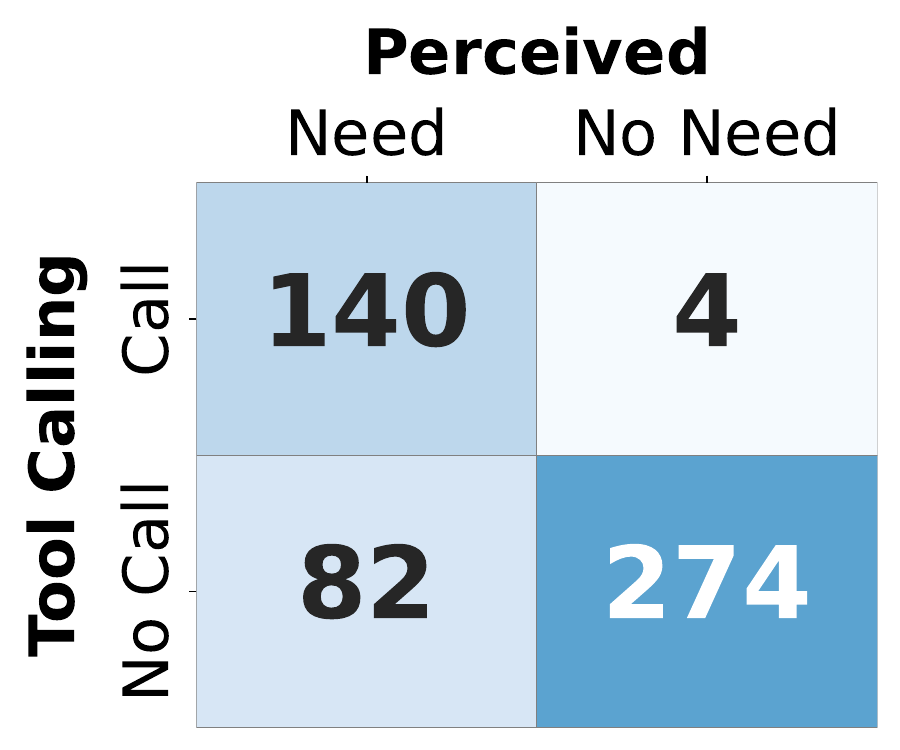}
    \caption{Perceived Need Prompt V1}
\end{subfigure}\hfill
\begin{subfigure}{0.32\linewidth}
    \centering
    \includegraphics[width=\linewidth]{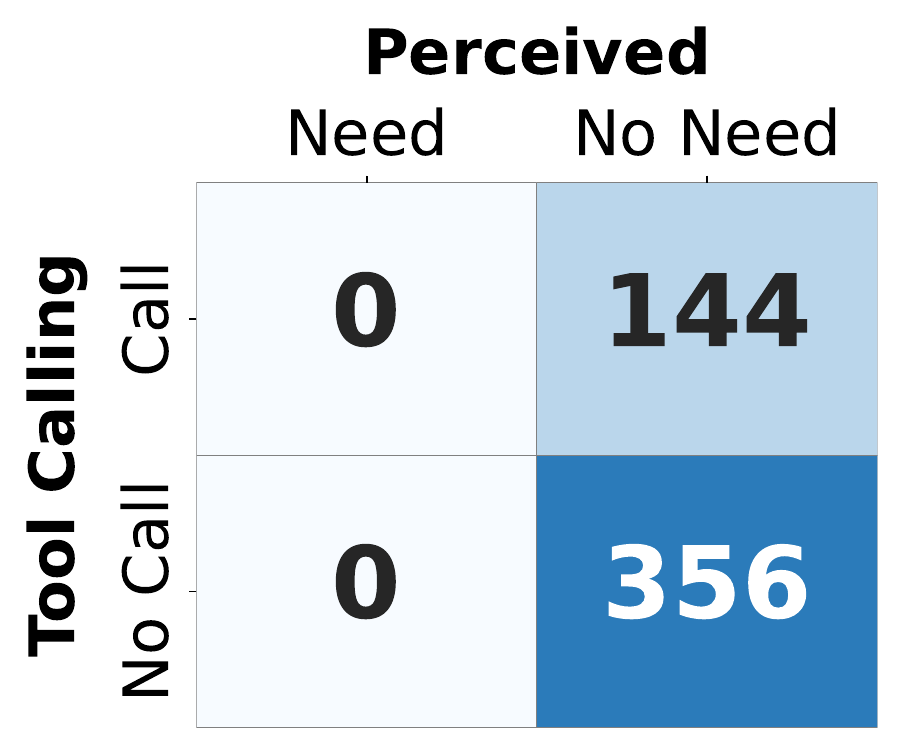}
    \caption{Perceived Need Prompt V2}
\end{subfigure}\hfill
\begin{subfigure}{0.32\linewidth}
    \centering
    \includegraphics[width=\linewidth]{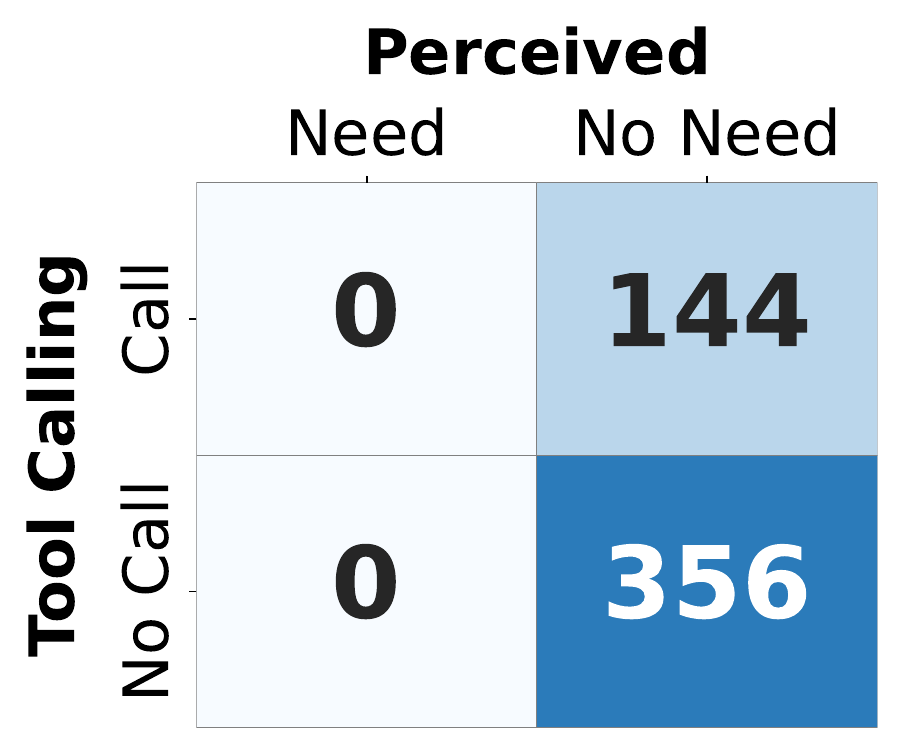}
    \caption{Perceived Need Prompt V13}
\end{subfigure}\hfill
\caption{\textbf{[Entity Task; Perplexity Search] Perceived need is only partially aligned with tool call (perceived utility).} Model: GPT-OSS-120B. The x-axis shows the model's perceived need, and the y-axis shows perceived utility / tool-call decisions.}
\label{fig:perplexity_figure_need_utility}
\vspace{-15pt}
\end{figure}

Figure~\ref{fig:perplexity_figure_need_utility} shows that after changing the web search backend, the model's self-perception of need and utility is still partly aligned. 

\begin{figure}
    \centering
    \includegraphics[width=0.9\linewidth]{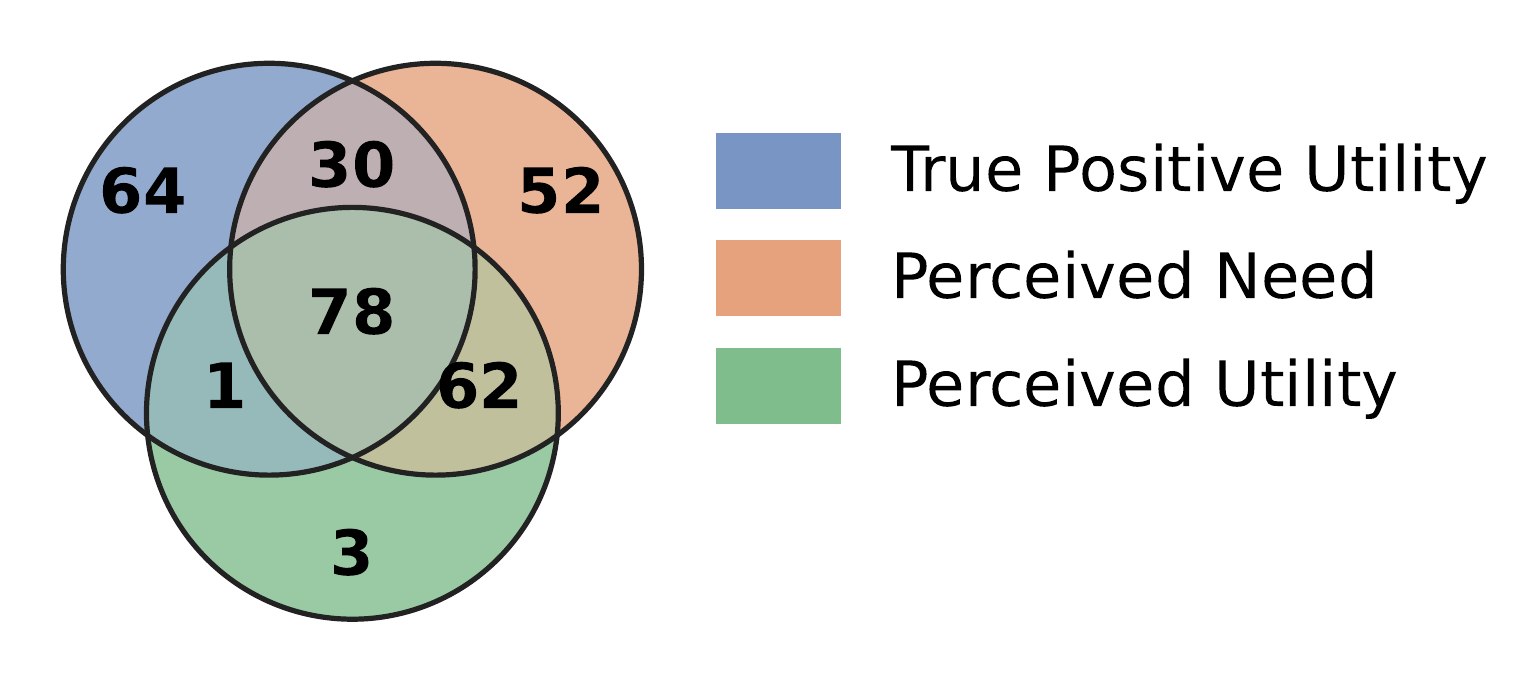}
    \caption{
    \textbf{[Entity Task; Perplexity Search] Perceived signals only partially align with true utility.}
    Venn diagrams of \textit{True Positive Utility}, \textit{Perceived Need}, and \textit{Perceived Utility} for GPT-OSS-120B on the entity task illustrate this misalignment.
    Ideally, \textit{Perceived Utility} $\subseteq$ \textit{Perceived Need} $\subseteq$ \textit{True Positive Utility}. However, this nesting is violated, indicating misalignment with true utility and helping to explain the suboptimal performance of \auto{}.
    }
    \label{fig:perplexity_venn-gpt}
\end{figure}

However, in Figure~\ref{fig:perplexity_venn-gpt}, we still observed that the perceived need and utility are not aligned with the true positive utility, which leads to non-optimal results with the Perplexity web search backend.

\begin{figure}[t]
\centering

\begin{subfigure}{0.48\linewidth}
    \centering
    \includegraphics[width=\linewidth]{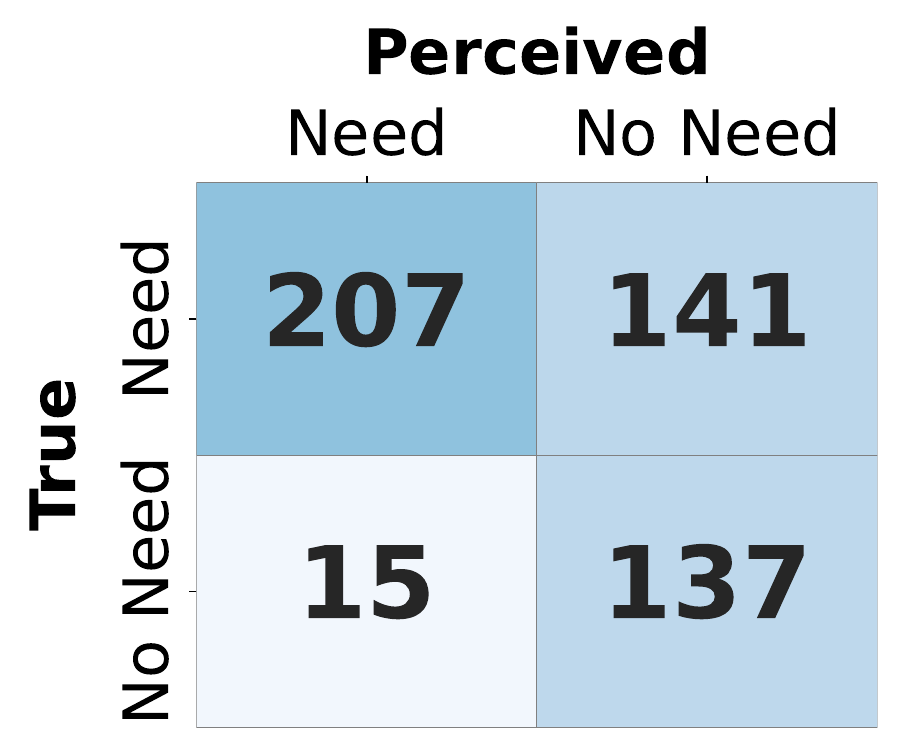}
    \caption{True Need VS Perceived Need}
\end{subfigure}

\begin{subfigure}{0.48\linewidth}
    \centering
    \includegraphics[width=\linewidth]{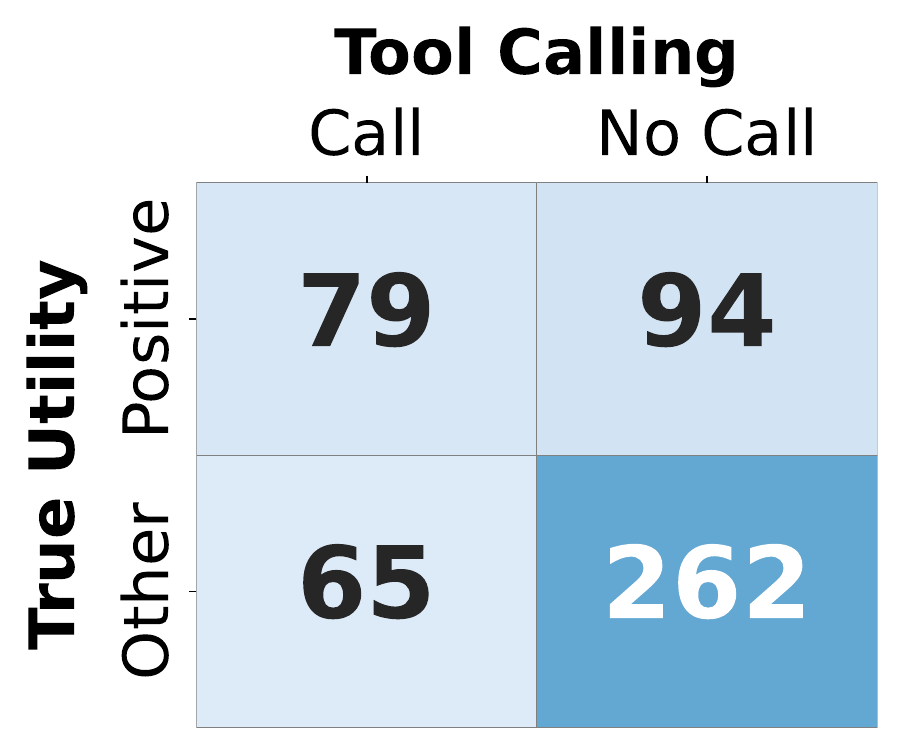}
    \caption{True Utility VS Perceived Utility}
\end{subfigure}

\caption{
\textbf{[Entity Task; Perplexity Search] The perceived need and utility are not aligned with the true need and utility.}
Entity Task; Top: perceived need matrices. Bottom: true vs.\ perceived utility across models.
}
\label{fig:perplexity_true_perceived}
\end{figure}

In the Figure~\ref{fig:perplexity_true_perceived}, we then show the breakdown of the misalignment through the normative lens and the descriptive lens.

\begin{figure}[t]
\vspace{-6pt}
\centering
\begin{subfigure}{\linewidth}
    \centering
    \includegraphics[width=\linewidth]{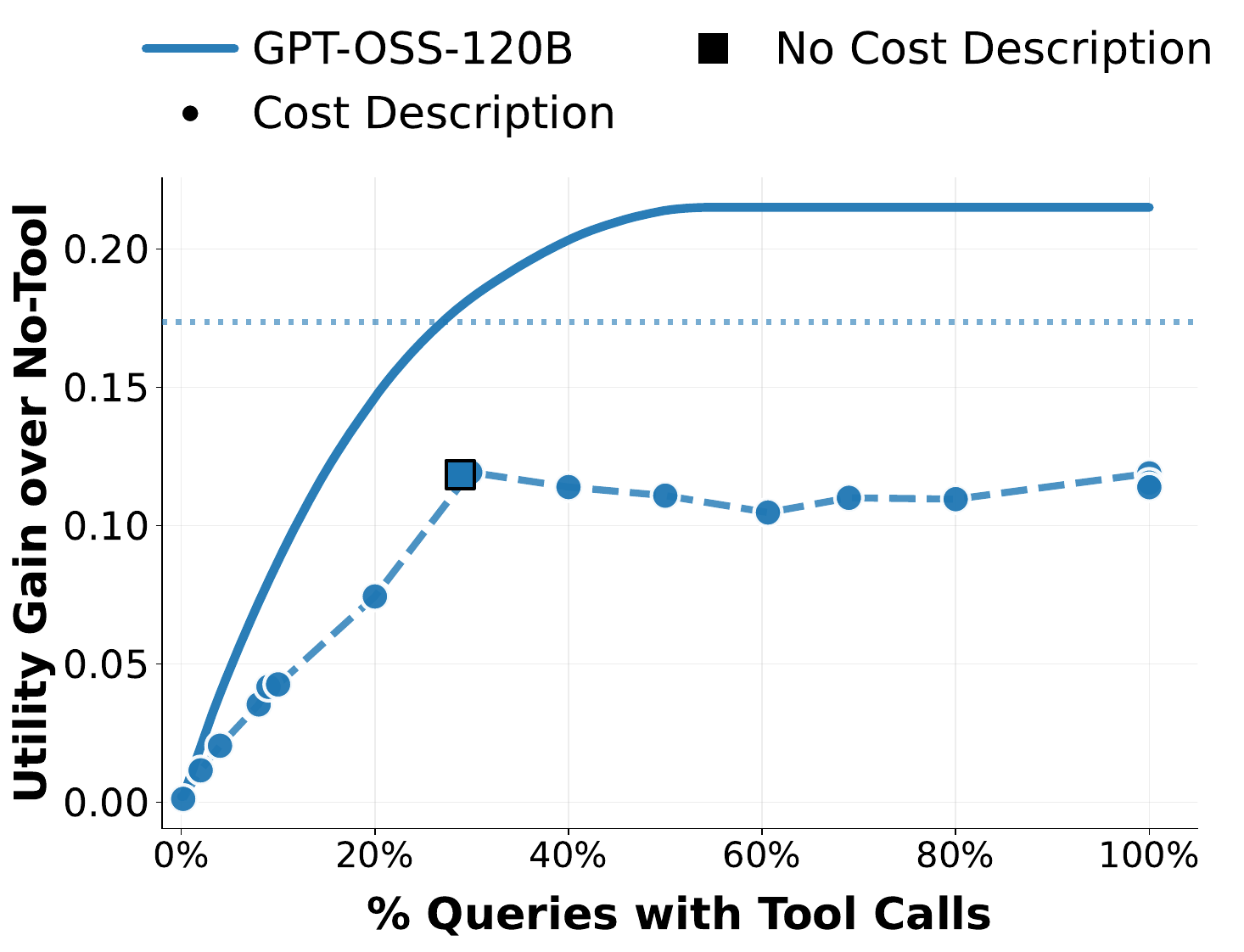}
    \caption{\textbf{Utility gain with hard stop after exceeding the budget.}}
\end{subfigure}\hfill
\begin{subfigure}{\linewidth}
    \centering
    \includegraphics[width=\linewidth]{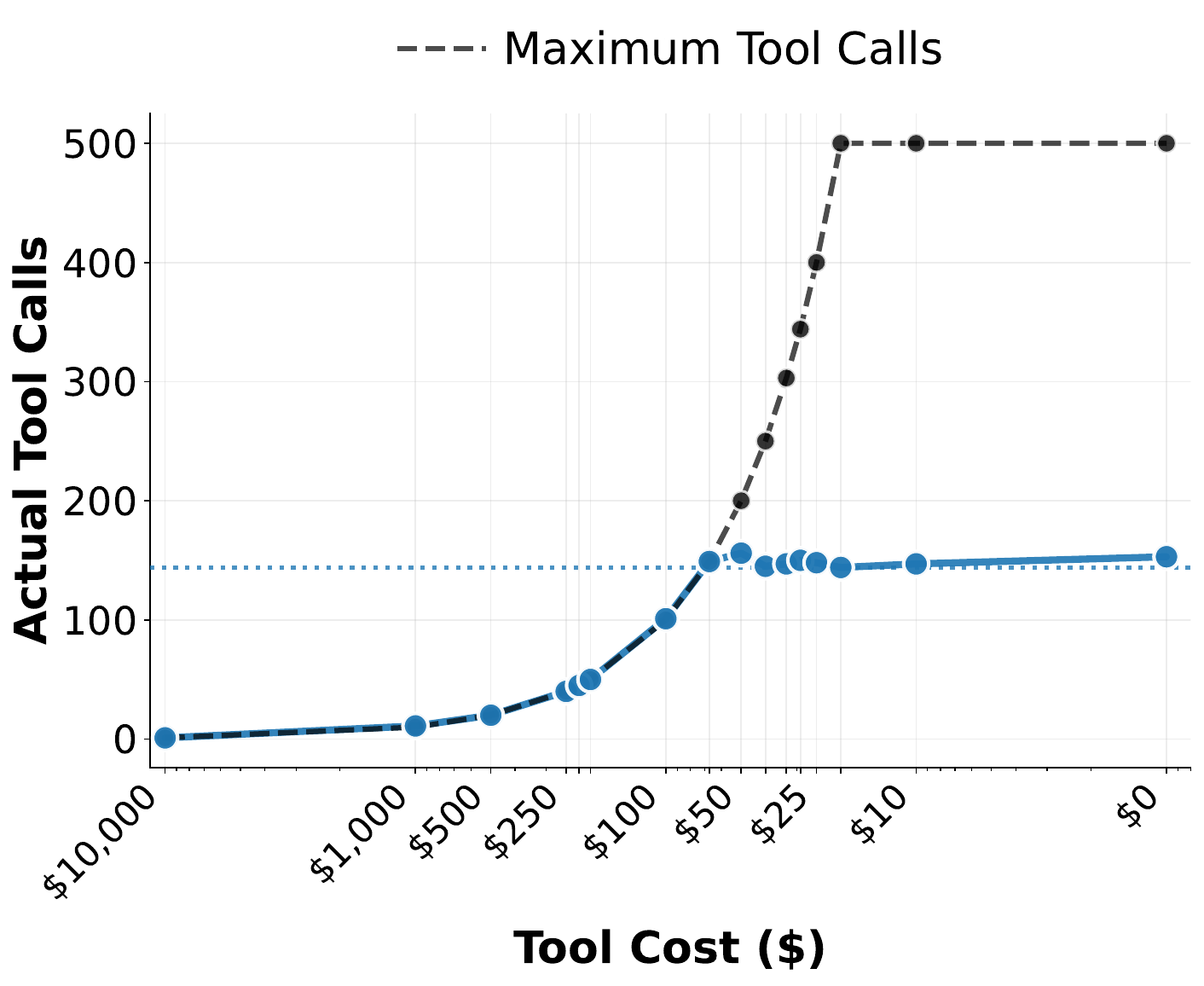}
    \caption{\textbf{Tool-calling behavior without hard stop.}}
\end{subfigure}
\caption{
\textbf{[Entity Task; Perplexity Search] Cost-aware tool use with explicit budget notification.}
\textbf{Left:} Utility gain over the no-tool baseline under varying cost constraints. Solid lines show oracle allocation (optimal top-$k$), dashed lines show model performance with cost information, squares denote no cost-awareness, and dotted lines indicate always-calling.
\textbf{Right:} Actual tool calls without budget enforcement. Models do not reliably reduce or stop calls as cost increases, despite being provided with cost and remaining budget.
}
\label{fig:perplexity_affordability_combined_v2}
\end{figure}

\begin{figure}[t]
\centering
\begin{subfigure}{\linewidth}
    \centering
    \includegraphics[width=\linewidth]{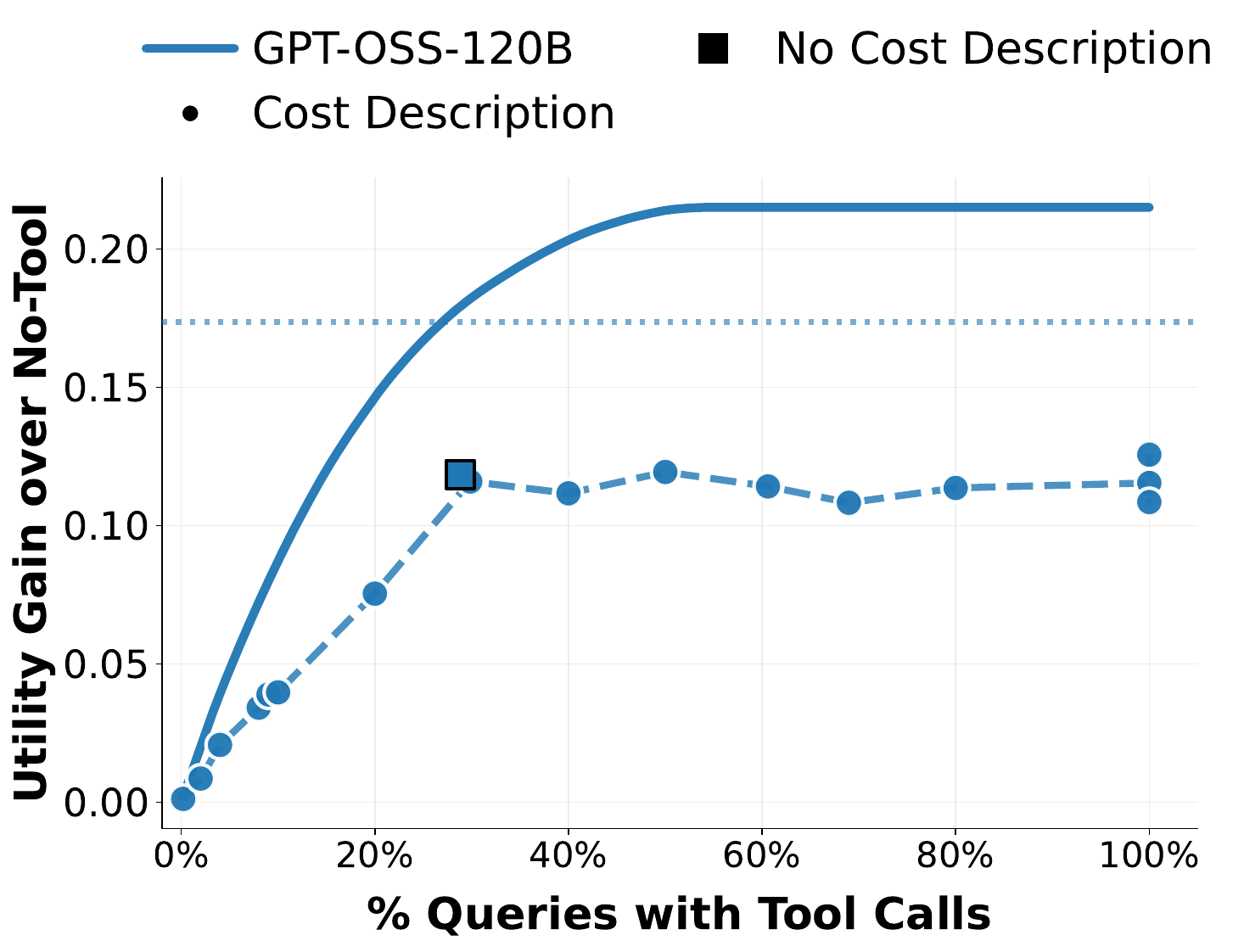}
    \caption{\textbf{Utility gain with hard stop after exceeding the budget.}}
\end{subfigure}\hfill
\begin{subfigure}{\linewidth}
    \centering
    \includegraphics[width=\linewidth]{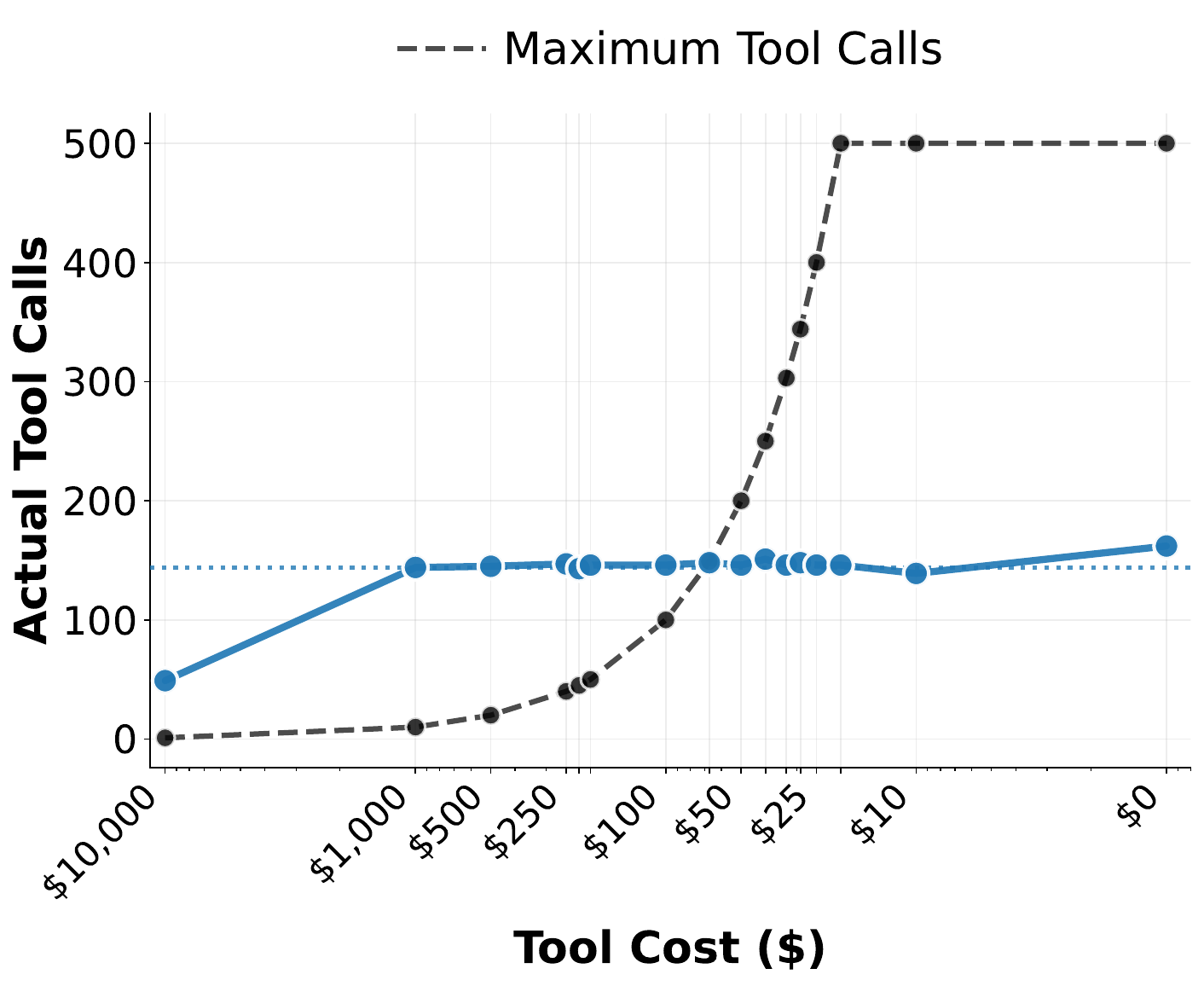}
    \caption{\textbf{Tool-calling behavior without hard stop.}}
\end{subfigure}
\caption{
\textbf{[Entity Task; Perplexity Search] Cost-aware tool use with implicit budget notification.}
\textbf{Left:} Utility gain over the no-tool baseline under varying cost constraints. Solid lines show oracle allocation (optimal top-$k$), dashed lines show model performance with cost information, squares denote no cost-awareness, and dotted lines indicate always-calling.
\textbf{Right:} Actual tool calls without budget enforcement. Models do not reliably reduce or stop calls as cost increases, despite being provided with cost and remaining budget.
}
\label{fig:perplexity_affordability_combined_v1}
\end{figure}

\section{InVivoQuery Task}

In this section, we show the additional results for the InVivoQuery task. 

\begin{figure*}
    \centering
    \includegraphics[width=\linewidth]{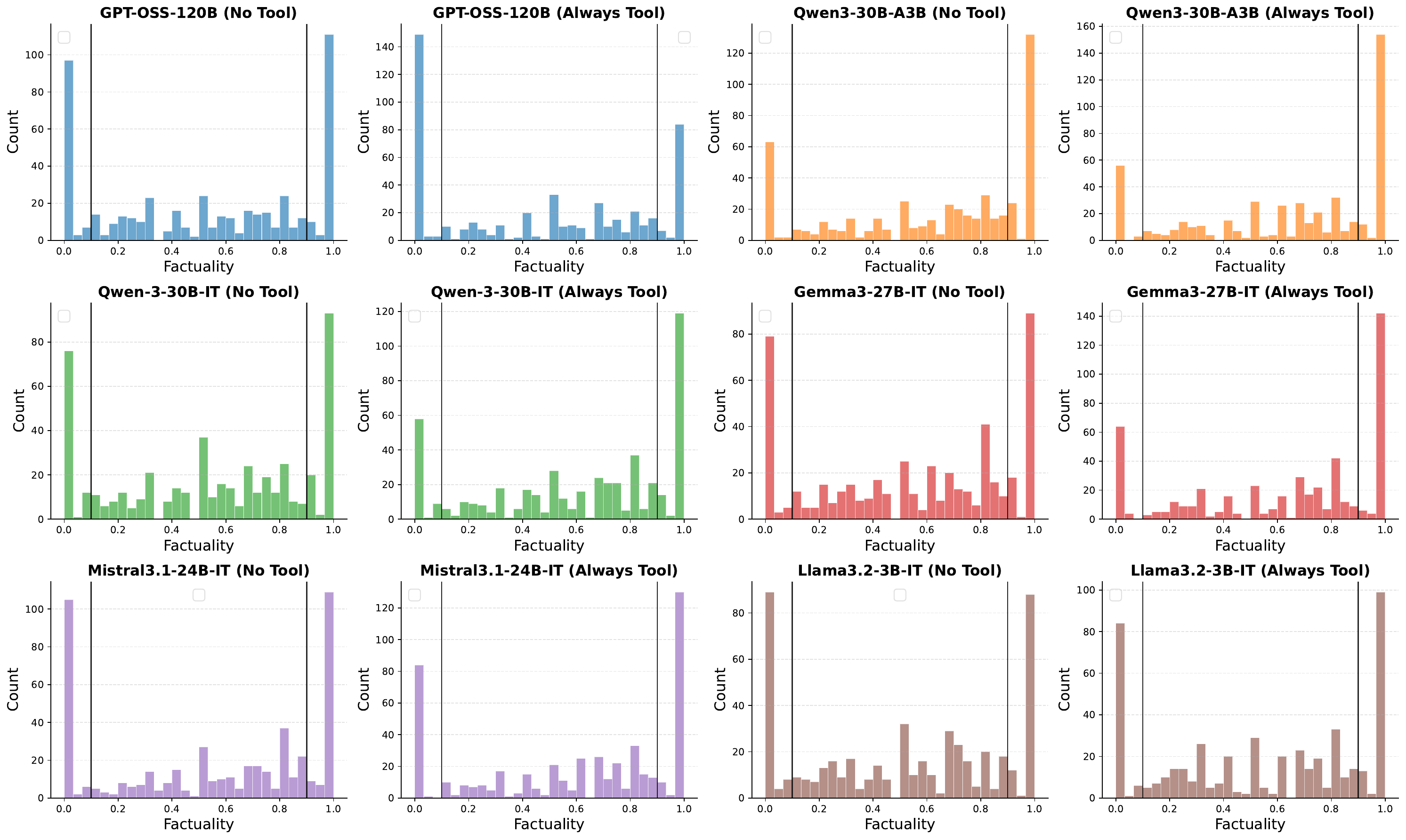}
    \caption{\textbf{[InVivoQuery Task] factuality distribution across different models.}}
    \label{fig:invivo_hist}
\end{figure*}

In Figure~\ref{fig:invivo_hist}, we show the factuality score distribution across all the models and entities for the InVivoQuery Task. Visualizing the distribution is important because aggregate metrics alone (e.g., mean or accuracy) can obscure underlying differences in model behavior. The distribution provides a more fine-grained view of how factuality scores are spread, revealing patterns such as skewness, variance, and the presence of extreme cases.
In particular, this figure allows us to examine how factuality shifts when tool use is enabled versus disabled. Rather than only observing average improvements, the distribution highlights whether gains are consistent across samples or driven by a subset of cases. 

\subsection{Normative Lens}

\begin{figure*}[t]
\begin{subfigure}{0.25\linewidth}
    \centering
    \includegraphics[width=\linewidth]{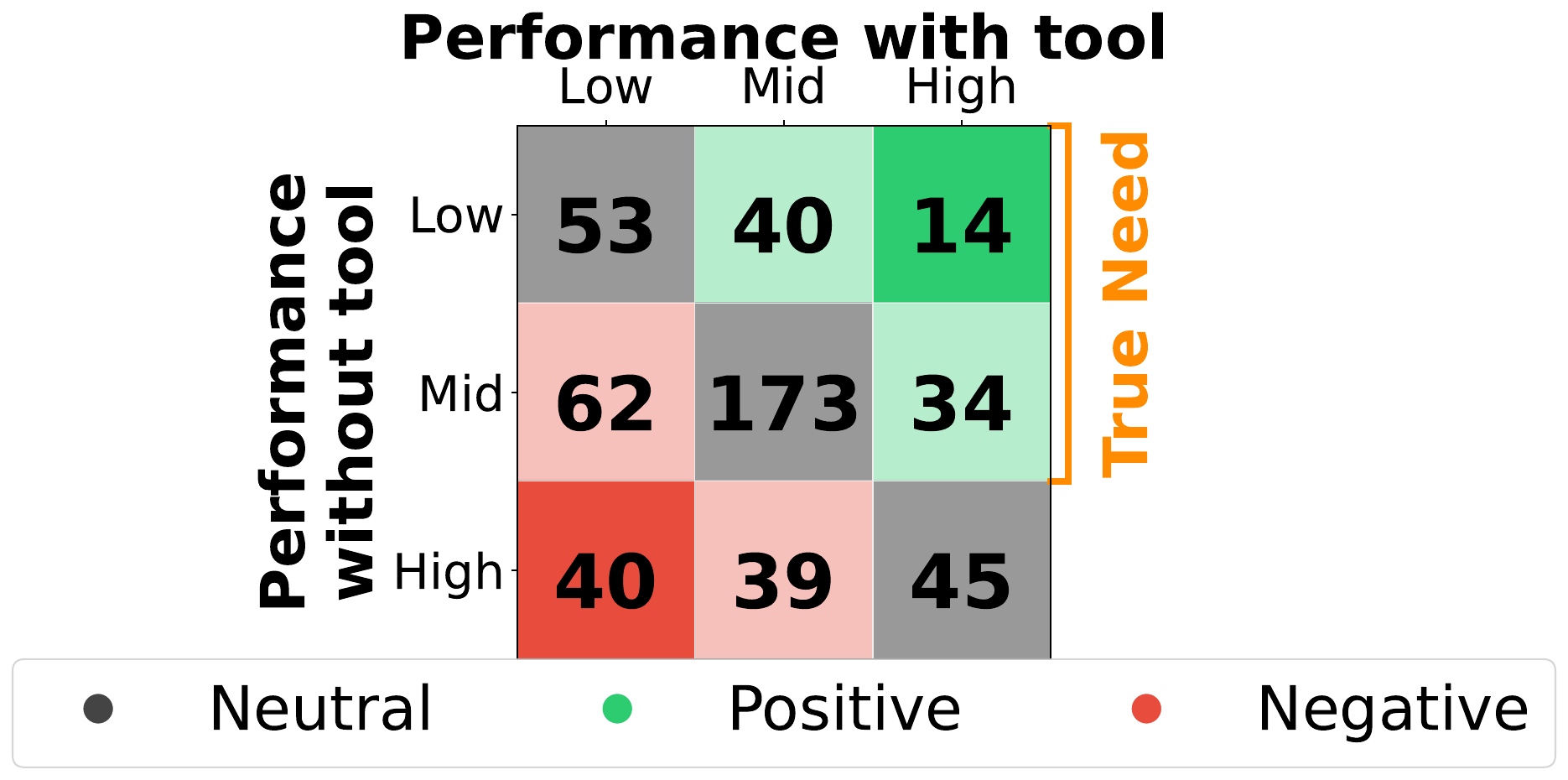}
    \caption{GPT-OSS-120B}
\end{subfigure}
\hfill
\begin{subfigure}{0.25\linewidth}
    \centering
    \includegraphics[width=\linewidth]{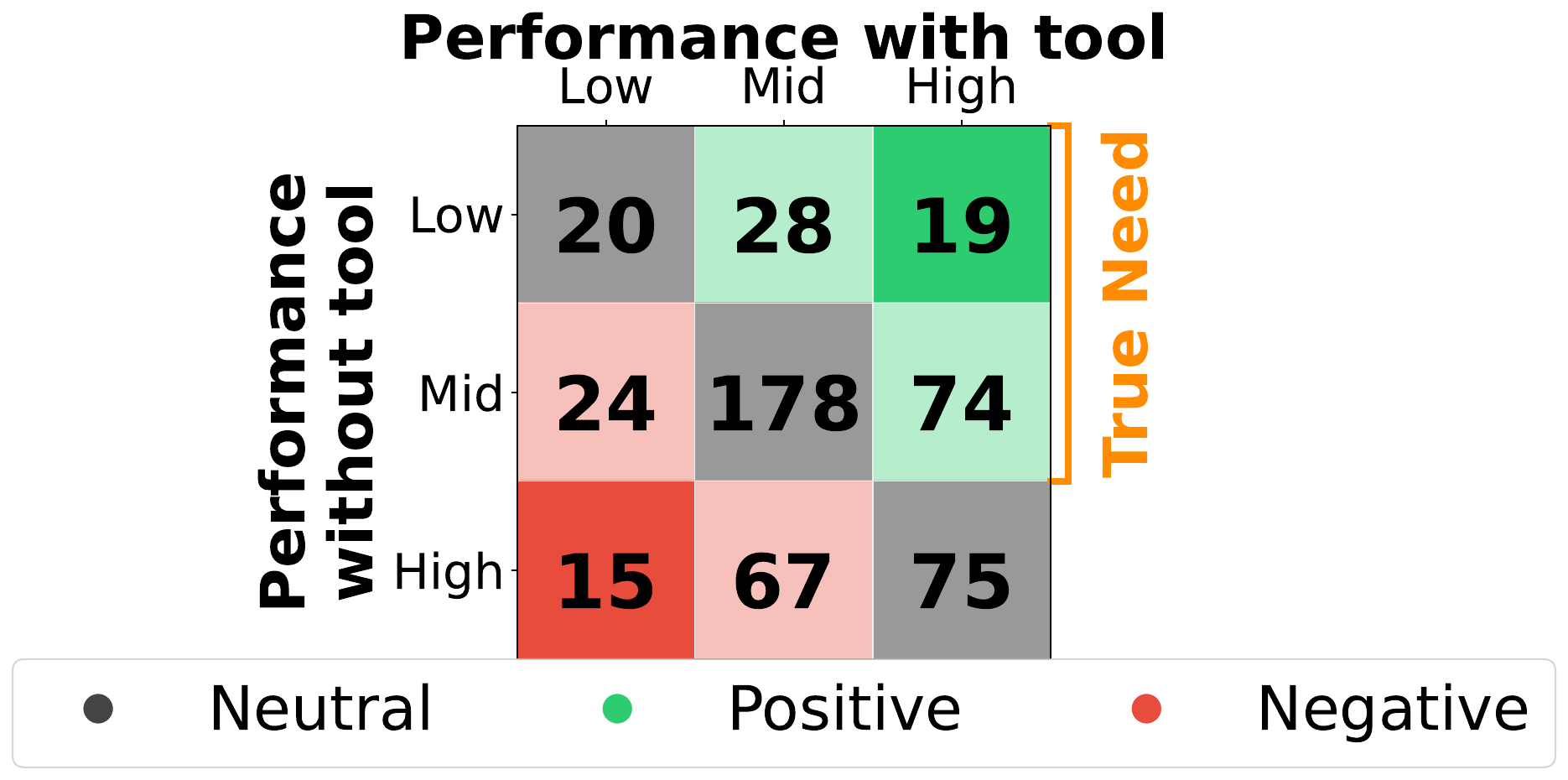}
    \caption{Qwen3-30B-A3B}
\end{subfigure}
\hfill
\begin{subfigure}{0.25\linewidth}
    \centering
    \includegraphics[width=\linewidth]{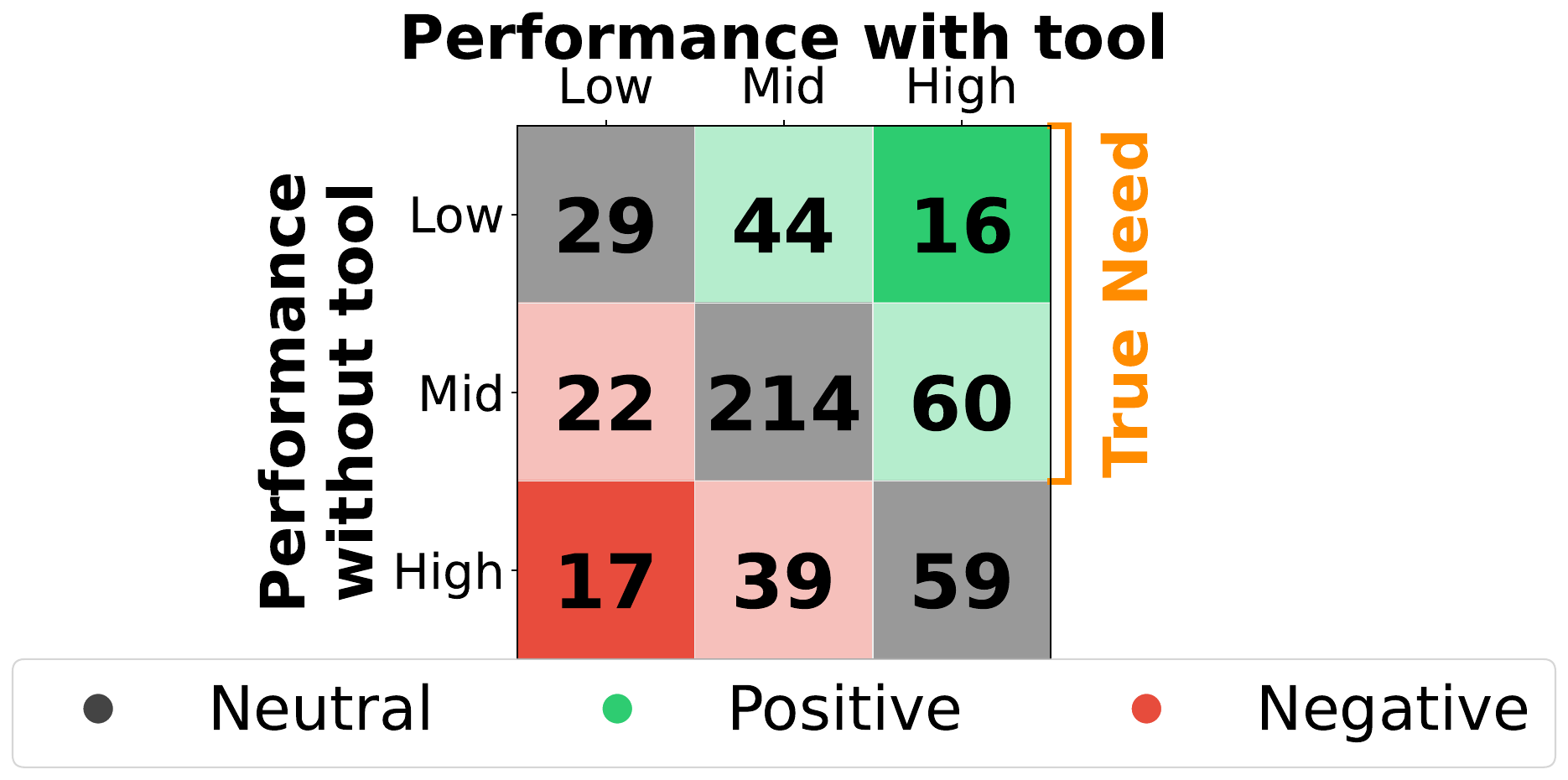}
    \caption{Qwen3-30B-A3B-Instruct}
\end{subfigure}
\begin{subfigure}{0.25\linewidth}
    \centering
    \includegraphics[width=\linewidth]{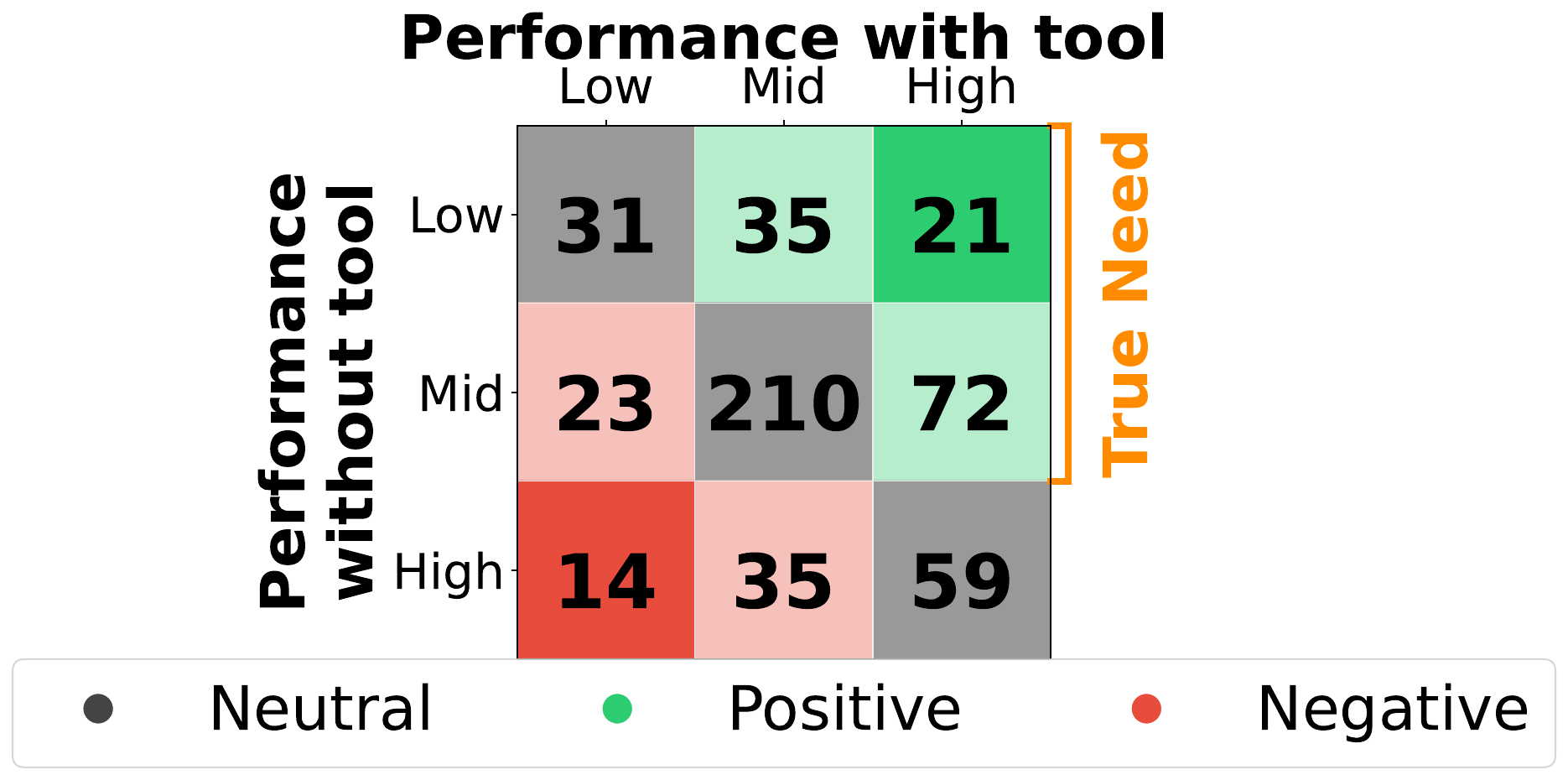}
    \caption{Gemma-3-27B-IT}
\end{subfigure}
\hfill
\begin{subfigure}{0.25\linewidth}
    \centering
    \includegraphics[width=\linewidth]{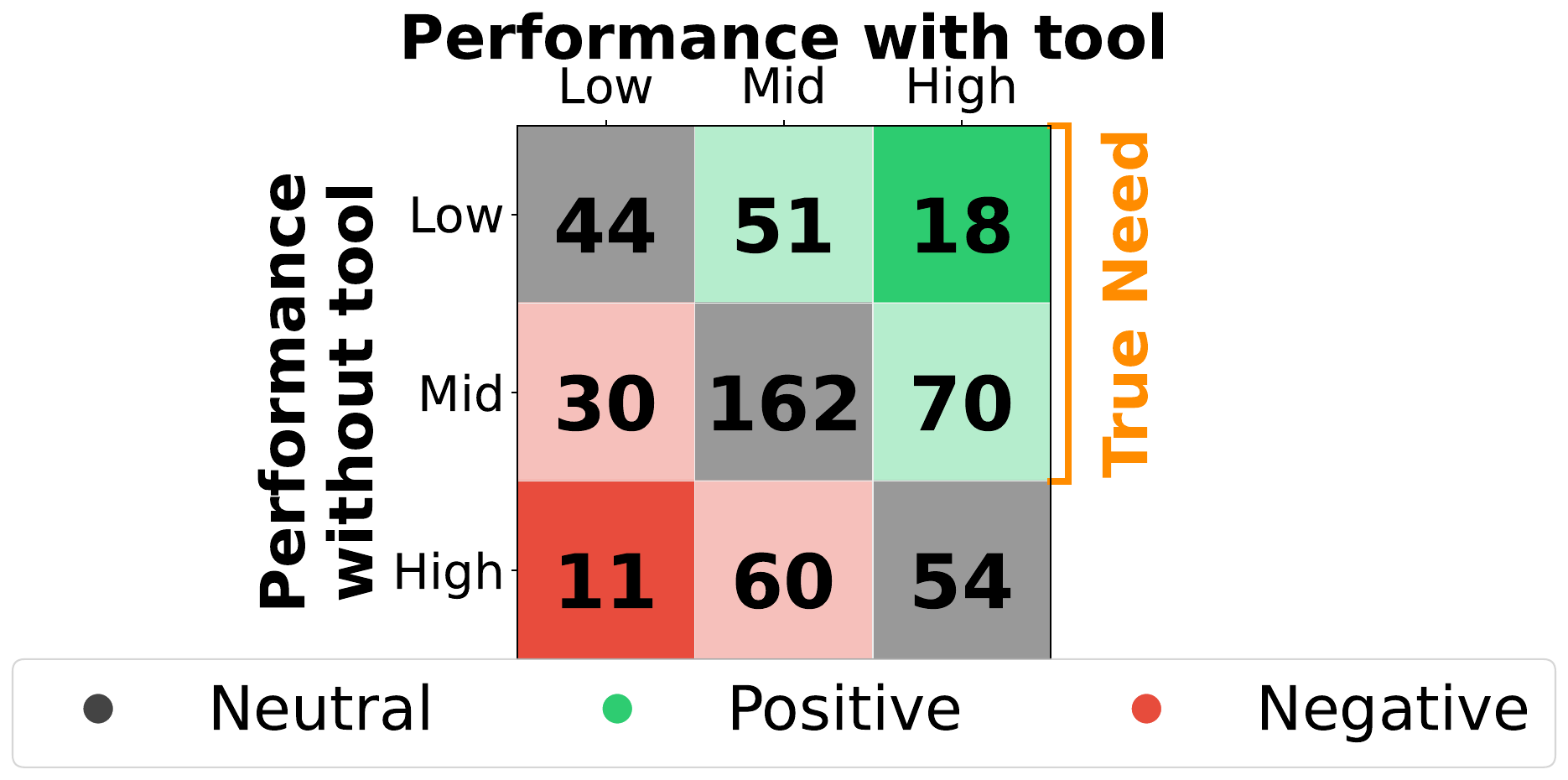}
    \caption{Mistral-3.1-24B-IT}
\end{subfigure}
\hfill
\begin{subfigure}{0.25\linewidth}
    \centering
    \includegraphics[width=\linewidth]{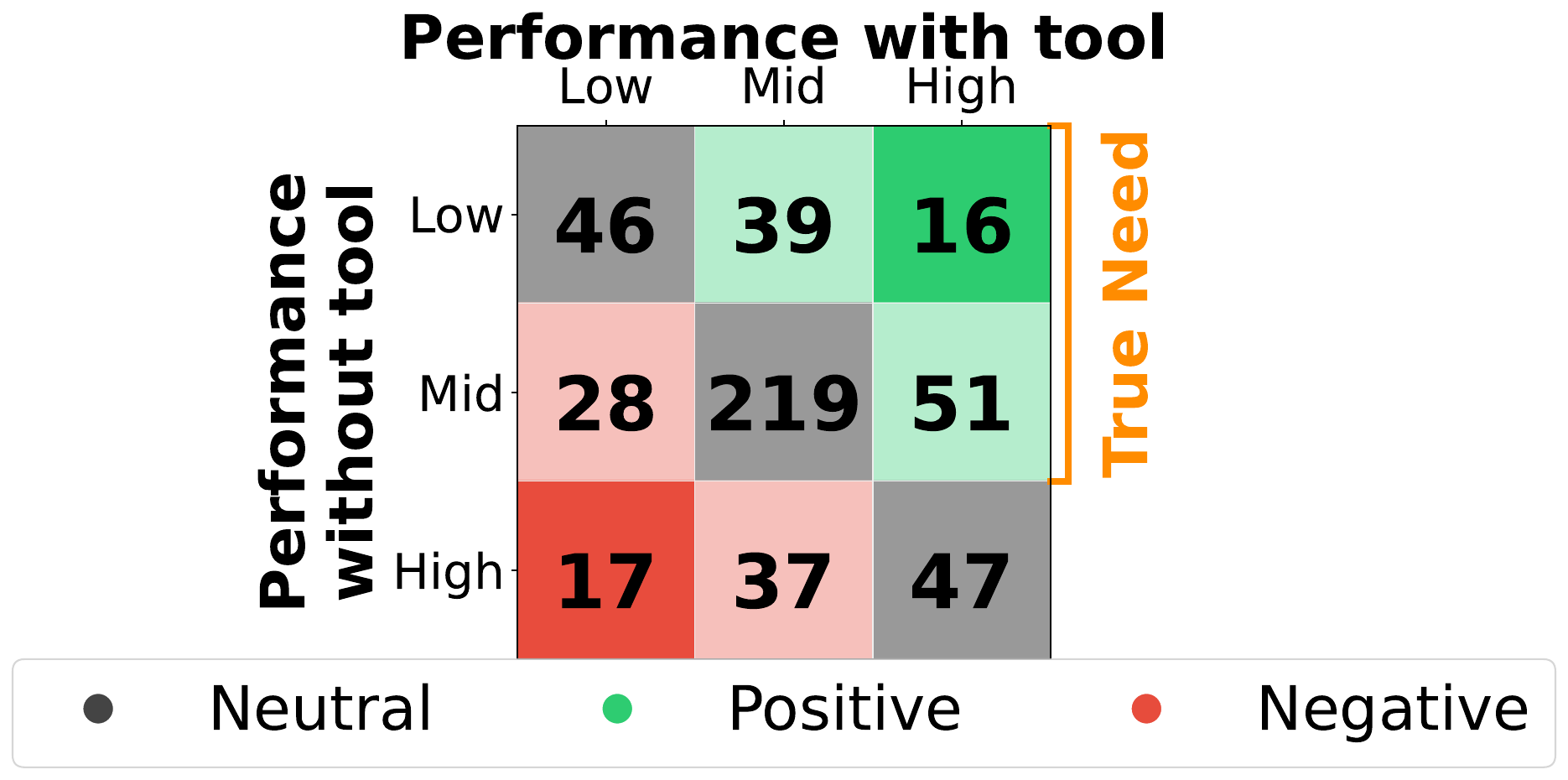}
    \caption{Llama-3.2-3B-IT}
\end{subfigure}
\hfill
\begin{subfigure}{0.25\linewidth}
    \centering
    \includegraphics[width=\linewidth]{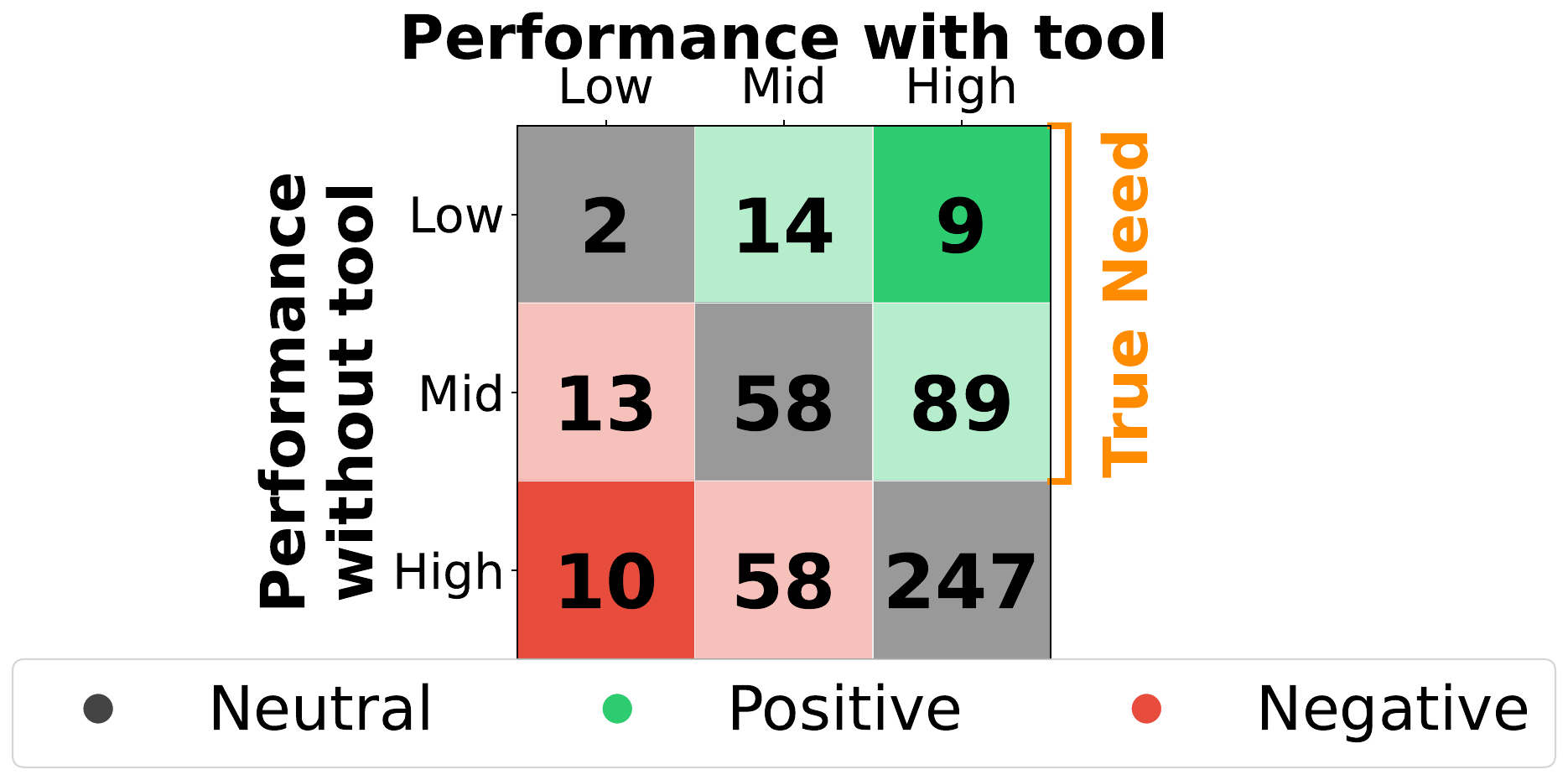}
    \caption{GPT-5.5}
\end{subfigure}

\caption{\textbf{[InVivoQuery Task] \notool vs.\ \withtool~ performance.}
        Rows group entities by the model’s factuality score \textit{without a tool} (reflecting parametric knowledge), while columns group scores when tool use is \textit{forced}. Each cell reports the count and the column percentage. Off-diagonal cells indicate performance shifts due to tool use: cells above the diagonal show cases where the tool has {\color{green}\textit{positive utility}}, while cells below the diagonal indicate cases where the tool has {{\color{red}\textit{negative utility}}}. The dashed bracket marks the region of {{\color{orange}\textit{True Need}}}, where \textit{Low} or \textit{Mid} No-Tool scores suggest insufficient parametric knowledge and thus a likely need for an external tool. }
\label{fig:all_invivo_actul_need_utility}
\end{figure*}

As shown in Figure~\ref{fig:all_invivo_actul_need_utility}, a consistent pattern emerges across all models: \textbf{tool use is most beneficial when it is truly needed, can be harmful when unnecessary, and is often redundant otherwise}. This observation highlights the importance of accurately predicting when to invoke external tools, as indiscriminate usage may introduce noise or errors rather than improving factuality.

\subsection{Descriptive Lens}

\begin{figure*}[t]
    \centering
    \includegraphics[width=0.78\textwidth]{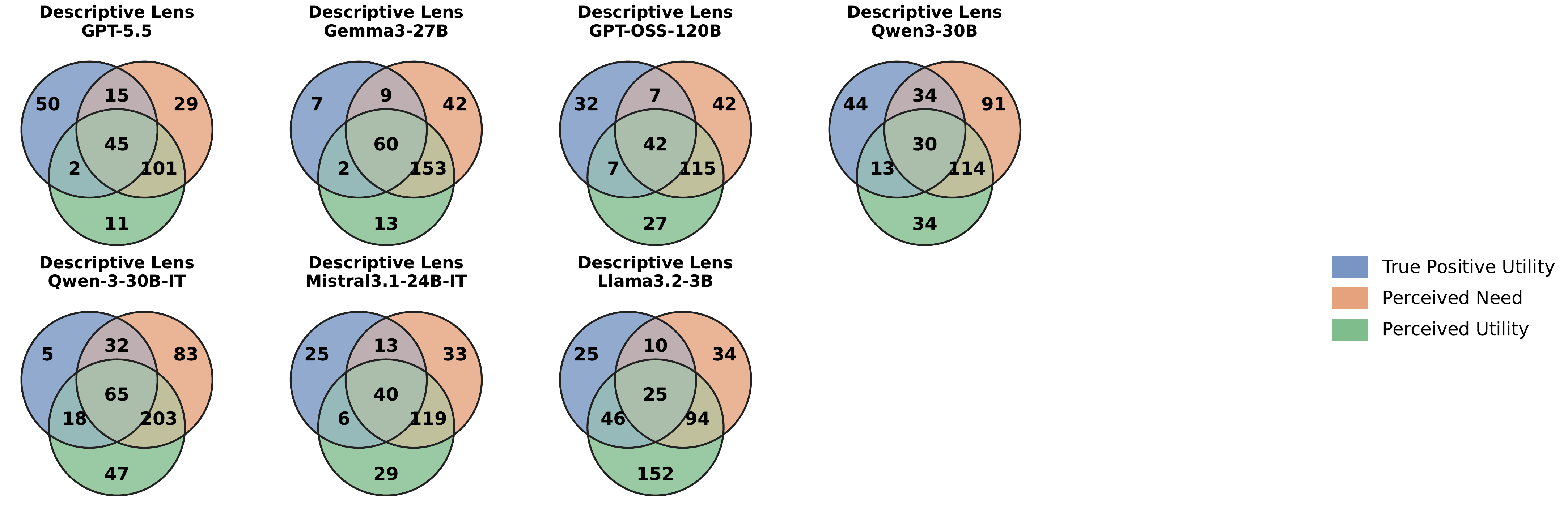}
    \caption{[InVivoQuery Task] Venn diagrams of \textbf{True Positive Utility, Perceived Need, and Perceived Utility}. Each panel shows their empirical overlap for one model. Calls outside true positive utility are non-beneficial; true-positive-utility cases outside perceived utility are missed opportunities. Perceived need is a separate self-assessment and need not be nested within either utility set.}
    \label{fig:venn-invivo}
\end{figure*}

As shown in Figure~\ref{fig:venn-invivo}, there is a consistent misalignment between the perceived need and utility and the true positive utility across all models. This discrepancy indicates that models often fail to accurately identify when tool use is genuinely beneficial. As a result, none of the models achieve optimal tool-calling performance, since effective tool use critically depends on correctly aligning perceived need with actual utility.

\begin{figure*}
\centering
\vspace{-10pt}

\begin{subfigure}{0.25\linewidth}
    \centering
    \includegraphics[width=\linewidth,height=0.105\textheight,keepaspectratio]{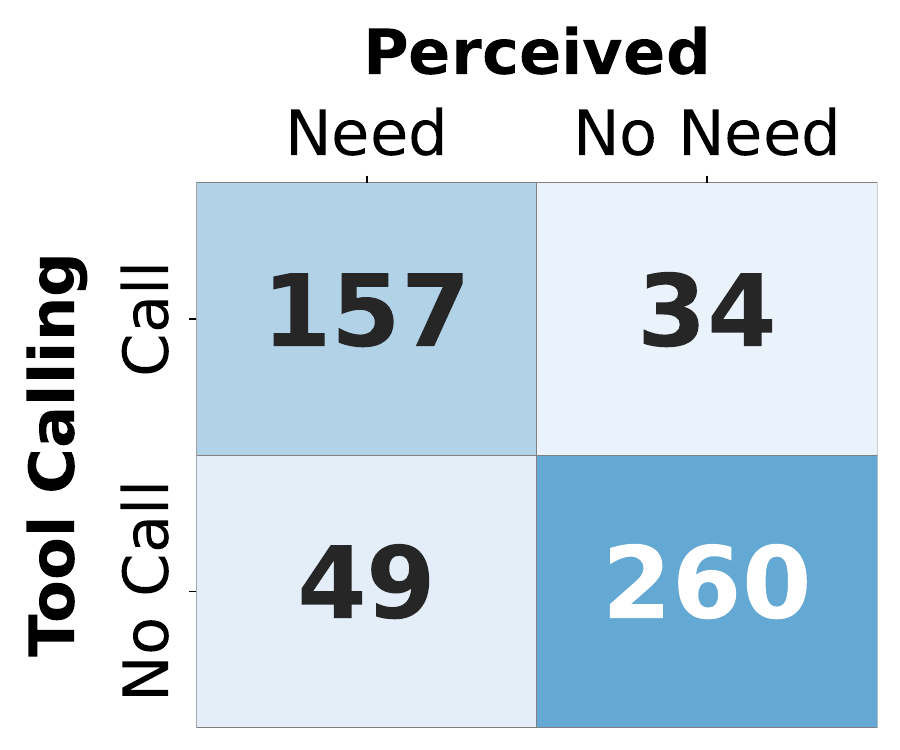}
    \caption{GPT-OSS-120B}
\end{subfigure}\hfill
\begin{subfigure}{0.25\linewidth}
    \centering
    \includegraphics[width=\linewidth,height=0.105\textheight,keepaspectratio]{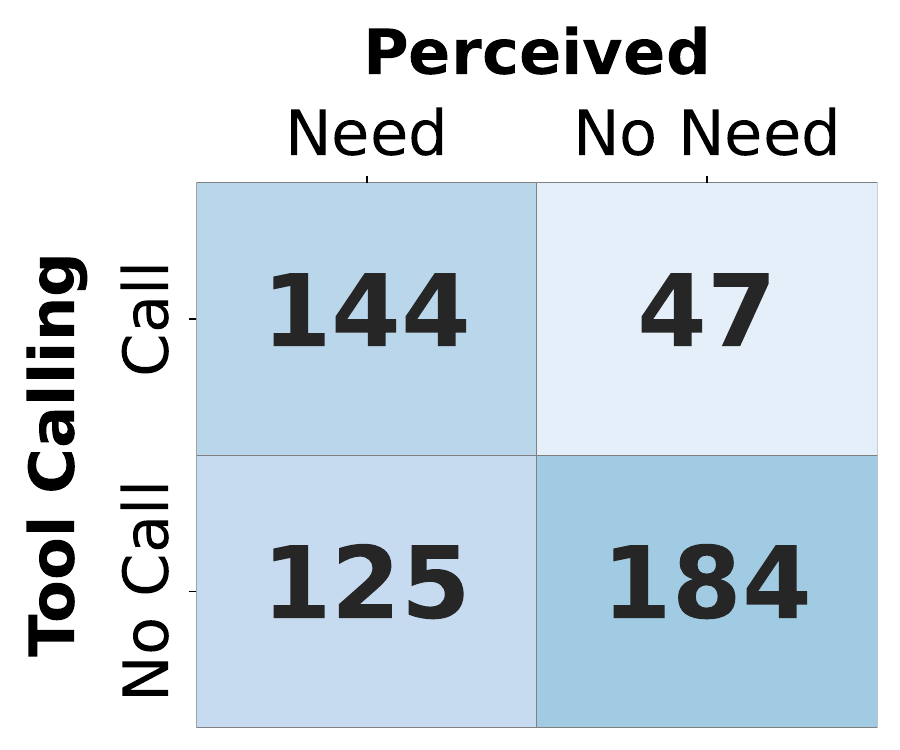}
    \caption{Qwen3-30B-A3B}
\end{subfigure}\hfill
\begin{subfigure}{0.25\linewidth}
    \centering
    \includegraphics[width=\linewidth,height=0.105\textheight,keepaspectratio]{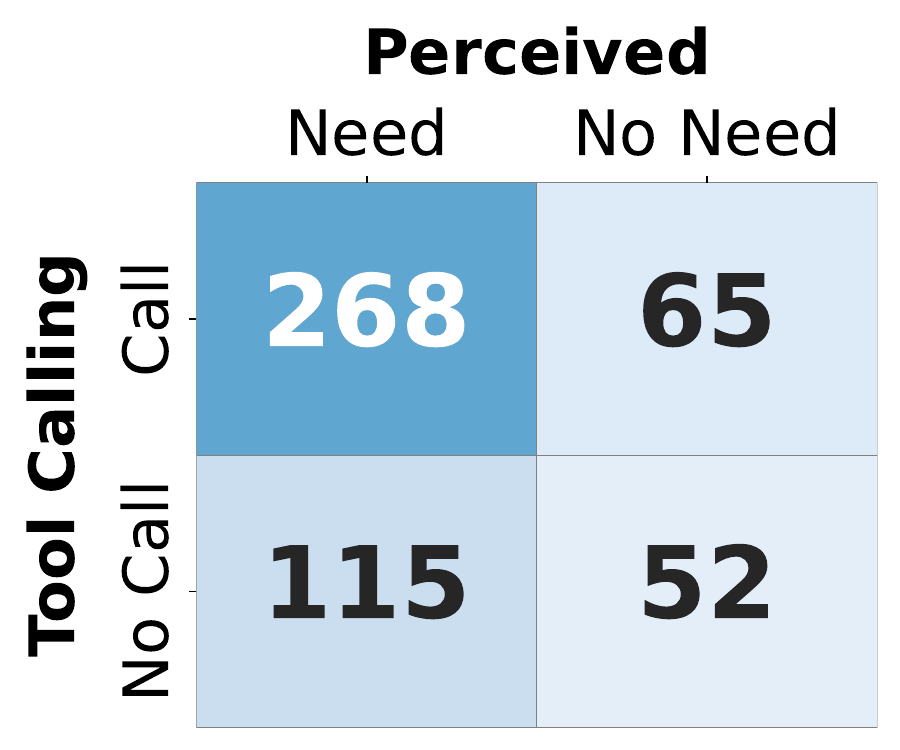}
    \caption{Qwen3-30B-IT}
\end{subfigure}\hfill
\begin{subfigure}{0.24\linewidth}
    \centering
    \includegraphics[width=\linewidth,height=0.105\textheight,keepaspectratio]{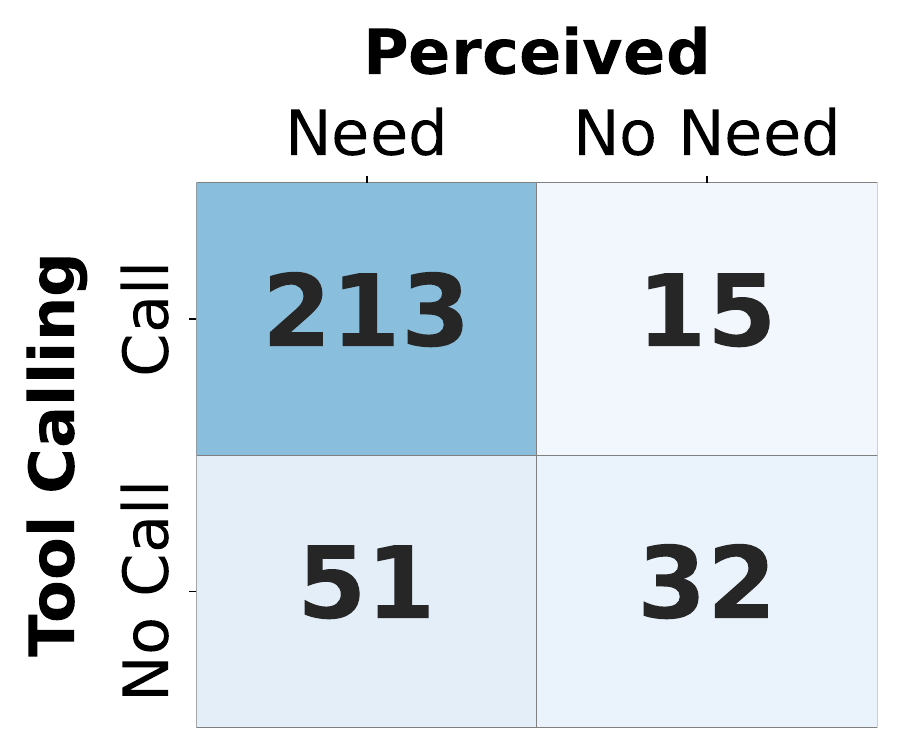}
    \caption{Gemma-3-27B-IT}
\end{subfigure}\hfill
\begin{subfigure}{0.24\linewidth}
    \centering
    \includegraphics[width=\linewidth,height=0.105\textheight,keepaspectratio]{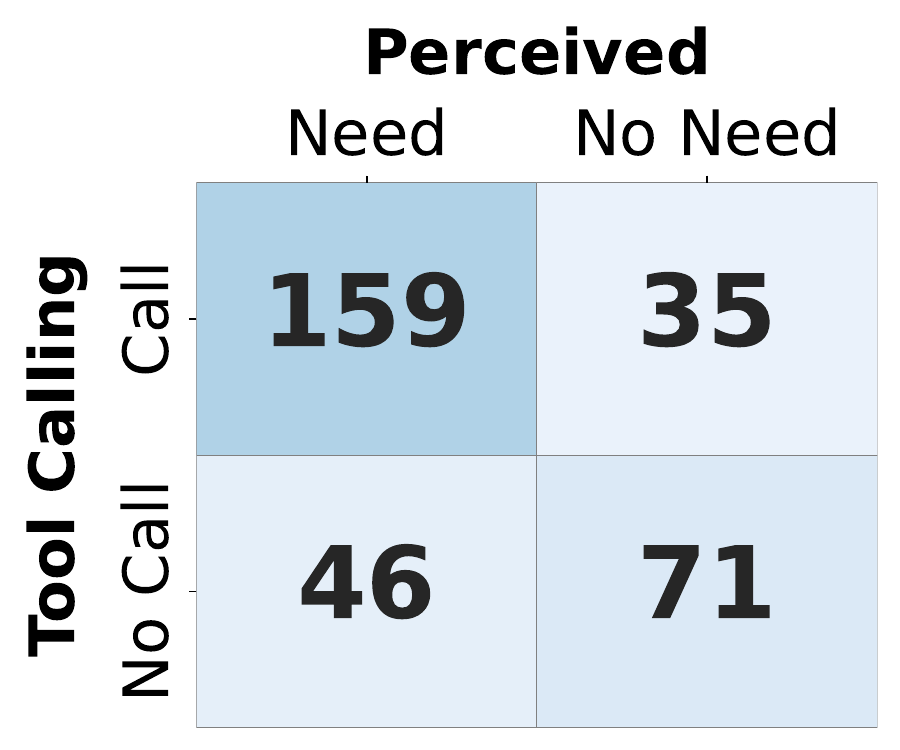}
    \caption{Mistral-3.1-24B-IT}
\end{subfigure}\hfill
\begin{subfigure}{0.24\linewidth}
    \centering
    \includegraphics[width=\linewidth,height=0.105\textheight,keepaspectratio]{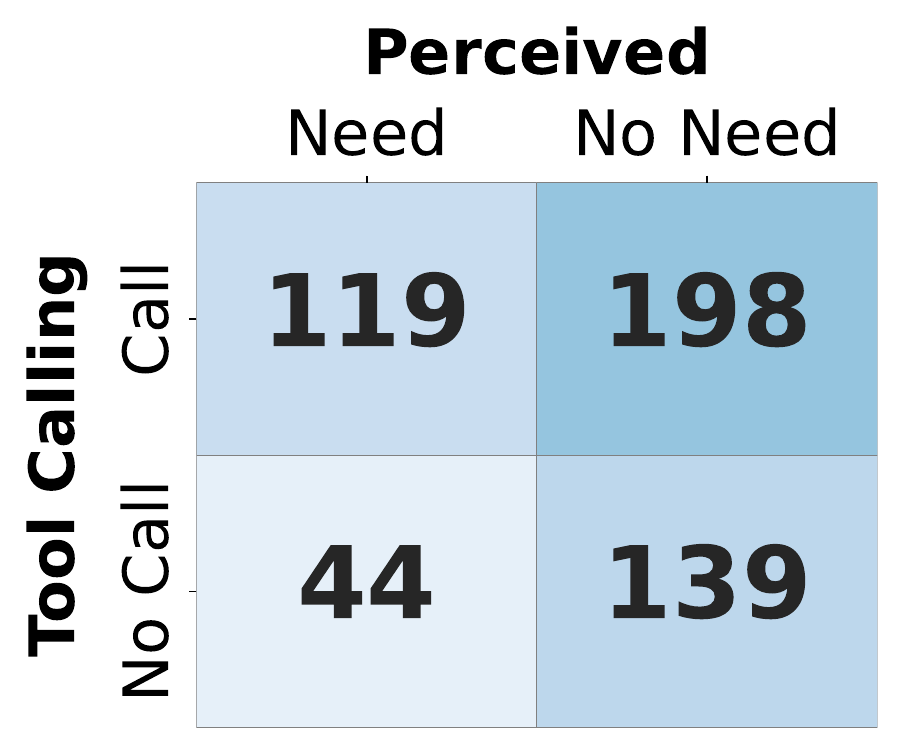}
    \caption{Llama-3.2-3B-IT}
\end{subfigure}\hfill
\begin{subfigure}{0.24\linewidth}
    \centering
    \includegraphics[width=\linewidth,height=0.105\textheight,keepaspectratio]{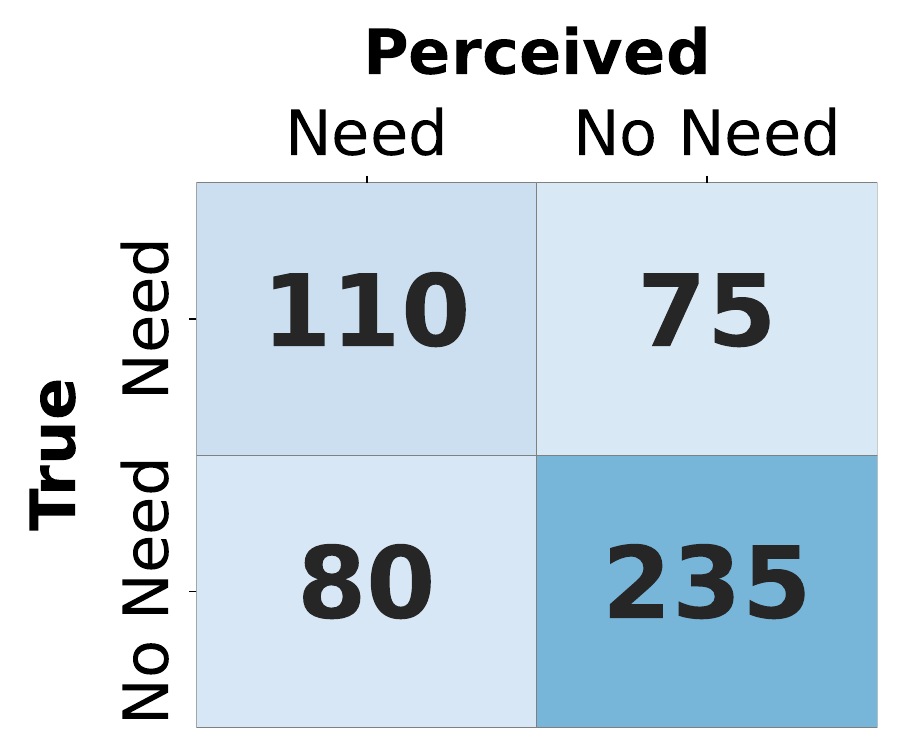}
    \caption{Llama-3.2-3B-IT}
\end{subfigure}

\textbf{(a) Perceived-need prompt v1}

\begin{subfigure}{0.25\linewidth}
    \centering
    \includegraphics[width=\linewidth,height=0.105\textheight,keepaspectratio]{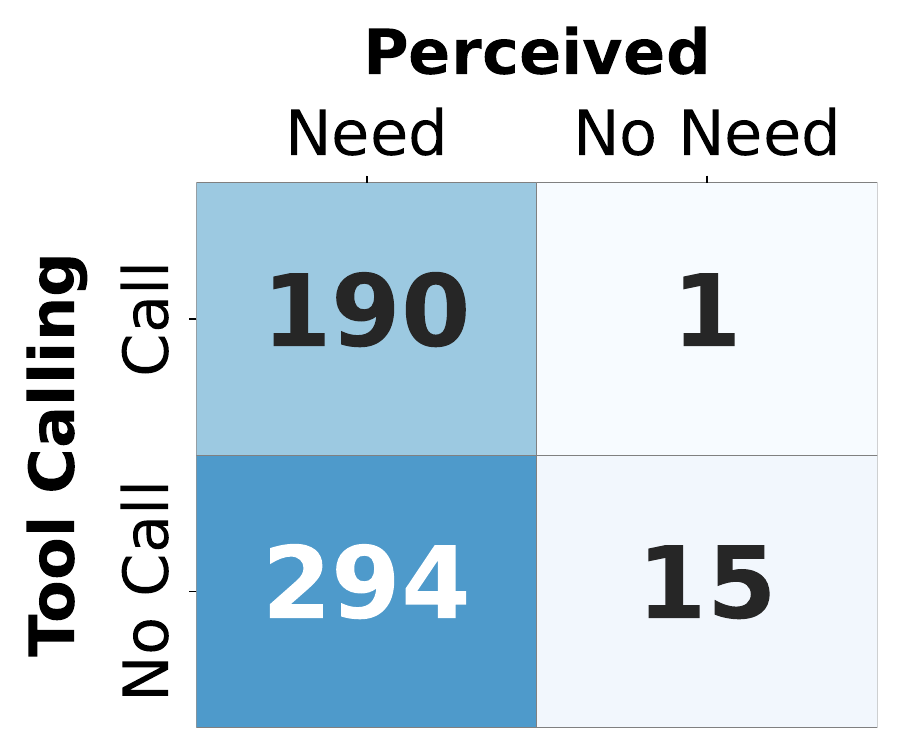}
    \caption{GPT-OSS-120B}
\end{subfigure}\hfill
\begin{subfigure}{0.25\linewidth}
    \centering
    \includegraphics[width=\linewidth,height=0.105\textheight,keepaspectratio]{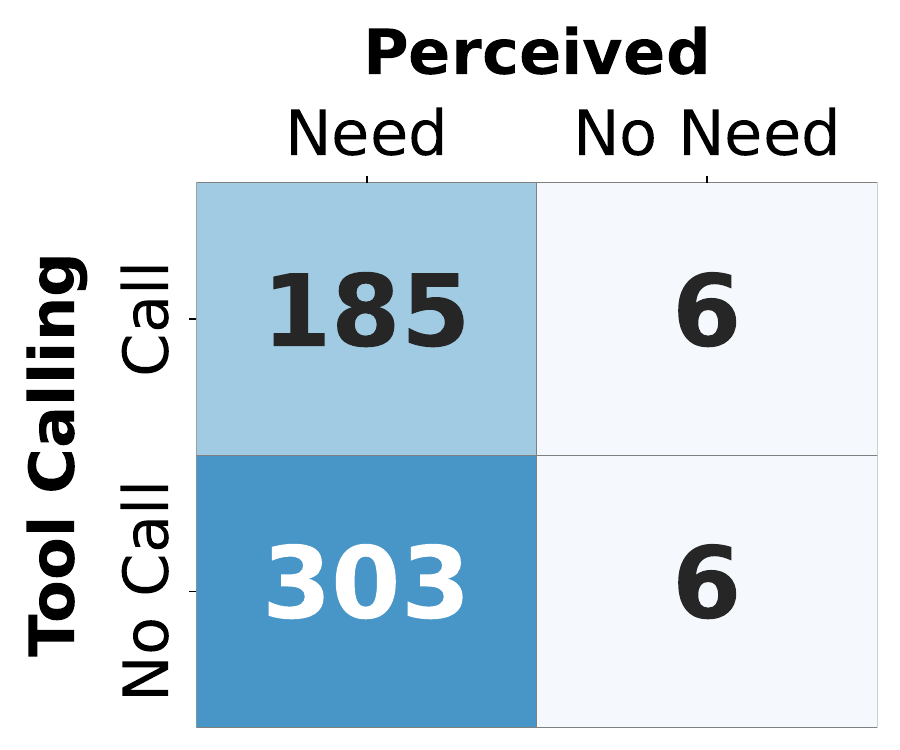}
    \caption{Qwen3-30B-A3B}
\end{subfigure}\hfill
\begin{subfigure}{0.25\linewidth}
    \centering
    \includegraphics[width=\linewidth,height=0.105\textheight,keepaspectratio]{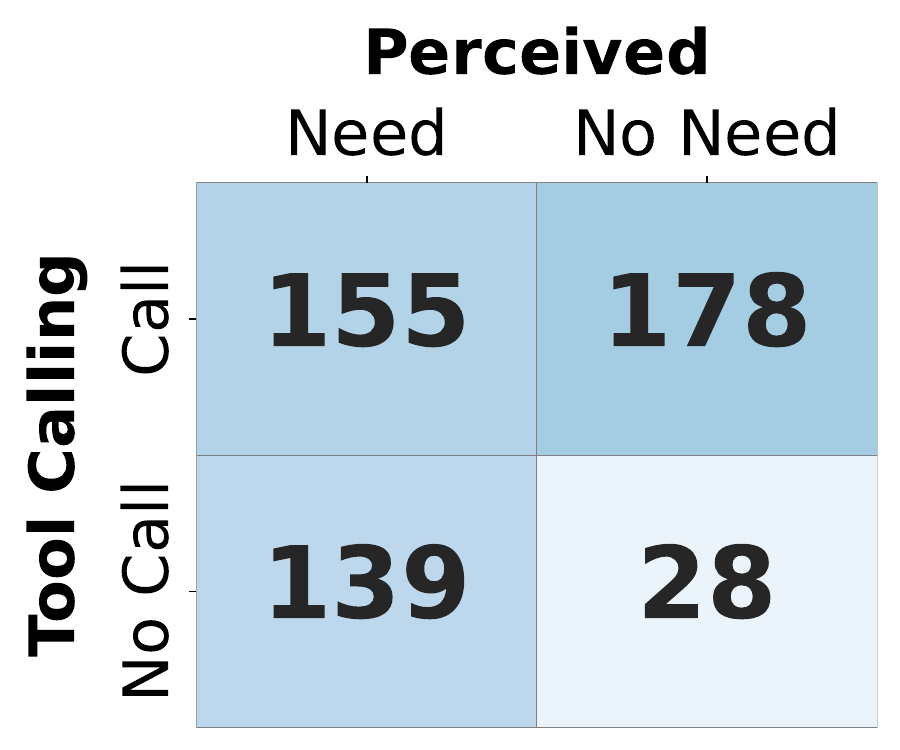}
    \caption{Qwen3-30B-IT}
\end{subfigure}\hfill
\begin{subfigure}{0.25\linewidth}
    \centering
    \includegraphics[width=\linewidth,height=0.105\textheight,keepaspectratio]{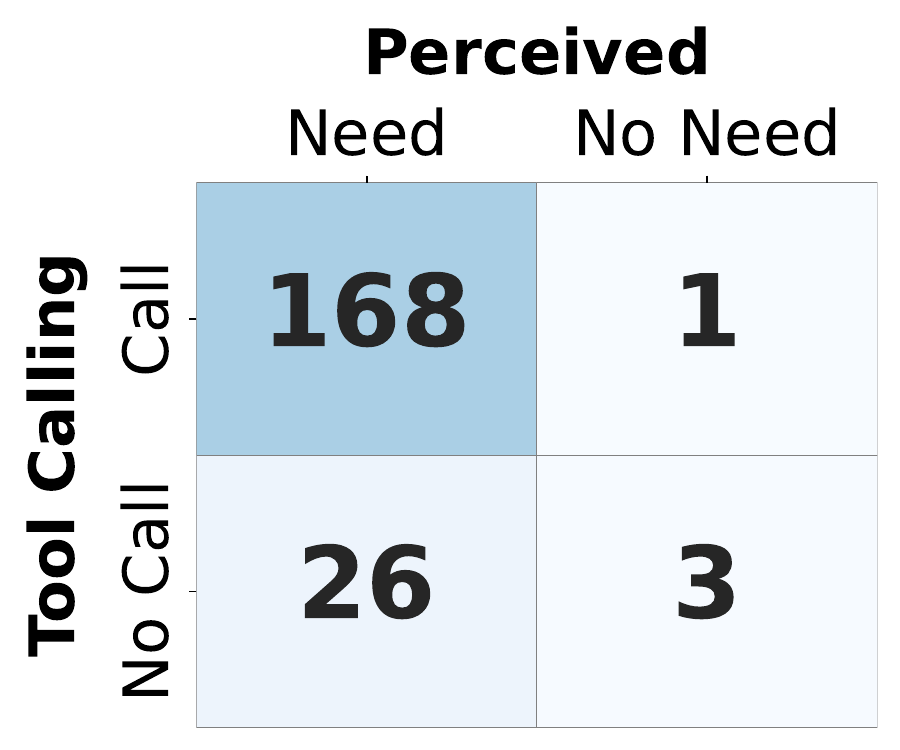}
    \caption{Gemma-3-27B-IT}
\end{subfigure}\hfill
\begin{subfigure}{0.25\linewidth}
    \centering
    \includegraphics[width=\linewidth,height=0.105\textheight,keepaspectratio]{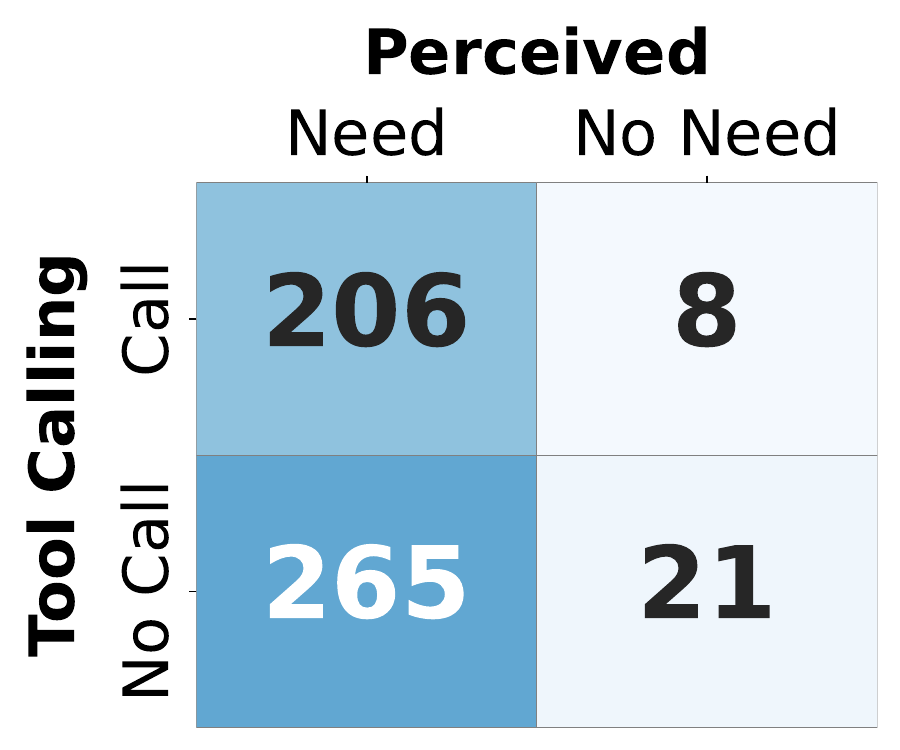}
    \caption{Mistral-3.1-24B-IT}
\end{subfigure}\hfill
\begin{subfigure}{0.25\linewidth}
    \centering
    \includegraphics[width=\linewidth,height=0.105\textheight,keepaspectratio]{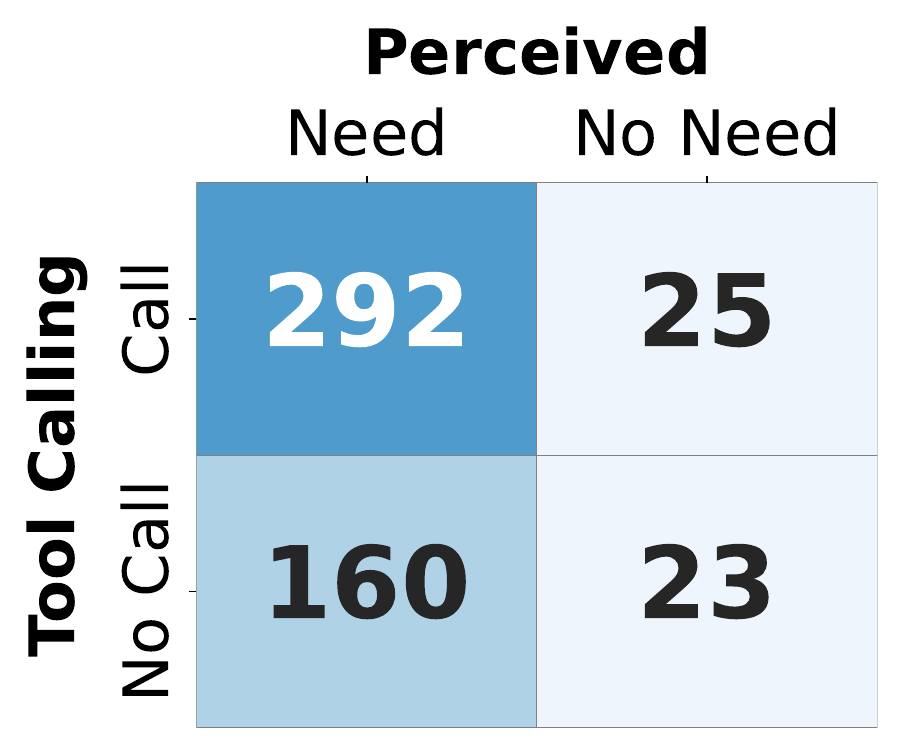}
    \caption{Llama-3.2-3B-IT}
\end{subfigure}

\textbf{(b) Perceived-need prompt v2}

\begin{subfigure}{0.25\linewidth}
    \centering
    \includegraphics[width=\linewidth,height=0.105\textheight,keepaspectratio]{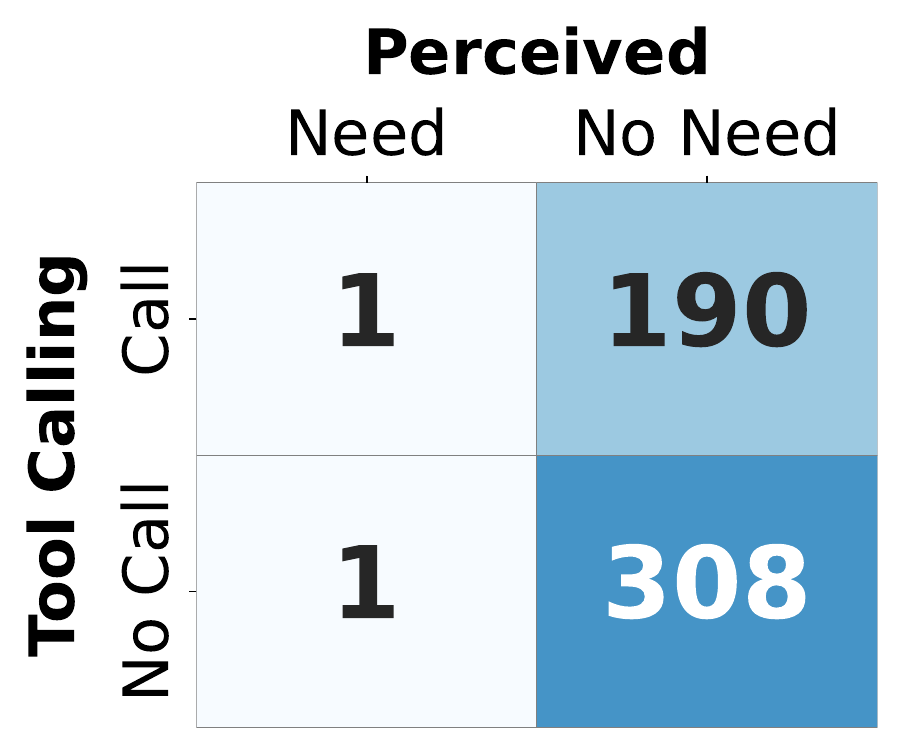}
    \caption{GPT-OSS-120B}
\end{subfigure}\hfill
\begin{subfigure}{0.25\linewidth}
    \centering
    \includegraphics[width=\linewidth,height=0.105\textheight,keepaspectratio]{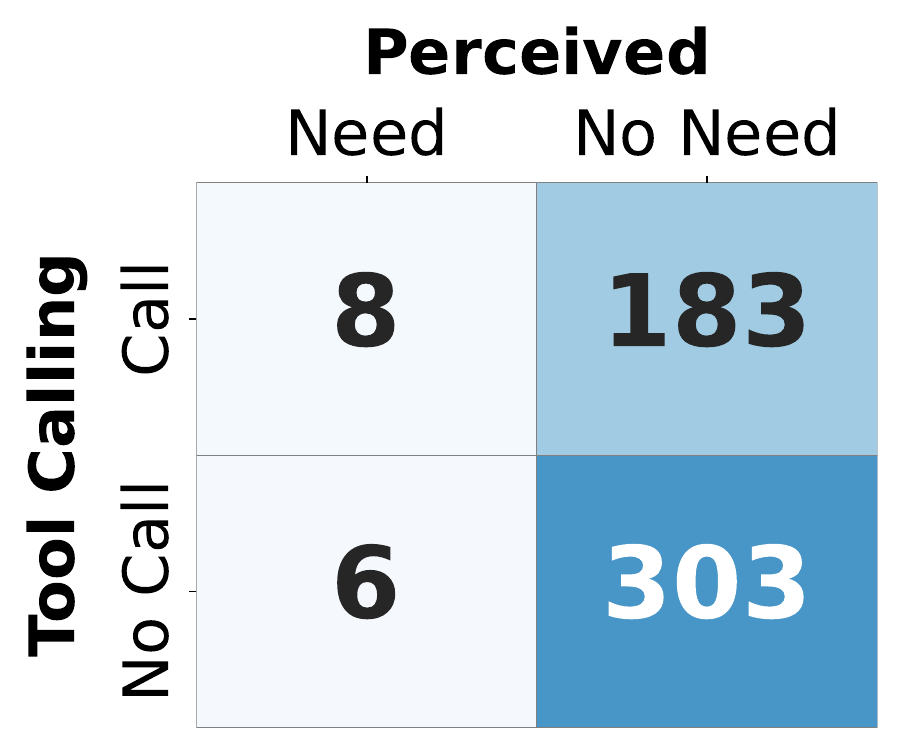}
    \caption{Qwen3-30B-A3B}
\end{subfigure}\hfill
\begin{subfigure}{0.25\linewidth}
    \centering
    \includegraphics[width=\linewidth,height=0.105\textheight,keepaspectratio]{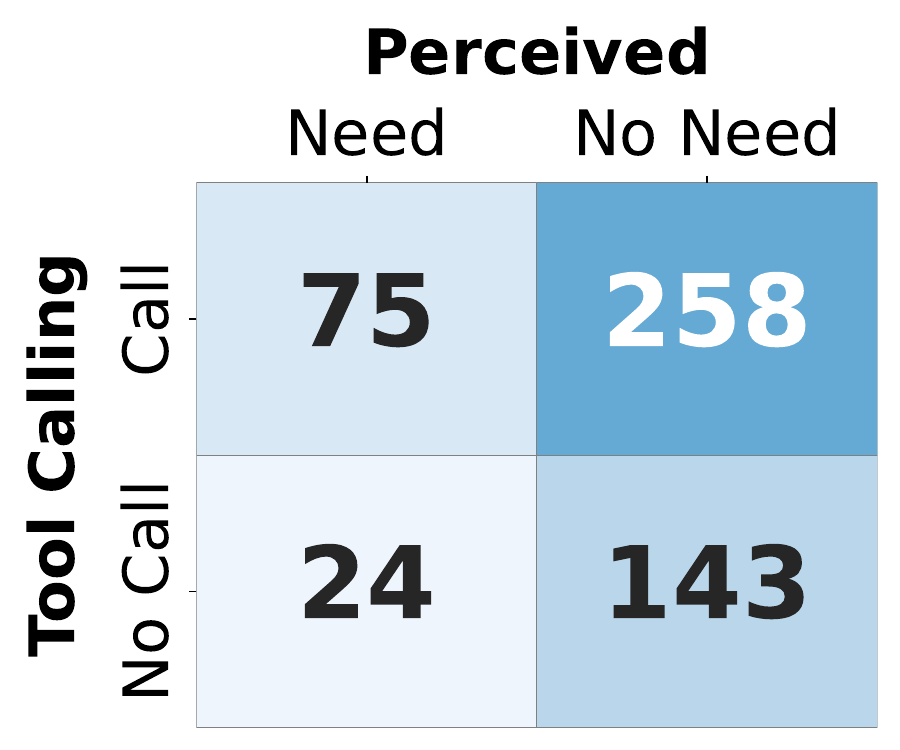}
    \caption{Qwen3-30B-IT}
\end{subfigure}\hfill
\begin{subfigure}{0.25\linewidth}
    \centering
    \includegraphics[width=\linewidth,height=0.105\textheight,keepaspectratio]{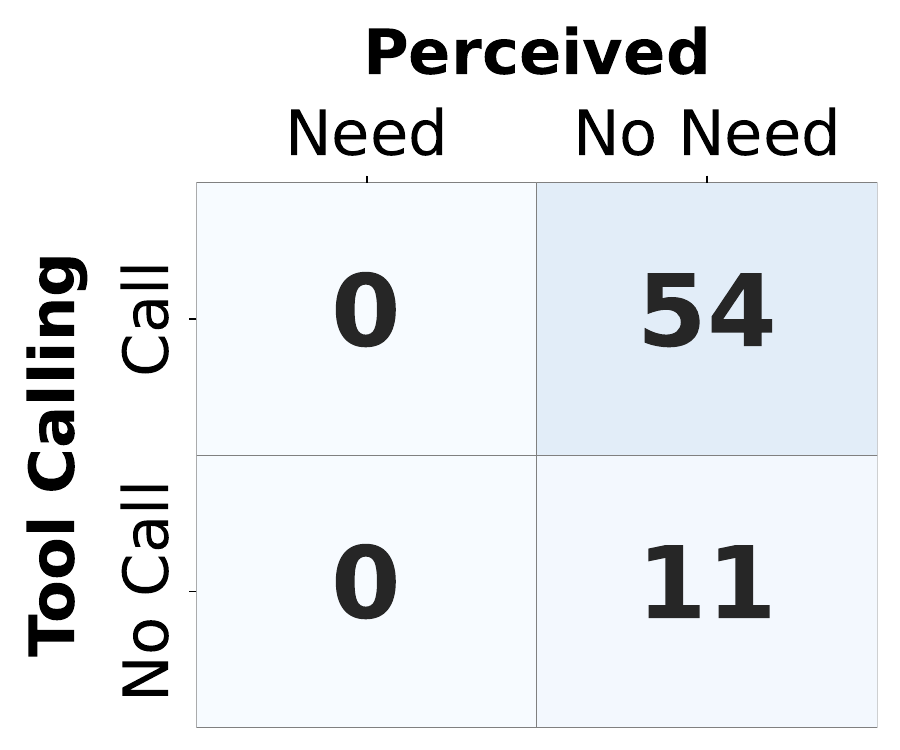}
    \caption{Gemma-3-27B-IT}
\end{subfigure}\hfill
\begin{subfigure}{0.25\linewidth}
    \centering
    \includegraphics[width=\linewidth,height=0.105\textheight,keepaspectratio]{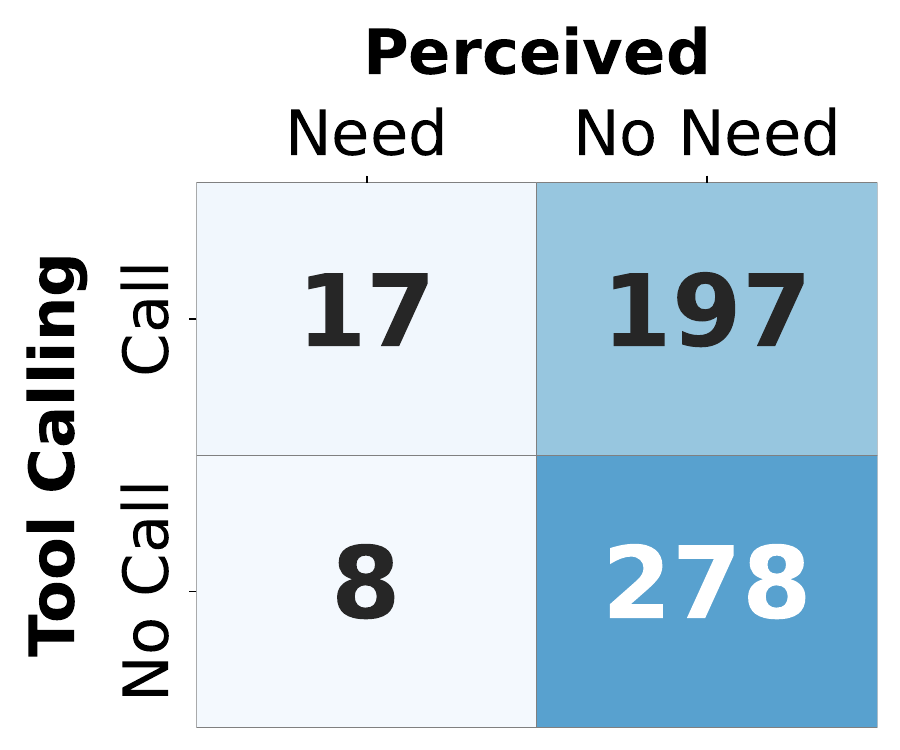}
    \caption{Mistral-3.1-24B-IT}
\end{subfigure}\hfill
\begin{subfigure}{0.25\linewidth}
    \centering
    \includegraphics[width=\linewidth,height=0.105\textheight,keepaspectratio]{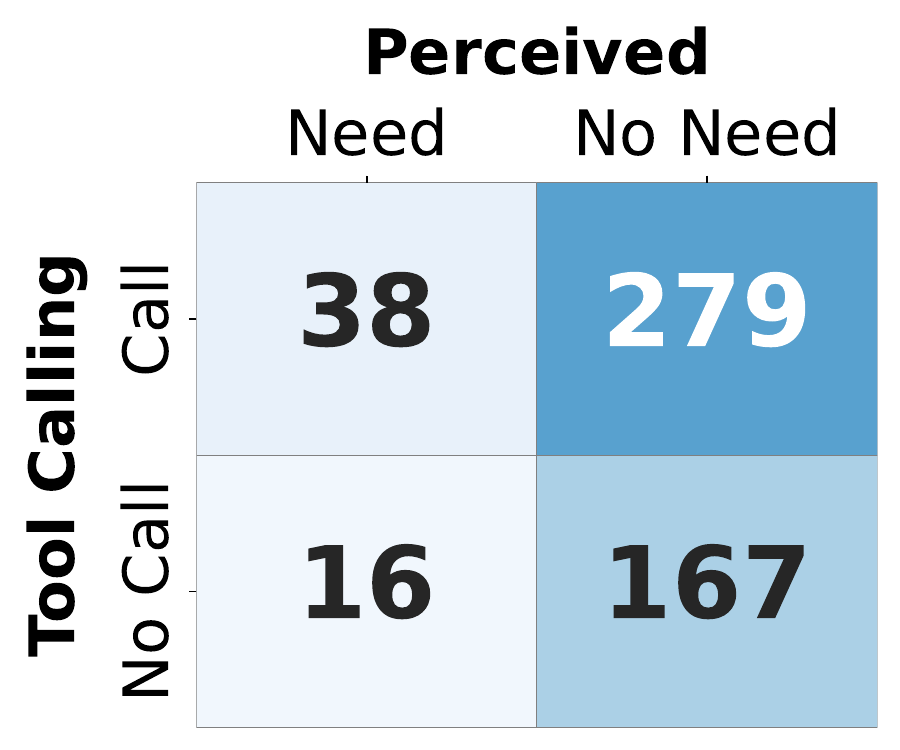}
    \caption{Llama-3.2-3B-IT}
\end{subfigure}

\textbf{(c) Perceived-need prompt v3}

\caption{\textbf{[InVivoQuery Task] perceived need is only partially aligned with tool use.} The x-axis shows perceived utility (number of entities predicted to need or not need external information), and the y-axis shows actual tool-use decisions. Percentages indicate how often the model follows its own prediction (call vs.\ not call). Results are shown for three prompt variants. Some responses are excluded due to parsing failures (i.e., missing explicit yes/no decisions), so the total count is less than 500.}

\label{fig:invivo_need_utility_combined_v12}
\vspace{-15pt}
\end{figure*}

Figure~\ref{fig:invivo_need_utility_combined_v12} shows two key observations. First, the model's self-perceived need for tool use is highly sensitive to the prompting format, where even small variations can lead to noticeably different outcomes. Second, across these variations, perceived need and utility (i.e., actual tool-calling decisions) are consistently related but not perfectly aligned. 

\begin{figure*}[t]
\centering
{\small\textbf{Need}\par}\vspace{2pt}
\begin{subfigure}{0.135\textwidth}\centering\includegraphics[width=\linewidth]{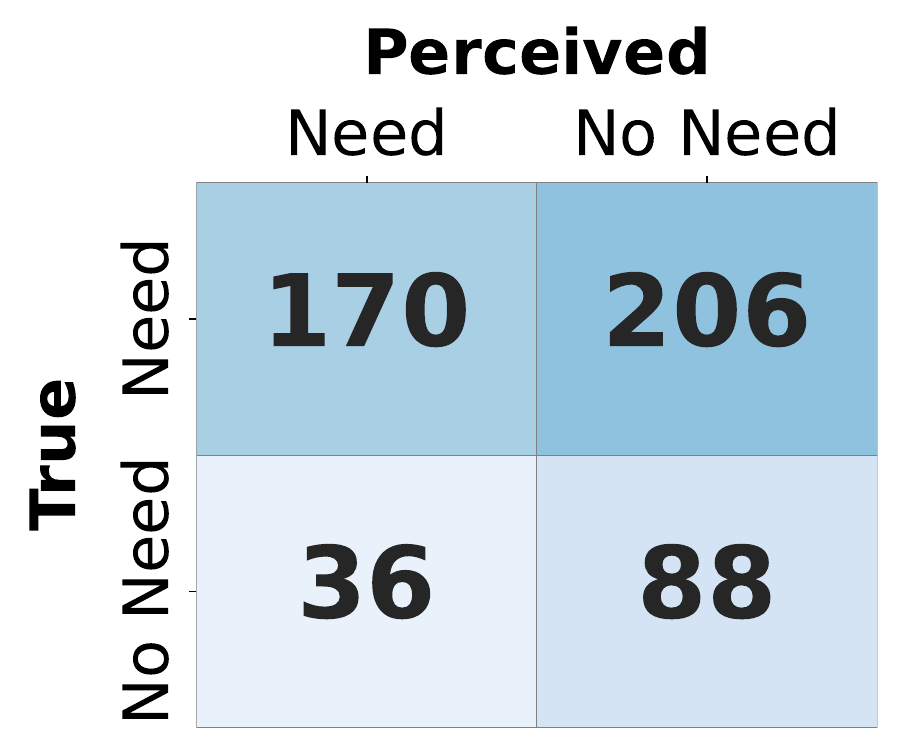}\caption{GPT-OSS-120B}\end{subfigure}\hfill
\begin{subfigure}{0.135\textwidth}\centering\includegraphics[width=\linewidth]{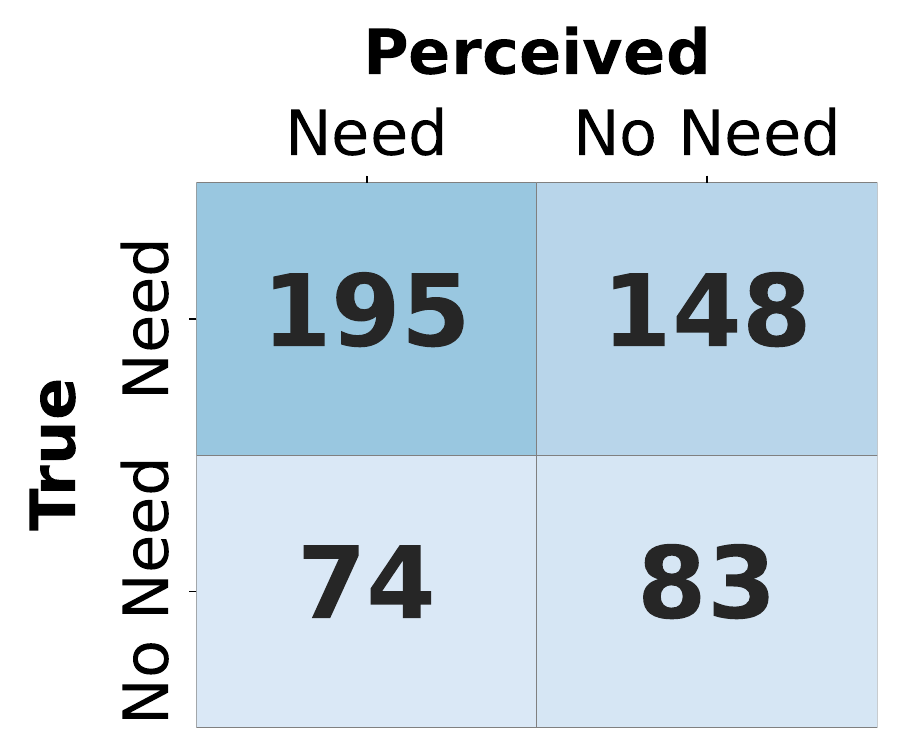}\caption{Qwen3-30B-A3B}\end{subfigure}\hfill
\begin{subfigure}{0.135\textwidth}\centering\includegraphics[width=\linewidth]{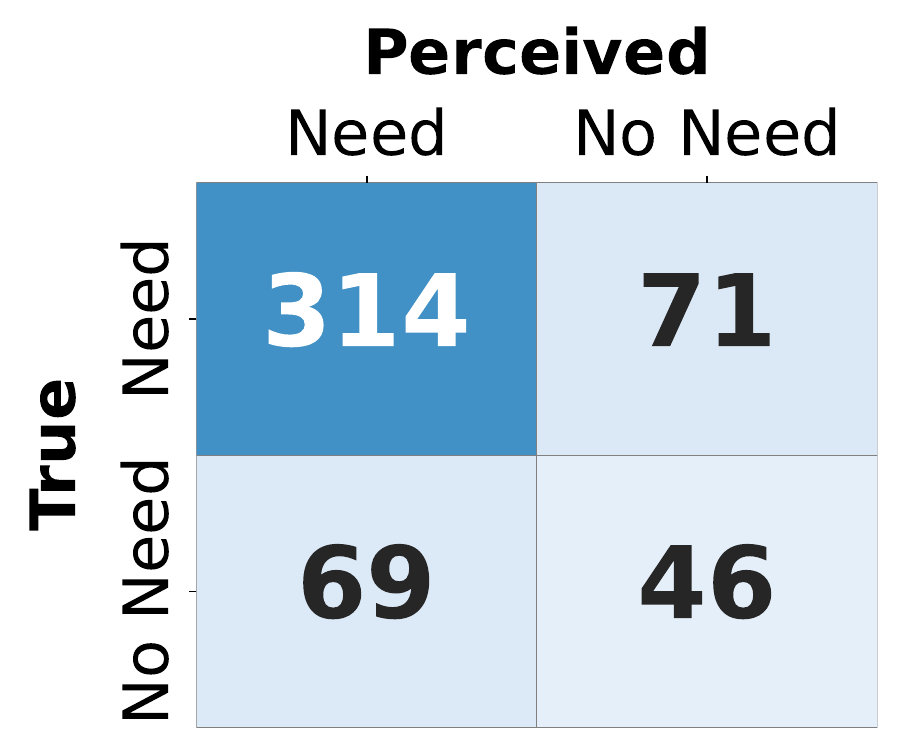}\caption{Qwen3-30B-A3B-IT}\end{subfigure}\hfill
\begin{subfigure}{0.135\textwidth}\centering\includegraphics[width=\linewidth]{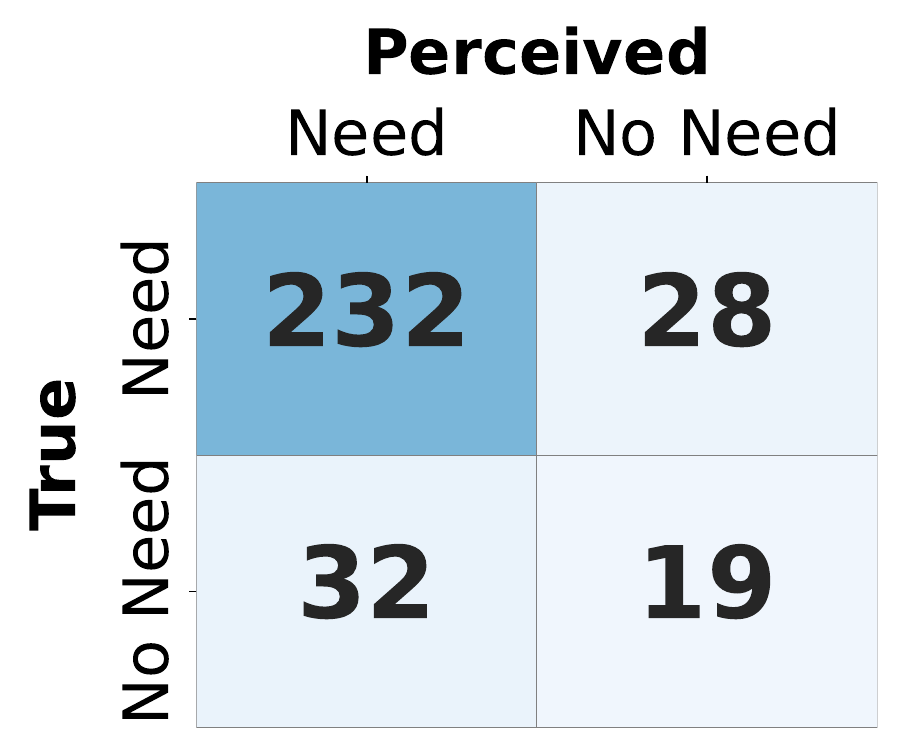}\caption{Gemma-3-27B-IT}\end{subfigure}\hfill
\begin{subfigure}{0.135\textwidth}\centering\includegraphics[width=\linewidth]{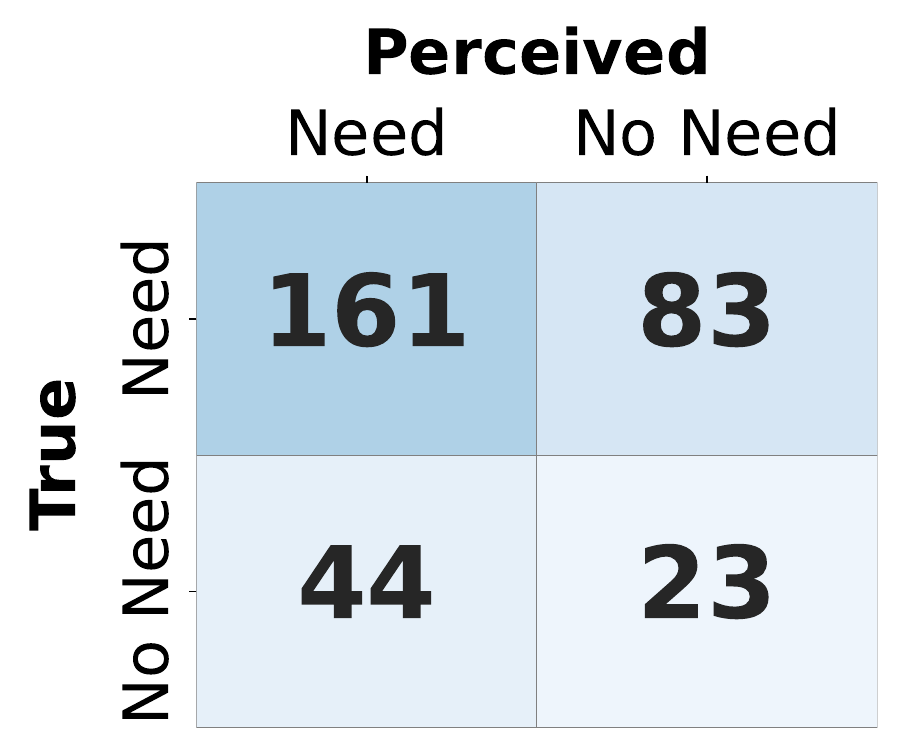}\caption{Mistral-3.1-24B-IT}\end{subfigure}\hfill
\begin{subfigure}{0.135\textwidth}\centering\includegraphics[width=\linewidth]{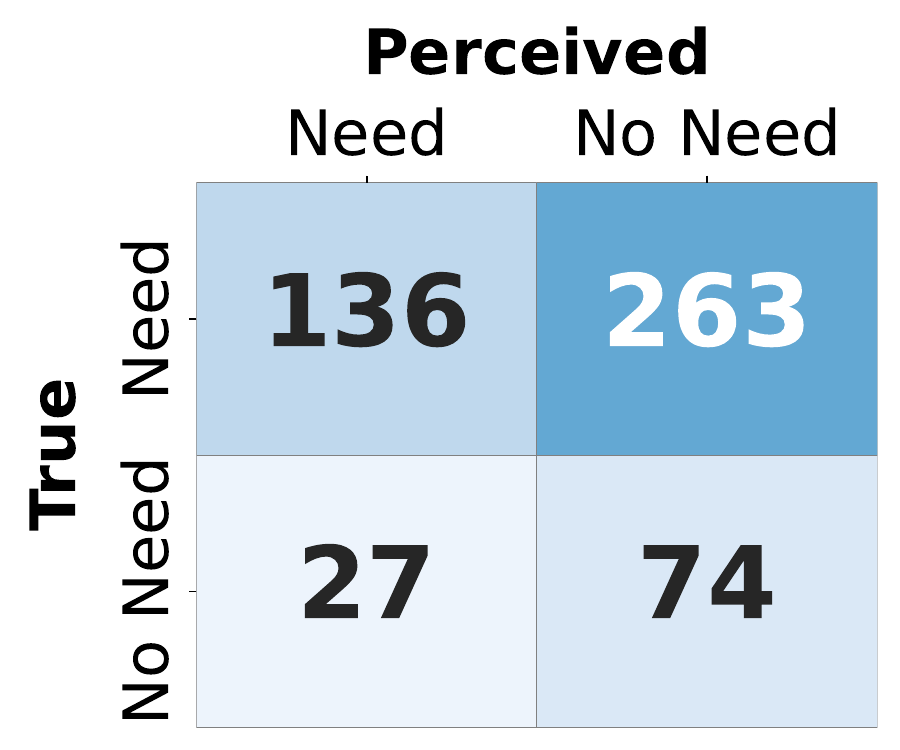}\caption{Llama-3.2-3B-IT}\end{subfigure}\hfill
\begin{subfigure}{0.135\textwidth}\centering\includegraphics[width=\linewidth]{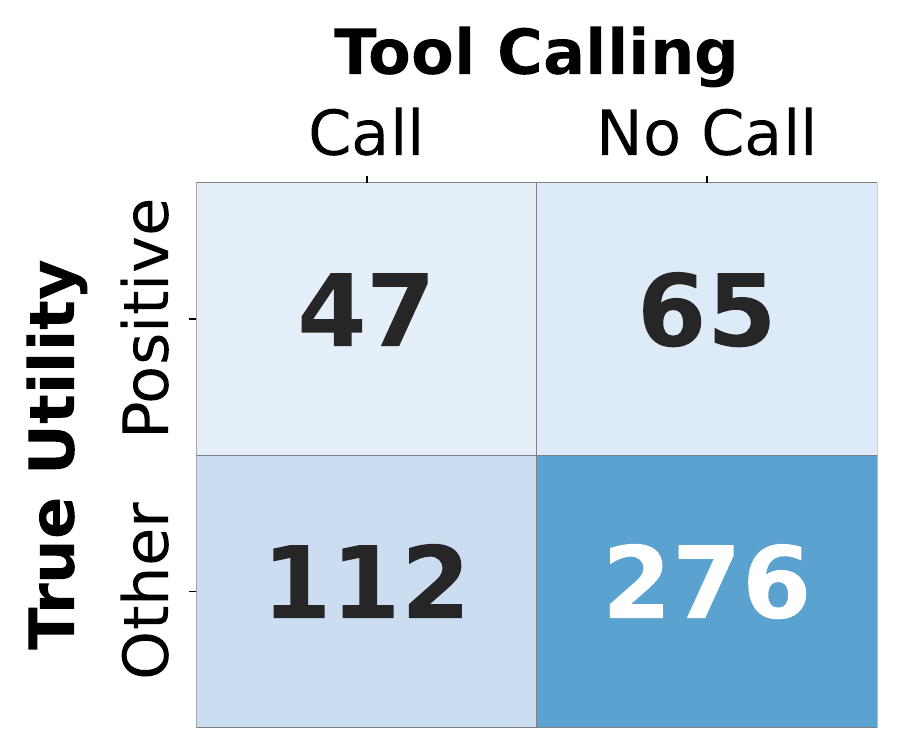}\caption{GPT-5.5}\end{subfigure}\hfill

\vspace{5pt}{\small\textbf{Utility}\par}\vspace{2pt}
\begin{subfigure}{0.135\textwidth}\centering\includegraphics[width=\linewidth]{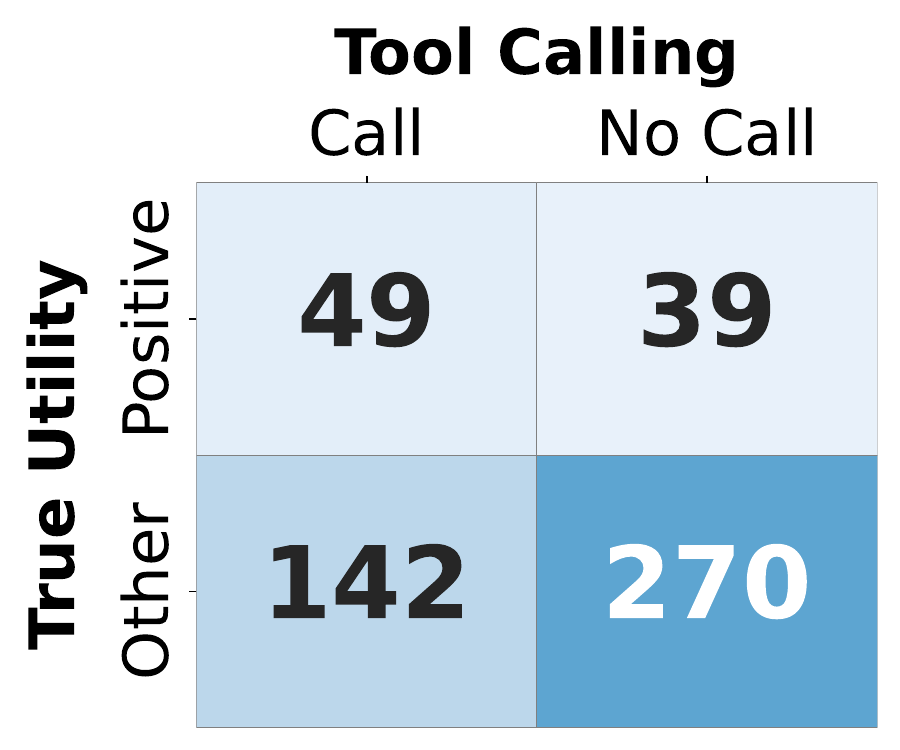}\caption{GPT-OSS-120B}\end{subfigure}\hfill
\begin{subfigure}{0.135\textwidth}\centering\includegraphics[width=\linewidth]{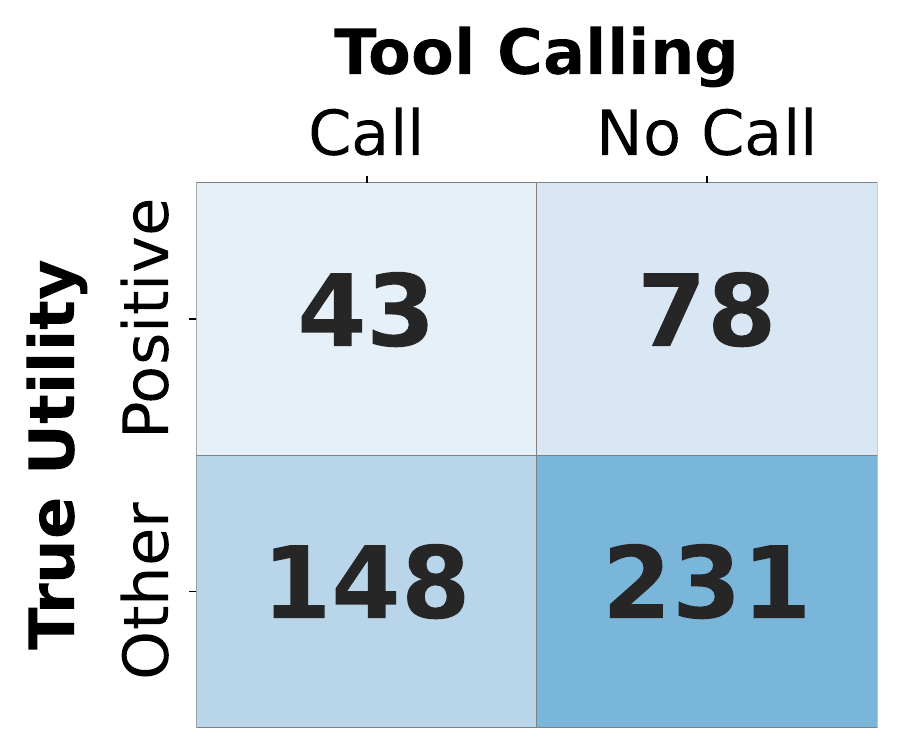}\caption{Qwen3-30B-A3B}\end{subfigure}\hfill
\begin{subfigure}{0.135\textwidth}\centering\includegraphics[width=\linewidth]{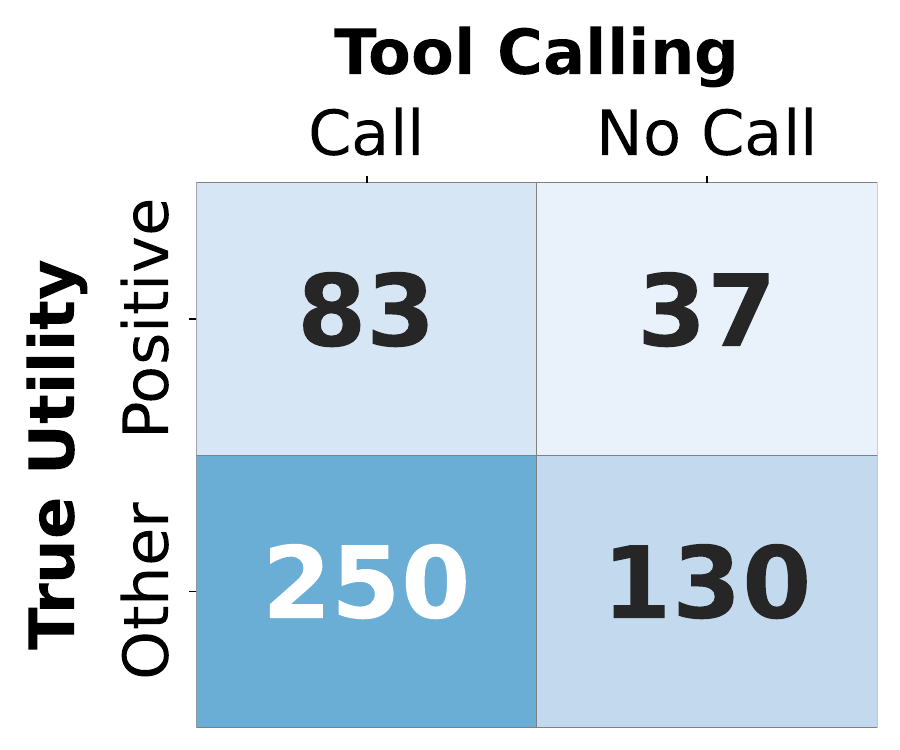}\caption{Qwen3-30B-A3B-IT}\end{subfigure}\hfill
\begin{subfigure}{0.135\textwidth}\centering\includegraphics[width=\linewidth]{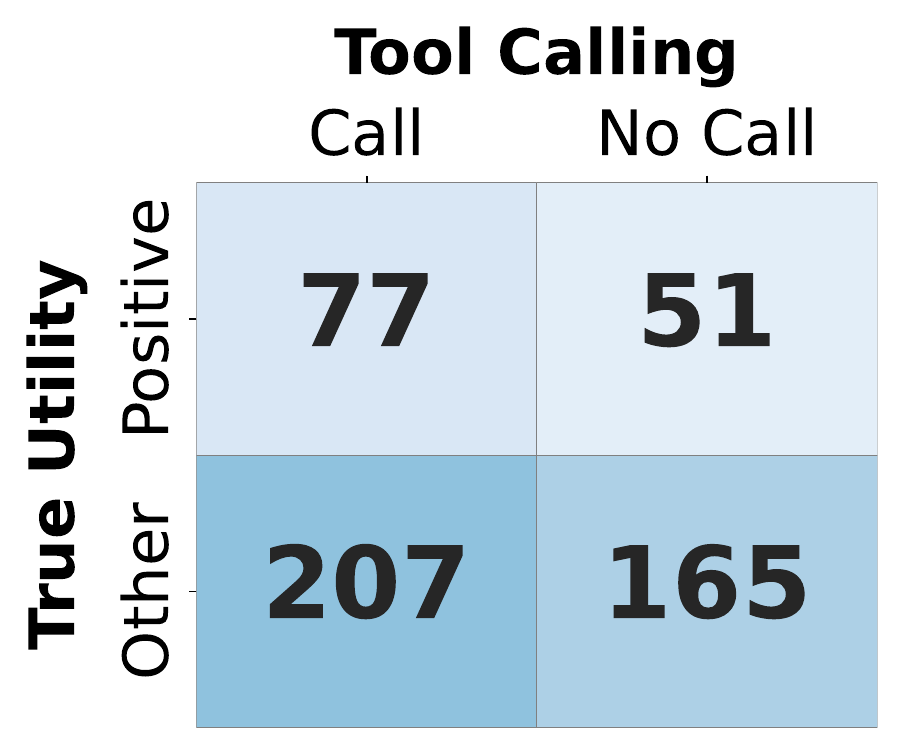}\caption{Gemma-3-27B-IT}\end{subfigure}\hfill
\begin{subfigure}{0.135\textwidth}\centering\includegraphics[width=\linewidth]{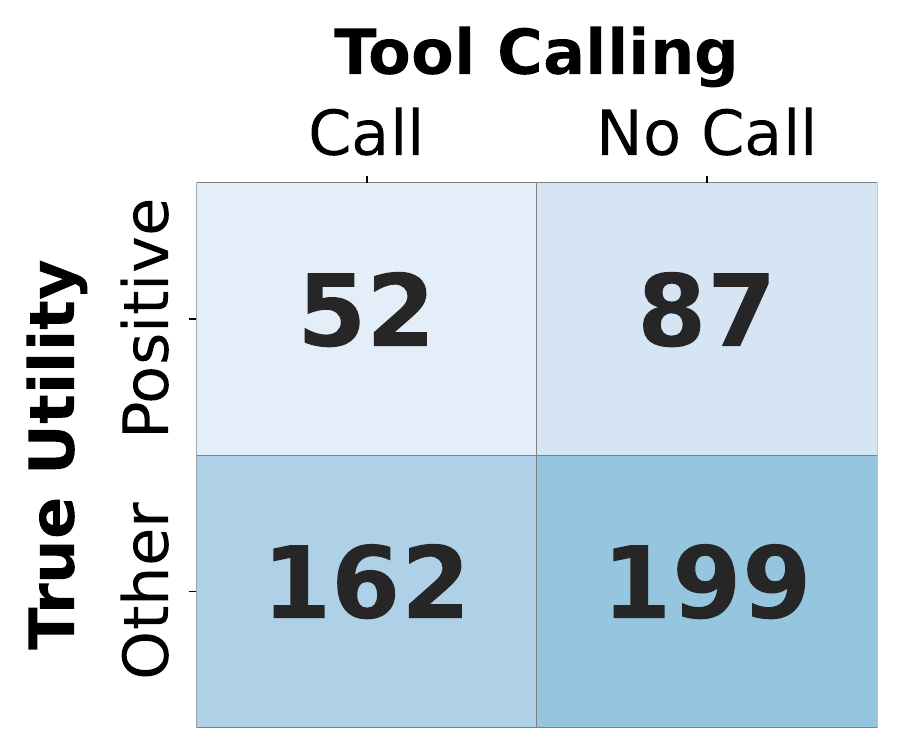}\caption{Mistral-3.1-24B-IT}\end{subfigure}\hfill
\begin{subfigure}{0.135\textwidth}\centering\includegraphics[width=\linewidth]{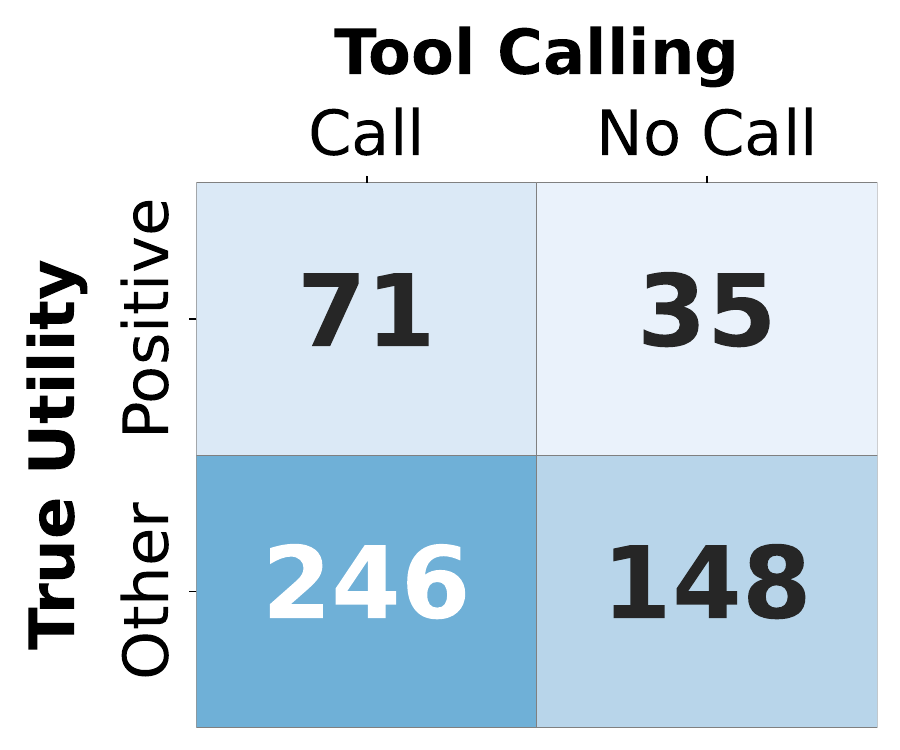}\caption{Llama-3.2-3B-IT}\end{subfigure}\hfill
\begin{subfigure}{0.135\textwidth}\centering\includegraphics[width=\linewidth]{invivo-invivo_gpt-5.5_perceived_utility_matrix.pdf}\caption{GPT-5.5}\end{subfigure}
\caption{\textbf{[InVivoQuery Task] The perceived need and utility are not aligned with the true need and utility.} Top: perceived need matrices. Bottom: true vs.\ perceived utility across models. The GPT-5.5 perceived-need panel was not available.}
\label{fig:invivo_true_perceived}
\label{fig:invivo_gpt55_true_perceived}
\end{figure*}

GPT-5.5 exhibits the same descriptive misalignment on InVivoQuery: its autonomous web-search decisions do not perfectly separate positive utility from negative or neutral utility. Its available true-utility versus perceived-utility result is included in Figure~\ref{fig:invivo_true_perceived}; a corresponding perceived-need result was not available, so we do not infer or synthesize that measurement.

\begin{figure}[t]
\vspace{-6pt}
\centering
\begin{subfigure}{\linewidth}
    \centering
    \includegraphics[width=\linewidth]{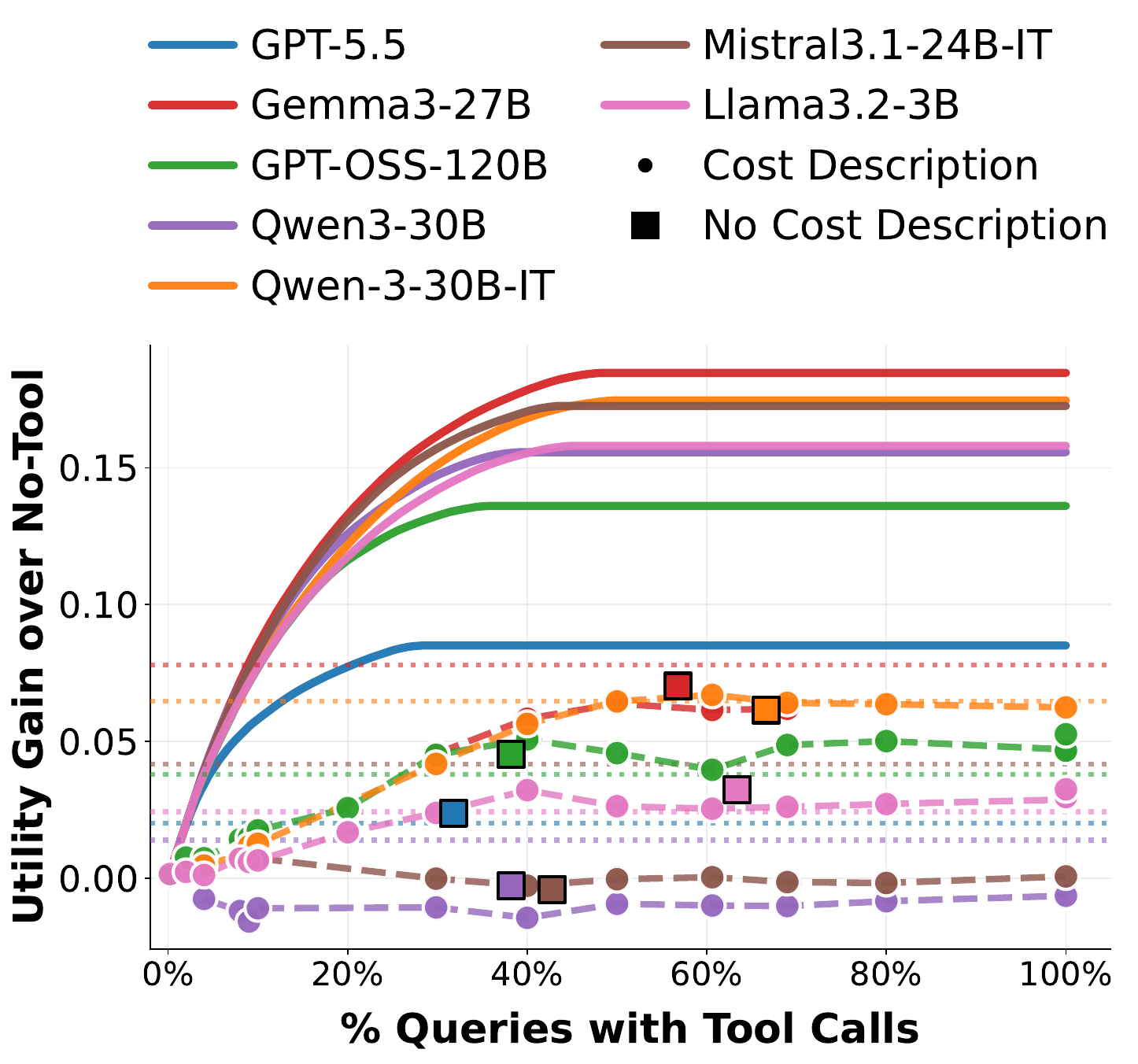}
    \caption{\textbf{Utility gain with hard stop after exceeding the budget.}}
\end{subfigure}\hfill
\begin{subfigure}{\linewidth}
    \centering
    \includegraphics[width=\linewidth]{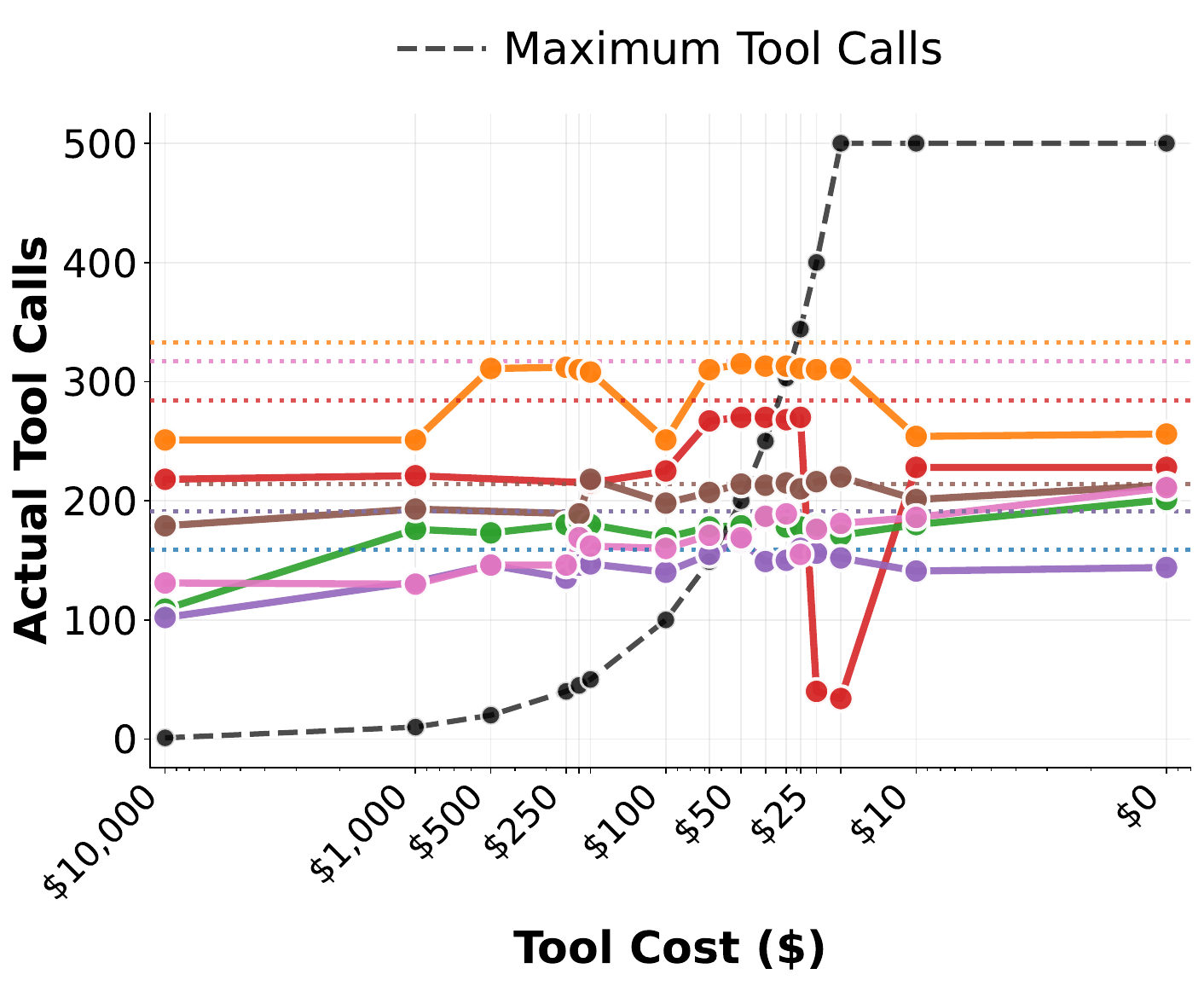}
    \caption{\textbf{Tool-calling behavior without hard stop.}}
\end{subfigure}
\caption{
\textbf{Cost-aware tool use on the Invivo Task with implicit budget notification.}
\textbf{Left:} Utility gain over the no-tool baseline under varying cost constraints. Solid lines show oracle allocation (optimal top-$k$), dashed lines show model performance with cost information, squares denote no cost-awareness, and dotted lines indicate always-calling.
\textbf{Right:} Actual tool calls without budget enforcement. Models do not reliably reduce or stop calls as cost increases, despite being provided with cost and remaining budget.
}
\label{fig:invivo_affordability_combined}
\vspace{-8pt}
\end{figure}

\begin{figure}
    \centering
    \includegraphics[width=\linewidth]{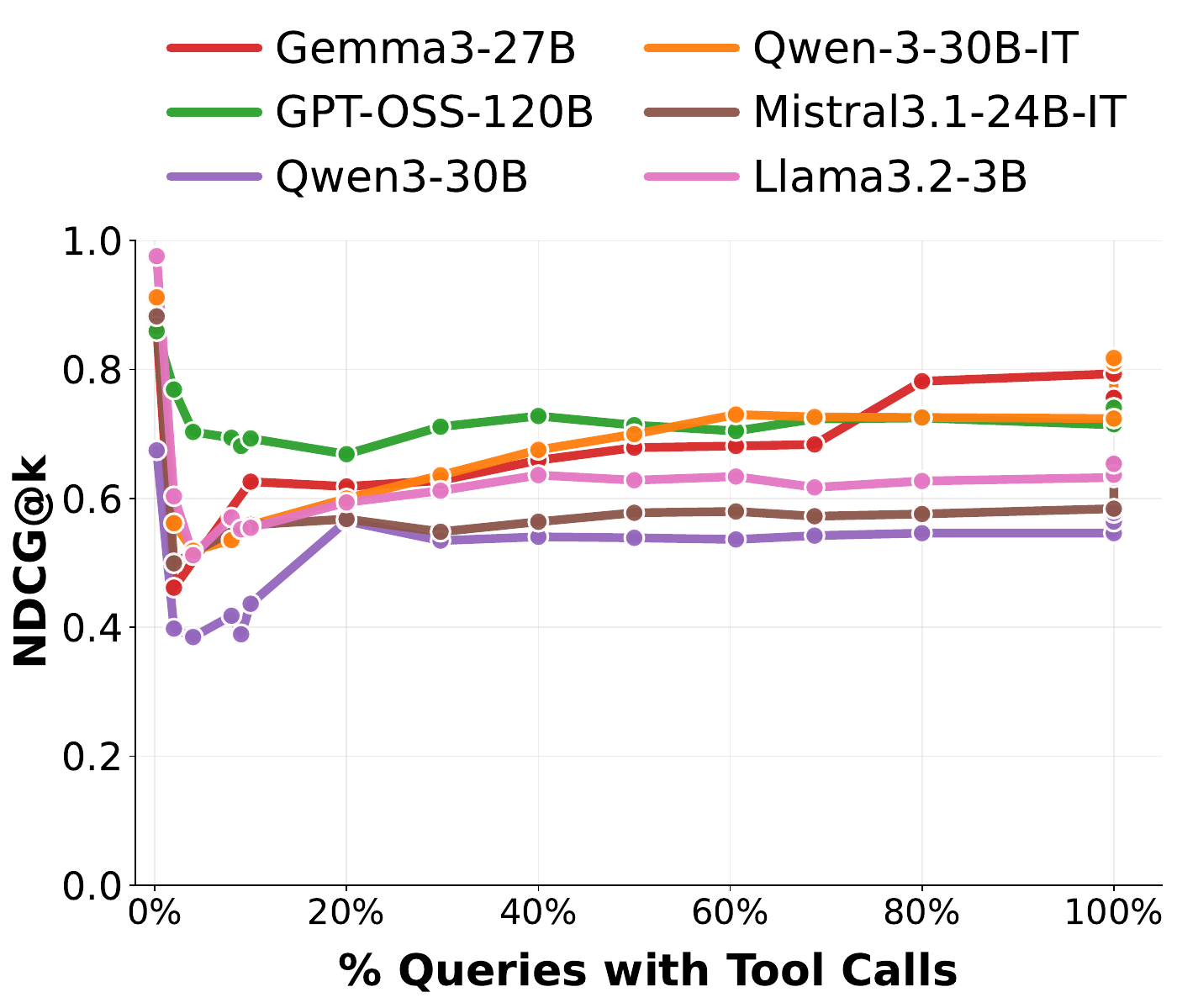}
    \caption{[InVivoQuery Task] The NDCG rank correlation under different budgets across different models. The correlation is low, which reflects that the models are not choosing the best utility gain tool calling. Cost prompt v1.}
    \label{fig:invivo-ndcg-v1}
\end{figure}

\begin{figure}[t]
\vspace{-6pt}
\centering
\begin{subfigure}{\linewidth}
    \centering
    \includegraphics[width=\linewidth]{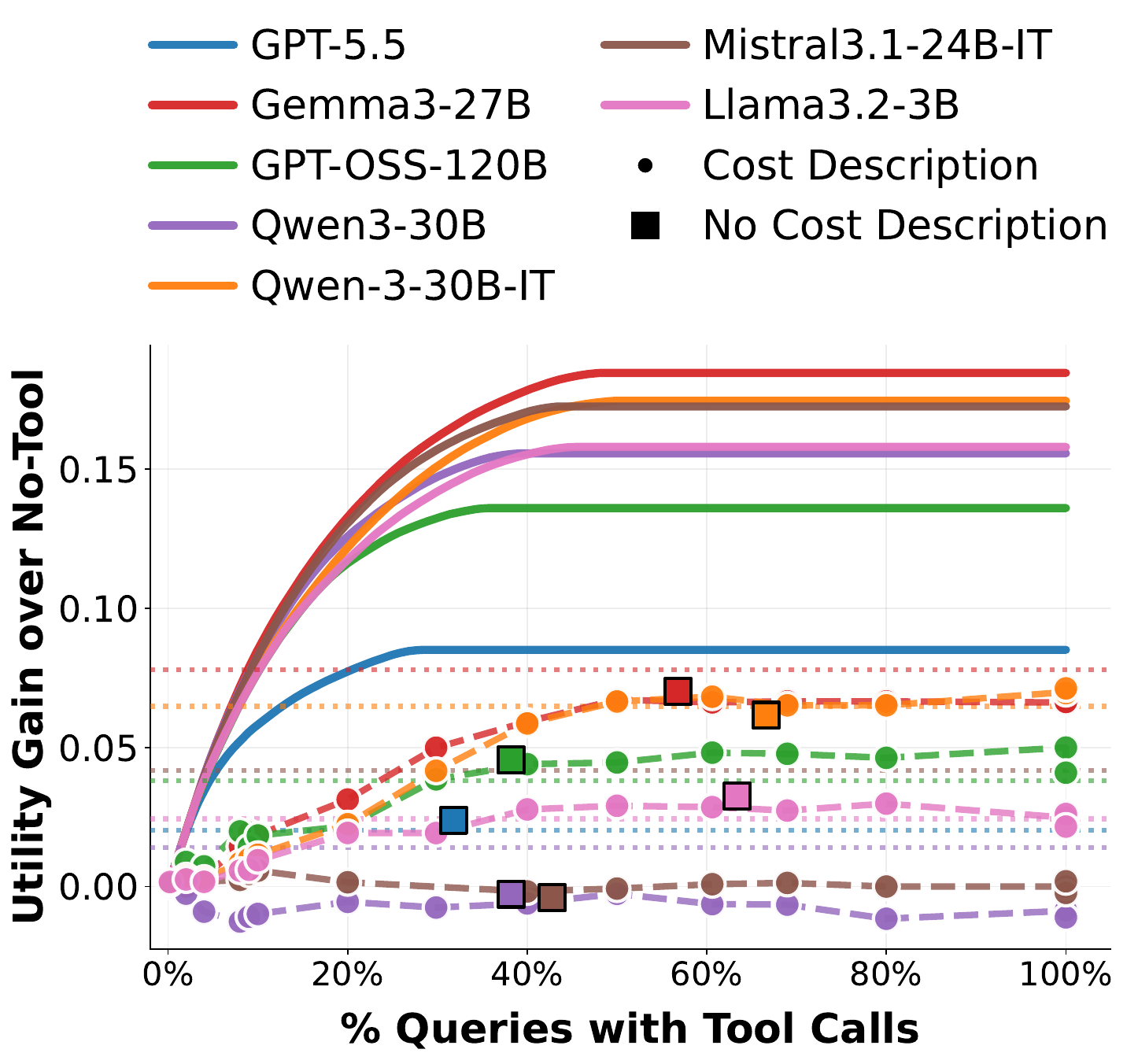}
    \caption{\textbf{Utility gain with hard stop after exceeding the budget.}}
\end{subfigure}\hfill
\begin{subfigure}{\linewidth}
    \centering
    \includegraphics[width=\linewidth]{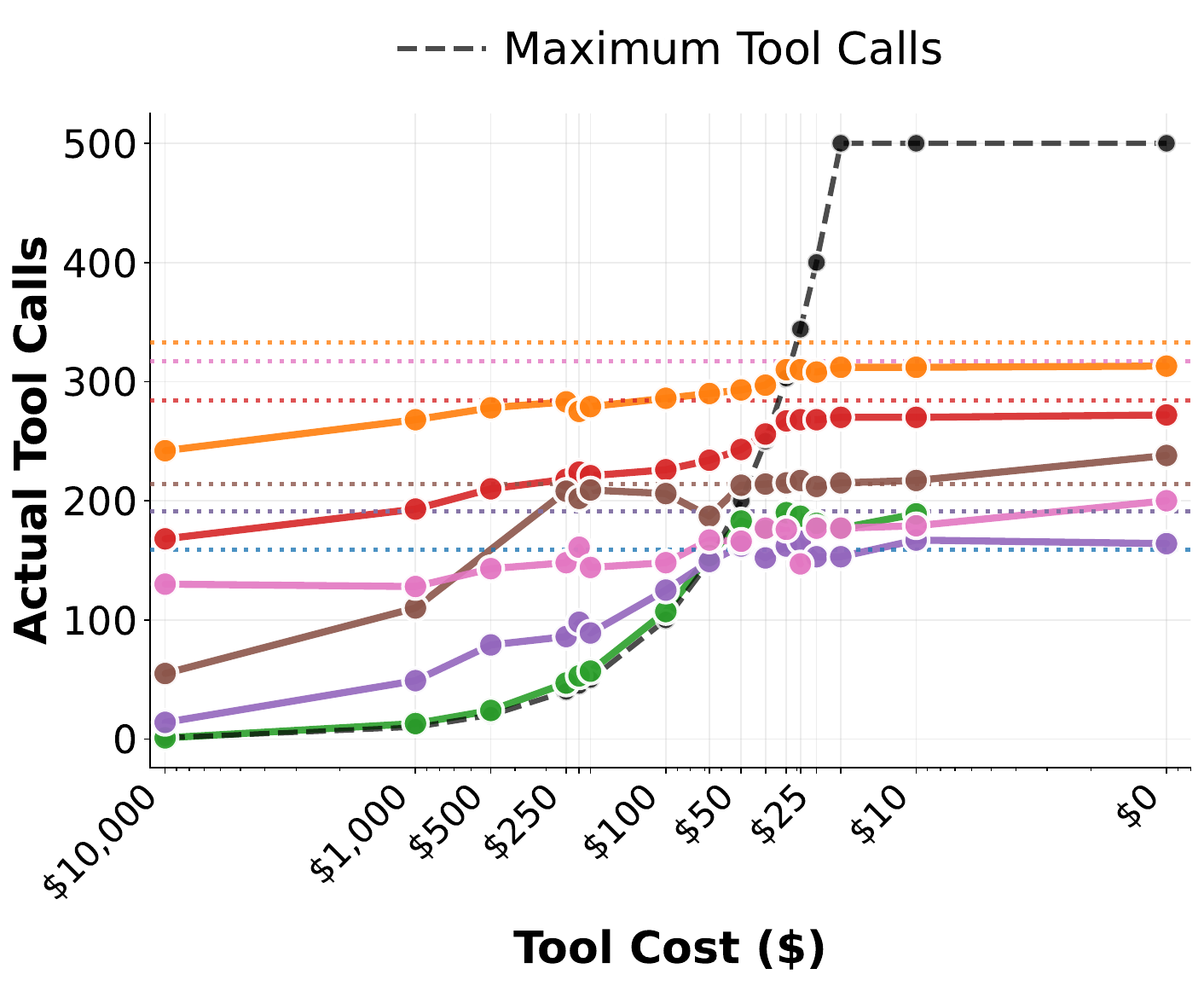}
    \caption{\textbf{Tool-calling behavior without hard stop.}}
\end{subfigure}
\caption{
\textbf{Cost-aware tool use on the Invivo Task with explicit budget notification.}
\textbf{Left:} Utility gain over the no-tool baseline under varying cost constraints. Solid lines show oracle allocation (optimal top-$k$), dashed lines show model performance with cost information, squares denote no cost-awareness, and dotted lines indicate always-calling.
\textbf{Right:} Actual tool calls without budget enforcement. Models do not reliably reduce or stop calls as cost increases, despite being provided with cost and remaining budget.
}
\label{fig:invivo_affordability_combined}
\vspace{-8pt}
\end{figure}

\begin{figure}
    \centering
    \includegraphics[width=\linewidth]{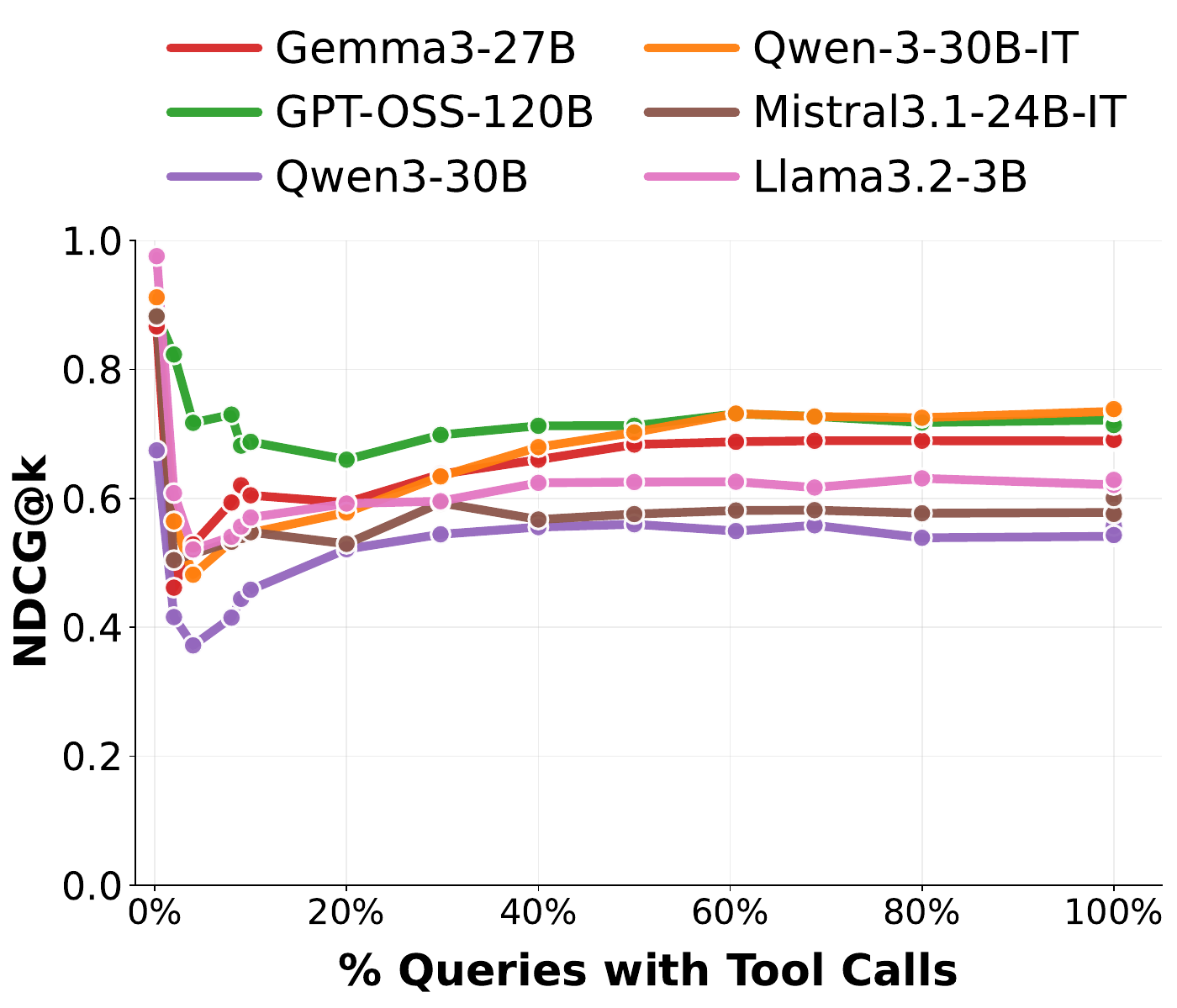}
    \caption{[InVivoQuery Task] The NDCG rank correlation under different budgets across different models. The correlation is low, which reflects that the models are not choosing the best utility gain tool calling. Cost prompt v2.}
    \label{fig:invivo-ndcg-v2}
\end{figure}

\subsection{Controller Framework}

\begin{figure}
    \centering
    \vspace{-10pt}
    \includegraphics[width=\linewidth]{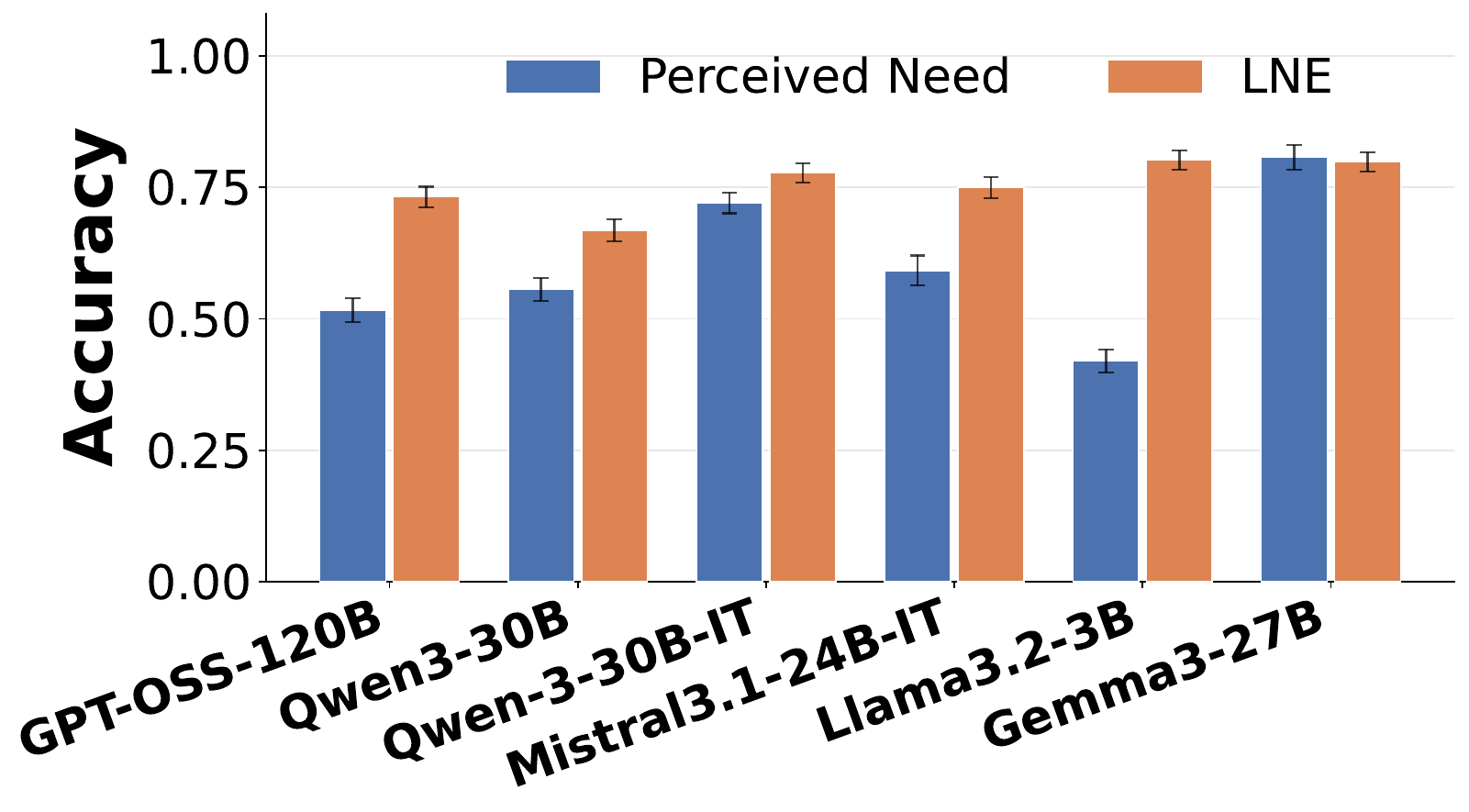}
    \caption{
       \textbf{InVivoQuery task: The LNE can predict the \textit{True Need} more accurately across most models, especially for small and weaker models.}
    }
    \label{fig: invivo_lne}
    \vspace{-10pt}
\end{figure}

\begin{figure}
    \centering

    \includegraphics[width=\linewidth]{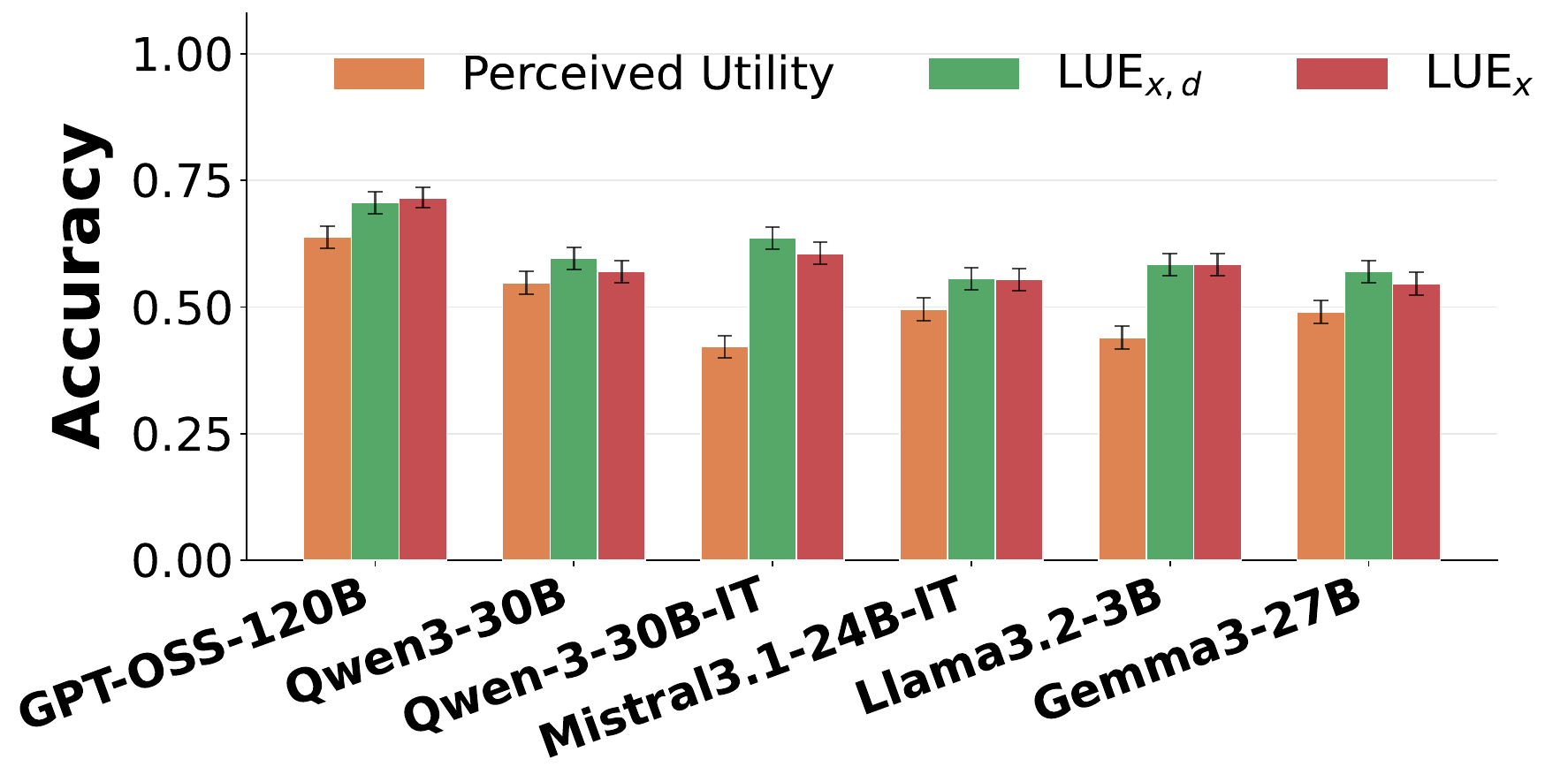}
    \caption{
       \textbf{The LUE can predict the \textit{True Utility} more accurately across most models, especially for small and weaker models.}
    }
    \label{fig: invivo_lue}

\end{figure}

Figure~\ref{fig: invivo_lue} reports the accuracy of $LUE_x$ and $LUE_{x,d_{\mathcal{F}}}$ on InVivoQuery.

\begin{figure}
    \centering
    \begin{subfigure}{\linewidth}
    \centering
    \includegraphics[width=\linewidth]{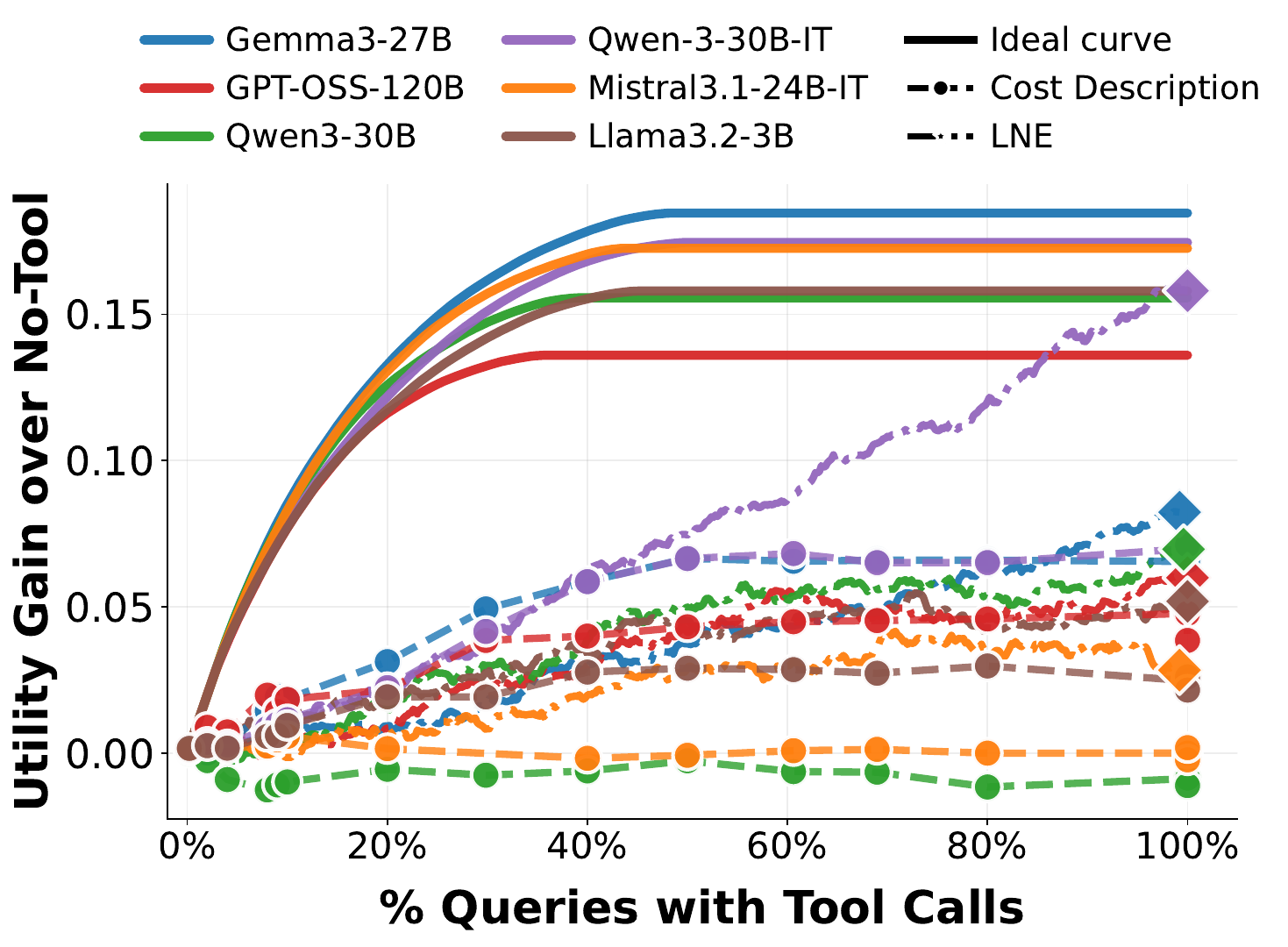}
    \caption{LNE}
    \end{subfigure}
    \begin{subfigure}{\linewidth}
    \centering
    \includegraphics[width=\linewidth]{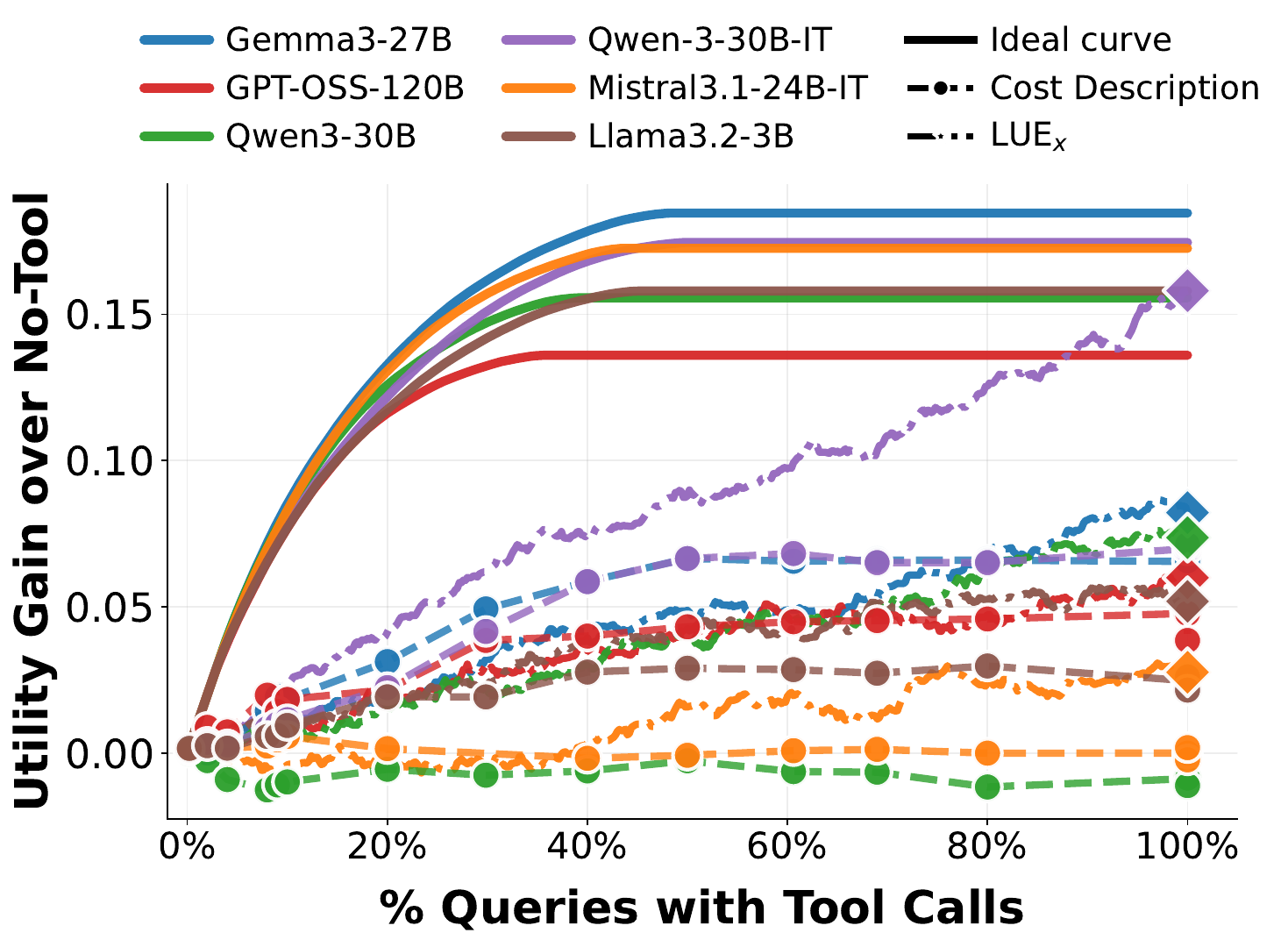}
    \caption{LUE$_x$}
    \end{subfigure}
    \begin{subfigure}{\linewidth}
    \centering
    \includegraphics[width=\linewidth]{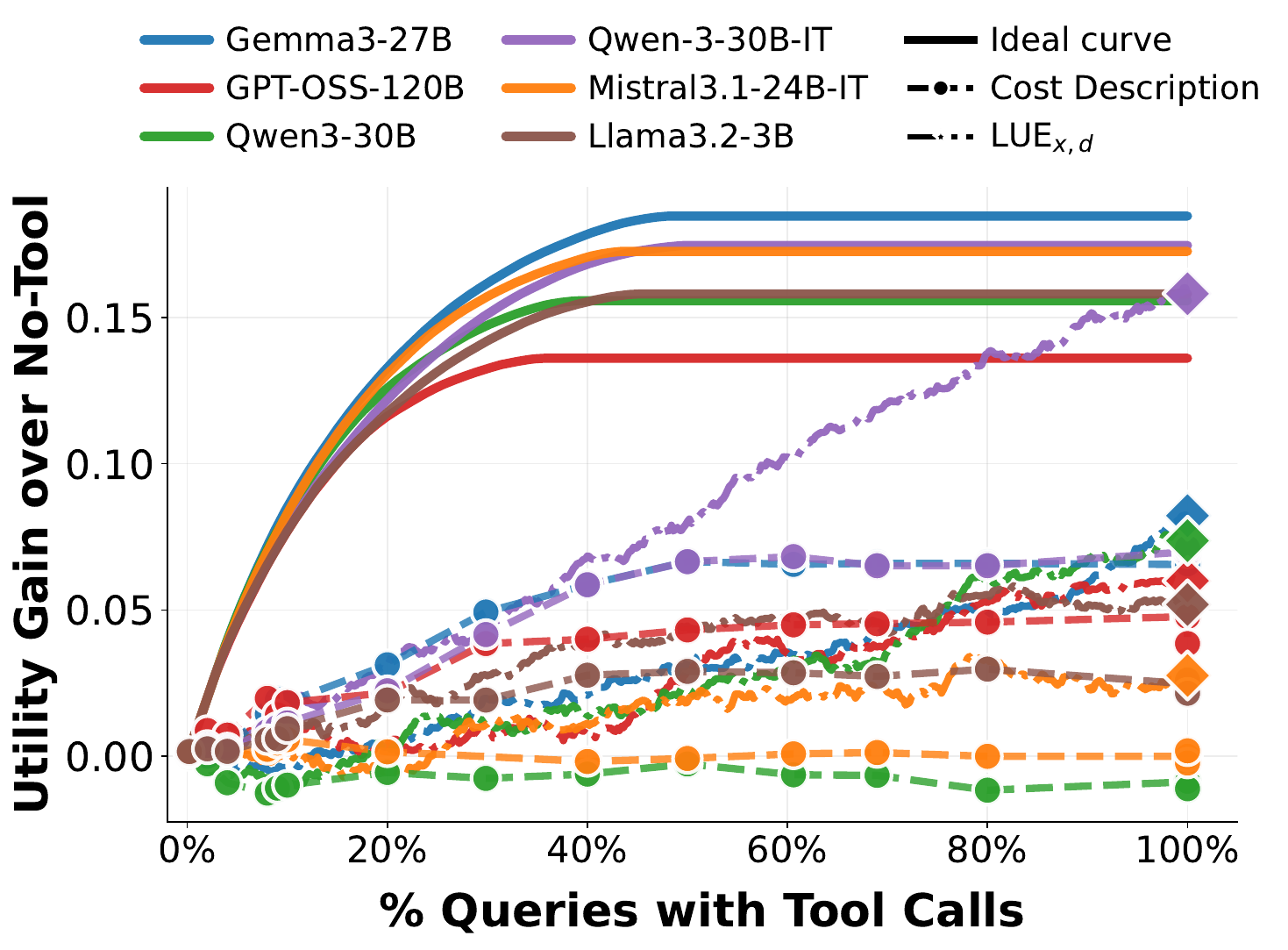}
    \caption{LUE$_{x,d}$}
    \end{subfigure}
\caption{[InVivoQuery Task] Tool-call decisions are guided by the latent need estimator’s predicted probabilities under a fixed budget constraint.}
\label{fig:prediction_cost_invivo}
\end{figure}

Tool-call decisions are guided by the latent need estimator’s predicted probabilities under a fixed budget constraint. In particular, we follow the predictor’s scores to rank instances by their likelihood of requiring tool use, and enable tool calling for the top-$k$ instances within a given budget. This strategy yields improved performance under the same budget compared to alternative allocation schemes. Figure~\ref{fig:prediction_cost_invivo} illustrates the effectiveness of this approach across different budget levels.

\section{Results for BFCL task}

In this section, we show the additional results for the BFCL task. 

\begin{figure*}
    \centering
    \includegraphics[width=\linewidth]{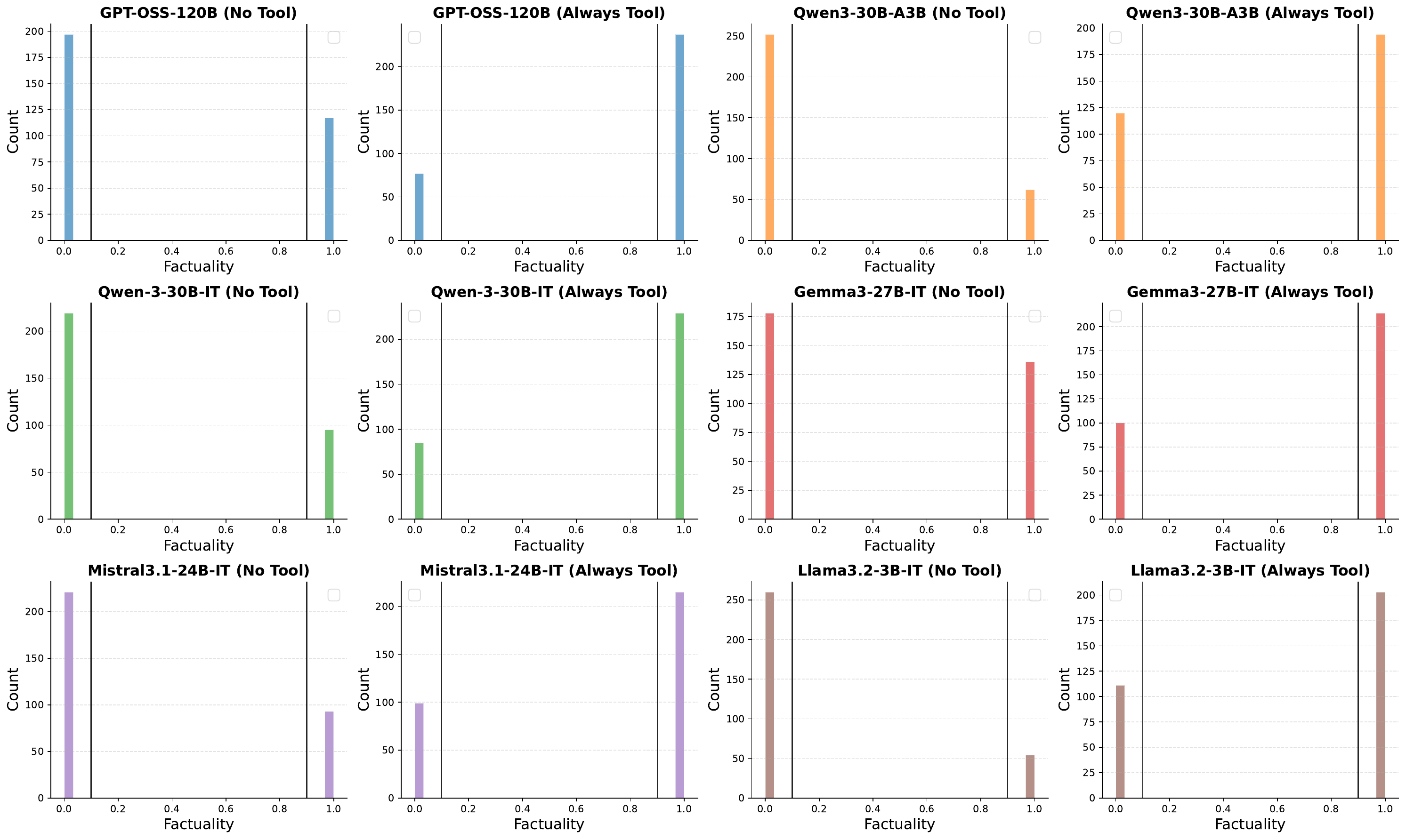}
    \caption{\textbf{[BFCL Task] factuality distribution across different models.}}
    \label{fig:bfcl_hist}
\end{figure*}

In Figure~\ref{fig:bfcl_hist}, we show the factuality score distribution across all the models and entities for the BFCL Task. Visualizing the distribution is important because aggregate metrics alone (e.g., mean or accuracy) can obscure underlying differences in model behavior. The distribution provides a more fine-grained view of how factuality scores are spread, revealing patterns such as skewness, variance, and the presence of extreme cases.
In particular, this figure allows us to examine how factuality shifts when tool use is enabled versus disabled. Rather than only observing average improvements, the distribution highlights whether gains are consistent across samples or driven by a subset of cases. 

\subsection{Normative Lens}

\begin{figure*}[t]
\centering

\begin{subfigure}{0.32\linewidth}
    \centering
    \includegraphics[width=\linewidth]{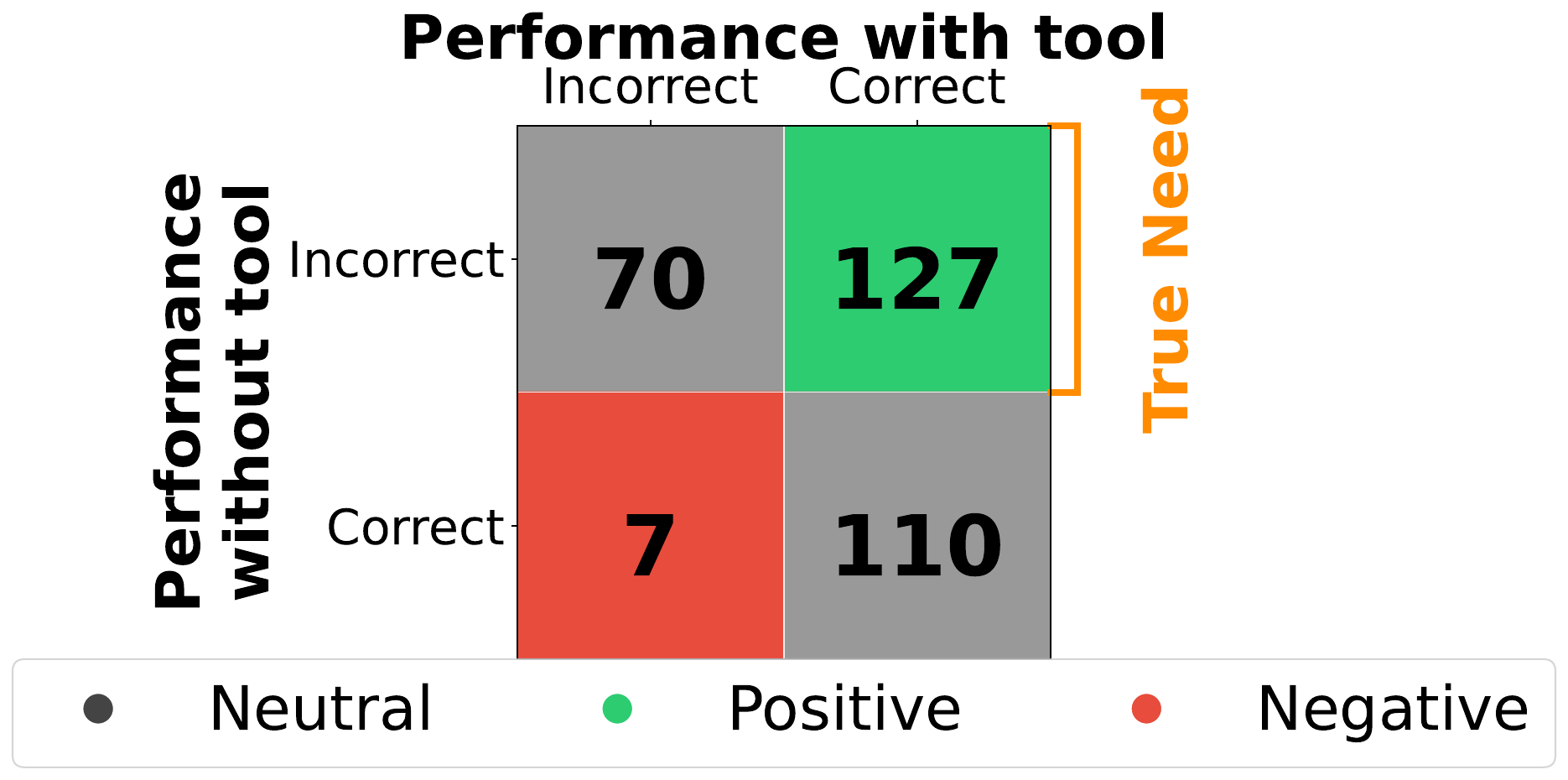}
    \caption{GPT-OSS-120B}
\end{subfigure}
\hfill
\begin{subfigure}{0.32\linewidth}
    \centering
    \includegraphics[width=\linewidth]{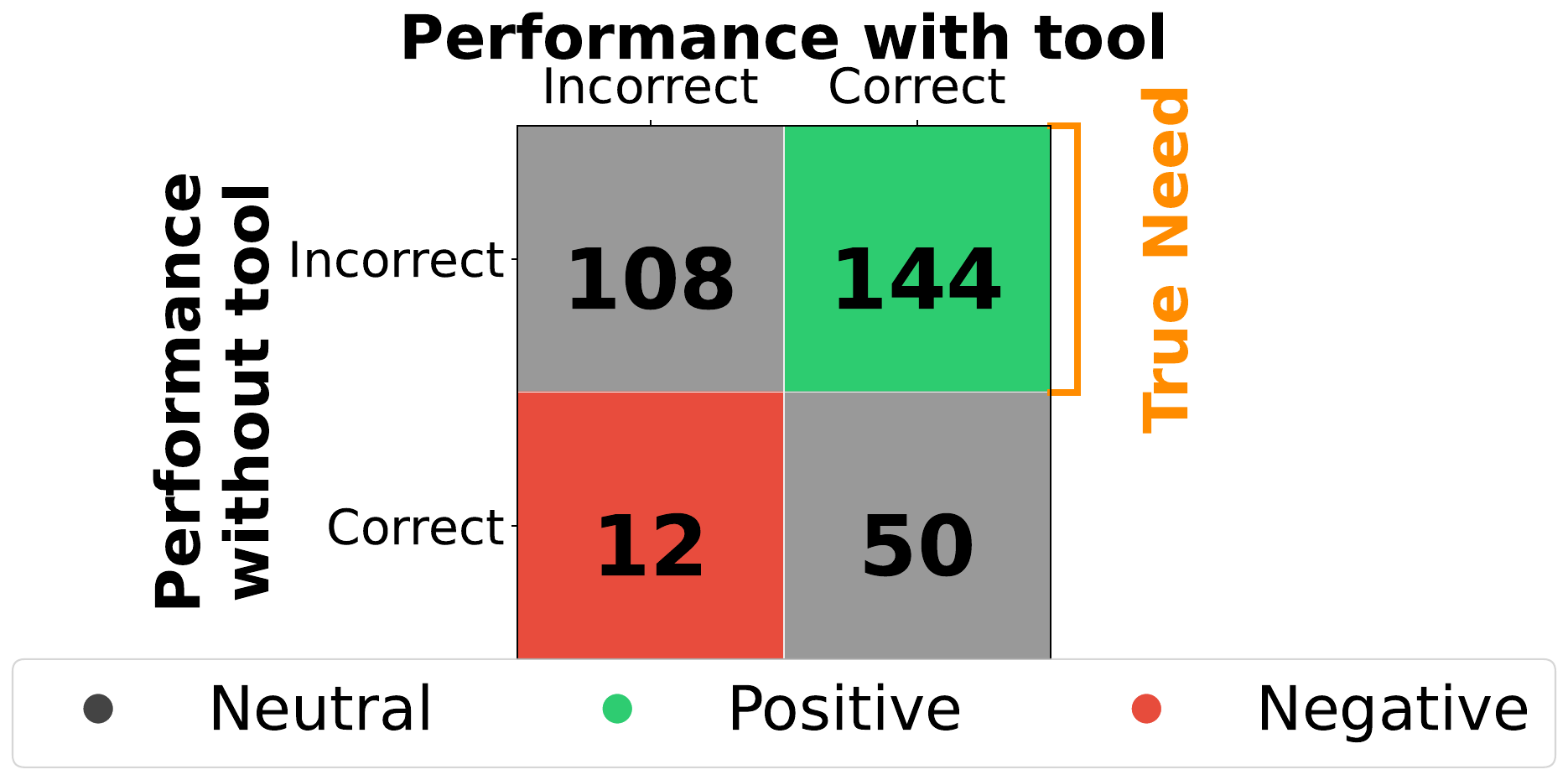}
    \caption{Qwen3-30B-A3B}
\end{subfigure}
\hfill
\begin{subfigure}{0.32\linewidth}
    \centering
    \includegraphics[width=\linewidth]{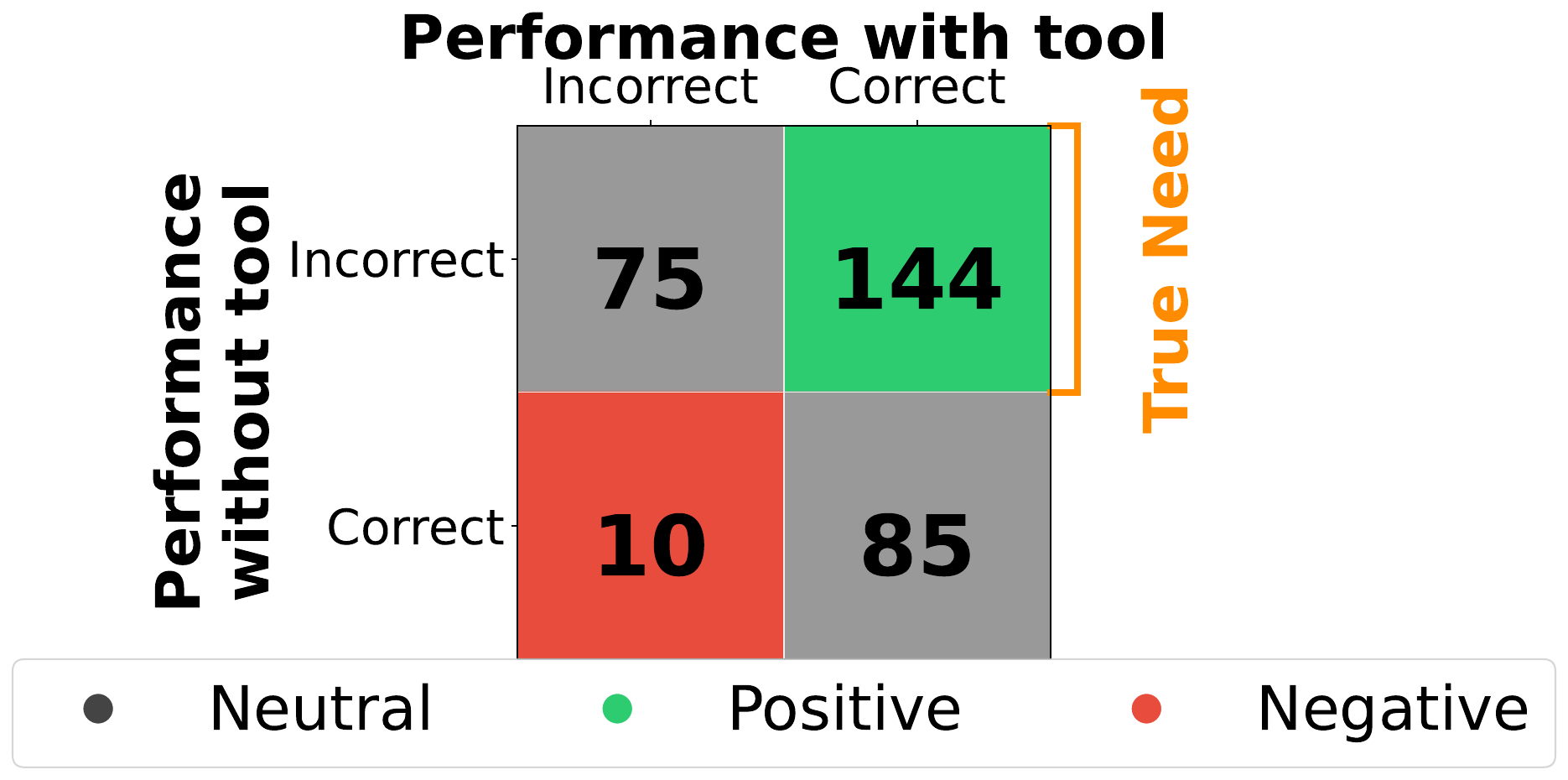}
    \caption{Qwen3-30B-A3B-Instruct}
\end{subfigure}

\begin{subfigure}{0.32\linewidth}
    \centering
    \includegraphics[width=\linewidth]{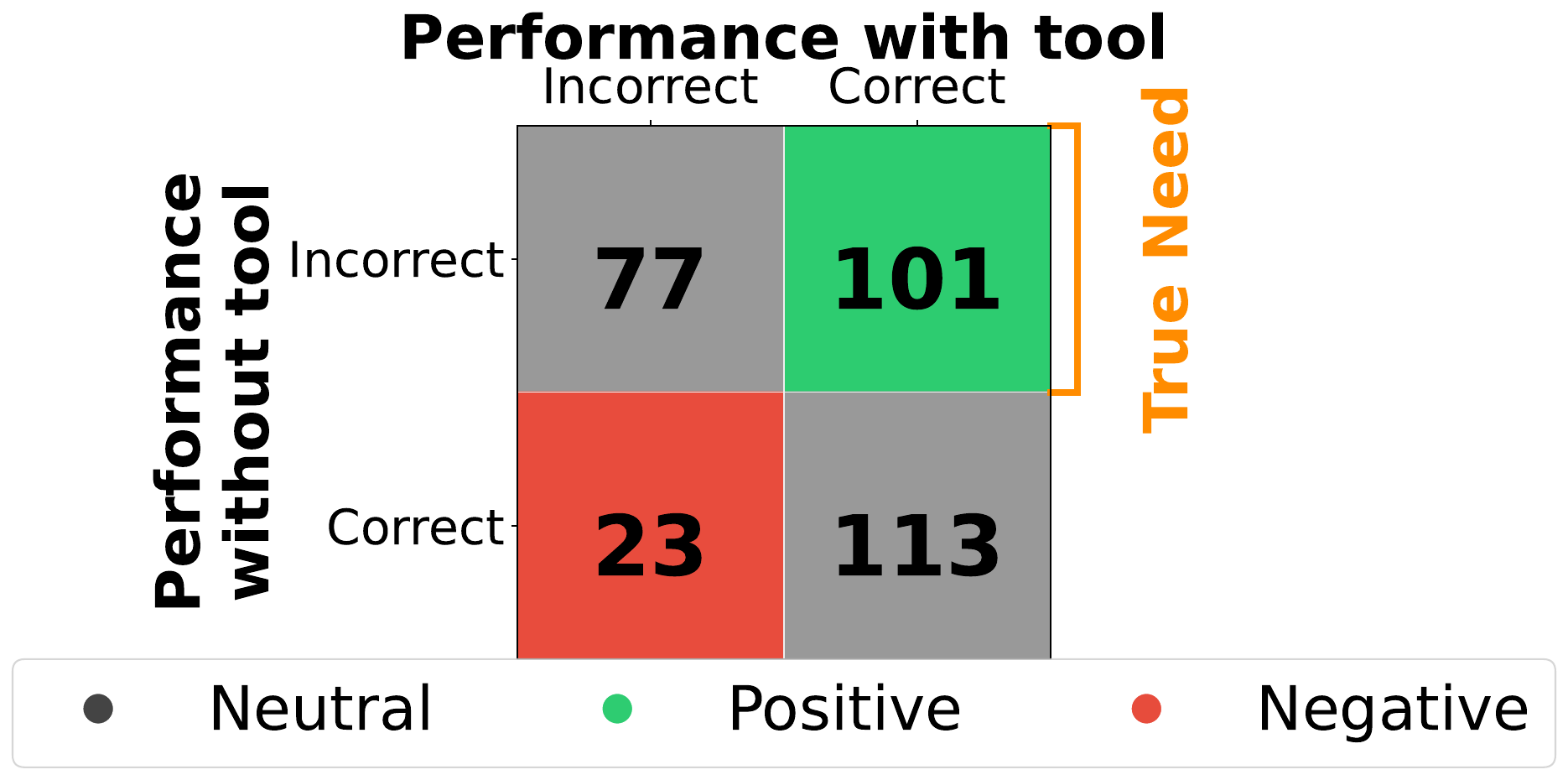}
    \caption{Gemma-3-27B-IT}
\end{subfigure}
\hfill
\begin{subfigure}{0.32\linewidth}
    \centering
    \includegraphics[width=\linewidth]{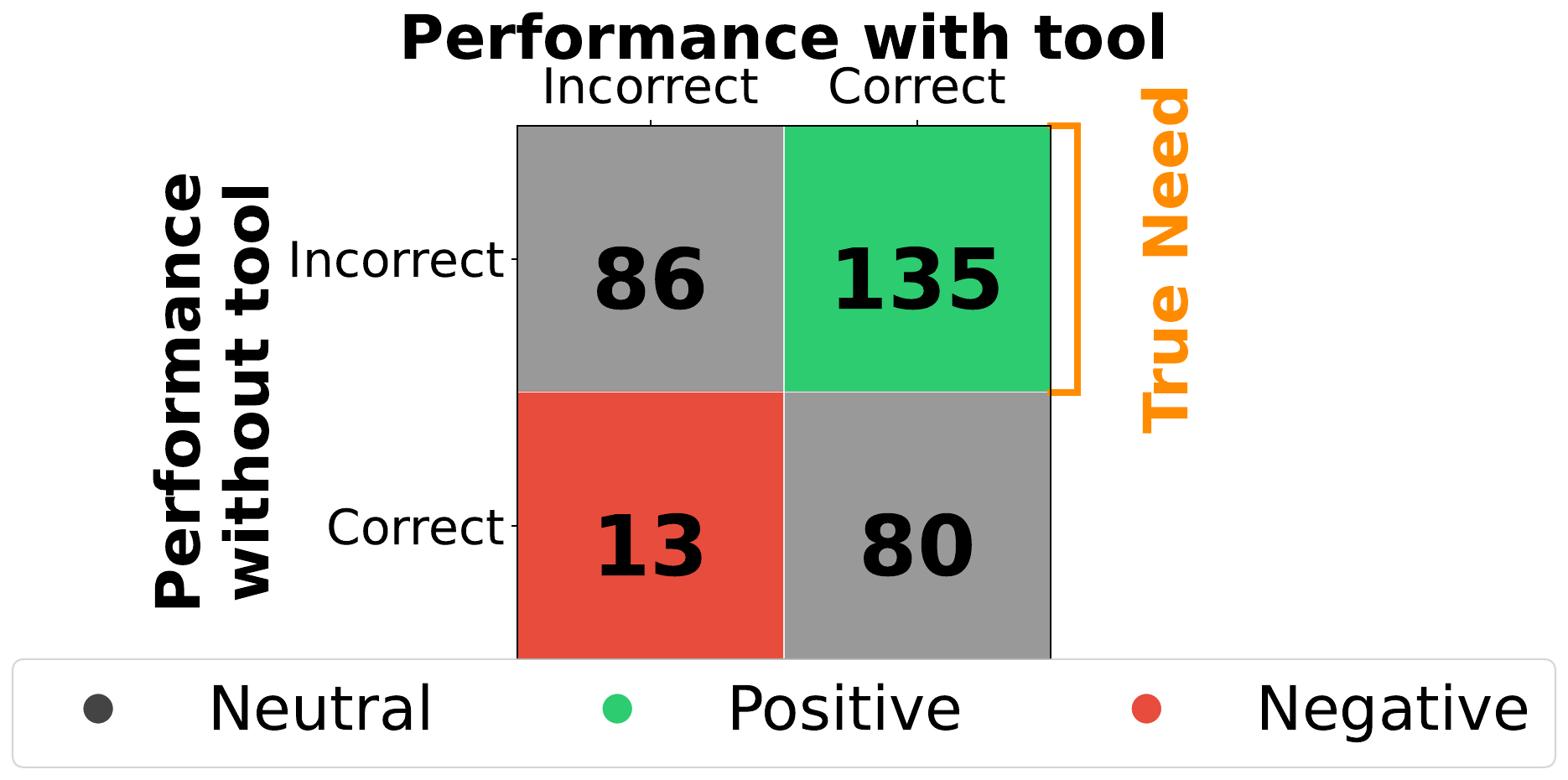}
    \caption{Mistral-3.1-24B-IT}
\end{subfigure}
\hfill
\begin{subfigure}{0.32\linewidth}
    \centering
    \includegraphics[width=\linewidth]{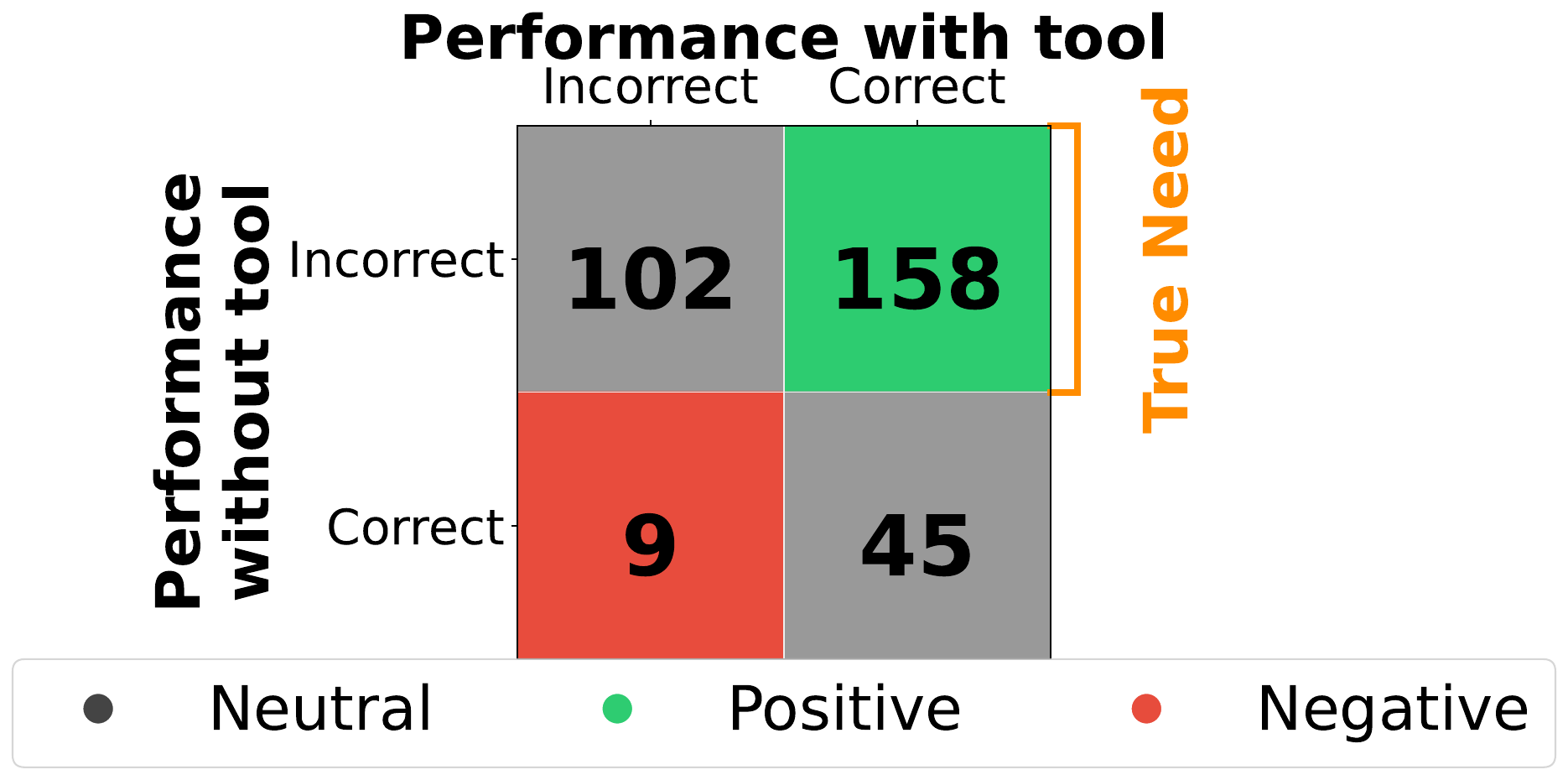}
    \caption{Llama-3.2-3B-IT}
\end{subfigure}
\hfill
\begin{subfigure}{0.32\linewidth}
    \centering
    \includegraphics[width=\linewidth]{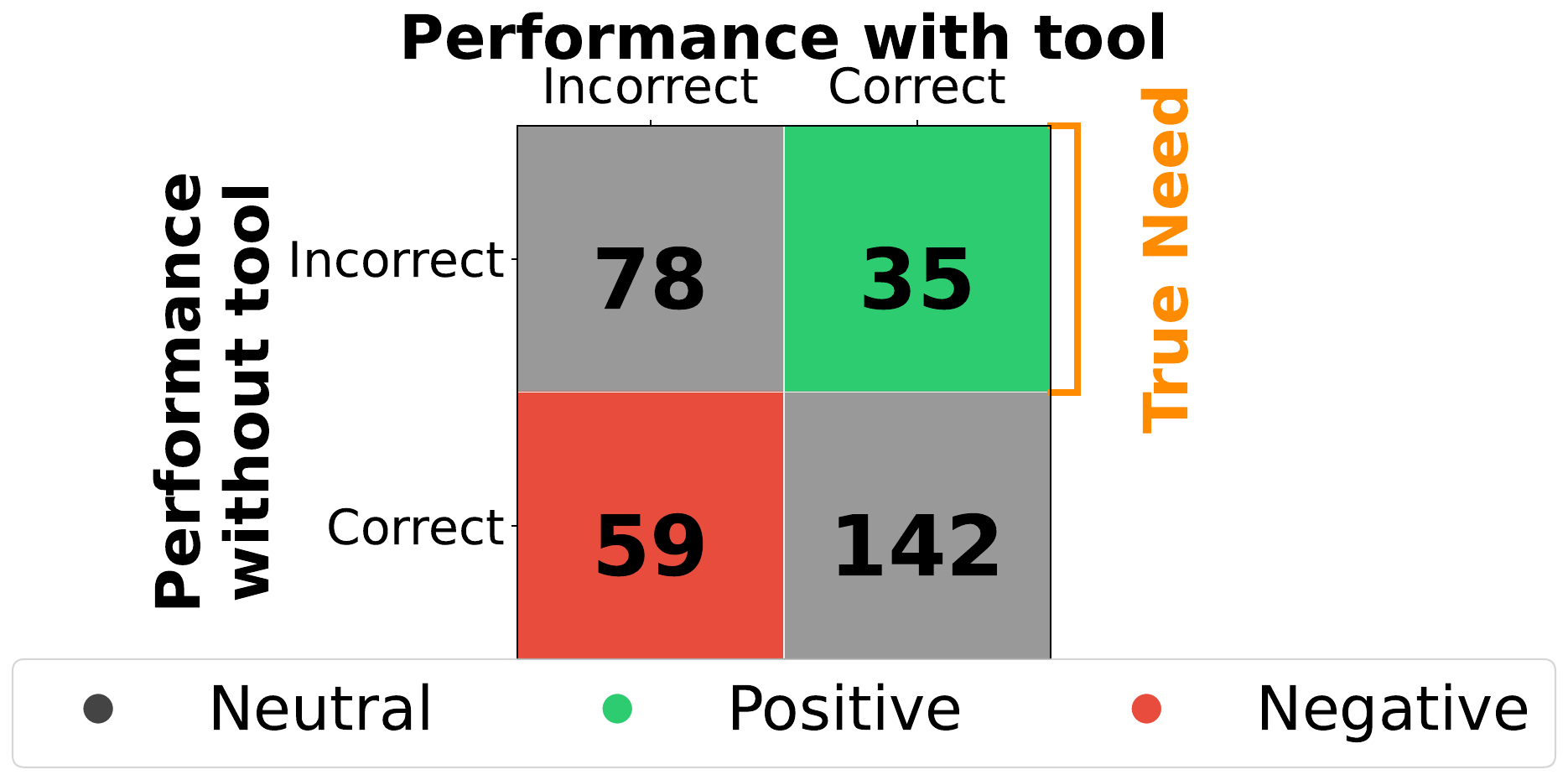}
    \caption{GPT-5.5}
\end{subfigure}

\caption{\textbf{[BFCL task]: No-Tool vs.\ Always-Tool performance.}
        Rows group entities by the model’s factuality score \textit{without a tool} (reflecting parametric knowledge), while columns group scores when tool use is \textit{forced}. Each cell reports the count and the column percentage. Off-diagonal cells indicate performance shifts due to tool use: cells above the diagonal show cases where the tool has {\color{green}\textit{positive utility}}, while cells below the diagonal indicate cases where the tool has {{\color{red}\textit{negative utility}}}. The dashed bracket marks the region of {{\color{orange}\textit{True Need}}}, where \textit{Incorrect} scores suggest insufficient parametric knowledge and thus a likely need for an external tool.}
\label{fig:all_bfcl_actul_need_utility}
\end{figure*}

As shown in Figure~\ref{fig:all_bfcl_actul_need_utility}, a consistent pattern emerges across all models: \textbf{tool use is most beneficial when it is truly needed, can be harmful when unnecessary, and is often redundant otherwise}. This observation highlights the importance of accurately predicting when to invoke external tools, as indiscriminate usage may introduce noise or errors rather than improving factuality.

\subsection{Descriptive Lens}

\begin{figure*}[t]
    \centering
    \includegraphics[width=0.78\textwidth]{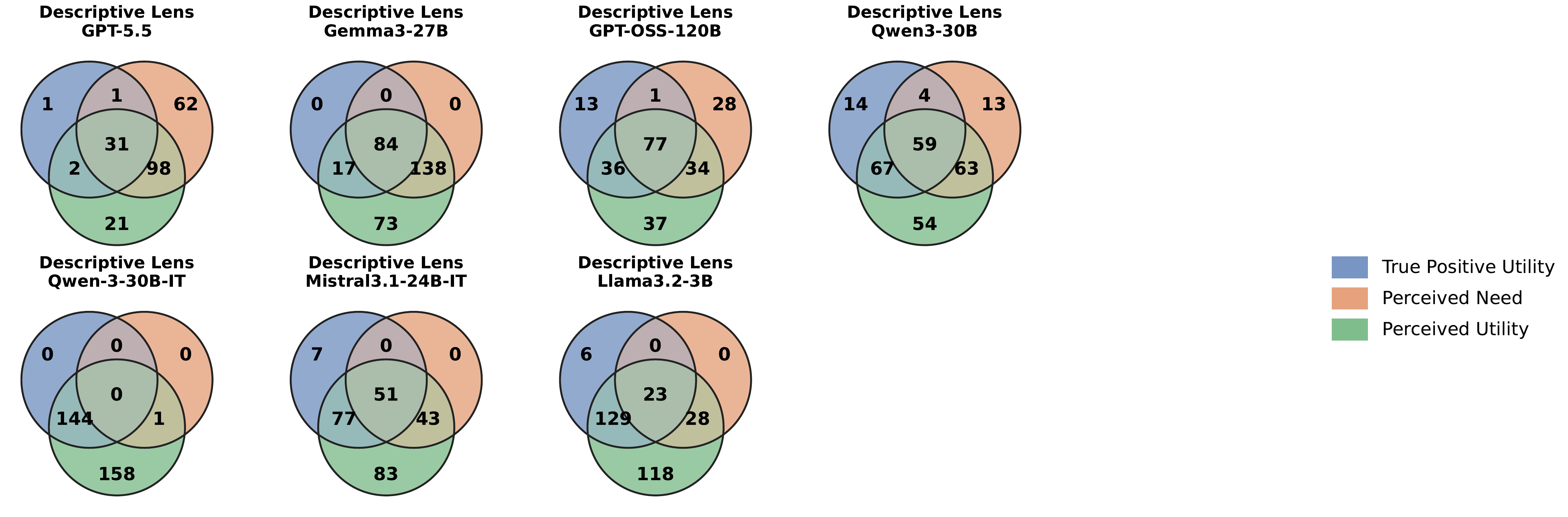}
    \caption{[BFCL Task] Venn diagrams of \textbf{True Positive Utility, Perceived Need, and Perceived Utility}. Each panel shows their empirical overlap for one model. Calls outside true positive utility are non-beneficial; true-positive-utility cases outside perceived utility are missed opportunities. Perceived need is a separate self-assessment and need not be nested within either utility set.}
    \label{fig:venn-bfcl}
\end{figure*}

As shown in Figure~\ref{fig:venn-bfcl}, there is a consistent misalignment between the perceived need and utility and the true positive utility across all models. This discrepancy indicates that models often fail to accurately identify when tool use is genuinely beneficial. As a result, none of the models achieve optimal tool-calling performance, since effective tool use critically depends on correctly aligning perceived need with actual utility.

\begin{figure*}
\centering

\begin{subfigure}{0.25\linewidth}
    \centering
    \includegraphics[width=\linewidth]{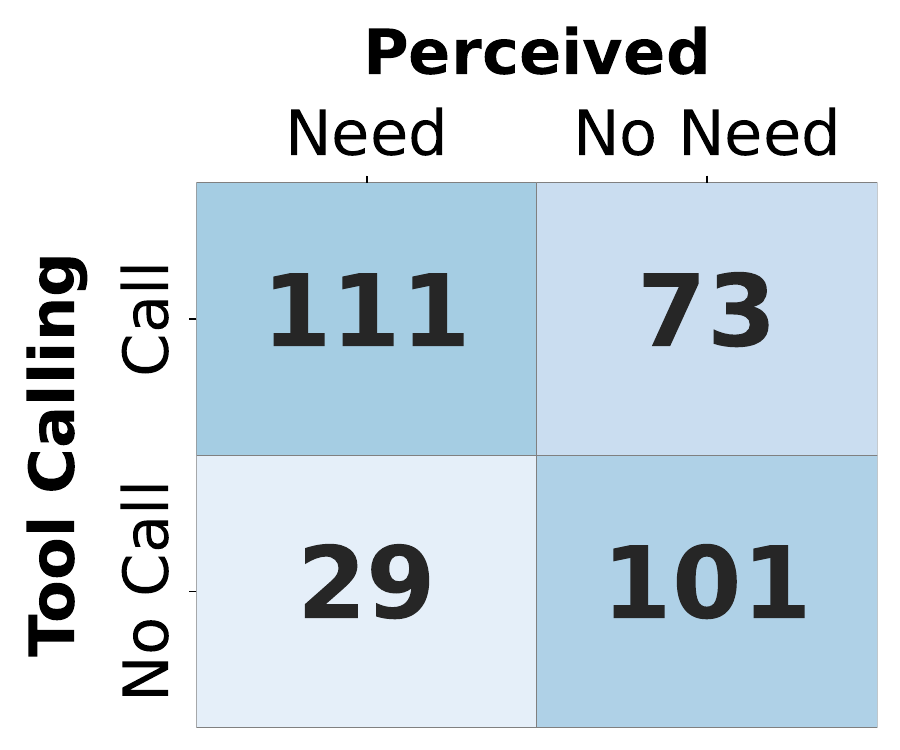}
    \caption{GPT-OSS-120B}
\end{subfigure}\hfill
\begin{subfigure}{0.25\linewidth}
    \centering
    \includegraphics[width=\linewidth]{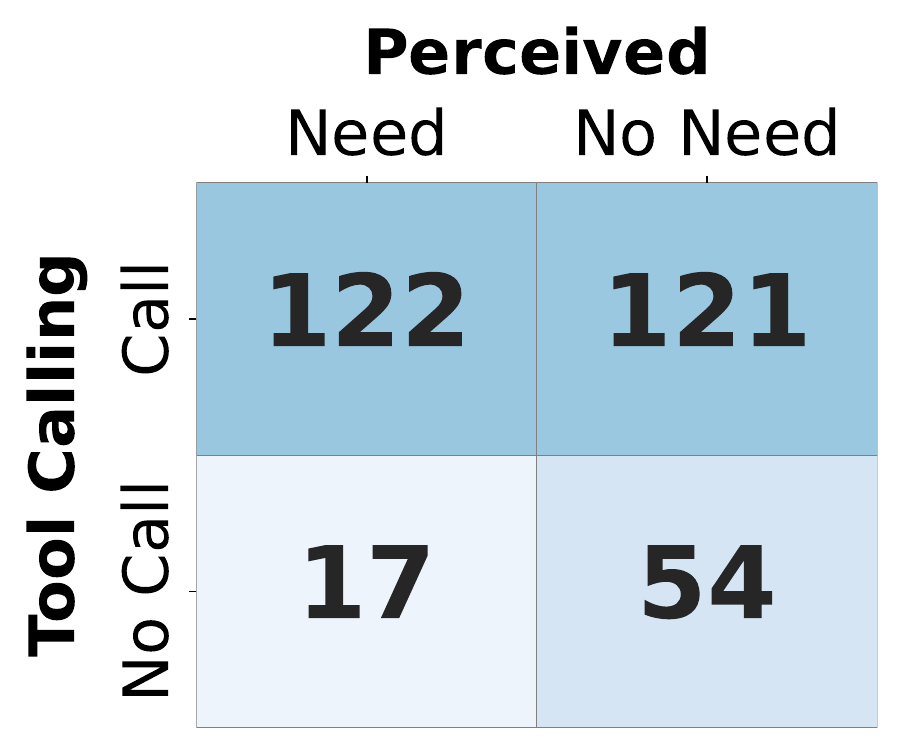}
    \caption{Qwen3-30B-A3B}
\end{subfigure}\hfill
\begin{subfigure}{0.25\linewidth}
    \centering
    \includegraphics[width=\linewidth]{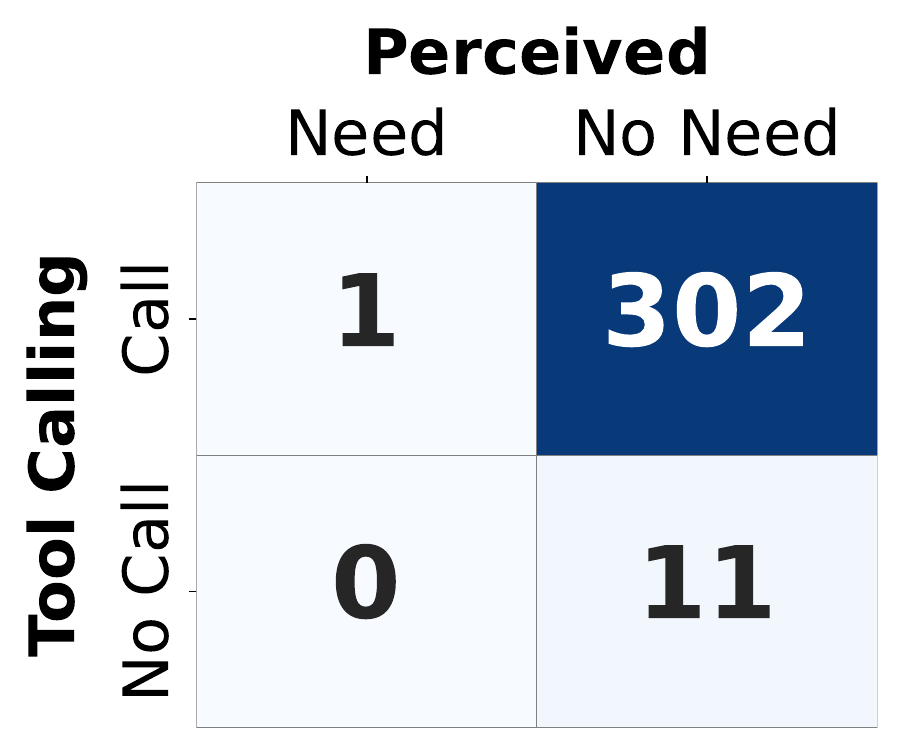}
    \caption{Qwen3-30B-A3B-IT}
\end{subfigure}\hfill
\begin{subfigure}{0.25\linewidth}
    \centering
    \includegraphics[width=\linewidth]{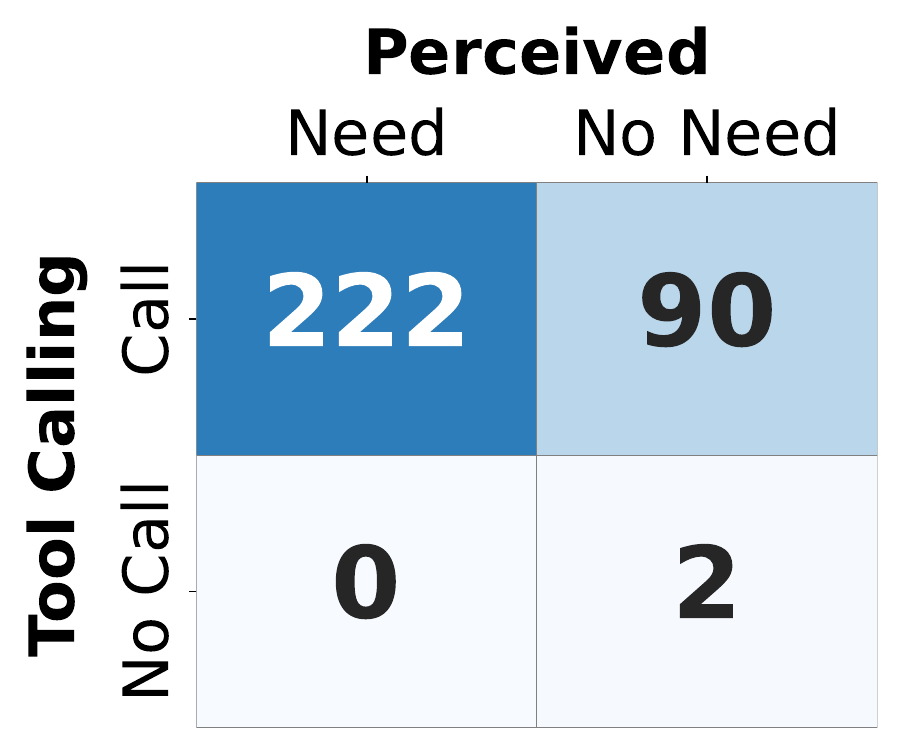}
    \caption{Gemma-3-27B-IT}
\end{subfigure}\hfill
\begin{subfigure}{0.25\linewidth}
    \centering
    \includegraphics[width=\linewidth]{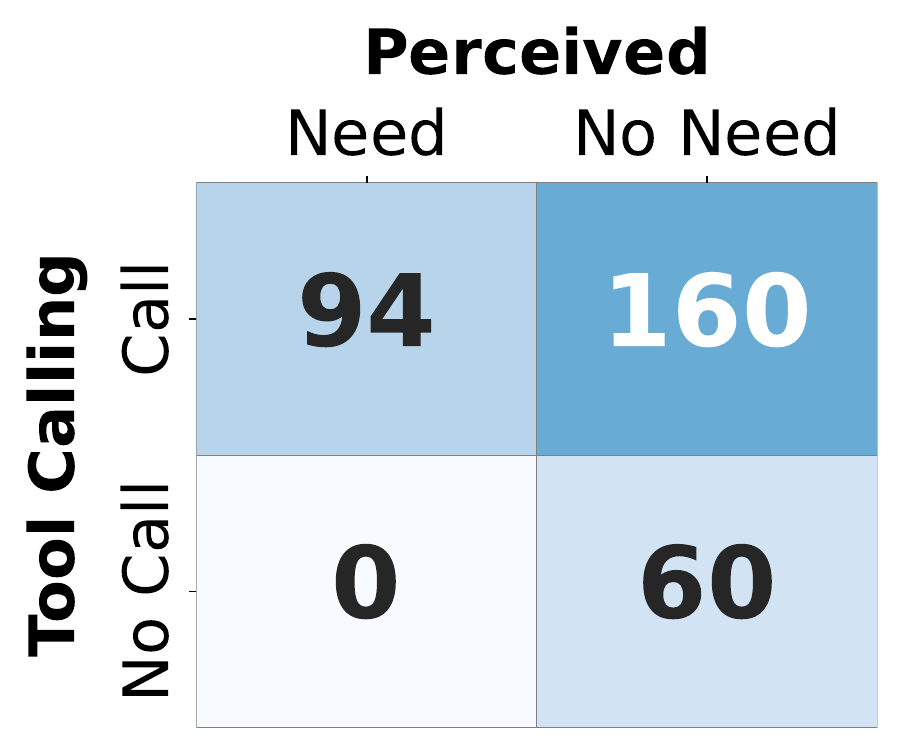}
    \caption{Mistral-3.1-24B-IT}
\end{subfigure}\hfill
\begin{subfigure}{0.25\linewidth}
    \centering
    \includegraphics[width=\linewidth]{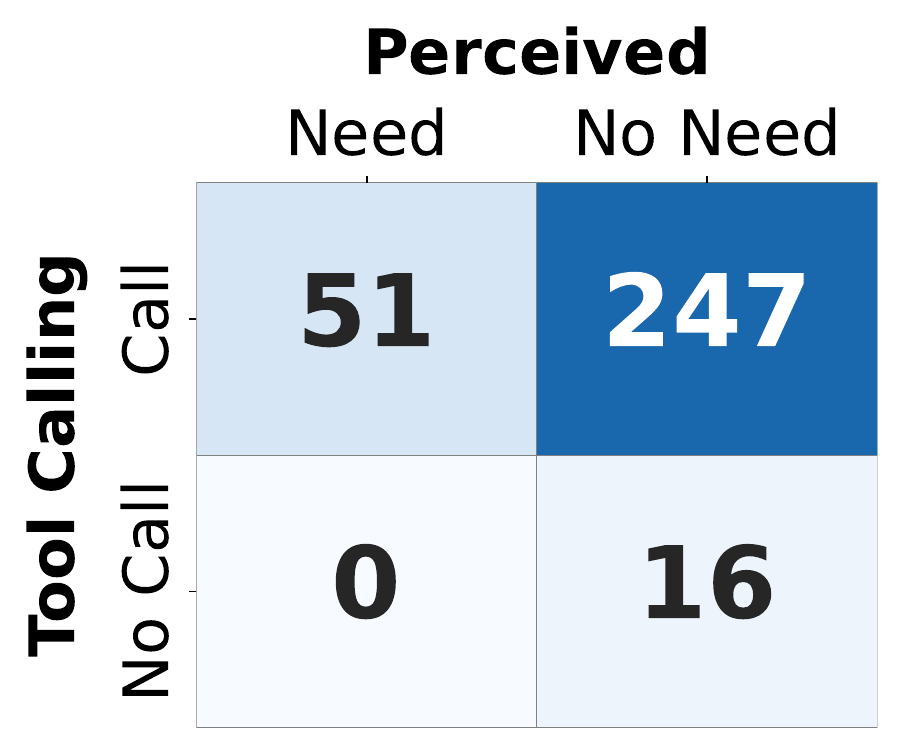}
    \caption{Llama-3.2-3B-IT}
\end{subfigure} \hfill
\begin{subfigure}{0.25\linewidth}
    \centering
    \includegraphics[width=\linewidth]{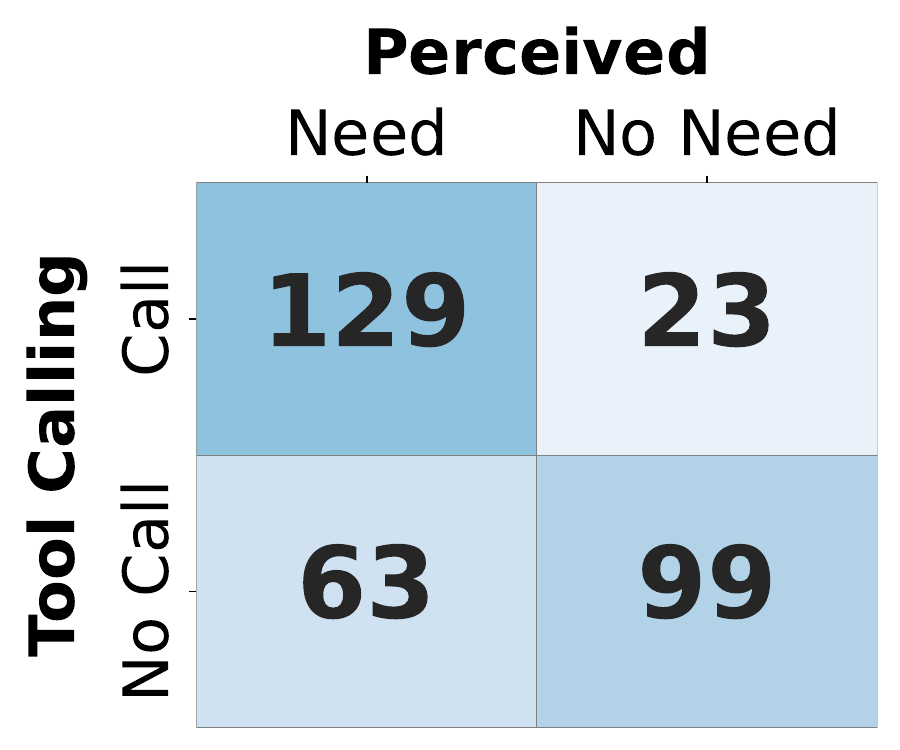}
    \caption{GPT-5.5}
\end{subfigure}

\vspace{6pt}
\textbf{(a) Perceived-need prompt v1}

\begin{subfigure}{0.25\linewidth}
    \centering
    \includegraphics[width=\linewidth]{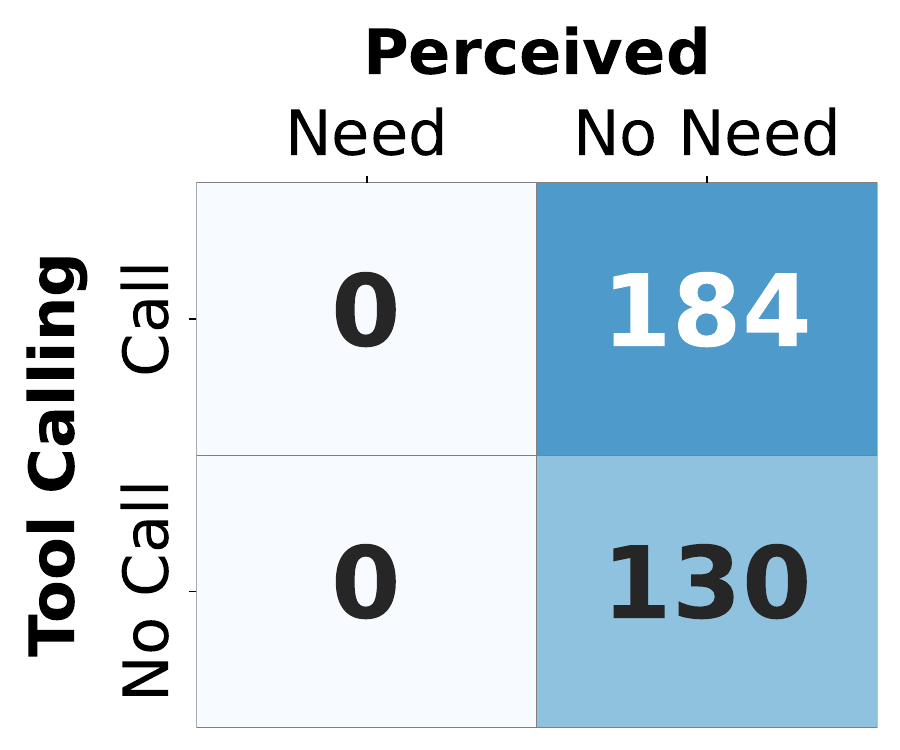}
    \caption{GPT-OSS-120B}
\end{subfigure}\hfill
\begin{subfigure}{0.25\linewidth}
    \centering
    \includegraphics[width=\linewidth]{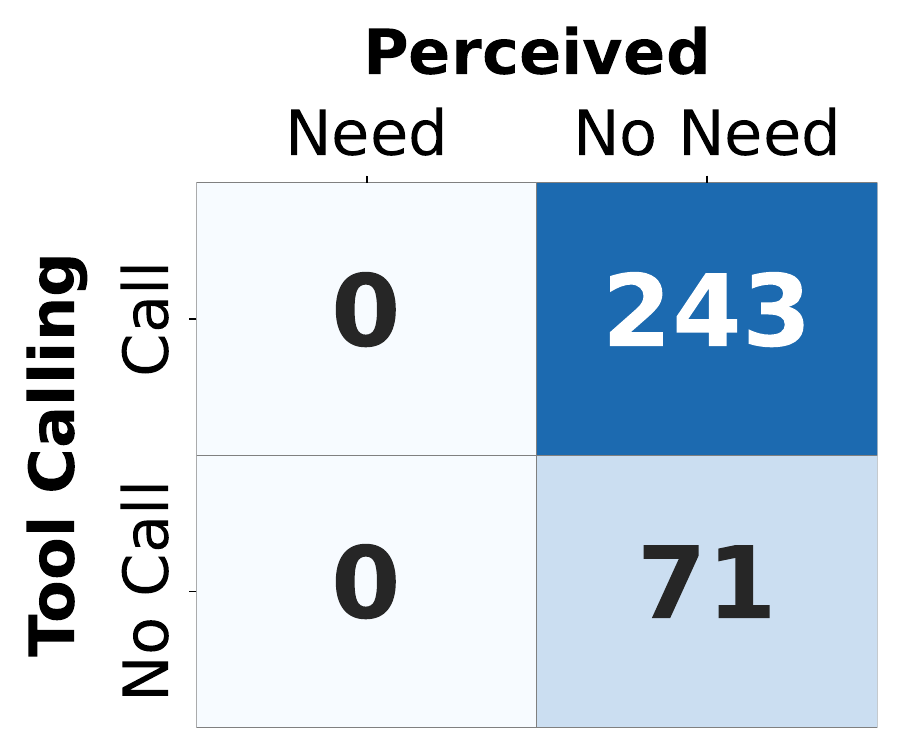}
    \caption{Qwen3-30B-A3B}
\end{subfigure}\hfill
\begin{subfigure}{0.25\linewidth}
    \centering
    \includegraphics[width=\linewidth]{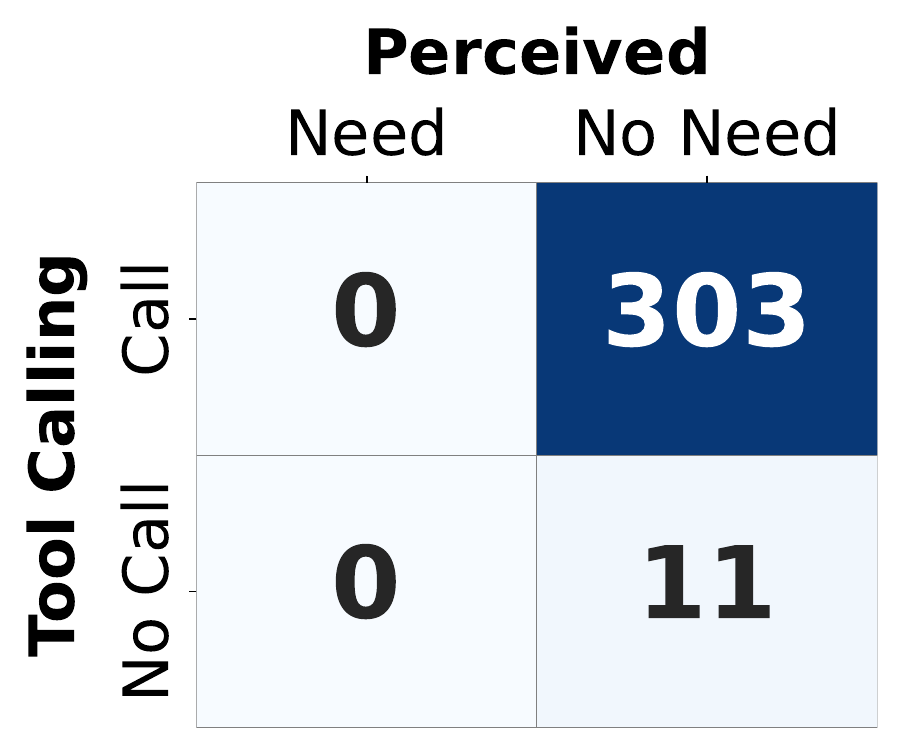}
    \caption{Qwen3-30B-A3B-IT}
\end{subfigure}\hfill
\begin{subfigure}{0.25\linewidth}
    \centering
    \includegraphics[width=\linewidth]{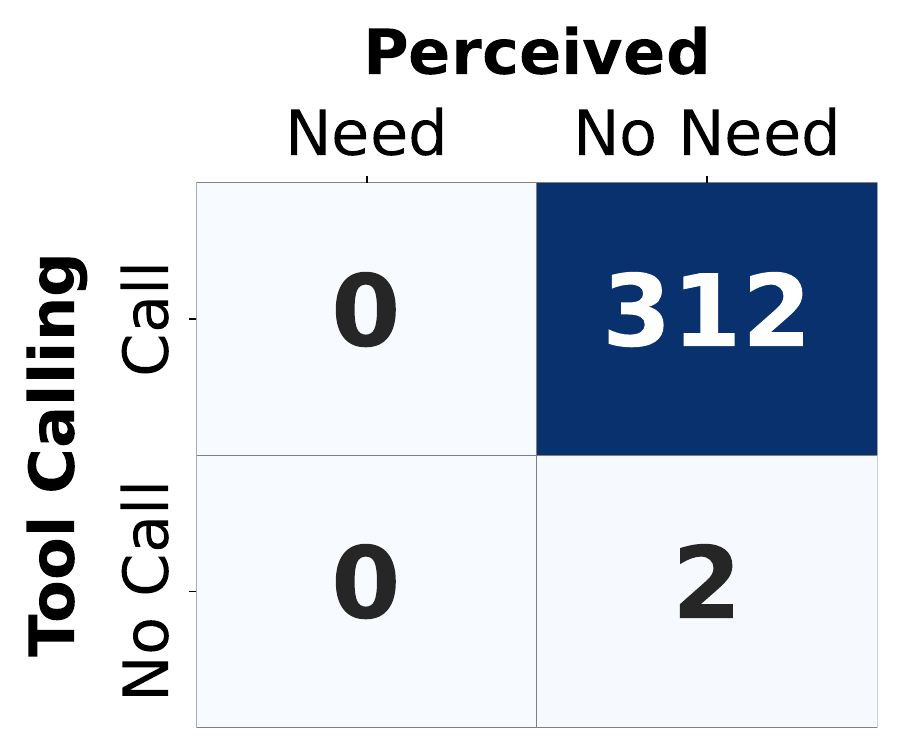}
    \caption{Gemma-3-27B-IT}
\end{subfigure}\hfill
\begin{subfigure}{0.25\linewidth}
    \centering
    \includegraphics[width=\linewidth]{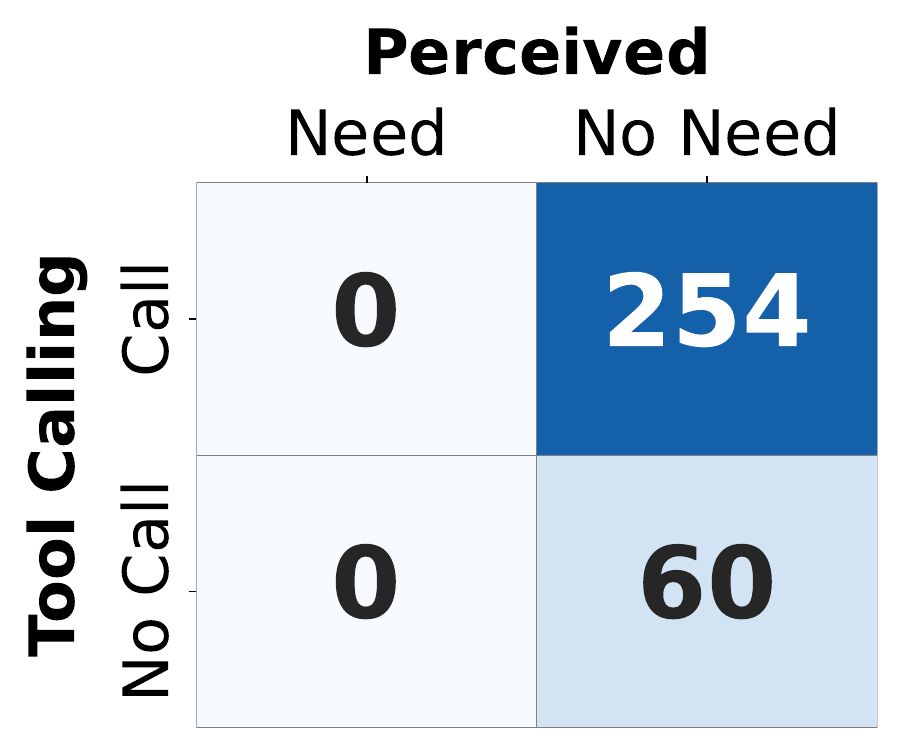}
    \caption{Mistral-3.1-24B-IT}
\end{subfigure}\hfill
\begin{subfigure}{0.25\linewidth}
    \centering
    \includegraphics[width=\linewidth]{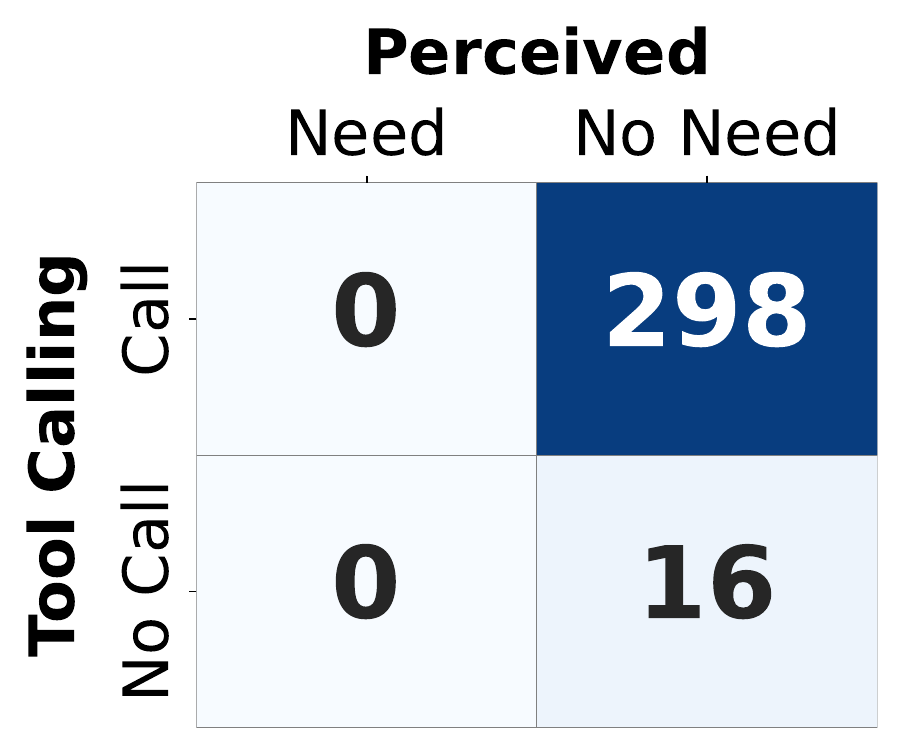}
    \caption{Llama-3.2-3B-IT}
\end{subfigure}

\vspace{6pt}
\textbf{(b) Perceived-need prompt v2}

\caption{\textbf{[BFCL Task]: perceived need is only partially aligned with tool use.} The x-axis shows perceived utility (number of entities predicted to need or not need external information), and the y-axis shows actual tool-use decisions. Percentages indicate how often the model follows its own prediction (call vs.\ not call). Results are shown for three prompt variants. Some responses are excluded due to parsing failures (i.e., missing explicit yes/no decisions), so the total count is less than 500.}

\label{fig:bfcl_need_utility_combined_v12}
\vspace{-15pt}
\end{figure*}

Figure~\ref{fig:bfcl_need_utility_combined_v12} shows two key observations. First, the model's self-perceived need for tool use is highly sensitive to the prompting format, where even small variations can lead to noticeably different outcomes. Second, across these variations, perceived need and utility (i.e., actual tool-calling decisions) are consistently related but not perfectly aligned. 

\begin{figure*}[t]
\centering
{\small\textbf{Need}\par}\vspace{2pt}
\begin{subfigure}{0.135\textwidth}\centering\includegraphics[width=\linewidth,height=0.105\textheight,keepaspectratio]{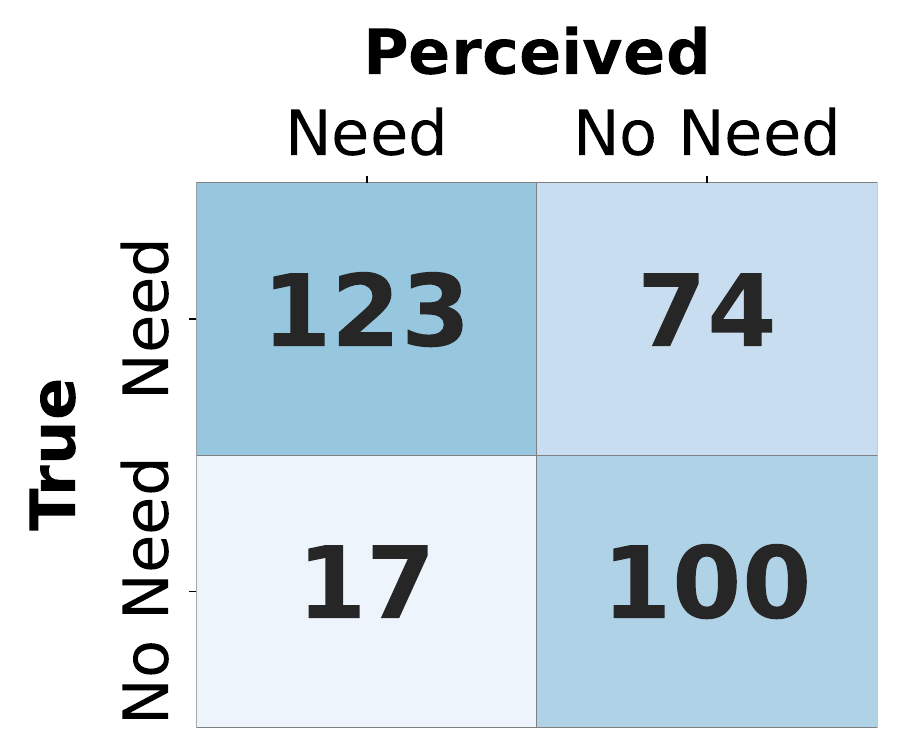}\caption{GPT-OSS-120B}\end{subfigure}\hfill
\begin{subfigure}{0.135\textwidth}\centering\includegraphics[width=\linewidth,height=0.105\textheight,keepaspectratio]{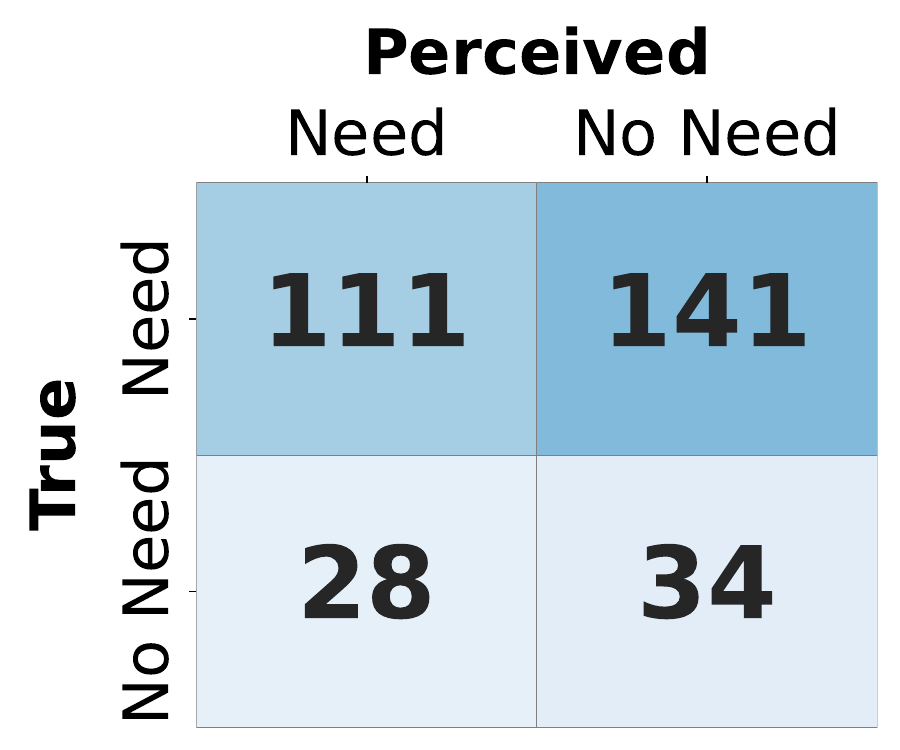}\caption{Qwen3-30B-A3B}\end{subfigure}\hfill
\begin{subfigure}{0.135\textwidth}\centering\includegraphics[width=\linewidth,height=0.105\textheight,keepaspectratio]{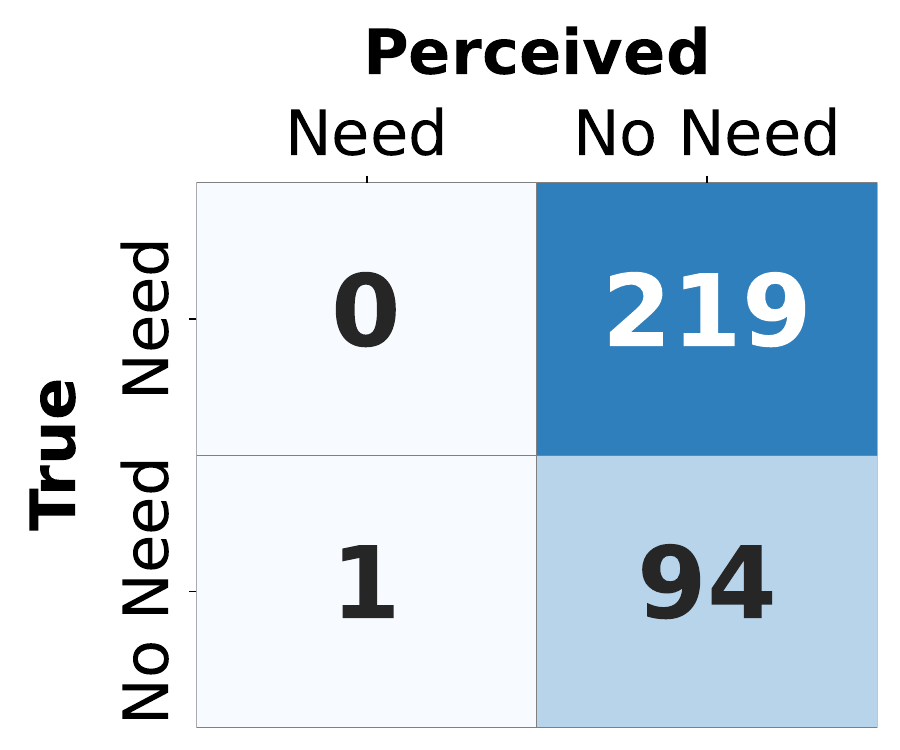}\caption{Qwen3-30B-A3B-IT}\end{subfigure}\hfill
\begin{subfigure}{0.135\textwidth}\centering\includegraphics[width=\linewidth,height=0.105\textheight,keepaspectratio]{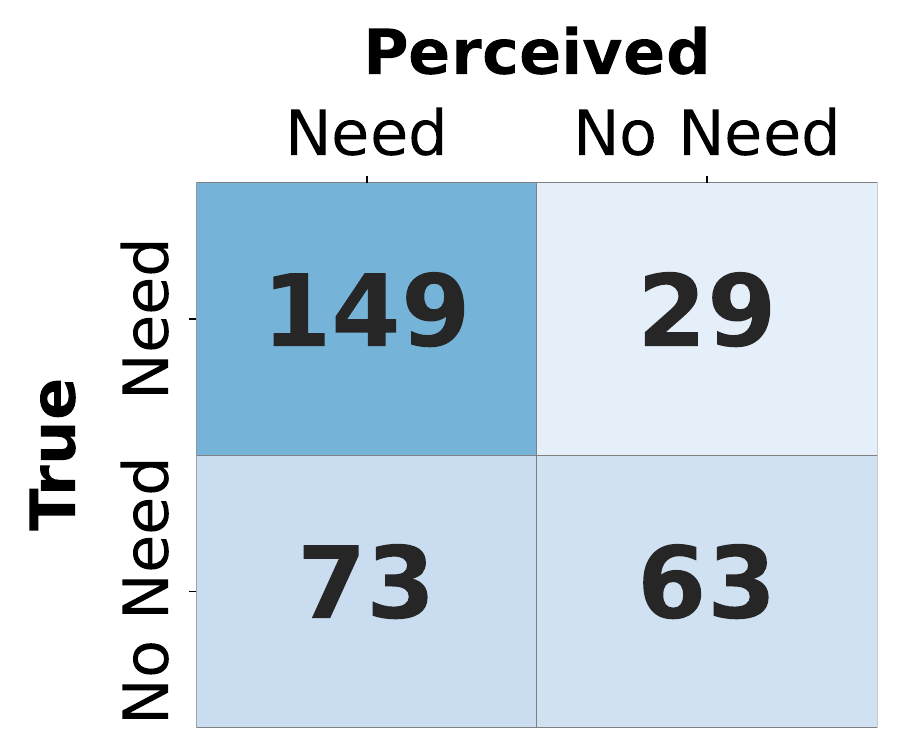}\caption{Gemma-3-27B-IT}\end{subfigure}\hfill
\begin{subfigure}{0.135\textwidth}\centering\includegraphics[width=\linewidth,height=0.105\textheight,keepaspectratio]{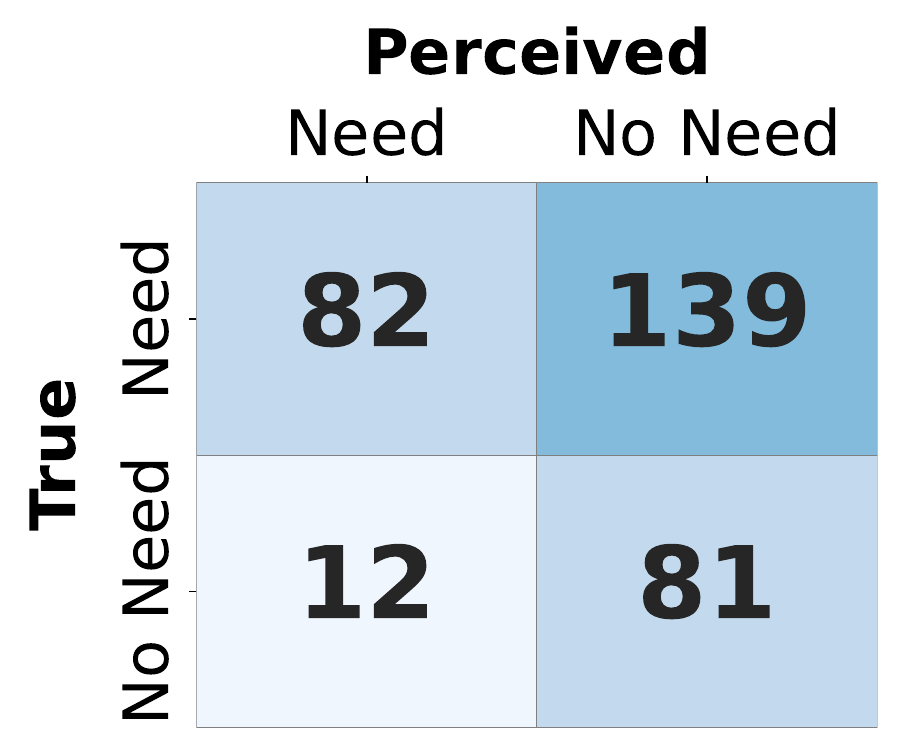}\caption{Mistral-3.1-24B-IT}\end{subfigure}\hfill
\begin{subfigure}{0.135\textwidth}\centering\includegraphics[width=\linewidth,height=0.105\textheight,keepaspectratio]{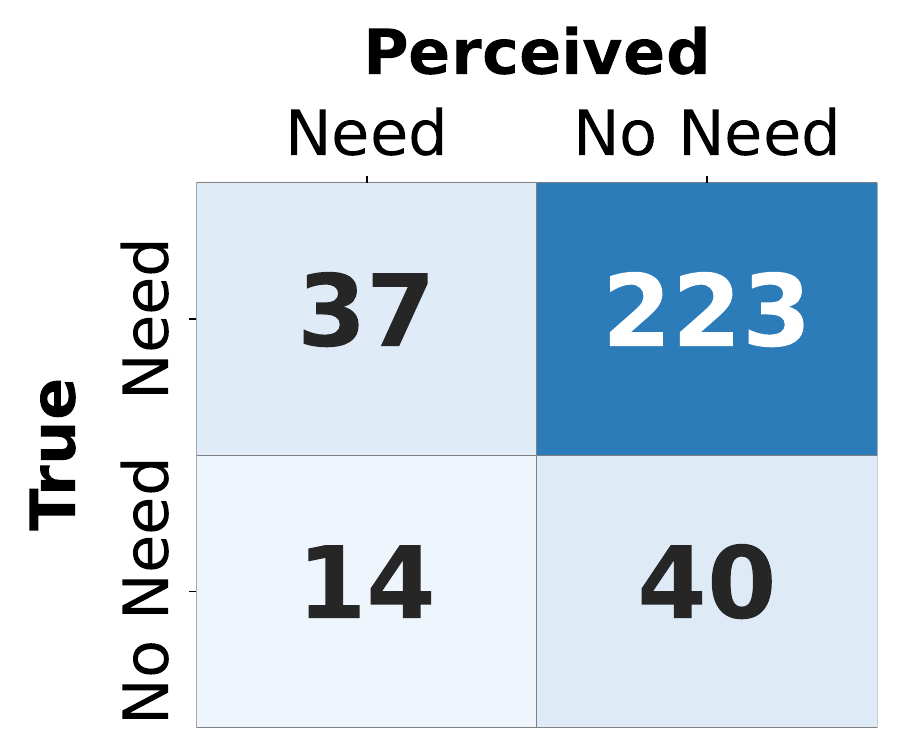}\caption{Llama-3.2-3B-IT}\end{subfigure}\hfill
\begin{subfigure}{0.135\textwidth}\centering\includegraphics[width=\linewidth,height=0.105\textheight,keepaspectratio]{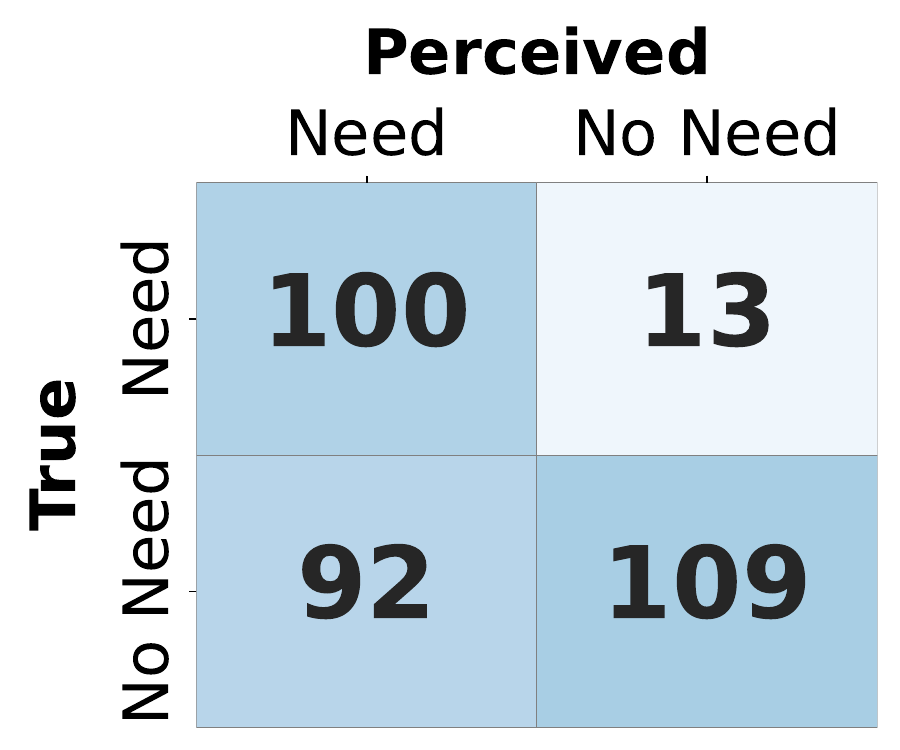}\caption{Llama-3.2-3B-IT}\end{subfigure}

\vspace{5pt}{\small\textbf{Utility}\par}\vspace{2pt}
\begin{subfigure}{0.135\textwidth}\centering\includegraphics[width=\linewidth,height=0.105\textheight,keepaspectratio]{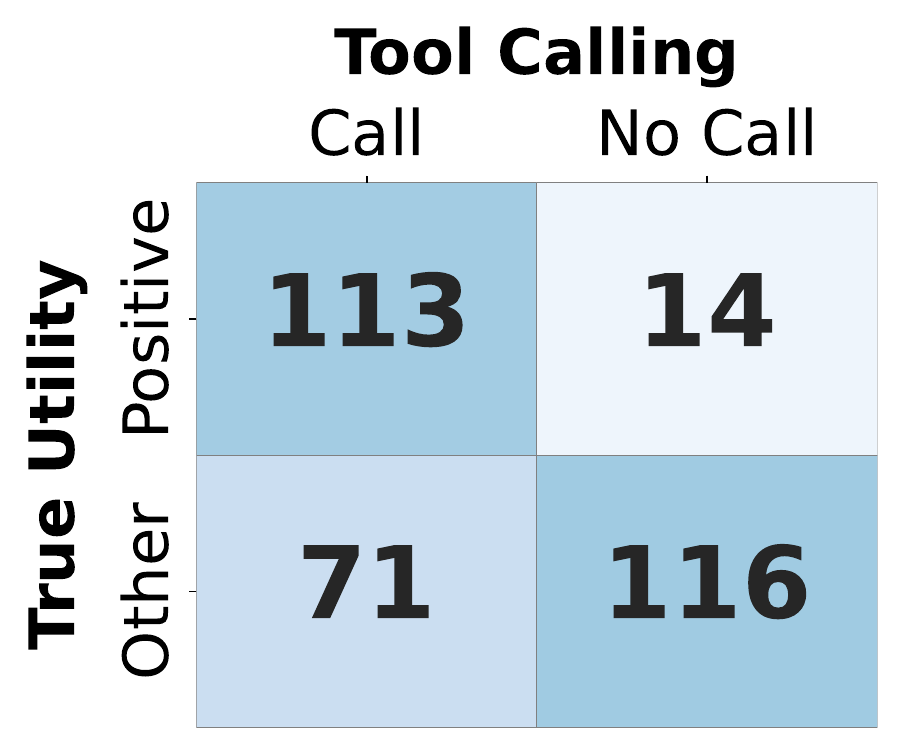}\caption{GPT-OSS-120B}\end{subfigure}\hfill
\begin{subfigure}{0.135\textwidth}\centering\includegraphics[width=\linewidth,height=0.105\textheight,keepaspectratio]{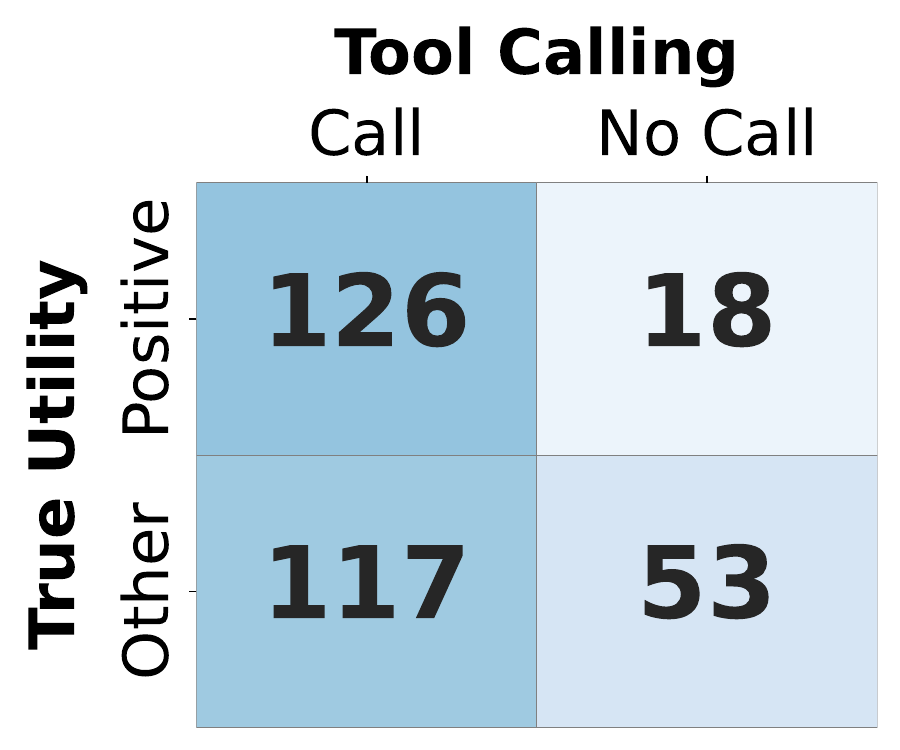}\caption{Qwen3-30B-A3B}\end{subfigure}\hfill
\begin{subfigure}{0.135\textwidth}\centering\includegraphics[width=\linewidth,height=0.105\textheight,keepaspectratio]{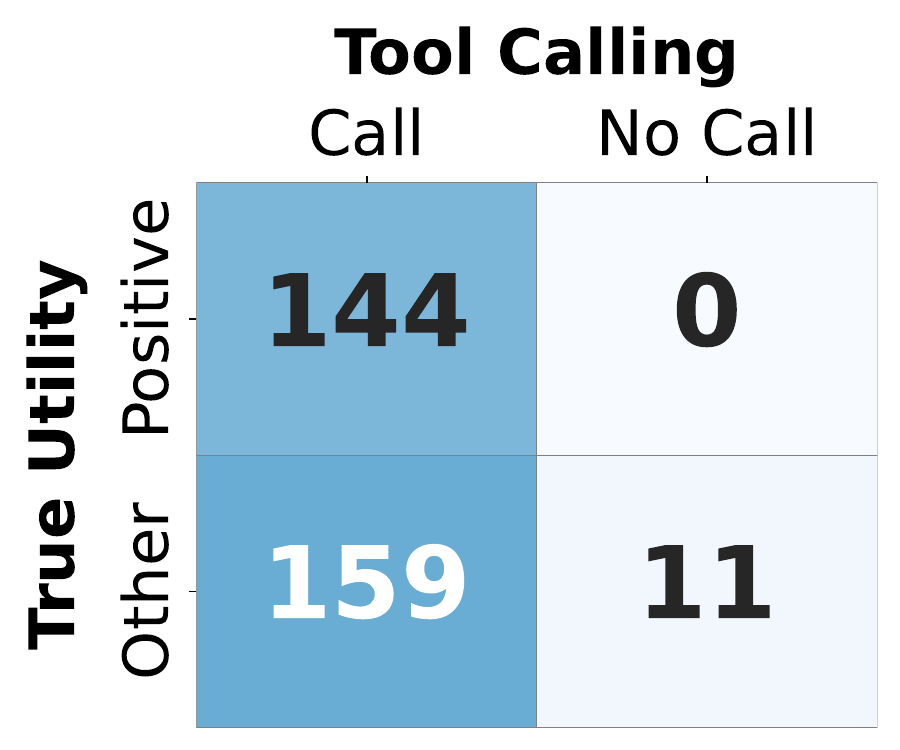}\caption{Qwen3-30B-A3B-IT}\end{subfigure}\hfill
\begin{subfigure}{0.135\textwidth}\centering\includegraphics[width=\linewidth,height=0.105\textheight,keepaspectratio]{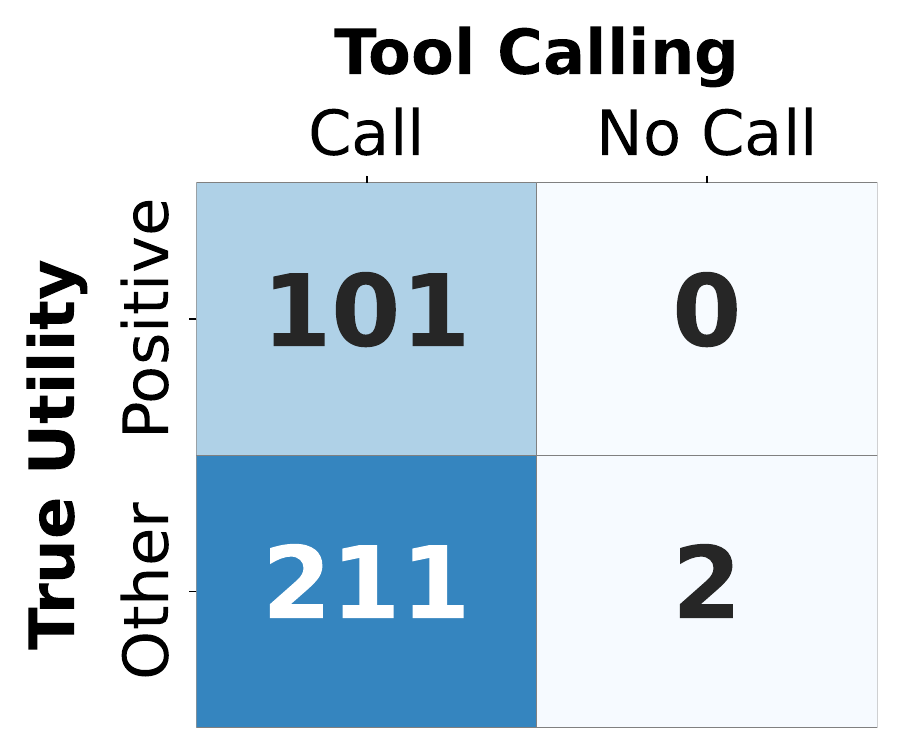}\caption{Gemma-3-27B-IT}\end{subfigure}\hfill
\begin{subfigure}{0.135\textwidth}\centering\includegraphics[width=\linewidth,height=0.105\textheight,keepaspectratio]{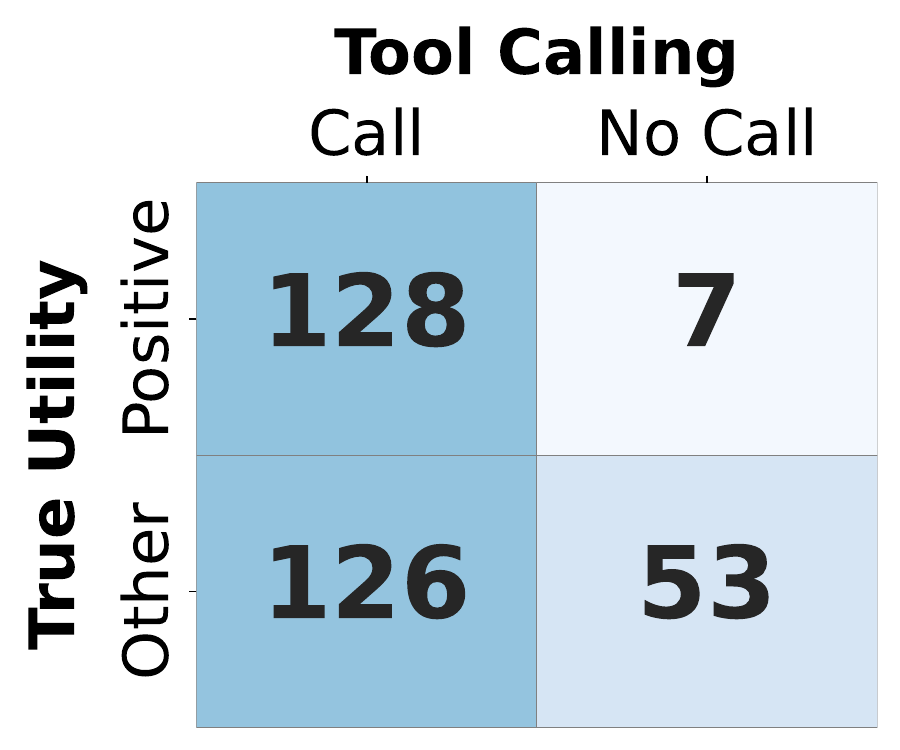}\caption{Mistral-3.1-24B-IT}\end{subfigure}\hfill
\begin{subfigure}{0.135\textwidth}\centering\includegraphics[width=\linewidth,height=0.105\textheight,keepaspectratio]{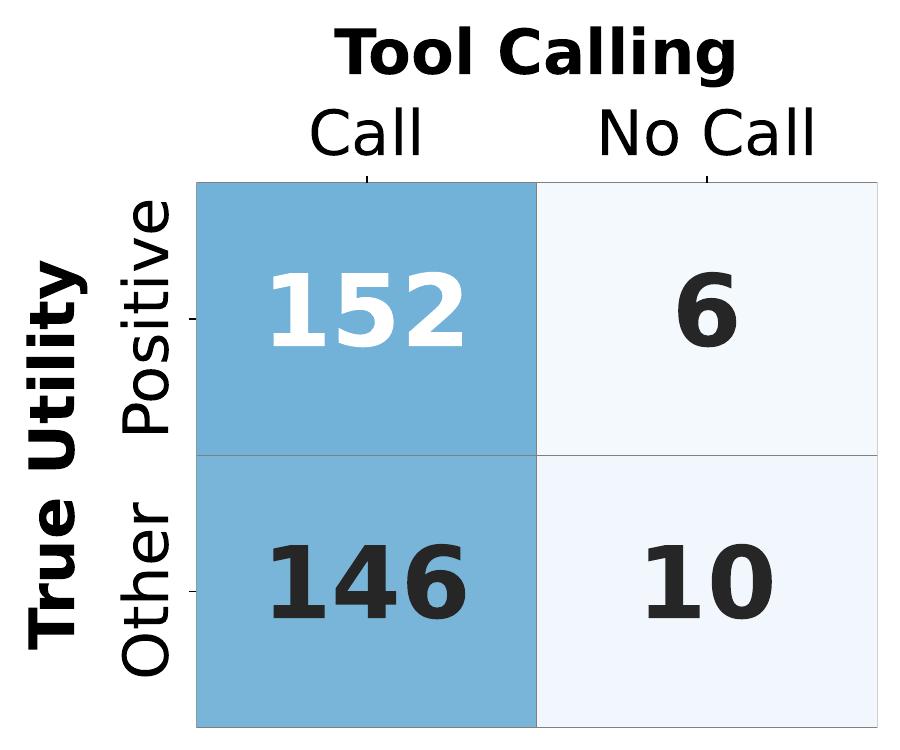}\caption{Llama-3.2-3B-IT}\end{subfigure}\hfill
\begin{subfigure}{0.135\textwidth}\centering\includegraphics[width=\linewidth,height=0.105\textheight,keepaspectratio]{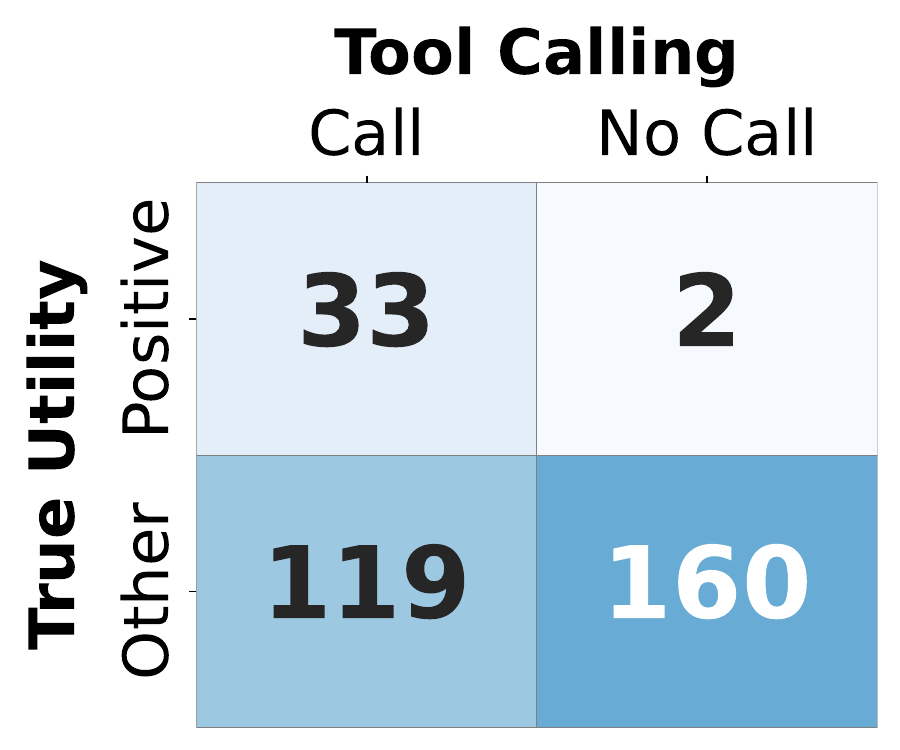}\caption{GPT-5.5}\end{subfigure}
\caption{\textbf{[BFCL Task] The perceived need and utility are not aligned with the true need and utility.} Top: perceived need matrices. Bottom: true vs.\ perceived utility across models. The GPT-5.5 perceived-need panel was not available.}
\label{fig:bfcl_true_perceived}
\label{fig:bfcl_gpt55_true_perceived}
\end{figure*}

GPT-5.5 also exhibits descriptive misalignment on BFCL: autonomous tool calling does not perfectly separate positive utility from negative or neutral utility. Its available true-utility versus perceived-utility result is included in Figure~\ref{fig:bfcl_true_perceived}; a corresponding perceived-need result was not available, so we do not infer or synthesize that measurement.

\begin{figure}[t]
\vspace{-6pt}
\centering

\begin{subfigure}{\linewidth}
    \centering
    \includegraphics[width=\linewidth]{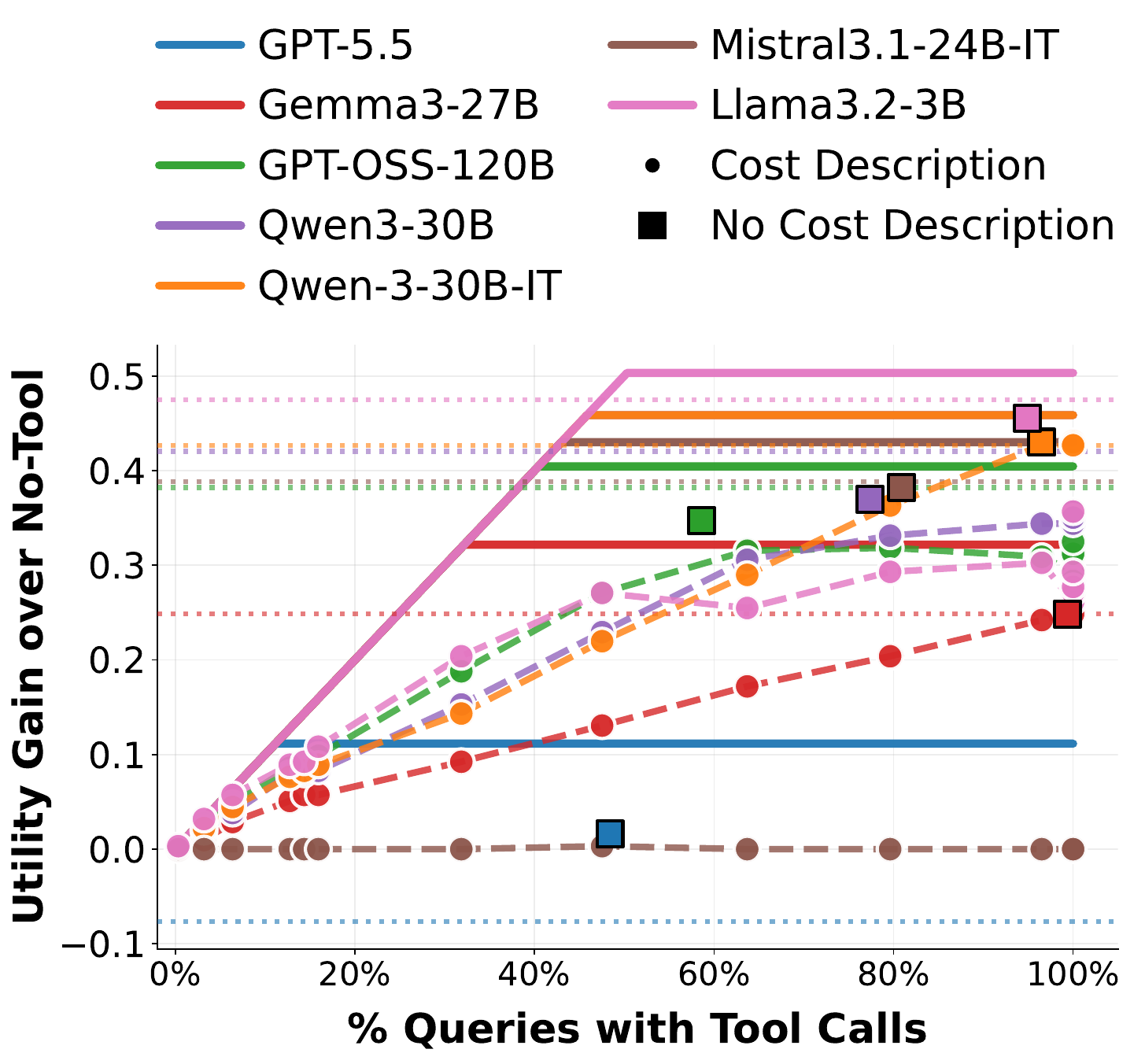}
    \caption{\textbf{Utility gain with hard stop after exceeding the budget.}}
\end{subfigure}\hfill
\begin{subfigure}{\linewidth}
    \centering
    \includegraphics[width=\linewidth]{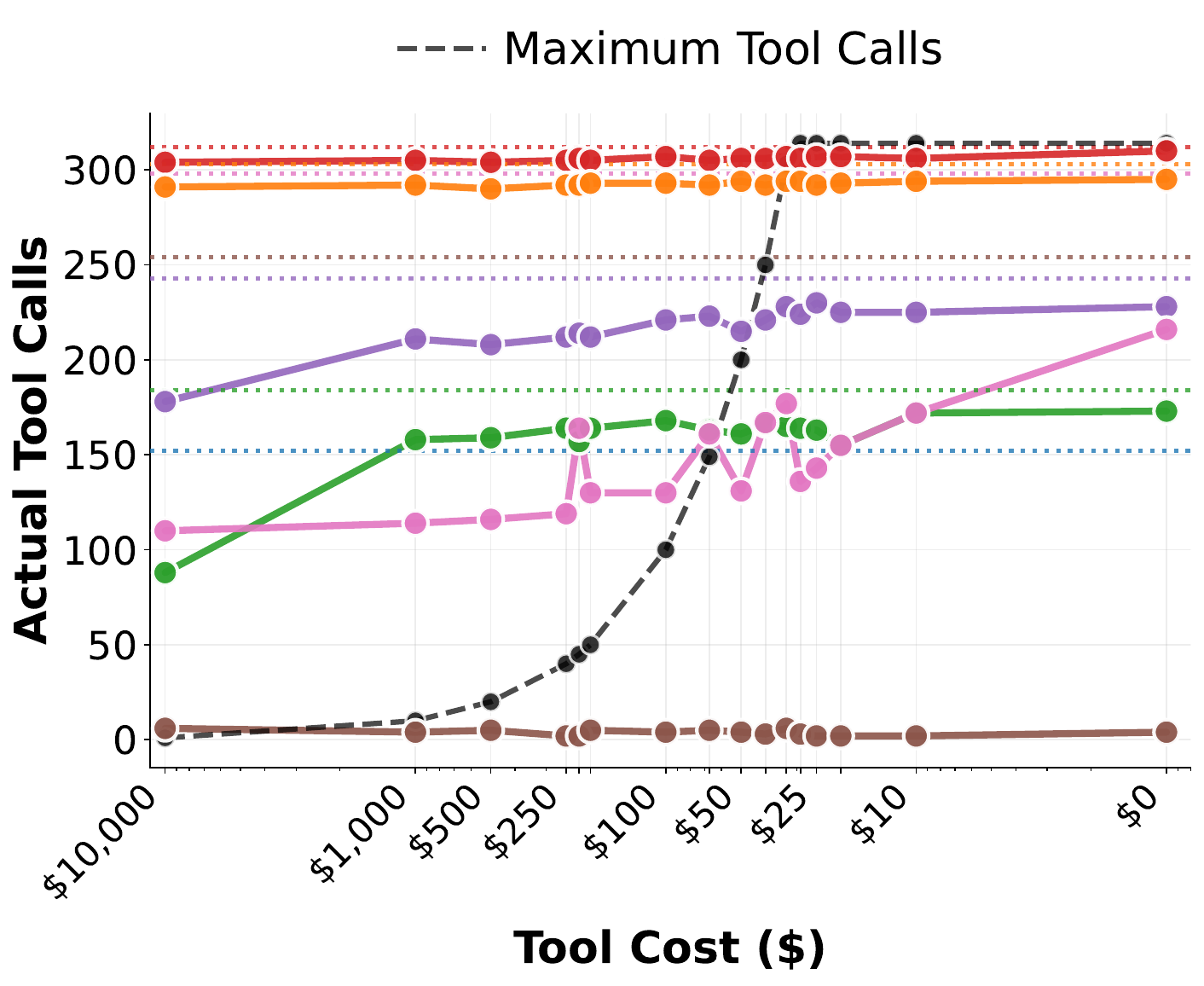}
    \caption{\textbf{Tool-calling behavior without hard stop.}}
\end{subfigure}
\caption{
\textbf{Cost-aware tool use on the BFCL Task with implicit budget notification..}
\textbf{Left:} Utility gain over the no-tool baseline under varying cost constraints. Solid lines show oracle allocation (optimal top-$k$), dashed lines show model performance with cost information, squares denote no cost-awareness, and dotted lines indicate always-calling.
\textbf{Right:} Actual tool calls without budget enforcement. Models do not reliably reduce or stop calls as cost increases, despite being provided with cost and remaining budget.
}
\label{fig:bfcl_affordability_combined}
\end{figure}

\begin{figure}
    \centering
    \includegraphics[width=\linewidth]{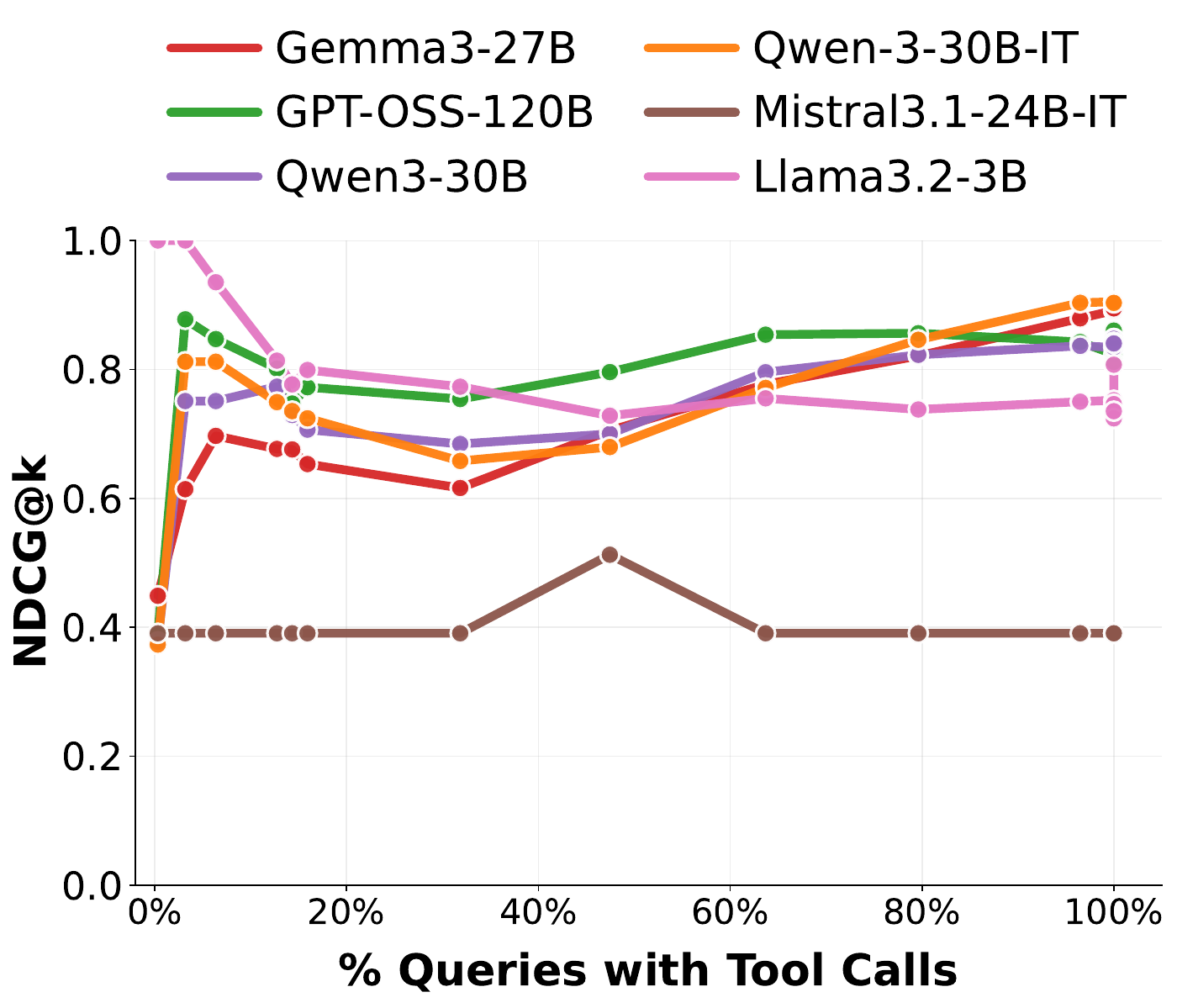}
    \caption{[BFCL Task] The NDCG rank correlation under different budgets across different models. The correlation is low, which reflects that the models are not choosing the best utility gain tool calling. Cost prompt v1.}
    \label{fig:bfcl-ndcg-v1}
\end{figure}

\begin{figure}[t]
\centering
\begin{subfigure}{\linewidth}
    \centering
    \includegraphics[width=\linewidth]{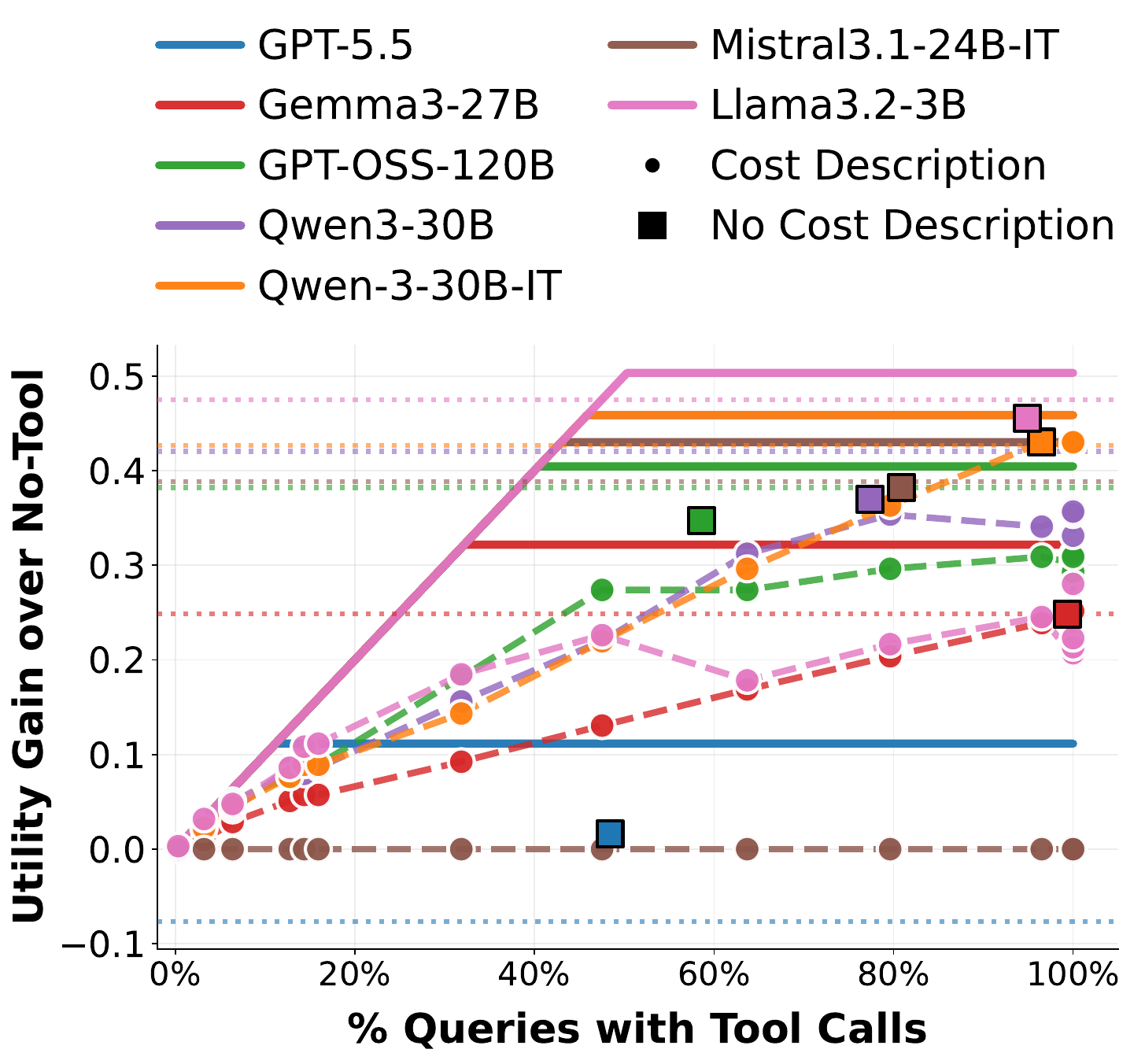}
    \caption{\textbf{Utility gain with hard stop after exceeding the budget.}}
\end{subfigure}\hfill
\begin{subfigure}{\linewidth}
    \centering
    \includegraphics[width=\linewidth]{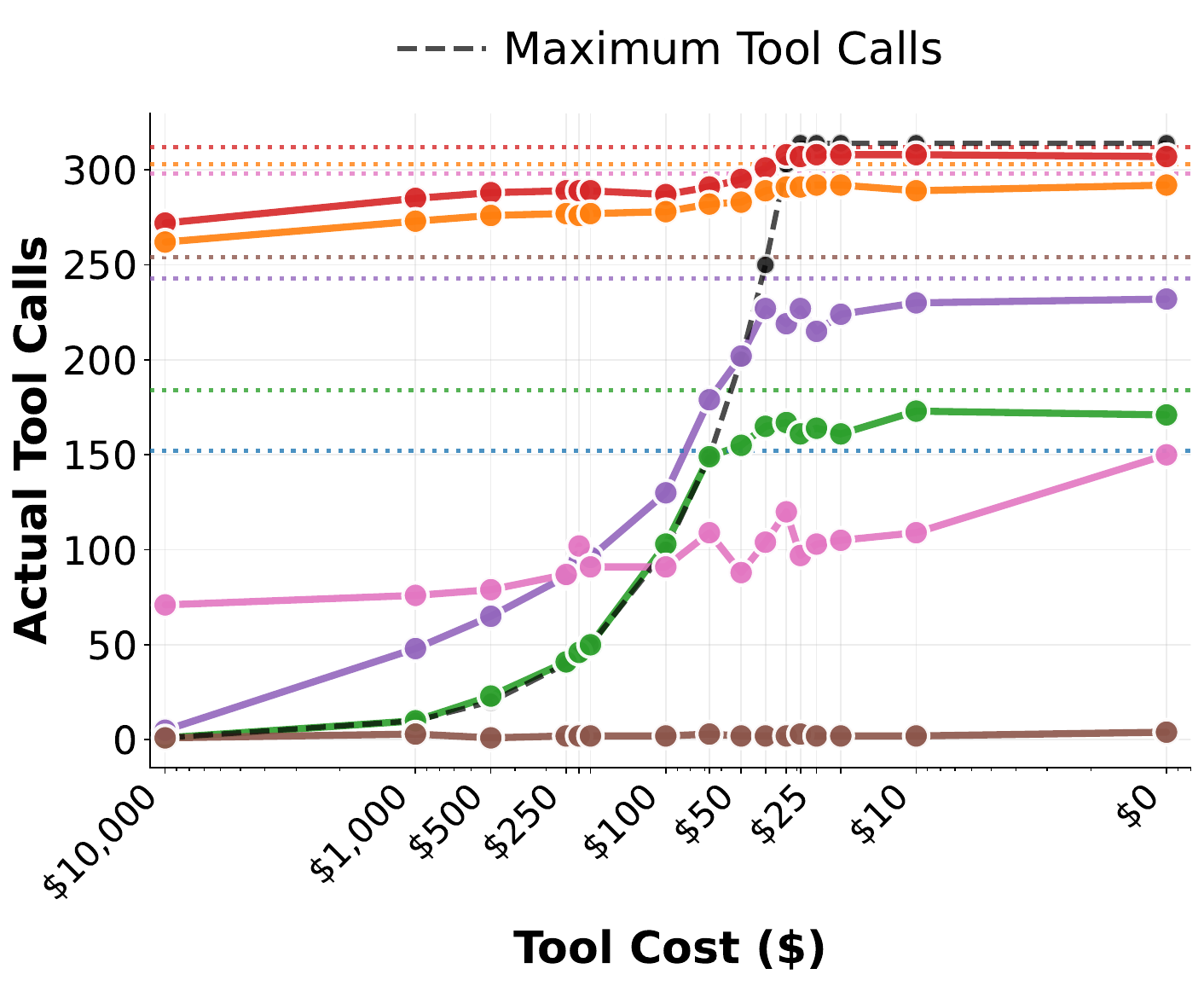}
    \caption{\textbf{Tool-calling behavior without hard stop.}}
\end{subfigure}
\caption{
\textbf{Cost-aware tool use on the BFCL Task with explicit budget notification.}
\textbf{Left:} Utility gain over the no-tool baseline under varying cost constraints. Solid lines show oracle allocation (optimal top-$k$), dashed lines show model performance with cost information, squares denote no cost-awareness, and dotted lines indicate always-calling.
\textbf{Right:} Actual tool calls without budget enforcement. Models do not reliably reduce or stop calls as cost increases, despite being provided with cost and remaining budget.
}
\label{fig:bfcl_affordability_combined_v2}
\vspace{-8pt}
\end{figure}

\begin{figure}
    \centering
    \includegraphics[width=\linewidth]{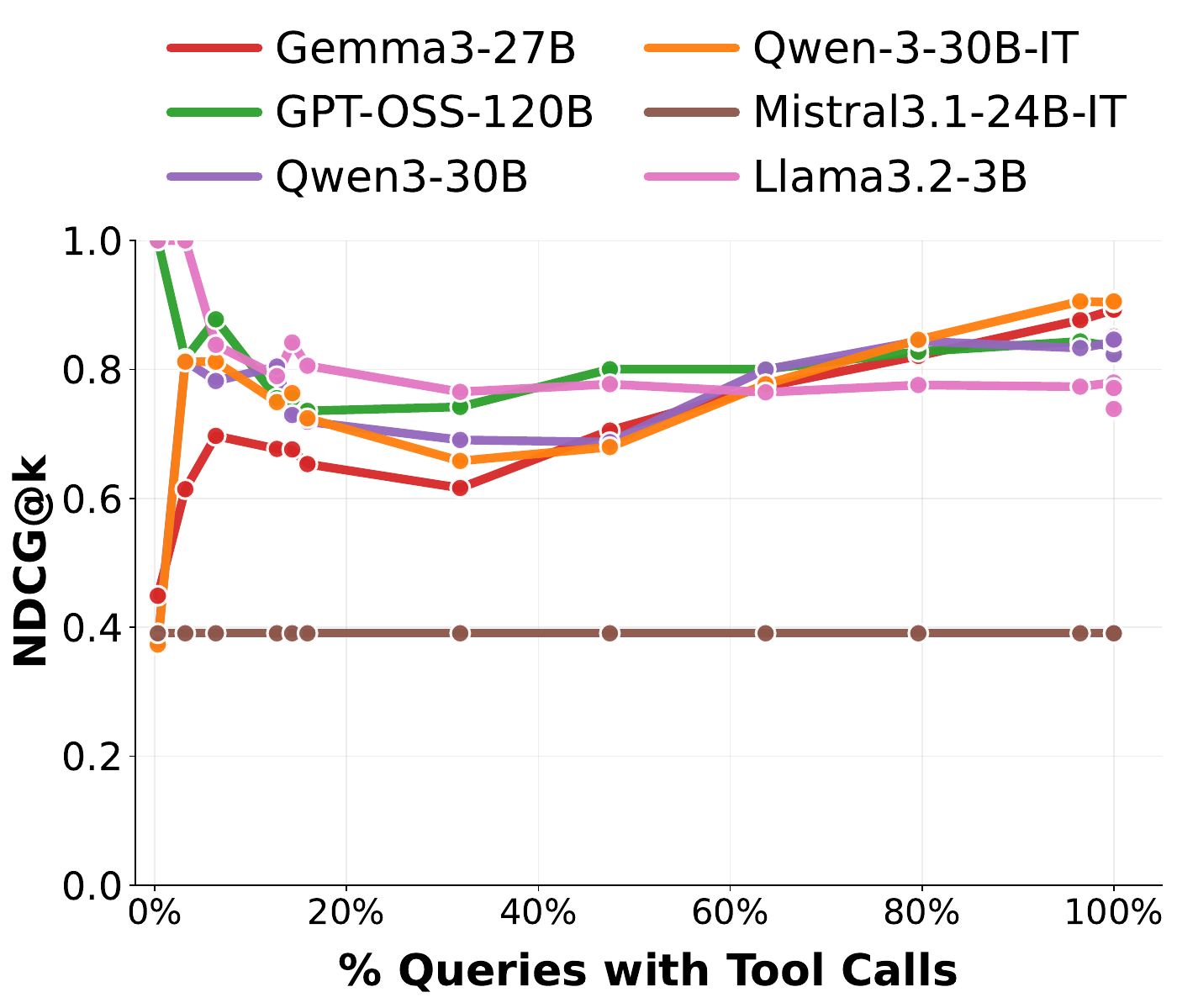}
    \caption{[BFCL Task] The NDCG rank correlation under different budgets across different models. The correlation is low, which reflects that the models are not choosing the best utility gain tool calling. Cost prompt v2.}
    \label{fig:bfcl-ndcg-v2}
\end{figure}

\subsection{Controller Framework}

\begin{figure}
    \centering
    \vspace{-10pt}
    \includegraphics[width=\linewidth]{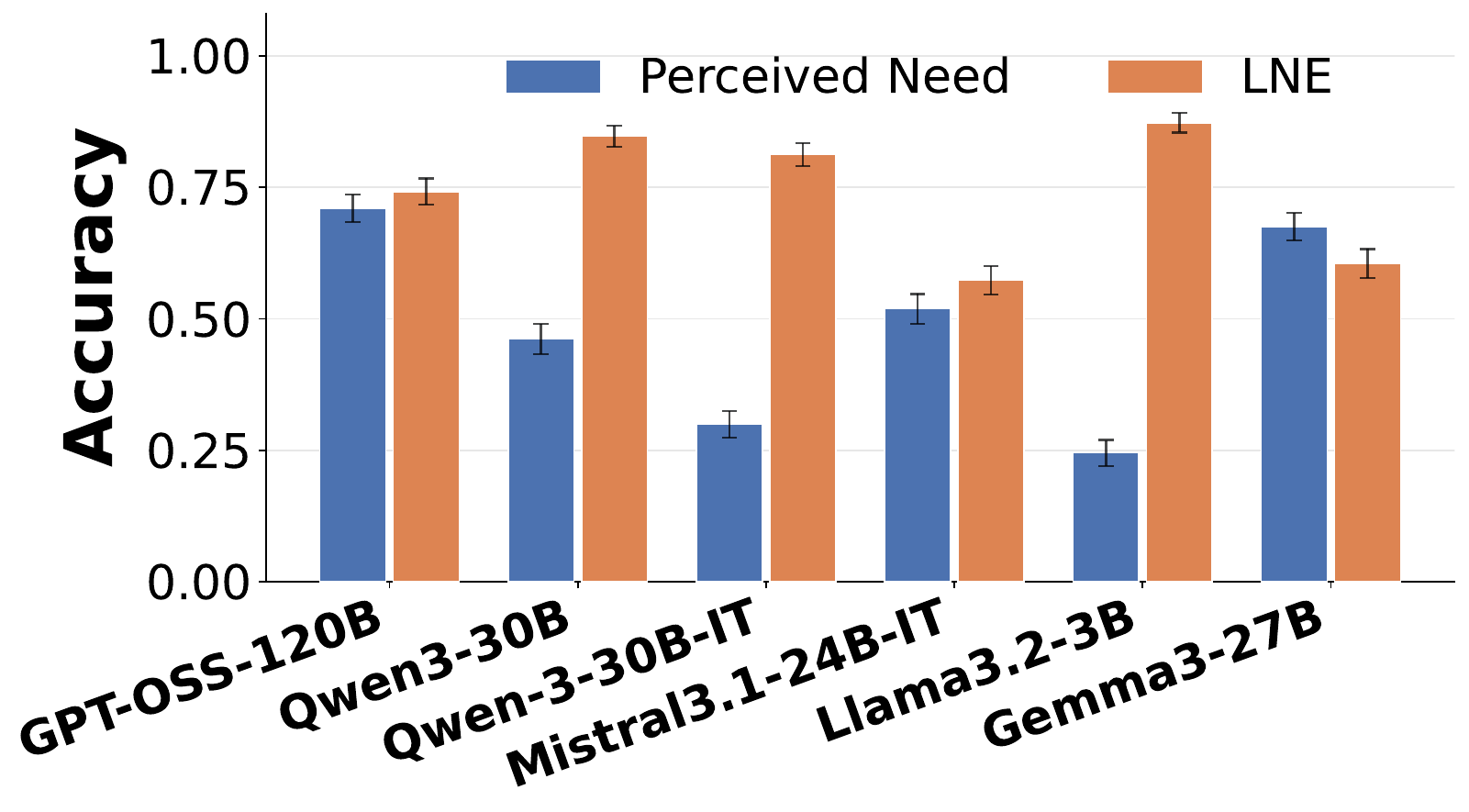}
    \caption{
       \textbf{BFCL task: The LNE can predict the \textit{True Need} more accurately across most models, especially for small and weaker models.}
    }
    \label{fig: bfcl_lne}
    \vspace{-10pt}
\end{figure}

\begin{figure}
    \centering

    \includegraphics[width=\linewidth]{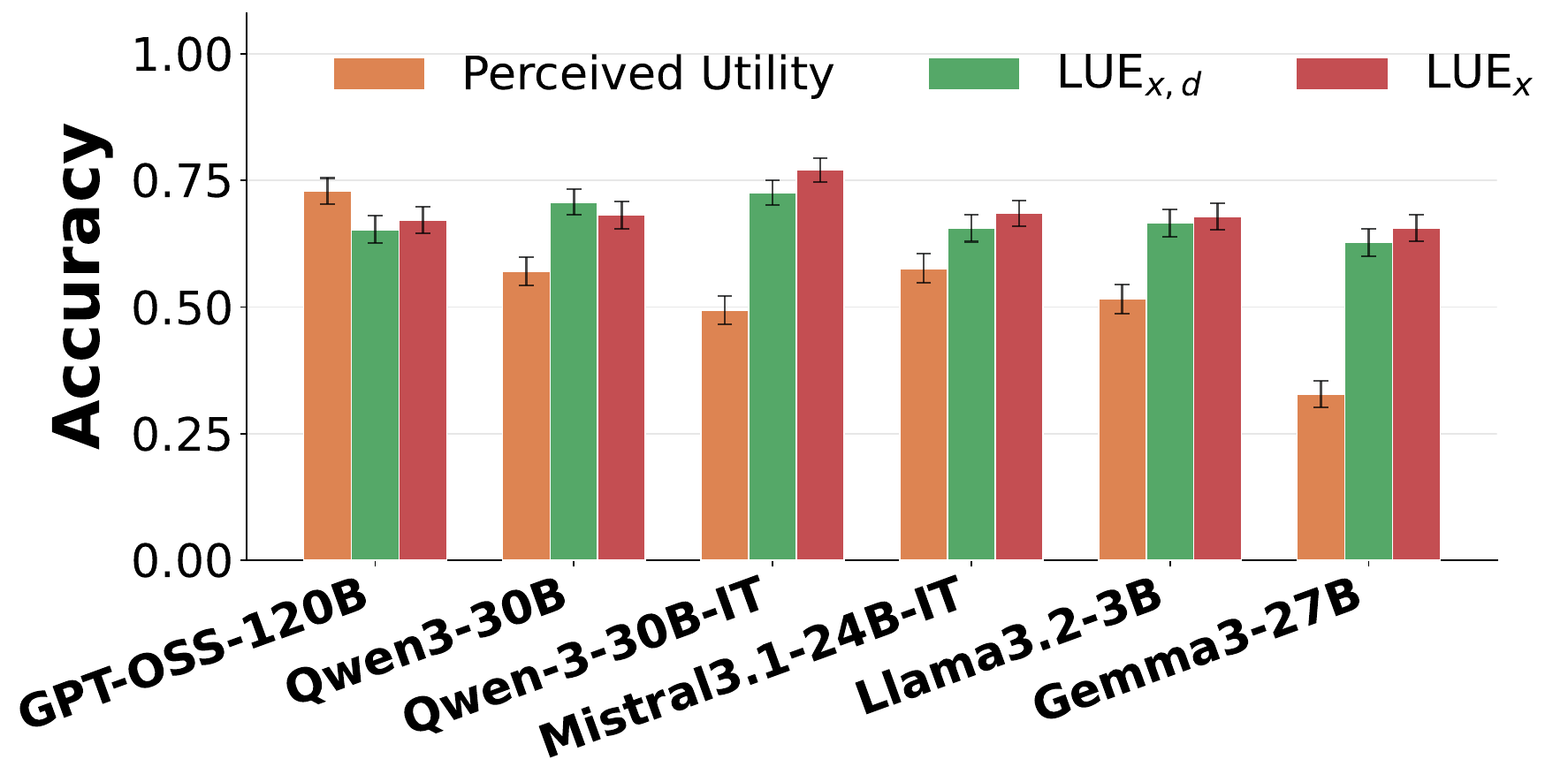}
    \caption{
       \textbf{The LUE can predict the \textit{True Utility} more accurately across most models, especially for small and weaker models.}.
    }
    \label{fig: bfcl_lue}

\end{figure}

\clearpage
\begin{figure}
    \centering
    \begin{subfigure}{\linewidth}
    \centering
    \includegraphics[width=\linewidth]{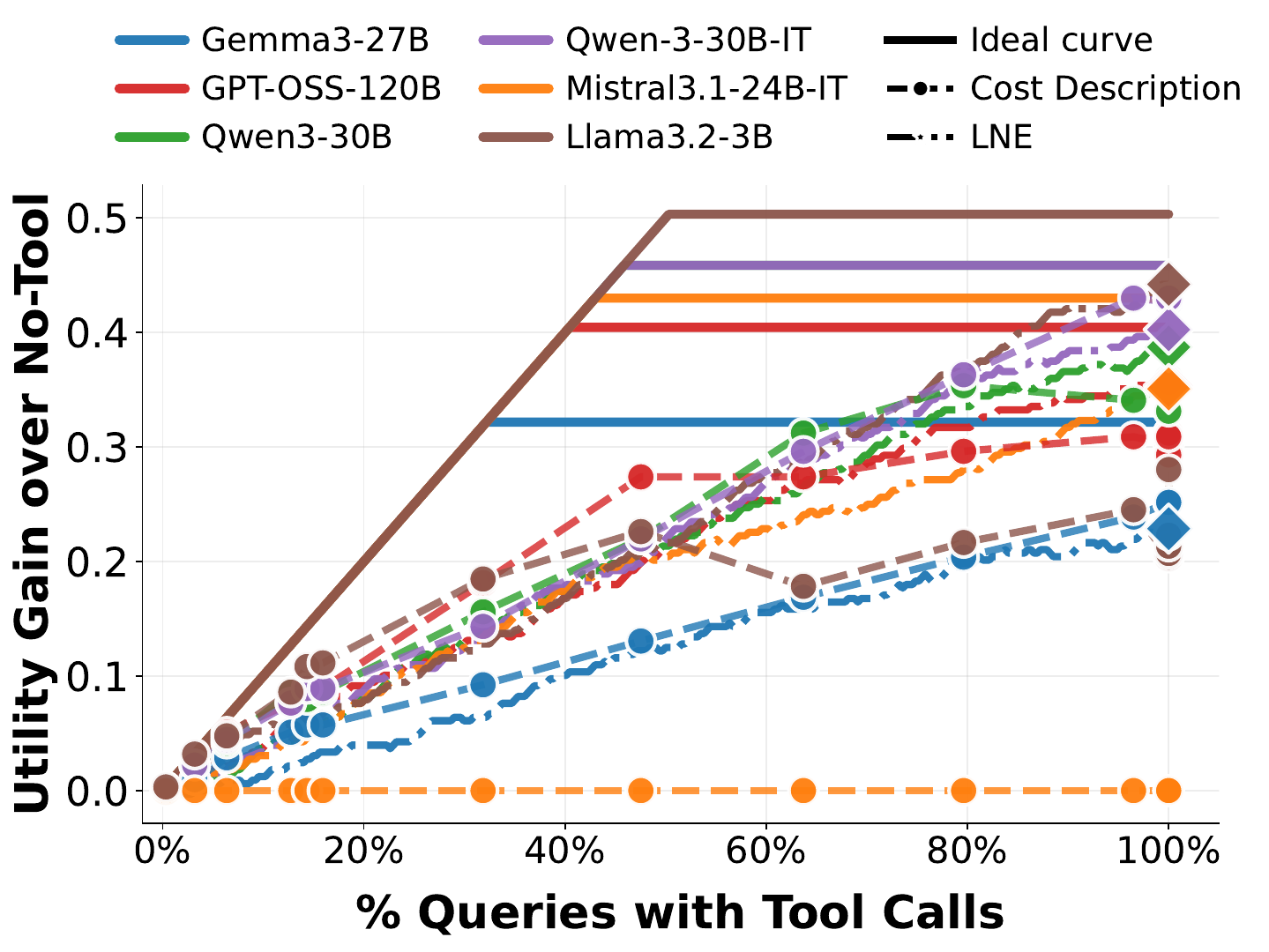}
    \caption{LNE}
    \end{subfigure}
    \begin{subfigure}{\linewidth}
    \centering
    \includegraphics[width=\linewidth]{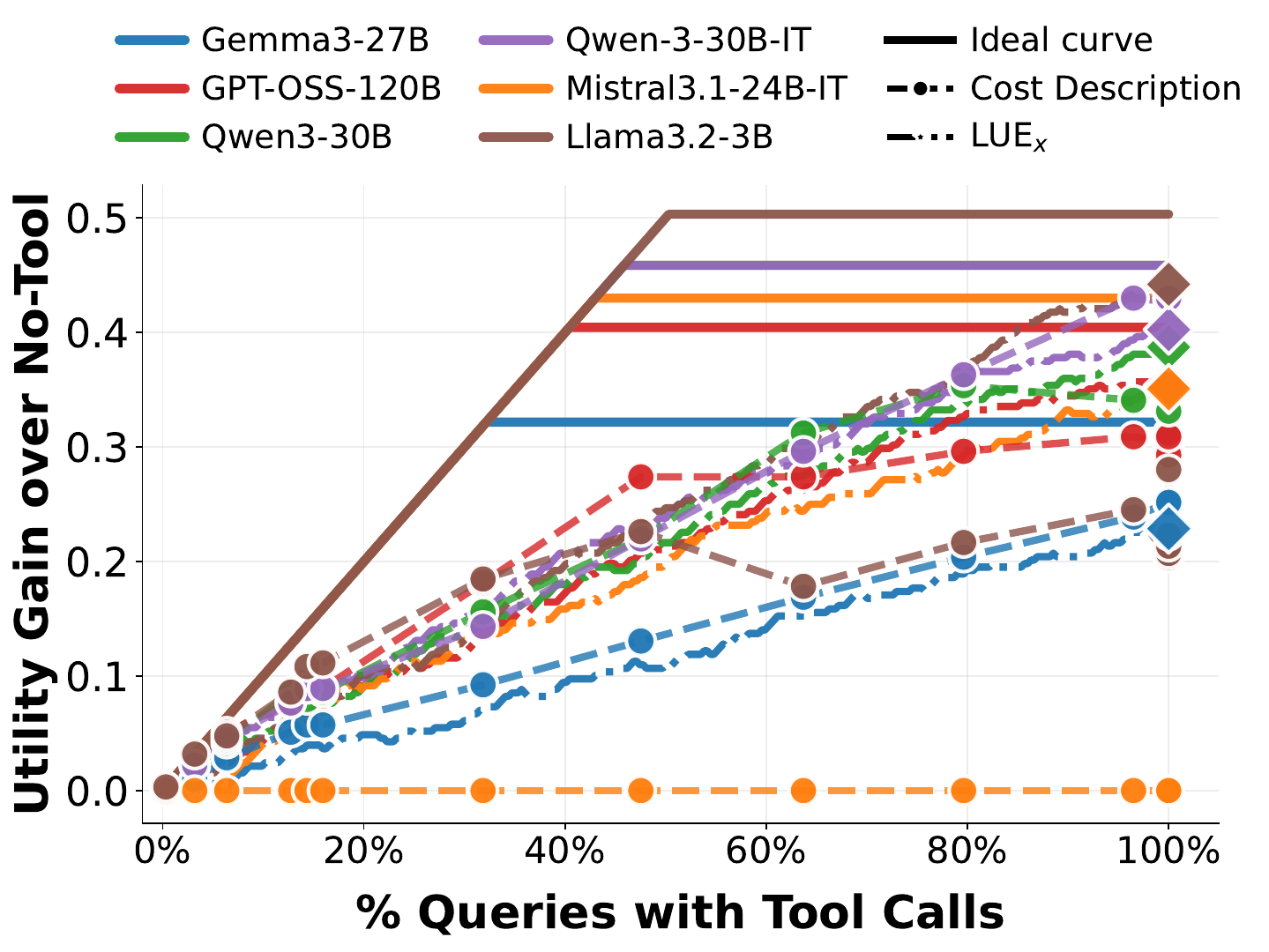}
    \caption{LUE$_x$}
    \end{subfigure}
    \begin{subfigure}{\linewidth}
    \centering
    \includegraphics[width=\linewidth]{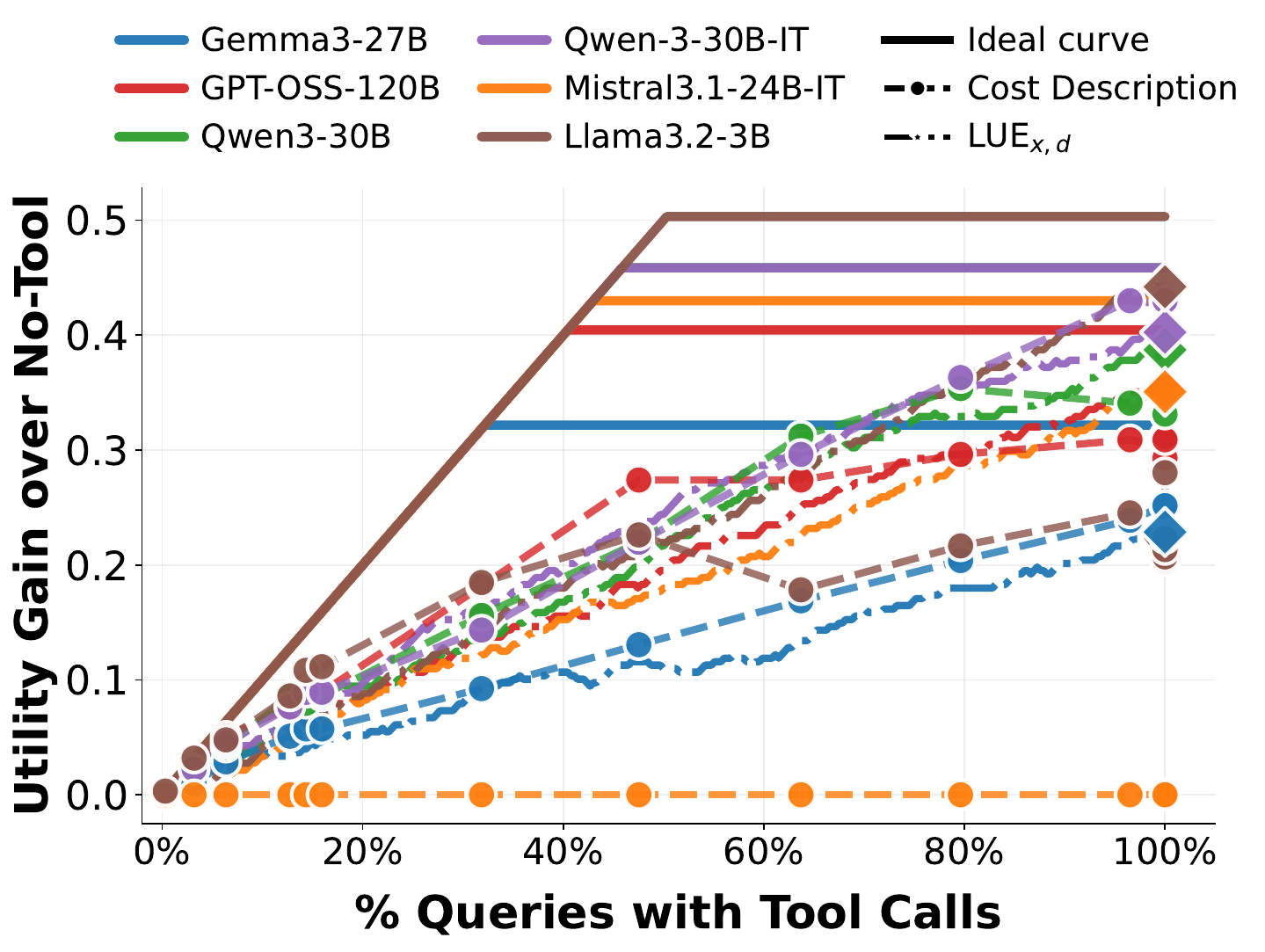}
    \caption{LUE$_{x,d}$}
    \end{subfigure}
\caption{[BFCL Task] Tool-call decisions are guided by the latent need estimator’s predicted probabilities under a fixed budget constraint.}
\label{fig:prediction_cost_bfcl}
\end{figure}

Tool-call decisions are guided by the latent need estimator’s predicted probabilities under a fixed budget constraint. In particular, we follow the predictor’s scores to rank instances by their likelihood of requiring tool use, and enable tool calling for the top-$k$ instances within a given budget. This strategy yields improved performance under the same budget compared to alternative allocation schemes. Figure~\ref{fig:prediction_cost_bfcl} illustrates the effectiveness of this approach across different budget levels.

\section{Calculator Task Results}
\label{sec:calculator_results}

\begin{figure*}[t]
    \centering
    \begin{subfigure}{0.32\textwidth}
        \centering
        \includegraphics[width=\linewidth]{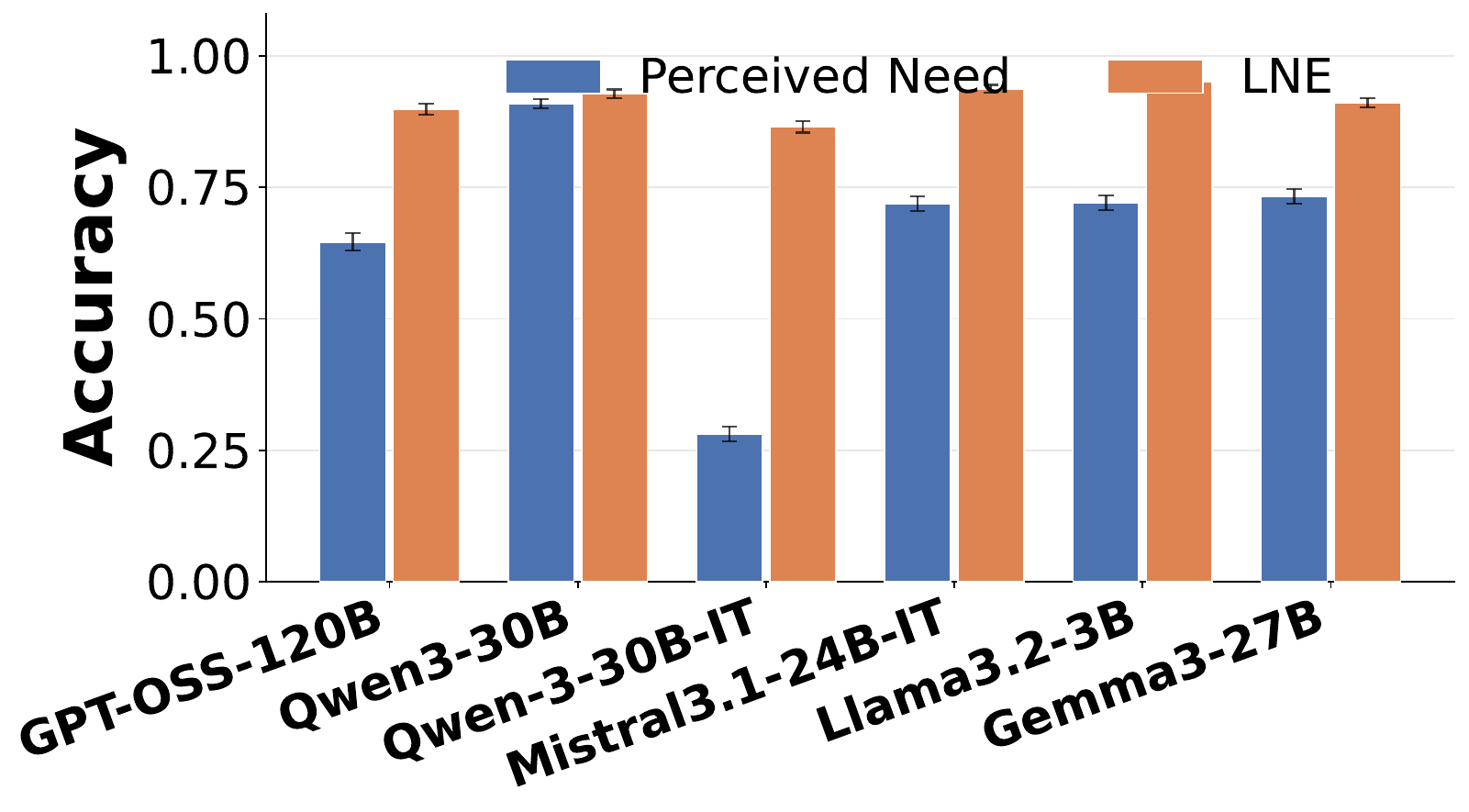}
        \caption{Synthetic Multiplication}
        \label{fig:synthetic_mult_lne_accuracy}
    \end{subfigure}\hfill
    \begin{subfigure}{0.32\textwidth}
        \centering
        \includegraphics[width=\linewidth]{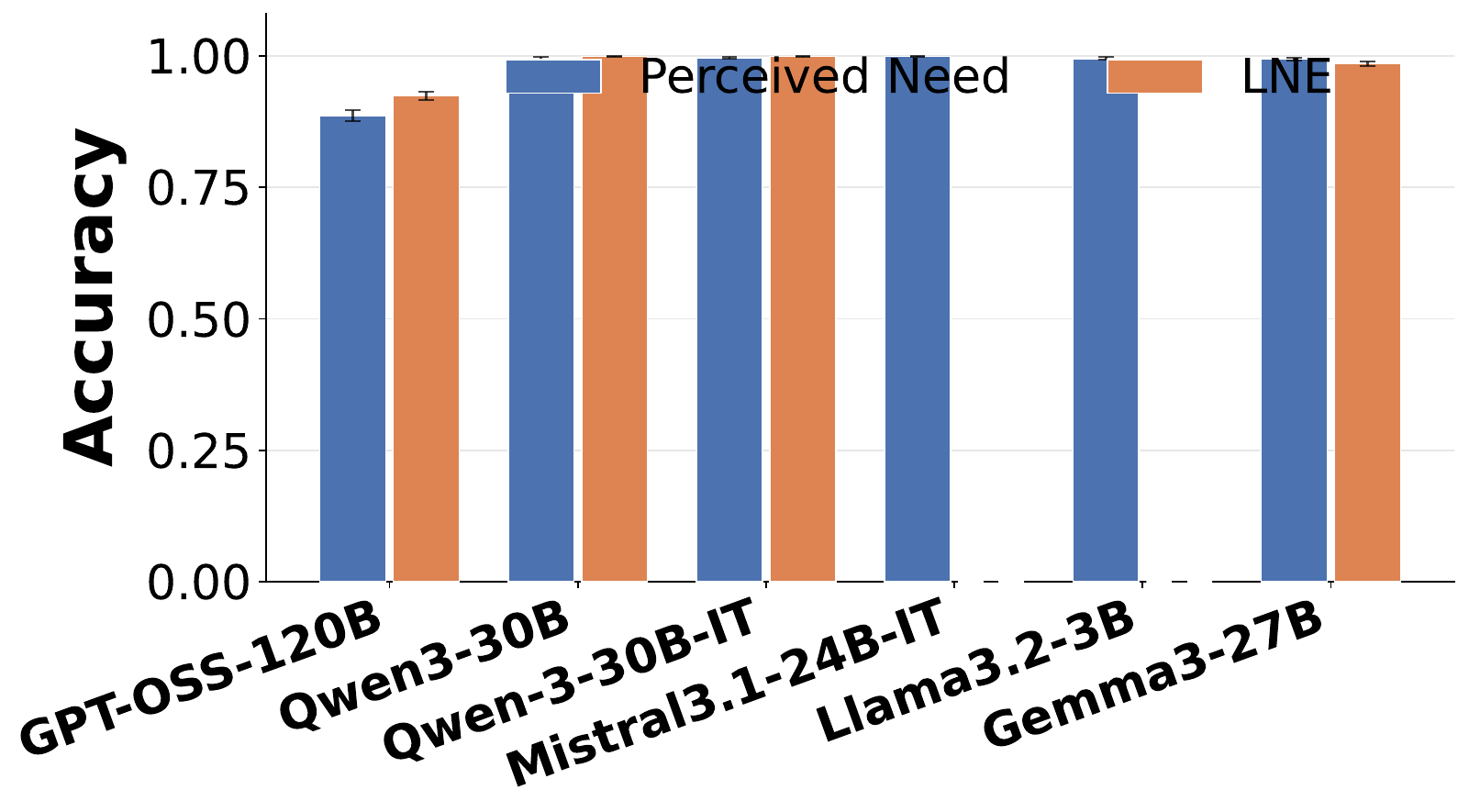}
        \caption{Synthetic Large-Digit Multiplication}
        \label{fig:synthetic_nn_lne_accuracy}
    \end{subfigure}
    \caption{\textbf{Calculator tasks: LNE (Predictor 1) accuracy for true-need prediction.} Accuracy is compared with each model's perceived-need judgment. Synthetic Large-Digit Multiplication model--task pairs with degenerate true-need labels are omitted as explained in Section~\ref{par:synthetic-nn-lne-omission}.}
    \label{fig:calculator_lne_accuracy}
\end{figure*}

This appendix reports need and utility results for GSM-Hard, Synthetic Multiplication, and Synthetic Large-Digit Multiplication. Figure~\ref{fig:calculator_lne_accuracy} summarizes LNE (Predictor 1) true-need prediction accuracy across the three tasks. All panels use exact-match correctness. In the normative matrices, rows indicate correctness without the calculator and columns indicate correctness with it; the upper-right cell is positive utility and the lower-left cell is negative utility. In the descriptive matrices, a tool call is the model's perceived positive utility.

\newcommand{\calcsevenpanels}[8]{%
\begin{figure*}[t]\centering
\begin{subfigure}{0.135\textwidth}\centering #2\caption{GPT-OSS}\end{subfigure}\hfill
\begin{subfigure}{0.135\textwidth}\centering #3\caption{Qwen3-A3B}\end{subfigure}\hfill
\begin{subfigure}{0.135\textwidth}\centering #4\caption{Qwen3-IT}\end{subfigure}\hfill
\begin{subfigure}{0.135\textwidth}\centering #5\caption{Gemma3-IT}\end{subfigure}\hfill
\begin{subfigure}{0.135\textwidth}\centering #6\caption{Mistral-IT}\end{subfigure}\hfill
\begin{subfigure}{0.135\textwidth}\centering #7\caption{Llama3.2-IT}\end{subfigure}\hfill
\begin{subfigure}{0.135\textwidth}\centering #8\caption{GPT-5.5}\end{subfigure}
\caption{#1}\end{figure*}}
\newcommand{\calcnormativepanels}[8]{%
\begin{figure*}[t]\centering
\begin{subfigure}{0.22\textwidth}\centering #2\caption{GPT-OSS}\end{subfigure}\hfill
\begin{subfigure}{0.22\textwidth}\centering #3\caption{Qwen3-A3B}\end{subfigure}\hfill
\begin{subfigure}{0.22\textwidth}\centering #4\caption{Qwen3-IT}\end{subfigure}\hfill
\begin{subfigure}{0.22\textwidth}\centering #5\caption{Gemma3-IT}\end{subfigure}\par\medskip
\begin{subfigure}{0.22\textwidth}\centering #6\caption{Mistral-IT}\end{subfigure}\hspace{0.04\textwidth}
\begin{subfigure}{0.22\textwidth}\centering #7\caption{Llama3.2-IT}\end{subfigure}\hspace{0.04\textwidth}
\begin{subfigure}{0.22\textwidth}\centering #8\caption{GPT-5.5}\end{subfigure}
\caption{#1}\end{figure*}}

\subsection{GSM-Hard}
\paragraph{Normative observation.}
The calculator produces positive utility on otherwise incorrect answers for every model, but GSM-Hard is not a pure calculation task: correct tool use still depends on identifying and executing the appropriate reasoning steps. Consequently, negative utility remains visible, particularly for Gemma3-IT, Mistral-IT, and Llama3.2-IT. Llama3.2-IT gains on only 10 instances while 219 previously correct answers become incorrect, a marked integration failure that differs from the generally reliable synthetic-task behavior.

\calcnormativepanels{\textbf{GSM-Hard: No-Tool vs. Always-Tool correctness.}\label{fig:gsmhard_actual_need_utility}}
{\includegraphics[width=\linewidth]{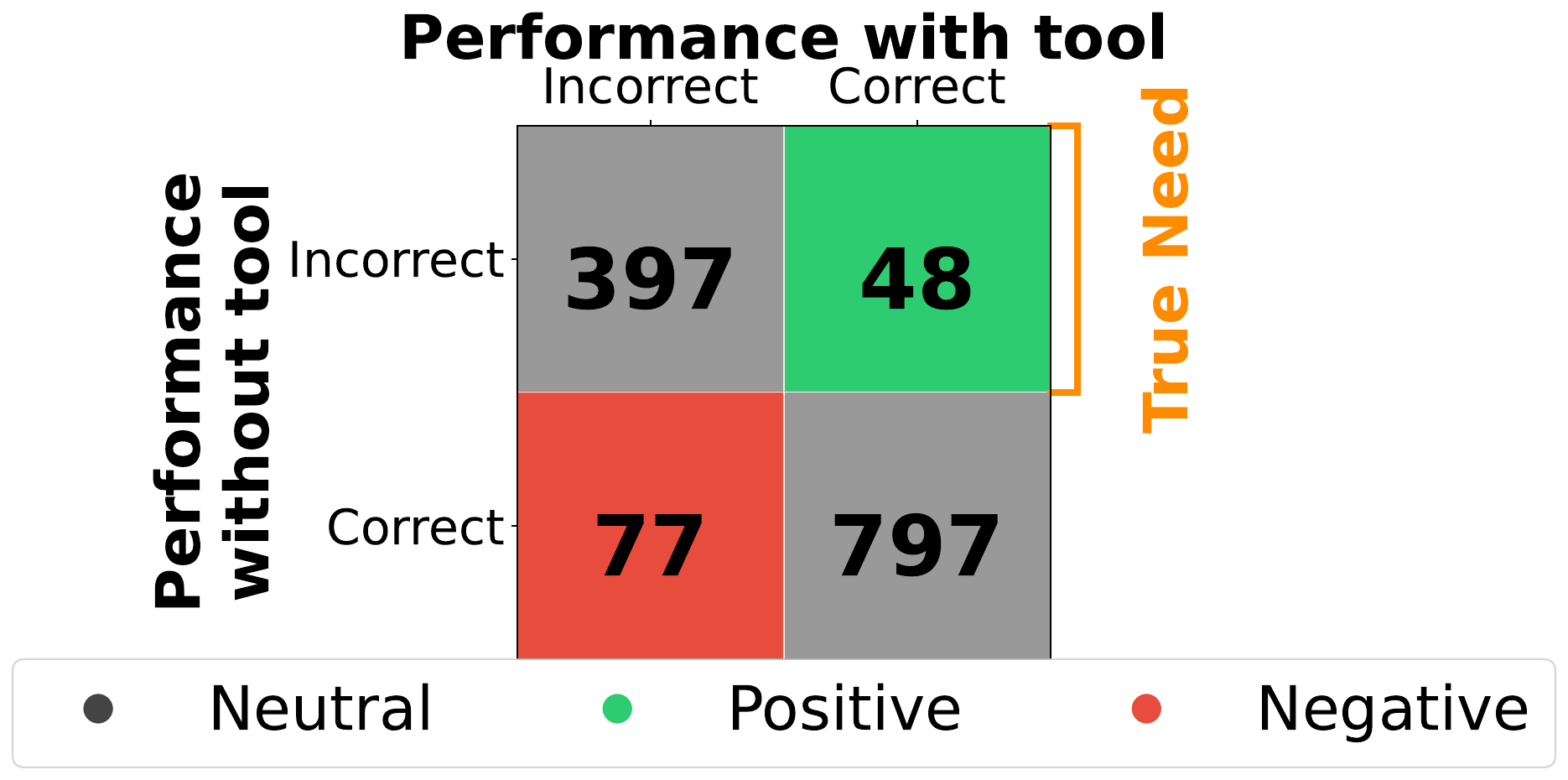}}
{\includegraphics[width=\linewidth]{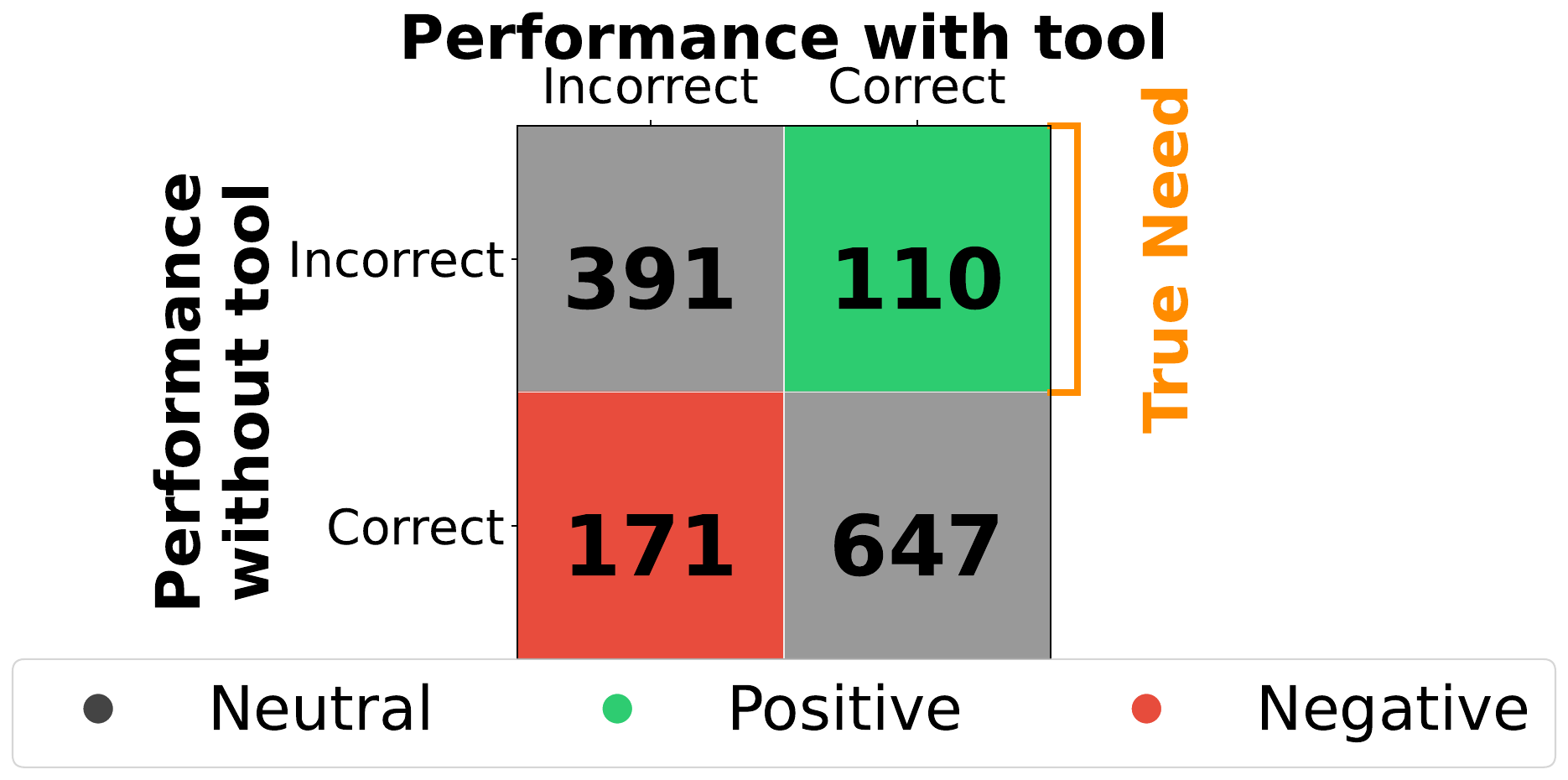}}
{\includegraphics[width=\linewidth]{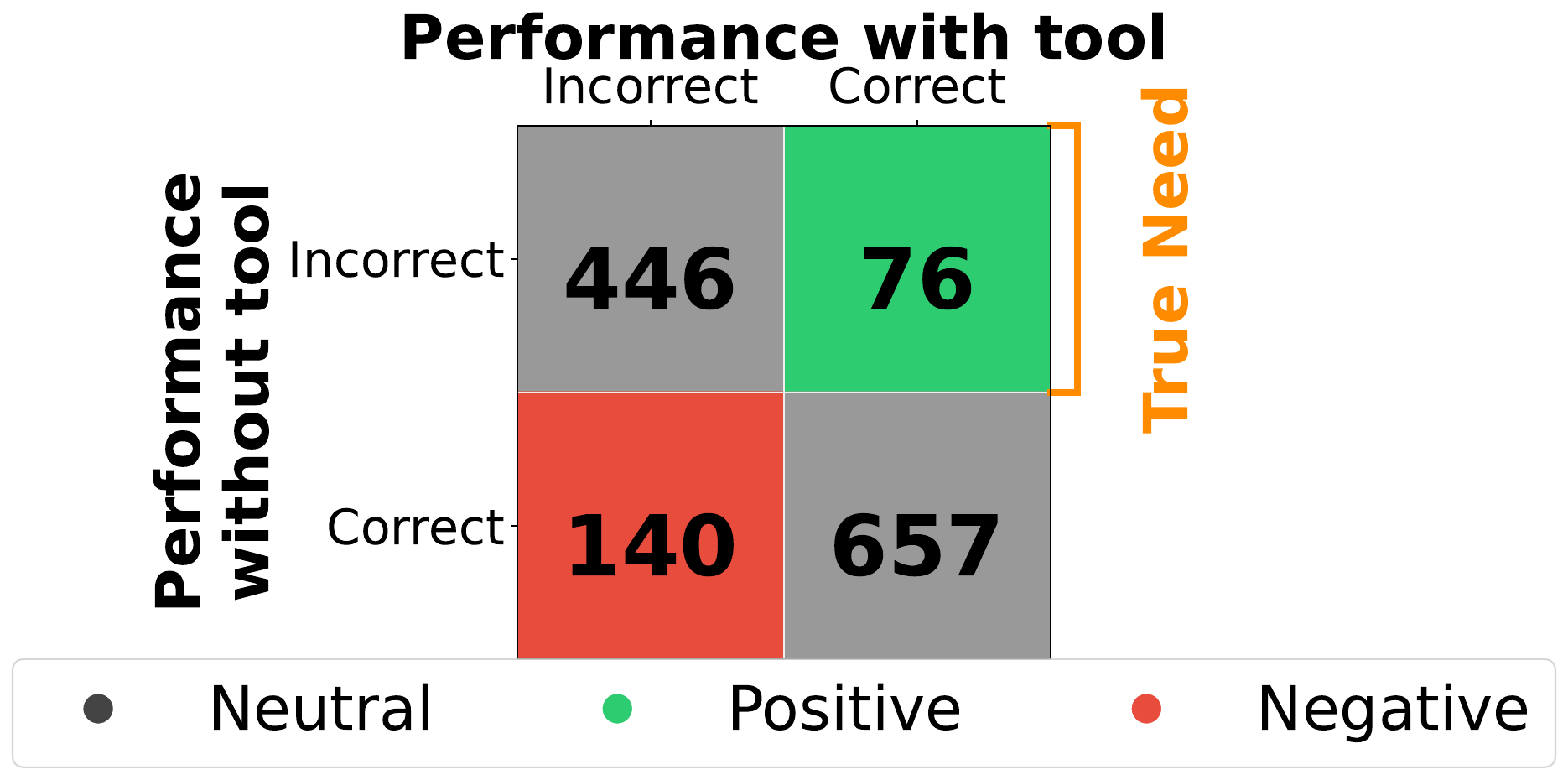}}
{\includegraphics[width=\linewidth]{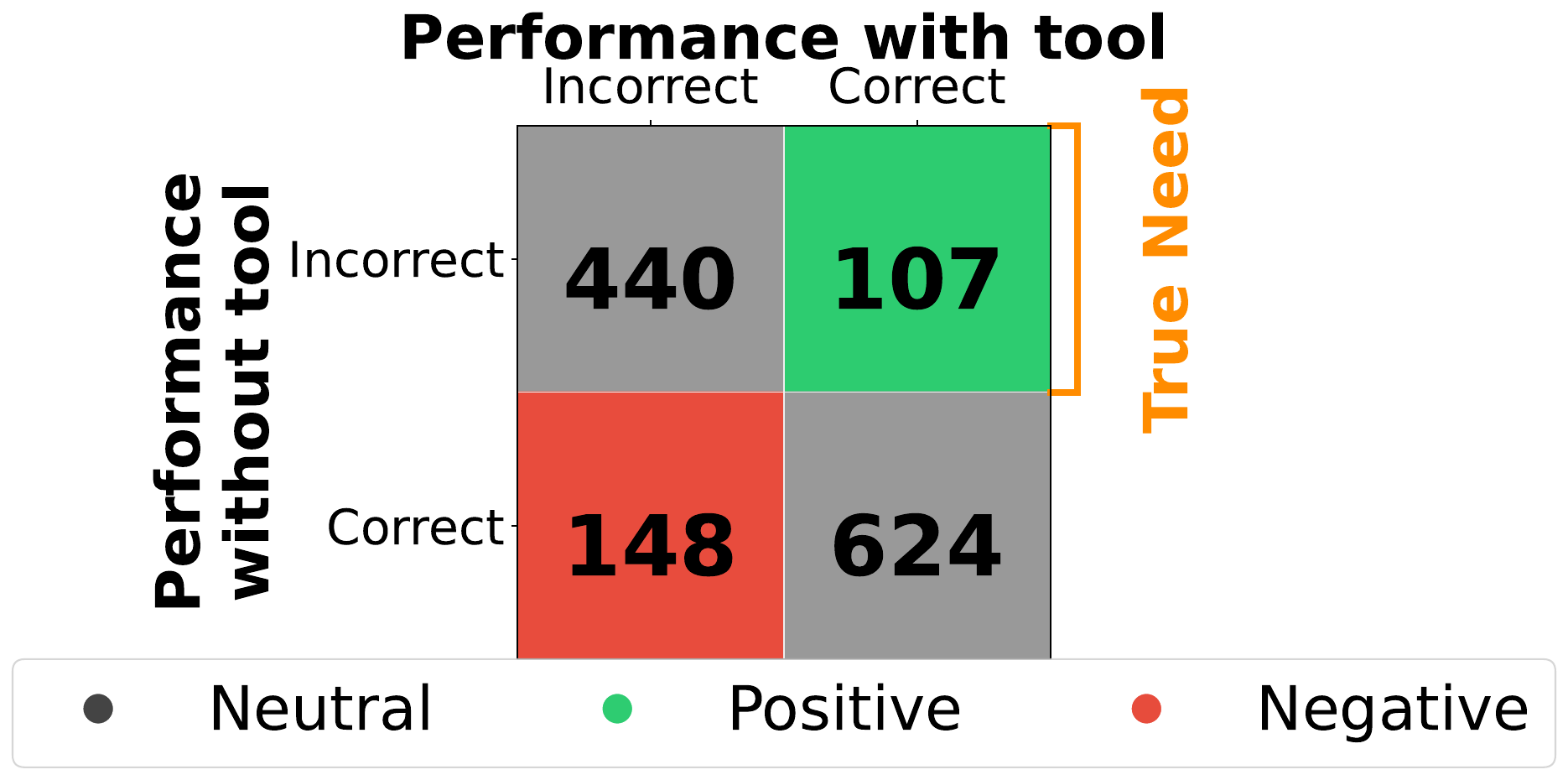}}
{\includegraphics[width=\linewidth]{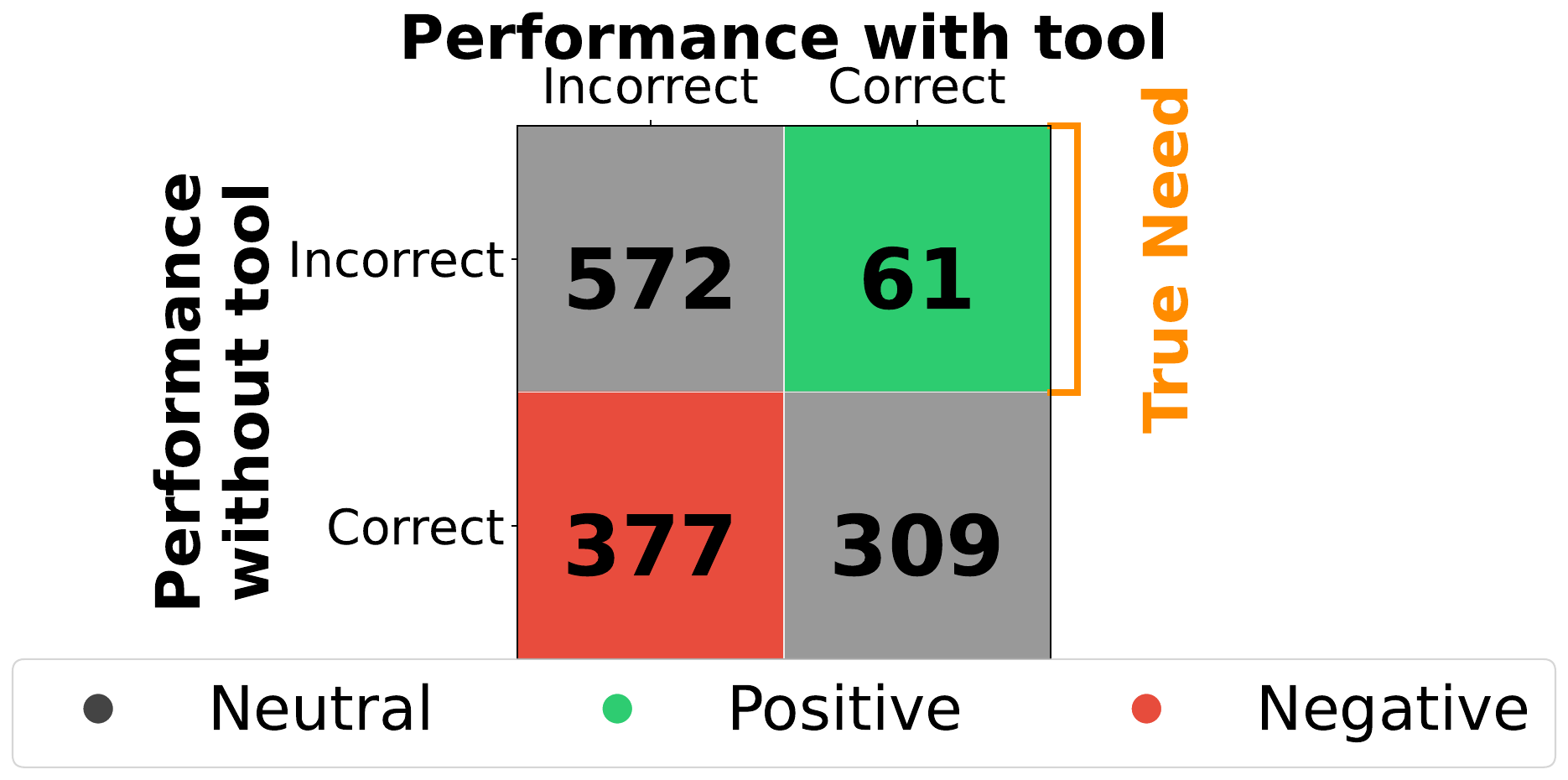}}
{\includegraphics[width=\linewidth]{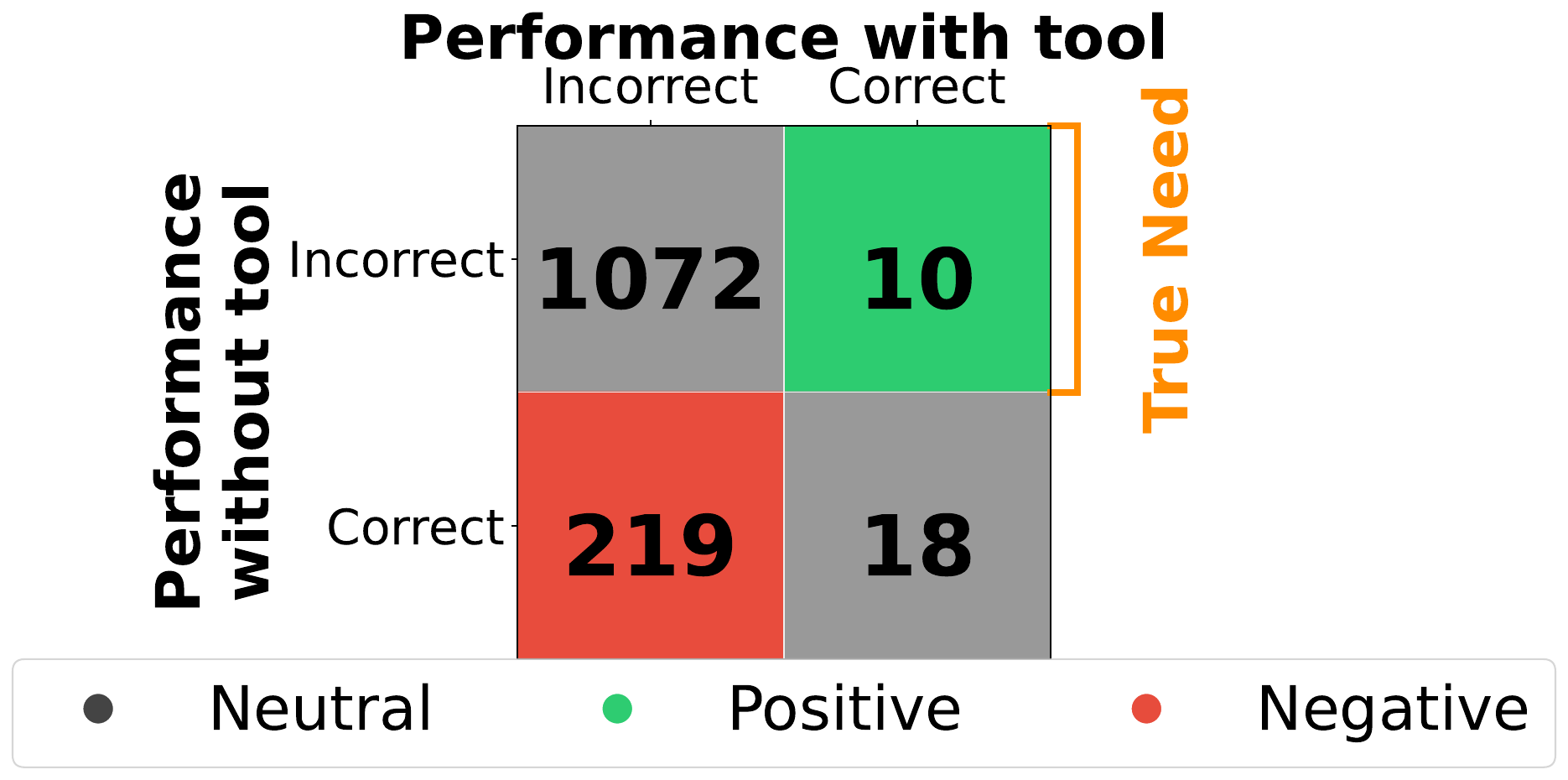}}
{\includegraphics[width=\linewidth]{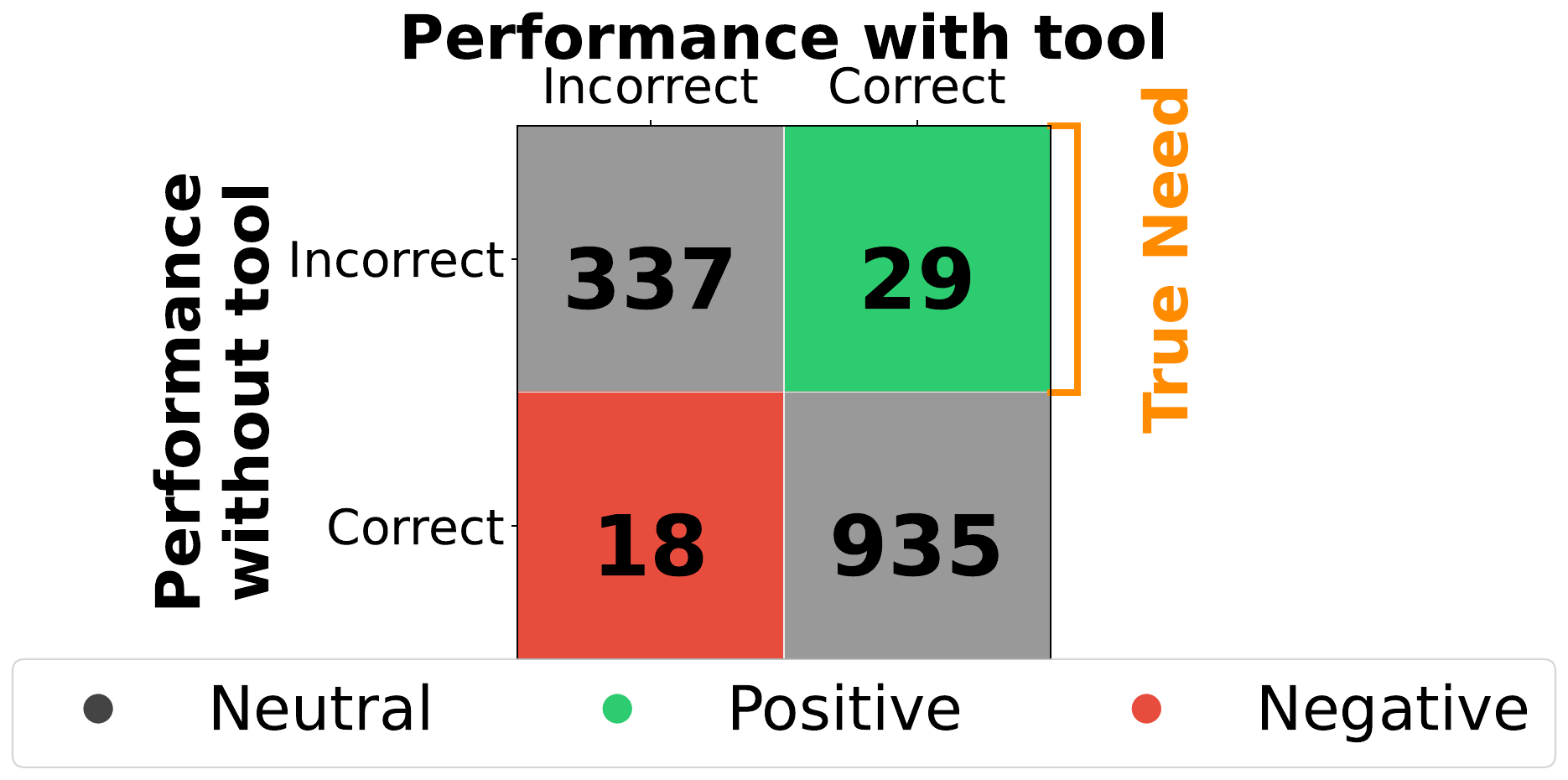}}

\paragraph{Descriptive observation.}
Tool calling is not calibrated to true positive utility. Gemma3-IT and Llama3.2-IT call on almost all non-positive-utility instances, while GPT-OSS, Mistral-IT, and GPT-5.5 omit many beneficial calls. GPT-5.5 also provides direct evidence that perceived need and calling are distinct: it reports need rarely, yet sometimes invokes the calculator while reporting no need.

\begin{figure*}[t]
    \centering
    \includegraphics[width=0.78\textwidth]{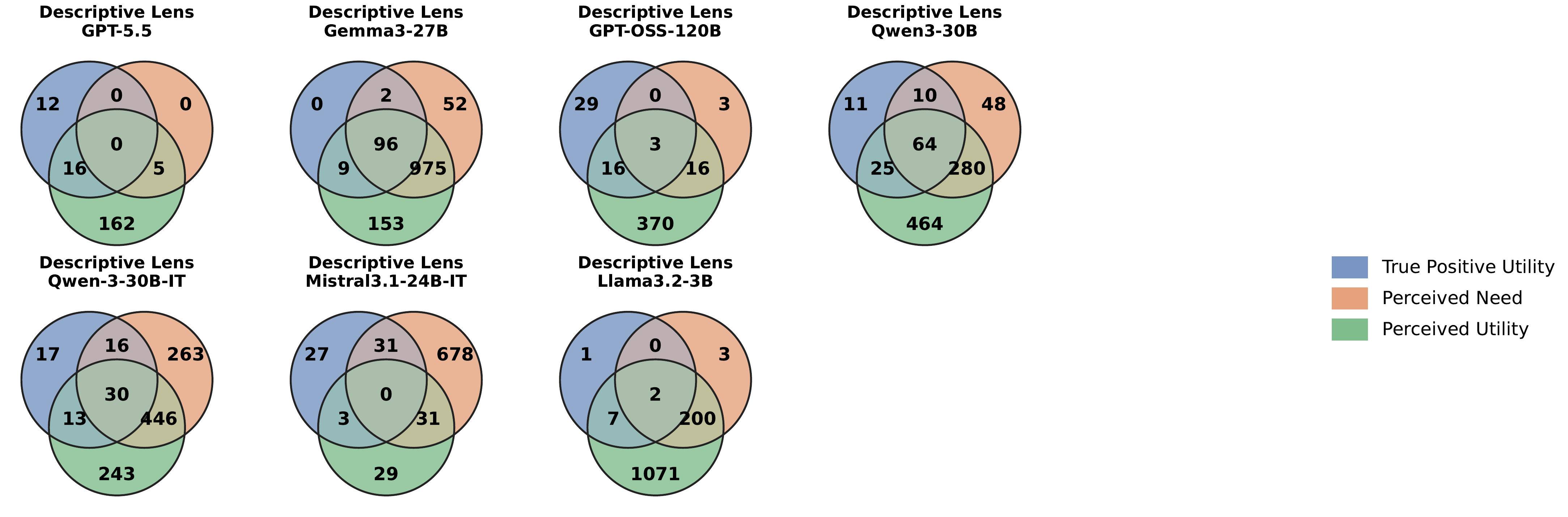}
    \caption{[GSM-Hard Task] Venn diagrams of \textbf{True Positive Utility, Perceived Need, and Perceived Utility}. Each panel shows their empirical overlap for one model. Calls outside true positive utility are non-beneficial; true-positive-utility cases outside perceived utility are missed opportunities. Perceived need is a separate self-assessment and need not be nested within either utility set.}
    \label{fig:venn-gsmhard}
\end{figure*}

\calcsevenpanels{\textbf{GSM-Hard: perceived need vs. calculator calling.}\label{fig:gsmhard_perceived_need_utility}}
{\includegraphics[width=\linewidth]{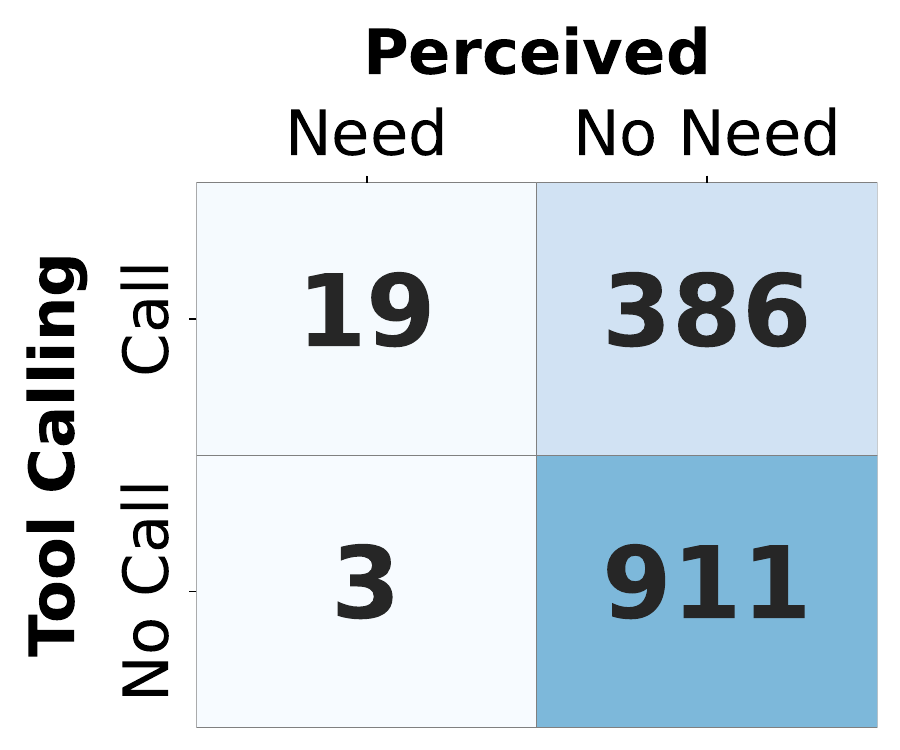}}{\includegraphics[width=\linewidth]{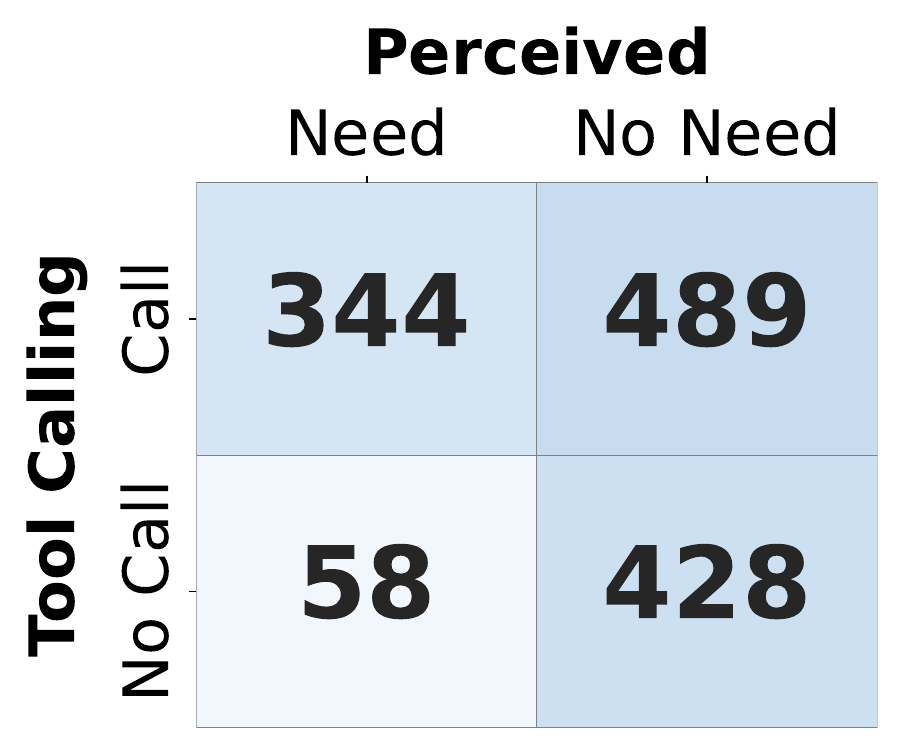}}{\includegraphics[width=\linewidth]{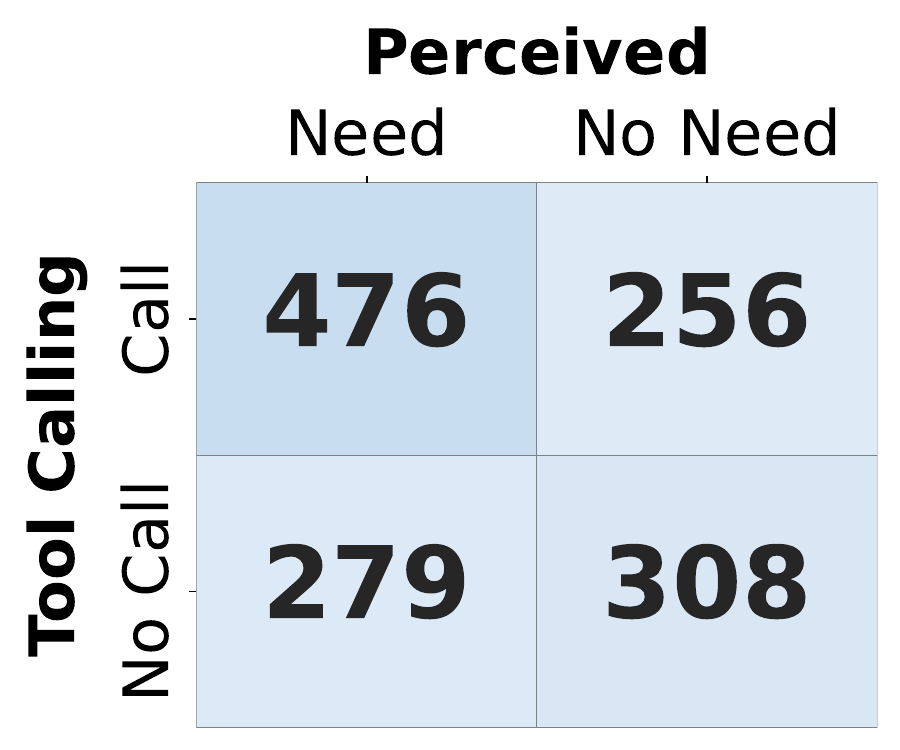}}{\includegraphics[width=\linewidth]{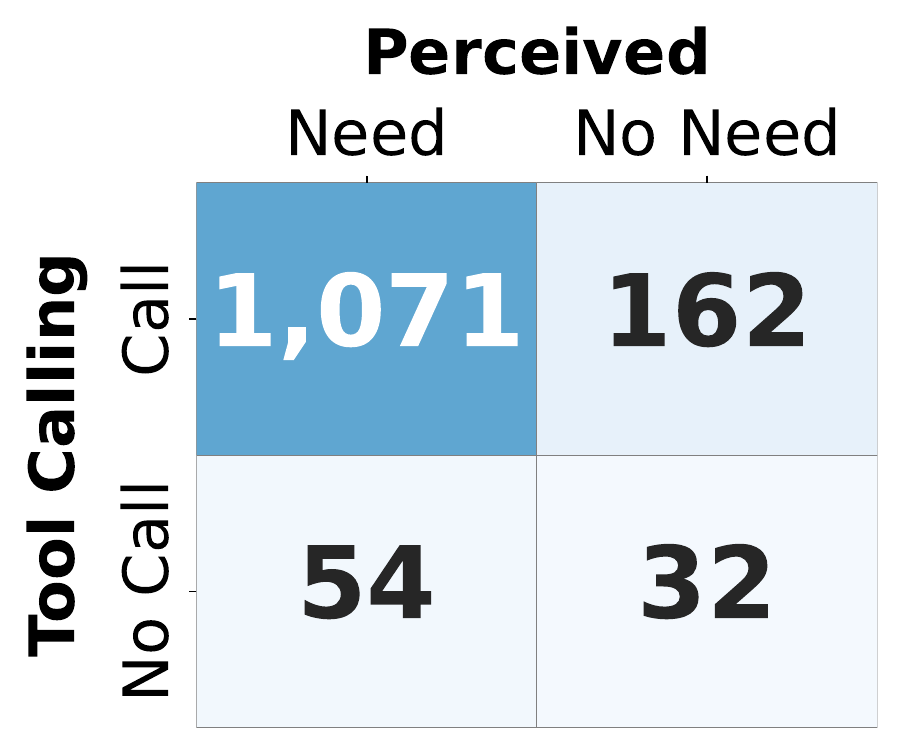}}{\includegraphics[width=\linewidth]{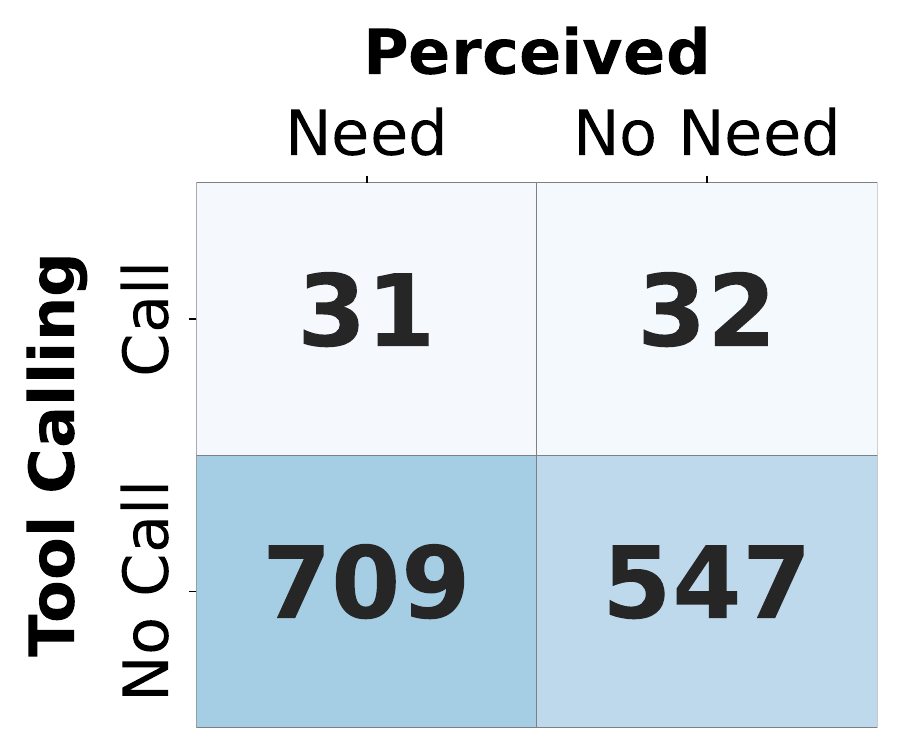}}{\includegraphics[width=\linewidth]{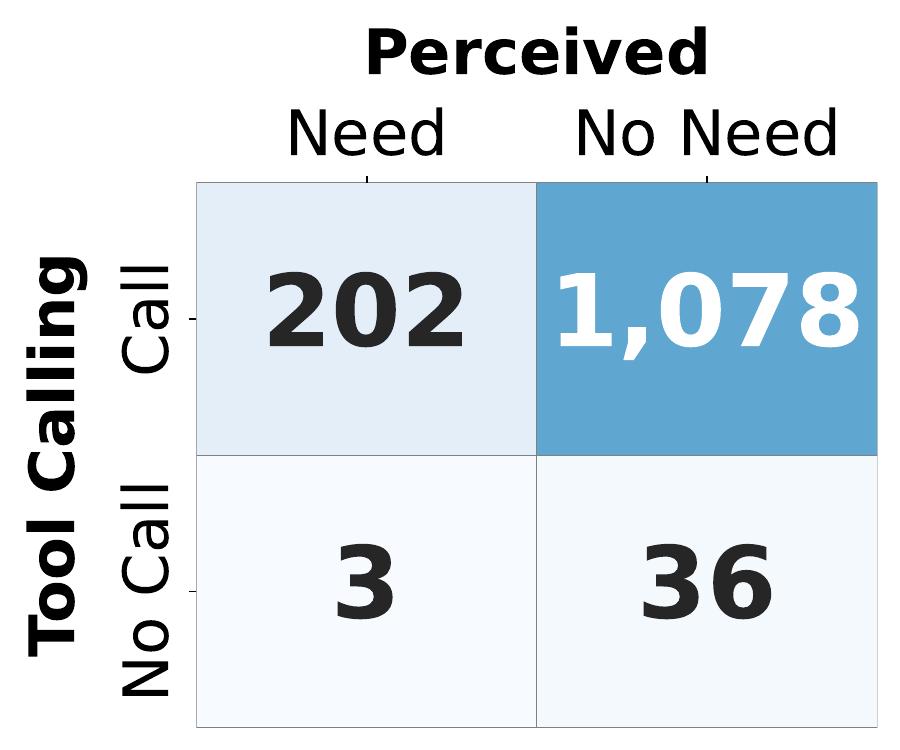}}
{\includegraphics[width=\linewidth]{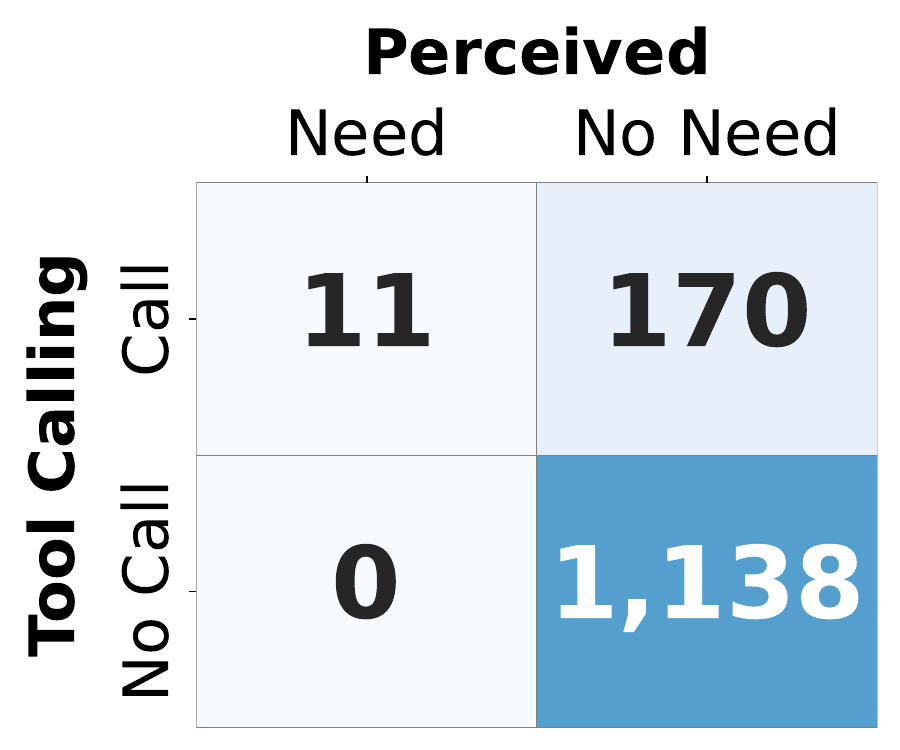}}

\calcsevenpanels{\textbf{GSM-Hard: perceived need.}\label{fig:gsmhard_perceived_need}}
{\includegraphics[width=\linewidth]{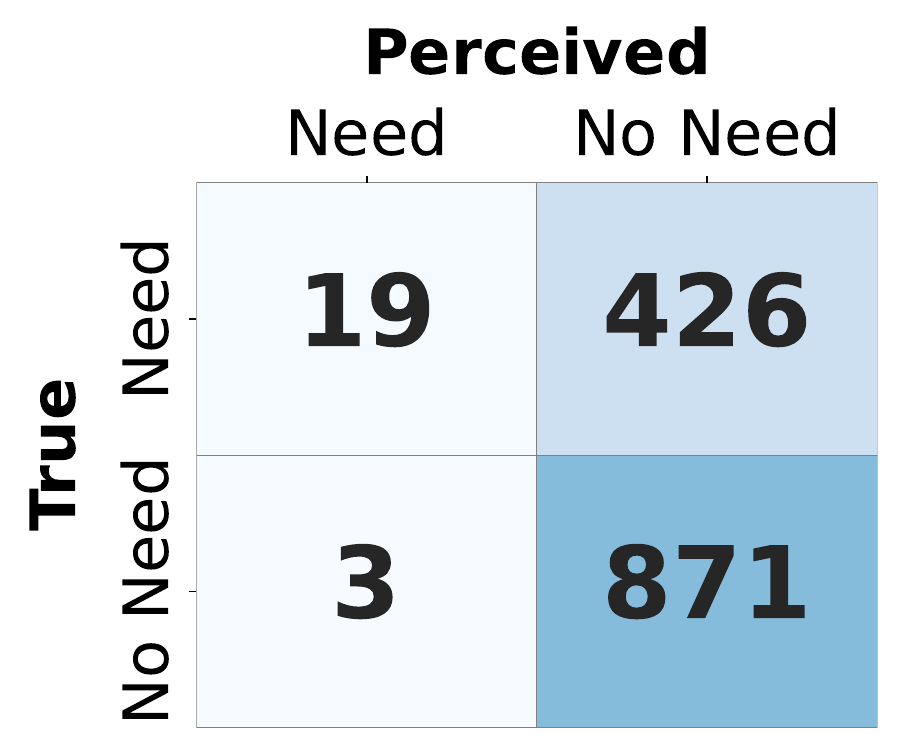}}{\includegraphics[width=\linewidth]{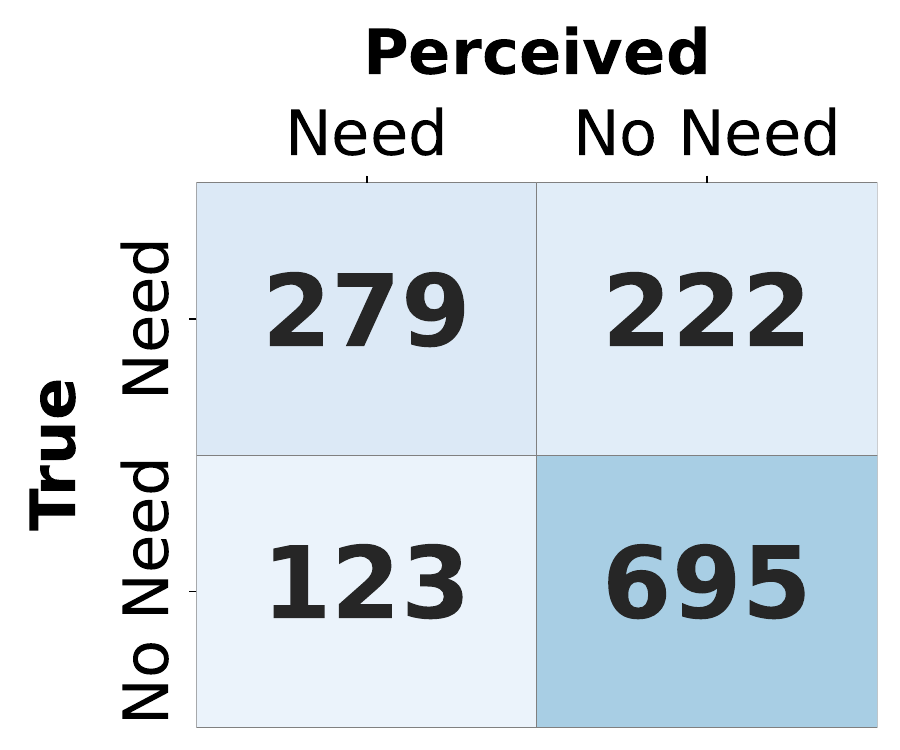}}{\includegraphics[width=\linewidth]{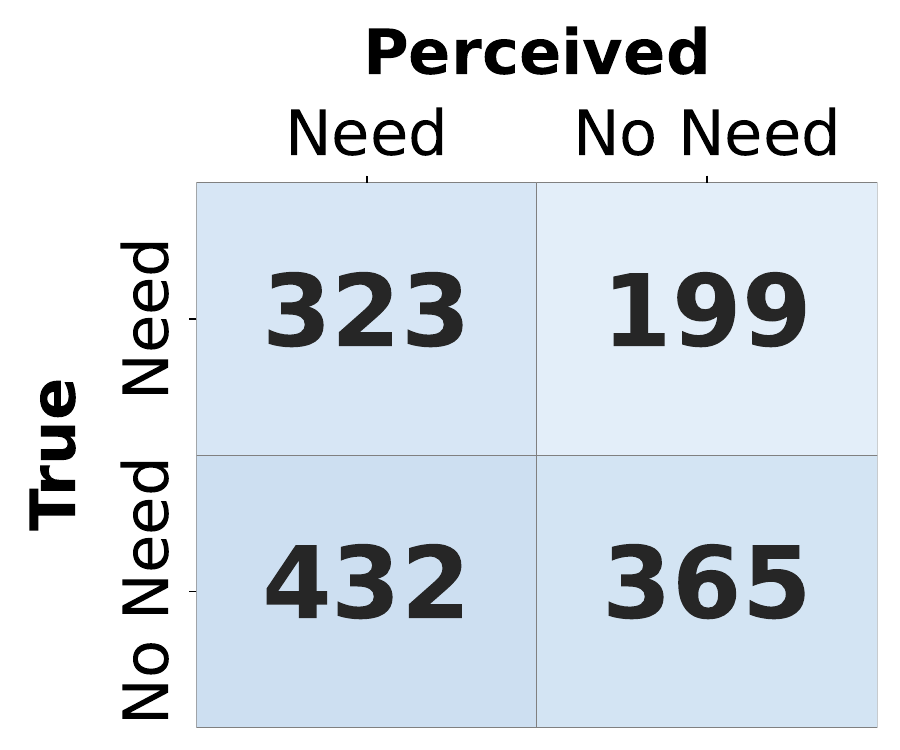}}{\includegraphics[width=\linewidth]{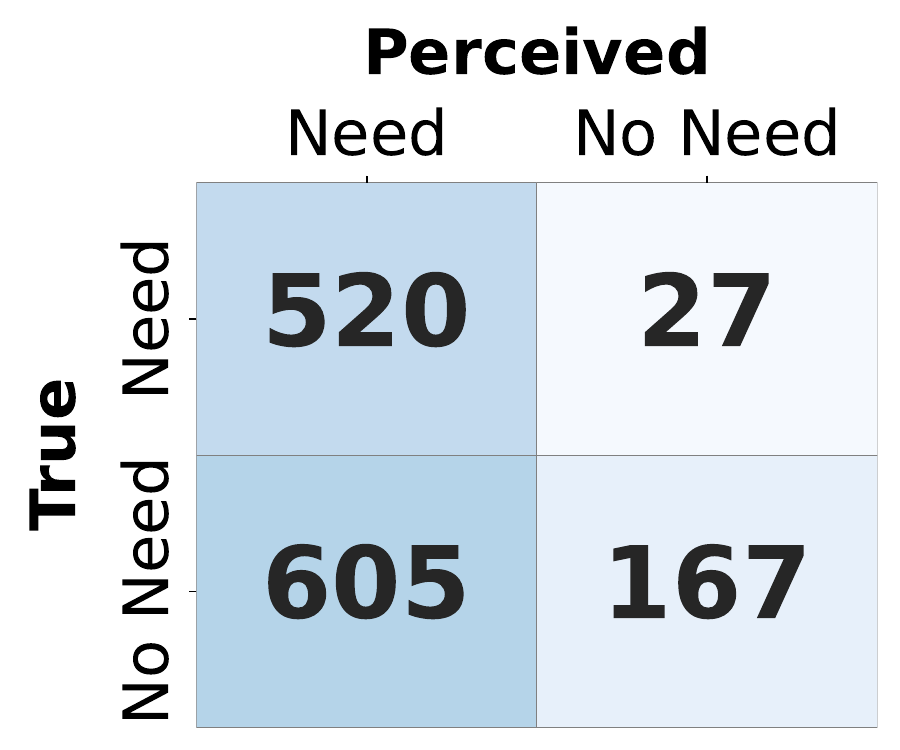}}{\includegraphics[width=\linewidth]{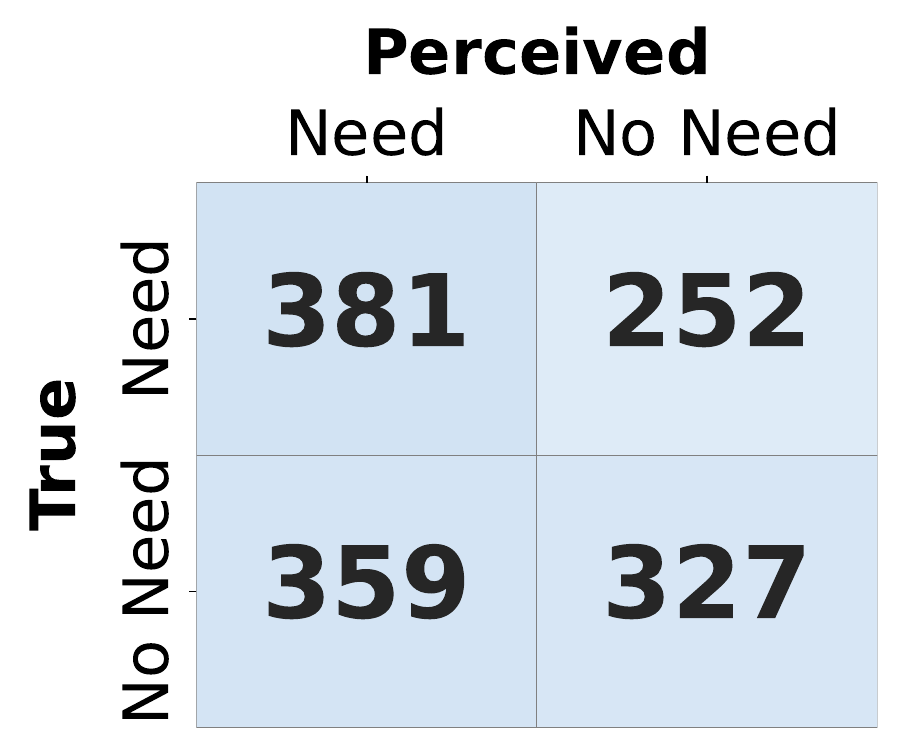}}{\includegraphics[width=\linewidth]{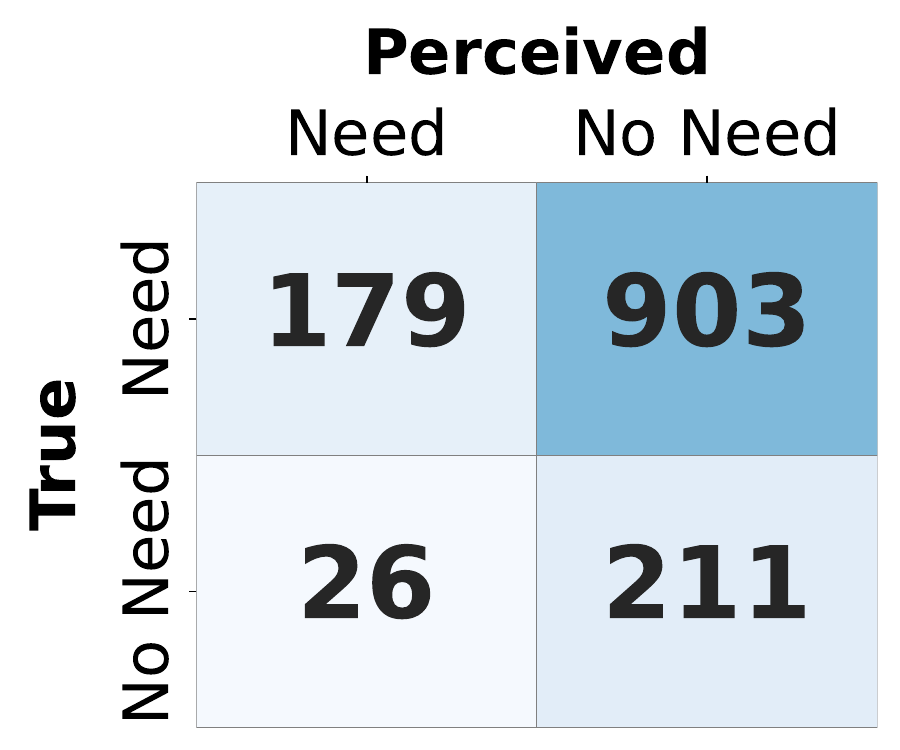}}
{\includegraphics[width=\linewidth]{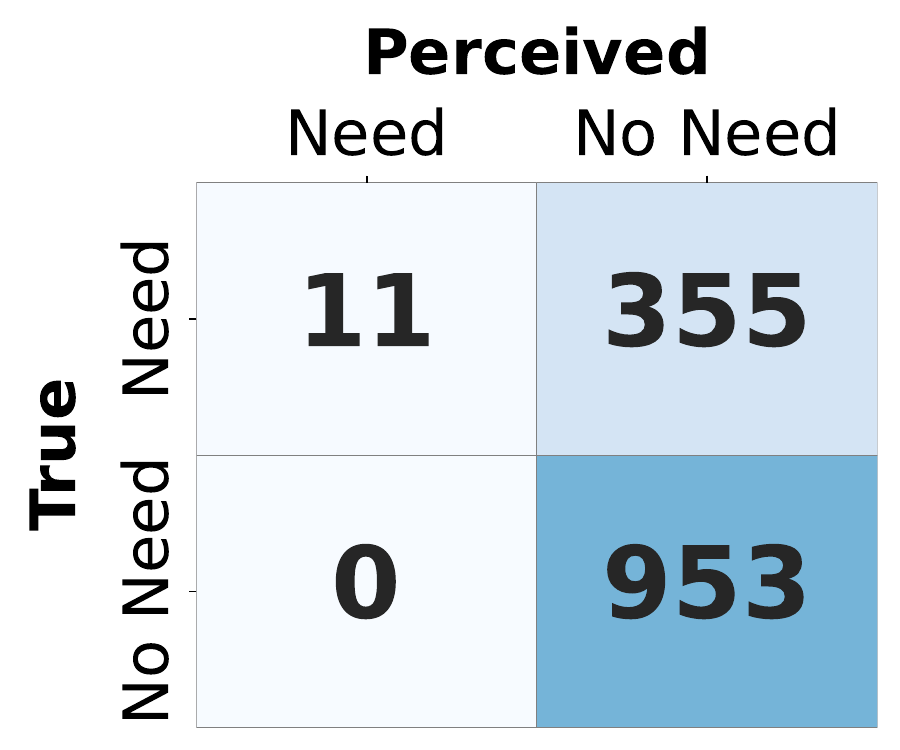}}

\calcsevenpanels{\textbf{GSM-Hard: true utility vs. perceived utility (calculator calling).}\label{fig:gsmhard_true_perceived}}
{\includegraphics[width=\linewidth]{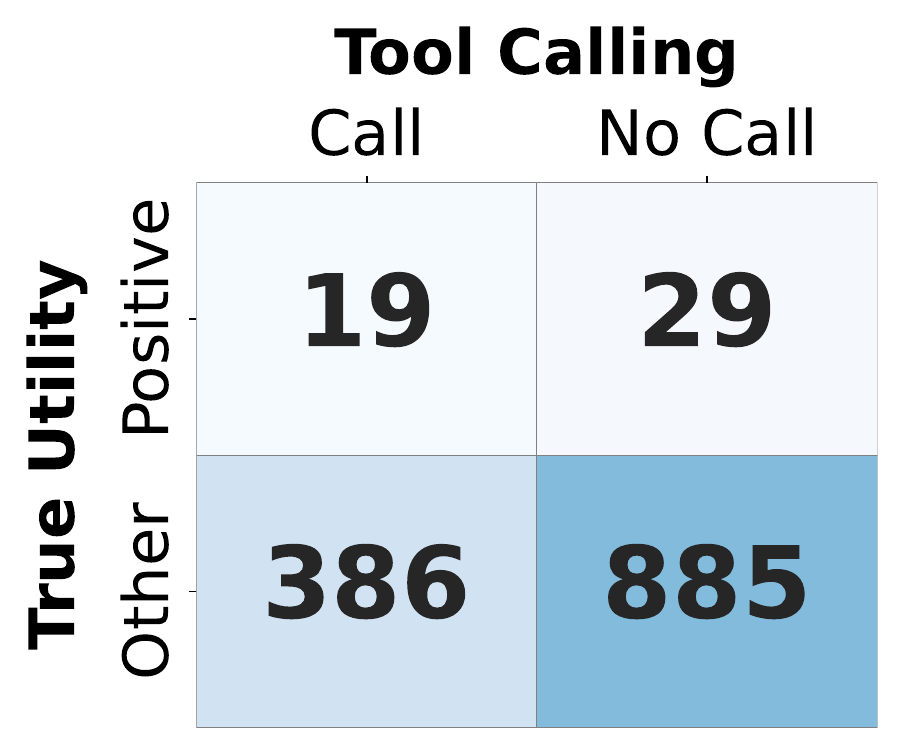}}{\includegraphics[width=\linewidth]{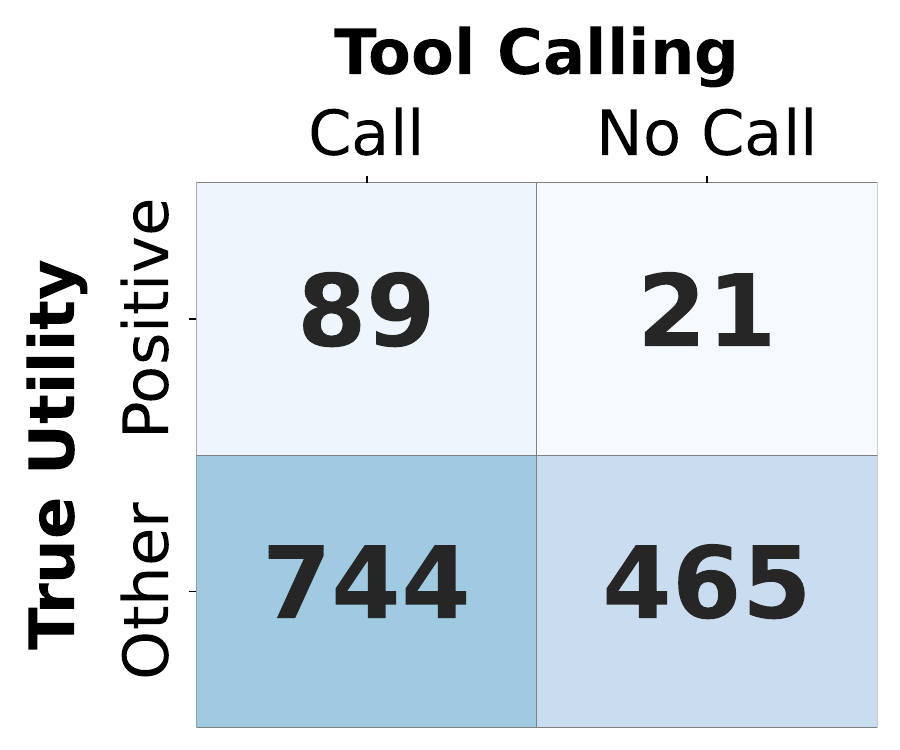}}{\includegraphics[width=\linewidth]{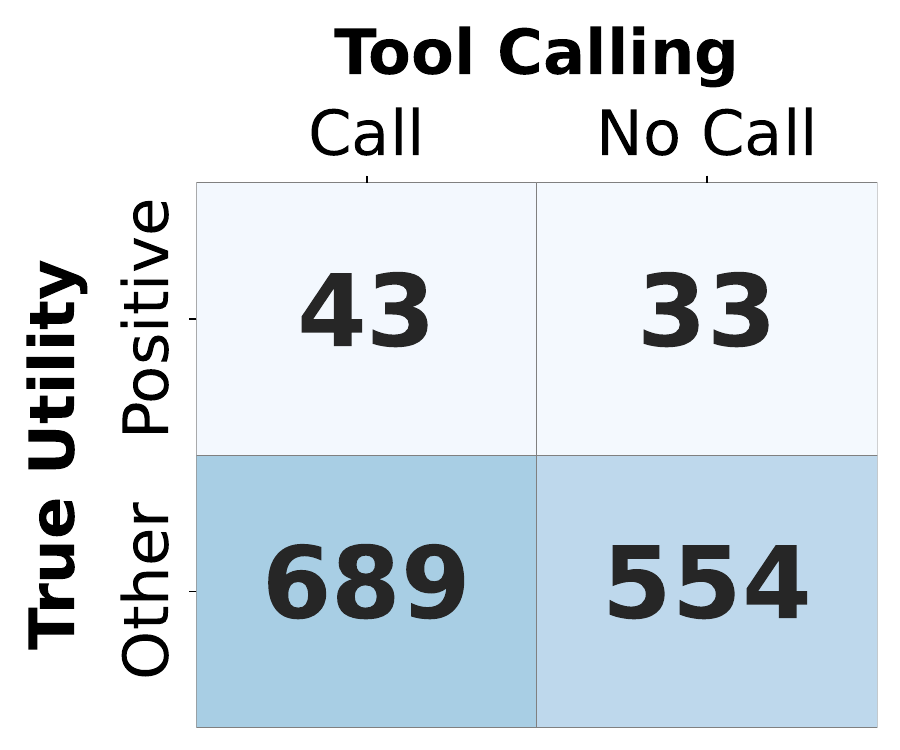}}{\includegraphics[width=\linewidth]{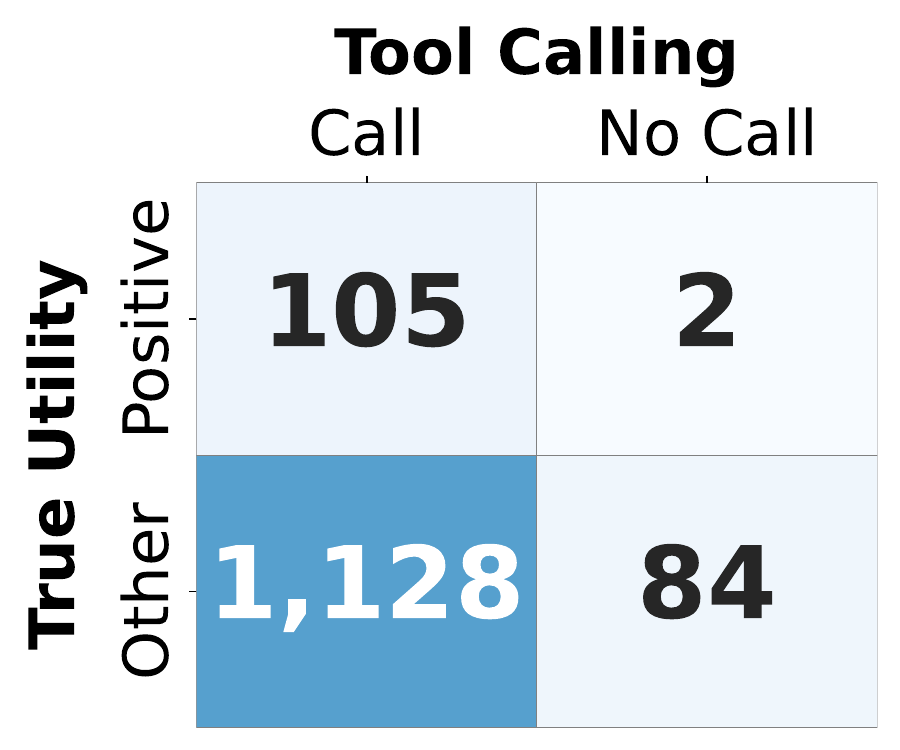}}{\includegraphics[width=\linewidth]{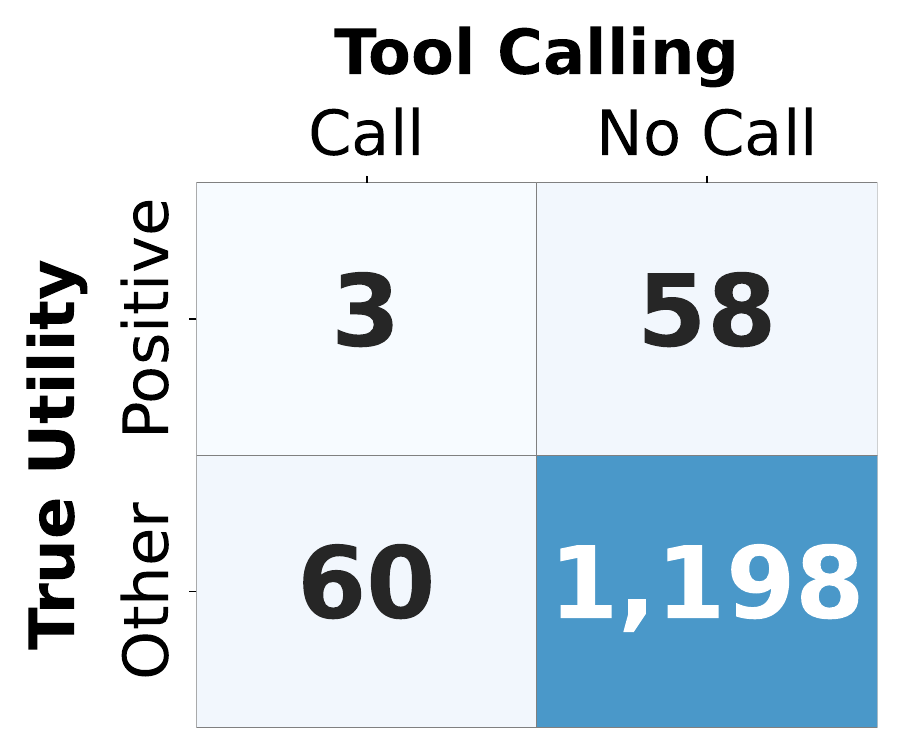}}{\includegraphics[width=\linewidth]{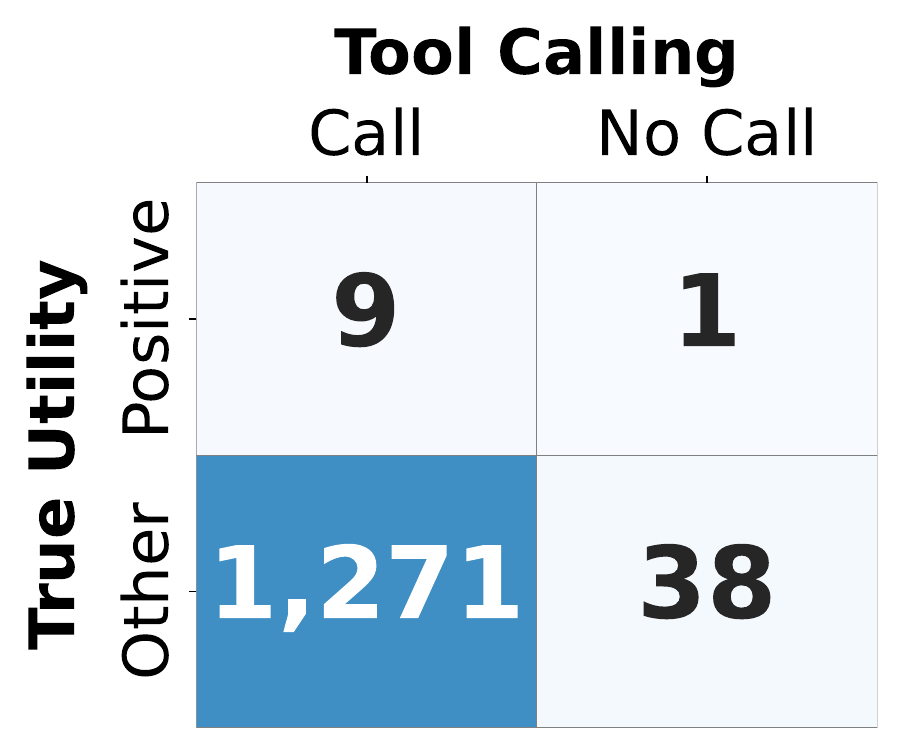}}{\includegraphics[width=\linewidth]{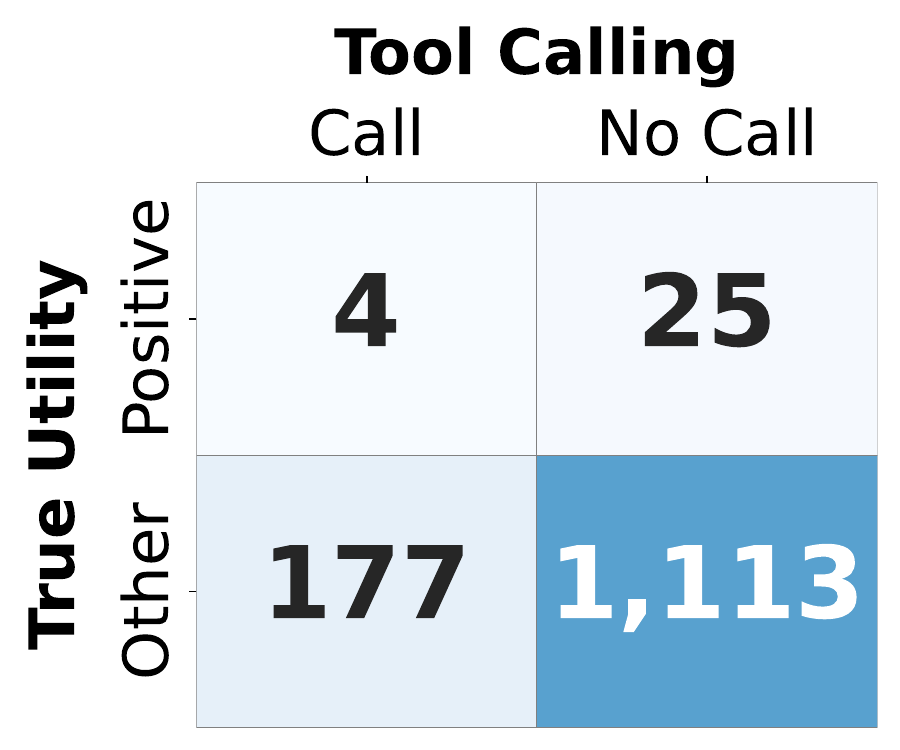}}

\subsection{Synthetic Multiplication}
\paragraph{Normative observation.}
For Gemma3-IT, GPT-OSS, Qwen3-A3B, and Qwen3-IT, calculator use corrects nearly every previously incorrect answer and almost never harms a correct one. This is a cleaner true-need/positive-utility alignment than on web search or GSM-Hard. Mistral-IT remains imperfect, and Llama3.2-IT is again anomalous: it realizes no positive-utility cases and changes 166 correct answers to incorrect ones.

\calcnormativepanels{\textbf{Synthetic Multiplication: No-Tool vs. Always-Tool correctness.}\label{fig:synthetic_mult_actual_need_utility}}
{\includegraphics[width=\linewidth]{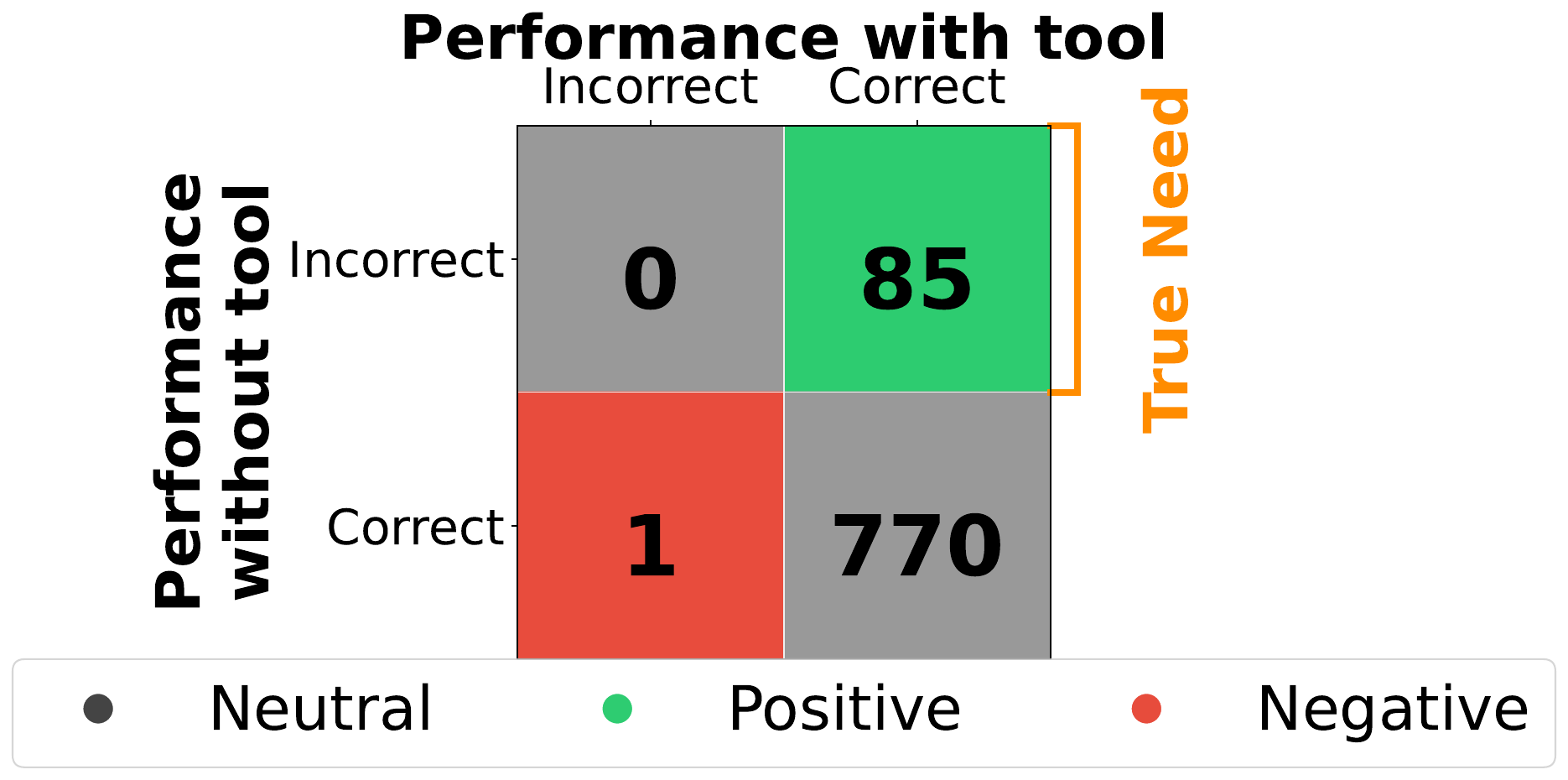}}{\includegraphics[width=\linewidth]{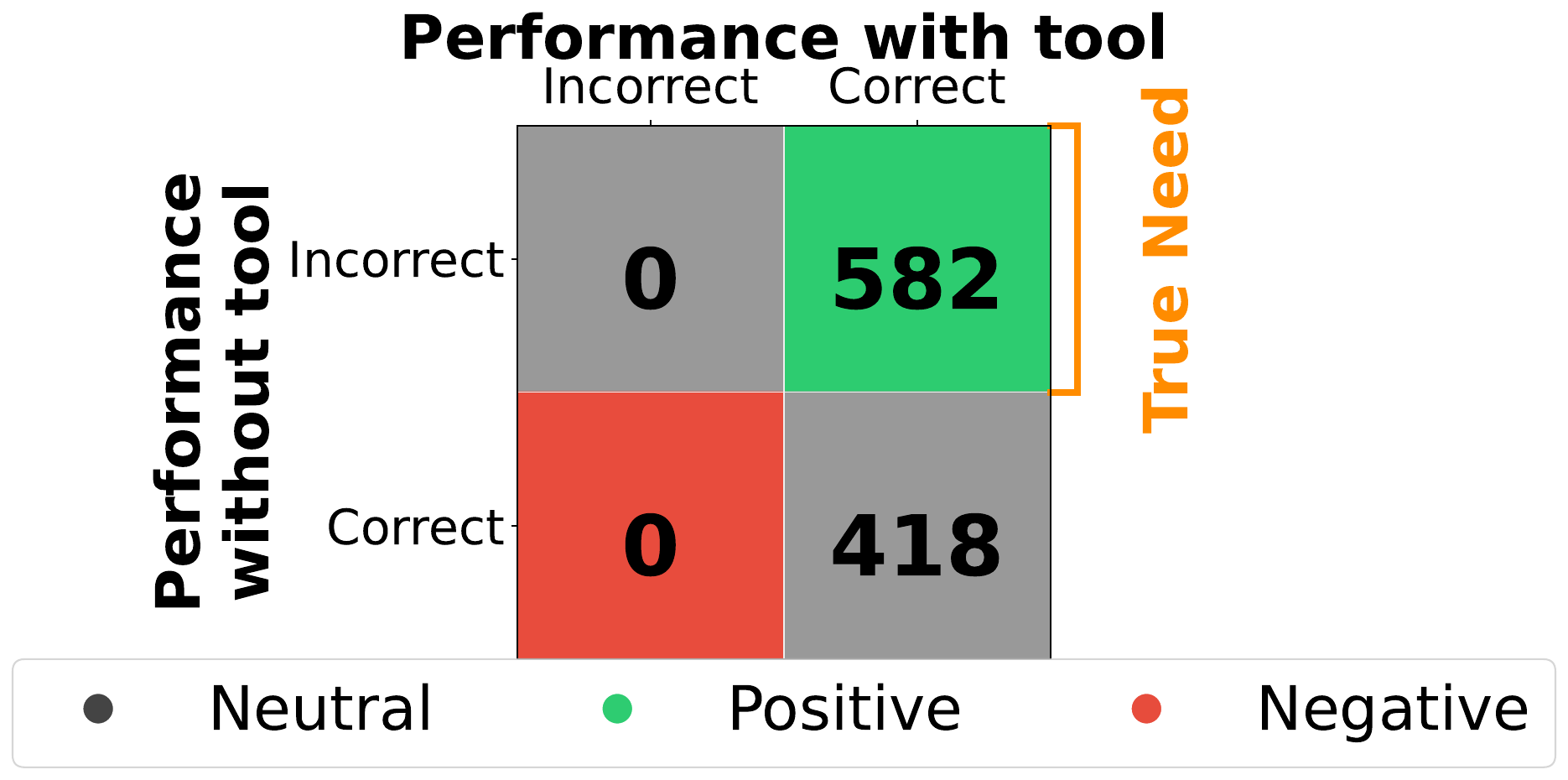}}{\includegraphics[width=\linewidth]{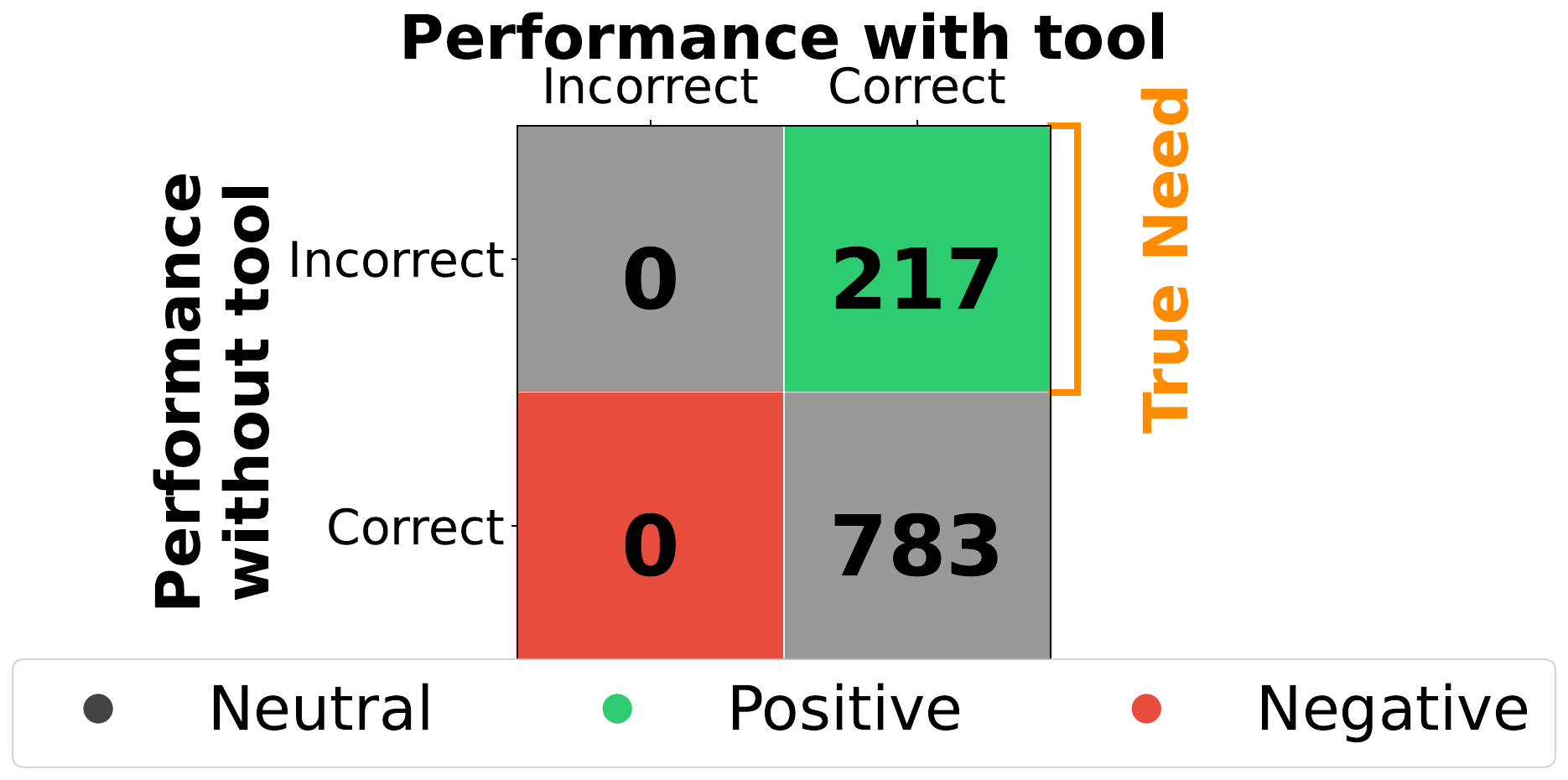}}{\includegraphics[width=\linewidth]{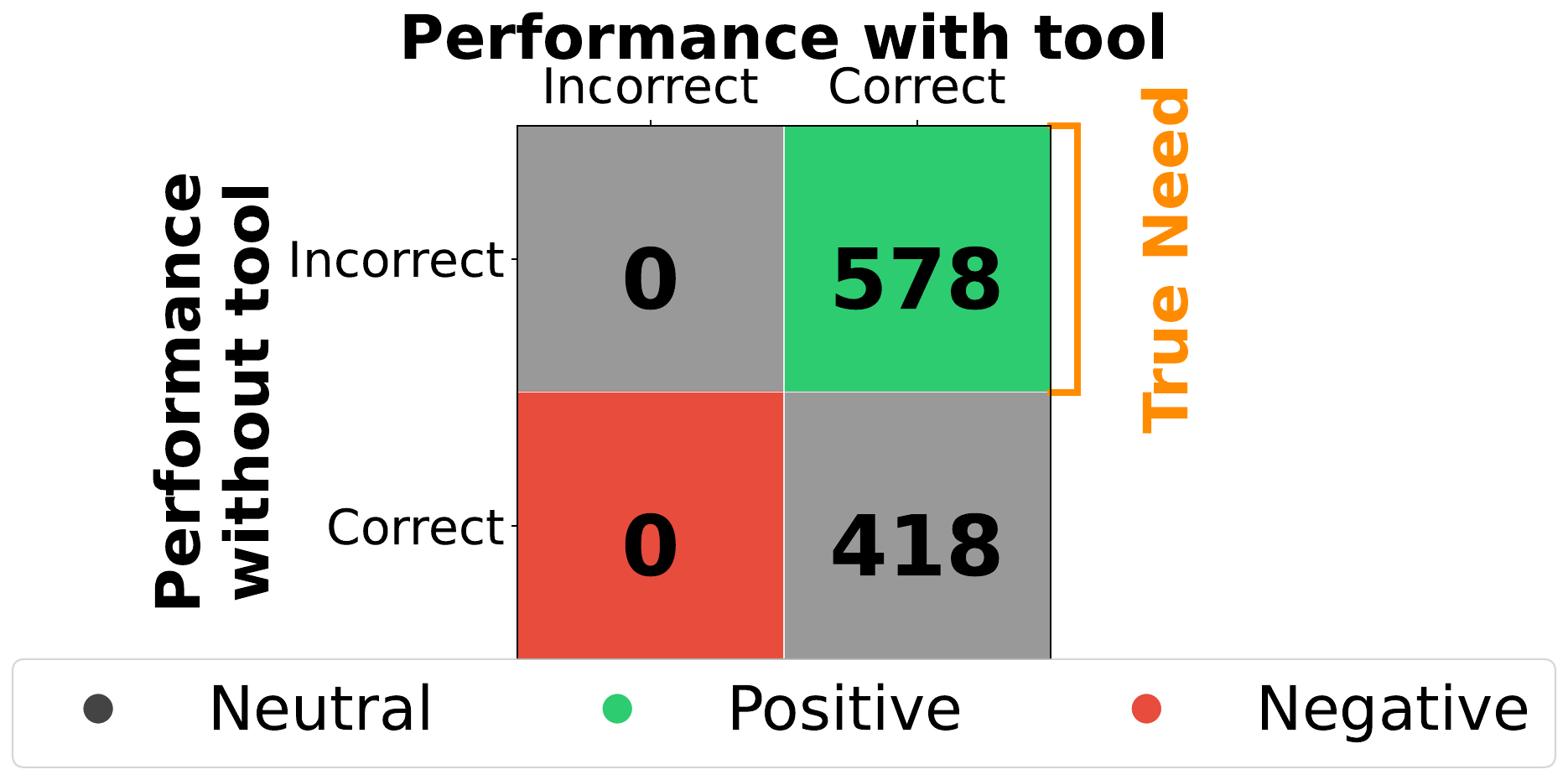}}{\includegraphics[width=\linewidth]{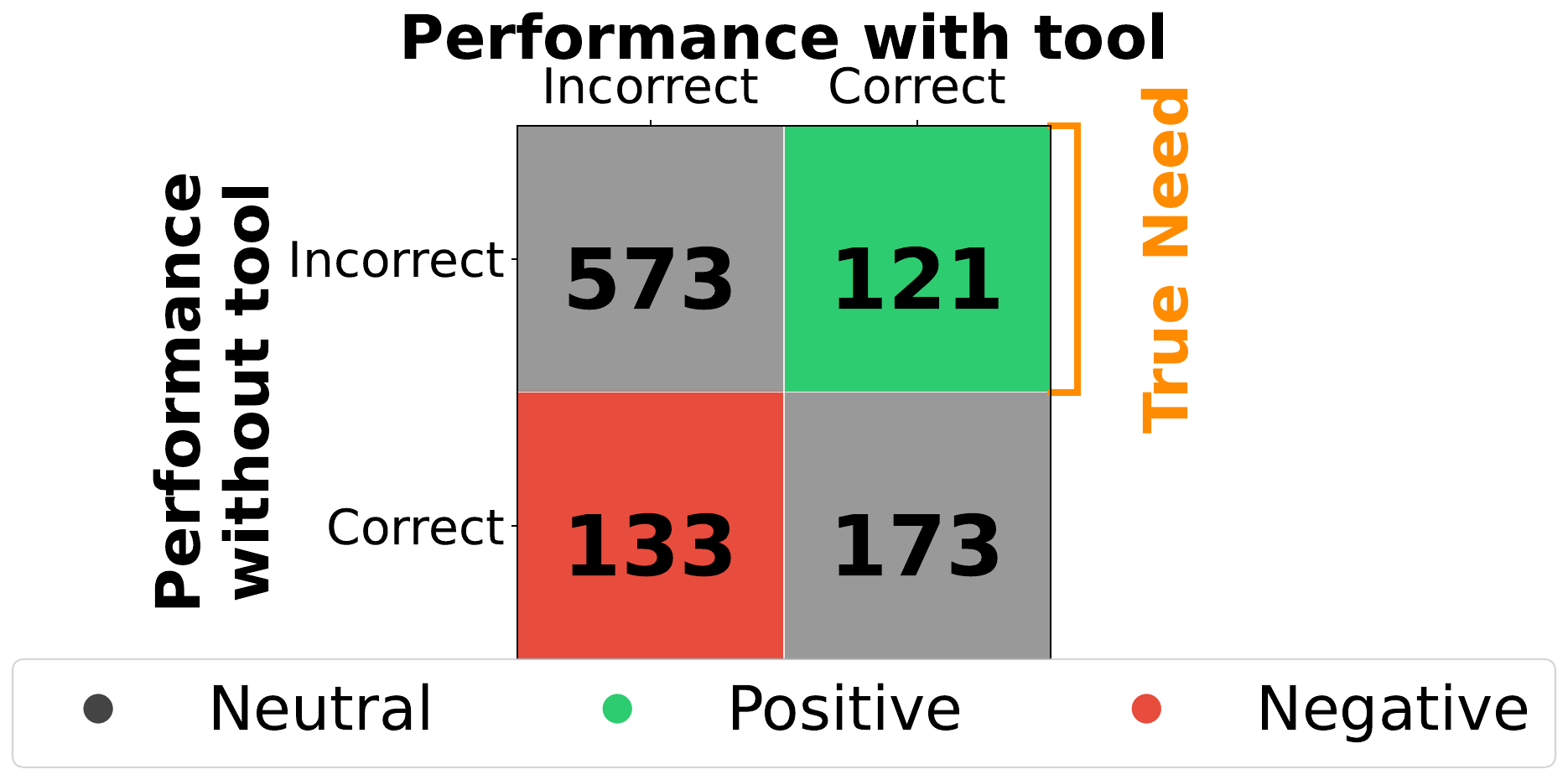}}{\includegraphics[width=\linewidth]{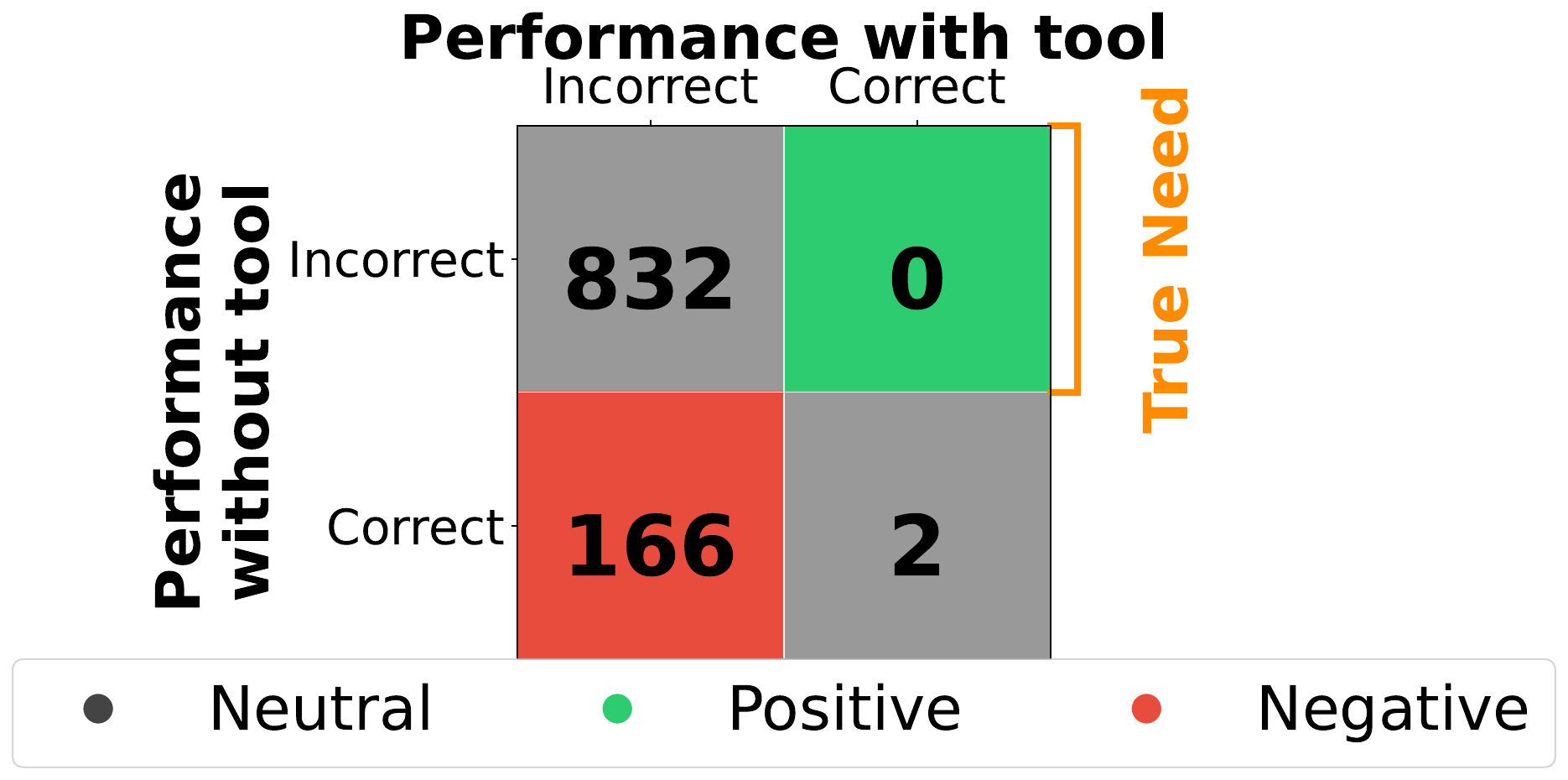}}{\includegraphics[width=\linewidth]{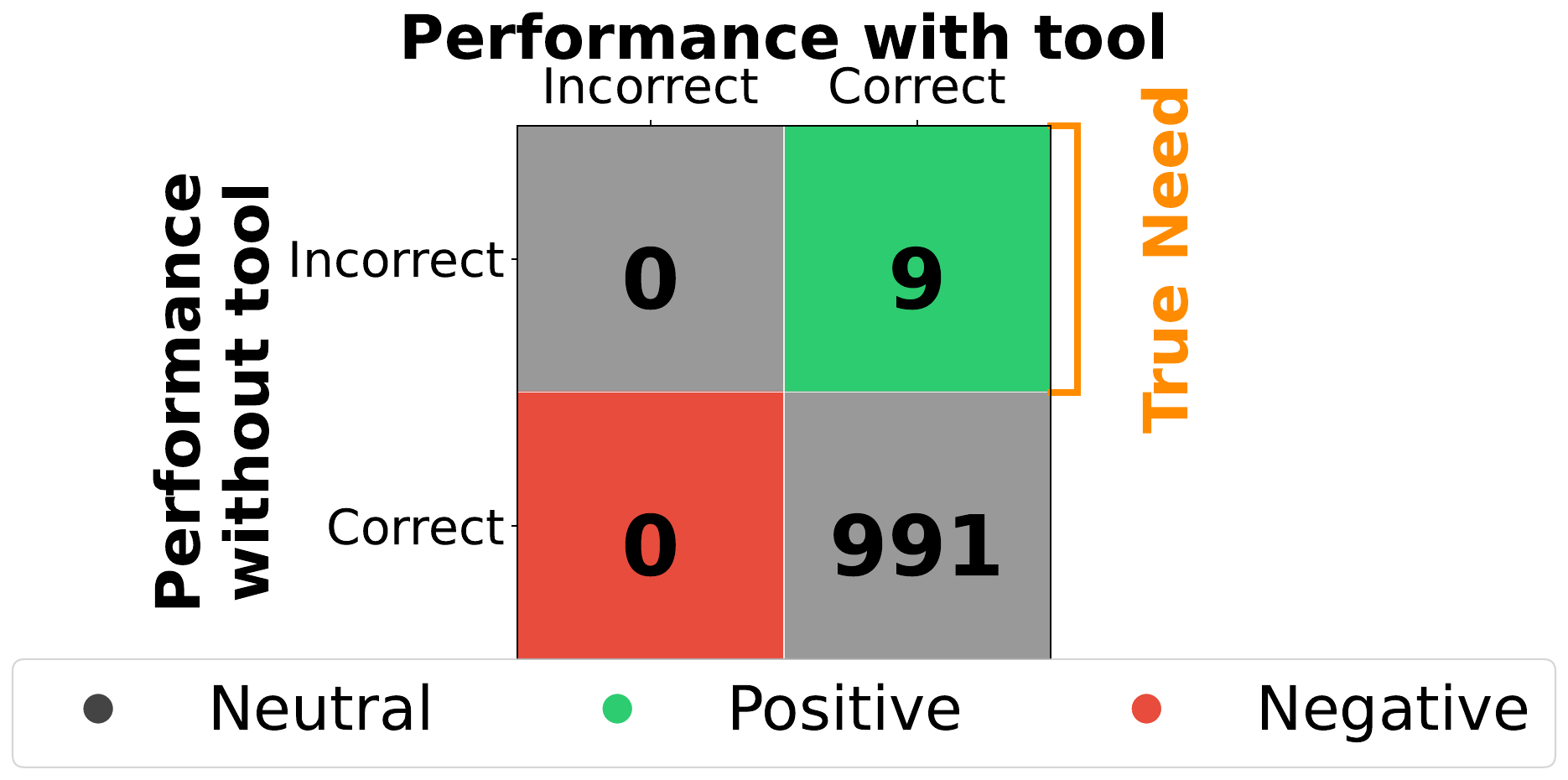}}

\paragraph{Descriptive observation.}
Most models over-call the deterministic calculator. Gemma3-IT and Qwen3-IT call on every evaluated instance; Qwen3-A3B and Llama3.2-IT call almost universally. GPT-OSS is better at capturing positive utility but still makes hundreds of non-beneficial calls, whereas Mistral-IT misses most beneficial calls. GPT-5.5's available perceived-need panel similarly shows that many calls occur despite a reported lack of need.

\begin{figure*}[t]
    \centering
    \includegraphics[width=0.78\textwidth]{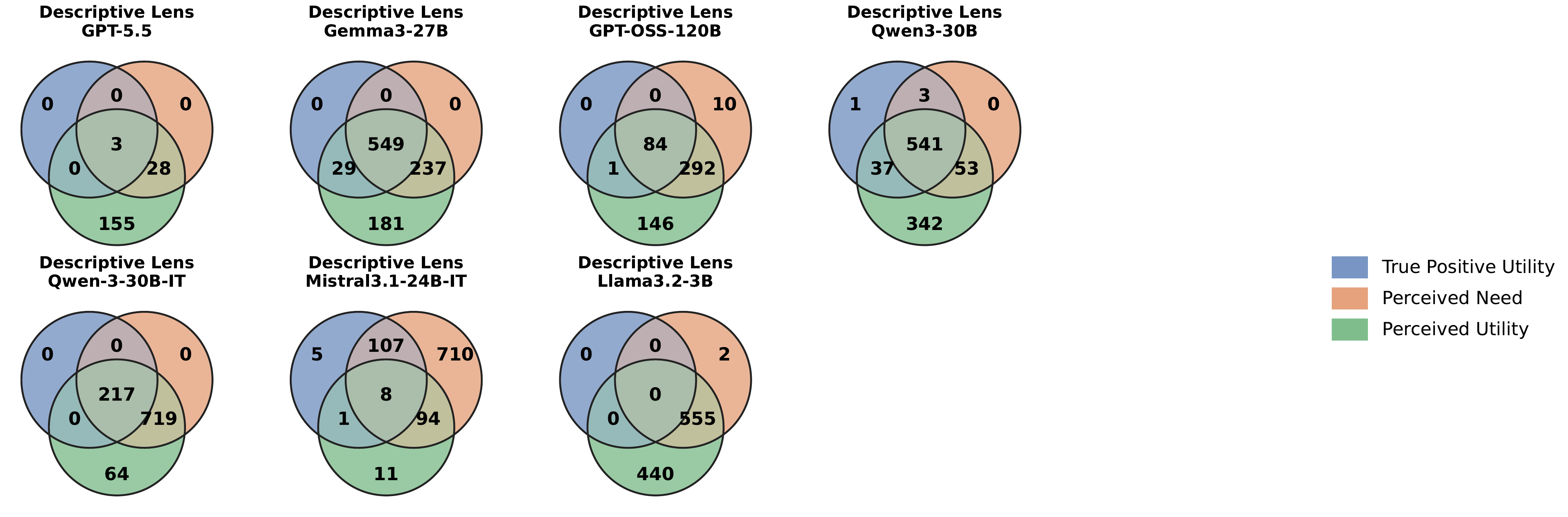}
    \caption{[Synthetic Multiplication Task] Venn diagrams of \textbf{True Positive Utility, Perceived Need, and Perceived Utility}. Each panel shows their empirical overlap for one model. Calls outside true positive utility are non-beneficial; true-positive-utility cases outside perceived utility are missed opportunities. Perceived need is a separate self-assessment and need not be nested within either utility set.}
    \label{fig:venn-synthetic-mult}
\end{figure*}

\calcsevenpanels{\textbf{Synthetic Multiplication: perceived need vs. calculator calling.}\label{fig:synthetic_mult_perceived_need_utility}}
{\includegraphics[width=\linewidth]{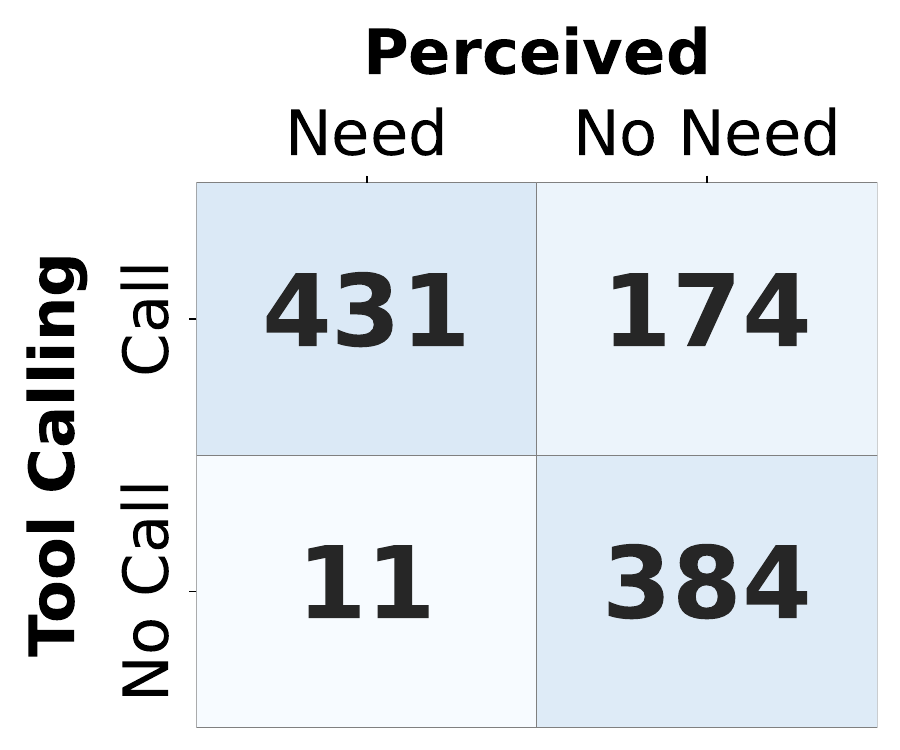}}{\includegraphics[width=\linewidth]{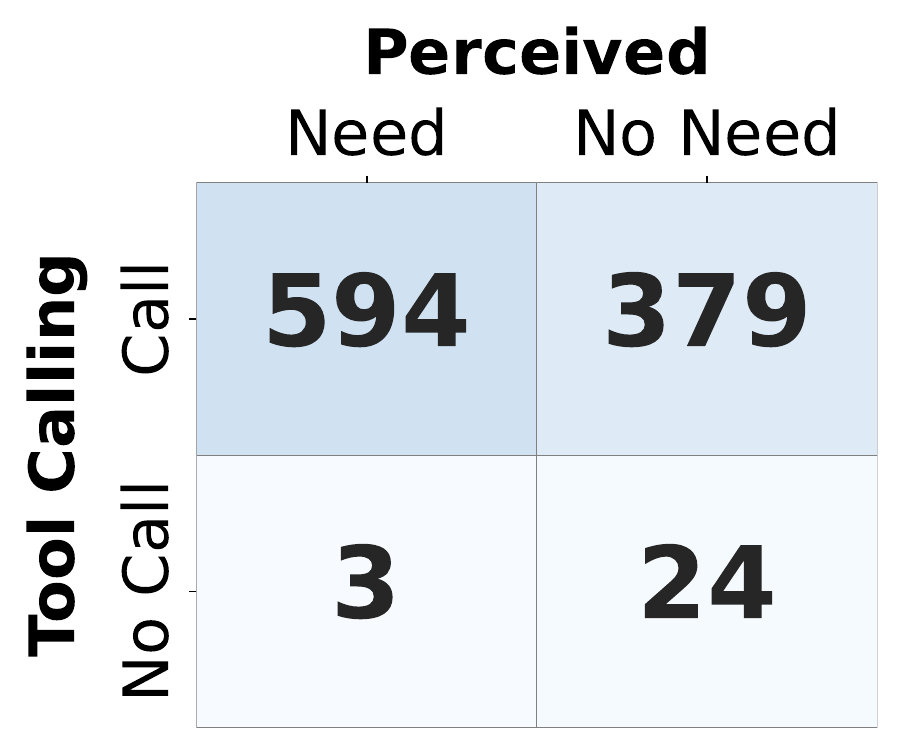}}{\includegraphics[width=\linewidth]{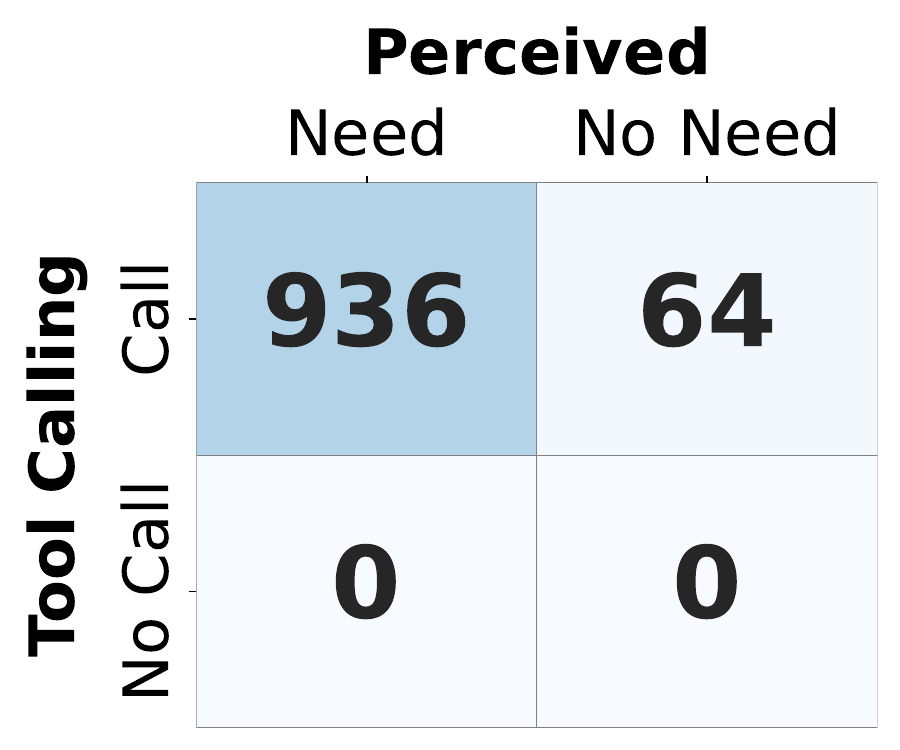}}{\includegraphics[width=\linewidth]{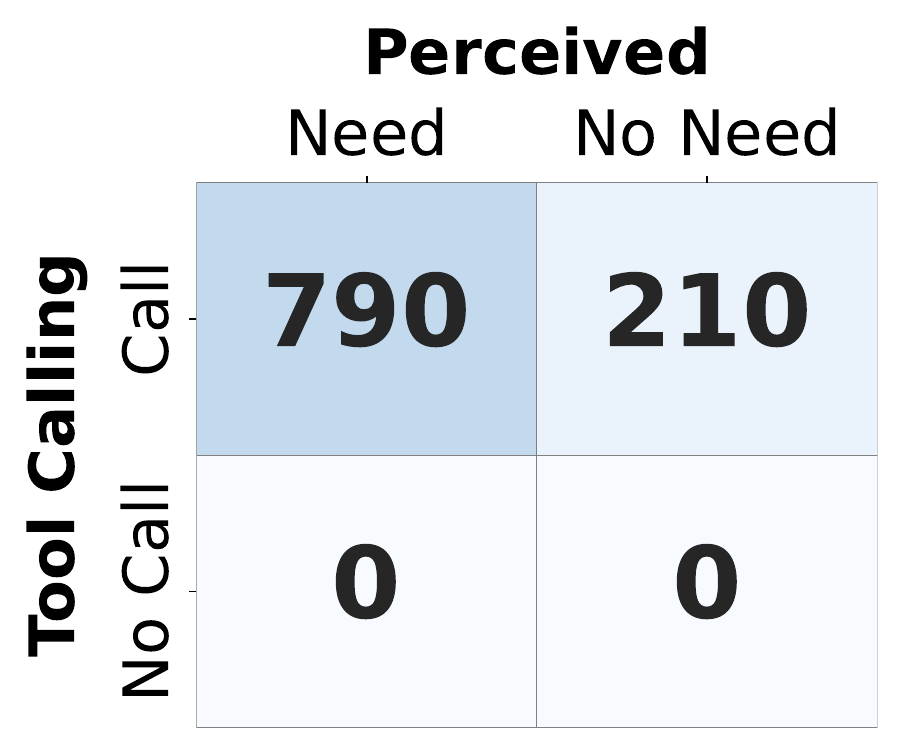}}{\includegraphics[width=\linewidth]{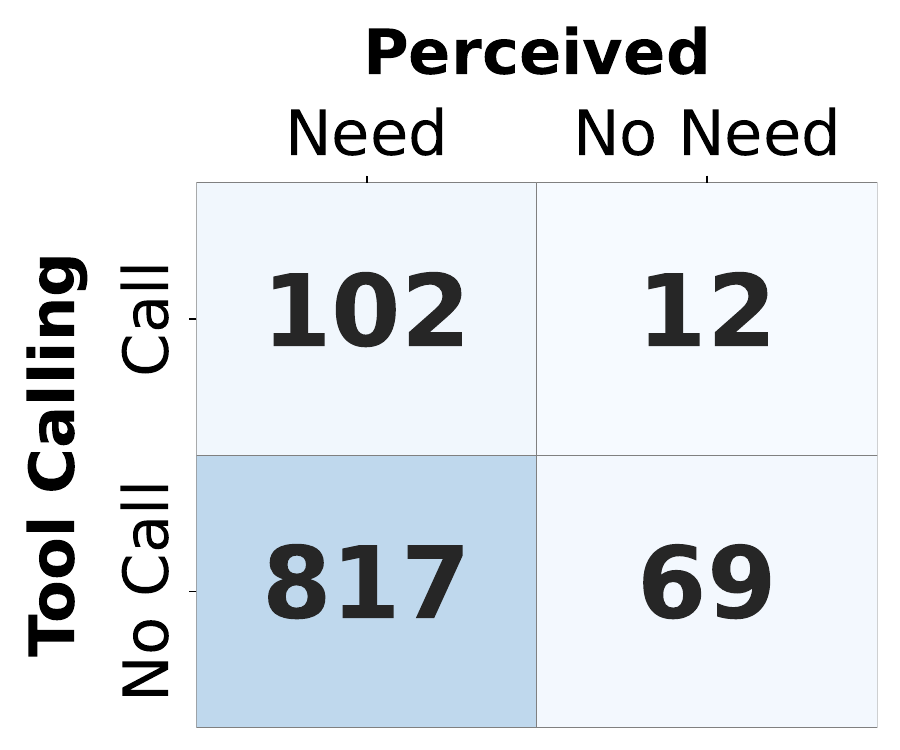}}{\includegraphics[width=\linewidth]{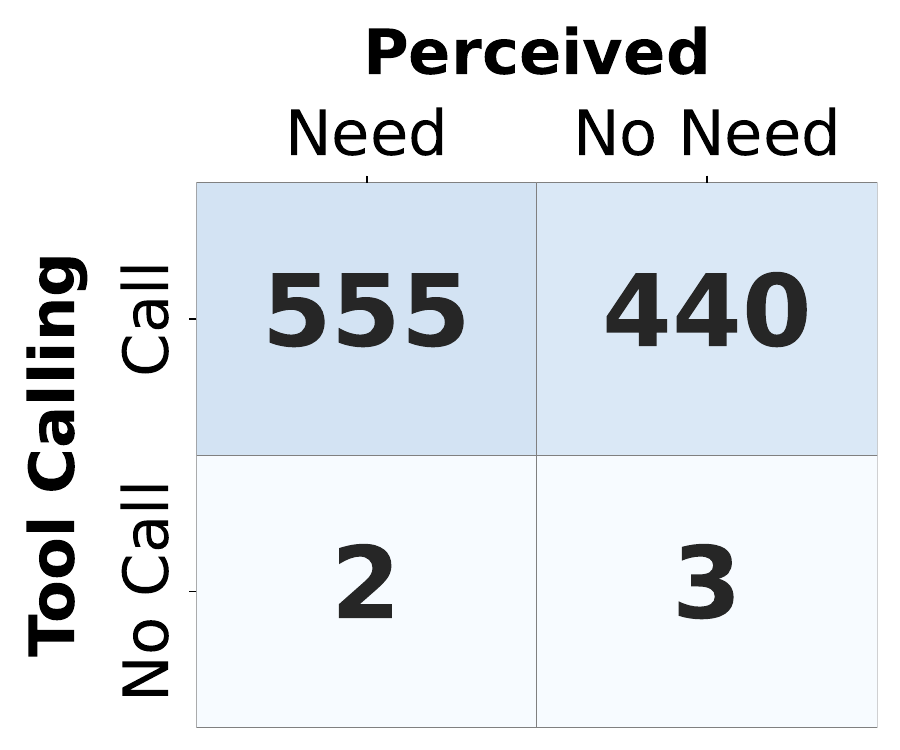}}
{\includegraphics[width=\linewidth]{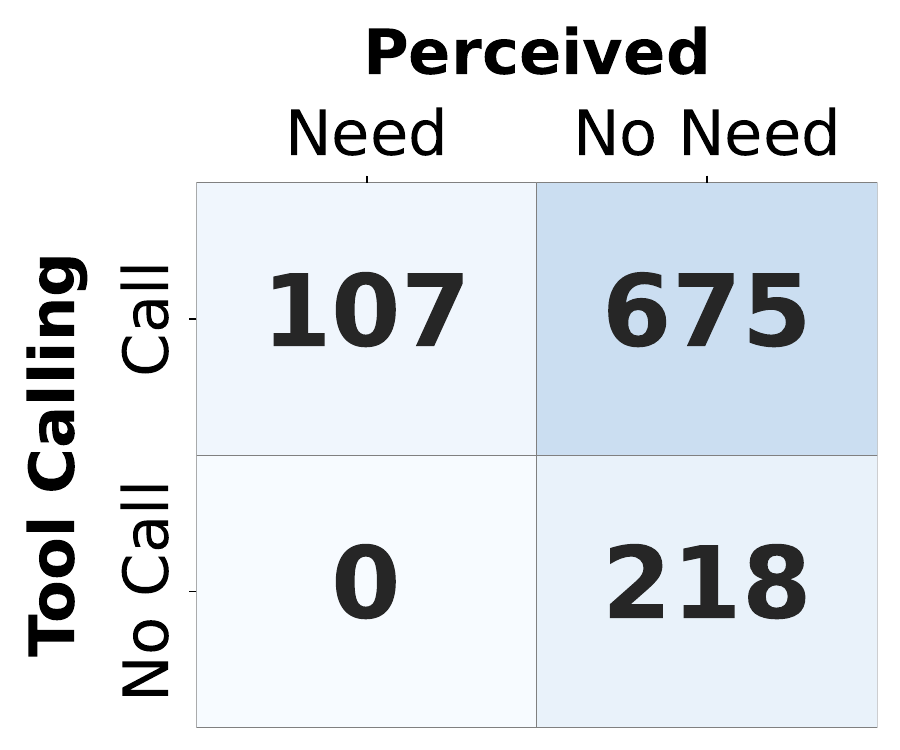}}

\calcsevenpanels{\textbf{Synthetic Multiplication: perceived need.}\label{fig:synthetic_mult_perceived_need}}
{\includegraphics[width=\linewidth]{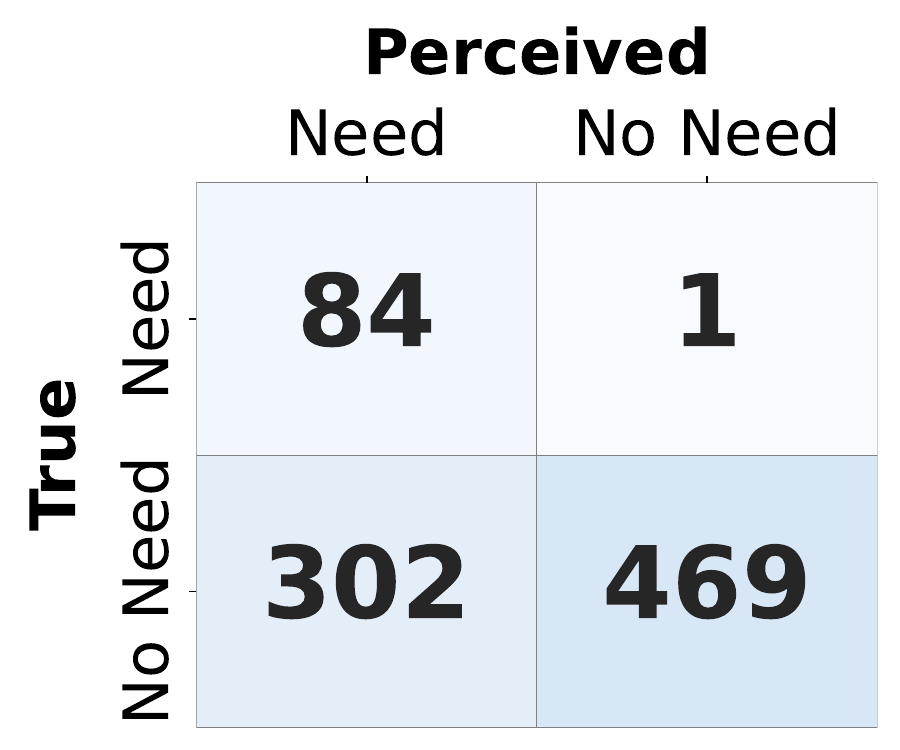}}{\includegraphics[width=\linewidth]{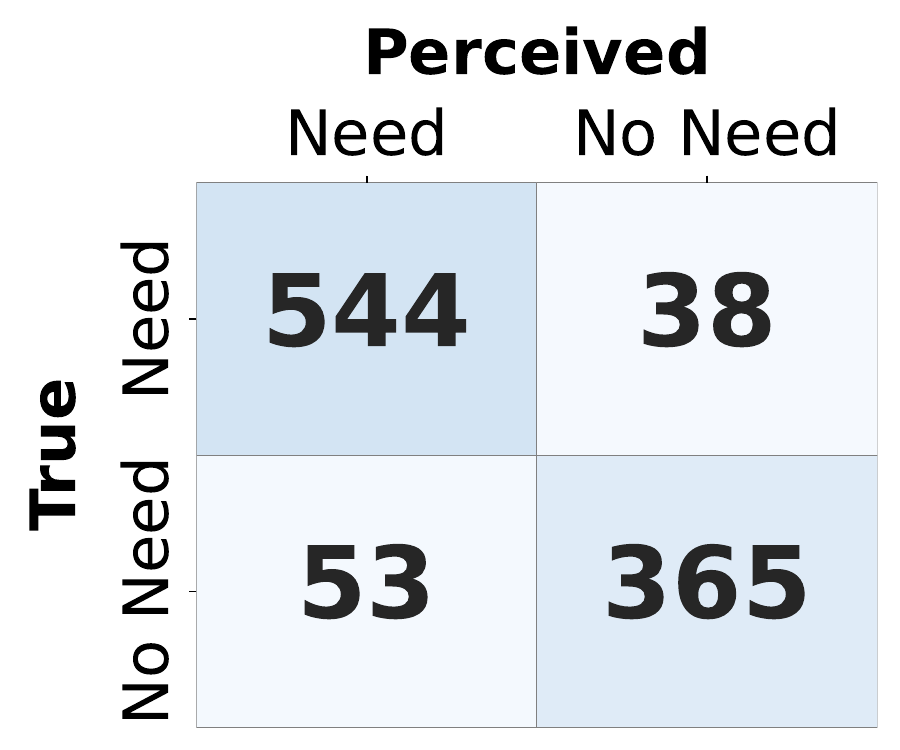}}{\includegraphics[width=\linewidth]{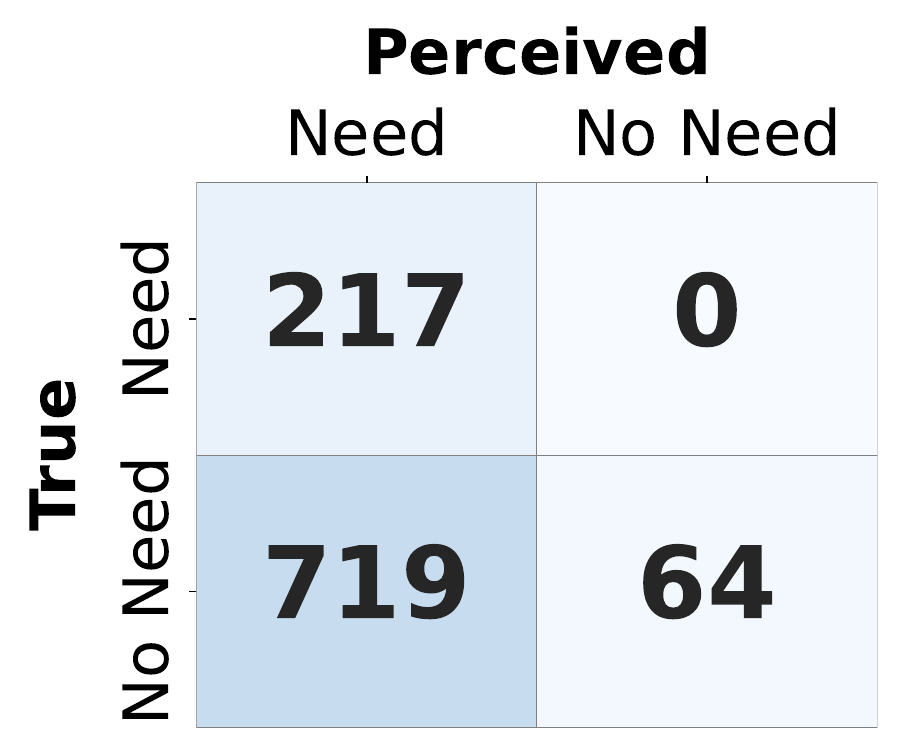}}{\includegraphics[width=\linewidth]{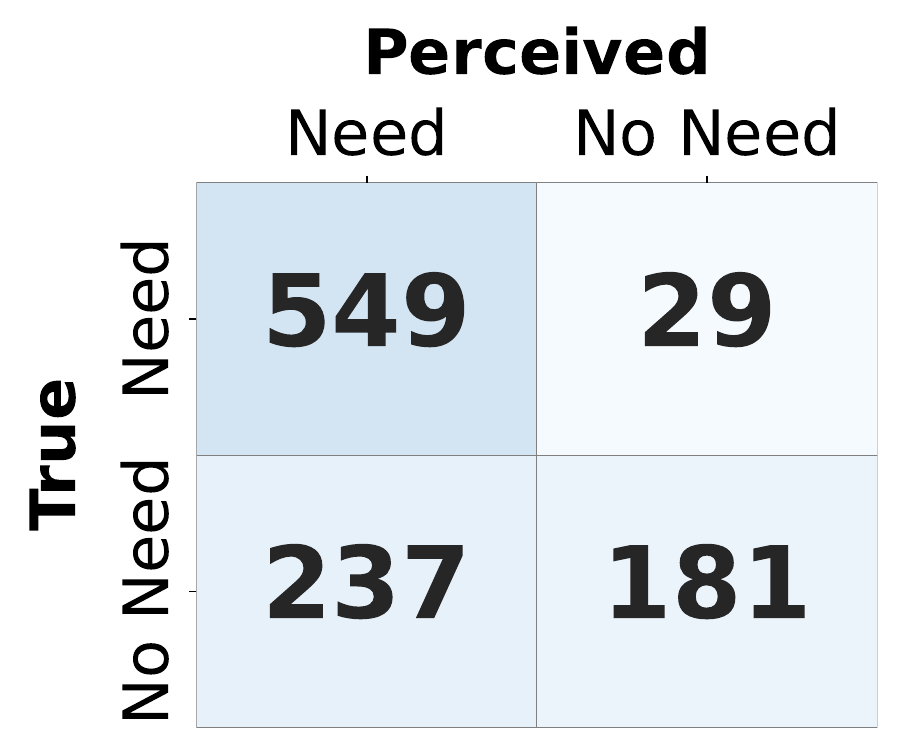}}{\includegraphics[width=\linewidth]{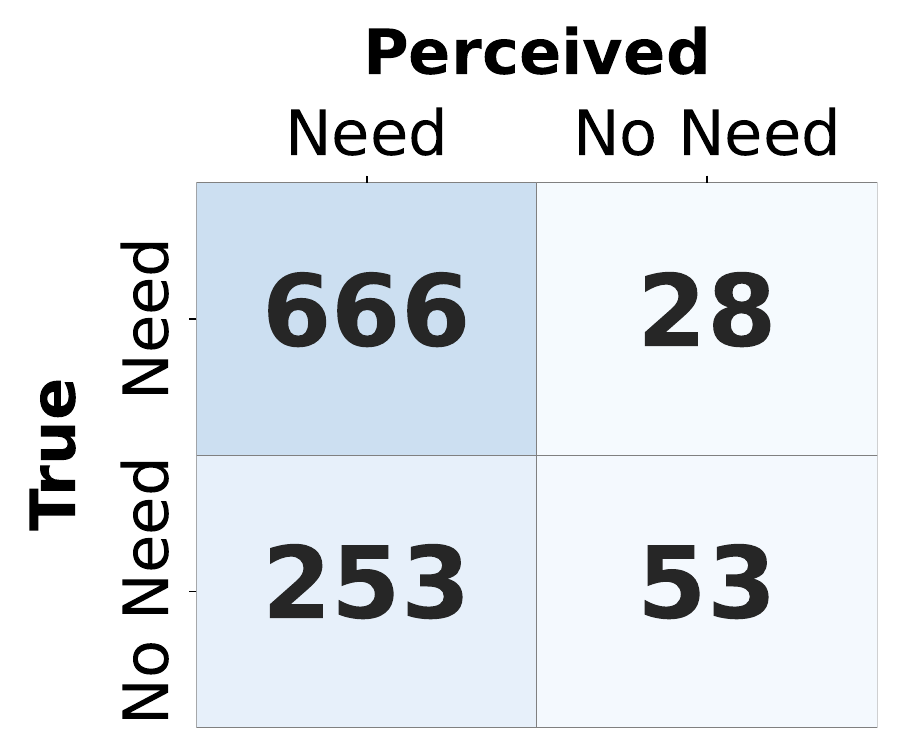}}{\includegraphics[width=\linewidth]{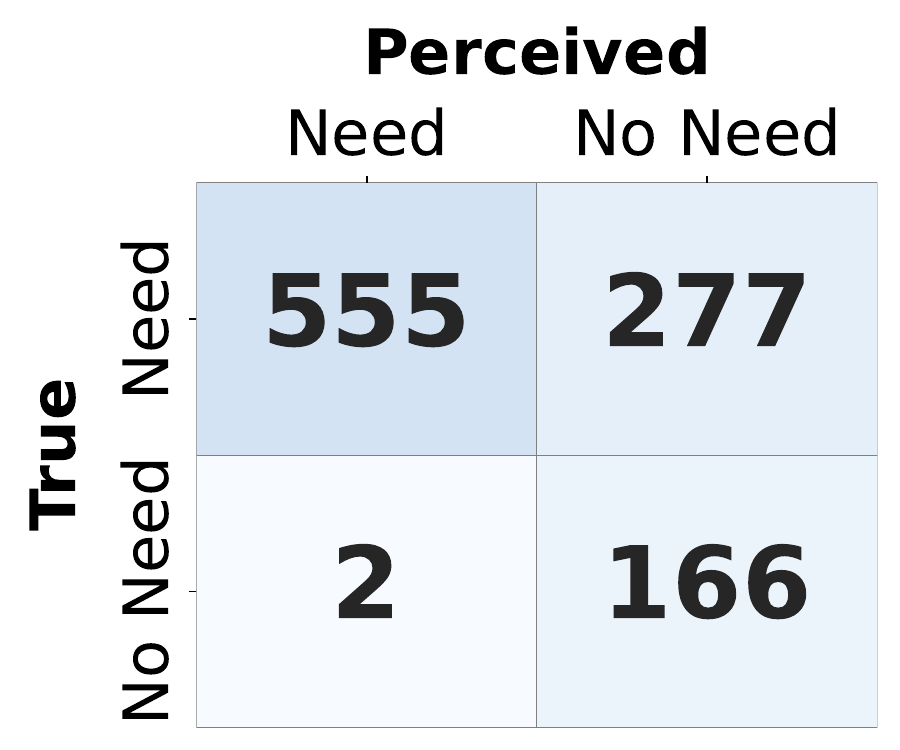}}
{\includegraphics[width=\linewidth]{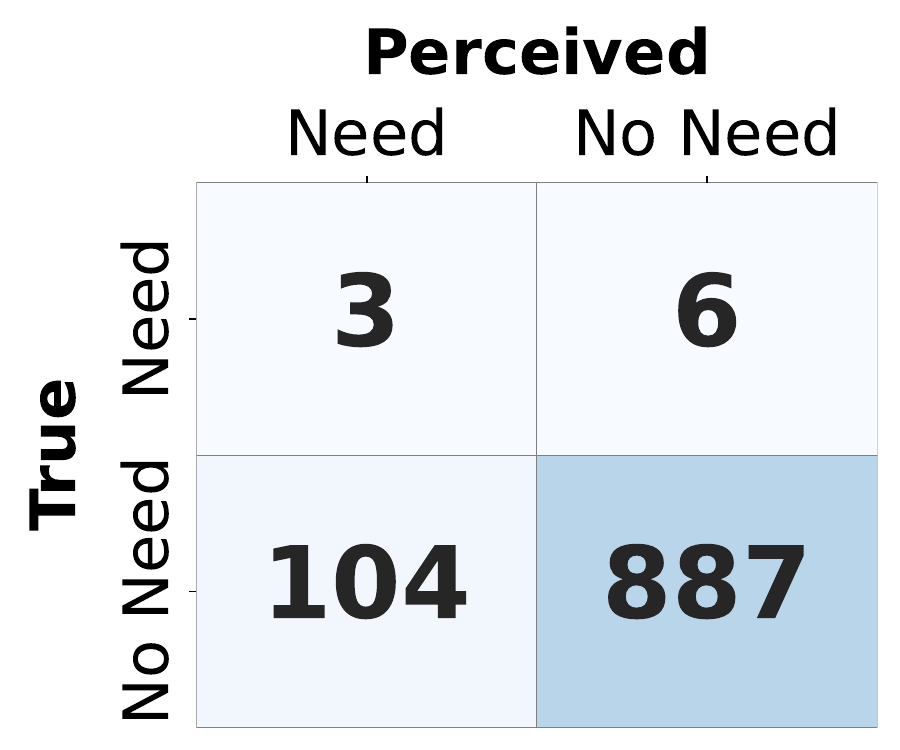}}

\calcsevenpanels{\textbf{Synthetic Multiplication: true utility vs. perceived utility (calculator calling).}\label{fig:synthetic_mult_true_perceived}}
{\includegraphics[width=\linewidth]{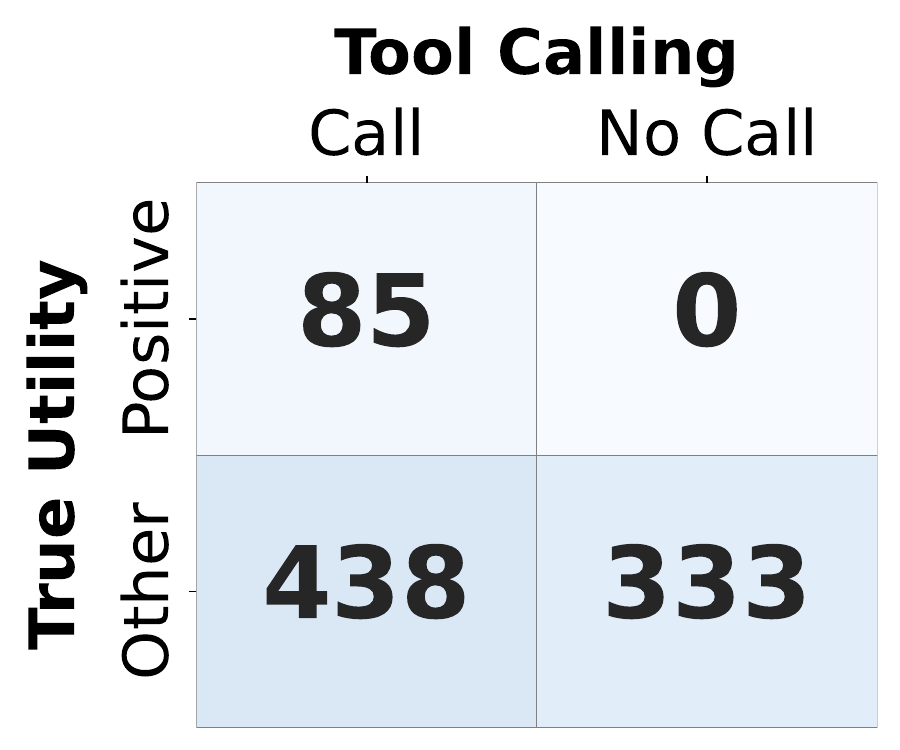}}{\includegraphics[width=\linewidth]{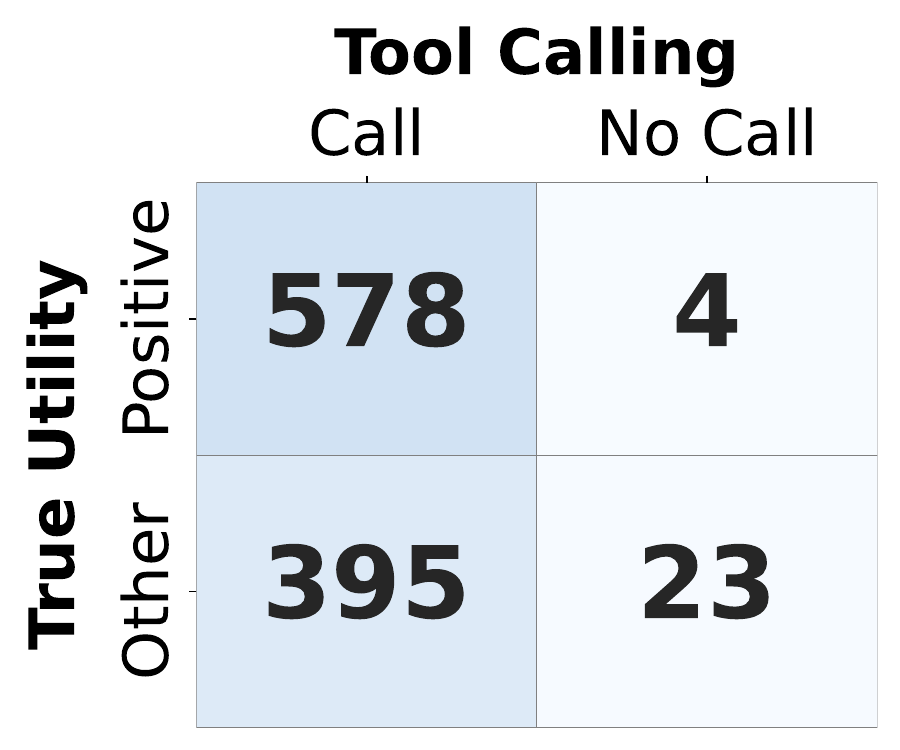}}{\includegraphics[width=\linewidth]{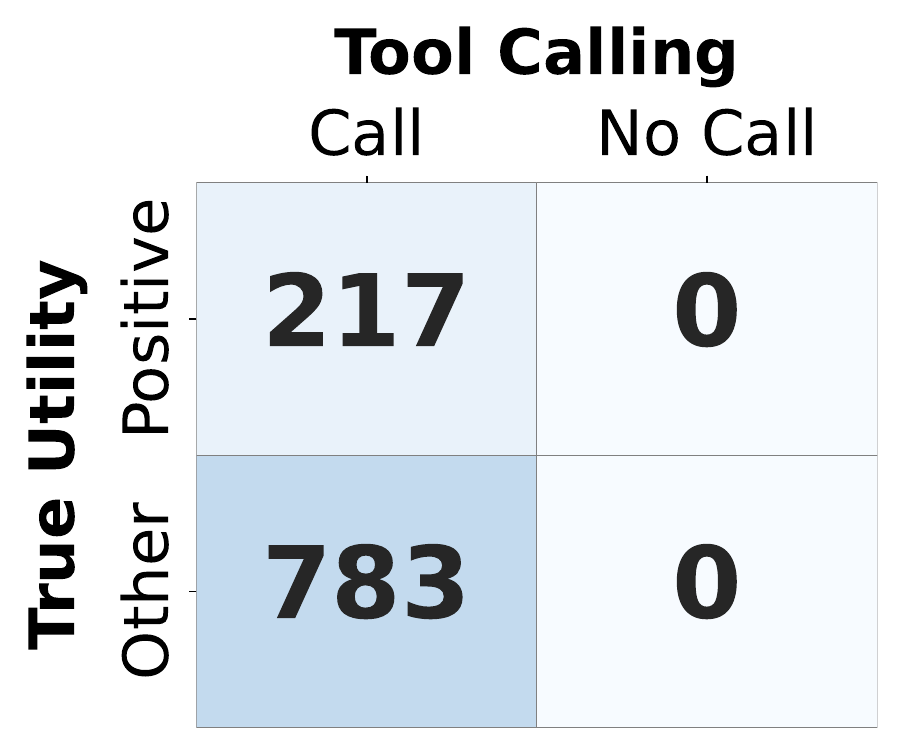}}{\includegraphics[width=\linewidth]{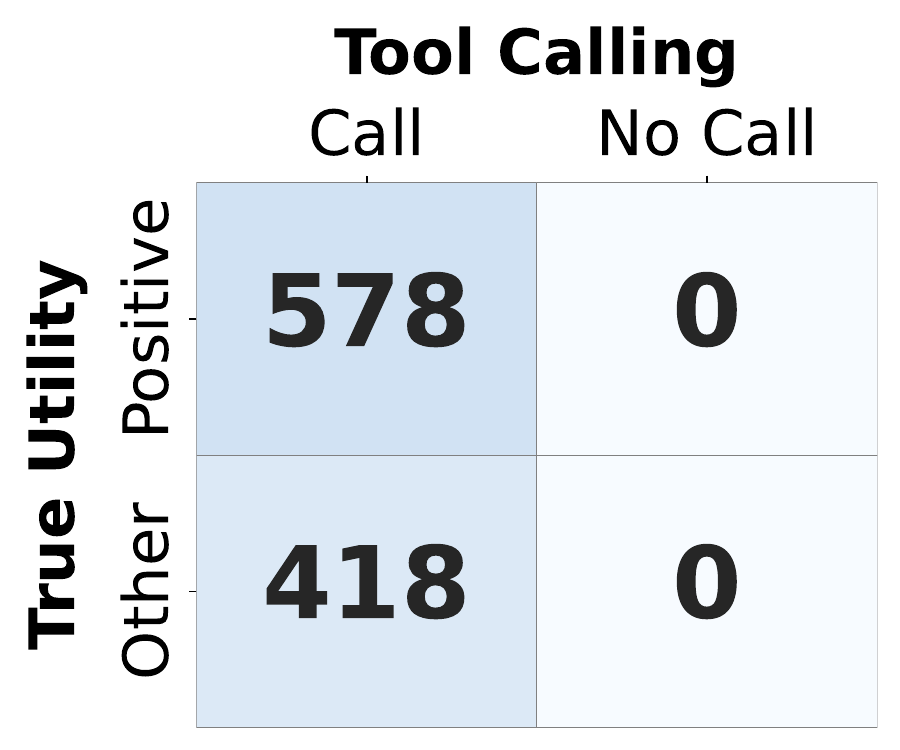}}{\includegraphics[width=\linewidth]{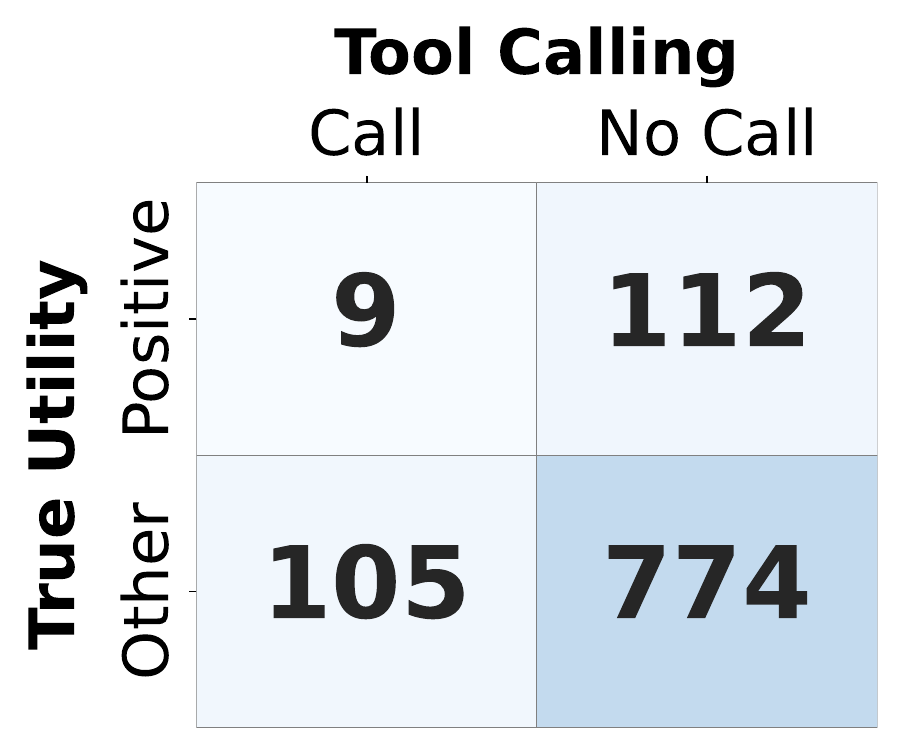}}{\includegraphics[width=\linewidth]{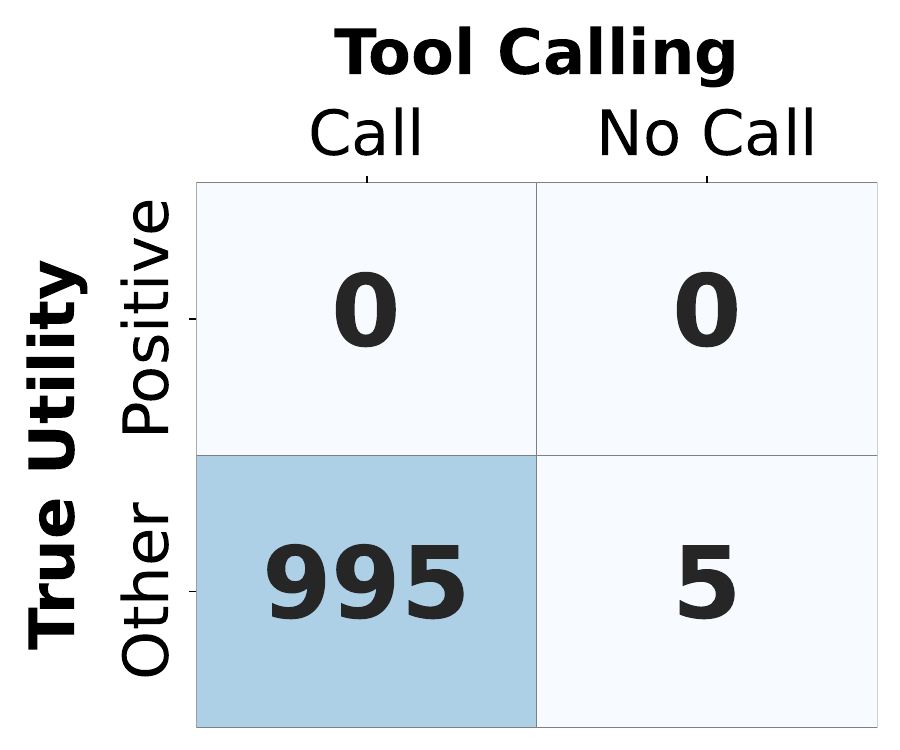}}{\includegraphics[width=\linewidth]{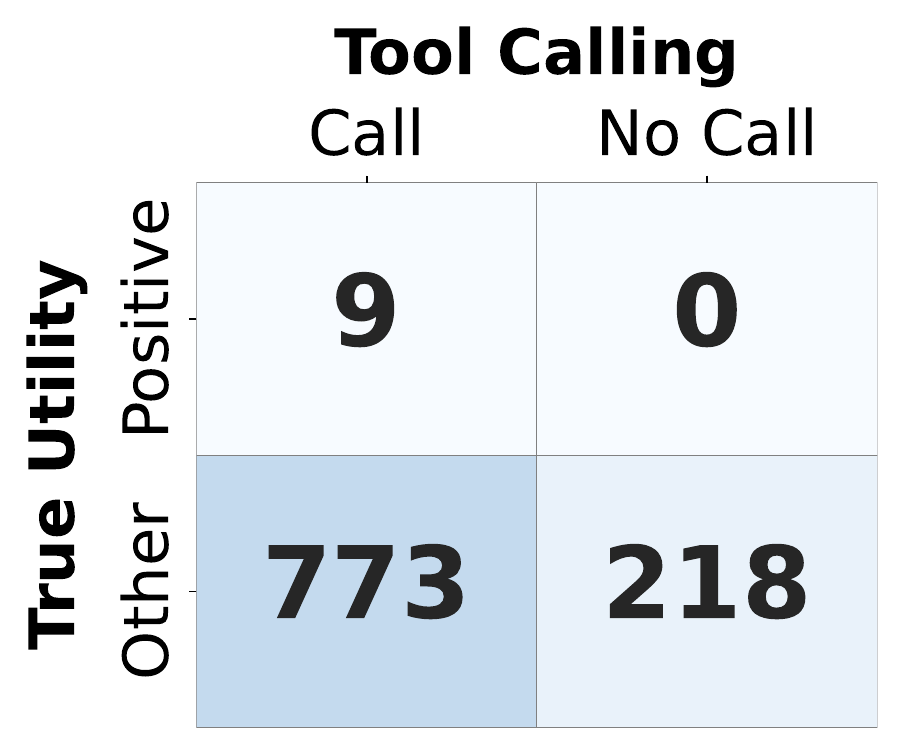}}

\subsection{Synthetic Large-Digit Multiplication}
\paragraph{Normative observation.}
This task creates near-universal true need for most open models. With the calculator, Gemma3-IT, GPT-OSS, Qwen3-A3B, and Qwen3-IT solve almost all examples, and Mistral-IT improves on 656 instances. Negative utility is nearly absent because unaided correct answers are rare. Llama3.2-IT is again different: all 1,000 no-tool answers are incorrect and the calculator condition corrects none, isolating a tool-use or output-integration failure rather than insufficient calculator capability.

\calcnormativepanels{\textbf{Synthetic Large-Digit Multiplication: No-Tool vs. Always-Tool correctness.}\label{fig:synthetic_nn_actual_need_utility}}
{\includegraphics[width=\linewidth]{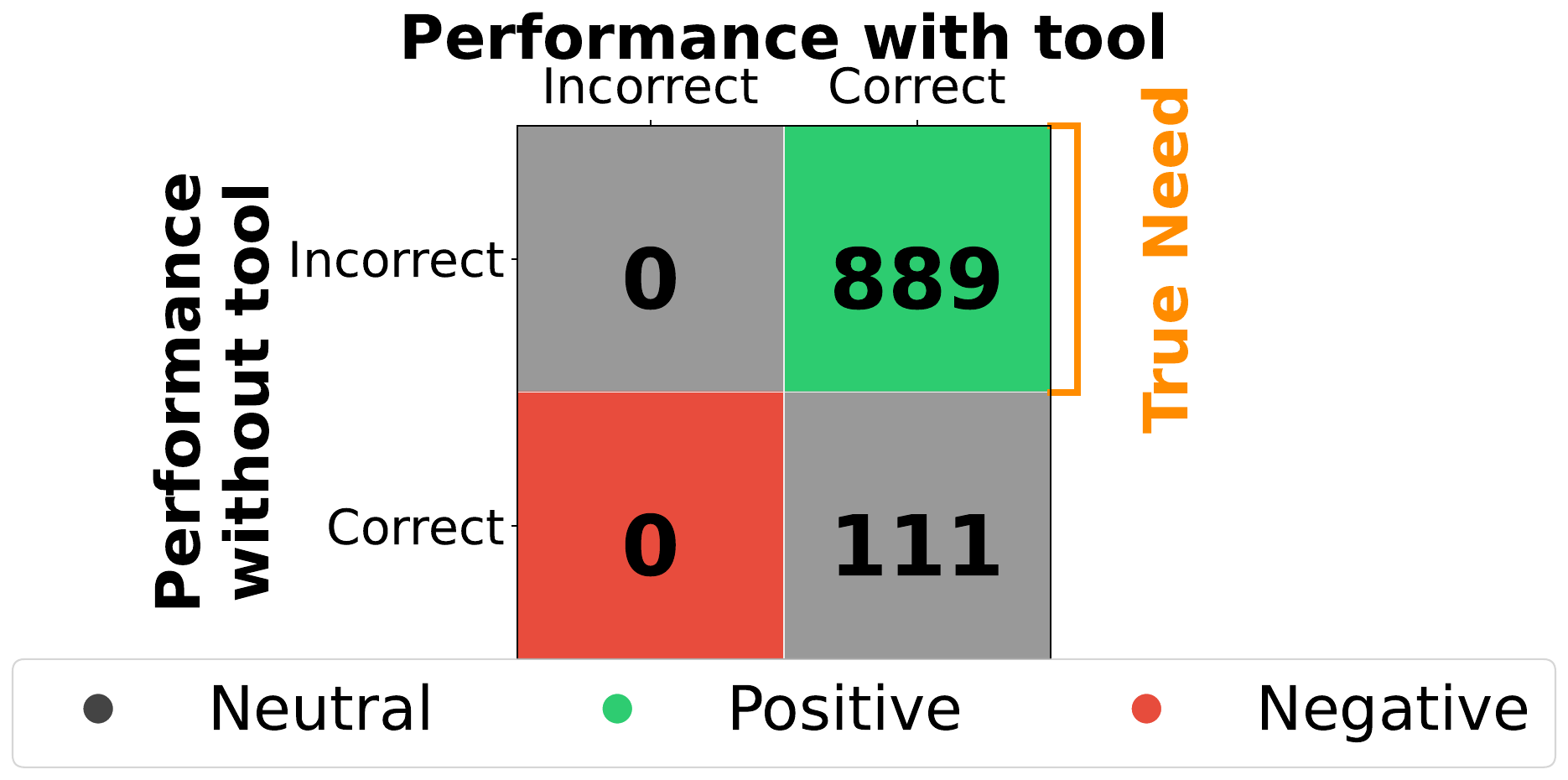}}{\includegraphics[width=\linewidth]{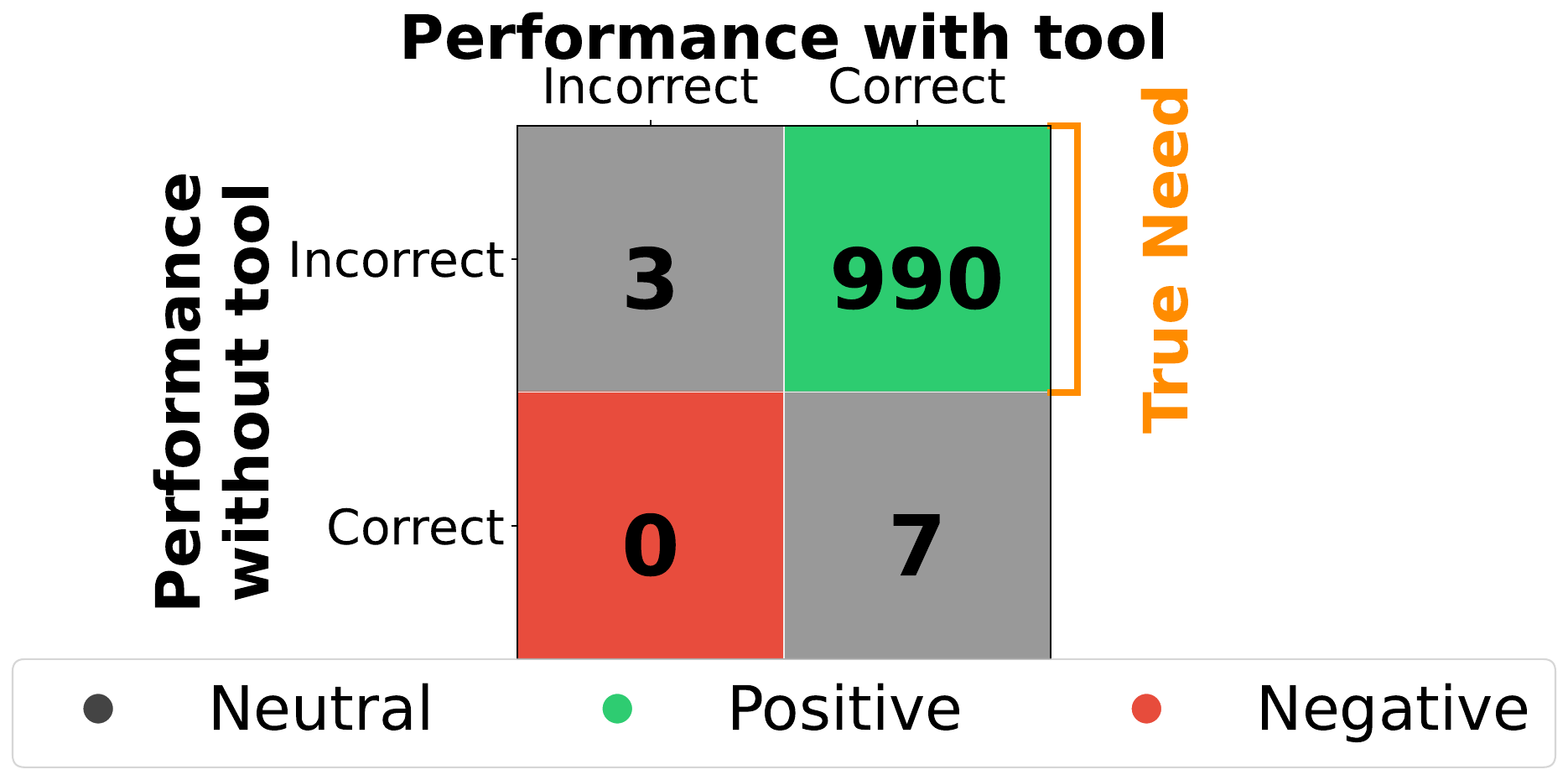}}{\includegraphics[width=\linewidth]{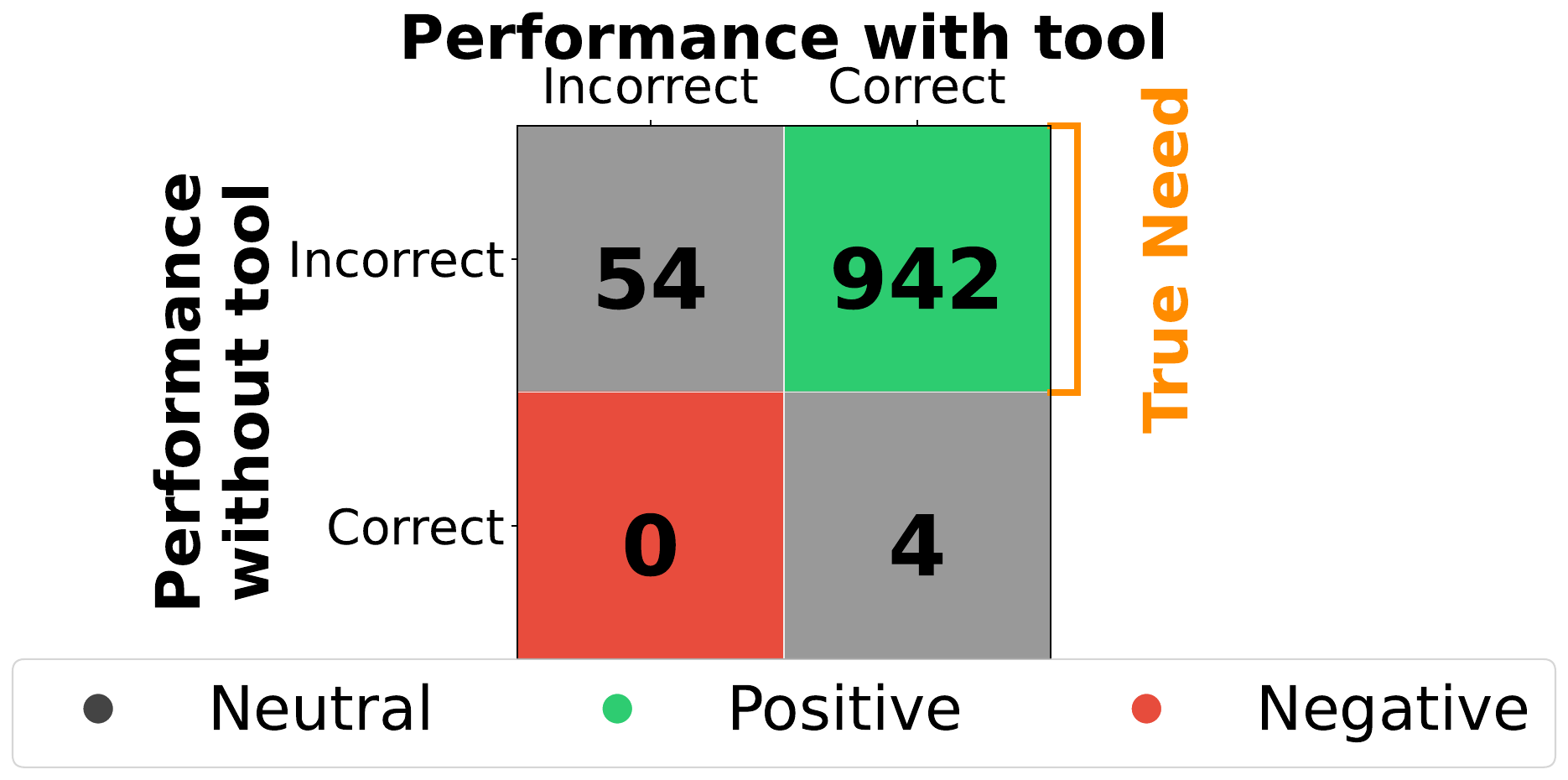}}{\includegraphics[width=\linewidth]{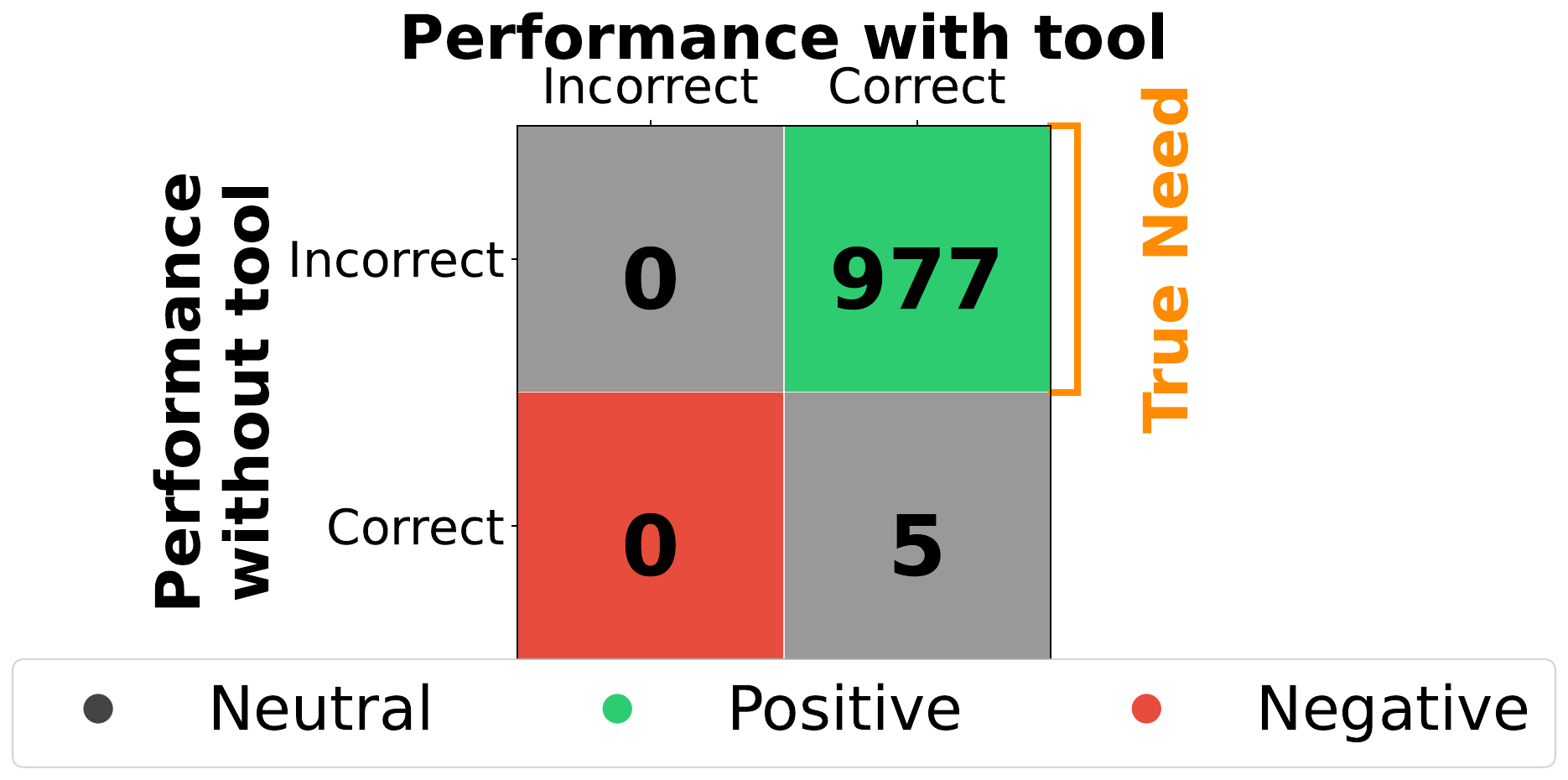}}{\includegraphics[width=\linewidth]{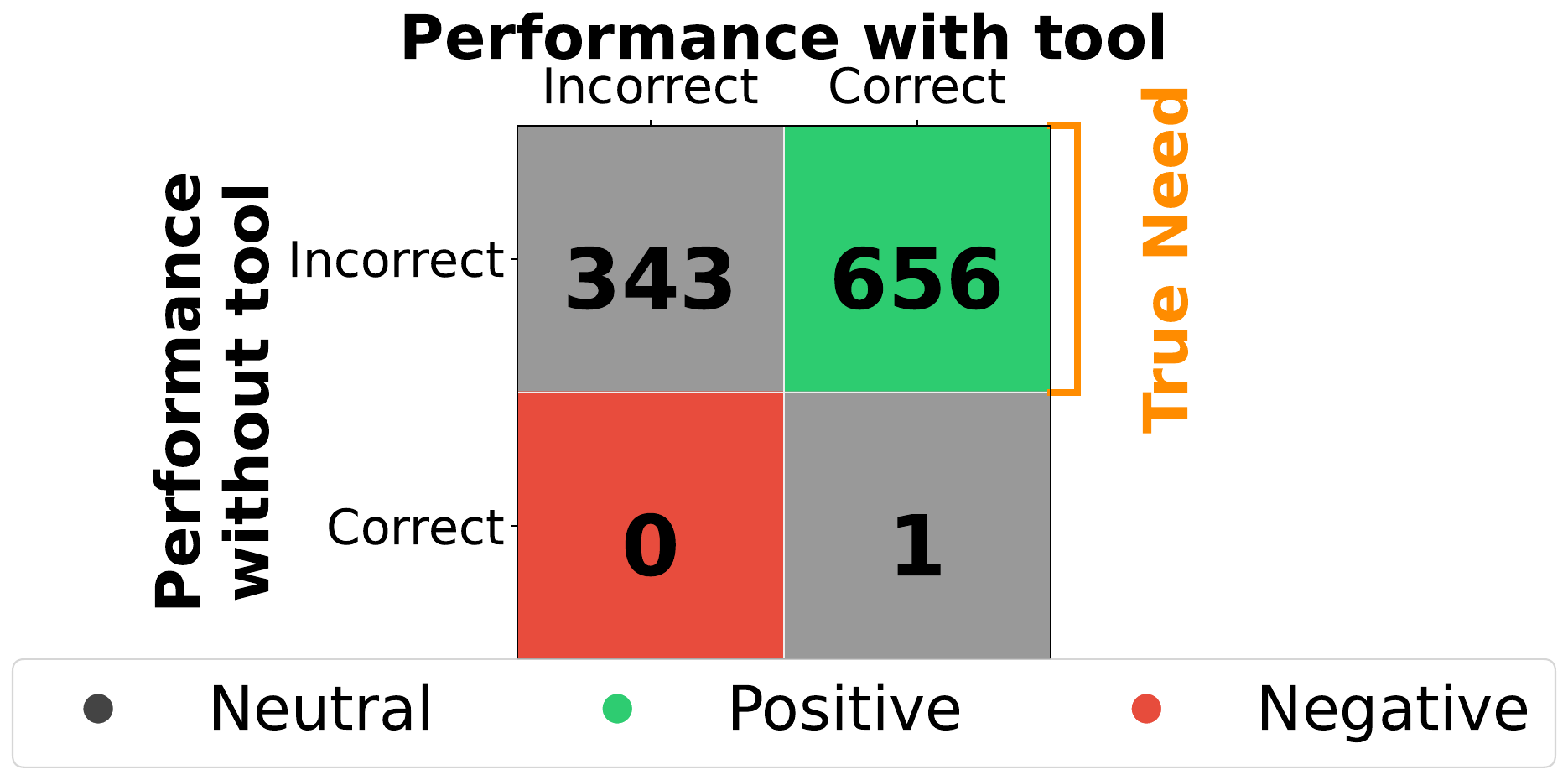}}{\includegraphics[width=\linewidth]{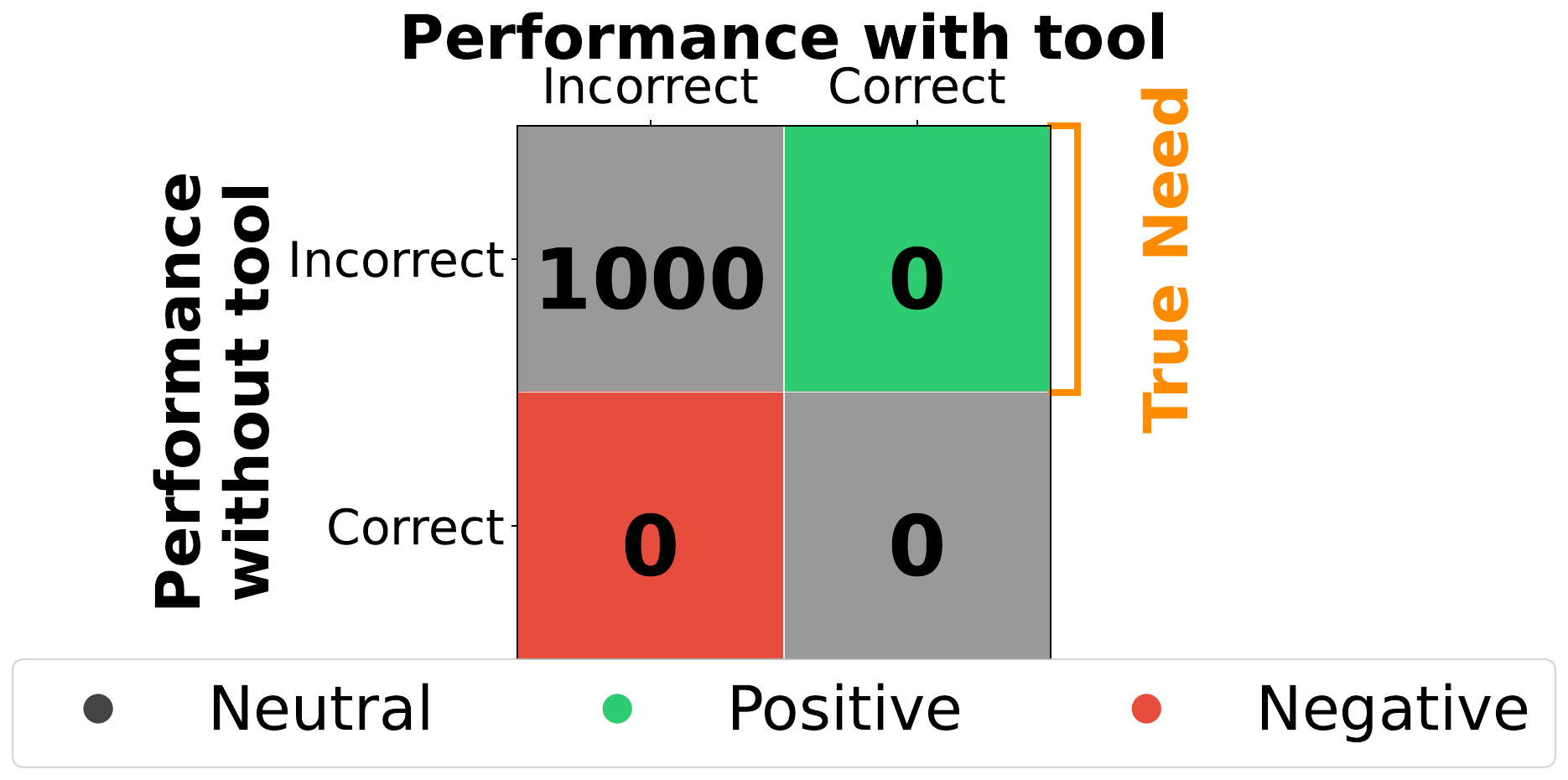}}{\includegraphics[width=\linewidth]{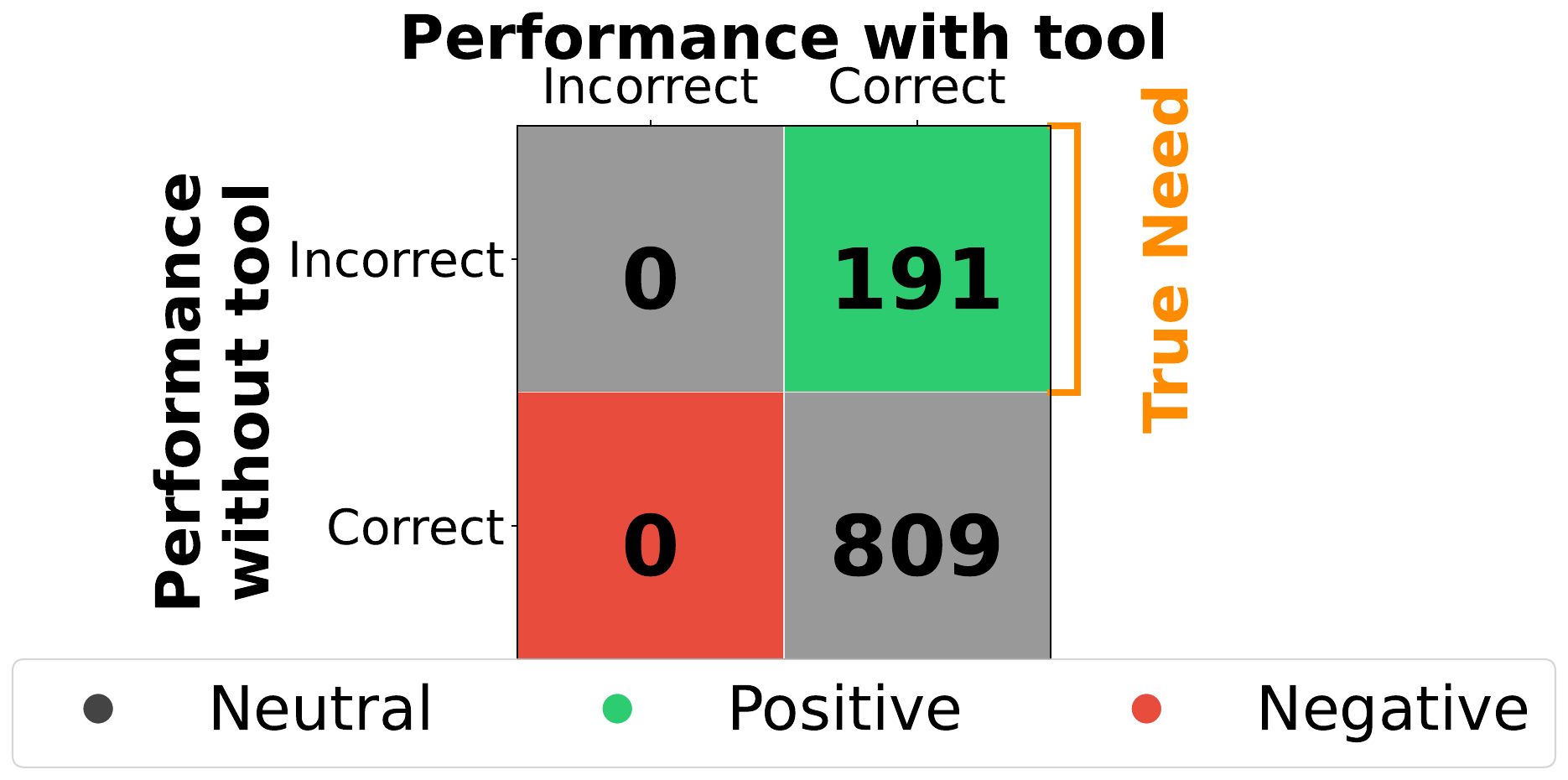}}

\paragraph{Descriptive observation.}
Calling is almost universal for most models and therefore captures most positive utility, but it is not selective. This distinction is most visible for Llama3.2-IT, which calls on all 1,000 examples despite realizing no positive utility. Mistral-IT behaves differently from the near-universal callers: it misses 305 of 656 positive-utility cases while calling on 204 non-positive cases.

\begin{figure*}[t]
    \centering
    \includegraphics[width=0.78\textwidth]{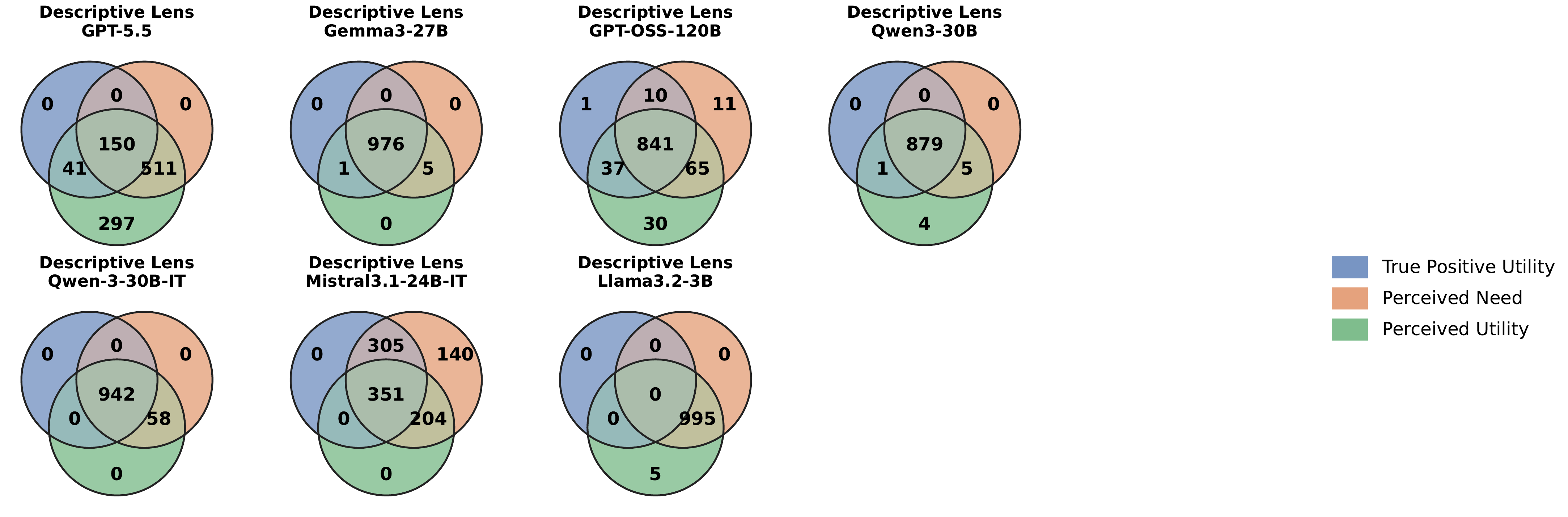}
    \caption{[Synthetic Large-Digit Multiplication Task] Venn diagrams of \textbf{True Positive Utility, Perceived Need, and Perceived Utility}. Each panel shows their empirical overlap for one model. Calls outside true positive utility are non-beneficial; true-positive-utility cases outside perceived utility are missed opportunities. Perceived need is a separate self-assessment and need not be nested within either utility set.}
    \label{fig:venn-synthetic-nn}
\end{figure*}

\paragraph{LNE omission for degenerate labels.}
\label{par:synthetic-nn-lne-omission}
LNE scores for Llama-3.2-3B-It and Mistral-3.1-24B-IT are omitted on Synthetic Large-Digit Multiplication because the per-sample supervision signal is degenerate for these models: labels are defined by whether the forced-calculator answer beats the no-tool answer, and Llama-3.2-3B-IT has zero such cases (0/1000) while Mistral-Small-3.1-24B-Instruct-2503 has only one (1/1000) --- too few positives to fit or cross-validate a classifier.

\calcsevenpanels{\textbf{Synthetic Large-Digit Multiplication: perceived need vs. calculator calling.}\label{fig:synthetic_nn_perceived_need_utility}}
{\includegraphics[width=\linewidth]{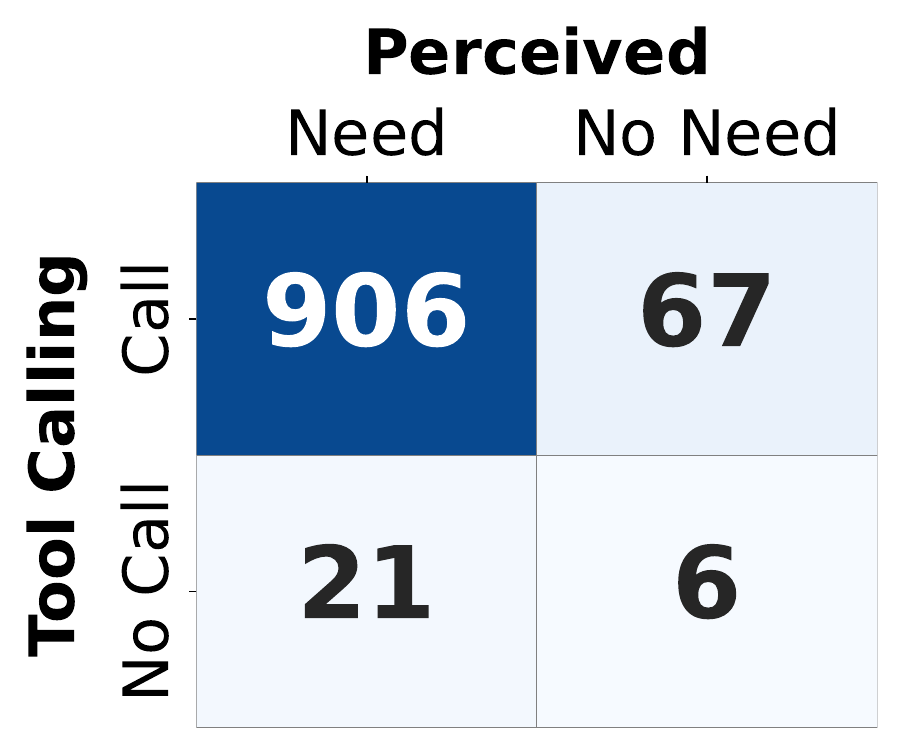}}{\includegraphics[width=\linewidth]{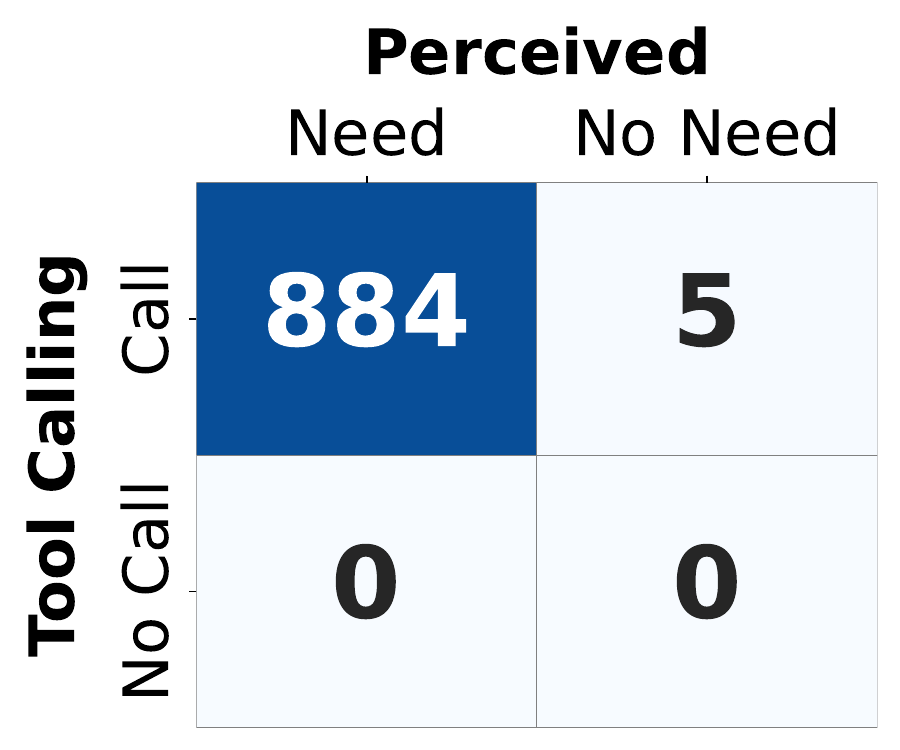}}{\includegraphics[width=\linewidth]{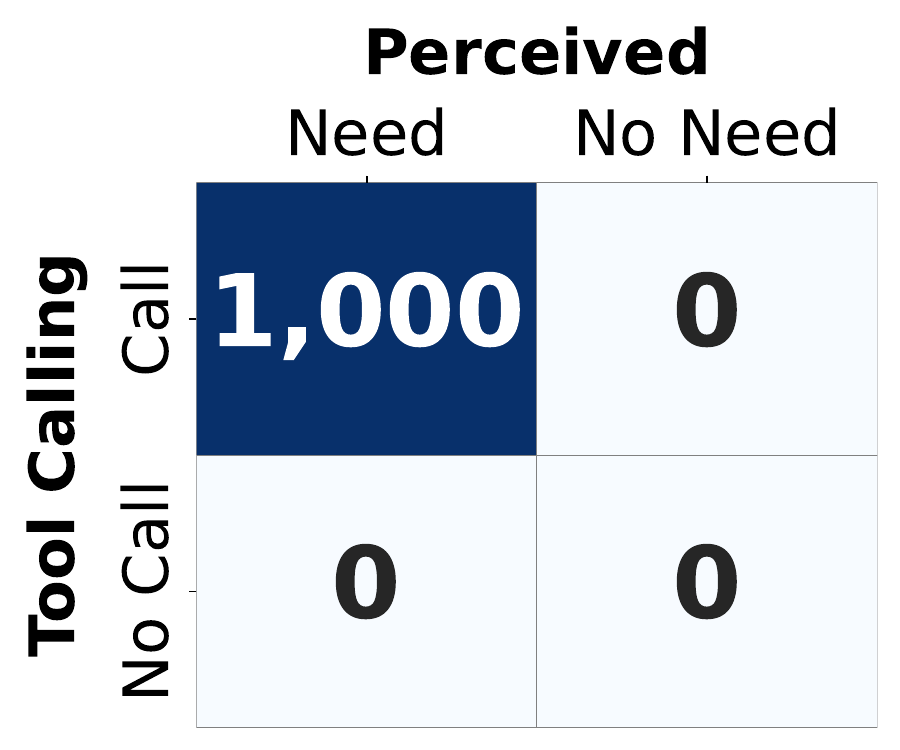}}{\includegraphics[width=\linewidth]{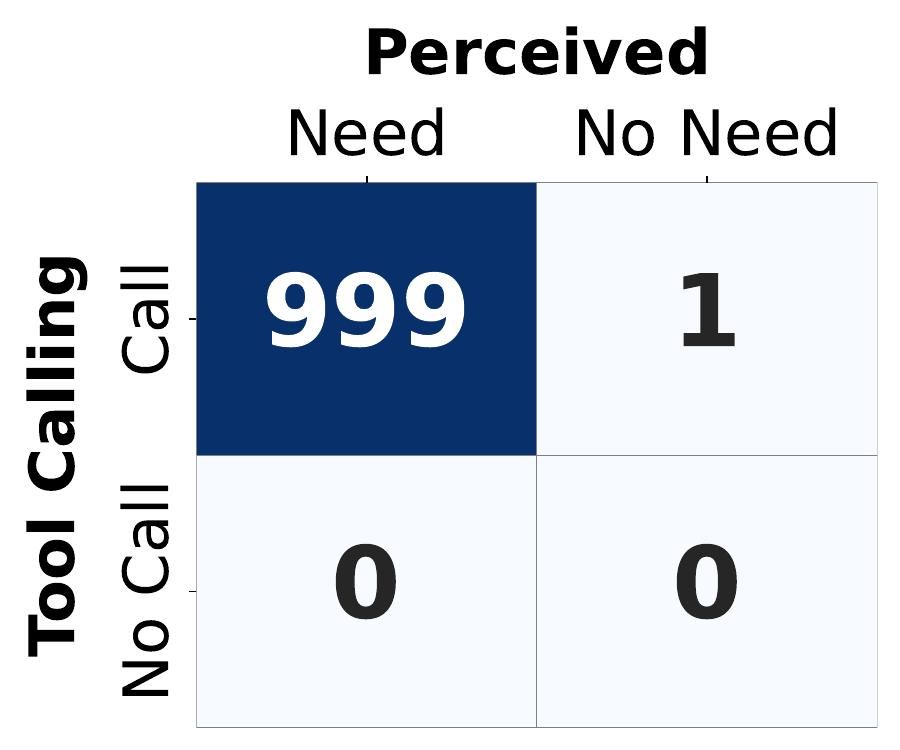}}{\includegraphics[width=\linewidth]{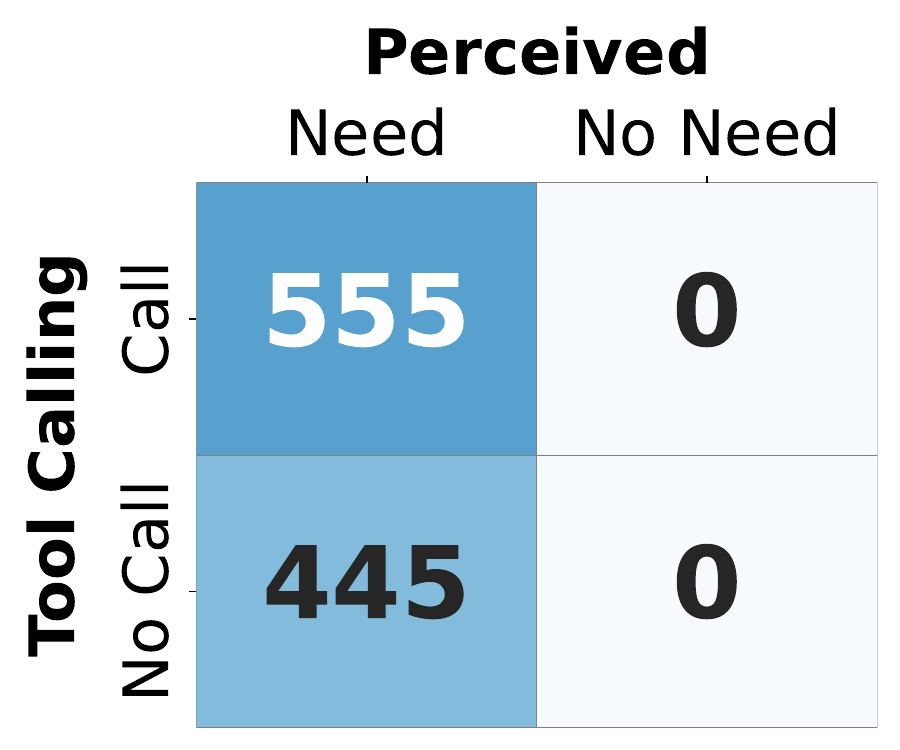}}{\includegraphics[width=\linewidth]{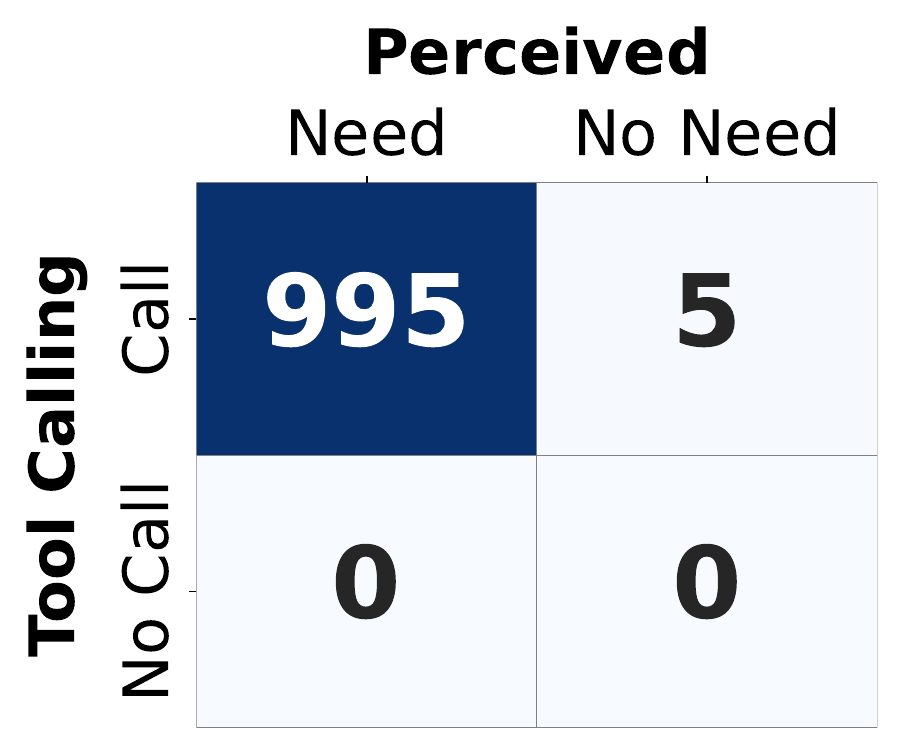}}{\includegraphics[width=\linewidth]{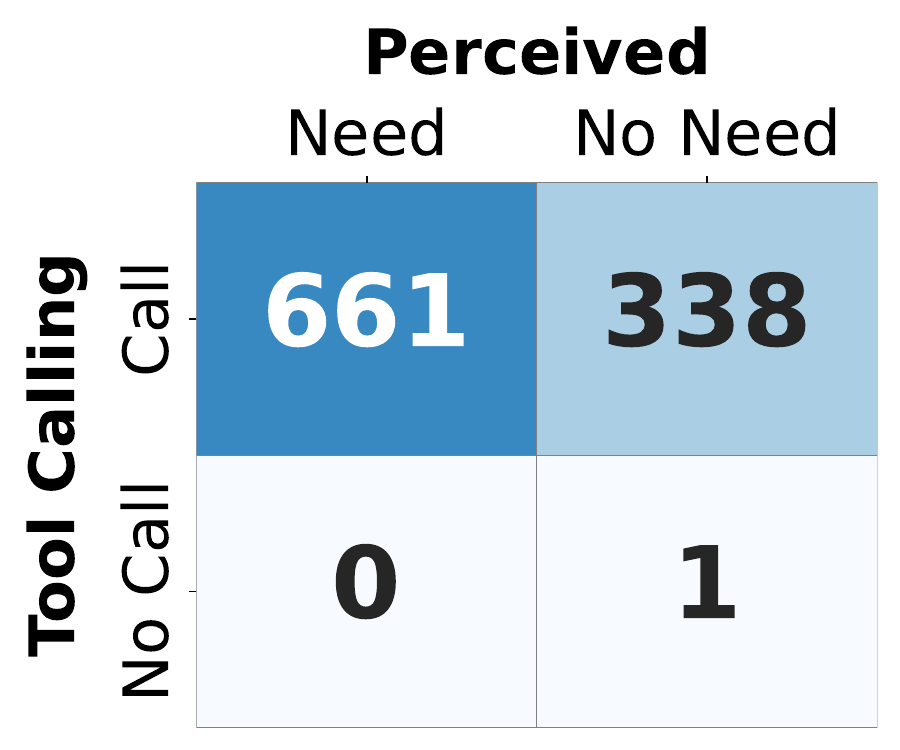}}

\calcsevenpanels{\textbf{Synthetic Large-Digit Multiplication: perceived need.}\label{fig:synthetic_nn_perceived_need}}
{\includegraphics[width=\linewidth]{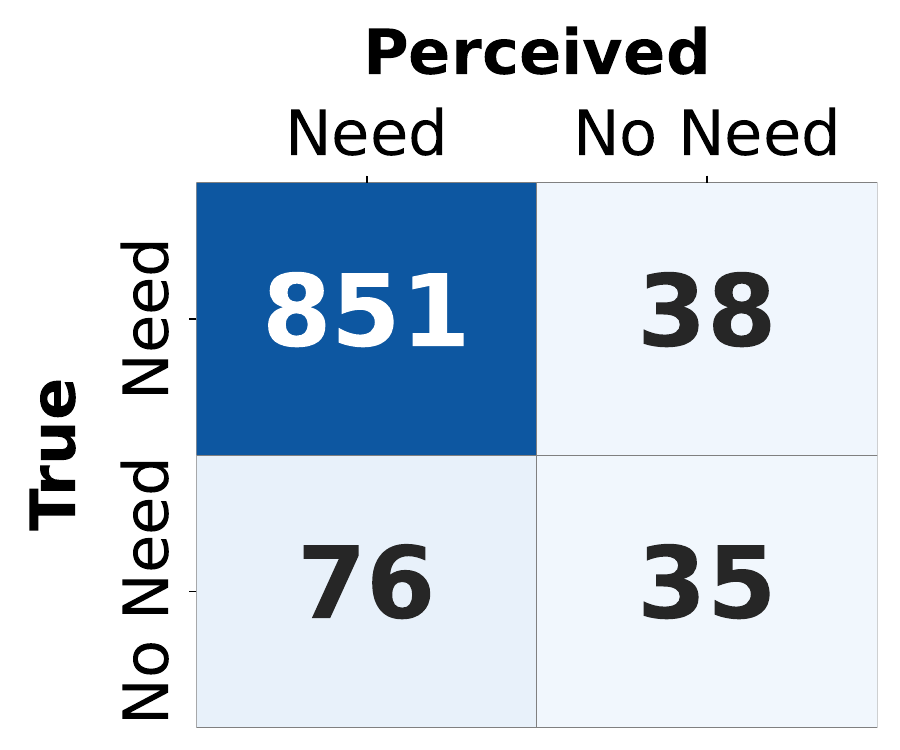}}{\includegraphics[width=\linewidth]{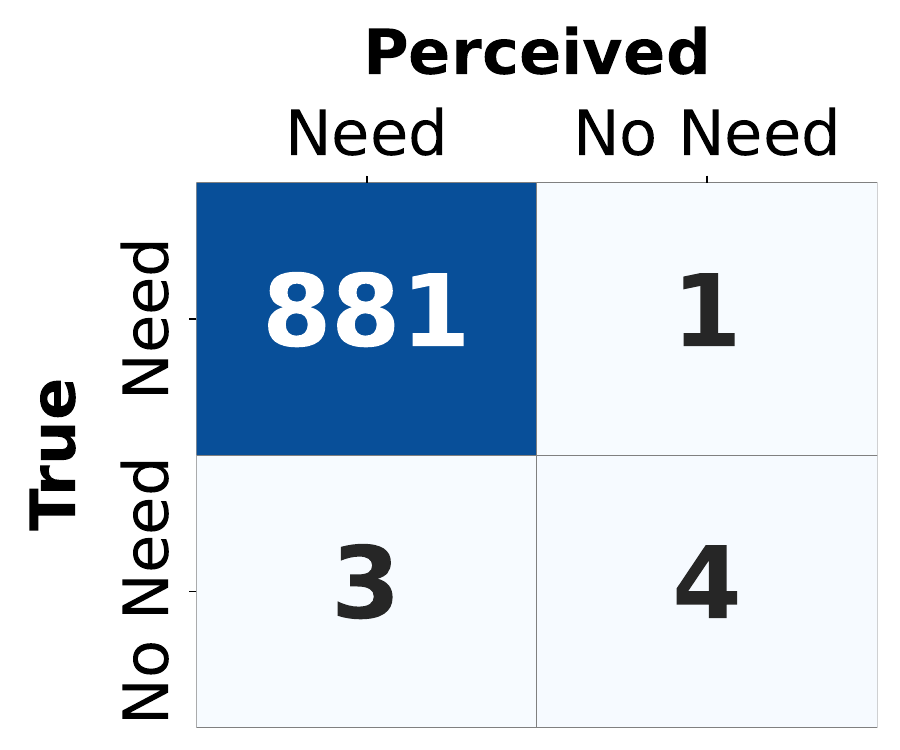}}{\includegraphics[width=\linewidth]{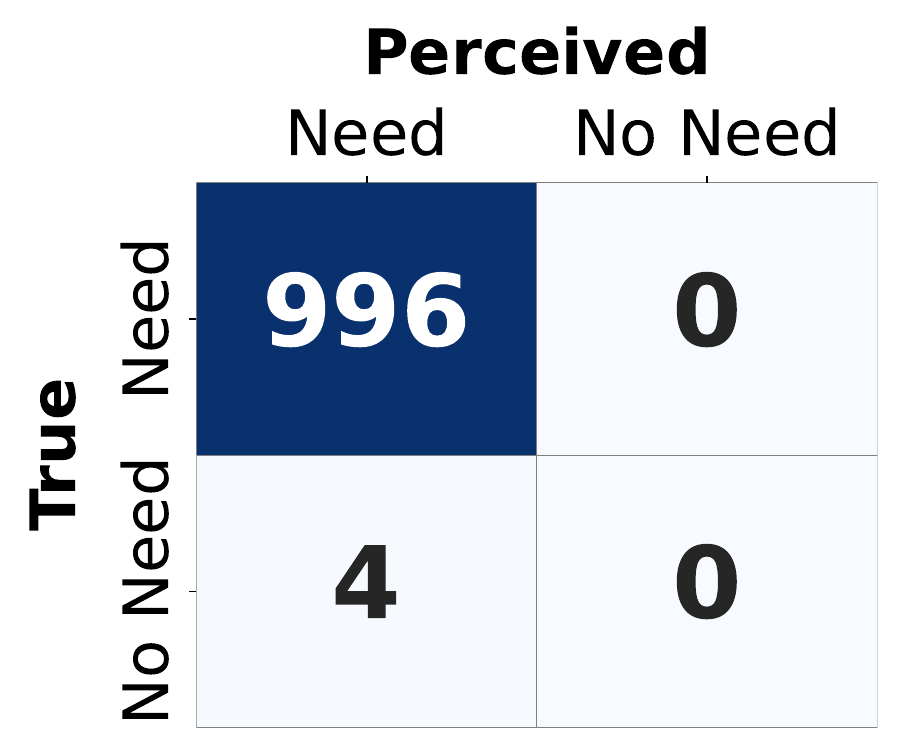}}{\includegraphics[width=\linewidth]{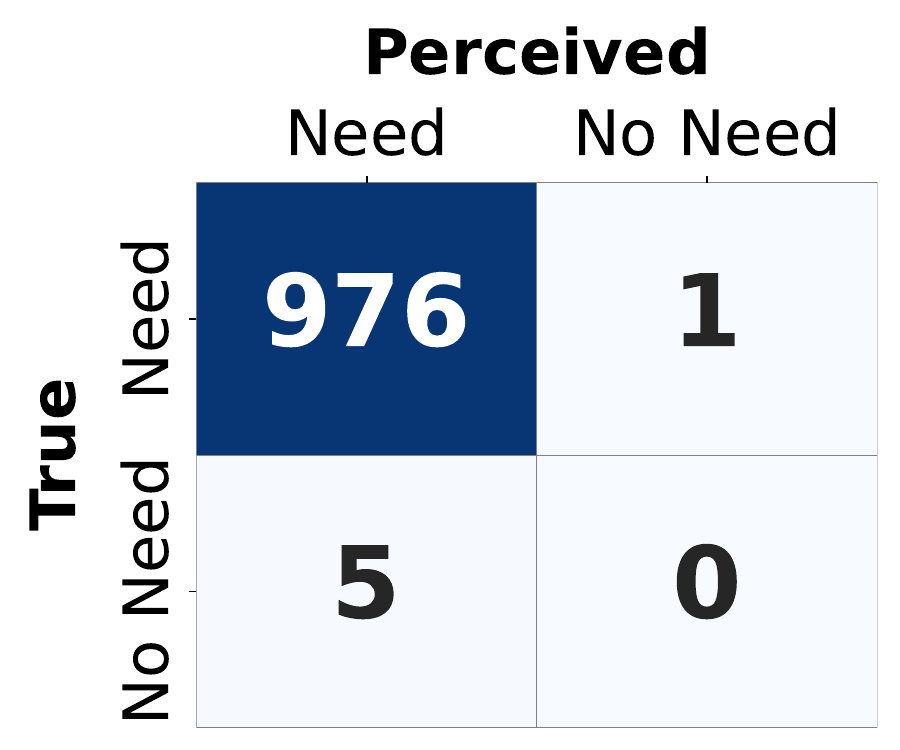}}{\includegraphics[width=\linewidth]{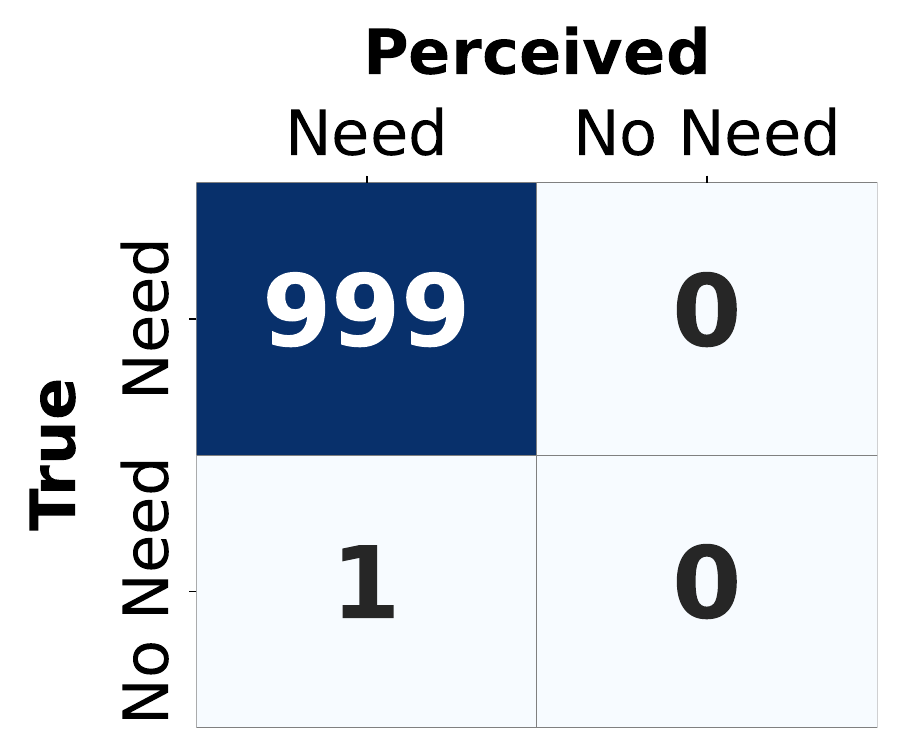}}{\includegraphics[width=\linewidth]{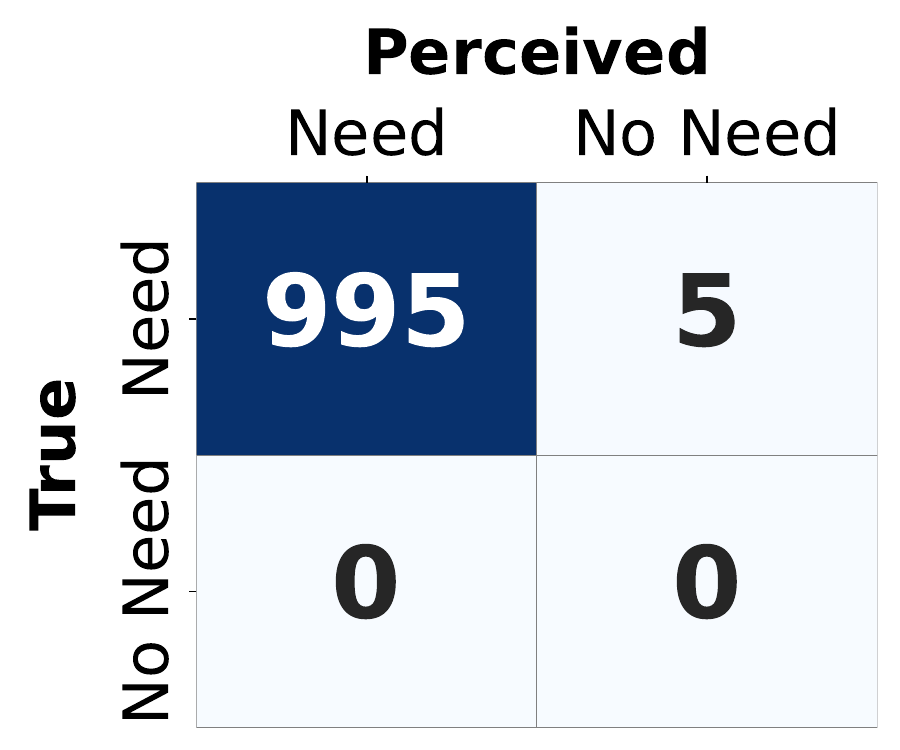}}{\includegraphics[width=\linewidth]{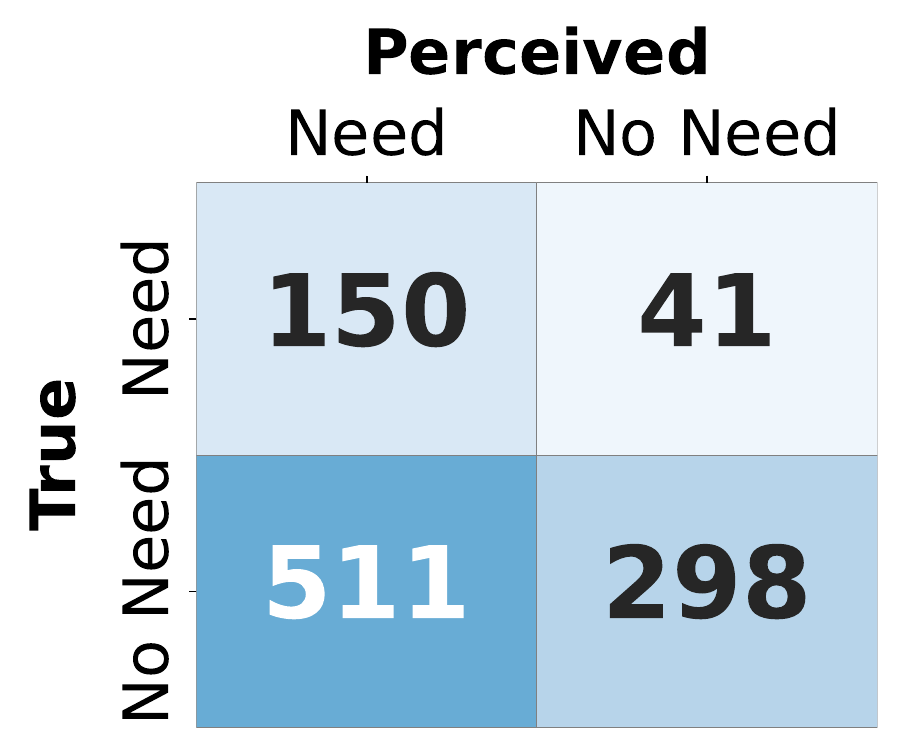}}

\calcsevenpanels{\textbf{Synthetic Large-Digit Multiplication: true utility vs. perceived utility (calculator calling).}\label{fig:synthetic_nn_true_perceived}}
{\includegraphics[width=\linewidth]{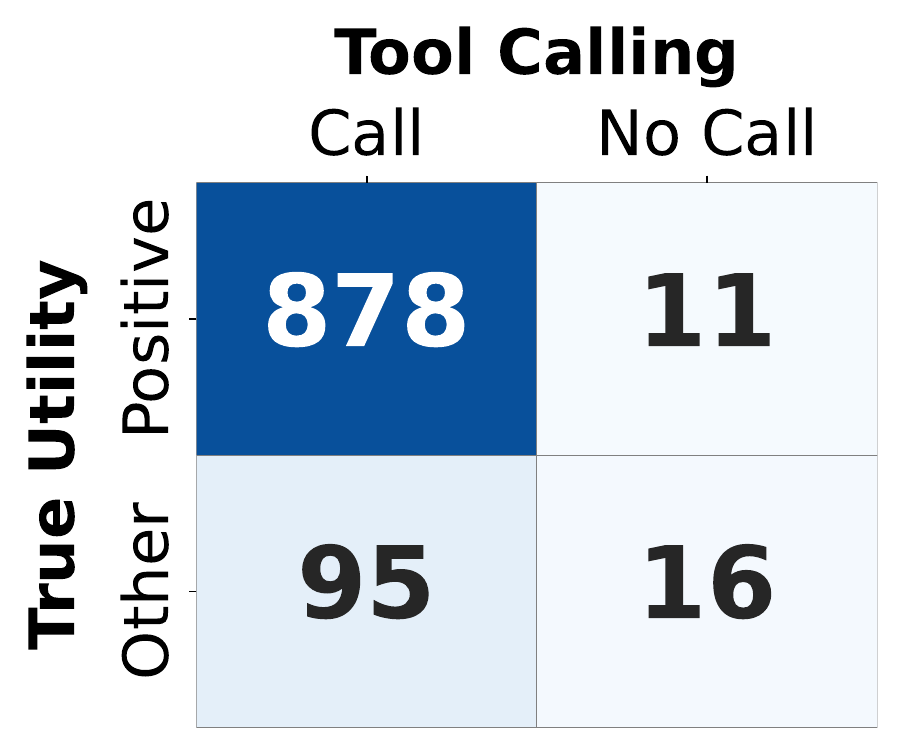}}{\includegraphics[width=\linewidth]{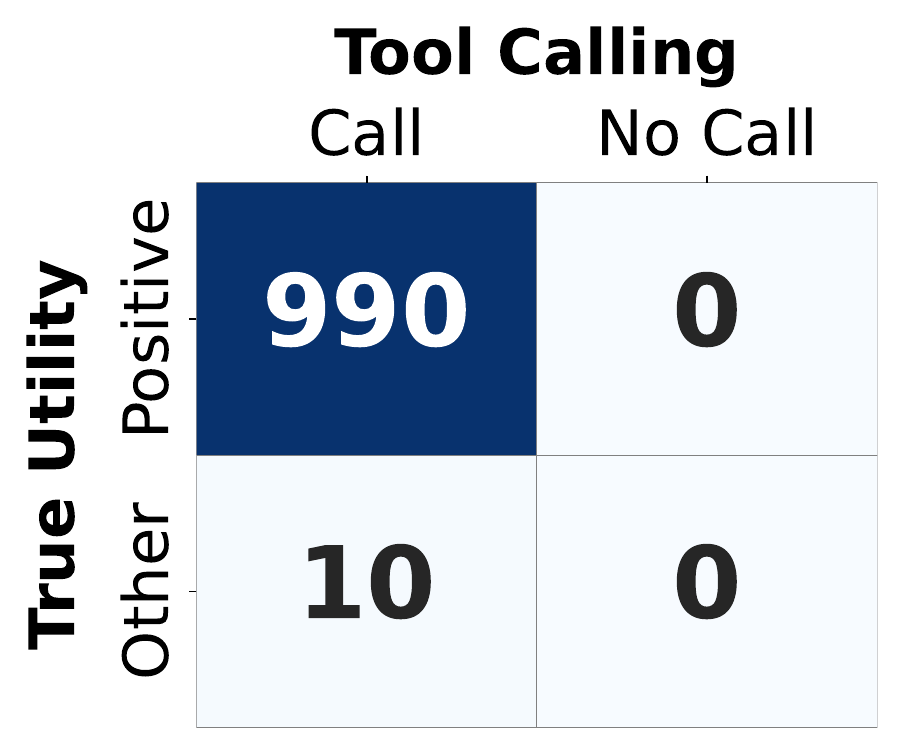}}{\includegraphics[width=\linewidth]{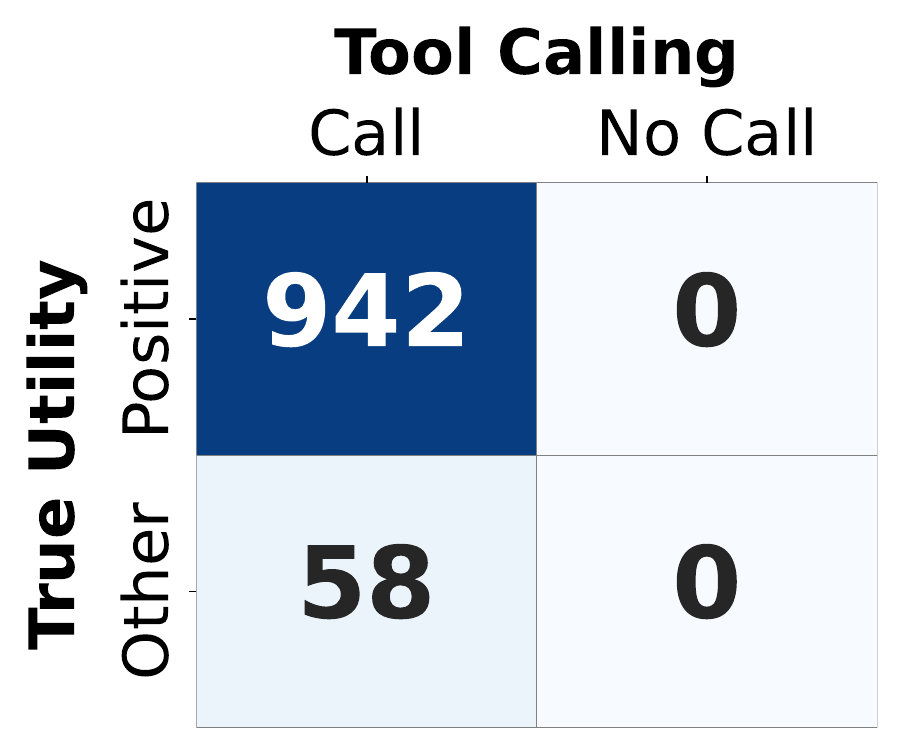}}{\includegraphics[width=\linewidth]{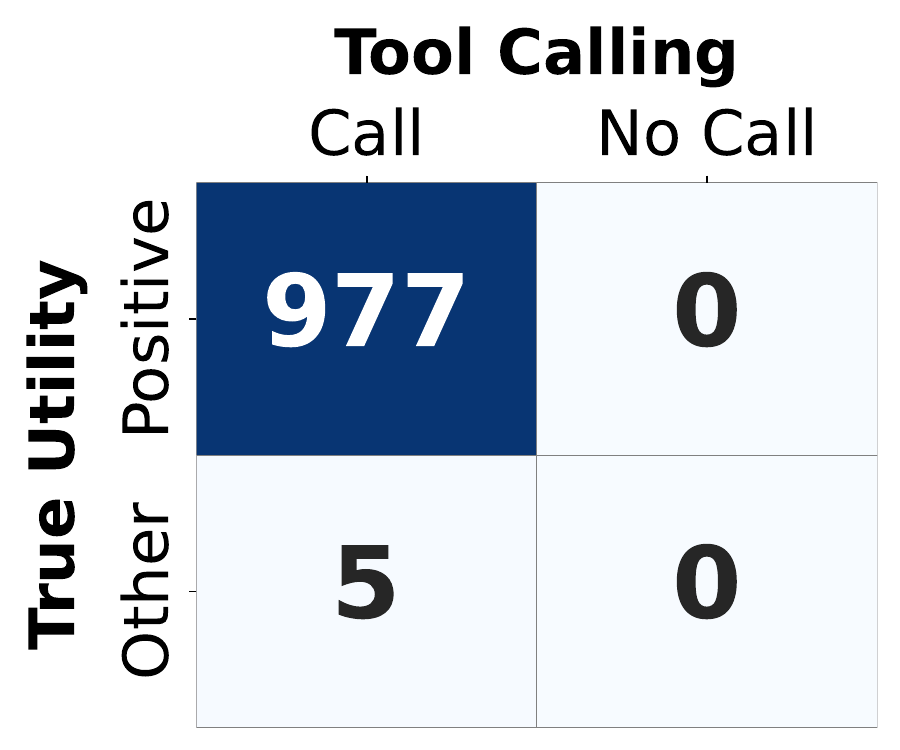}}{\includegraphics[width=\linewidth]{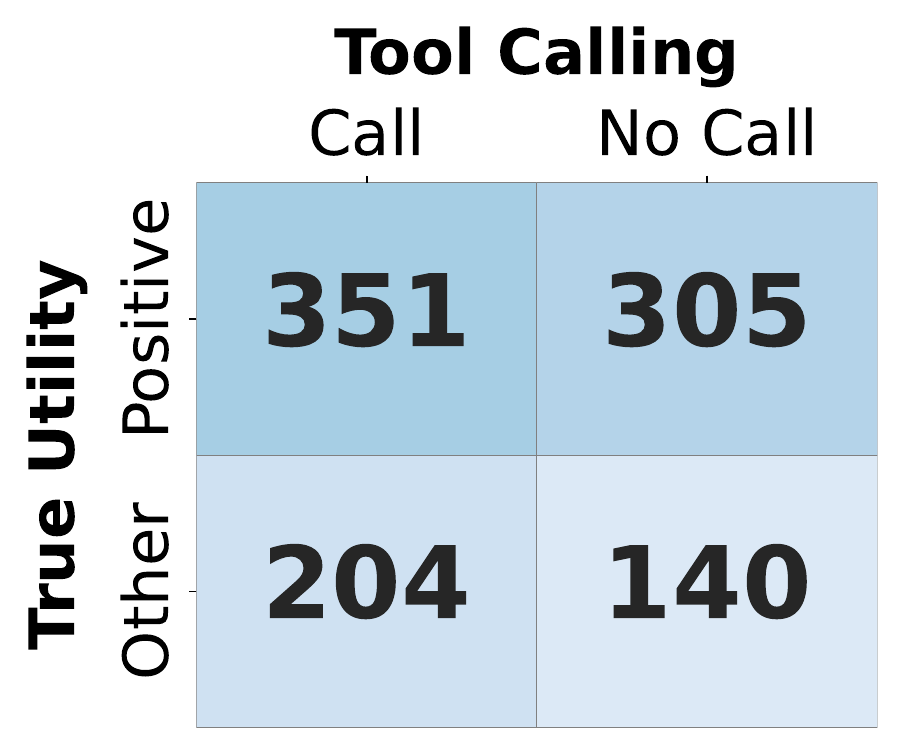}}{\includegraphics[width=\linewidth]{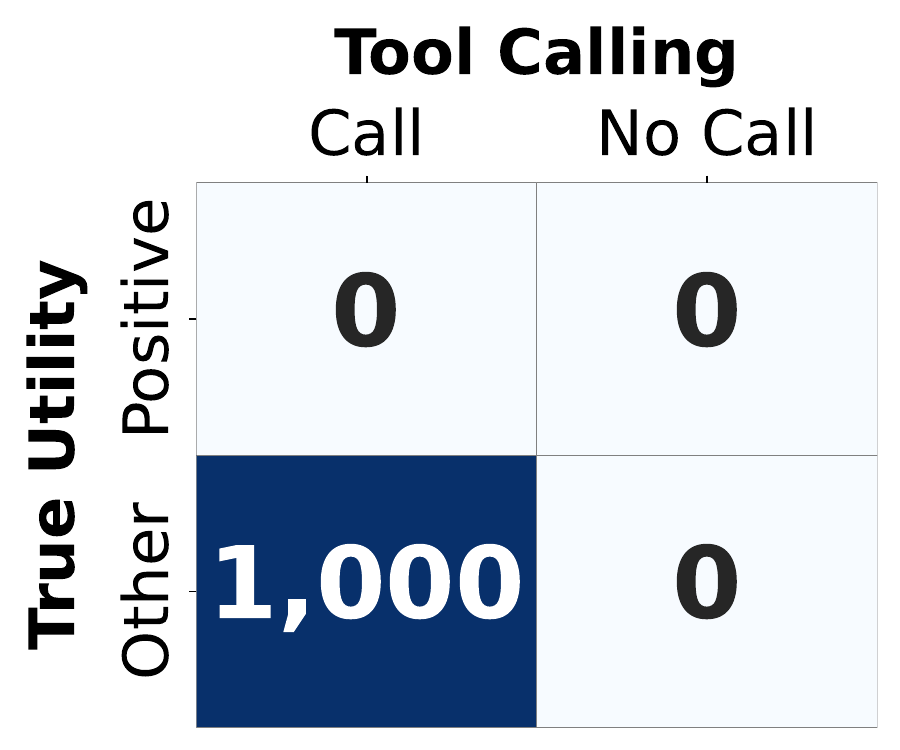}}{\includegraphics[width=\linewidth]{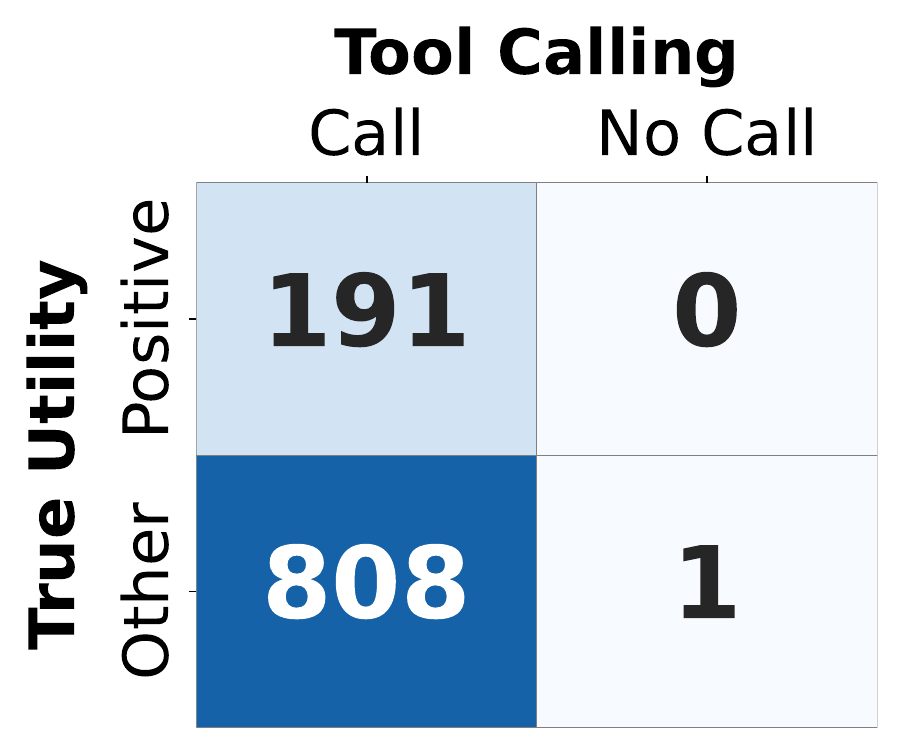}}

\let\calcsevenpanels\relax
\let\calcnormativepanels\relax

\end{document}